\documentclass[11pt]{gsasthesis} % 10,11 and 12pt fonts allowed

\usepackage{etex} % extend the number of registers

\usepackage[margin={1.2in}]{geometry}
\usepackage[titletoc]{appendix}

\usepackage{rotating}
\usepackage{listings}
\usepackage{tikz}
\usepackage{longtable}
\usepackage{adjustbox}
\usepackage[square,numbers,sort&compress]{natbib}

\usepackage[sc]{mathpazo}
\usepackage{courier}

\usepackage[utf8]{inputenc}
\usepackage[T1]{fontenc}

\usepackage[english]{babel}
\usepackage{blindtext}

\usepackage{amsmath}
\usepackage{amssymb}
\usepackage{amsthm}

\newtheorem{theorem}{Theorem}[chapter]
\newtheorem{lemma}[theorem]{Lemma}
\newtheorem{corollary}[theorem]{Corollary}
\newtheorem{proposition}[theorem]{Proposition}
\newtheorem{definition}[theorem]{Definition}

\newtheorem{example}[theorem]{Example}
\newtheorem{remark}[theorem]{Remark}

\newtheorem{defn}[theorem]{Definition} % alias for paper2 compatibility
\newtheorem{thm}[theorem]{Theorem} % alias for chapter4 (ICLR paper)
\newtheorem{prop}[theorem]{Proposition} % alias for chapter4 (ICLR paper)
\newtheorem{Theorem}[theorem]{Theorem} % alias for chapter4 (NeurIPS paper)
\newtheorem{Claim}[theorem]{Claim} % for appendix proofs
\newtheorem{Lemma}[theorem]{Lemma} % alias for appendix proofs

\usepackage{algorithm}
\usepackage{algpseudocode}  % provides \State, \If, \EndIf etc. (replaces algorithmic)

\usepackage{wrapfig}
\usepackage{epigraph}
\usepackage{bm}
\usepackage{bbm}
\usepackage{breqn}
\usepackage{subcaption}  % provides subfigure environment with \begin{subfigure}[b]{width}
\usepackage{sidecap}
\usepackage{colonequals}
\usepackage{enumitem}
\usepackage{mdframed}

\usepackage{dsfont}
\usepackage{mathtools}
\usepackage{xcolor}
\usepackage{hyperref}
\usepackage{array,booktabs,tabularx,multirow,colortbl}

\newcommand{\KMultiAcc}{\text{KMAcc}}

\newcommand{\NN}{\mathbb{N}}
\newcommand{\fbin}{\mathcal{F}_{\mathsf{bin}}}
\newcommand{\fsim}{f_\mathsf{sim}}
\newcommand{\rC}{\mathrm{C}}
\newcommand{\cP}{\mathcal{P}}
\newcommand{\simplexwater}{\texttt{SimplexWater}}
\newcommand{\heavywater}{\texttt{HeavyWater}}
\newcommand{\Dgap}{\mathsf{D}_{\mathsf{gap}}}
\newcommand{\cX}{\mathcal{X}}
\newcommand{\cS}{\mathcal{S}}
\newcommand{\cA}{\mathcal{A}}
\newcommand{\rd}{R_d}
\newcommand{\rp}{R_p}

\newcommand{\ber}{\mathsf{Ber}}

\newcommand\independent{\protect\mathpalette{\protect\independenT}{\perp}}
\def\independenT#1#2{\mathrel{\rlap{$#1#2$}\mkern2mu{#1#2}}}
\newcommand{\indep}{\independent}
\newcommand{\MCBoost}{\text{MCBoost}}
\newcommand{\LSBoost}{\text{LSBoost}}

\newcommand{\bDelta}{\Delta}
\newcommand{\bgamma}{\boldsymbol{\gamma}}
\newcommand{\bh}{\boldsymbol{h}}

\newcommand{\bq}{\boldsymbol{q}}

\newcommand{\ens}{\textup{ens}}
\newcommand{\blambda}{\boldsymbol{\lambda}}
\newcommand{\defined}{\triangleq}

\newcommand{\hatRm}{\hat{\calR}_m}

\newcommand{\bI}{\boldsymbol{I}}

\newcommand{\br}{\boldsymbol{r}}

\newcommand{\bb}{\boldsymbol{b}}
\newcommand{\bg}{\boldsymbol{g}}

\newcommand{\ones}{\mathbf{1}}
\newcommand{\floor}[1]{\lfloor #1 \rfloor}
\newcommand{\ceil}[1]{\lceil #1 \rceil}

\newcommand{\Reals}{\mathbb{R}}

\newcommand{\exampleend}{\hfill$\blacklozenge$}
\DeclareMathOperator*{\argmin}{\arg\!\min}
\DeclareMathOperator*{\argmax}{\arg\!\max}

\newcommand{\bk}{\mathbf{K}}
\newcommand{\BR}{\mathbb{R}}
\newcommand{\BE}{\mathbb{E}}
\newcommand{\BP}{\mathbb{P}}

\newcommand{\BN}{\mathbb{N}}
\newcommand{\frakC}{\mathfrak{C}}
\newcommand{\define}{:=}
\newcommand{\calX}{\mathcal{X}}
\newcommand{\calY}{\mathcal{Y}}
\newcommand{\calH}{\mathcal{H}}
\newcommand{\calC}{\mathcal{C}}
\newcommand{\calV}{\mathcal{V}}
\newcommand{\calD}{\mathcal{D}}

\newcommand{\calL}{\mathcal{L}}
\newcommand{\calP}{\mathcal{P}}
\newcommand{\calS}{\mathcal{S}}
\newcommand{\calA}{\mathcal{A}}
\newcommand{\calB}{\mathcal{B}}

\newcommand{\calT}{\mathcal{T}}
\newcommand{\calZ}{\mathcal{Z}}

\newcommand{\calR}{\mathcal{R}}
\newcommand{\bX}{\boldsymbol{X}}
\newcommand{\bx}{\boldsymbol{x}}
\newcommand{\by}{\boldsymbol{y}}
\newcommand{\bc}{\boldsymbol{c}}
\newcommand{\bA}{\boldsymbol{A}}
\newcommand{\bB}{\boldsymbol{B}}
\newcommand{\bd}{\boldsymbol{d}}
\newcommand{\bxi}{\boldsymbol{\xi}}
\newcommand{\bK}{\boldsymbol{K}}

\newcommand{\indicator}{\mathds{1}}
\newcommand{\ExpVal}[2]{\mathbb{E}_{#1}\left[ #2 \right]}
\newcommand{\EE}[1]{\mathbb{E}\left[ #1 \right]}

\newcommand{\empCstar}{c^\star_{k,\calD_0,f}}
\newcommand{\empC}{c^\star_{k,\calD_0,f}}
\newcommand{\wc}{\, \cdot \,}
\newcommand{\Co}{ \ : \ }
\renewcommand{\tilde}{\widetilde}
\newcommand{\TV}{\mathsf{TV}}
\DeclareMathOperator{\supp}{supp}
\newcommand{\remarkend}{\hfill$\blacklozenge$}
\allowdisplaybreaks

\usepackage{microtype}

\usepackage[stable,multiple]{footmisc}

\RequirePackage[font=small,format=plain,labelfont=bf,textfont=it]{caption}
\title{Trustworthy AI: Ensuring Reliability and Accountability from Models to Agents} % needs to match title on DAC
\author{Xuan Long} % full name as it appears on your GSAS record, needs
\degreename{Doctor of Philosophy}
\degreefield{Applied Mathematics} % Official name of subject as listed in GSAS
\department{John A. Paulson School Of Engineering And Applied Sciences} % official name of department
\degreemonth{March} % Month of Defense (i.e. month when DAC was signed)
\degreeyear{2026} % Year the DAC was signed
\principaladvisor{Professor Flavio Calmon}

\begin{document}

%%%%%%%%%%%%%%%% FRONTMATTER %%%%%%%%%%%%%%%%

\pagenumbering{roman} % GSAS wants roman page numbers for frontmatter

% the following four pages are required in that order. The first two pages are
% not allowed to have page numbers, this is taken care of in the class file.
\thesistitlepage
\copyrightpage
\begin{abstract}
In this thesis, we develop algorithms with theoretical guarantees for ensuring reliability and accountability of Machine Learning (ML) systems. As ML systems evolve from predictive models to generative models and autonomous agents, the landscape of trustworthy AI has shifted. This thesis introduces tools grounded in information theory, optimization, and statistical learning to mitigate bias, reduce arbitrary decisions, ensure content provenance, and evaluate LLM-driven agents in autonomous settings.

Towards mitigating bias and arbitrariness in traditional ML models, we introduce a kernel-based method to achieve multiaccuracy across complex subpopulations that traditional demographic categories may overlook. We also develop methods to address predictive multiplicity, where equally accurate models yield conflicting individual predictions. 

We ensure the accountability in generative AI through watermarking large language models (LLMs). We characterize the information-theoretic trade-off between watermark detection and text distortion and derive optimal watermarking strategies by leveraging optimal transport and coding theory. Empirical evaluations show our watermarks achieve a superior detection-quality tradeoff across language generation and coding tasks.

Finally, we evaluate autonomous LLM agents in multi-agent environments through the first simulator of a fully LLM-driven supply chain. While agents can outperform human experts, reducing costs by up to 67\%, we identify systemic risks such as costly tail events. 
\end{abstract}

% Center headings for table of contents, LOT, and LOF and make them smaller so
% that "Abstract", "Acknowledgments" and "Contents" all look alike. Comment out
% if you want the default. If you want more control, use the "tocloft" package.
\renewcommand{\contentsname}{\protect\centering\protect\Large Contents}
\renewcommand{\listtablename}{\protect\centering\protect\Large List of Tables}
\renewcommand{\listfigurename}{\protect\centering\protect\Large List of Figures}

\tableofcontents % Table of contents

% The rest of the front matter: Lists of tables, figures, dedication and
% acknowledment is optional. Comment out whatever you don't like
\listoftables
\listoffigures
\begin{acknowledgments}
A professor-friend once told me that I was the only happy Ph.D. student she knew. This was in December 2024, when I attended NeurIPS for the third time in Vancouver. I did not miss a single NeurIPS during my five years as a Ph.D. student, thanks to my collaborators and Flavio’s generous support. 

What she said was definitely a hyperbole, but looking around me, it unfortunately reflects an element of truth. I did not start there, though. Like many others, I once took the games of life and that self-identity of mine very seriously, living under deluded thoughts and disregarding my human nature in pursuit of things I had been told were important. A paradigm shift was brewing when I was fortunate enough to encounter some setbacks --- mainly health- and interpersonal-related issues --- which culminated in the middle of my Ph.D. I managed to rebuild the support system both within myself and around me, and I rebounded.

I came to the conclusion that happiness is a choice and a state of mind, completely unrelated to external circumstances. While the point is not to cling to desire, maintaining this mindset long enough, paradoxically, helped me get things done. In fact, many people throughout history have expressed this idea in different ways: Nassim Nicholas Taleb calls it "antifragile" (\textit{Antifragile}, 2012) Xuanzang describes it as a "mind free of hindrances" (\textit{the Heart Sutra}, 7th century CE), and Laozi simply describes it as water(\textit{Dao De Jing}, 4th century BCE). 

Pursuing a PhD is a dual journey of self-evolution, both academic and personal. It is a process of constantly challenging one’s own hypotheses, pushing forward while falling into rabbit holes with research ideas, and striving to keep pace with the ever-evolving academic currents. Parallel to this intellectual journey are the internal landscapes of self-doubt, relationships with collaborators and the wider academic community, and all the ties we have with the world. 

This process is humbling. I am grateful for the classical texts of Chinese philosophy and Chan; their endless streams of wisdom and compassion have served as a compass, helping me navigate complicated waters. As a meditation practitioner, I am also thankful for the teachers, environments, and communities that have cultivated my practice. 

This thesis is a milestone of my journey as an academic, a role which, like any other in society, bears unique responsibilities and commitments. I hope to move forward on this path with the same curiosity and humility that defined these fruitful years. I am excited to see how this journey continues to unfold.

I am grateful to my family and my partner for their unwavering support and unconditional love. A special thank you to my uncle and aunt for providing me with a place to call home during my eight years of education in the United States. I also wish to thank my friends, from my undergraduate years at NYU to my doctoral studies at Harvard. Thank you for the wonderful memories of traveling together, bonding over food, dancing in studios and on stages, creating music together, and celebrating birthdays and holidays. Life is so much more colorful thanks to you.

I would also like to thank all my collaborators and members of the lab, your drive for excellence never fails to motivate me to do better. I am fortunate to have crossed paths with you.

I would like to thank Professor Madhu Sudan and Professor Himabindu Lakkaraju for their guidance and for witnessing and supporting my growth from my qualifying exam to my thesis defense. I would like to thank all the professors and teachers I have had during my 20 years of education. Thank you for sharing your passion for your subjects and for inspiring my own.

I wish to express heartfelt gratitude to three key mentors and advisors who have shaped my academic and personal trajectory. 

First, Professor Robert V. Kohn of NYU Courant. It is with a heavy heart that I learned of his passing as I write these words. His kindness, support, and encouragement kick-started my academic path. 

I am deeply indebted to Professor Flavio du Pin Calmon for graciously accepting me as a Ph.D. student in his lab and for his support throughout my time at Harvard, including shouldering the uncertainties of funding behind the scenes. His mentorship has been instrumental in shaping how I think about and conduct research. 

Finally, I would like to thank Professor David Simchi-Levi, whom I first contacted through a cold email, for recognizing the potential impact of the research direction I intended to pursue. His generosity and guidance transformed the initial ideas into impactful work that would not have been possible without his invaluable input. 

To all who have supported or challenged me throughout this journey, I extend my sincere
gratitude.

\end{acknowledgments}
\begin{dedication}
This thesis is dedicated to sunshine, air, and water, to joy and love in the world, and to all who have supported or challenged me in this journey.
\end{dedication}

%%%%%%%%%%%%%%%% MAIN BODY %%%%%%%%%%%%%%%%
\pagenumbering{arabic} % reset page numbering and switch to arabic

% Introductory chapter. Comment out if you don't have an intro chapter, but I
% think most committees expect you to have one.
% Don't number the intro chapter, but add to to the table of contents
% \addcontentsline{toc}{chapter}{Introduction}
\chapter{Introduction}\label{ch:intro}

Machine learning is no longer a fragmented landscape of isolated tools; instead, they have been integrated into the core decision-making pipelines across domains, shaping who is hired, who receives medical care, and who is approved for credit. In the past few years, large language models (LLMs) have expanded the scope of AI from \emph{prediction} to \emph{generation}---producing synthetic text at scale---and, more recently, to \emph{action}, through agentic systems that reason, plan, and make sequential decisions in dynamic environments. These shifts raise a central question that motivates this thesis:

\begin{quote}
\emph{How do we ensure that AI systems remain trustworthy as they evolve from models that predict, to models that generate, to agents that act?}
\end{quote}

The title of this thesis, \textit{Trustworthy AI: Ensuring Reliability and Accountability from Models to Agents}, reflects two core requirements that become increasingly urgent across this progression. \textbf{Reliability} demands that system outputs be stable and dependable—especially when multiple “equally good” models can disagree, or when errors concentrate in subtle subpopulations. \textbf{Accountability} demands that system outputs be attributable and auditable—especially when synthetic content can be indistinguishable from human text, and when autonomous agents can produce emergent dynamics whose failures are costly and difficult to diagnose.

% [TBD] To address these challenges, this thesis adopts a perspective inspired by a classical information-theoretic methodology: \emph{build an abstract model of the system, analyze its properties and fundamental limits under clear assumptions, and translate the analysis into principled mechanism designs}. 

% [TBD] This thesis follows that blueprint across three settings—predictive models, generative models, and multi-agent systems—using tools from information theory, optimization, and statistical learning.

\paragraph{From accuracy to trustworthiness.}
In many deployments of ML systems, optimizing for high performance alone is not enough to guarantee trust. In high-stakes settings, model failures are often \emph{structured}: model errors can disproportionately affect particular population groups; an arbitrary designer choice during model development can change decisions for the same individual. In the LLM era, high-fidelity synthetic content from generative models can be used for misinformation or fraud. For AI agents making business decisions, strong average performance is not sufficient; true reliability depends on robust behavior in the tail, where rare but costly failures occur. 

Moving from predictive models to generative models and agents changes not only the \emph{capabilities} of AI, but also the \emph{trustworthiness issues} we must anticipate and address. Prior research has largely emphasized model evaluation using average accuracy, bias over predefined groups, or single-agent benchmarks. This thesis identifies the reliability and accountability gaps that emerge beyond these metrics and develops principled solutions across models and agents.

\begin{itemize}
\item \textbf{Predictive models:} High average accuracy does not guarantee trustworthy outcomes. Average performance can mask systematic failures on vulnerable subpopulations, and equally accurate models may produce conflicting predictions for the same individuals (\emph{predictive multiplicity}), introducing \emph{arbitrariness} into decision-making. Existing fairness frameworks largely overlook this individual-level instability. This thesis formalizes arbitrariness as a distinct dimension of reliability and reveals its tension with fairness interventions, expanding the traditional fairness–accuracy paradigm.
\item \textbf{Generative models:} As LLM outputs become indistinguishable from human writing, trust increasingly depends on our ability to track content provenance. Without reliable attribution mechanisms, generative models can be misused at scale. Existing watermarking methods lack consideration for a prominent failure mode of watermarking---when the token distributions have low entropy. This thesis fills that gap by grounding watermark design in an optimization framework that considers worst-case scenarios, establishing principled detection–quality tradeoffs and constructing provably detectable, zero-distortion schemes.
\item \textbf{Autonomous agents:} Most prior work evaluates LLMs in isolation, yet real-world deployments involve interacting agents in dynamic environments. In such systems, reliability is determined by emergent collective behavior, where feedback loops and stochastic decisions can generate rare but consequential tail events despite strong average performance. Existing benchmarks do not capture these systemic risks. This thesis addresses this gap by developing a multi-agent testbed that enable the study of multi-agent decision-making in the high-stake domain of supply-chain management. Furthermore, we study guardrail design, information sharing, and orchestration that enables better coordination among LLM agents.

% \item \textbf{Autonomous agents:} trust can fail at the \emph{system level} when multiple agents interact over time; collective behavior can exhibit high variance and rare “tail events,” even if average performance looks strong.
\end{itemize}

\paragraph{Thesis scope.}
This thesis develops theory and algorithms that target these risks along three pillars:
\begin{enumerate}
\item \textbf{Reliability of predictive ML models:} diagnosing model bias and improving model performance across rich collections of subpopulations, while reducing individual-level prediction arbitrariness.
\item \textbf{Accountability of generative models:} 
 designing rigorous text watermarking mechanisms for LLMs with provable detection-distortion guarantees, targeting the low-entropy regime typical in LLM token distributions.
\item \textbf{Trustworthiness of autonomous AI agents:} building testbeds that simulate emergent multi-agent dynamics and identifying mechanisms that improve performance while revealing systemic risks.
\end{enumerate}

\section*{Overview and Main Contributions}

Here, we provide an overview and our main contributions of each chapter. 

\subsection*{Individual Arbitrariness and Group Fairness}
% % individual arbitrariness and group fairness
% Machine learning tasks may admit multiple competing models that achieve similar performance yet produce arbitrary outputs for individual samples---a phenomenon known as predictive multiplicity. In this thesis, we demonstrate that fairness interventions in machine learning optimized solely for group fairness and accuracy can exacerbate predictive multiplicity. Consequently, state-of-the-art fairness interventions can mask high predictive multiplicity behind favorable group fairness and accuracy metrics. We argue that a third axis of ``arbitrariness'' should be considered  when deploying models to aid decision-making in applications of individual-level impact.
% To address this challenge, we propose an ensemble  algorithm applicable to any fairness intervention that provably ensures  more consistent predictions. 

A central reliability challenge in predictive ML is \emph{predictive multiplicity}: many learning problems admit multiple models with similar performance, yet these models can disagree substantially on individual predictions. This disagreement is not merely a technical nuisance—it translates into arbitrariness for individuals subject to the model’s decision. In this thesis, we show that interventions designed to improve \emph{group fairness} can paradoxically \emph{increase} predictive multiplicity. As a result, a system can appear “good” under accuracy and group fairness metrics while hiding severe instability at the individual level.

This motivates a third evaluation axis—\textbf{arbitrariness}—that complements (rather than replaces) accuracy and group fairness when models are deployed in human-facing decisions. To address this, we develop an \textbf{ensemble-based algorithm} that can be applied on top of existing fairness interventions and that \textbf{provably increases individual-level predictive consistency}. 

This chapter is based on the following paper\citep{long2023individual}:
\begin{itemize}
    \item Carol Long, Hsiang Hsu, Wael Alghamdi, and Flavio Calmon. Individual arbitrariness and group fairness. Advances in Neural Information Processing Systems \textbf{(NeurIPS)}, 2023.
\end{itemize}

\subsection*{Kernel Multiaccuracy}
% %kernel multiaccuracy
Predefined demographic groups often overlook the subpopulations most impacted by model errors, leading to a growing emphasis on data-driven methods that pinpoint where models underperform. The emerging field of multi-group fairness addresses this by ensuring models perform well across a wide range of group-defining functions, rather than relying on fixed demographic categories.

We demonstrate that recently introduced notions of multi-group fairness can be equivalently formulated as \textbf{integral probability metrics (IPM)}. IPMs are the common information-theoretic tool that underlie definitions such as multiaccuracy, multicalibration, and outcome indistinguishably. Building on this connection, we develop a simple and powerful approach to achieving \textbf{multiaccuracy over an infinite-dimensional function class} defined by a \textbf{reproducing kernel Hilbert space (RKHS)}: first perform a kernel regression of a model's errors, then subtract the resulting function from a model's predictions. We combine these results to develop a post-processing method that improves multiaccuracy with respect to bounded-norm functions in an RKHS, enjoys provable performance guarantees, and, in binary classification benchmarks, achieves favorable multiaccuracy relative to competing methods. 

This chapter is based on the following paper\citep{long2025kernel}:
\begin{itemize}
    \item Carol Long, Wael Alghamdi, Alexander Glynn, Yixuan Wu, and Flavio Calmon. Kernel multiaccuracy. In 6th Symposium on Foundations of Responsible Computing (\textbf{FORC} 2025), 2025.
\end{itemize}

Taken together, Chapters 1 and 2 broaden the notion of “reliability” from a single performance metric into a set of complementary desiderata for machine learning models: stability at the individual level (low arbitrariness) and uniformly strong performance across rich, data-driven subpopulations (multiaccuracy).

\subsection*{Watermarking Large Language Models}

Large language models (LLMs) are now able to produce text that is indistinguishable from human-generated content. This has fueled the development of watermarks that imprint a “signal” in LLM-generated text with minimal perturbation of an LLM’s output. 

\paragraph{Information-Theoretic Trade-offs and Optimized Couplings.} We first provide an analysis of text watermarking in a one-shot setting. Through the lens of hypothesis testing with side information, we formulate and analyze the fundamental trade-off between watermark detection power and distortion in generated textual quality. A key design component becomes the \textbf{coupling} between (i) side information available to the detector (e.g., derived from a hash of prior tokens) and (ii) a randomized partitioning of the model vocabulary that induces watermark structure.

Within a worst-case analysis over next-token distributions that satisfy a \textbf{min-entropy constraint}, we characterize the \textbf{optimal coupling and randomization strategy}. This yields a \textbf{closed-form expression for the detection rate} under the proposed scheme and a precise quantification of the detection-distortion trade-off in a max-min sense. Finally, We numerically compare the proposed scheme with the theoretical optimum.

\paragraph{\heavywater{} and \simplexwater{}: Distortion-free LLM Watermarks for Low-Entropy Distributions.}
A practical obstacle for watermarking is that LLM next-token distributions are often \textbf{near-deterministic}: a large fraction of the LLM token distributions place much of the probability mass on a single token. In fact, over 90\% of next-token distributions across Q\&A and coding tasks exhibit the low-entropy property, with more than half of the probability mass on a single token. In this low-entropy regime, many watermarking techniques either become difficult to detect or degrade text quality. 

To address this regime, we introduce an \textbf{optimization framework} for watermark design that explicitly targets how random side information should be used to achieve the best detection--quality trade-off. The framework motivates two new watermarking schemes, \heavywater{} and \simplexwater{}, which are tunable (enabling customization interpolation between detectability and distortion), model-agnostic (applicable to any LLM), and side-information-agnostic (compatible with standard side information generation mechanisms).

We evaluate these methods across several benchmarks, demonstrating \textbf{superior detection with minimal downstream quality degradation} across generation tasks. The theoretical analysis also reveals a surprising structural insight: for widely used \emph{binary scoring} detectors, designing optimally detectable watermarks is equivalent to constructing \textbf{codes with large Hamming distances}. This thesis thus connects watermark design to \textbf{coding theory}, offering both theoretical and algorithmic leverage.

This chapter is based on the following papers\cite{long2025optimized,tsur2025heavywater}:
\begin{itemize}
    \item Carol Long, Dor Tsur, Claudio Mayrink Verdun, Hsiang Hsu, Haim H Permuter, and Flavio Calmon. Optimized couplings for watermarking large language models. In International Symposium of Information Theory (\textbf{ISIT}), 2025.
    \item Dor Tsur, Carol Long, Claudio Mayrink Verdun, Hsiang Hsu, Chen-Fu Chen, Haim Permuter, Sajani Vithana, and Flavio P Calmon. Heavywater and simplexwater: Watermarking low-entropy text distributions. Advances in Neural Information Processing Systems (\textbf{NeurIPS}), 2025.
\end{itemize}

\subsection*{Autonomous Multi-Agent System for Supply Chain Management}

\paragraph{Why agents require a different notion of trust.}
When an LLM is embedded in a loop—observing state, reasoning, taking actions, and affecting downstream state—its probabilistic outputs become \emph{decisions}, and errors can compound over time. Moreover, many high-impact applications require \emph{multiple} agents that coordinate or compete across a shared environment. In such settings, trustworthiness is determined less by any single agent’s capabilities and more by the \textbf{emergent collective dynamics}.

\paragraph{A testbed for autonomous supply chains: the GenAI Beer Game.}
To study these dynamics, we develop the \textbf{first online simulator} in which an entire classical supply chain environment—the Beer Distribution Game—is managed by \textbf{LLM-driven agents} at every facility (retailer, wholesaler, distributor, factory). This testbed enables controlled experimentation on multi-stage, multi-agent decision-making: agents must forecast demand and make sequential replenishment decisions while minimizing system-wide cost.

Using this environment, we identify \textbf{universal strategies and mechanisms} that reliably nudge agent behavior toward better performance. 
Using advanced reasoning models with proper orchestration and guardrails, we show that modern LLM agents can \textbf{outperform human experts}, substantially reducing supply chain costs. 
However, the testbed also exposes critical risks that average-case evaluations miss. In particular, performance can be \textbf{highly variable}, with rare \textbf{tail events} in which a single run incurs dramatically higher costs due to output variability and cascading interactions. These results motivate an agent-specific view of trust: it must include not only mean performance, but also variance, tail risk, and failure analysis for collective behavior in a multi-agent system.

This chapter is based on the following paper\citep{long2025supply}:
\begin{itemize}
    \item Carol Long, David Simchi-Levi, Andre Calmon, and Flavio Calmon. When Supply Chains Become Autonomous. In Harvard Business Review (\textbf{HBR}), 2025.
\end{itemize}

\section{Thesis Organization}

The remainder of this thesis develops these ideas in depth, proceeding in three parts: (i) reliability of predictive models (model bias beyond predefined groups, arbitrariness under fairness constraints), (ii) accountability of generative models (watermarking theory and algorithms, including the low-entropy regime), and (iii) trustworthiness of autonomous multi-agent systems (testbeds and empirical observation of the tail risks). To summarize, as AI systems become more capable and more autonomous, trustworthiness must be engineered at multiple layers: decision consistency for individuals, performance guarantees across rich subpopulations, verifiable provenance for synthetic content, and reliability mechanisms that curtail tail risks for multi-agent systems.

\chapter{Individual Arbitrariness and Group Fairness}
\label{ch:2}

\epigraph{What I call the Rashomon Effect is that there
is often a multitude of different descriptions (...).}{Leo Breiman, \textit{Statistical Modeling: The Two Cultures}\cite{breiman2001statistical}}

\section{Introduction}\label{sec:introduction}

Machine learning models are increasingly used to make high-stakes decisions in domains such as hiring, lending, and criminal justice. In these contexts, ensuring that models are fair and do not perpetuate existing societal biases is of paramount importance. A plethora of fairness-aware learning algorithms have been developed to address this challenge, aiming to satisfy various statistical fairness criteria \citep{hardt2016equality, dwork2015preserving, chouldechova2017fair, pleiss2017fairness}. However, the focus on the trade-off between fairness and accuracy has largely overlooked a critical third dimension: \textbf{arbitrariness}.

Non-arbitrariness is an important facet of non-discriminatory decision-making. Substantial arbitrariness exists in the training and selection of machine learning (ML) models. 
By simply varying hyperparameters of the training process (e.g., random seeds in model training), we can produce models with arbitrary outputs on individual input samples~\citep{breiman2001statistical,marx2020predictive, hsu2022rashomon, cooper2023variance}. 
The phenomenon where distinct models exhibit similar accuracy but arbitrary individual predictions is called \textit{predictive multiplicity}\footnote[1]{In the rest of the chapter, we informally use the terms ``arbitrariness'' and ``predictive multiplicity'' exchangeably to refer to the phenomenon of inconsistent predictions caused by the randomness in model training.}~\citep{marx2020predictive}.  The arbitrary variation of outputs due to unjustified choices made during training can disparately impact individual samples, i.e., predictive multiplicity is not equally distributed across inputs of a model. When deployed in high-stakes domains (e.g., medicine, education, resume screening), the arbitrariness in the ML pipeline may target and cause systemic harm to specific individuals by excluding them from favorable outcomes  \citep{creel2022algorithmic,black2022model,watson2022predictive}. 

Popular fairness metrics in the ML literature do not explicitly capture non-arbitrariness. A widely recognized notion of non-discrimination is \emph{group fairness}. Group fairness is quantified in terms of, for example, statistical parity~\citep{dwork2015preserving}, equal opportunity, equalized odds~\citep{hardt2016equality}, and  variations such as multi-accuracy~\citep{kim2019multiaccuracy} and multi-calibration~\citep{hebert2018multicalibration}. Broadly speaking, methods that control for group fairness aim to guarantee comparable performance of a model across population groups in the data. The pursuit of group fairness has led to hundreds of fairness interventions that seek to control for performance disparities while preserving accuracy \citep{hort2022bia}.

The central question we tackle in this chapter is: Do models corrected for group fairness exhibit less arbitrariness in their outputs? We answer this question in the \emph{negative}. We demonstrate  that state-of-the-art fairness interventions may improve group fairness metrics at the expense of exacerbating arbitrariness. The harm is silent: the increase in arbitrariness is masked by favorable group fairness and accuracy metrics. Our results show that arbitrariness lies beyond the fairness-accuracy frontier: predictive multiplicity should be accounted for \emph{in addition} to usual group-fairness and accuracy metrics during model development.

\begin{figure}[!tb]
\centering
\includegraphics[width=1.0\textwidth]{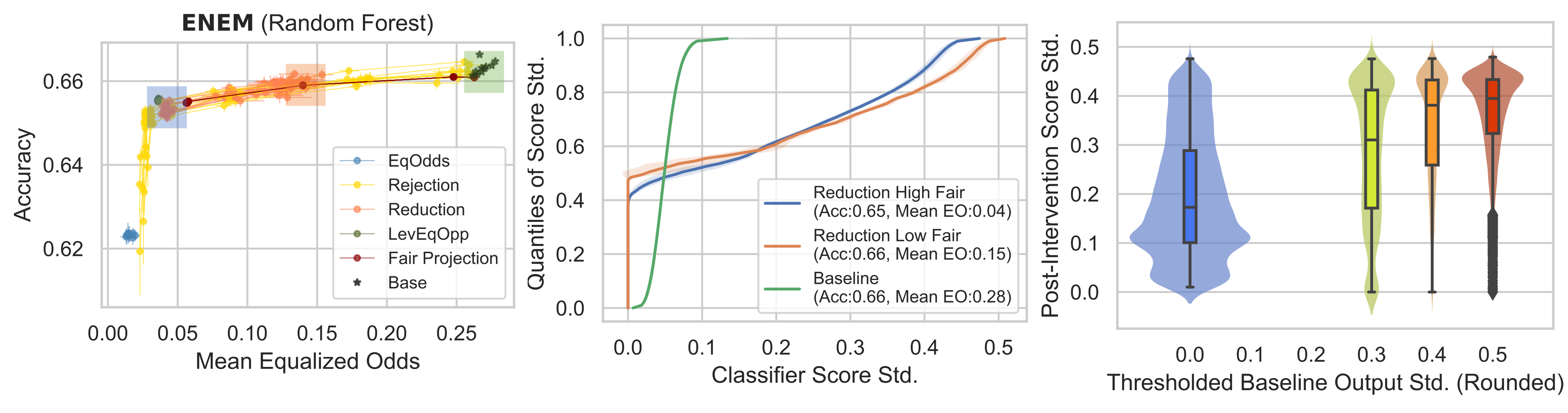}
\caption{\footnotesize
Accuracy-fairness frontier does not reveal arbitrariness in competing models. \textbf{Left}: Fairness-Accuracy frontier of baseline and fair models corrected by 5 fairness interventions; point clouds generated by different random seed choices.
\textbf{Middle}: The cumulative distribution functions (CDF) of per-sample score std. across classifiers at different intervention levels (see Definition~\ref{def:: score SD}). For each sample, std. is measured across competing scores produced by classifiers initialized with different random seeds. A \emph{wider} CDF indicates \emph{more} disparity of the impact of arbitrariness on different individuals. \textbf{Right}: The distribution of score std. relative to the thresholded baseline model. Removing samples that receive very low score std. both from thresholded baseline and fair classifiers, the largest group (blue area) in this violin plot are those individuals for which std. increases from 0 to a large positive value (median around 0.15). Hence, significant arbitrariness is introduced by the fairness intervention, in addition to and separate from the effects of thresholding the baseline.}
\label{fig::Enem quantile}
\end{figure}

Figure~\ref{fig::Enem quantile} illustrates how fairness interventions can increase predictive multiplicity. Here, state-of-the-art fairness interventions are applied\footnote{See Section~\ref{sec:exp} for a detailed description of the experiment and dataset.} to a baseline random forest classifier to ensure group fairness (mean equalized odds \citep{hardt2016equality}, see Definition~\ref{eq::EO})  in a student performance binary prediction task. We produce multiple baseline classifiers by varying the random seed used to initialize the training algorithm. Each baseline classifier achieves comparable accuracy and fairness violation. They also mostly agree in their predictions: for each input sample, the standard deviation of output scores across classifiers is small (see Definition~\ref{def:: score SD}). After applying a fairness intervention to each randomly initialized baseline classifier, we consistently reduce  group fairness violations at a small accuracy cost, as expected. However, predictive multiplicity changes significantly post-intervention: for roughly half of the students, predictions are consistent across seeds, whereas for 20\% of the students, predictions are comparable to a coin flip. For the latter group, the classifier output depends on the choice of a random seed instead of any specific input feature. The increase in predictive multiplicity is masked by the fairness-accuracy curve, does not impact all samples equally, and is consistent across datasets and learning tasks.

At first, the increase in predictive multiplicity may seem counter-intuitive: adding fairness constraints to a learning task should reduce the solution space, leading to less disagreement across similarly-performing classifiers relative to an unconstrained baseline. We demonstrate that, in general, this is not the case. For a given hypothesis class, the non-convex nature of group fairness constraints can in fact \emph{increase} the number of feasible classifiers at a given fairness and accuracy level. We show that this phenomenon occurs even in the simple case where the  hypothesis space is comprised of threshold classifiers over one-dimensional input features, and the optimal baseline classifier is unique. To address this challenge, we demonstrate -- both theoretically and through experiments -- that ensembling classifiers is an effective strategy to counteract this multiplicity increase.

The main contributions of this work include\footnote{Proofs and additional experiments are included in Appendice \ref{apdx:proofs_ch2}.}:
\begin{enumerate}
    \item We demonstrate that the usual ``fairness-accuracy'' curves can systematically mask an increase of predictive multiplicity.  Notably, applying state-of-the-art fairness interventions can incur higher arbitrariness in the ML pipeline.
    \item We show that multiplicity can be arbitrarily high even if group fairness and accuracy are controlled, when we do not have perfect classifiers. Hence,  fairness interventions optimized solely for  fairness and accuracy cannot, in general, control predictive multiplicity. We also provide examples of why fairness constraints may exacerbate arbitrariness.
    \item We propose an ensemble algorithm that reduces multiplicity while maintaining fairness and accuracy. We derive convergence rate results to show that the probability of models disagreeing drops exponentially as more models are added to the ensemble.
    \item We demonstrate the multiplicity phenomena and benchmark our ensemble method through comprehensive experiments using state-of-the-art fairness interventions across real-world datasets.
\end{enumerate}

\subsection{Related Works}
\paragraph{Multiplicity, its implications, and promises.} 
Recent works have investigated various factors that give rise to multiplicity. 
\citet{d2022underspecification} studied how under-specification presents challenges to the credibility of modern machine learning algorithms. 
More precisely, under-specified optimization problems in machine learning admit a plethora of models that all attain similar performance, and which model to deploy in practice may ultimately depend on arbitrary choices of the randomization made during training procedure \citep{bahri2021locally}.
The arbitrariness of the model could potentially harm the reproducibility of model predictions \citep{bhojanapalli2021reproducibility}, and hence the credibility of the conclusion made thereof. 

\citet{creel2022algorithmic} thoroughly explored the notion of arbitrariness in machine learning and discuss how high-multiplicity predictions can lead to systematized discrimination in society through ``algorithmic leviathans.'' %When a single prediction model, out of the Rashomon set, is adopted across domains, certain individuals may be deprived of opportunities across the board.
Multiplicity in prediction and classification can also have beneficial effects.
\citet{black2022model,semenova2019study}, and \citet{fisher2019all} view multiplicity of equally-performing models as an opportunity to optimize for additional criteria such as generalizability, interpretability, and fairness. 
\citet{coston2021characterizing} develop a framework to search over the models in the Rashomon set for a better operation point on the accuracy-fairness frontier. 
However, they do not discuss the potential predictive multiplicity cost of existing fairness interventions nor propose algorithms to reduce this cost.
%\citet{coston2021characterizing} searches over the Rashomon set to get a single model that has a minimal disagreement with other equally good models.

%The works most similar to ours are \citet{cooper2023variance} and \citet{coston2021characterizing}, although neither considers the increase in arbitrariness as a result of fairness intervention methods. 
The work most similar to ours is \citep{cooper2023variance}.
\citet{cooper2023variance} consider the problem of predictive multiplicity as a result of using different splits of the training data. Therein, they quantify predictive multiplicity by prediction variance, and they propose a  bagging strategy \citep{breiman1996bagging} to combine models.
Our work considers a different problem where predictive multiplicity is exacerbated by group-fairness interventions. 
Our work is also different from \citet{cooper2023variance} as we fix the dataset when training models and consider multiplicity due to randomness used during training. % for ensembling and use bootstrapping to plot error bars, rather than combining bootstrapped classifiers. 
In this sense, our ensemble algorithm is actually a voting ensemble \citep{witten2002data} (see Section~\ref{sec:ensemble}); see also ensembling and reconciliation strategies proposed by \citet{black2021selective} and \citet{roth2022reconciling} that aim to create more consistent predictions among competing models. To the best of the authors' knowledge, we are the first to measure and report the arbitrariness cost of  fairness interventions.

\paragraph{Hidden costs of randomized algorithms.}
%https://arxiv.org/pdf/2302.07185.pdf
Recent works~\citep{ganesh2023impact,krco2023mitigating, kulynych2023arbitrary} examine the potential detrimental consequences of randomization in the ML pipeline.
In their empirical study, Ganesh et al.\citep{ganesh2023impact} observe that group fairness metrics exhibit high variance across models at different training epochs of Stochastic Gradient Descent (SGD). The authors point out that random data reshuffling in SGD makes empirical evaluation of fairness (on a test set) unreliable, and they attribute this phenomenon to the volatility of predictions in minority groups. Importantly, they do not incorporate fairness interventions in their experiments. In contrast, we apply fairness interventions to baseline models. Specifically, we examine the variance in predictions among models with similar fairness and accuracy performances. In addition to the observations made by Ganesh et al., our theoretically-grounded study reveals the different paths that lead to group-fairness, i.e., that arbitrariness can be an unwanted byproduct of imposing fairness constraints. \citet{krco2023mitigating} empirically study if fairness interventions reduce bias equally across groups, and examine whether affected groups overlap across different fairness interventions. 
In contrast, our work examines the \emph{multiplicity cost} of group fairness and its tension with individual-level prediction consistency, rather than the \emph{fairness cost} of  randomness in the ML pipeline. 
Another work on the hidden cost of randomized algorithms is given by \citet{kulynych2023arbitrary}, who report that  well-known differentially-private training mechanisms can exacerbate predictive  multiplicity. % due to apply an additive Gaussian noise on either the gradient or model outputs.

In an early work \citep{lipton2018does}, Lipton et al. indirectly points to the potential arbitrary decision on individuals as a result of imposing group fairness constraints. They give an illustrative example using synthetic hiring data to show that a fair model resorts to using irrelevant attribute (hair length) to make hiring decision in order to achieve near-equal hiring rate for men and women.

\section{Problem Formulation}\label{sec:setup}

We explain the setup and relevant definitions in this section.

% \textbf{Prediction tasks.} 
\paragraph{Prediction tasks.} 
We consider a binary  classification setting with training examples being triplets %$(\bX,S,Y)\stackrel{iid}{\sim}P_{\bX,S,Y} $
$(\bX,S,Y)$ with joint distribution $P_{\bX,S,Y} $. Here, $\bX$ is an $\mathbb{R}^d$-valued feature vector, $S$ is a discrete random variable supported on $[K]\defined \{1,\cdots,K\}$  representing $K$ (potentially overlapping) group memberships, and $Y$ is a binary (i.e., $\{0,1\}$-valued) random variable denoting class membership.\footnote{We note that our setup can be readily extended to multi-class prediction.}
We consider probabilistic classifiers in a hypothesis space $\calH$, where each $h\in \calH$ is a mapping $h:\mathbb{R}^d\to [0,1]$. 
Each value of a classifier $h(\bx)$ aims to approximate $P_{Y|\bX=\bx}(1)$. 
The predicted labels $\hat{y}$ can be obtained by thresholding the scores, e.g., $\hat{y} = \indicator \{h(\bx) \ge 0.5\}$, where $\indicator\{\, \cdot\, \}$ is the indicator function. 
Finally, we denote by $\bDelta_c$ the probability simplex over $c$ dimensions. 

\paragraph{Randomized training procedures and the Rashomon set.} 
% We are interested in detecting whether a pool of classifiers (deployed for, and performing similarly in the same prediction task) have conflicting predictions non-uniformly across individuals.
% in measuring the uniformity (across samples) of conflict in the predictions produced by a pool of classifiers. 
We assume access to the following:
\begin{enumerate}
    \item a training dataset of $n$ i.i.d samples $\calD \defined \{(\bx_i, s_i, y_i) \}_{i = 1}^n$ drawn from $P_{\bX,S,Y}$; 
    \item a randomized training procedure $\calT$; and 
    \item an induced distribution $\calT(\calD)$ on the hypothesis class of predictors $\calH$. 
\end{enumerate}
We denote a sampled classifier by $h \sim \calT(\calD)$, which can be sampled, for example, using different random seeds at the beginning of the execution of procedure $\calT$ on $\calD$. For concreteness, the above data may for example correspond to the following practical setting. 
\begin{example}
    The dataset $\calD$  can be comprised of resumes of individuals applying for a job, and the training procedure $\calT$ is an algorithm to predict whether to extend an interview opportunity for an applicant. For example, $\calT$ can be a neural network with unspecified hyperparameters (e.g., random seed that needs to be chosen at the outset); alternatively, $\calT$ can be the same pre-trained neural network composed with a fairness intervention method. The classifiers considered will be the last layer of the neural network (or the classifier after fairness enhancement), which will belong to a hypothesis class $\calH$ determined by the chosen neural network architecture. By varying the random seed, say, $m$ times, we would obtain \emph{independent} classifiers, denoted by $h_1,\cdots,h_m \stackrel{i.i.d.}{\sim} \calT(\calD)$. \exampleend
\end{example}

We are interested in detecting whether competing classifiers (i.e., deployed for, and performing similarly in the same prediction task) have conflicting predictions non-uniformly across individuals. Next, we define the set of competing models obtained from the randomized training procedure $\calT$.

For a loss function $\ell:[0,1]\times \{0,1\}\to \BR^+$, finite dataset $\calD \subset \BR^d\times [K]\times \{0,1\}$, and classifier $h:\BR^d\to [0,1]$, we let the empirical loss incurred by $h$ on $\calD$ be denoted by $\ell(h; \calD) \defined |\calD|^{-1}\sum_{(\bx,s, y)\in \calD}\ell(h(\bx),y)$. The (\emph{empirical}) \emph{Rashomon set} ($\epsilon$-level set) of competing models is defined as the set of models with loss lower than $\epsilon$~\citep{hsu2022rashomon}, i.e., $\calR(\calH,\calD,\epsilon) \triangleq \{h\in \calH \ : \  \ell(h;\calD)\leq \epsilon\}$. We extend the definition of the Rashomon set to take into consideration the effect of the randomized algorithm $\calT$, as follows.

\begin{defn}[Empirical Rashomon Set of Randomized Training Procedure]
Fix a finite dataset $\calD$, a hypothesis class $\calH$, and a randomized training procedure $\calT$ inducing the distribution $\calT(\calD)$ on $\calH$. 
Given a loss function $\ell:[0,1]\times\{0,1\}\to \Reals^+$ and a  parameter $\epsilon>0$, we define the \emph{empirical Rashomon set with $m$ models induced by $\calT$}  as the collection of $m$ classifiers independently sampled from $\calT(\calD)$ and having empirical loss less than $\epsilon$:
\begin{equation}
   \hatRm(\calT, \calD, \epsilon) \triangleq \left\{ h_1, \cdots, h_m \in \calH \ : \  h_1, \cdots, h_m \stackrel{i.i.d.}{\sim} \calT(\calD) \text{ and } \ell(h_j;\calD) \le \epsilon \ \forall j \in [m] \right\}.  
\end{equation}
\end{defn}

Here, $\epsilon$ is an approximation parameter that determines the size of the set. The set $\hatRm(\calT, \calD,\epsilon)$ can be viewed as an approximation of the \emph{Rashomon set} of ``good'' models \citep{breiman2001statistical,watson2022predictive,marx2020predictive}, and indeed we have the inclusion $\hatRm(\calT, \calD,\epsilon) \subset \calR(\calH, \calD,\epsilon)$ where $\calH=\supp(\calT(\calD))$.  Note that the set $\hatRm(\calT,\calD,\epsilon)$ is itself \emph{random} even for a fixed dataset $\calD$, where the source of randomness is coming from the distribution $\calT(\calD)$. In the sequel, we omit the arguments of $\hatRm(\calT, \calD,\epsilon)$ when they are clearly implied from context.

There are various metrics to quantify predictive multiplicity across models in  $\hatRm$  by either considering their output scores \citep{watson2022predictive} or thresholded predictions \citep{marx2020predictive}.
We focus on two metrics: 1)~\emph{ambiguity} for evaluating the predictive multiplicity of thresholded predictions, and 2)~\emph{cumulative distribution function (CDF) of standard deviation (std.) of output scores} when model outputs are in the interval $[0,1]$ (interpreted as the probability of the positive class). Those two metrics are defined %(for arbitrary finite sets of models $\calR$) 
as follows. 

\begin{defn}[Ambiguity \citep{marx2020predictive}] 
\label{def:: ambiguity}
    Fix a dataset $\calD = \{(\bx_i, s_i, y_i) \}_{i\in [n]} \subset \BR^d\times [K] \times \{0,1\}$ and a finite set of models $\calR \subset \calH$. Let $f(r) \defined \indicator\{r \ge 0.5\}$ be the thresholding function. The \emph{ambiguity} of a dataset over the set of models $\calR$ is the proportion of points in the dataset that can be assigned a conflicting prediction by a competing classifier within $\calR$:
    \begin{equation}
        \alpha\left(\calD,\calR\right) \triangleq \frac{1}{|\calD|}\sum_{i \in [n]} \ \max_{h,h'\in \calR} \ \indicator \left\{ f(h(\bx_i)) \neq f(h'(\bx_i)) \right\}.
    \end{equation}
\end{defn}

To define the CDF of std. of scores, we first delineate what we mean by empirical std. of scores.
\begin{defn}[Std. of Scores] 
\label{def:: score SD}
    Fix a finite set of models $\calR = \{h_j\}_{j\in [m]}\subset \calH$. The empirical standard deviation (std.) of scores for a sample $\bx\in \BR^d$ relative to $\calR$ is defined by 
    \begin{equation}
        s(\bx,\calR) \triangleq \sqrt{\frac{1}{m-1}\sum_{j\in [m]} (h_j(\bx)-\bar{\mu}_{\bx})^2},
    \end{equation}
    where $\bar{\mu}_{\bx} \triangleq \frac1m \sum_{j\in [m]} h_j(\bx)$ denotes the empirical mean (over $\calR$) of the scores.
\end{defn}
Further, to understand the std. of scores of the population, we consider the empirical cumulative distribution function 
of the std. of the scores, defined as follows.
\begin{defn}[Quantiles of std. of Scores]
    Fix a dataset $\calD = \{(\bx_i, s_i, y_i) \}_{i\in [n]} \subset \BR^d\times [K] \times \{0,1\}$ and a finite set of models $\calR \subset \calH$. We define the empirical cumulative distribution function of the std. of the scores by (where $s(\bx,\calR)$ is the empirical std. as in Definition~\ref{def:: score SD})
    \begin{equation}
        \hat{F}_{\calD,\calR}(t) \triangleq \frac{1}{|\calD|}\sum_{i \in [n]} \ \indicator \left\{s(\bx_i,\calR) \leq t \right\}.
    \end{equation}
\end{defn}

\begin{example}
    Consider a resume screening task where the algorithm decides whether to extend an interview opportunity. If  $\hat{F}_{\calD,\calR}(0.5)=90\%$, then for 10\% of the individuals in the dataset, the predictions produced by the competing models are arbitrary and conflicting: regardless of the mean scores, with an std. of at least 0.5, there would exist models with scores falling above and below the one-half threshold, so the thresholded output can be both 0 (no interview) and 1 (offer interview). \exampleend
\end{example}

\paragraph{A note on related metrics.} 
An alternative measurement of score variation is \emph{Viable Prediction Range} as defined in~\citep{watson2022predictive}, which measures the difference in max and min scores among competing models on each individual. For thresholded scores, the original definition of \emph{Ambiguity} \citep{marx2020predictive} considers the proportion of a flip in prediction with respect to a baseline model (from empirical risk minimization with fixed hyperparameters and randomness). Since we consider randomized training procedures with no clear baseline model, the definition for ambiguity above is a variation of the original.

\paragraph{Group fairness.} 
We consider three  group fairness definitions for classification tasks---statistical parity (SP), equalized odds (EO), and overall accuracy equality (OAE) \citep{dwork2015preserving,hardt2016equality,pleiss2017fairness,chouldechova2017fair}. OAE and Mean Equalized Odds (MEO) are defined below as they are used in the next sections, and we refer the reader to Appendix~\ref{sec:: discussion fairness} for the remaining definitions. 

\begin{defn}[Overall Accuracy Equality, OAE] \label{def::OAE} Let $\widehat{Y}$ be the predicted label obtained, e.g., from thresholding the scores of a classifier $h:\BR^d\to [0,1]$. 
    The predictor $\widehat{Y}$ satisfies overall accuracy equality (OAE) if its accuracy is independent of the group attribute: for all groups $s,s'\in [K]$,
    \begin{equation}
        \Pr(\widehat{Y} = Y \mid S = s) = \Pr(\widehat{Y} = Y \mid S = s').
    \end{equation} %, and all labels $a\in [c]$ . 
\end{defn}

For binary classification, SP boils down to requiring the average predictions to be equal across groups, while EO requires true positive rates (TPR) and false positive rates (FPR) to be calibrated. In this chapter, we consider mean equalized odds (MEO): the average of absolute difference in FPR and TPR for binary groups $S\in \{0,1\}$. We consider binary group since this is the setup for most fairness intervention methods. 
\begin{defn}[Mean Equalized Odds, MEO~\citep{hardt2016equality, bellamy2019ai}] Let $\widehat{Y}$ be the predicted label, $S\in \{0,1\}$ denotes binary group membership. Mean Equalized Odds is the average odds difference for binary groups:

\begin{equation} \label{eq::EO}
    \textsc{MEO} \defined \frac{1}{2}\left( |\textsc{TPR}_{S=0}-\textsc{TPR}_{S=1}| +|\textsc{FPR}_{S=0}-\textsc{FPR}_{S=1}| \right),
\end{equation}
where $\textsc{TPR}_{S=s} \defined \Pr(\widehat{Y} = 1 \mid Y=1, S = s)$ and $\textsc{FPR}_{S=s} \defined \Pr(\widehat{Y} = 1 \mid Y=0, S = s)$.
\end{defn}

To examine whether current fairness intervention methods lead to an exacerbation of multiplicity, we survey state-of-the-art intervention methods, including Reductions~\citep{agarwal2018reductions}, Fair Projection~\citep{alghamdi2022beyond}, Reject Options~\citep{kamiran2012decision}, and EqOdds~\citep{hardt2016equality}. We offer a brief discussion of their mechanism in Appendix \ref{sec:: discussion fairness}.

\section{Orthogonality of Fairness and Arbitrariness}

We discuss next why arbitrariness is a third axis not captured by fairness and accuracy. Models with similar fairness and accuracy metrics can differ significantly in predictions. Moreover, a set of fair and approximately accurate models can attain maximal predictive multiplicity. 
We also explore through an example one fundamental reason why adding fairness constraints can lead to more arbitrariness. 

\begin{wrapfigure}{R}{0.5\textwidth}
    \begin{center}
        \includegraphics[width=0.48\textwidth]{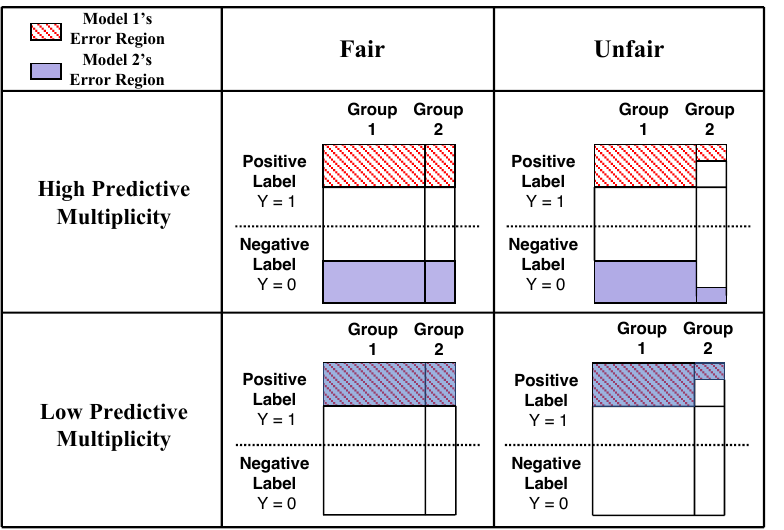}
    \end{center}
    \caption{\footnotesize Illustration on two models being fair/unfair and exhibit high/low predictive multiplicity through the models' error regions in each of the 4 cases. The metrics for fairness and predictive multiplicity are Overall Accuracy Equality (Definition~\ref{def::OAE}) and ambiguity (Definition~\ref{def:: ambiguity}), respectively.}
\label{fig::diagram}
\end{wrapfigure}

\begin{example}[Ambiguity $\neq$ OAE]
 Overall Accuracy Equality (OAE, Definition \ref{def::OAE}) does not capture the ambiguity of model outputs (Definition  \ref{def:: ambiguity}). Consider two hypothetical models that are fair/unfair and exhibit high/low predictive multiplicity in Figure \ref{fig::diagram}. Here, in each panel, the rectangle represents the input feature space and the shaded regions represent the error region of each model.

In the top left panel, both Model 1 and 2 have equal accuracy for both groups, since the proportion of the error regions (red stripes and pink shade) for both groups are the same. Hence, both models are considered group-fair in terms of OAE. However, the error regions of the two models are disjoint. Since ambiguity is measured by the percentage of the samples that receive conflicting predictions from either models, samples from the union of the two error regions contribute to ambiguity. Hence, Model 1 and 2 bear high predictive multiplicity despite being group-fair. 

In the lower right panel, Model 1 and 2 attain low fairness and low predictive multiplicity. Both models have higher accuracy for Group 2 than Group 1, so they are both unfair. The error regions completely overlap, which means that the two models are making the same error---ambiguity is 0. \exampleend
\end{example}

The schematic diagram in Figure~\ref{fig::diagram} shows that predictive multiplicity is not captured by OAE. Indeed, ambiguity of a collection of models is a \emph{global} property (i.e., verified at the collection level), whereas OAE is a \emph{local} property (i.e., verified at the classifier level). Hence, one should not \emph{a priori} expect that a set of competing models each satisfying OAE would necessarily comprise a Rashomon set with favorable ambiguity. 

We prove the orthogonality of OAE and Statistical Parity (SP) from ambiguity formally in the proposition below, where we show that it is possible to construct classifiers with very stringent accuracy and perfect fairness constraint, albeit with maximal ambiguity. We determine the Rashomon set using the 0-1 loss: 
\begin{equation*}
    \ell_{\text{0-1}}(h;\calD) = \frac{1}{|\calD|} \sum_{(\bx_i,s_i,y_i)\in \calD} \ \indicator\left\{ \hat{y}_i \neq y_i \right\},
\end{equation*}
where $\hat{y}_i\in \{0,1\}$ is the class membership of $\bx_i$ predicted by $h$. We prove the following orthogonality in Appendix~\ref{apdx:proofs_ch2}. 

\begin{proposition}[Orthogonality of OAE/SP and Ambiguity] \label{proposition: orthogonality OAE}
    Fix any empirical loss value $0<\epsilon \le \frac12$ and any number of models $m>\frac{1}{\epsilon}$. Then, for some finite dataset $\calD\subset \BR^d \times [K] \times \{0,1\}$, there is a realization of the empirical Rashomon set $\hat{\calR}_m = \{h_j\}_{j\in [m]}$ satisfying the following simultaneously:
    \begin{enumerate}
        \item Each $h_j$ has 0-1 loss upper bounded by $\ell_{\textup{0-1}}(h_j;\calD)\le \epsilon$;

        \item Each $h_j$ satisfies OAE perfectly, or each $h_j$ satisfies SP perfectly; 

        \item The collection $\hat{\calR}_m$ has the worst ambiguity, i.e., $\alpha(\calD,\hat{\calR}_m)=100\%$. 
    \end{enumerate}
\end{proposition}
\begin{remark} \label{remark:OAE orthogonality}
    For OAE, such Rashomon set $\hat{\calR}_m$  exists for \emph{any} dataset $\calD$ satisfying the two conditions:
    \begin{enumerate}
        \item with $n_k$ denoted the number of samples in $\calD$ belonging to group $k\in [K]$, the greatest common divisor of the $n_k$ is at least $(m-1)/(m\epsilon - 1)$. 

        \item if $(\bx,s,y),(\bx,s',y')\in \calD$ share the same feature vector $\bx$, then $y=y'$ too.
    \end{enumerate} 
    The requirement that the $n_k$ share a large enough common divisor is used in the proof to guarantee \emph{perfect} OAE. One could relax this requirement at the cost of nonzero OAE violation.
\end{remark}

Proposition~\ref{proposition: orthogonality OAE} implies that there exists a dataset and competing models for which all samples receive conflicting predictions. Specifically, we can construct a large enough set of competing models ($m>\frac{1}{\epsilon}$) such that 100\% of the samples in the dataset can receive conflicting predictions from this set of perfectly fair models with respect to OAE. 

In the next example, we demonstrate that, counter-intuitively, adding a fairness constraint can enlarge the set of optimal models, thereby increasing predictive multiplicity. This points to a fundamental reason why adding fairness constraints can lead to more arbitrariness in model decisions.

\begin{figure}[!tb]
\centering
\includegraphics[width=0.75\textwidth]{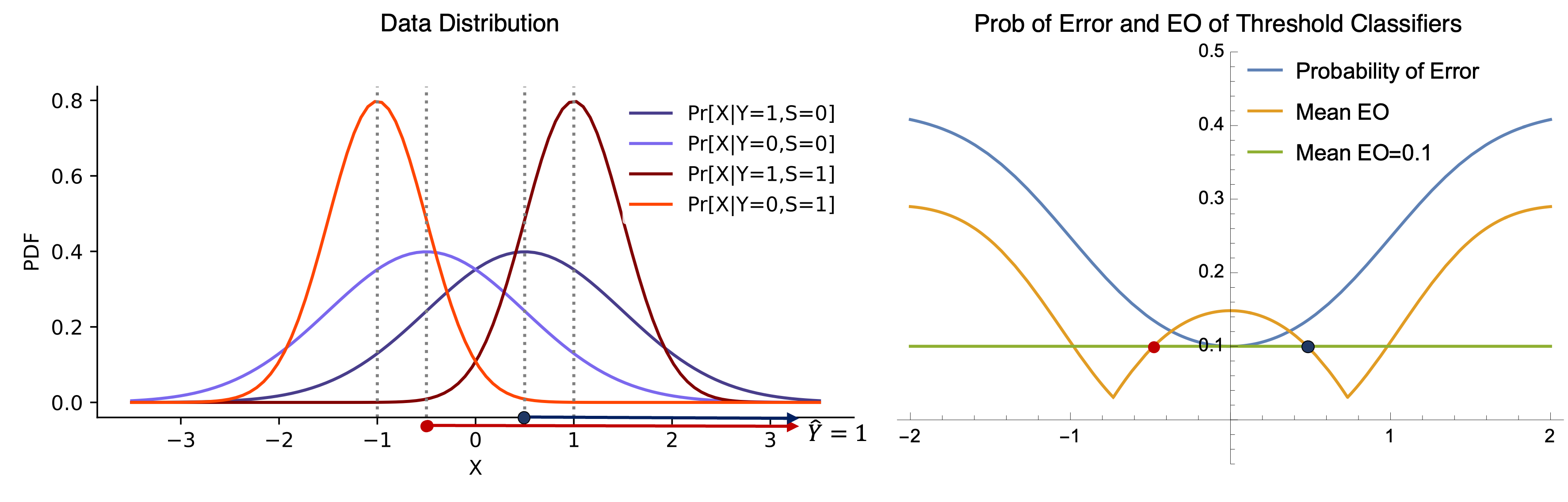}
\caption{\footnotesize
Data distribution of a population with two groups used in Example 2 (\textbf{Left}). In \textbf{Right}, without the \textrm{Mean EO} constraint~\eqref{eq::EO} (green line), there is a unique optimal classifier (with threshold 0) that attains the smallest probability of error (blue line). Adding the \textrm{Mean EO} constraint enlarges the set of optimal threshold classifiers to two classifiers (red and blue dots) with  indistinguishable accuracy and fairness levels (\textbf{Right}) but different decision regions. We illustrate the decision regions of each classifier as red and blue arrows on the \textbf{Left}.}
\label{fig::normal example}
\end{figure}

\begin{example}[Arbitrariness of Threshold Classifiers with Fairness Constraint] Given a data distribution of a population with two groups (Figure \ref{fig::normal example} \textbf{Left}), we want to build a threshold classifier that predicts the true label. Without fairness consideration, the optimal threshold is 0 -- i.e., assigning positive predictions to samples with $X>0$ and negative predictions to $X \leq 0$ minimizes the probability of error (Figure \ref{fig::normal example} \textbf{Right}). This optimal model is unique, thus there is no arbitrariness in model predictions. Suppose we add a fairness constraint that requires \textrm{Mean EO}$\leq 0.1$, the previously optimal classifier at 0 (with \textrm{Mean EO}=0.15, \textbf{Right}) does not meeting the fairness criteria. Searching over the threshold classifiers that minimizes the probability of error while lying \textrm{Mean EO}below the  constraint yields two equally optimal models (red and blue dots \textbf{Right}) with distinct decision regions (red and blue arrows \textbf{Left}). 
Even in this simple hypothesis class, the addition of fairness constraints yields multiple models with indistinguishable fairness and accuracy but with distinct decision regions. The arbitrary selection between these points can lead to arbitrary outputs to points near the boundary.
\end{example}

\section{Ensemble Algorithm for Arbitrariness Reduction}\label{sec:ensemble}

To tackle the potential arbitrariness cost of fairness intervention algorithms, we present a disparity-reduction mechanism through ensembling. We provide theoretical guarantees and numerical benchmarks to demonstrate that this method significantly reduces the predictive multiplicity of fair and accurate models. 

In a nutshell, given competing models $h_1,\cdots,h_m$, we argue that the disparity in their score assignment can be reduced by considering a convex combination of them, defined as follows. 
\begin{defn}[Ensemble Classifier] \label{def:ensemble}
    Given $m$ classifiers $\{ h_1,\cdots,h_m:\BR^d\to [0,1]\}$ and a vector $\blambda \in \bDelta_m$, we define the \emph{$\blambda$-ensembling} of the $h_j$ to be the convex combination $\bh^{\ens,\blambda}\defined \sum_{j\in [m]} \lambda_j h_j$.
\end{defn}

\subsection{Concentration of Ensembled Scores}

We prove in the following result that \emph{any} two different ensembling methods agree for fixed individuals with high probability. Recall that we fix a dataset $\calD$ and a set of competing models $\calT(\calD)$ coming from a stochastic training algorithm $\calT$ (see Section~\ref{sec:setup}). All proofs are provided in Appendix~\ref{apdx:proofs_ch2}.

\begin{theorem}[Concentration of Ensembles' Scores] \label{Thm:concentration}
Let $h_1,\hdots,h_m;\tilde{h}_1,\hdots,\tilde{h}_m \overset{iid}{\sim} \calT(\calD)$ be $2m$ models drawn from $\calT(\calD)$, and $\bh^{\ens,\blambda},\tilde{\bh}^{\ens,\bgamma}$ be the ensembled models (constructed with $\{h_1,\hdots,h_m\}$ and $\{\tilde{h}_1,\hdots,\tilde{h}_m\}$ respectively) for $\blambda,\bgamma \in \bDelta_m$ (see Definition~\ref{def:ensemble}) satisfying $\|\blambda\|_2^2,\|\bgamma\|_2^2\le c/m$ for an absolute constant $c$. 
For every $\bx\in \BR^d$ and $\nu \ge 0$, we have the exponentially-decaying (in $m$) bound
\begin{equation}\label{eq:score_variation_prob}
    \BP\left( \left| \bh^{\textup{ens},\blambda}(\bx) - \tilde{\bh}^{\textup{ens},\bgamma}(\bx) \right| \ge \nu \right) \le 4 e^{-\nu^2m/(2c)}.
\end{equation}
In particular, for any validation set $\calD_{\textup{valid.}}\subset \BR^d$ of size $|\calD_{\textup{valid.}}|=n$, we have the uniform bound
\begin{equation}
    \BP\left( \left| \bh^{\textup{ens},\blambda}(\bx) - \tilde{\bh}^{\textup{ens},\bgamma}(\bx) \right| < \nu \ \text{ for all } \bx\in \calD_{\textup{valid.}}\right) > 1 - 4 n e^{-\nu^2 m /(2c)}.
\end{equation}
\end{theorem}

\subsection{Concentration of Predictions Under Ensembling}

The above theorem implies that we can have a dataset of size that is exponential in the number of accessible competing models and still obtain similar scoring for \emph{any} two ensembled models (uniformly across the dataset). 

In practice, one cares more about the agreement of the final prediction of the classifiers. The following result extends Theorem~\ref{Thm:concentration} to the concentration of thresholded classifiers. For this, we need to define the notion of \emph{confident classifiers}.

\begin{defn}[Confident Classifier] \label{def:strong}
    Fix a probability measure $P_{\bX}$ over $\BR^d$ and constants $\delta,\theta\in [0,1]$. We say that a classifier $h:\BR^d\to [0,1]$ is \emph{$(P_{\bX},\delta,\theta)$-confident} if $\BP\left( \left| h(\bX) - \frac12 \right| < \delta \right) < \theta$. 
    % \begin{equation}
    %     \BP\left( \left| h(\bX) - \frac12 \right| < \delta \right) < \theta.
    % \end{equation}
\end{defn}
In other words, $h$ is a confident classifier if it is ``more sure'' of its predictions. 
We observe in experiments that models corrected by fairness interventions have scores concentrated around 0 and 1. 

Using confident classifiers, we are able to extend Theorem~\ref{Thm:concentration} to thresholded ensembles, as follows.

\begin{theorem} \label{thm::score variation}
    Let $\bh^{\ens,\blambda},\tilde{\bh}^{\ens,\bgamma}$ be as in Theorem~\ref{Thm:concentration}, and assume that both ensembled classifiers are $(P_{\bX},\delta,\theta)$-confident in the sense of Definition~\ref{def:strong}. Let $f(t)\defined \indicator \{t\ge 0.5\}$ be the thresholding function. For any set $\calD_{\textup{valid.}} \subset \BR^d$ of size $|\calD_{\textup{valid.}}|=n$, we may guarantee the probability of agreement in the predictions for all samples under the two ensembles to be at least
    \begin{equation}
        \BP\left( f(\bh^{\ens,\blambda}(\bx)) = f(\tilde{\bh}^{\ens,\bgamma}(\bx)) \ \text{ for every } \bx\in \calD_{\textup{valid.}} \right) \ge 1 - \left( 4 e^{-2\delta^2 m/c  }+2\theta \right)n.
    \end{equation}
\end{theorem}
We note that in the fairness-intervention setting, the set $\calD_{\textup{valid.}}$ in the above theorem would be chosen as the subset of samples having the same group attribute. Thus, the size $n_0$ of $\calD_{\textup{valid.}}$ would be significantly smaller than the total size of the dataset, and the parameter $\theta$ then can be required to be moderately small. 
\begin{remark}
    In Appendix \ref{sec:: optimize param}, we discuss how to optimize the ensembling parameters $\blambda$. In the next section, we will stick to the uniform ensembling: $\bh^{\ens,\blambda} = \frac{1}{m}\sum_{j \in [m]} h_j$, i.e., $\blambda = \frac{1}{m}\mathbf{1}$. This simple uniform ensemble suffices to illustrate the main goal of this chapter: that arbitrariness can be a by-product of fairness intervention methods, and ensembling can mitigate this unwanted effect.
\end{remark}

\section{Experimental Results}\label{sec:exp}
We present empirical results to show that arbitrariness is masked by favorable group-fairness and accuracy metrics for multiple fairness intervention methods, baseline models, and datasets \footnote{Code can be found at \url{https://github.com/Carol-Long/Fairness_and_Arbitrariness}}. We also demonstrate the effectiveness of the ensemble in reducing the predictive multiplicity of fair models. 

\paragraph{Setup and Metrics.}
We consider three baseline classifiers (\textsc{Base}): random forest (RF), gradient boosting (GBM), and logistic regression (LR), implemented by Scikit-learn \citep{pedregosa2011scikit}. By varying the random seed, we obtain 10 baseline models with comparable performance. Then, we apply various state-of-the-art fairness methods (details in Appendix \ref{sec:: discussion fairness}) on the baseline models to get competing fair models.

On the test set, we compute mean accuracy, \textrm{Mean EO} (Definition \ref{eq::EO}), and predictive multiplicity levels on competing models before and after fairness interventions. We use ambiguity (Definition \ref{def:: ambiguity}) and score standard deviations (Definition \ref{def:: score SD}) as metrics for predictive multiplicity.

\paragraph{Datasets.}
We report predictive multiplicity and benchmark the ensemble method on three datasets -- two datasets in the education domain: the high-school longitudinal study (HSLS) dataset \citep{ingels2011high,jeong2022fairness} and the ENEM dataset
 \citep{cury2022instituto} (see \citet{alghamdi2022beyond} Appendix B.1), and the UCI Adult dataset\citep{lichman2013uci} which is based on the US census income data. The ENEM dataset contains Brazilian college entrance exam scores along with student demographic information and socio-economic questionnaire answers (e.g. if they own a computer). After pre-processing, the dataset contains ~1.4 million samples with 139 features. Race is used as the group attribute $S$, and Humanities exam score is used as the label $Y$. Scores are quantized into two classes for binary classification. The race feature $S$ is binarized into White and Asian ($S = 1$) and others ($S = 0$). The experiments are run with a smaller version of the dataset with 50k samples. Complete experimental results can be found in Appendix \ref{sec:: additional experiments}.

\begin{figure}[!tb]
\centering
\includegraphics[width=1.0\textwidth]{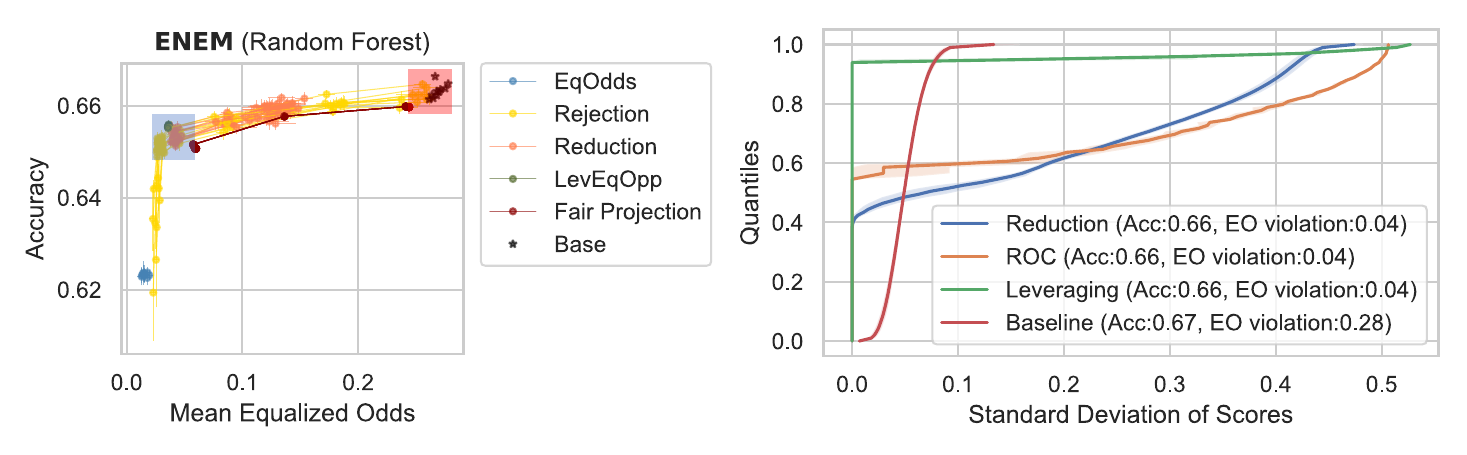}
\caption{\footnotesize
Quantile plot on high-fairness bin for various fairness interventions v.s. baseline on ENEM. \textbf{Left}: Fairness-Accuracy frontier. \textbf{Right}: Fair models produce larger score std. at top percentiles compared to the baseline model (horizontal axis computed via~\eqref{eq::EO}).  (\textsc{Rejection} and \textsc{Leveraging} output thresholded scores directly.)}
\label{fig::enem_multiple}
\end{figure}

\paragraph{Results that Reveal Arbitrariness.} 
We juxtapose the fairness-accuracy frontier and metrics for predictive ambiguity to reveal arbitrariness masked by favorable group-fairness and accuracy metrics in Figure \ref{fig::Enem quantile} and \ref{fig::enem_multiple}. Starting with 10 baseline classifiers by varying the random seed used to initialize the training algorithm, we apply the fair interventions \textsc{Reduction} \citep{agarwal2018reductions}, \textsc{Rejection} \citep{kamiran2012decision}, \textsc{Leveraging} \citep{chzhen2019leveraging} to obtain point clouds of models with comparable fairness and accuracy metrics. 
In Figure \ref{fig::enem_multiple}, we take models that achieve very favorable accuracy and \textrm{MEO} metrics (in blue rectangle in \textbf{Left}) and plot the std. of scores to illustrate predictive multiplicity \textbf{Right}. Group fairness violations are greatly reduced (from 0.28 in baseline to 0.04 in fair models) at a small accuracy cost (from 67\% in baseline to 66\% in fair models). 
However, there is higher arbitrariness. 

Compared to baseline (red curve), fair models corrected by \textsc{Reduction} and \textsc{ROC} produce lower score arbitrariness for the bottom 50\% but much higher arbitrariness for the top 50\% of samples; importantly, the induced arbitrariness becomes \emph{highly nonuniform across different individuals} after applying the two fairness intervention. We observe that \textsc{Leveraging} produce models that agree on ~90\% of the samples, thereby not inducing concerns of arbitrariness.

Remarkably, arbitrariness does not vary significantly among models with different fairness levels. We consider two sets of models trained with high and low fairness constraints using \textsc{Reduction} in Figure \ref{fig::Enem quantile}.

\begin{figure}[!tb]
\centering
\includegraphics[width=0.9\textwidth]{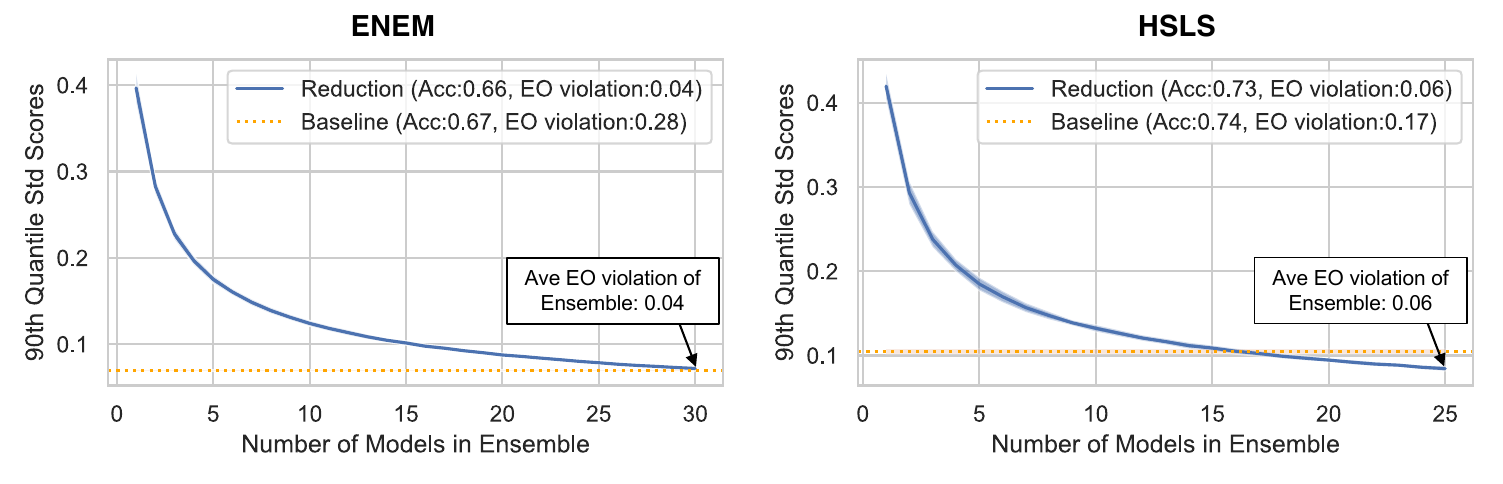}
 \caption{\footnotesize
 Standard deviation of ensembled models trained on ENEM and HSLS with baseline random forest classifiers. We fix the high-fairness bin and vary the number of models $m$ in each ensemble. As we increase the number of ensembles, score std. (on 10 ensembles) drops and meets the score std. of 10 baseline RFC when $m=30$ on ENEM and $m = 17$ on HSLS. (\textrm{Mean EO} is computed using~\eqref{eq::EO}.}
 \label{fig::ensemble convergence}
\end{figure}

\paragraph{Results on the Effectiveness of Ensembling.}
We pair our proofs in Section 4 with experiments that demonstrate the concentration of scores of ensembled models. In Figure \ref{fig::ensemble convergence} \textbf{Left}, taking the competing models in the high-fairness bins corrected with \textsc{Reduction} that achieve an \textrm{Mean EO} violation of $0.04$ but very high score std. for half of the samples (blue rectangle in Figure \ref{fig::enem_multiple}), we ensemble the models with increasing number of models per ensemble ($m$) ranging from 1 to 30. For each $m$, we measure std. of scores in 10 such ensembles. The top percentile std. of the ensembled fair models drops to baseline with 30 models. Similar convergence occur on the HSLS dataset. Importantly, the ensembled models are still fair, the \textrm{Mean EO} violations of the ensembled models remain low.

\section{Conclusion}

We demonstrate in this chapter that arbitrariness is a facet of responsible machine learning that is orthogonal to existing fairness-accuracy analyses. Specifically, fairness-vs-accuracy frontiers are insufficient for detecting arbitrariness in the predictions of group-fair models: two models can have the same fairness-accuracy curve while at the same time giving widely different predictions for subsets of individuals. We demonstrate this undesirable phenomenon both theoretically and experimentally on state-of-the-art fairness intervention methods. Furthermore, towards mitigating this arbitrariness issue, we propose an ensemble algorithm, where a convex combination of several competing models is used for decision-making instead of any of the constituent models. We prove that the scores of the ensemble classifier concentrate, and that the ensuing predictions can be made to concentrate under mild assumptions. Importantly, we exhibit via real-world experiments that our proposed ensemble algorithm can reduce arbitrariness while maintaining fairness and accuracy.

\paragraph{Limitations.}
The proposed framework for estimating the predictive multiplicity of fairness interventions requires re-training multiple times, limiting its applicability to large models. 
We consider model variation due to randomness used during training. In practice, competing models may exist due to inherent uncertainty (i.e., a non-zero confidence interval) when evaluating model performance on a finite test set. In this regard, models with comparable average performance can be produced by searching over this Rashomon set even if training is deterministic (e.g., equivalent to solving a convex optimization problem). 
\paragraph{Future directions.}
An interesting future direction is to explore the multiplicity cost of fairness interventions in such deterministic settings.
Furthermore, our ensembling strategy may not guarantee that the ensemble classifier ensures fairness constraints due to the non-convex nature of such constraints. Though we empirically observe that fairness constraints are indeed satisfied by the ensemble model, proving such guarantees theoretically would be valuable.

\paragraph{Societal impacts.}
While fairness intervention algorithms can effectively reduce the disparate impact among population groups, they can induce predictive multiplicity in individual samples. The increase in predictive multiplicity does not impact all individuals equally. 
If predictive multiplicity caused by fairness interventions is not accounted for, some individuals will bear the brunt of arbitrary decision-making---their predictions could be arbitrary upon re-training the classifier using different random initializations, leading to another level of disparate treatment to certain population groups.

\chapter{Multi-Group Fairness: Kernel Multiaccuracy}\label{ch:3}

\section{Introduction}

Machine learning (ML) models can be inaccurate or miscalibrated on underrepresented population groups defined by categorical features such as race, religion, and sex \cite{barocas2023fairness}. 
Equitable treatment of groups defined by categorical features is a central aspect of the White House's ``Blueprint for an AI Bill of Rights'' \cite{hine2023blueprint}. 
Over the past decade, hundreds of  fairness metrics and interventions  have been introduced to quantify and control an ML model's  performance disparities   across pre-defined population groups \cite{friedler2019comparative,hort2022bia}. Examples of group-fairness-ensuring  interventions include post-processing \cite{hardt2016equality,kamiran2012decision,alghamdi2022beyond} or retraining  \cite{agarwal2018reductions} a model.

Although common, using pre-determined categorical features for measuring ``fairness'' in ML poses several limitations.  Crucially, we design group attributes based on our preconception of where discrimination commonly occurs and whether group-denoting information can be readily measured and obtained. A more complex structure of unfairness can easily elude group-fairness interventions. For instance, \cite{kearns2018preventing} demonstrates that algorithms designed to ensure fairness on binary group attributes can be maximally unfair across more complex, intersectional groups---a phenomenon termed ``fairness gerrymandering.'' Recently, \cite{long2023individual} shows that group fairness interventions do not control for---and may exacerbate---arbitrary treatment at the individual and subgroup level.

\begin{figure*}[!tb]
  \centering
  \resizebox{\linewidth}{!}{\begin{tabular}{ccc}
     \includegraphics[]{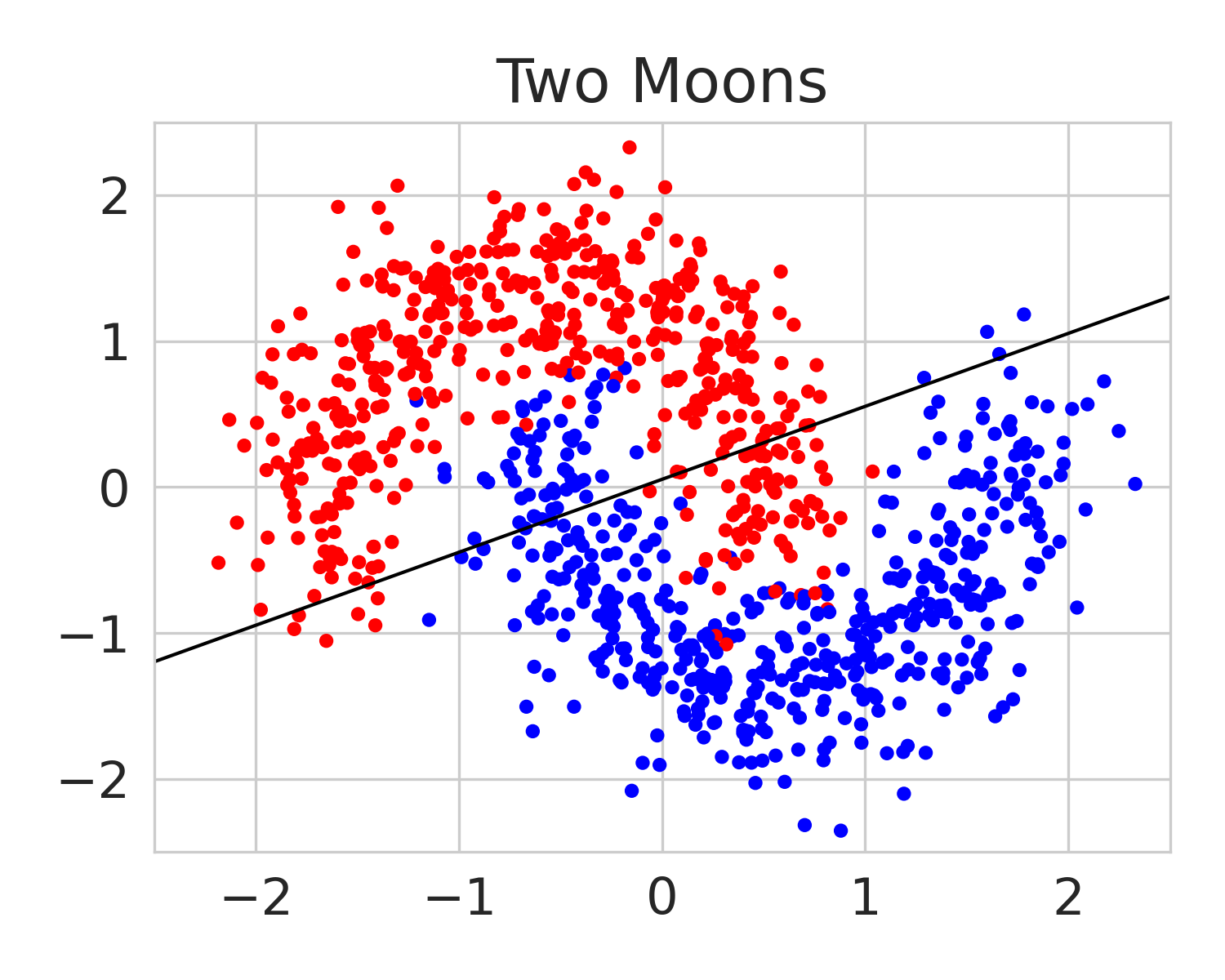} &
     \includegraphics[]{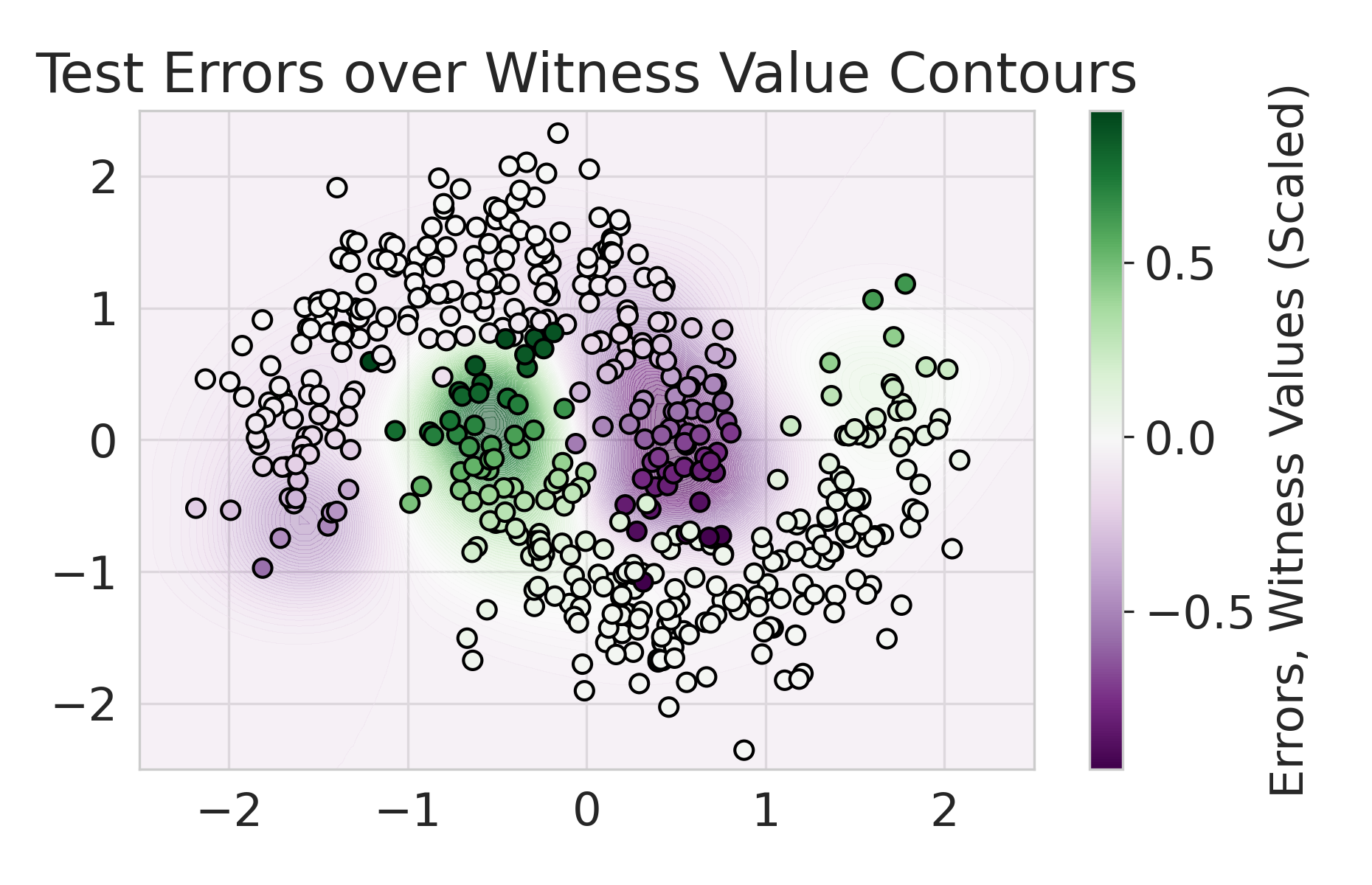} &
    \includegraphics[]{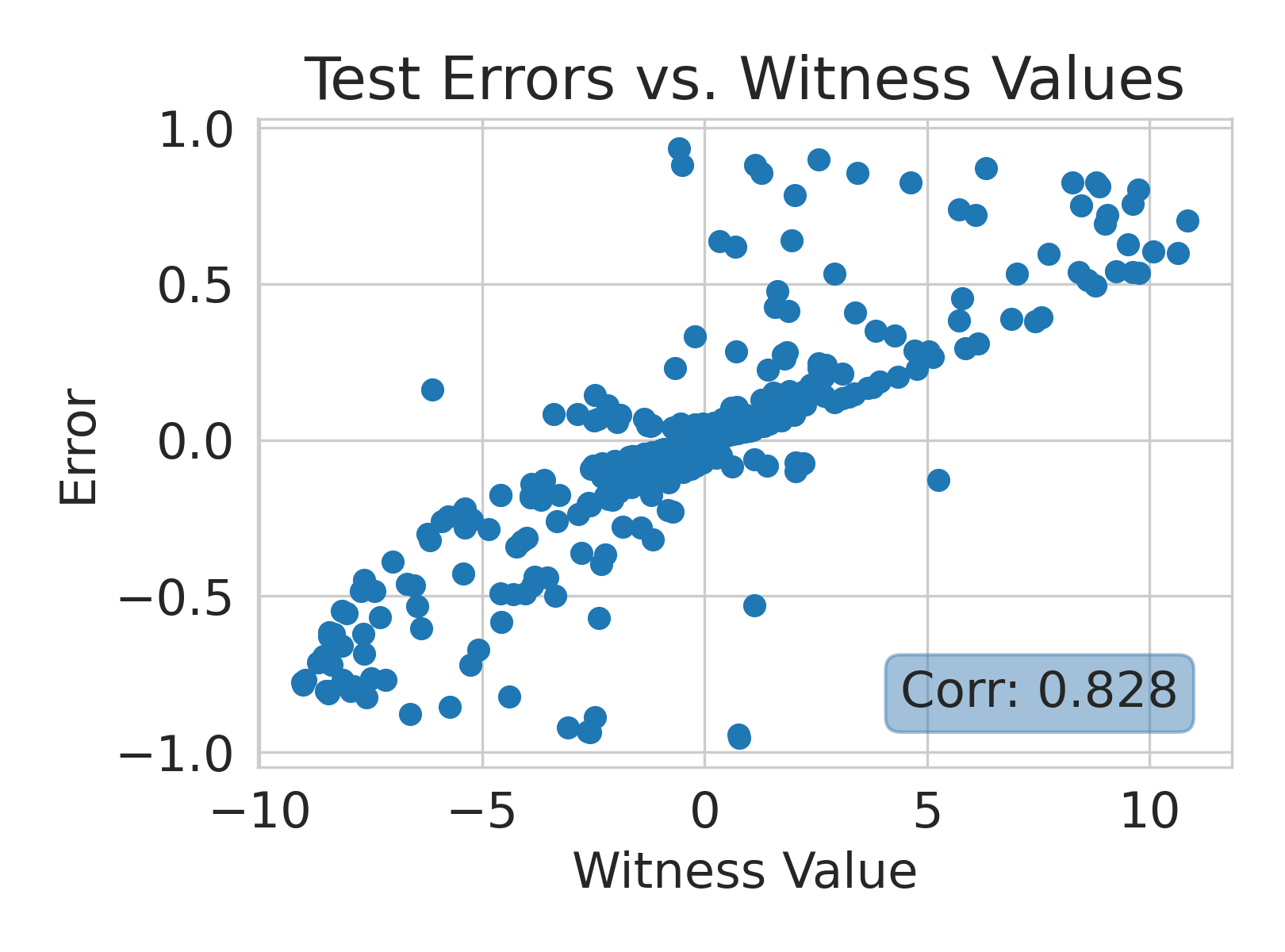}
     \end{tabular}}
  
  \caption{Witness function values are highly correlated with errors of the model. \textbf{Left:} Visualization of the moon dataset, with the logistic regression classifier decision boundaries displayed. \textbf{Middle:} Witness function values (Definition \ref{def:witness} with rbf kernel) $\empCstar$ is plotted as a contour under the error of the classifiers on test samples $y-f(\bx)$. The colored dots denote the error for each test sample $y-f(x)$. For samples where the model is most erroneous (dark green and dark purple dots), the predicted witness values are high (dark contour underneath). \textbf{Right:} The error $y-f(x)$ is plotted against the witness values $\empCstar$, with a Pearson correlation coefficient of 0.828.}
\label{fig::moon_circle_results}
\end{figure*}

The paradigm of fairness over categorical groups is an instance of embedded human bias in ML: tools are developed to fit a pre-defined metric on predefined groups, and once a contrived audit is passed, we call the algorithm ``fair.'' Defining groups by indicator functions over categorical groups is not expressive enough, and the most discriminated groups may not be known \emph{a priori}. This fact has fueled recent calls for new data-driven methods that uncover groups where a model errs the most. 
In particular, the burgeoning field of multi-group fairness, and definitions such as multicalibration and multiaccuracy \cite{hebert2018multicalibration, kim2019multiaccuracy,dwork2021outcome}, are important steps towards a more holistic view of fairness in ML, requiring a model to be calibrated on a large, potentially uncountable number of group-denoting functions instead of pre-defined categorical groups \cite{hebert2018multicalibration}. 

Multi-group fairness notions trade the choice of pre-determined categorical features for selecting a \emph{function class} over features. Here, the group most correlated with a classifier's errors (multiaccuracy) or against which a classifier is most miscalibrated (multicalibration) is indexed by a function in this class. \cite{hebert2018multicalibration} describes the class as being computable by a circuit of a fixed size. More concretely, \cite{kim2019multiaccuracy} and \cite{globus2023multicalibration} take this class to be linear regression, ridge regression, or shallow decision trees. 

We build on this line of work by considering a more general class of functions given by a Reproducing Kernel Hilbert Space (RKHS), defined on an infinite-dimensional feature space \cite{sriperumbudur2011universality}. In fact, an RKHS with a universal kernel is a dense subset of the space of continuous functions \cite{widmann2019calibration}. Surprisingly, by leveraging results from  information  and statistical learning theory \cite{muller1997integral, sriperumbudur2012empirical}, we show that the multi-group fairness problem in an RKHS is tractable: the most biased group has a closed form up to a proportionality constant. This leads to an exceedingly simple algorithm (\KMultiAcc, Algorithm \ref{alg:kernel_multiaccuracy}), which first identifies the function in the RKHS that correlates the most with error $y-f(\bx)$ (called the witness function), and then improves multiaccuracy by subtracting this function from the original predictions. 
As an example, Figure \ref{fig::moon_circle_results} illustrates that the error of a logistic regression model on the Two Moons synthetic dataset shows a strong correlation with the witness function values.

\subsection{Our Contributions}

The main contributions of this work include:
\begin{enumerate}
    \item We show that multiaccuracy, multicalibration, and outcome indistinguishability are integral probability metrics (IPMs), a well-studied family of statistical distance measures. When the groups or distinguishers lie in an RKHS, these IPMs have closed-form estimators, characterized by a witness function that achieves the supremum.  
    
    \item We introduce a consistent estimator for multiaccuracy, which flags the most discriminated group in terms of a function in Hilbert space, effectively revealing the previously unknown group that suffers the most from inaccurate predictions.
    
    \item We propose an algorithm, \KMultiAcc{} (Algorithm \ref{alg:kernel_multiaccuracy}), which provably corrects the given predictor's scores against its witness function. Empirically, our algorithm improves multiaccuracy and multicalibration after applying a standard score quantization technique, without the need for the iterative updates required by competing boosting-based models. 
    
    \item We conduct extensive experiments on both synthetic and real-world tabular datasets commonly used in fairness research. We show competitive or improved performance compared to competing models, both in terms of multi-group fairness metrics and AUC.
\end{enumerate}

\subsection{Related Literature}

\textbf{Multiaccuracy and Multicalibration.} 
Multiaccuracy and multicalibration, which emerged from theoretical computer science, ensure fairness over the set of computationally identifiable subgroups \cite{hebert2018multicalibration}. 
Multiaccuracy aims to make classification errors uncorrelated with subgroups, while multicalibration additionally requires predictions to be calibrated.
\cite{hebert2018multicalibration} and \cite{kim2019multiaccuracy} ensure multiaccuracy and multicalibration via a two-step process: identify subgroups with accuracy disparities, then apply a transformation to the classification function to boost accuracy over those groups---a method akin to weak agnostic learning \cite{feldman2009distributionspecificAB}. 
Subsequent works \cite{gopalan2021omnipredictors,gopalan2022loss,globus2023multicalibration} connect multicalibration to the general framework of loss minimization, introducing new techniques including reducing squared multicalibration error and projection-based error corrections \cite{globus2023multicalibration,deng2023happymap}. Recent developments include online multicalibration algorithms across Lipschitz convex loss functions \cite{garg2024oracle} and via a game-theoretic approach \cite{haghtalab2024unifying}. In addition, \cite{zhang2024fair} adopts multicalibration for multi-dimensional outputs for fair risk control. 

A common thread across work on multigroup fairness is to define subgroups in terms of function classes instead of pre-determined discrete combinations of group-denoting features \cite{kim2019multiaccuracy,dwork2021outcome, gopalan2021omnipredictors, kim2022ua}.
 Examples of such function classes include ``learnable" classes (in the usual statistical learning sense) \cite{kim2019multiaccuracy} and the set of indicator functions  \cite{dwork2023pseudorandomness}. Practical implementations of multigroup-fairness ensuring algorithms include \MCBoost{} \cite{kim2019multiaccuracy}, which uses ridge regression and decision tree regression, and \LSBoost{} \cite{globus2023multicalibration}, which uses linear regression and decision trees.
 Here,  we use both methods as benchmarks. 
 Unlike prior work, we consider the class of functions to be an RKHS and show that this class yields closed-form expressions for the function that correlates the most with error, allowing an efficient multiaccuracy intervention.

\textbf{Kernel-Based Calibration Metrics.}
Calibration ensures that probabilistic predictions are neither over- nor under-confident \cite{widmann2022calibration}.
Prior works have formulated calibration errors for tasks such as classification \cite{dawid1982well, niculescu2005predicting}, regression \cite{song2019distribution}, and beyond \cite{widmann2022calibration}. Calibration constraints may be directly incorporated into the training objective of a model \cite{kumar2018trainable}.
\cite{kumar2018trainable, widmann2019calibration,marx2023calibration,calibration_MMD} have adopted RKHS as the class of functions to ensure calibration.
We build on this prior work and develop kernel-based metrics and consistent estimators focused on multi-group fairness. 

\textbf{Integral Probability Measures (IPMs).}
\cite{dwork2021outcome} introduces outcome indistinguishability to unify multiaccuracy, multicalibration through a pseudo-randomness perspective---whether one can(not) tell apart ``Nature's'' and the predictor's predictions. We provide an alternative unifying perspective through distances between Nature's and the predictor's distributions. As discussed in \cite{dwork2021outcome}, outcome indistinguishability is closely connected to statistical distance (total variation distance), which, in turn, is one instantiation of an  IPM \cite{muller1997integral}, an extensively studied concept in statistical theory that measures the distance between two distributions with respect to a class of functions. \cite{sriperumbudur2012empirical} provides estimators for IPM defined on various classes of functions, which we apply to develop a consistent estimator for multiaccuracy.

\subsection{Notation}
We consider a pair of random variables $\bX$ and $Y$,  taking values in $\calX$ and $\calY$ respectively, where $\calX$ denotes the input features space to a prediction task and $\calY\subset \BR$ the output space. Often, we will take $\calY=\{0,1\}$, i.e., binary prediction.  
The pair $(\bX,Y)$ is distributed according to a fixed unknown joint distribution (\emph{Nature}'s distribution) $P_{\bX,Y}$ with marginals $P_{\bX}$ and $P_Y$.
In  binary prediction, we refer to a measurable function $f:\calX \to [0,1]$ as a \emph{predictor}. The predictor $f$ gives rise to a conditional distribution $Q_{Y\mid \bX=\bx}(1) \define f(\bx)$. We think of $Q_{Y\mid \bX}$ as an estimate of Nature's distribution, i.e., $Q_{Y\mid \bX=\bx}(1)\approx P_{Y\mid \bX=\bx}(1)$. 
The induced joint distribution %when $Y\mid \bX=\bx$ is sampled from 
for $Q_{Y\mid \bX=\bx}$ is denoted by $Q_{\bX,Y}\define P_{\bX} Q_{Y\mid \bX}$; this joint distribution $Q_{\bX,Y}$ will be referred to as the \emph{predictor}'s distribution.
%We denote the joint distribution induced by $f$ as $Q_{X,Y}=P_{X}Q_{Y|X}$. Note 
The marginal distribution $P_{\bX}$ is the same for both $Q_{\bX,Y}$ and  $P_{\bX,Y}$; only the conditional distribution $Q_{Y|\bX}$ changes due to using $f$. 

Given a measurable function $c$ and a random variable $Z\sim P$, we interchangeably denote expectation by $\BE_P[c] = \BE[c(Z)] = \BE_{Z \sim P}[c(Z)] \define \int_{\calZ} c(z) dP(z)$ depending on what is clearer from context. If $\calD$ is a finite set of i.i.d. samples, then we denote the empirical average by $\BE_{\calD}[c]=\BE_{Z\sim \calD}[c(Z)] \define |\calD|^{-1}\sum_{z\in \calD}c(z)$. %, which is itself a random variable. 

\section{Multi-Group Fairness as  Integral Probability Metrics}

We show the connection between IPMs~\cite{muller1997integral, sriperumbudur2012empirical}---a concept rooted in statistical learning theory---and multi-group fairness notions such as \textit{multiaccuracy}, \textit{multicalibration} \cite{hebert2018multicalibration}, and  \textit{outcome indistinguishability} \cite{dwork2023pseudorandomness}. The key property allowing for these connections is that the multi-group fairness notions and IPMs are both variational forms of measures of deviation between probability distributions. IPMs give perhaps the most general form of such variational representations, and we recall the definition next.

\begin{definition}[Integral Probability Metric \cite{muller1997integral, sriperumbudur2012empirical}] \label{defn::IPM}
Given two probability measures $P$ and $Q$ supported on $\calZ$ and a collection of functions $\mathfrak{C}\subset \{c : \calZ\to \BR\}$. We define the \emph{integral probability metric} (IPM) between $P$ and $Q$ with respect to $\frakC$ by
\begin{align}
    \gamma_{\mathfrak{C}}(P, Q) & \define %\sup_{c\in\mathfrak{C}}\left|\int_\calX c d\mathbb{P}-\int_\calX c d\mathbb{Q}\right|\\
    %&
    \sup_{c\in\mathfrak{C}} \ \left|\ExpVal{Z \sim P}{c(Z)}-\ExpVal{Z\sim Q}{c(Z)}\right|. \label{eq:def IPM}
\end{align}    
\end{definition}
\begin{example}
    IPMs recover other familiar metrics on probability measures, such as the total variation (statistical distance) metric. Indeed, when $\frakC$ is the unit $L^\infty$ ball of real-valued functions, i.e., $\frakC=\{c:\calZ \to \BR \Co \sup_{z\in \calZ} \ |c(z)|\le 1\}$, then $\gamma_{\frakC}(P,Q) = \TV(P,Q)$. 
\end{example}

As the example above shows, the complete freedom in choosing the set $\frakC$ allows IPMs the ability to subsume existing metrics on probability measures. We show that the expressiveness of IPMs carries through to multi-group fairness notions. Later, in Section~\ref{section:optimal}, we instantiate our IPM framework for multiaccuracy to the particular case when $\frakC$ is the unit ball in an infinite-dimensional Hilbert space, which then recovers another familiar metric on probability measures called the \emph{maximum mean discrepancy} (MMD) or \emph{kernel distance}.

\subsection{Multi-group Fairness Notions}

We recall the definitions of multiaccuracy and multicalibration from \cite{kim2019multiaccuracy, kim2022ua}, where the guarantees are parametrized by a class of real-valued functions $\mathfrak{C} \subset \{c:\calX \to \BR\}$. We call $\frakC$ herein the set of \emph{calibrating} functions.  Intuitively,
multi-group notions ensure that $c(\bX)$ for every group-denoting function $c\in \mathfrak{C}$ is uncorrelated with a model's errors $Y-f(\bX)$.

\begin{definition}[Multiaccuracy \cite{kim2019multiaccuracy,kim2022ua}]
     \label{Definition::multiaccuracy}
     Fix a collection of functions\footnote{The range is $[-1,1]$ in \cite{kim2019multiaccuracy} and $\mathbb{R}^+$ in \cite{kim2022ua}. We extend the range to $\mathbb{R}$.} $\mathfrak{C} \subset \{c:\calX \to \mathbb{R}\}$ and a distribution $P_{\bX,Y}$ supported on $\calX \times \calY$. A predictor $f:\calX \to [0,1]$ is ($\mathfrak{C},\alpha$)-\emph{multiaccurate} over $P_{\bX,Y}$ if for all $c\in\mathfrak{C}$ the following inequality holds:
     \begin{equation} \label{eq:multi-acc def}
         \mu(c,f,P_{\bX,Y}) \define \left| \ExpVal{}{c(\bX)(f(\bX)-Y)} \right| \leq \alpha
     \end{equation}
\end{definition}

Multicalibration proposed by~\cite{hebert2018multicalibration}  requires the predictor to be unbiased \emph{and} calibrated against groups denoted by functions in $\frakC$.

\begin{definition}[Multicalibration \cite{hebert2018multicalibration, kim2022ua, deng2023happymap}] 
\label{Definition::multicalibration}
    Fix a collection of functions $\mathfrak{C} \subset \{c: \calX \times [0,1] \to \mathbb{R}\}$ and a distribution $P_{\bX,Y}$ supported on $\calX \times \calY$. Fix a predictor $f:\calX \to [0,1]$ such that $f(\bX)$ is a discrete random variable.\footnote{Alternatively, one can consider a quantization of $f(\bX)$ such as done in \cite{globus2022multicalibrated}.} 
    We say that $f$ is ($\mathfrak{C},\alpha$)-\emph{multicalibrated} over $P_{\bX,Y}$ if for all $c\in\mathfrak{C}$ and $v\in \supp(f(\bX))$:
    \begin{equation}
        \left| \ExpVal{}{c(\bX, f(\bX))(f(\bX)-Y)\mid f(\bX)=v} \right| \leq \alpha
    \end{equation}
\end{definition}
As discussed in \cite{dwork2021outcome}, multi-group fairness constraints are equivalent to a broader framework of learning called outcome indistinguishability (OI). The object of interest is the distance between the two distributions---the ones induced by the predictor and by Nature.

\begin{definition}[Outcome Indistinguishability \cite{dwork2021outcome,dwork2023pseudorandomness}]
\label{definition:OI}
    Fix a collection of functions $\mathfrak{C} \subseteq \{c: \calX \times [0,1]\times \calY \to \BR\}$ and a distribution $P_{\bX,Y}$ supported on $\calX \times \calY$. We say that a predictor $f:\calX \to [0,1]$ is ($\mathfrak{C},\alpha)$-\emph{outcome-indistinguishable} if for all $c\in\mathfrak{C}$,
    \begin{align*}
    \left| \ExpVal{(\bX,Y)\sim P_{\bX,Y}}{c(\bX, f(\bX), Y)}-\ExpVal{(\bX,Y)\sim Q_{\bX,Y}} {c(\bX, f(\bX), Y)} \right|\leq \alpha,
    \end{align*}
    where we define the induced distribution by the predictor $Q_{\bX,Y}\define P_{\bX}Q_{Y\mid \bX}$ for $Q_{Y\mid \bX}(1) \define f(1)$. 
    % Recall that $P_{\bX,Y}$ and $Q_{\bX,Y}$ are Nature's and predictor's induced distributions, respectively.
\end{definition}

Total Variation (statistical) distance, one instantiation of an  IPM \cite{muller1997integral}, provides sufficient conditions for OI (\cite{dwork2021outcome}). We establish this broader connection next.

\subsection{Equivalence Between Multi-group Fairness Notions and IPMs} \label{section:IPM}

Since multiaccuracy, multicalibration, and outcome indistinguishability all pertain to finding the largest distance between distributions with respect to a collection of functions, we can unify them in terms of IPMs. First, we show that ensuring a predictor's multiaccuracy with respect to a set of calibrating functions $\frakC$ is equivalent to ensuring an upper bound on the IPM between Nature's and the predictor's distribution with respect to a modified set of function $\tilde{\frakC}$, given explicitly in the following result. 

\begin{proposition}[Multiaccuracy as an IPM] \label{proposition: MA as IPM}
\label{prop::Multiaccuracy-IPM}
    Fix a collection of functions $\mathfrak{C} \subset L^1(\calX)$, and let $\calY=\{0,1\}$. Fix a predictor $f:\calX \to [0,1]$ inducing the distribution $Q_{\bX,Y}$. Denote the modified set of functions
    \begin{equation}
        \widetilde \frakC = \left\{\tilde c: \calX\times \calY \to \mathbb{R}~\middle|~\tilde{c}(\bx,y) = (-1)^{1-y} \cdot c(\bx)/2 ~\mbox{for } c\in \frakC \right\}.
    \end{equation}  
    Then, for any $\alpha \ge 0$, the predictor $f$ is ($\mathfrak{C},\alpha$)-multiaccurate if and only if the IPM between Nature's distribution and the predictor's distribution is upper bounded by $\alpha$:
    \begin{equation}
        \gamma_{\widetilde{\mathfrak{C}}}(P_{\bX,Y}, Q_{\bX,Y}) \le \alpha.
    \end{equation}
\end{proposition}
% \begin{proof}
%     See Appendix~\ref{appendix: MA as IPM}.
% \end{proof}
\begin{proof}
  Let $(\bxi,Y)$ be an identical copy of $(\bX,Y)$. Using the notation in~\eqref{eq:multi-acc def} in the definition of multiaccuracy (Definition~\ref{Definition::multiaccuracy}), we have that for every $c\in \frakC$
\begin{align}
    \mu(c,f,P_{\bX,Y}) &= \left|\BE\left[ c(\bxi)(f(\bxi)-Y) \right] \right| \\
    &= \left| \BE\left[ \BE\left[ c(\bxi)(Y-f(\bxi) ) \mid \bxi \right] \right] \right| \\
    &= \left| \BE\left[ c(\bxi) \left( P_{Y\mid \bX=\bxi}(1) - Q_{Y\mid \bX=\bxi}(1) \right) \right] \right| \\
    % &= \left| \BE\left[ c(\bxi) \left( \frac12 \left( P_{Y\mid \bX=\bxi}(1) - Q_{Y\mid \bX=\bxi}(1) \right) \right.\right.\right.\\
    %    &\quad\quad \left.\left.\left.- \frac12 \left( P_{Y\mid \bX=\bxi}(0) - Q_{Y\mid \bX=\bxi}(0) \right) \right) \right] \right| \\
    &= \left| \BE\left[ \frac{c(\bxi)}{2}P_{Y\mid \bX=\bxi}(1) -  \frac{c(\bxi)}{2}P_{Y\mid \bX=\bxi}(0) \right] \right.\\
        &\quad\quad \left. -\BE\left[ \frac{c(\bxi)}{2}Q_{Y\mid \bX=\bxi}(1) -  \frac{c(\bxi)}{2}Q_{Y\mid \bX=\bxi}(0) \right] \right| \\
    &= \left| \BE_{P_{\bX,Y}}\left[ \tilde{c} \right] - \BE_{Q_{\bX,Y}}\left[ \tilde{c} \right] \right|,
\end{align}
where $\tilde{c}(\bx,y) \define (-1)^{1-y} \cdot c(\bx)/2$. 
By definition of multiaccuracy, we have that $f$ is $(\frakC,\alpha)$-multiaccurate if and only if $\sup_{c\in \frakC} \ \mu(c,f,P_{\bX,Y}) \le \alpha$.
This is equivalent, by the above, to having the IPM bound $\gamma_{\tilde{\frakC}}(P_{\bX,Y},Q_{\bX,Y})\le \alpha$, where $\tilde{\frakC}$ is as defined in the proposition statement, i.e., it is the collection of modified functions $\tilde{c}$ as $c$ ranges over $\frakC$. 
\end{proof}

Expressing multiaccuracy as an IPM bound will allow us to rigorously accomplish two goals: 1) quantifying multiaccuracy from finitely many samples of $P_{\bX,Y}$, and 2) correcting a given predictor $f$ to be multiaccurate. These two goals are the subject of Section~\ref{section:optimal}. Similarly, multicalibration and OI can be expressed as IPMs.

\begin{proposition}[Multicalibration as an IPM] 
\label{prop::Multicalibration-IPM}
Fix a collection of functions 
$\mathfrak{C} \subseteq \{c: \calX \to \mathbb{R}\}$, and let $\calY=\{0,1\}$. Fix a %\footnote{W.l.o.g. we choose $\{-1,1\}$ for ease of presentation, since mapping back to $\{0,1\}$ is injective.} 
predictor $f:\calX \to [0,1]$ inducing the distribution $Q_{\bX,Y}$. Moreover, let $\eta_y \define (-1)^{1-y}$. 
Let $d:[0,1]\to \mathcal{V}\subset[0,1]$, $\left|\mathcal{V}\right|< \infty$ be a discrete, finite quantization of $[0,1]$, where   $P_{\bX}(d(f(\bX))= v)>0$ for all $v\in \mathcal{V}$. 
Define the set of functions 
$$\widetilde \frakC_v \define \left\{\tilde{c}: \calX\times \calY \times \calV \to \mathbb{R} ~\middle|~ \tilde{c}(\bx,y,v) =  \frac{c(\bx)\indicator_{f(\bX)=v}\eta_y}{2P_{\bX}(f(\bX)=v)} ~\mbox{for some } c\in \frakC \right\}.$$  

Then  $f$ is ($\mathfrak{C},\alpha$)-multicalibrated if and only if $\gamma_{\widetilde{\mathfrak{C}}_v}(P_{\bX,Y}, Q_{\bX,Y})\leq \alpha$ for every $v\in \mathcal{V}$.   
\end{proposition}
% \begin{proof}
%     See Appendix \ref{appendix: MC as IPM}.
% \end{proof}
\begin{proof}
    Let $(\bxi,Y)$ be an identical copy of $(\bX,Y)$. Using the notation in the definition of multicalibration (Definition~\ref{Definition::multicalibration}), we have that for every $c\in \frakC$, $v\in\calV$
\begin{align}
        &\ExpVal{}{c(\bxi)(Y-f(\bxi))|f(\bxi)=v}\\
        = &\ExpVal{}{\frac{c(\bxi)\indicator_{f(\bxi)=v}}{P_{\bX}(f(\bxi)=v)}(P_{Y|\bX =\bxi}(1)-Q_{Y|\bX = \bxi}(1))}\\
        % = &\ExpVal{}{\frac{c(\bxi)\indicator_{f(\bxi)=v}}{2P_{\bX}(f(\bxi)=v)}(P_{Y|\bX = \bxi}(1)-Q_{Y|\bX=\bxi}(1))  \\
        % &\quad -\frac{c(\bxi)\indicator_{f(\bxi)=v}}{2P_{\bX}(f(\bxi)=v)}(P_{Y|\bX=\bxi}(0)-Q_{Y|\bX=\bxi}(0))}\\
        = &\ExpVal{}{\frac{c(\bxi)\indicator_{f(\bxi)=v}}{2P_{\bX}(f(\bxi)=v)}(P_{Y|\bX = \bxi}(1) -P_{Y|\bX=\bxi}(0))} \\
        &\quad - \ExpVal{}{\frac{c(\bxi)\indicator_{f(\bxi)=v}}{2P_{\bX}(f(\bxi)=v)}(Q_{Y|\bX=\bxi}(1)-Q_{Y|\bX=\bxi}(0))}\\
        = &\ExpVal{P_{\bX,Y}}{\tilde{c}}- \ExpVal{Q_{\bX,Y}}{\tilde{c}} 
\end{align}
where $\displaystyle\tilde{c}(\bx,y,v) \define  \frac{c(\bx)\indicator_{f(\bX)=v}\eta_y}{2P_{\bX}(f(\bX)=v)}$.
% \begin{equation*}
%     \tilde{c}(\bx,y,v) \triangleq \begin{cases}
%     \displaystyle\frac{c(\bx)\indicator_{f(\bx)=v}}{2P_{\bX}(f(\bx)=v)} &\mbox{if } y = 1,\\
%     -\displaystyle\frac{c(\bx)\indicator_{f(\bx)=v}}{2P_{\bX}(f(\bx)=v)} &\mbox{if } y = 0.
%     \end{cases}
% \end{equation*}
\end{proof}

\begin{proposition}[OI as an IPM] \label{prop::OI-IPM}
Let $\frakC\subset \{c:\calX\times [0,1]\times \{0,1\} \to \BR \}$ be a collection of functions, and 
fix a predictor $f:\calX \to [0,1]$ inducing the distribution $Q_{\bX,Y}$ on $\calX\times \{0,1\}$ via composing with $P_{\bX}$. Define the set of function
\begin{equation}
    \tilde{\frakC} = \left\{ \tilde{c} : \calX\times \{0,1\} \to \BR \mid \tilde{c}(\bx,y) = c(\bx,f(\bx),y) \text{ for some } c\in \frakC \right\}.
\end{equation}
Then, for any $\alpha\ge 0$, $f$ is ($\mathfrak{C},\alpha$)-OI if and only if $\gamma_{\tilde{\mathfrak{C}}}(P_{\bX,Y}, Q_{\bX,Y})\leq \alpha.$
\end{proposition}

% \begin{proof} Follows directly from definition of OI \ref{definition:OI}.
%  % See Appendix \ref{appendix: OI as IPM}.
% \end{proof}

\begin{proof}
From Definition \ref{defn::IPM}, if $\gamma_{\tilde{\mathfrak{C}}}(P_{\bX,Y}, Q_{\bX,Y})\leq \alpha$,

\begin{align}
    \gamma_{\tilde{\mathfrak{C}}}(P_{\bX,Y}, Q_{\bX,Y}) \define &%\sup_{c\in\mathfrak{C}}\left|\int_\calX c d\mathbb{P}-\int_\calX c d\mathbb{Q}\right|\\
    %&
    \sup_{\tilde{c}\in \tilde{\mathfrak{C}}} \ \left|\ExpVal{(\bX,Y) \sim P_{\bX,Y}}{\tilde{c}(\bX,Y)}-\ExpVal{(\bX,Y) \sim Q_{\bX,Y}}{\tilde{c}(\bX,Y)}\right|\\
    & =\sup_{c\in \mathfrak{C}} \ \left|\ExpVal{(\bX,Y) \sim P_{\bX,Y}}{c(\bX,f(\bX),Y)}\right.\\
    & \quad \left. -\ExpVal{(\bX,Y) \sim Q_{\bX,Y}}{c(\bX,f(\bX),Y)}\right|\\
    & \leq \alpha
\end{align} 
By Definition \ref{definition:OI}, $f$ is
($\mathfrak{C},\alpha$)-OI. The other direction is analogous.
\end{proof}

\section{ Multiaccuracy in Hilbert Space} \label{section:optimal}

% \fc{[The next paragraph needs rephrasing. Be simple and direct. The below should be a start:

We develop a theoretical framework and an algorithm for quantifying and ensuring $(\frakC,\alpha)$-multiaccuracy. We consider the group-denoting functions $\frakC_k$ to be the unit ball in an infinite-dimensional Hilbert space, namely, an RKHS $\calH_k$ defined by a given kernel $k$ (Definition~\ref{definition:RKHS}). %In addition to the expressivity of $\frakC_k$ of calibration functions, 
The proposed set of calibration functions $\frakC_k$ can easily exhibit and exceed the expressivity of group-denoting indicator functions. 
Surprisingly, despite the expressiveness of $\frakC_k$, we show that the calibration function that maximizes multiaccuracy error, i.e. the witness function $c_k^\star$ (Definition~\ref{def:witness}), has a closed form---in contrast to when $\frakC$ is, for example, a set of decision trees \cite{globus2023multicalibration, kim2019multiaccuracy}. This enables us to derive a procedure for ensuring multiaccuracy (\KMultiAcc{}, Algorithm \ref{alg:kernel_multiaccuracy}).

%We derive an explicit formula for the witness function $c_k^\star$  (Definition~\ref{def:witness}) and propose an optimization problem to find the closest multiaccurate predictor $g^\star$ to a given predictor $f$ \eqref{eq:multiaccuracy optimization simplified}. Our algorithm, \KMultiAcc{}, shifts $f$ by adding a multiple of $c^\star$ and clipping the result to $[0,1]$. We prove that \KMultiAcc is optimal for solving~\eqref{eq:multiaccuracy optimization simplified}.

\subsection{Calibration Functions in RKHS and its Witness Function for Multiaccuracy}

Our choice of calibrating functions $\frakC$ is the set of functions with bounded norm in an RKHS. First, recall that an RKHS can be defined via \emph{kernal functions}, as follows\footnote{The characterizing property of a real RKHS is that it is a Hilbert space $\calH$ of functions $c:\calX\to \BR$ for which every evaluation map $c\mapsto c(\bx)$ is a continuous function from $\calH$ to $\BR$ for each fixed $\bx\in \calX$.}.

\begin{definition}[Reproducing kernel Hilbert space (RKHS)] \label{definition:RKHS}
    Let $\calH\subset \{c:\calX \to \BR \}$ be a real Hilbert space with inner product $\langle \wc, \wc \rangle_{\calH}$, and fix a function $k:\calX^2 \to \BR$. We say that $\calH$ is a \emph{reproducing kernel Hilbert space} with kernel $k$ if it holds that $k(\wc,\bx)\in \calH$ for all $\bx\in \calX$ and $\langle c,k(\wc,\bx) \rangle_{\calH} = c(\bx)$ for all $c\in \calH$ and $\bx\in \calX$. 
    We denote $\calH$ by $\calH_k$ if $k$ is given.
\end{definition}

We use the structure of the RKHS as our group-denoting functions. Thus, for a prescribed multiaccuracy level $\alpha$, we will need to restrict attention to elements of $\calH_k$ whose norm satisfies a given bound. To normalize, we choose the unit ball in $\calH_k$ as our set of calibration functions, i.e. 
    \begin{equation} \label{eq:C RKHS def}
        \frakC_k \define \left\{c\in \calH_k \Co \|c\|_{\calH_k} \ \le 1 \right\}.
    \end{equation}
    
% \begin{definition}[Calibration functions in an RKHS] 
%     Given a kernel function $k:\calX^2 \to \BR$ and the induced RKHS $\calH_k$, we define the set of calibration functions $\frakC_k$ to be the unit ball in $\calH_k$, i.e., 
%     \begin{equation} \label{eq:C RKHS def}
%         \frakC_k \define \left\{c\in \calH_k \Co \|c\|_{\calH_k} \ \le 1 \right\}.
%     \end{equation}
% \end{definition}

We note that when the class of functions $\frakC$ is the unit ball in an RKHS, the induced IPM $\gamma_{\frakC}(P,Q)$ is called the \emph{maximum mean discrepancy} (MMD)~\cite{sriperumbudur2012empirical}. 

Of particular importance are calibration functions $c\in \frakC$ that attain the maximal multiaccuracy error (the LHS of~\eqref{eq:multi-acc def}). Such functions, called \emph{witness functions} \cite{kim2016examples}, encode the multiaccuracy definition without the need to consider the full set $\frakC$.

\begin{definition}[Witness function for multiaccuracy] \label{def:witness}
    For a fixed set of calibration functions $\frakC \subset \{c:\calX \to \BR\}$, predictor $f:\calX \to [0,1]$, and distribution $P_{\bX,Y}$, we say that $c^\star\in \frakC$ is a \emph{witness function} for multiaccuracy of $f$ with respect to $\frakC$ if it attains the maximum on the LHS in~\eqref{eq:multi-acc def}:
    \begin{equation}
       \mu(c^\star,f,P_{\bX,Y}) = \max_{c\in \frakC} \ \mu(c,f,P_{\bX,Y}).
    \end{equation}
\end{definition}

While an RKHS can encompass a broader class of functions than shallow decision trees or linear models, finding the function in the RKHS that errs the most (i.e., the witness function as per Definition~\ref{def:witness}) is surprisingly simple. 
Firstly, it can be shown that for the IPM $\gamma_{\frakC_k}(P,Q)$ (where $\frakC_k$ is the unit ball in $\calH_k\subset \{c:\calZ \to \BR\}$), the function $c\in \frakC_k$ that maximizes the RHS of~\eqref{eq:def IPM} is in closed form, up to a multiplicative constant~\cite{gretton2006kernel, kim2016examples}
\begin{equation} \label{eq::witness function}
    c^\star(z) \propto \ExpVal{\zeta\sim P}{k(z,\zeta)}-\ExpVal{\zeta\sim Q}{k(z,\zeta)}.
\end{equation}

By the connection between IPM and multiaccuracy, we can similarly find the closed form of the witness function for multiaccuracy(Definition~\ref{def:witness}). 
\begin{proposition}[Witness function for multiaccuracy] \label{prop::witness_multiaccuracy}
    Given a the kernel function $k:\calX^2\to \BR$ and distribution $P_{\bX,Y}$ over $\calX\times \{0,1\}$. We assume that $\calH_k\subset L^1(\calX)$\footnote{$L^1(\calX)$ denotes the space of real-valued functions that are integrable against $P_{\bX}$, i.e. $L^1(\calX) \define \left\{ c:\calX \to \BR \Co \BE\left[\left|c(\bX)\right|\right] < \infty \right\}$.}. Fix a predictor $f:\calX\to [0,1]$ satisfying $\BE[k(\wc,\bX)f(\bX)]\in \calH_k$. Then, there exists a unique (up to sign) witness function for multiaccuracy of $f$ with respect to $\frakC_k$ (as per Definition~\ref{def:witness}), and it is given by
    \begin{equation} \label{eq:c star def}
        c_{k,f}^\star(\bx) \define \ExpVal{}{\theta \cdot (Y-f(\bX))k(\bx,\bX)},
    \end{equation}
    where $\theta\in \BR$ is a normalizing constant so that $\|c_{k,f}^\star\|_{\calH_k}=1$.
\end{proposition}
% \begin{proof}
%     See Appendix~\ref{appendix:witness}.
% \end{proof}
\begin{proof}
% First, we show that multiaccuracy can be equivalently written in terms of an IPM using Proposition~\ref{prop::Multiaccuracy-IPM}. For this, we only need to verify that $\frakC_k \subset L^1(\calX)$. Fix an arbitrary $c(\wc)=\sum_{i\in \BN} c_ik(\bx_i,\wc)\in \frakC_k$ for some $c_i\in \BR$ and $\bx_i\in \calX$. By continuity of the evaluation functionals on $\calH_k$, we obtain that $h_n(\bx) \define \sum_{i=1}^n c_ik(\bx_i,\bx) \to c(\bx)$ pointwise for each $\bx\in \calX$ as $n\to \infty$~\cite[Chapter~1, Corollary~1]{RKHS_book}. Let $h(\bx)\define \sum_{i=1}^\infty |c_ik(\bx_i,\bx)|$. Thus, by the first condition in Assumption~\ref{assumption:witness}, and using Tonelli's theorem, we have that $\BE[|c(\bx)|]\le \sum_{i\in \BN} |c_i| \BE[|k(\bx_i,\bX)|] < \infty$, as desired. Thus, we may apply Proposition~\ref{prop::Multiaccuracy-IPM} to obtain that $(\frakC_k,\alpha)$-multiaccuracy of $f$ is equivalent to the IPM bound $\gamma_{\tilde{\frakC}}(P_{\bX,Y},Q_{\bX,Y})\le \alpha$, where $\tilde{\frakC}$ and $Q_{\bX,Y}$ are as constructed in Proposition~\ref{prop::Multiaccuracy-IPM}. Next, we use the definition of IPMs to deduce the fomula for the witness function. 

First, by continuity of the evaluation functionals on $\calH_k$, we obtain that $h_n(\bx) \define \sum_{i=1}^n c_ik(\bx_i,\bx) \to c(\bx)$ pointwise for each $\bx\in \calX$ as $n\to \infty$~\cite[Chapter~1, Corollary~1]{RKHS_book}. Let $h(\bx)\define \sum_{i=1}^\infty |c_ik(\bx_i,\bx)|$. 
Next, applying Proposition~\ref{prop::Multiaccuracy-IPM}, $(\frakC_k,\alpha)$-multiaccuracy of $f$ is equivalent to the IPM bound $\gamma_{\tilde{\frakC}}(P_{\bX,Y},Q_{\bX,Y})\le \alpha$, where $\tilde{\frakC}$ and $Q_{\bX,Y}$ are as constructed in Proposition~\ref{prop::Multiaccuracy-IPM}. Next, we use the definition of IPMs to deduce the formula for the witness function. 
%and using Tonelli's theorem, we have that $h\in L^1(\calX)$. Furthermore, $|h_n(\bx)|\le h(\bx)$ for all $n\in \BN$ and $\bx\in \calX$. 

We rewrite the function inside the maximization definition of $\gamma_{\tilde{\frakC}}(P_{\bX,Y},Q_{\bX,Y})$ as an inner product in $\calH$. Fix $c$ as above. Then, with $\tilde{c}(\bx,y)\define (-1)^{1-y}c(\bx)/2$, we have that
\begin{align}
    2\BE_{P_{\bX,Y}}[\tilde{c}] &= \BE_{P_{\bX,Y}}\left[ (-1)^{1-Y}c(\bX) \right] \\
    &= \BE_{P_{\bX,Y}}\left[ (-1)^{1-Y} \left\langle c, k(\wc,\bX) \right\rangle_{\calH} \right] \\
    &= \BE_{P_{\bX,Y}}\left[ (-1)^{1-Y} \langle \sum_{i\in \BN} c_i k(\bx_i,\wc), k(\wc,\bX) \rangle_{\calH} \right]  \\
    &= \BE_{P_{\bX,Y}}\left[ (-1)^{1-Y} \sum_{i\in \BN} c_i \langle  k(\bx_i,\wc), k(\wc,\bX) \rangle_{\calH} \right] \label{eq:witness continuity} \\
    % &= \BE_{P_{\bX,Y}}\left[ (-1)^{1-Y} \sum_{i\in \BN} c_i   k(\bx_i, \bX) \right] \\
    &= \sum_{i\in \BN} c_i \ \BE_{P_{\bX,Y}}\left[ (-1)^{1-Y}   k(\bx_i, \bX) \right], \label{eq:witness assumption}
\end{align}
where~\eqref{eq:witness continuity} follows by continuity of the inner product and~\eqref{eq:witness assumption} by Fubini's theorem since $\calH_k\subset L^1(\calX)$.
%in view of Assumption~\ref{assumption:witness}.
The same steps follow for $Q_{\bX,Y}$ in place of $P_{\bX,Y}$, and subtracting the ensuing two equations we obtain
\begin{align}
    \BE_{P_{\bX,Y}}[\tilde{c}] - \BE_{Q_{\bX,Y}}[\tilde{c}] &= \sum_{i\in \BN} c_i \BE_{P_{\bX,Y}}\left[ (Y-f(\bX))k(\bx_i,\bX) \right]\\
    &= \langle c, \BE_{P_{\bX,Y}}\left[ (Y-f(\bX))k(\wc,\bX) \right] \rangle_{\calH}.
\end{align}
Therefore, the maximizing function is given up to a normalizing constant by
\begin{equation*}
    c_{k,f}^\star(\bx) \propto \BE_{P_{\bX,Y}}\left[ (Y-f(\bX))k(\bx,\bX) \right].
\end{equation*}
\end{proof}

In the presence of finitely many samples, one must resort to numerical approximations of the witness function.

\begin{definition}[Empirical Witness Function] 
\label{definition:emp witness}
    Let $\calD_0$ be a finite set of i.i.d. samples from $P_{\bX,Y}$. We define the \emph{empirical witness function} as the plug-in estimator of~\eqref{eq:c star def}:
    \begin{equation} \label{eq:emp c star def}
        c_{k,\calD_0,f}^\star(\bx) = \ExpVal{(\bX,Y)\sim \calD_0}{\hat{\theta} \cdot (Y-f(\bX))k(\bx,\bX)},
    \end{equation}
    where $\hat{\theta}\in \BR$ is a normalizing constant so that $\|c_{k,\calD_0,f}^\star\|_{\calH}=1$.
\end{definition}

Observe that given a training dataset $\{\bx_i\}_{i=1}^n$, the witness function for a new sample $\bx$ is proportional to the sum of the error of $\bx_i$ weighted by $k(\bx, \bx_i)$---the distance between $\bx_i$ and the new sample $\bx$ in the kernel space. The witness function is performing a kernel regression of a model's errors. From the definition of the witness function, it attains the supremum in the IPM, which measures the distance between Nature and the Predictor's distribution. Hence, if a new sample $\bx$ attains a high witness function value, $f(\bx)$ is likely erroneous.

We call the multiaccuracy error when $\frakC$ comes from an RKHS the \emph{kernel multiaccuracy error}, defined with the witness function $c_{k,f}^\star(\bX)$ which attains the maximum error.

\begin{definition}[Kernel Multiaccuracy Error (KME)] \label{defn::multiaccuracy_error}
Let $\frakC_k$ be the set of calibration functions in the RKHS $\calH_k$ as defined in~\eqref{eq:C RKHS def}. Given a predictor $f$, the \emph{kernel multiaccuracy error} (KME) is defined as 
  \begin{align}
  \gamma_k(f,P_{\bX,Y}) \define 
\left|\ExpVal{}{c_{k,f}^\star(\bX)(Y-f(\bX))}\right|.
\end{align}  
\end{definition}

\noindent The empirical version has the plug-in estimator of the witness function $c_{k,f}^\star$.

\begin{definition}[Empirical KME] \label{definition:emp KME}
    Given a test set of freshly sampled i.i.d. datapoints $\calD$, we define the \emph{empirical} KME by
    \begin{align}
      \gamma_k(f,\calD) \define 
    \left|\ExpVal{(\bX,Y)\sim \calD}{c_{k,\calD_0,f}^\star(\bX)(Y-f(\bX))}\right|.
    \end{align}  
    %where the witness function $\cstar(x)$ and its empirical version $\empCstar(\bx)$ are defined in Proposition \ref{prop::witness_multiaccuracy}.
\end{definition}

\begin{remark}[Overcoming the Curse of Dimensionality] \label{remark::overcome_curse_dimentionality}
     One important observation is that the MMD estimator depends on the dataset $\calD$ only through the kernel $k$. Hence, once $k(\bx_i,\bx_j)$ is known, the complexity of the estimator is independent of the dimensionality of $\bX$---e.g., for $\bX\in \mathbb{R}^d$, sample complexity does not scale exponentially with $d$ (see the end of Section~2.1 in~\cite{sriperumbudur2012empirical}. \remarkend
\end{remark}

We give the consistency and rate of convergence of KME---the finite-sample estimation of KME converges to the true expectation, following a direct application of~\cite[Corollary~3.5]{sriperumbudur2012empirical}. 

\begin{theorem}[Consistency of the KME Estimator,~{\cite[Corollary~3.5]{sriperumbudur2012empirical}}]
    Suppose the kernel $k$ is measurable and satisfies $\sup_{\bx\in\calX}k(\bx,\bx)\leq C < \infty$. Then, with probability at least $1-2e^{-\tau}$ over the choice of i.i.d. samples $\calD$ from $P_{\bX,Y}$ and for every predictor $f:\calX\to [0,1]$, there is a constant $A=A(f,P_{\bX,Y})$ such that the inequality %there is a constant $A=A(P_{\bX,Y},C,f)$ such that for all $n$ and $m$ and every set $\calD$ 
\begin{align}
    |\gamma_k(f,\calD)-\gamma_k(f,P_{\bX,Y})| \le A (\frac{1+\sqrt{\tau}}{\sqrt{|\calD|}}+\frac{\tau}{|\calD|}),
\end{align}
In addition, we have the almost-sure convergence $\gamma_k(f,\calD) \rightarrow \gamma_k(f,P_{\bX,Y})$ as $|\calD| \rightarrow \infty$. 
\end{theorem}

%We provide further discussions on the curse of dimensionality, and the consistency and rate of convergence of KME in Appendix \ref{appndix:remarks}.

Next, we proceed to show an algorithm, \KMultiAcc{}, that corrects a given predictor $f$ of its multiaccuracy error using the empirical witness function. 

\subsection{\KMultiAcc{}: Proposed Algorithm for Multiaccuracy}
%Existing methods adopt the strategy of boosting, where multiple classifiers are trained to achieve low calibration error for different level sets \cite{kim2019multiaccuracy,globus2023multicalibration}. 
We propose a simple algorithm \KMultiAcc{} (Algorithm \ref{alg:kernel_multiaccuracy}) that corrects the original predictor from multiaccuracy error. Notably, \KMultiAcc{} does not require iterative updates, unlike all competing boosting or projection-based models \cite{kim2019multiaccuracy,globus2023multicalibration,deng2023happymap}. 
In a nutshell, \KMultiAcc{} first identifies the function in the RKHS that correlates the most with the predictor's error $y-f(\bx)$ (called the witness function) and subtracts this function from the original prediction to get a multi-group fair model. The first step is surprisingly simple---as we have shown above, the witness function of an RKHS has a closed form up to a proportionality constant. The second step is an additive update followed by clipping.

As outlined in Algorithm \ref{alg:kernel_multiaccuracy}, the algorithm takes in a pre-trained base predictor $f$, a proportionality constant $\lambda$, and a (testing) dataset $\calD$ on which the model is evaluated. Additionally, to define the witness function and the RKHS, the algorithm is given a dataset reserved for learning the witness function $\calD_0$ and a kernel function $k$. With these, for each sample, the algorithm first computes the witness function value, and subtracts away the witness function value multiplied by $\lambda$, the proportionality constant which we learn from data (described in the next paragraph). Finally, we clip the output to fall within $[0,1]$. 

\textbf{Learning the Proportionality Constant.} There are multiple approaches to obtaining the proportionality constant that scales the witness function appropriately. As an example, we adopt a data-driven approach to find $\lambda$. We use a validation set to perform a grid search on $[0,1]$ to get the $\lambda$ that produces a predictor $g'(\bx) = f(\bx)+ \lambda \empCstar(\bx)$ that is closest to $f$ in terms of $L-2$ distance, while also satisfying a $\alpha-$multiaccuracy with a specified $\alpha$\footnote{When the multiaccuracy constraint cannot be met, we output the $\lambda$ that achieves the lowest multiaccuracy error using the witness values of $f$.}.

\begin{algorithm}
\caption{\KMultiAcc}
\label{alg:kernel_multiaccuracy}
\begin{algorithmic}
\Require base predictor $f:\calX\to[0,1]$, constant $\lambda\ge 0$, finite datasets $\calD_0, \calD \subset \calX\times\calY$, kernel function $k:\calX^2\to\BR$, and empirical witness function (Definition~\ref{definition:emp witness}) $\empC(x) = \BE_{(\bX,Y)\sim\calD_0}[\theta\cdot(Y-f(\bX))k(x,\bX)]$.
\For{$(x,y) \in \calD$}
    \State $g'(x) = f(x) + \lambda \empC(x)$
    \State $g(x) = \max(0, \min(g'(x), 1))$
\EndFor
\State \textbf{Output} $g$
\end{algorithmic}
\end{algorithm}

% \begin{algorithm}
%    \caption{\KMultiAcc{}}
%    \label{alg:kernel_multiaccuracy}
% \begin{algorithmic}
%    \STATE {\bfseries Input:} base predictor $f:\calX\to [0,1]$, constant $\lambda \ge 0$, finite datasets $\calD_0,\calD\subset \calX\times \calY$, kernel function $k:\calX^2 \to \BR$, and empirical witness function (Definition~\ref{definition:emp witness}) $c_{k,\calD_0,f}^\star(\bx) = \sum_{(\bx',y) \in \calD_{0}} \theta \cdot (y-f(\bx'))k(\bx',\bx)$. 
%    \vspace{1mm}
%    % $\empCstar$
%    % \STATE {\bfseries where} $\empCstar(\bx') = \sum_{(\bx,y) \in \calD_{0}} (y-f(\bx))k(\bx,\bx').$ 
%     \FOR{$(\bx,y)\in \calD$}  
%     \vspace{1mm}
%        \STATE $g'(\bx) = f(\bx)+ \lambda \empCstar(\bx)$
%     \vspace{1mm}
%        \STATE $g(\bx)= \max(0, \min(g'(\bx),1))$
%        % \begin{cases}
%        %  0, & \text{if } g'(\bx)<0, \\
%        %  g'(\bx), & \text{if }  g'(\bx)  \in [0,1], \\
%        %  1, & \text{if } g'(\bx) >1.
%        % \end{cases}
%     \vspace{1mm}
%     \ENDFOR
%     \vspace{1mm}
%    \STATE {\bfseries Output} $g$ 
% \end{algorithmic}
% \end{algorithm}

\begin{remark}[One-Step Update]
    For the linear kernel $k(x,x')=x^Tx'$, we show that \KMultiAcc{} yields a $0$-multiaccurate predictor in a single step. While this property does not extend to nonlinear kernels, we observe empirically that the one-step update in \KMultiAcc{} significantly reduces the empirical KME for RBF kernels. See Appendix \ref{apdx:proofs_ch3}.\ for a detailed discussion.
\end{remark}

We discuss in the following section a theoretical framework that gives rise to \KMultiAcc{} and the grid-search approach.

\subsection{Theoretical Framework for \KMultiAcc{}}

We formulate an optimization that, given a prediction $f:\calX \to [0,1]$ that is not necessarily multiaccurate, finds the ``closest'' predictor $g:\calX\to [0,1]$ that is corrected for multiaccuracy with respect to the empirical witness function $c_{k,\calD_0,f}^\star$ of $f$. Specifically, we consider the mean-squared loss to obtain the problem:
\begin{align}
    \underset{g:\calX \to [0,1]}{\text{minimize}} & \quad \frac12 \BE_{(\bX,Y)\sim \calD}\left[ (f(\bX)-g(\bX))^2 \right]  \label{eq:multiaccuracy optimization simplified} \\ 
    \text{subject to} & \quad  \left| \BE_{(\bX,Y)\sim \calD}\left[ c^\star_{k,\calD_0,f}(\bX)(g(\bX)-Y) \right] \right|  \le \alpha. \nonumber
\end{align}
where $\calD_0$ and $\calD$ are sets of i.i.d. samples that are sampled independently of each other. 

A closer look at~\eqref{eq:multiaccuracy optimization simplified} shows that it is a quadratic program (QP)\footnote{Please find details of QP formulation in Appendix \ref{appendix: QP}}. 
Thus, we can solve this QP through its dual problem to obtain a closed-form formula for the solution $g^\star$. The following formula follows by applying standard results on QP~\cite[Chapter~4.4]{boyd2004convex}. 

\begin{theorem} \label{theorem::optimal_correction}
    Fix two independently sampled sets of i.i.d. samples $\calD_0$ and $\calD$ from $P_{\bX,Y}$ with $|\calD|=n$, and let $\boldsymbol{f}=\boldsymbol{f}_{\calD}$, $\by=\by_{\calD}$, $\bc = \bc_{\calD_0,\calD}$, $\bA=\bA_{\calD_0,\calD}$ and $\boldsymbol{b}=\boldsymbol{b}_{\calD_0,\calD}$ be the fixed vectors and matrix as defined in~\eqref{eq:QP constraint 0}--\eqref{eq:QP constraint 2}. Denote an optimization variable $\calL = (\lambda_{+},\lambda_{-},\bxi_{+}^T,\bxi_{-}^T)^T\in \BR^{2n+2}$ and let $\bB = \frac{1}{2}\bA \bA^T \in \BR^{(2n+2)\times (2n+2)}$ and $\bd = \boldsymbol{b} - \bA \boldsymbol{f}$. Let $\calL^\star = (\lambda_{+}^\star,\lambda_{-}^\star, (\bxi_{+}^\star)^T,(\bxi_{-}^\star)^T)^T$ be the unique solution to the QP
    \begin{equation}
        \underset{\calL\ge \mathbf{0}}{\textup{minimize}} \ \calL^T \bB \calL + \bd^T \calL.
    \end{equation}
    Then, the predictors solving the optimization~\eqref{eq:multiaccuracy optimization simplified} are determined by their restriction to $\calD$ as
    \begin{equation}
        g(\bx_i) = f(\bx_i) + \lambda^\star c_{k,\calD_0,f}^\star(\bx_i) + \xi_i^\star
    \end{equation}
    where $\lambda^\star \define \frac{1}{n}(\lambda_{-}^\star - \lambda_{+}^\star)$ and $\bxi^\star \define  \bxi_{-}^\star - \bxi_{+}^\star$. Furthermore, the value of $\bxi^\star$ may be chosen\footnote{To see this, note that, thinking of $\lambda_{-}$ and $\lambda_{+}$ as constants, the optimization over a single pair $(\xi_{-,i},\xi_{+,i})$ takes the form of minimizing $(\xi_i+\lambda c_i)^2/2+f_i\xi_i+\xi_{+,i}$ over $\xi_{+,i}\ge 0$ and $\xi_i\ge -\xi_{+,i}$. The optimal value for this can be easily seen to be $\xi_i = -\lambda c_i - f_i$ if $\lambda c_i + f_i <0$, or $\xi_i=0$ if $\lambda c_i+f_i \in [0,1]$, or $1-\lambda c_i - f_i$ if $\lambda c_i + f_i > 1$. This translates to clipping $g$ to be within $[0,1]$.} so that $g$ is projected onto $[0,1]$. In particular, applying \KMultiAcc{} (Algorithm~\ref{alg:kernel_multiaccuracy}) with the value $\lambda=\lambda^\star$ attains a solution to~\eqref{eq:multiaccuracy optimization simplified}.
\end{theorem}

\section{Experiments} \label{section:experiments}

\begin{figure*}[!tb]
     \centering
     \resizebox{1.0\linewidth}{!}{\begin{tabular}{c}
     \includegraphics[]{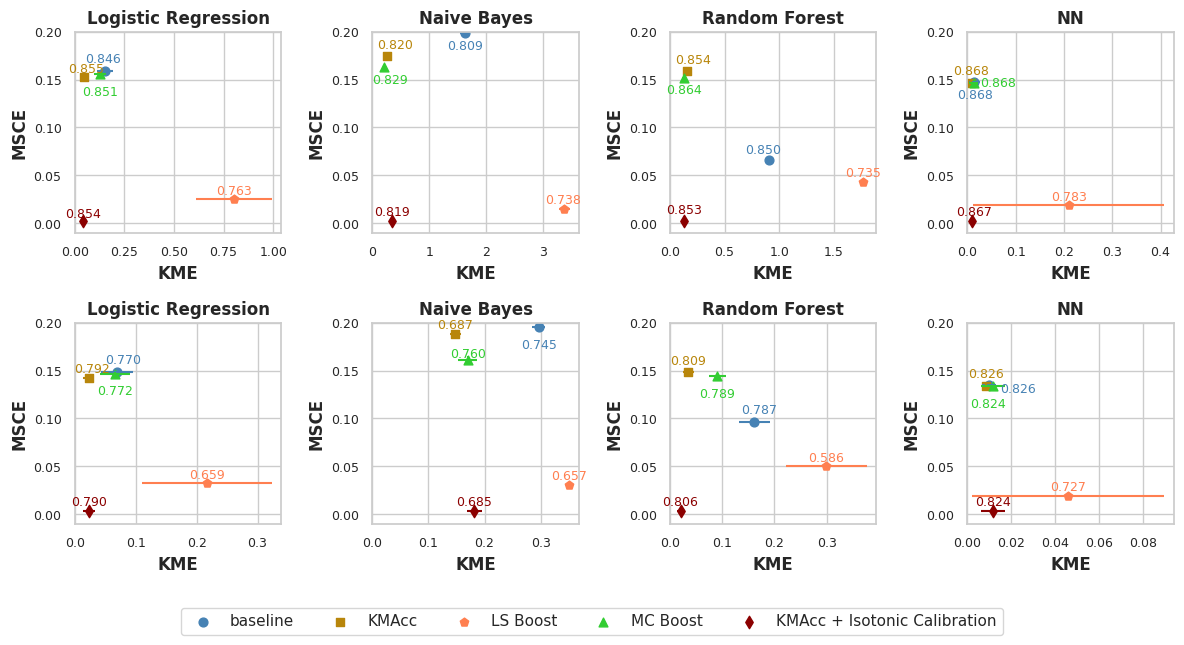}   
     \end{tabular}}
      \caption{
      Multiaccuracy error (KME, Definition~\ref{definition:emp KME}) vs. calibration error (MSCE, Definition~\ref{Definition::MSCE}) for \KMultiAcc (our method), competing methods (\LSBoost~and \MCBoost), and \KMultiAcc~with isotonic calibration, a standard score quantization technique. 
      Predictor performances are measured as AUC and labeled next to each method.
      \KMultiAcc~achieves improved or comparable KME and AUC to the baselines and \MCBoost~(with the exception of Naive Bayes baseline classifier). Notably, \KMultiAcc~+ isotonic calibration significantly improves MSCE while maintaining KME and AUC, with more favorable results than both \LSBoost and \MCBoost.
      Results are shown for the FolkTables Income of WA dataset (first row) and the Health Coverage of WI dataset (second row), with error bars on both axes\protect\footnotemark. }\label{fig::fig2}
\end{figure*}
 \footnotetext{The MSCE standard deviation is often imperceptible}
We benchmark our proposed algorithm, \KMultiAcc~(Algorithm~\ref{alg:kernel_multiaccuracy}), on four synthetic datasets and eight real-world tabular datasets\footnote{Implementation of \KMultiAcc can be found at \url{https://github.com/Carol-Long/KMAcc}}. We demonstrate \KMultiAcc's competitive or improved performance among competing interventions, both in multi-group fairness metrics and in AUC. 
Full experimental results are provided in Appendix \ref{apdx:proofs_ch3}.

\subsection{Datasets}
We provide experimental results from the US Census dataset FolkTables. We conduct 4 binary classification tasks, including ACSIncome, ACSPublicCoverage, ACSMobility, ACSEmployment, using two different states for each of these tasks. In addition, we generate four  synthetic datasets using the \texttt{sklearn.datasets} class in Scikit-Learn~\cite{pedregosa2011scikit}---moons, concentric circles, blobs with varied variance, and anisotropically distributed data. 

\subsection{Competing Methods}
We benchmark our method against \LSBoost{}\footnote{We use the official implementation of \LSBoost{} available at \url{https://github.com/Declancharrison/Level-Set-Boosting}.} by \cite{globus2023multicalibration} and \MCBoost{}\footnote{We use the official implementation of \MCBoost{} available at \url{https://osf.io/kfpr4/?view_only=adf843b070f54bde9f529f910944cd99}.} by \cite{kim2022ua}, which are (to the best of our knowledge) the two existing algorithms of multi-group fairness with usable Python implementations. 

The mechanism of \LSBoost{} is the following: over a number of level set partitions, each called $v$, \LSBoost{} finds a function $c_v^{t+1} \in \calC$ through a squared error regression oracle before updating a function $f_{t+1}$ as a rounding of the values to each level set using a successive updating of indicator values as to which set they lie in under the previous $f_t$ combined with the learned $c^{t+1}_v$: $\hat{f} = \sum_{v \in [1/m]} \mathbf 1 [f_t(x) = v]\cdot c_v^{t+1}(x)$, and $f = \text{Round}(\hat{f}_{t+1}, m)$. This updating continues so long as an error term measured by the expectation of the squared error continues to decrease at a rate above a parameterized value. $\calC$ is taken to be linear regression or decision trees.

The \MCBoost{} algorithm performs an iterative multiplicative weights update applied to successively learned functions. Starting with an initial predictor $p_0$, it learns a series of grouping functions $c(x)\in \calC$, that maximize multiaccuracy error. The algorithm now stores a set of both calibration points $S$ and validation points $V$, at each step generating the set $S_t$ by, $\forall (x, y) \in S$, having $(x, y - p_t(x)) \in S_t$. Then, using the weak agnostic learner on $S_t$, it produces a function $c$ which has its multiaccuracy checked on the validation set $V$ with the empirical estimate of the multiaccuracy error before enacting a multiplicative weights update $f_{t+1}(\bx) = e^{-\eta h_{t, S}(\bx)} \cdot f_t(\bx)$ if the multiaccuracy error is large. 
There are three different classes $\cal{C}$ it might draw $c$ from -- either taking sub-populations parameterized by some number of intersections of features, using ridge regression, or using shallow decision trees.

\subsection{Performance Metrics}
We evaluate the performance of baseline and multi-group fair models across three metrics: Kernel Multiaccuracy Error (KME, Definition \ref{defn::multiaccuracy_error}), Area Under the ROC Curve (AUC), and Mean-Squared Calibration Error (MSCE), where MSCE is defined as follows.

\begin{definition}[Mean-Squared Calibration Error (MSCE),~\cite{globus2023multicalibration}]
\label{Definition::MSCE}
The Mean-Squared Calibration Error (MSCE) over a dataset $\calD$ of a predictor $f:\calX\to [0,1]$ with a countable range $R(f)$ is defined by
\begin{align*}
    \textup{MSCE}(f,\calD) \define \sum_{v\in R(f)} \Pr_{(\bX,Y)\sim\calD}[f(\bX)=v]\cdot \left( \ExpVal{(\bX,Y)\sim\calD}{(Y-v) \mid f(\bX)=v} \right)^2,
\end{align*}
\end{definition}
Our algorithm optimizes for KME and utility, while \LSBoost{}~\cite{globus2023multicalibration} optimizes for MSCE. Hence, both of these metrics are reported. \MCBoost{}~\cite{kim2019multiaccuracy} optimizes for multiaccuracy error (without considering calibration functions in the kernel space) and classification accuracy. We report AUC since it captures the models' performance across all classification thresholds.

\subsection{Methodology} \label{section:methodology}

To implement and benchmark \KMultiAcc{}, we proceed through the following steps.

\textbf{Data splits.}
We assume access to a set of i.i.d. samples $\calD'$ drawn from $P_{\bX,Y}$, where $P_{\bX,Y}$ is a distribution over $\calX\times \calY = \BR^m \times \{0,1\}$. We randomly partition $\calD'$ into four disjoint subsets: $\calD_{\text{train}}$ (for training the baseline predictor $f$), $\calD_0$ for computing the witness function $\empCstar$, $\calD_{\text{val}}$ for finding the proportionality constant $\lambda^\star$), and finally $\calD_{\text{test}}$ for benchmarking the performance of \KMultiAcc~in a test set against the state-of-the-art methods. 

\textbf{Baseline predictor $f$.}
 Using the training data $\calD_{\text{train}}$, we learn a baseline classifier $f$. Our algorithm treats this function $f$ as a black box. For our experiments, we use four distinct supervised learning classification models as a baseline: Logistic Regression, 2-layer Neural Network, Random Forests, and Gaussian Naive Bayes, all implemented by Scikit-learn~\cite{pedregosa2011scikit}. We train these on $\calD_{\text{train}}$, values that are not used in learning our witness or in \KMultiAcc. 

\textbf{Learning the witness function.}
We take as our class of calibration functions the unit ball $\mathfrak{C}_k$ in the RKHS $\mathcal{H}_k$ (Equation~\ref{eq:C RKHS def}) with the kernel $k$ being the RBF kernel, given explicitly for a parameter $\gamma>0$ by
\begin{equation}
    k_\gamma(\bx, \bx') = \exp\left(- \gamma \| \bx - \bx' \|_2^2\right)
\end{equation}
The value of $\gamma$ is a hyperparameter that we finetune using $\calD_0$. We conduct a grid search over the parameter $\gamma$ to find a $\gamma^\star$ such that $\empCstar$ has maximal correlation with the errors $y-f(\bx)$, thus obtaining $\empCstar \in \mathfrak{C}_k$ in terms of $f$, $\gamma$,  and $\mathcal{D}_0$ (see Proposition~\ref{prop::witness_multiaccuracy}). 
To carry out this step, we run grid search on $\gamma$ using K-fold validation on the data $\calD_0$. %For each $\gamma$, we find the average correlation between the value of $\empCstar(\mathcal{D}_{0, x})$ and $\mathcal{D}_{0, y} - f(\mathcal{D}_{0, x})$.% 
% Then, we find the $\gamma^\star$ that maximizes the average correlation between $\empCstar$ and the errors, and obtain the final function $\empCstar$. \wa{[discuss here or in conclusion modifications of how to find $\gamma^\star$]}
The value of the normalizing constant $\theta$ in Proposition~\ref{prop::witness_multiaccuracy} (for attaining $\|\empCstar\|_{\calH_k}=1$) can be skipped in this step for the sake of finding the optimal multiaccurate predictor $g^\star$ solving~\eqref{eq:multiaccuracy optimization simplified}, because $\theta$ can be subsumed in the value of the optimal parameter $\lambda^\star$. %Finally, it is worth noting that our results on both the synthetic data and experiments indicate that the RBF kernel tends to be a reasonable choice that produces a useable witness function, but further investigation of other kernels, and the implications of the selection of a given kernel, is an intriguing avenue of future research.

\textbf{Performing \KMultiAcc.} Using $\mathcal{D}_{\text{val}}$, we perform a simple grid search to find the smallest $\lambda$ such that the multiaccuracy condition (Definition \ref{defn::multiaccuracy_error}) is met (alternatively, one could solve the QP detailed in Theorem \ref{theorem::optimal_correction}).
%with $\mathbf{c}= \empCstar(\mathcal{D})$, and $\mathbf{f}= f$. 
Running \KMultiAcc{} (Algorithm \ref{alg:kernel_multiaccuracy}), we update $f$ using $\lambda$ and $\empCstar$ to obtain the multi-group fair classifier $g^\star=f+\lambda\empCstar$. 
%In practice, as solving the QP for an explicit $\lambda^\star$ can require large amounts of memory, we may grid search for an appropriate $\lambda^\star$ instead, by testing what value of $\lambda$ produces the smallest change in predictions such that the multiaccuracy condition (Definition \ref{defn::multiaccuracy_error}) is met  on the validation data. 

\subsection{Results}

\begin{figure*}[!htb]
  \centering
  \resizebox{\linewidth}{!}{\begin{tabular}{cccc}
     \includegraphics[]{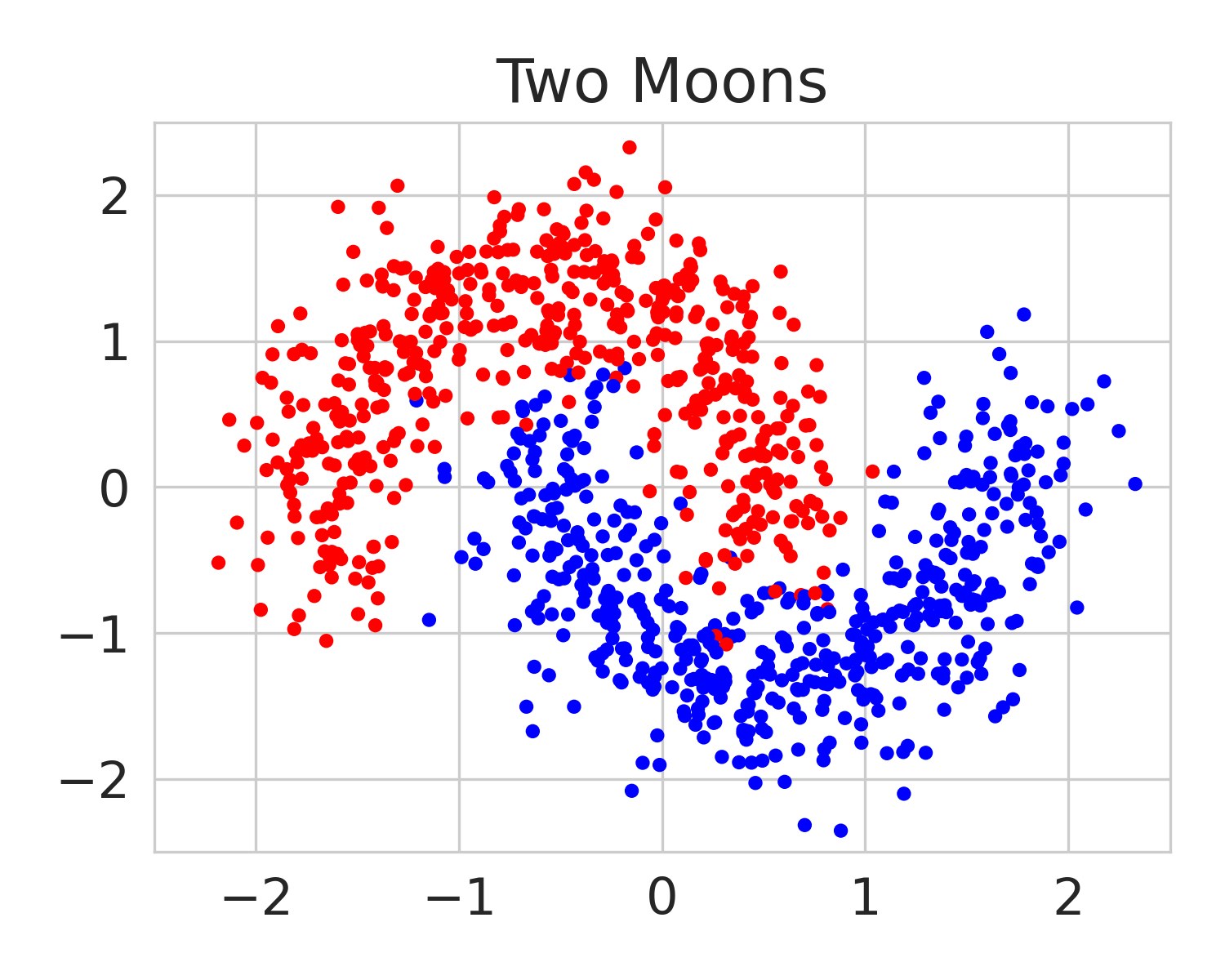} &
     \includegraphics[]{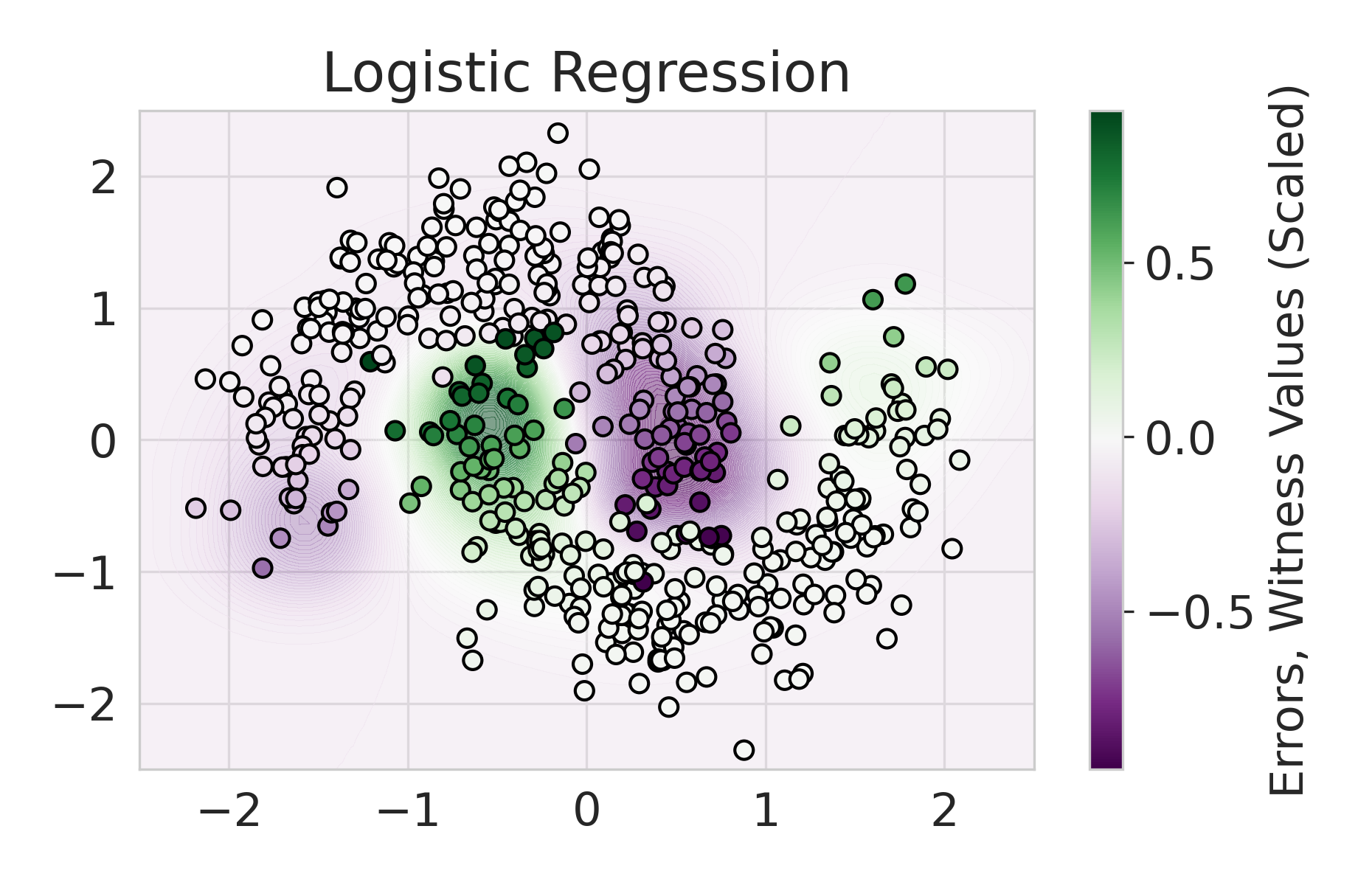} &
    \includegraphics[]{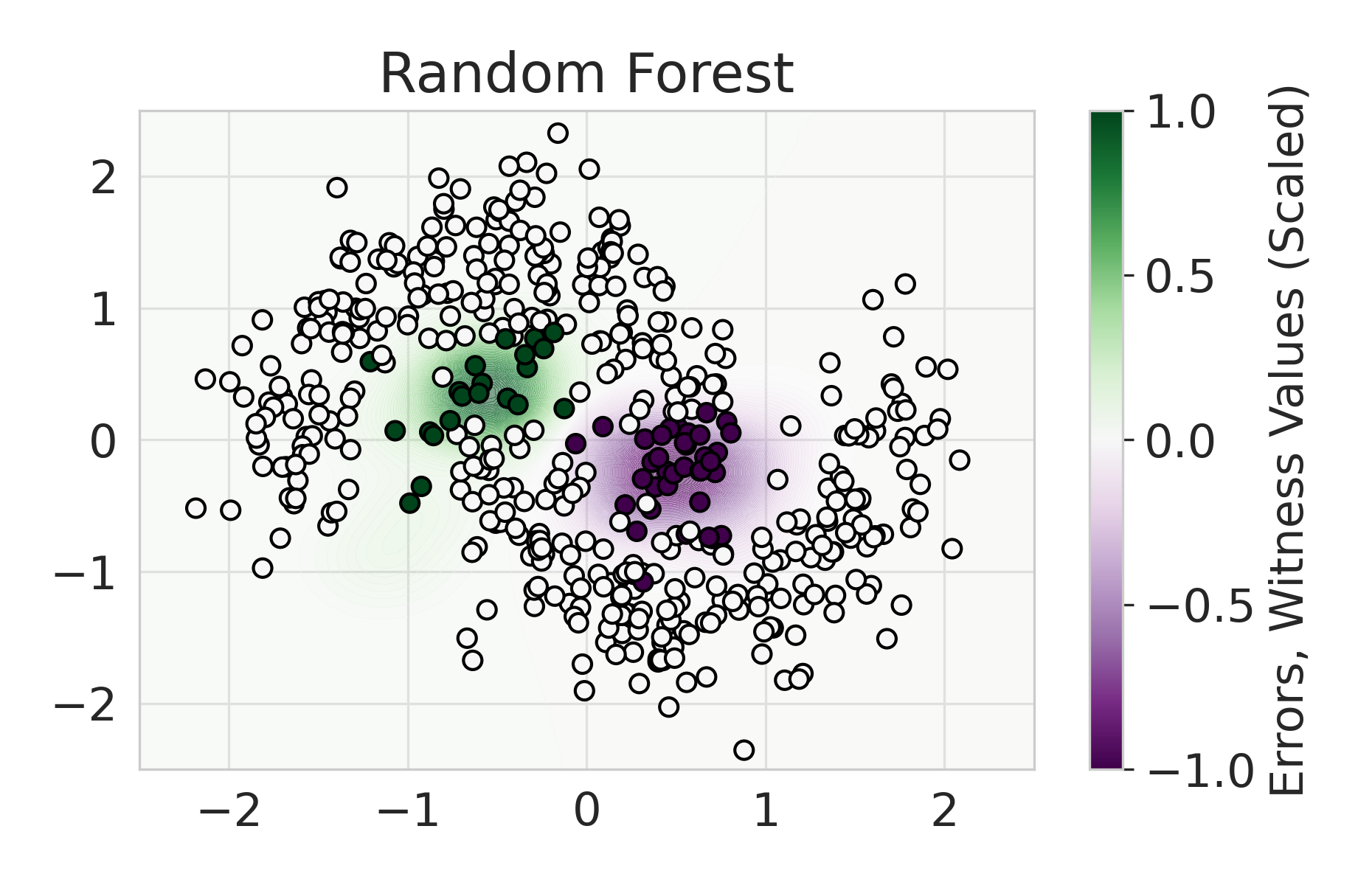} &
    \includegraphics[]{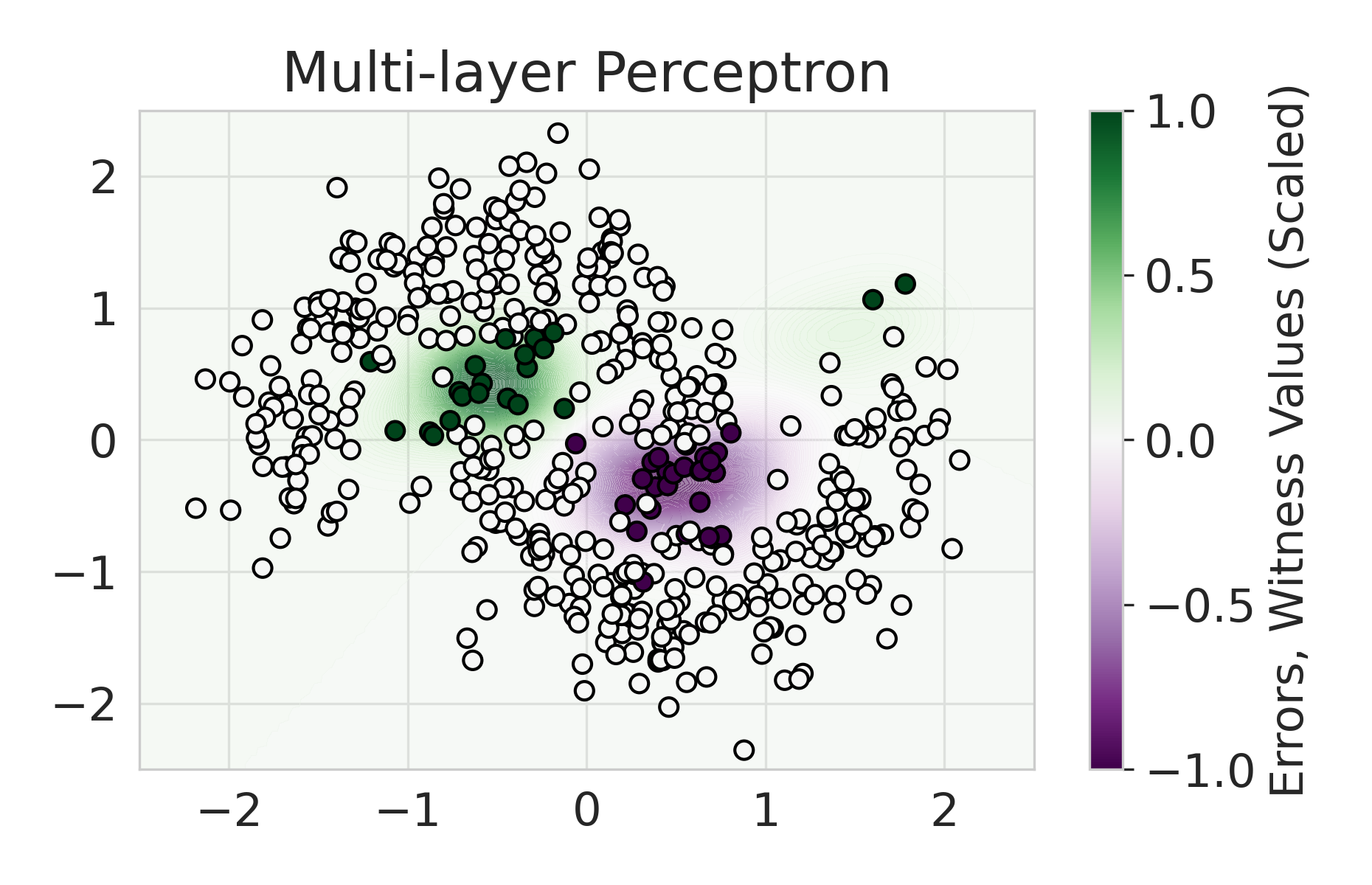}\\
     \includegraphics[]{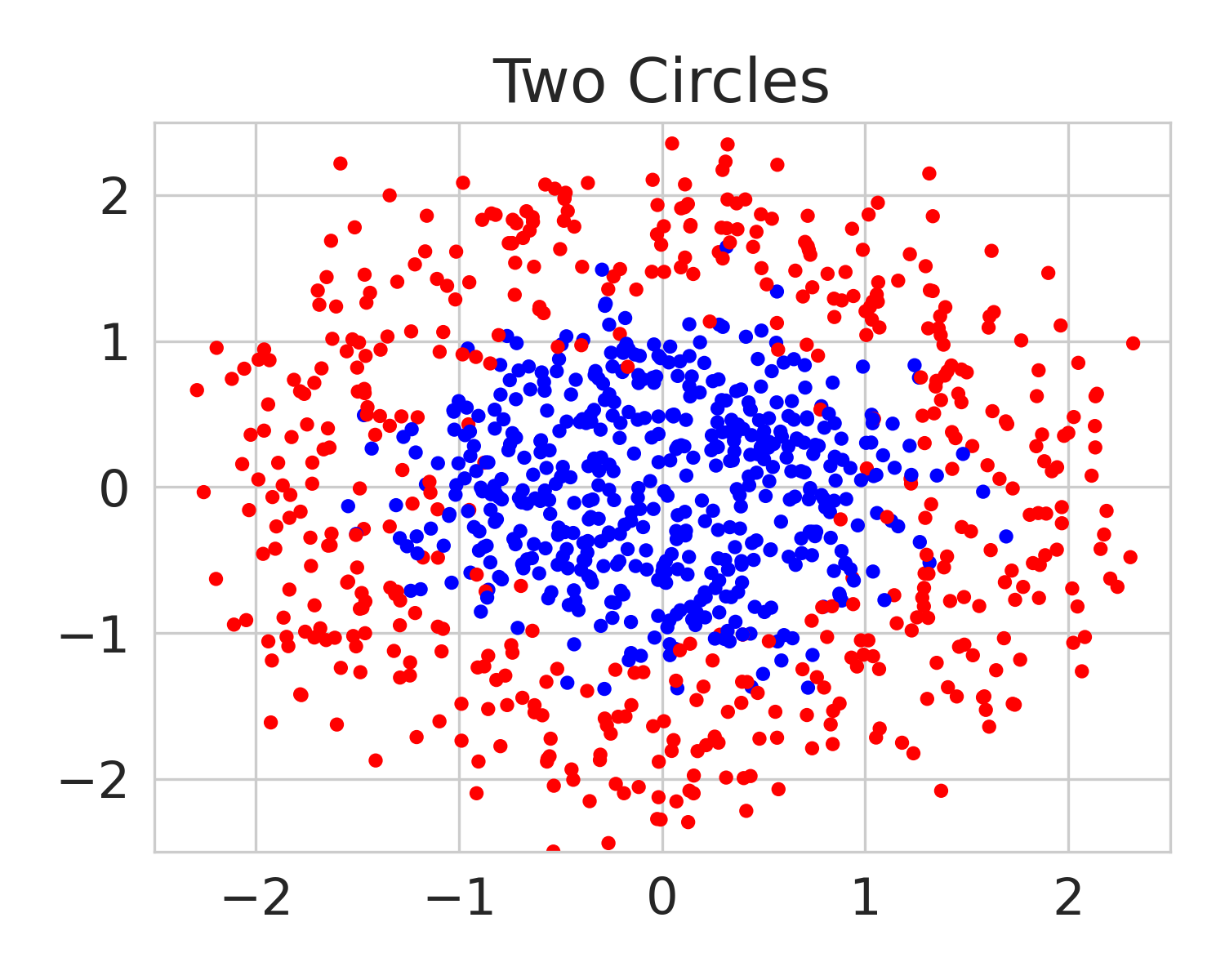} &
     \includegraphics[]{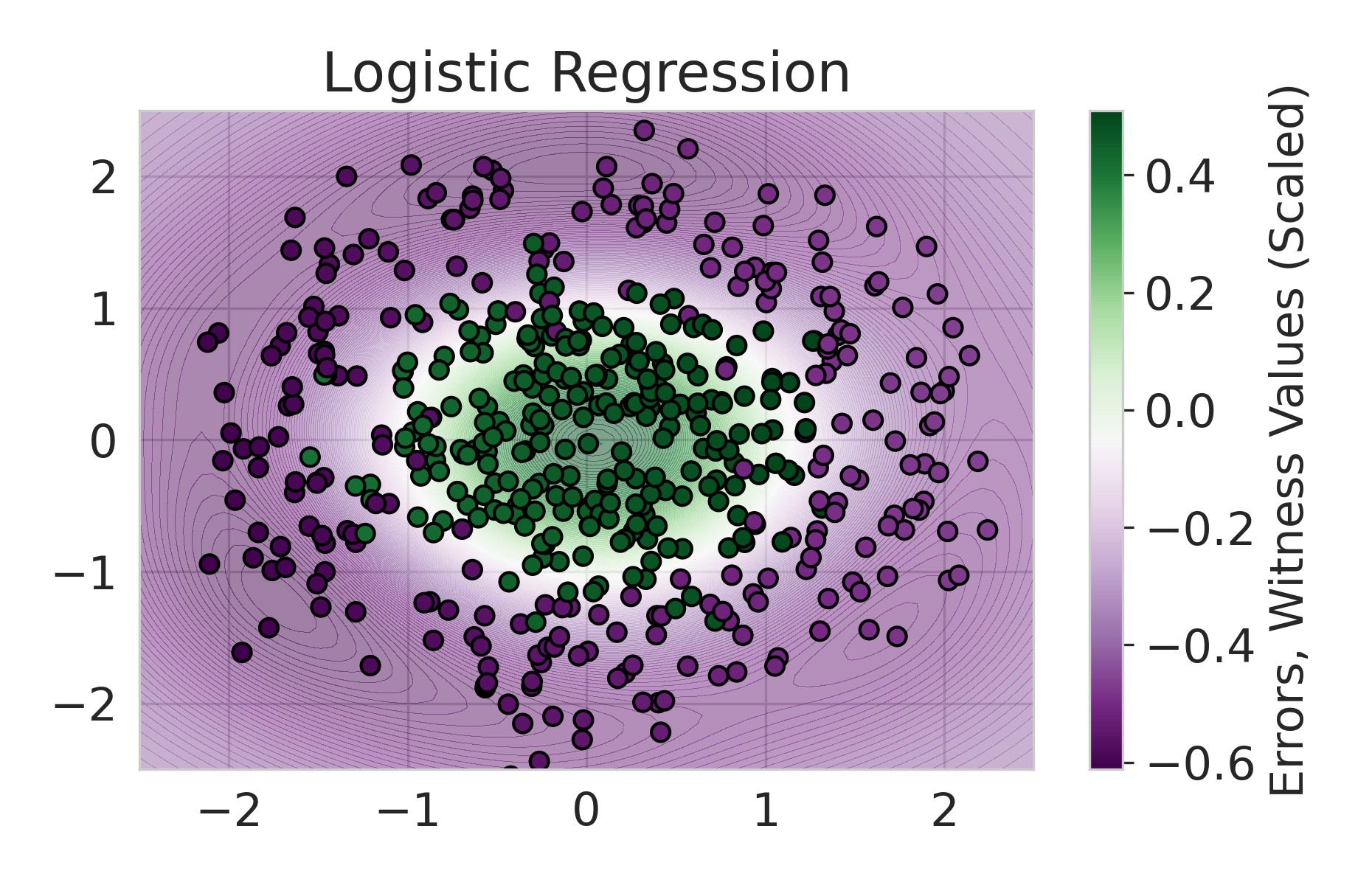} &
    \includegraphics[]{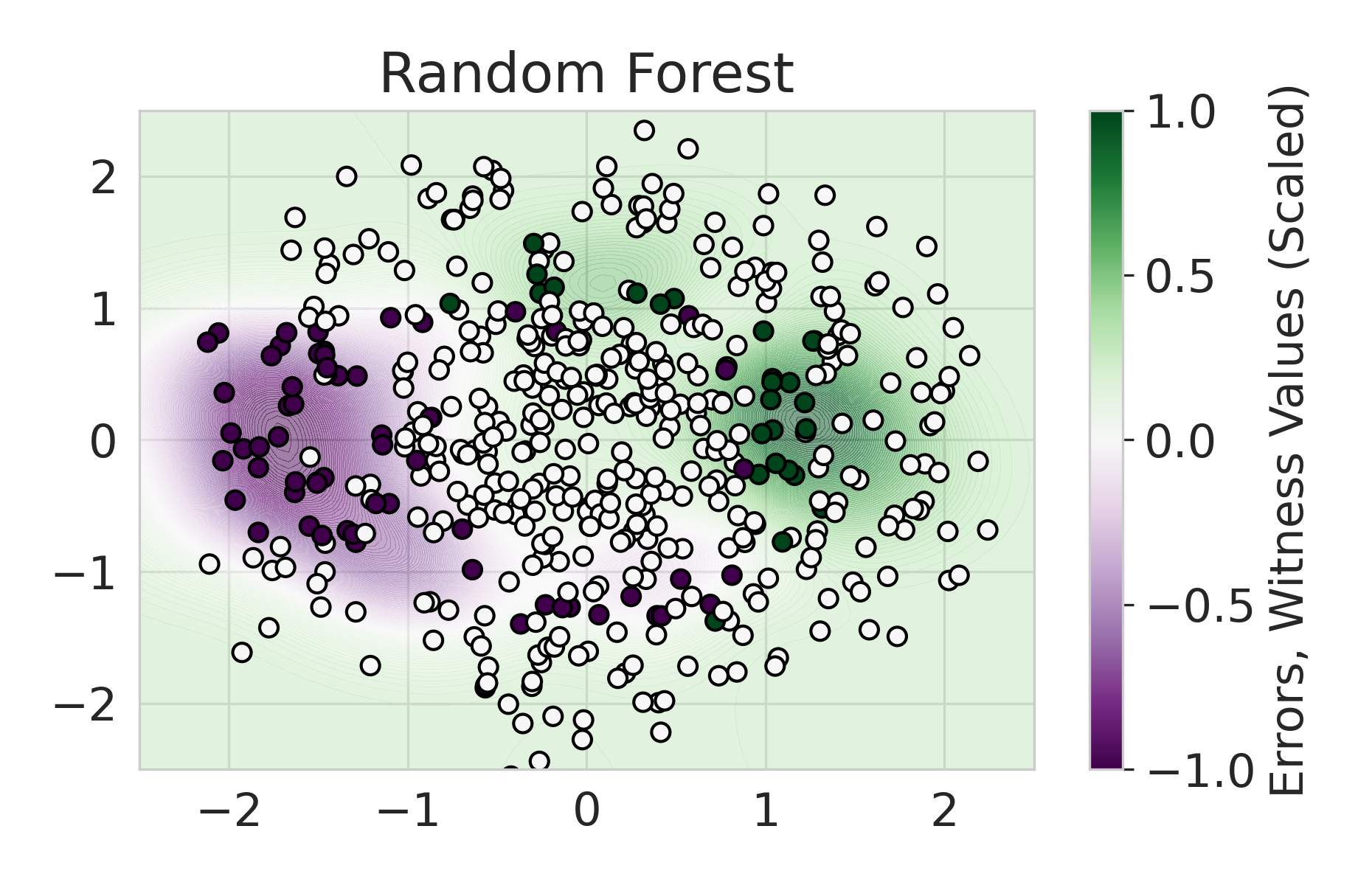} &
    \includegraphics[]{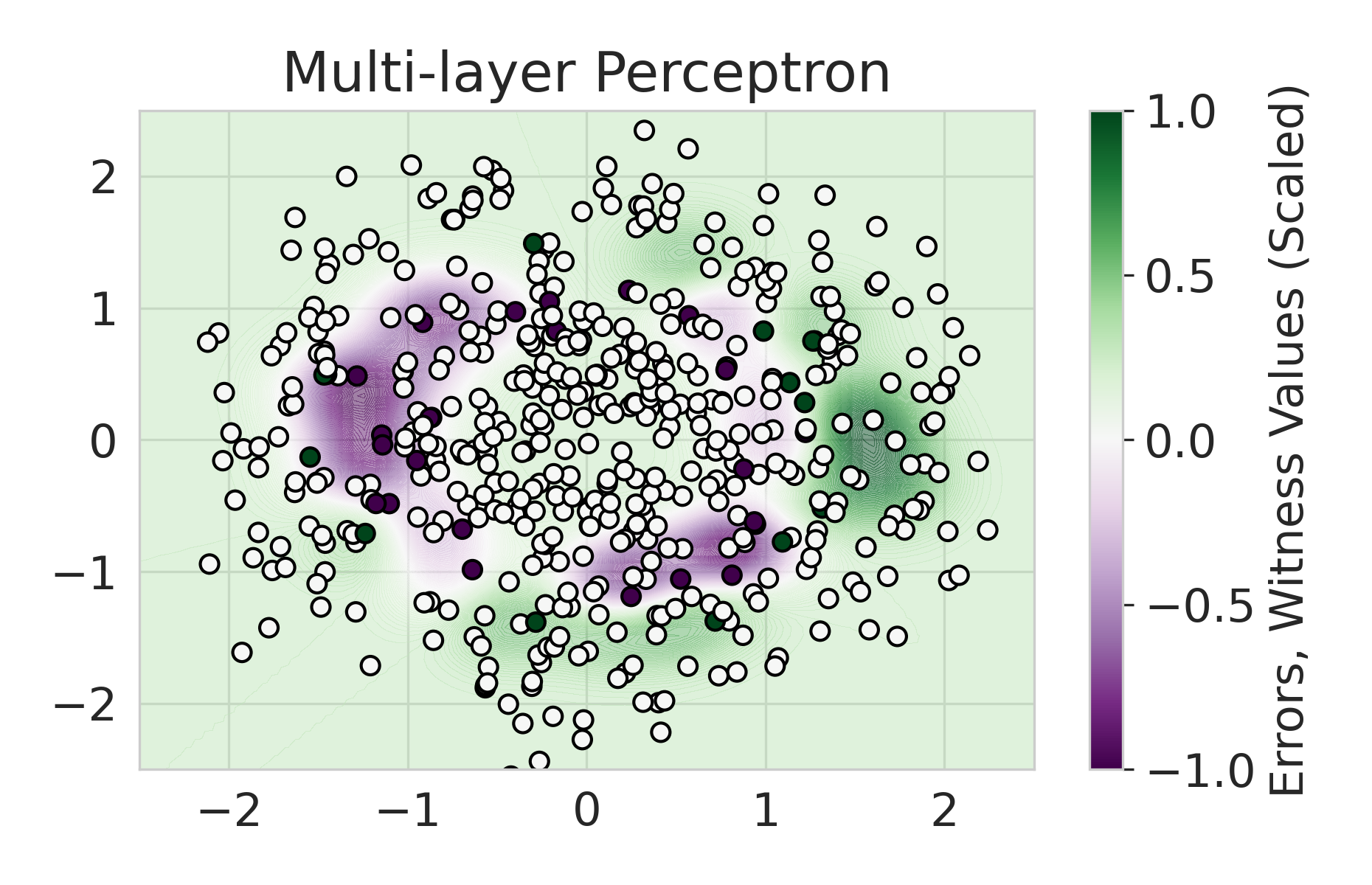}\\
     \includegraphics[]{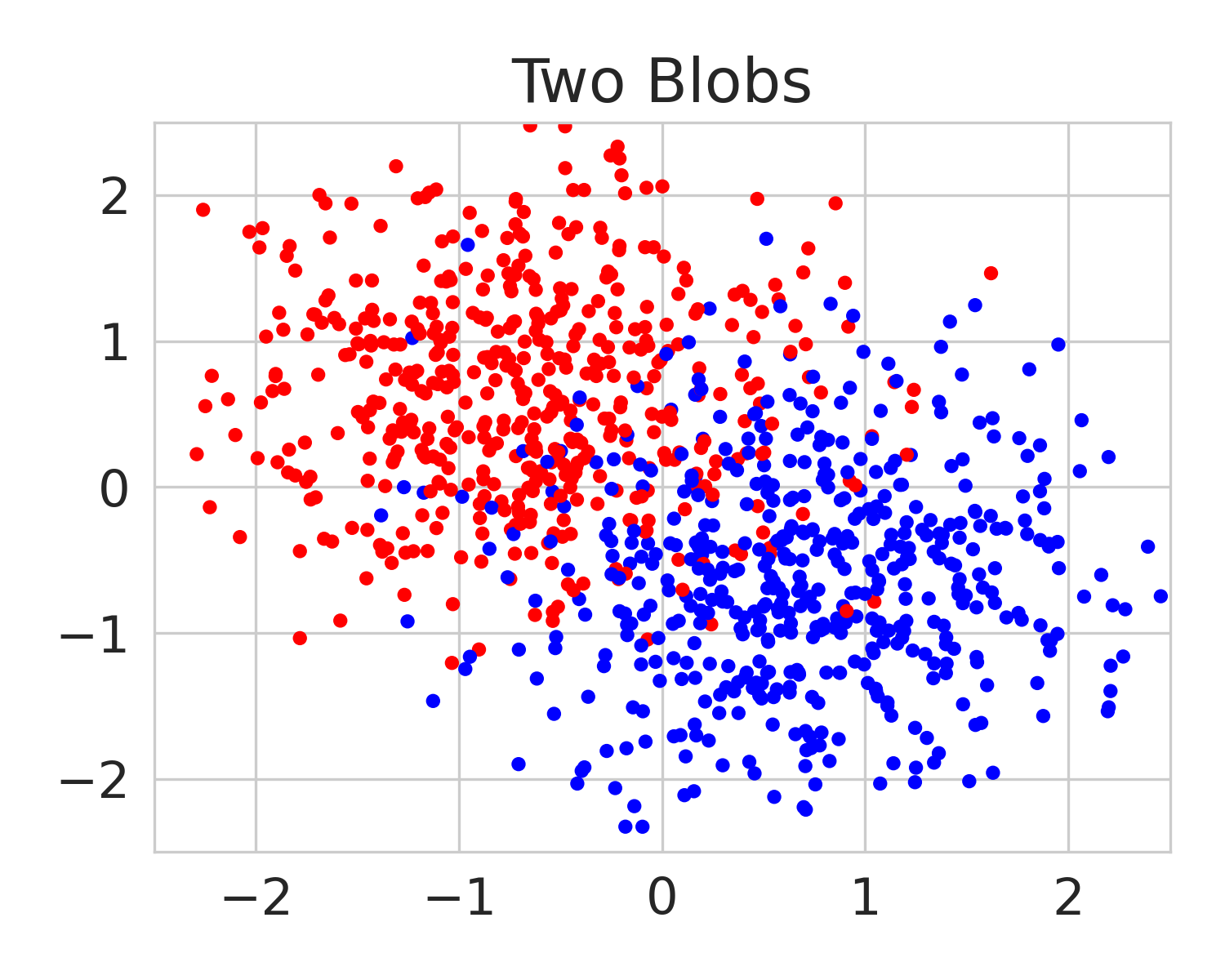} &
     \includegraphics[]{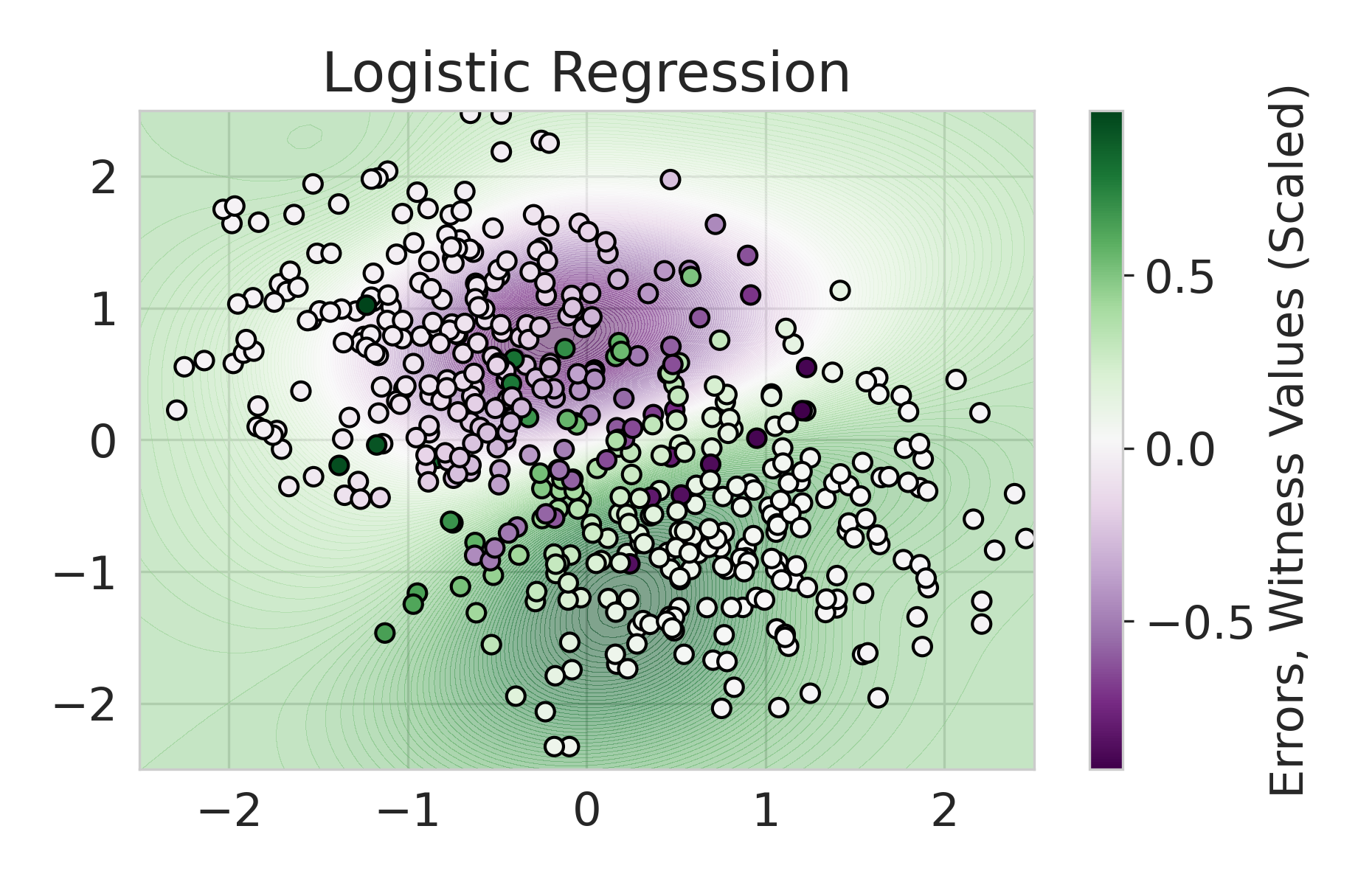} &
    \includegraphics[]{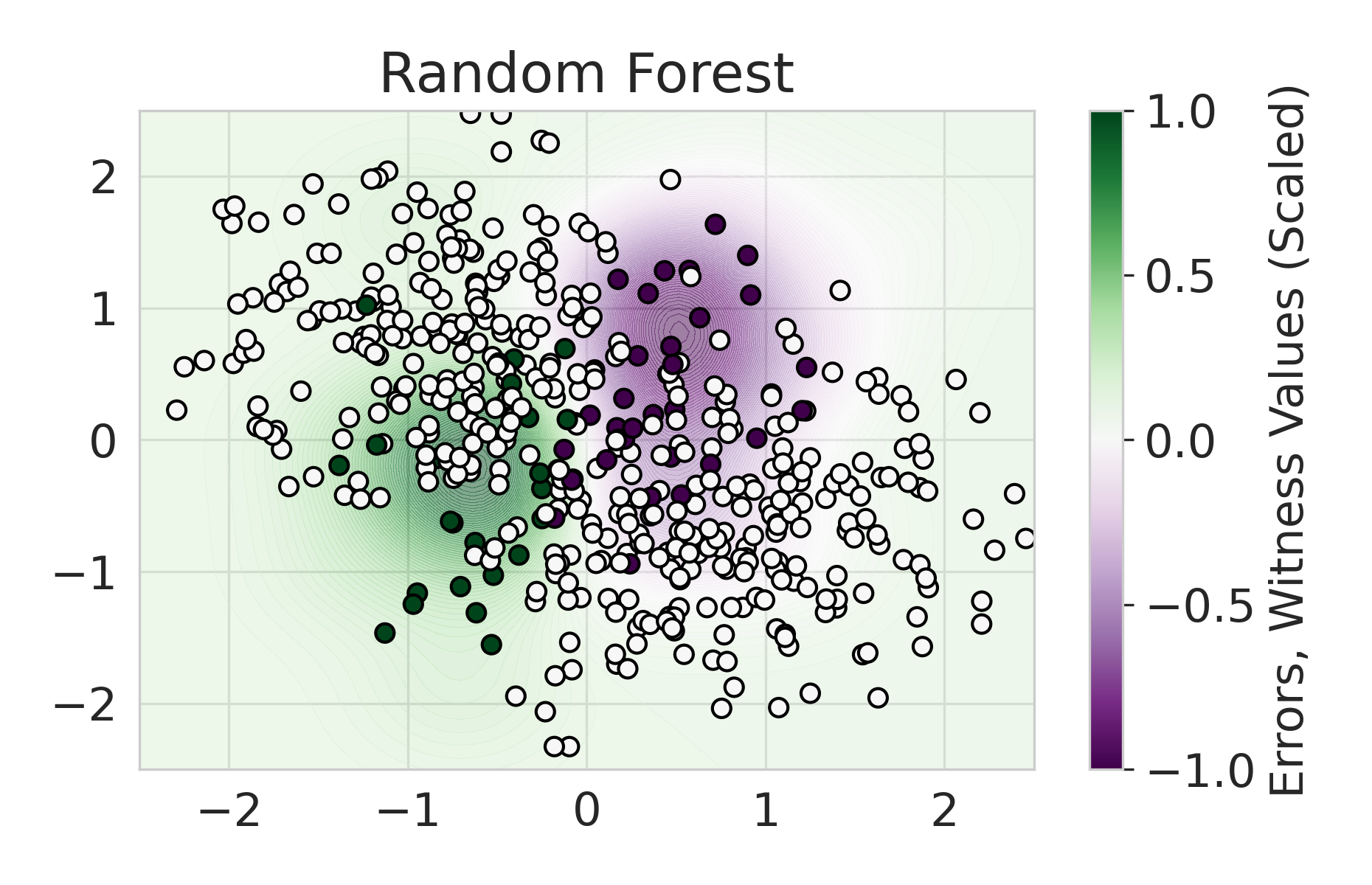} &
    \includegraphics[]{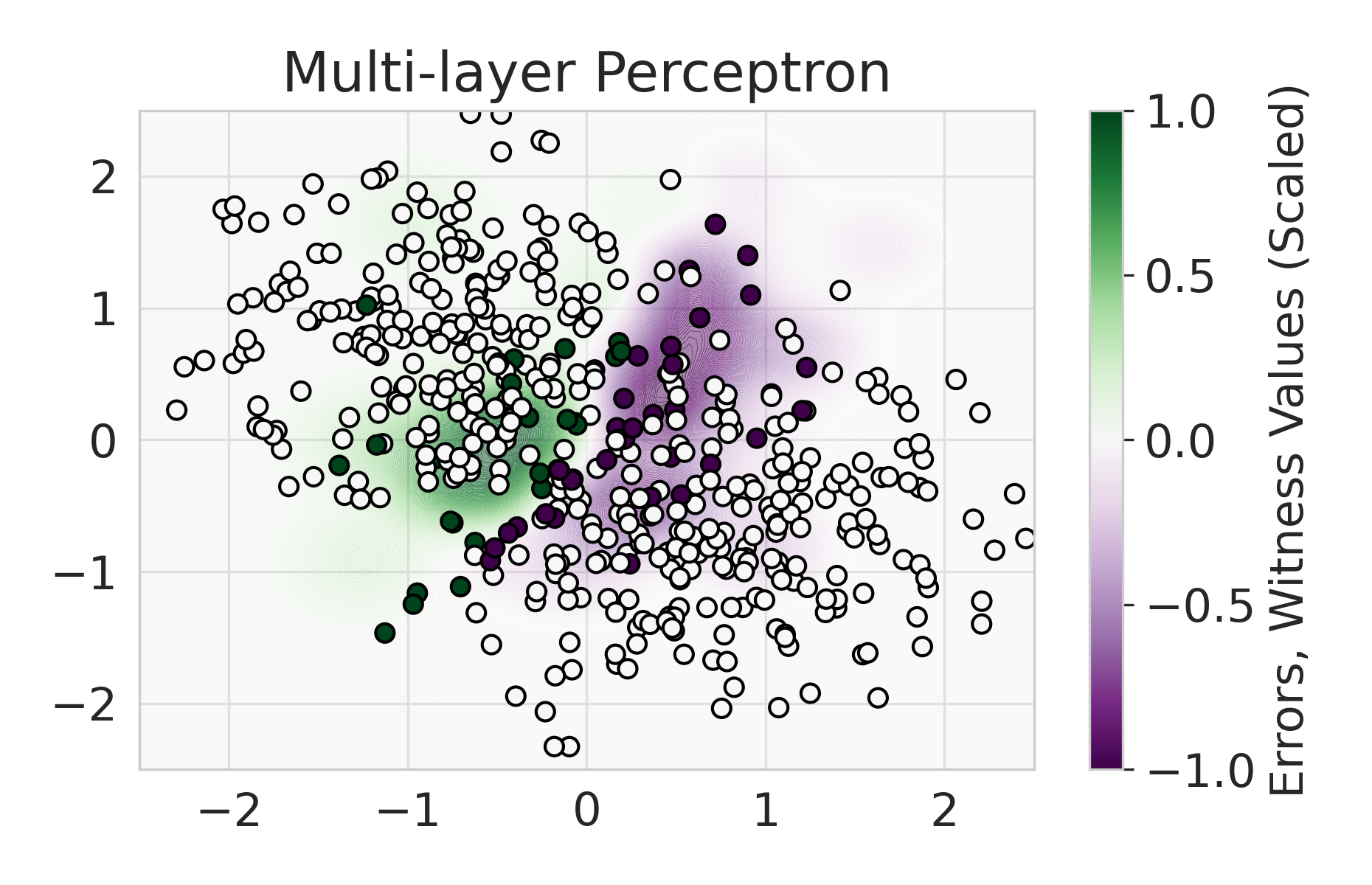}\\
     \includegraphics[]{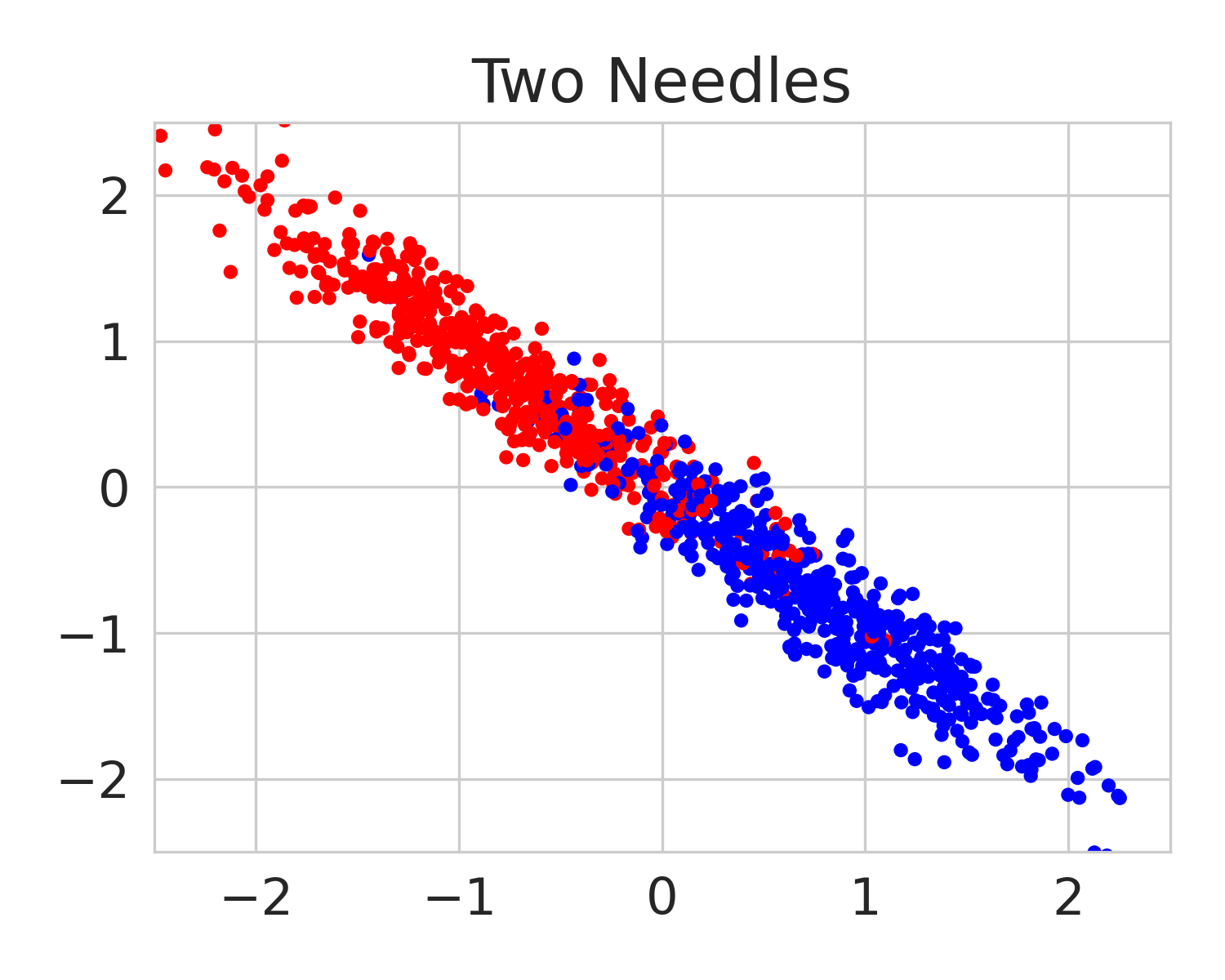} &
     \includegraphics[]{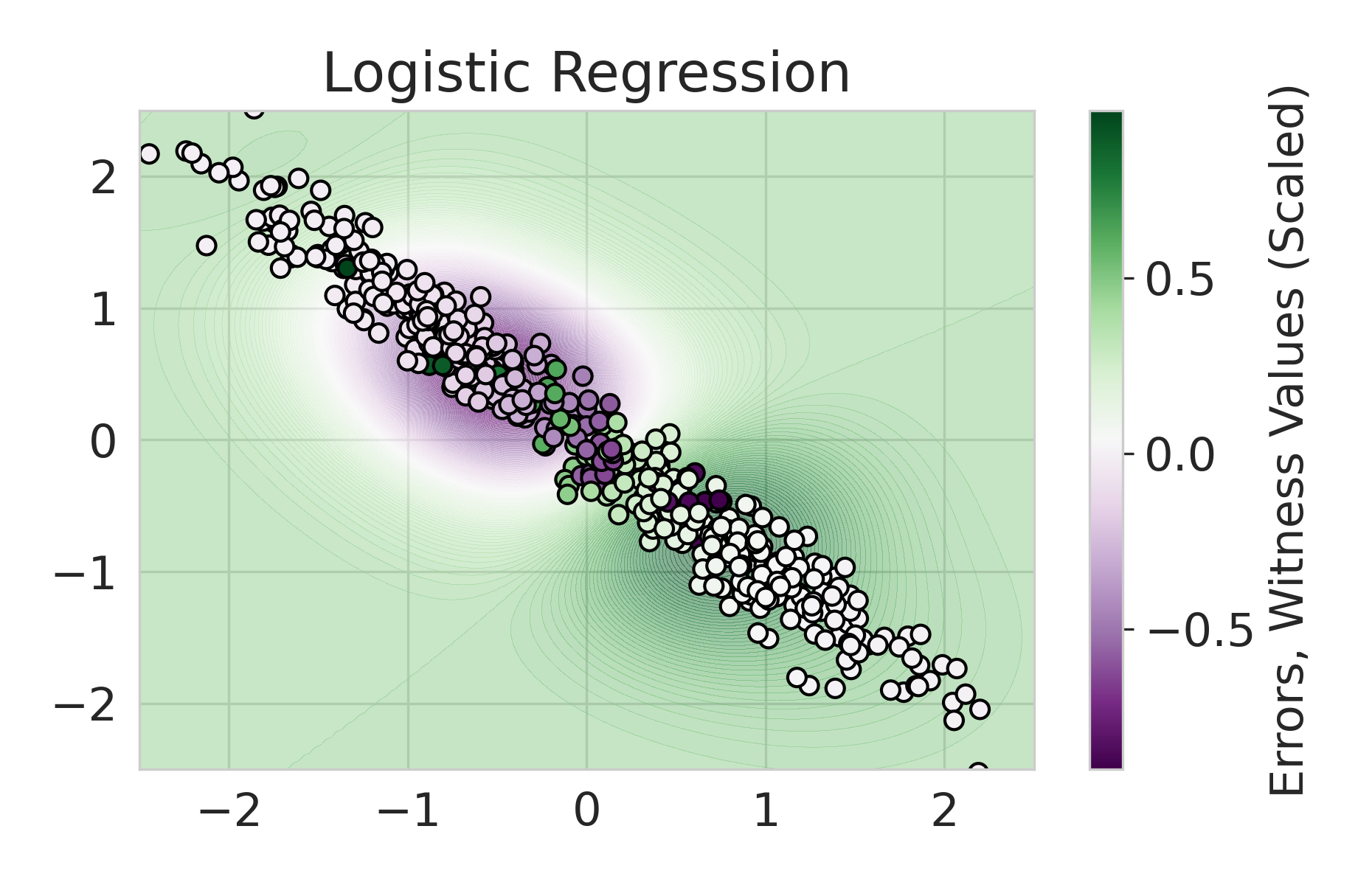} &
    \includegraphics[]{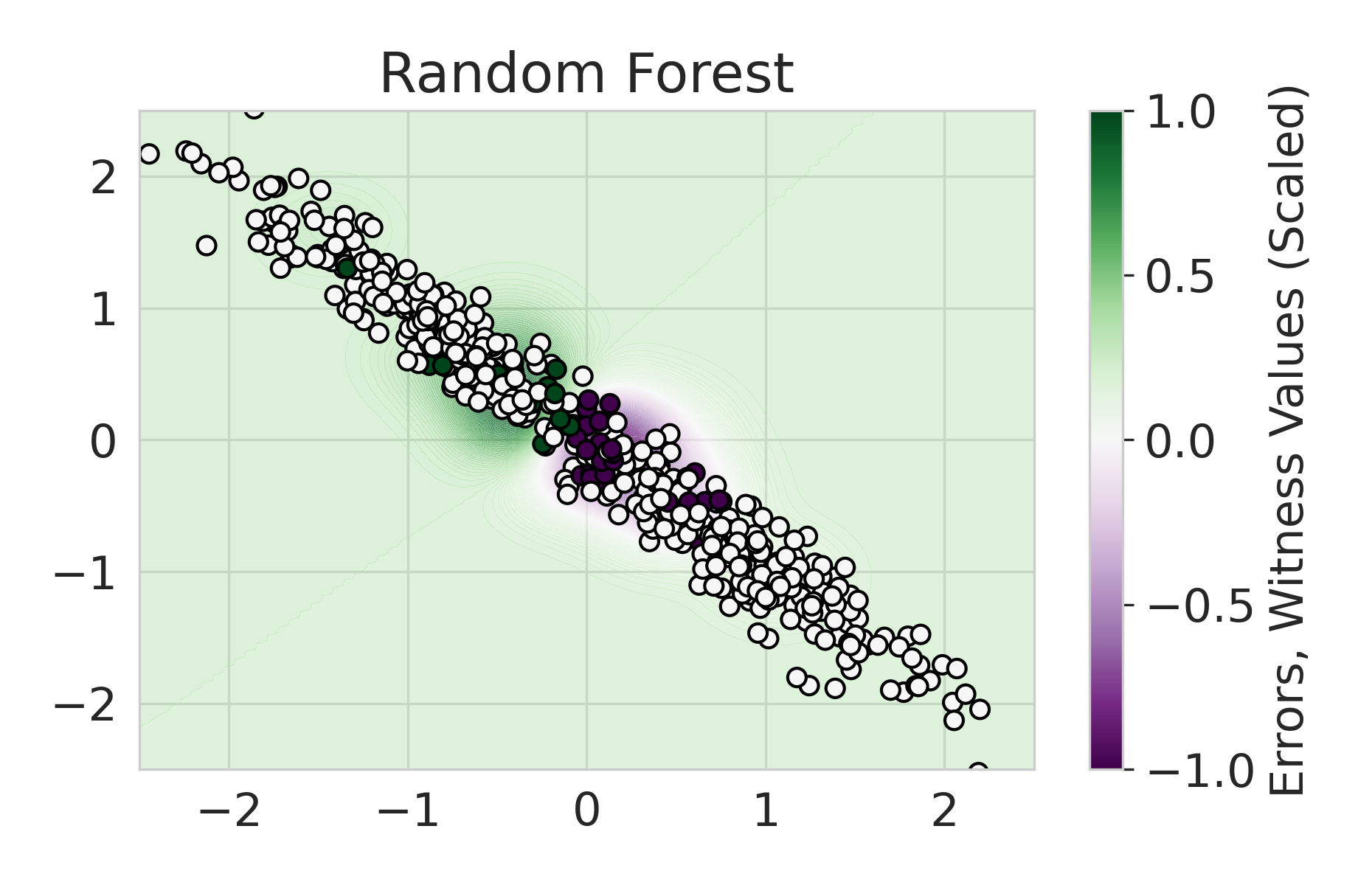} &
    \includegraphics[]{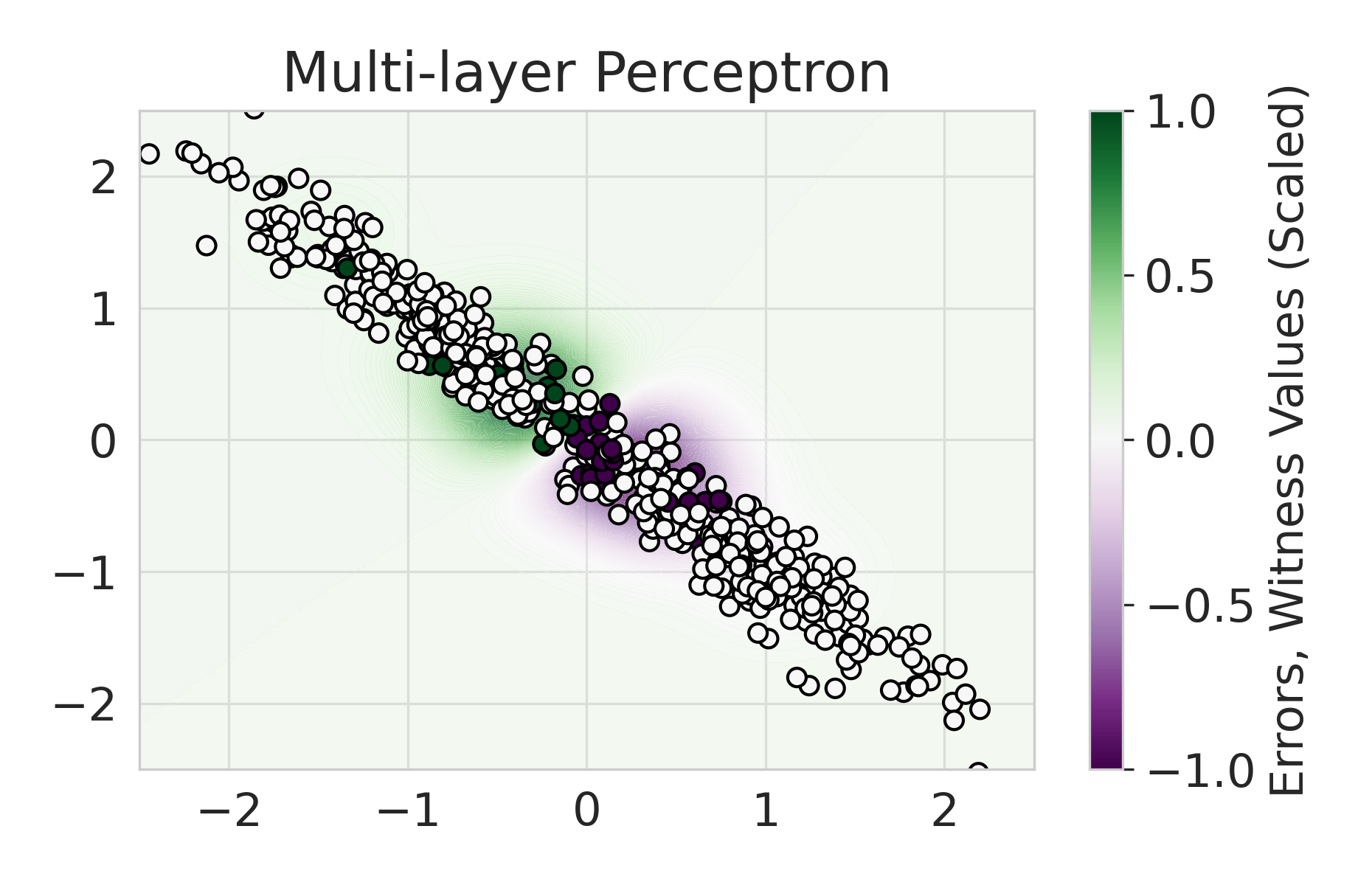}
     \end{tabular}}
  
  \caption{Test errors over witness value contours using the RBF kernel. \textbf{First Column: }Visualization of the moon, concentric-circle, blob, and needle datasets. Red and blue represent the true labels. \textbf{Second Column: }Classification via a logistic regression classifier. Witness function values (Definition \ref{def:witness}) $\empCstar$ is plotted as a contour under the error of the classifiers on test samples $y-f(\bx)$. The witness function values are highly correlated with the errors of the predictor. Dark green and dark green dots mark where the classifier is most erroneous for the blue class and the red class, respectively. The linear regression model is not capable of classifying concentric circles, resulting in almost the entire blue class being misclassified. Similarly, we show a similar pattern between classifier error and the witness function values for a random forest classifier (\textbf{Third Column}) and  a multi-layer perceptron classifier \textbf{Fourth Column}. 
  %Visualization of the witness function for the moon-circle dataset, 2-concentric-circle dataset, 2 blobs with varied variance dataset, and anistrophically distributed dataset. Witness function values $c^*(x)$ is plotted as contour under the error of the classifiers on test samples $y-f(\bx)$. The dark green and dark purple dots are samples where the linear regression model is most erroneous. The second column are the results from logistic regression, the third column are the results from random forest regression, and the fourth column are the results from a 2-layer neural network, all using the rbf kernel.
  }
\label{fig::rbf-synthetic_results}
\end{figure*}

With the process described in Section~\ref{section:methodology}, we test \KMultiAcc across various baseline classifiers using implementations in Scikit-Learn~\cite{pedregosa2011scikit}. In each US Census dataset, we execute five runs of each model on which we report, showing the mean value of each metric alongside error bars.

Firstly, on synthetic datasets, we demonstrate that the witness function is a good predictor of classifier error. In Figure \ref{fig::moon_circle_results}, we train a logistic regression classifier on the moon and circle datasets to perform binary classification. 
The classifier has an accuracy of 0.85 and AUC of 0.94, and most errors occur in the middle where the red and blue classes are not linearly separable. Indeed, samples with high errors in scores $y-f(x)$ also receive high predicted errors in terms of witness function values.
Indeed, the scatter plot (\textbf{Right Column}) illustrates the linear correlation between test error and witness value, with a high Pearson correlation coefficients of 0.828. Complete results using additional baseline models (Random Forest and Multi-layer Perceptron) are shown in Figure \ref{fig::rbf-synthetic_results}.

On US Census datasets and as demonstrated in Figure \ref{fig::fig2}, \KMultiAcc~achieves the lowest KME relative to competing models without sacrificing AUC, and \KMultiAcc~paired with isotonic calibration achieves the lowest multi-group metrics (KME and MSCE) while maintaining competitive AUC. In Figure \ref{fig::fig2}, baseline models (blue circle) have high MSCE, and most have non-negligible KME, with the exception of neural networks. 
Post-processing the baseline models using \KMultiAcc~ (yellow rectangle), we see a significant reduction in KME from the baseline (shifting to the left of the plot), and in a majority of experiments, the post-processed models achieve, on the test set, the pre-specified KME constraint with $\gamma_k(g,P_{\bX,Y}) < .01$. To target low calibration error (measured by MSCE on the y-axis), we apply off-the-shelf isotonic calibration on top of \KMultiAcc. We observe that applying \KMultiAcc$+$Isotonic Calibration (red diamond) to baseline results in low errors on both axes (KME and MSCE). Across all baselines and experiments, applying either \KMultiAcc or \KMultiAcc$+$Isotonic Calibration does not degrade the predictive power of the models---the AUCs (labeled next to each model) of models corrected by the proposed methods either stay relatively unchanged or improved.

Competing method \MCBoost{} achieves effective reduction in KME with minimal improvement on MSCE, without sacrificing AUC. We note that \KMultiAcc+Isotonic Calibration enjoys comparable or better performance with regards to \MCBoost{} on KME and better performance on MSCE, while eliminating the need for iterative updates to minimize miscalibration that is required in \MCBoost{}.
\LSBoost{}(orange polygon) achieves low MSCE while worsening both KME and AUC. 
%By applying a standard score quantization technique, either isotonic calibration or K-Means (grey cross in Appendix), on top of \KMultiAcc can effectively achieve lower MSCE without loss of predictive power as measured by AUC.

\section{Discussion and Conclusion}
%summary
We connect the multi-group notions to Integral Probability Measures (IPM), providing a unifying statistical perspective on Multiaccuracy, Multicalibration, and OI. 
This perspective leads us to a simple yet powerful algorithm (\KMultiAcc) for achieving multiaccuracy with respect to a class of functions defined by an RKHS.
\KMultiAcc{} boils down to first predicting the error of the classifier using the witness function, and then subtracting the error away.
This algorithm enjoys provable performance guarantees and empirically achieves favorable accuracy and multi-group metrics relative to competing methods.

A limitation of our empirical analysis in comparison to other methods is that we optimize over the calibration function class being the unit-ball RKHS with the RBF kernel, which may not be the set of calibration functions for which other benchmarks achieve the lowest multiaccuracy or multicalibration error on.
Furthermore, while the proposed method achieves favorable multicalibration results, this algorithm does not have provable guarantees for multicalibration. Developing a multicalibration-ensuring algorithm through the IPM perspective is an exciting future direction.

%discussion
To conclude, this work contributes to the greater effort of reducing embedded human bias in ML fairness. To this end, we adopt RKHS as the expressive group-denoting function class to ensure multi-group notions on, rather than using predefined groups. It remains an open question to explore the structure of the witness function---the most biased group-denoting function in the RKHS--- and its relationship to the predefined group attributes, which may inform us of the intersectionality and the structure of errors in ML models.

\chapter{Watermarking Large Language Models}
\label{ch:4}

\section{Information-Theoretic Trade-offs and Optimized Couplings}

\subsection{Introduction}

Large language models (LLMs) are now able to produce text that is indistinguishable from human-generated content. This has fueled the development of watermarks that imprint a ``signal'' in LLM-generated text with minimal perturbation of an LLM's output. This section provides an analysis of text watermarking in a one-shot setting through the lens of hypothesis testing with side information.

A large language model (LLM) is a generative model that, given a string of input tokens, outputs a probability distribution $Q_X$ for the next token $X$ in the sequence. The emergence of LLMs that generate text that is largely indistinguishable from humans has led to the creation of trustworthy text generation algorithms \cite{huang2024trustllm} that create safe \cite{bai2022constitutional}, interpretable \cite{geva2021transformer}, and authentic \cite{lin2022truthfulqa} content. This work focuses on \emph{watermarking}: the process of embedding a ``signal'' at the token level in LLM-generated text. The goal of a watermark is to enable automated detection of AI-generated content, providing proof of its authenticity (or lack thereof) and potentially of its origin. The past two years have witnessed the creation of increasingly sophisticated LLM watermarking schemes \cite{kirchenbauer2023watermark,christ2024undetectable,kuditipudi2023robust,zhaoprovable,aaronson2023watermark,he2024universally,bahri2024watermark,dathathri2024scalable,yang2023watermarking,ren2024subtle,huunbiased2024,zhao2024permute,chao2024watermarking,qu2024provably,xie2024debiasing,liuadaptive}.

A hallmark of existing LLM watermarks is their reliance on either distorting or coupling the next-token distribution $Q_X$ with a random variable $S$ drawn from a known distribution $P_S$. Here, $S$ represents shared randomness known both by the watermark generator and detector. For instance, \cite{kirchenbauer2023watermark} -- which ignited the recent interest in LLM watermarking in the machine learning community -- distorts $Q_X$ by randomly choosing a set of tokens (as determined by $S$) to be on a ``green list,'' i.e., a subset of tokens that are favored during generation, and increasing the mass of those tokens accordingly. The detector then counts the number of tokens in a sequence that appears on the green list and declares the text watermarked (i.e., AI-generated) if this count exceeds a threshold. However, such a distortion of the LLM distribution may impair the textual quality. Alternative approaches include \cite{aaronson2023watermark,kuditipudi2023robust,he2024universally,chao2024watermarking}, which instead couple $Q_X$ with the distribution $P_S$. Such couplings enable ``distortion-free'' watermarks that (averaged over $P_S$) do not change the expected next-token distribution, yet are still detectable.

The exact nature of the shared randomness $S$ between the model and the detector varies across watermark implementations. $S$ can be, for example, generated from the hash of previous tokens in a sequence \cite{kirchenbauer2023watermark} (where a hash function converts the token history into a fixed-size value that deterministically produces pseudo-random bits) or sophisticated tournament-like sampling strategies \cite{dathathri2024scalable}. For our theoretical analysis, we abstract away the exact generation process of the shared randomness $S$.

At a high level, existing LLM watermarks perform two steps when generating a sequence of tokens $\{X_i\}_{i=1}^n$ given shared randomness $\{S_i\}_{i=1}^n$:
\begin{enumerate}
    \item \emph{Watermark Generation:} For the $i$-th generated token and given $S_i$ and the predicted next token distribution $Q_{X}$, draw the next token by sampling from $X_i\sim \tilde{Q}_{X|S_i}.$
    \item \emph{Detection:} Given a sequence $\{(X_i,S_i)\}_{i=1}^n$, compute the statistic
        $T_n=\frac{1}{n}\sum_{i=1}^n f(X_i,S_i)$ for some function $f:\cX\times\cS\to[0,1]$,
    and declare that the sequence $\{X_i\}_{i=1}^n$ is watermarked if $T_n\geq \tau$.
\end{enumerate}
Importantly, a crucial assumption of current LLM watermarking schemes is that the function $f$ \underline{\emph{does not}} assume knowledge of the token distribution $Q_{X^n}$. This allows watermarks that are directly detectable from the sequence $\{(X_i,S_i)\}_{i=1}^n$, i.e., directly from generated text, without accessing the underlying LLM. If the distribution of the generated tokens $Q_{X^n}$ was known, then a standard likelihood ratio test (LRT) would suffice for watermark detection. What makes LLM watermarking distinct from existing information-theoretic watermarking schemes (e.g., \cite{gel1980coding,willems2000informationtheoretical,chen2000design,moulin2003information,martinian2005authentication,villan2006text}) are the assumptions that (i) the source distribution is unknown to the watermark detector and (ii) watermarking is performed on a per-token (vs. sequence) level.

\subsection{Main Contributions}\label{sec:main_contributions}
Motivated by the success of token-level schemes for LLM watermarking, we provide an in-depth analysis of a single-token watermarking process, i.e., when $n=1$.
Specifically, we study how to generate a coupling $\tilde{Q}_{X,S}$ and the corresponding detection function $f$ that maximizes the probability of detection of the watermark, while controlling the quality of the text.
The latter is controlled through the distortion relative to $Q_X$ -- a quantity we call \emph{perception}, following recent trends in the information theory literature on the source coding problem \cite{blau2019rethinking,theis2021coding,chen2022rate}. We refer to this setting as \emph{one-shot watermarking}. We jointly optimize $\tilde{Q}_{X,S}$ and $f$ given a perception constraint, with the case $\bar{Q}_X=Q_X$ corresponding to the \emph{perfect perception} setting. 
We focus on one-shot watermarking since, as mentioned above, existing schemes are constrained to watermark on a token-by-token basis. Moreover, small gains in single-token watermark detection compound to exponential gains in detection accuracy in threshold tests applied across multiple tokens.

We begin with an information-theoretic formulation for one-shot watermarking. We quantify the fundamental trade-off between watermark detection vs. perception when the underlying next-token distribution $Q_X$ is known with the side information $P_S$ uniformly distributed. This analysis yields a fundamental upper bound on one-shot watermark performance; see Theorems \ref{thm:opt_cornerpoints} and \ref{thm:universal_ub}. Interestingly, when the watermark does not change the next-token probability (i.e., perfect perception), optimizing a one-shot watermark is equivalent to maximizing the TV-information $ \mathsf{TV}\left(Q_{X,S}\|Q_XP_S\right)$ across the coupling $Q_{X|S}$ -- a non-convex optimization problem \cite[Section~7]{polyanskiy2025information}. This formulation embeds TV-information with a new operational interpretation. 

We optimize one-shot watermarks when $Q_X$ is unknown to the detector but satisfies a min-entropy constraint, i.e., 
$\|Q_X\|_\infty \leq \lambda$ (Eq. \eqref{eq:maxmin}), which corresponds to $H_\infty(Q_X) \geq -\log(\lambda)$. Operationally, lower values of $\lambda$ correspond to higher entropy token distributions with greater uncertainty, while higher values of $\lambda$ indicate more concentrated distributions where the next token is more predictable.
Moreover, we optimize for detection tests of the form $\ones[f(X)=S]$,  where $f:\calX\to\cS$ forms a partition of $\cX$.  

Motivated by the fact that deterministic token partitions lead to low detection probabilities, we introduce randomness to $f$. In Theorem \ref{thm:optimal_maxmin_detection}, we analyze the probability of detection of such detection tests under the worst-case token distribution.  
We pair our analysis with a characterization of the optimal design of the partition randomization.
In Theorem \ref{thm:approx_maxmin_detection}, we consider a simplified token partition strategy and show that it yields a near-optimal detection probability.
Together, we provide a complete characterization of the minimax detection rate for a given vocabulary size, side information, and min-entropy constraint under the optimal and near-optimal partition randomization strategies. 
Lastly, we provide numerical results of the Correlated Channel (CC) benchmarked against Gurobi-based optimum coupling\cite{gurobi}  and the red/green watermark\cite{kirchenbauer2023watermark}.

\textbf{Related Work.} 
Watermarking has been extensively studied in information theory \citet{chen2000design,moulin2003information,martinian2005authentication}, particularly through the Gelfand-Pinsker (GP) channel \citet{gel1980coding,villan2006text,willems2000informationtheoretical}.
These approaches typically focus on watermarking sequences via joint typicality and assume perfect knowledge of the underlying source distribution. 
The  work of \citet{kirchenbauer2023watermark} led to various developments in watermarking schemes \cite{aaronson2023watermark,he2024universally, bahri2024watermark,dathathri2024scalable,yang2023watermarking,ren2024subtle,huunbiased2024,zhao2024permute,chao2024watermarking,qu2024provably,xie2024debiasing,liuadaptive}, with several approaches focusing on distortion-free methods, e.g., \citet{kuditipudi2023robust,huunbiased2024,zhao2024permute,christ2024undetectable}. In particular, \citet{chao2024watermarking} proposes a watermark using error-correcting codes leading to correlated channels similar to the ones we find via optimizing couplings. 
In \citet{huang2023towards}, the optimal Type-II error for bounded Type-I error is analyzed by comparing watermarking schemes to the uniformly most powerful watermark with knowledge of $Q_X$. 
The authors of \citet{he2024universally} characterize the universal Type II error while controlling the worst-case Type-I error by optimizing the watermarking scheme and detector.
While these works operate on a token-level basis, they focus on the effect of a given strategy along a sequence.
In contrast, we focus on a preliminary step and aim to answer the simple yet important question -- \textit{What is the optimal coupling when watermarking a single token?}

\begin{figure}[!tb]
    \centering
    \includegraphics[trim={115pt 170pt 183pt 100pt}, clip, width=.6\linewidth]{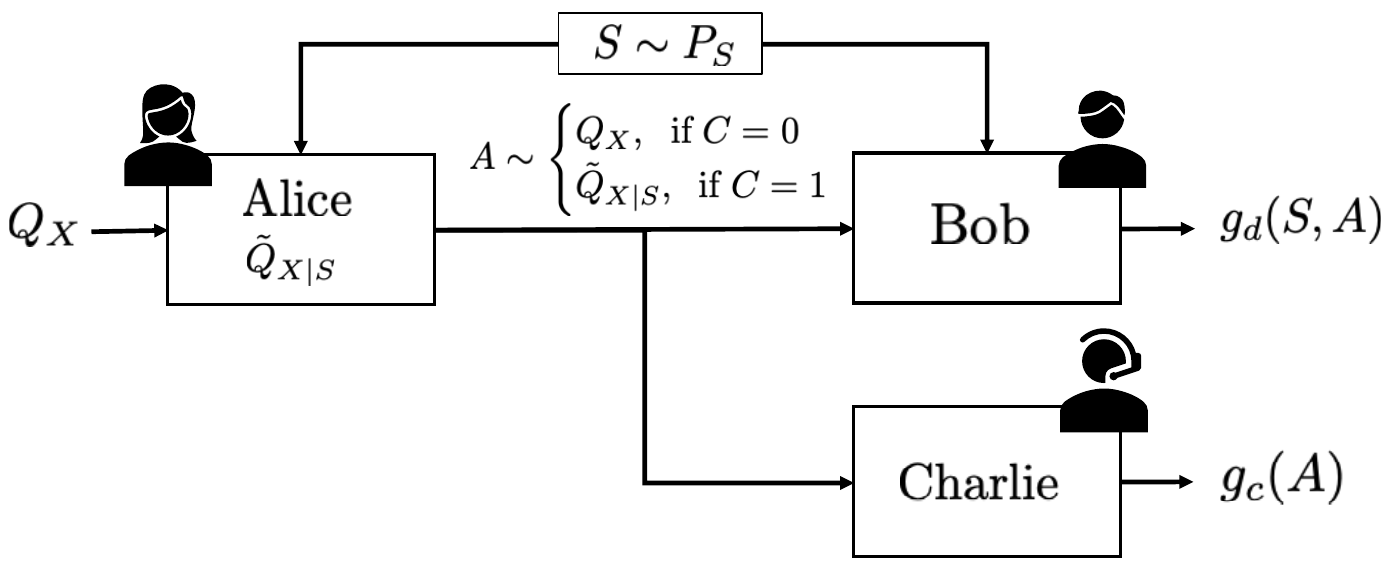}
    \caption{Watermarking problem as a hypothesis test with side information.}
    \label{fig:operational_problem}
\end{figure}

\subsection{Optimal One-Shot Watermarking}
In this section, we formulate the watermarking problem, derive the resulting optimization problem, and discuss the optimal solution structure.
We focus on the fundamental trade-off between detection probability and perceptual quality.
As mentioned above, while the optimal approach to watermarking considers sequence-to-sequence schemes, due to the autoregressive nature of token generation in LLMs most popular schemes focus on token level strategies \citet{kirchenbauer2023watermark,aaronson2023watermark,he2024universally,dathathri2024scalable}.
As a first step towards token-level watermarking of sequences, we provide an extensive analysis of the one-shot setting.
We discuss the extension to a token-level scheme in the sequential case in Section \ref{sec:sequential}.

\subsubsection{Problem Setting}

We consider a hypothesis test using the private side information setting and textual quality of the model as the ability of an external observer to detect the watermark without access to the side information. Formally, let $Q_X$ be the LLM distribution over some finite vocabulary of $|\cX|=m$ tokens. 
We consider \textit{Alice} (the watermarker), whose goal is to convey a single token to \textit{Bob} (the detector), which, in turn, tries to detect whether the token is watermarked or not.
Alice and Bob share some random side information\footnote{Side information often corresponds to a secret shared key; see, e.g., \cite{kuditipudi2023robust, zhao2024sok}.} $S\sim P_S$ with $|\cS|=k$.
Furthermore, we consider \textit{Charlie} (average observer), which tries to detect the existence of the watermark but does not have access to the side information.
The setting is depicted in Figure \ref{fig:operational_problem}.

On Alice's end, the watermark design boils down to the construction of the conditional distribution $Q_{X|S}$. We consider a Bayesian setting, in which Alice transmits a token according to a uniform prior: 
\begin{equation}\label{eq:sample_a}
    A = 
    \begin{cases}
        X\sim Q_X &\mbox{if }C=0,\\
        \tilde{X}\sim Q_{X|S} &\mbox{if }C=1.
    \end{cases}, \quad C\sim\mathsf{Ber}\left(\frac{1}{2}\right)
\end{equation}
where $C\indep(X,\tilde{X},S)$.
To detect the watermark, Bob performs the following hypothesis test
\begin{align*}
&H_0: A\sim Q_X\\
&H_1: A\sim Q_{X|S}.
\end{align*}
We assume that Charlie is aware of the watermarking mechanism but is not aware of the specific sample of $S$. Therefore, Charlie performs an hypothesis test with a corresponding alternative hypothesis, i.e.
\begin{align*}
&H_0: A\sim Q_X\\
&H_1: A\sim\bar{Q}_X,
\end{align*}
where $\bar{Q}_X\triangleq\mathbb{E}_{S}[ Q_{X|S}]$ is the watermark distribution averaged w.r.t. the side information $S$.

\subsubsection{A Detection-Perception Perspective}

Given the hypothesis test formulation, we recast the problem of watermarking as a trade-off between two measures: Bob's \textit{detection} and Charlie's \textit{perception} probabilities. Motivated by recent advances in lossy source-coding \citet{blau2019rethinking,theis2021coding,chen2022rate}, we adopt the notion of perceptual qualities of the data, which is quantified through a discrepancy measure between the two distributions, e.g. $f$-divergences, rather than a metric calculated directly on the random variables.

We define two fundamental metrics that capture the trade-off between detection capability for Bob and imperceptibility for Charlie. For Bob's detection capability, we weigh true negative (TN) detections with prior $\pi_0$ and true positive (TP) detections with prior $\pi_1=1-\pi_0$. The tests are defined as follows:
\begin{definition}[Watermark Tests and Error Probabilities]
    A watermarking scheme comprises of a detection test $g_d: \mathcal{X} \times \mathcal{S} \to \{0,1\}$, such that for $(A,S)\in\cX\times\cS$, we respectively define the detection probability with prior $\pi=(\pi_0,\pi_1)$ as
    \begin{align*}
    \rd \triangleq \mathbb{E}_{\pi}\left[\Pr(g_d(S,A)=C)\right].
    \end{align*}
    Perception probability $\rp$ is similarly defined with a test $g_c: \mathcal{X} \to \{0,1\}$ and a uniform prior $\pi_0=1/2$.
\end{definition}

Optimally, we aim to optimize detection $\rd$ while lowering $\rp$, which indicate Charlie's low perception of the watermark. The metrics detection and perception are formalized next.

\subsubsection{Characterizing Optimal Trade-off}

Following the Neyman-Pearson Lemma \citet{lehmann1986testing}, the likelihood ratio gives the optimal test statistic, and ($\rd$,$\rp$) have a simple form in terms of $\mathsf{E}_\gamma$ (or hockey-stick) divergence. The next proposition is a direct result of the well-known connection between $\mathsf{E}_\gamma$ and hypothesis testing; see, e.g., \citet{polyanskiy2010channel,polyanskiy2010channel_,liu2016e}.

\begin{prop}\label{prop:tv_prob}
Fix $(P_S,Q_X,\tilde{Q}_{X|S})$ and error prior $\pi$. Let $\gamma = \frac{\pi_1}{\pi_0}$. Using the LRT, the optimal detection and perception probabilities are given by 
\begin{align}
\rd &= \pi_1 + \pi_0 \mathsf{E}_\gamma\left(\tilde{Q}_{X|S}P_S, Q_XP_S\right), \\
    \rp &= \frac{1}{2} + \frac{1}{2}\TV\left(\tilde{Q}_X, Q_X\right).
\end{align}
\end{prop}

\begin{remark} The $\mathsf{E}_\gamma$ divergence characterizes the error of hypothesis tests with specified priors on TP and TN rates. 
It can be defined as\footnote{Some works include a residual term $(1-\gamma)_+$ \citet{asoodeh2020contraction}, which we omit for convenience as it does not affect the optimization problem.} in \citet{liu2016e}
$$\mathsf{E}_{\gamma}(P,Q) \triangleq \max_{\calA} [P(\calA)-\gamma Q(\calA)],$$ where $\cA$ are rejection regions, $P(\calA)$ and $Q(\calA)$ are $1-$TN rate and TP rate, respectively.
When. $\pi_0=\pi_1$ and $\gamma = 1$, detection probability boils down to the total variation (TV) distance, in which case, we have $\rd = \frac{1}{2} + \frac{1}{2}\TV(\tilde{Q}_{X|S},Q_{X}|P_S)$, where $\TV(\tilde{Q}_{X|S}P_S,Q_{X}P_S)=\TV(\tilde{Q}_{X|S},Q_{X}|P_S)$.
\end{remark}

Our hypothesis testing framework employs priors $\pi_0$ and $\pi_1$ to explicitly weight the importance of different error types in the detection process.  Setting $\pi_0 = \pi_1 = \frac{1}{2}$ gives equal importance to both errors, whereas asymmetric values prioritize either minimizing false positives (incorrectly flagging human content as AI-generated) or false negatives (failing to detect AI-generated content). This Bayesian framework provides a principled approach to designing watermark schemes with detection rates optimized for specific operational requirements, where the relative costs of different error types may vary significantly across applications.

Due to Jensen's inequality, for any fixed $(P_S,\tilde{Q}_{X|S})$, we have $\rp \leq \rd$, i.e., Bob's access to the shared side information allows for a potentially higher detection probability.
% The expression for Bob's detection probability shows that his access to side information allows him to achieve potentially better detection performance compared to Charlie, as the conditional total variation distance can be larger than the unconditional one -- a consequence of data processing inequality for the total variation where the process is the marginalization over $S$. This consequence can be formalized into the following lemma:
% \begin{lem}\label{lemma:rp_ub_rd}
%     The optimal perception rate is upper bounded by the optimal detection rate, i.e., 
%     $$
%     \rp(\delta, \pBm) \leq \rd(\delta, \pBm)
%     $$
% \end{lem}
Generally, for any perception constraint $\alpha_p\in[1/2,1]$, the optimal detection probability is given by the solution to the following optimization:
\begin{equation}\label{eq:curve_opt_iid}
    \sup_{\tilde{Q}_{X|S}}{\mathsf{E}_\gamma\left(\tilde{Q}_{X|S}, Q_X|P_S\right)},\quad \textrm{s.t.}\quad  \TV\left(\tilde{Q}_{X},Q_{X}\right)\leq \alpha_p.
\end{equation}
We are interested in characterizing the $(\rd,\rp)$ trade-off region, which amounts to solving \eqref{eq:curve_opt_iid} as a function of $\alpha_p$.

Note that \eqref{eq:curve_opt_iid} is a non-convex optimization problem.
However, in what follows, we characterize the several corner points of the optimal curve (i.e., $\rp=0.5$), which, in turn, gives insight into the structure of the $(\rd,\rp)$ region within the box $[\frac{1}{2},1]^2$.

We provide a complete characterization of the fundamental limits of detection probability under zero perception (where $\tilde{Q}_X=Q_X$). The following result establishes tight bounds on the optimal detection probability in this regime
\begin{thm}[Zero perception bounds]\label{thm:opt_cornerpoints}
     Fix $Q_X$ and let $P_S$ be uniform over $\cS$, $|\cS|\leq|\cX|$ and let $\pi_1=\frac{1}{2}$. Then, for $\rp=\frac{1}{2}$, we have
    \begin{equation}
    \label{eq:tv_opt_both}
    \frac{1}{2} \leq \sup_{\tilde{Q}_{X|S}}\rd \leq \max\left(\frac{1}{2},1-\frac{\gamma}{2k}\right).
    \end{equation}

\end{thm}
\noindent The upper bound emerges from jointly optimizing over both the coupling $\tilde{Q}_{X|S}$ and $Q_X$. This optimization reduces to a convex problem over the probability simplex, which we recast as counting the optimally assigning elements of $\cX$. The lower bound is achieved when $Q_X$ is a singleton.

% The upper bound is obtained by taking the supremum w.r.t. both $\tilde{Q}_{X|S}$ and the LLM distribution $Q_X$. If follows by showing we result with a convex function over the simplex, which then boils down to the problem of counting optimal assignments of elements of $\cX$. The lower bound is obtained when $Q_X$ is a singleton.

Beyond characterizing the zero-distortion endpoints, we derive an upper bound on the detection probability that holds across all perception levels.
The bound is given as follows: 
\begin{thm}[Uniform Detection Upper Bound]\label{thm:universal_ub}
Let $Q_{\mathsf{min}}\triangleq\min_{x\in\cX}Q_X(x)$. For any $\rp\geq 0$ we have $\rd\leq1-\frac{\gamma Q_{\mathsf{min}}}{2}$.
\end{thm}
This bound emerges from analyzing a simple strategy of replacing each token with the least likely symbol in the LLM's vocabulary.
The structure of the optimization \eqref{eq:curve_opt_iid} results in a nonconvex region, which generally lacks a closed form.
This non-convexity is demonstrated in our experimental results, see Section \ref{sec:CCexp}, where exact solvers are used to compute the trade-off region.
In light of this challenge, we will next derive a simple and tractable watermarking scheme.

\subsection{A One-Shot Watermarking Scheme}
\label{sec:cc_scheme}
While the optimal test that maximizes Bob's detection accuracy is the LRT, it is infeasible in practical scenarios where Bob is not assumed to have access to $Q_X$.
To make use of the shared side information, Bob and Alice look for a mechanism that couples $S$ with the token distribution.
This can be done by applying a map $f:\cX\to\cS$. 
Alice uses $(f(X),S)$ to construct a watermarked distribution, and Bob uses $(f(A),S)$ to detect its presence.
We note that a map $f$ creates a partition of $\cX$ into $\cS$ bins.
When $k=2$, this can be interpreted as a partition of $\cX$ into a rejection region and its complement.
We note that considering deterministic mappings is insufficient, as for $S\sim\mathsf{Unif}([1:k])$, the detection probability is $\frac{1}{k}$, independent of the choice of $(f,Q_X)$. Therefore, we introduce randomness into our partitioning approach by making the function $f$ stochastic rather than deterministic. Specifically, we define a randomized mapping that varies the way tokens are assigned to each partition based on additional random variables that both Alice and Bob can access.

\subsubsection{Optimal Randomized Partition -- Correlated Channel}

We randomize $f$ by introducing a set of $m$ $\cS$-valued random variables denoted $B^m$.
We assume that $B^m$ is publicly available to all parties and is therefore not considered a part of the private side information $S$.
% \dt{Having $B^m$, the mapping $f$ is simply the assignment of each $x\in\cX$ into a corresponding element of $B^m$, i.e., for any $x\in\cX$, $f(x,B^m)=B_x$.}
Our goal is therefore to couple the side information with the randomized mapping $f(X,B^m)$.
This boils down to finding a coupling of $Q_X$ and $S$ through the design of partition randomness $P_{B^m}$ and conditional distribution $ Q_{X|S}$.
We look for such $(P_{B^m}. Q_{X|S})$ that are optimal under the worst choice of token distribution $Q_X$ within a given class.
Our problem is therefore formally given by the following max-min expression
%\cl{check! if we want to keep max_$\tilde{Q}$, it should be inside max min}
\begin{align}\label{eq:maxmin}
    \rd^\star(\lambda) &\triangleq \max_{P_{B^m}} \min_{\substack{Q_X \in \Delta_m\\ \|Q_X\|_\infty \leq \lambda}} \mathbb{E}\left[\rd(Q_X,B^m)\right],
\end{align}
where $\|Q_X\|_\infty = \max_{x\in\cX}Q(x)$. As discussed in Section \ref{sec:main_contributions}, we consider the constraint $\{Q_X \in \Delta_m, \|Q_X\|_\infty \leq \lambda\}$ which enables a more comprehensive analysis by allowing us to adjust the parameter $\lambda$. This flexibility provides insights across various scenarios: smaller $\lambda$ values yield higher entropy token distributions with greater uncertainty, while larger $\lambda$ values produce more deterministic distributions with reduced uncertainty about the next token.

According to \eqref{eq:maxmin}, given a fixed pair $(P_{B^m},Q_X)$, we maximize $R_d(Q_X,B^m)$ by designing the coupling of $(f(X,B^m),S)$.
We consider the mapping of the form\footnote{We consider a vocabulary $\cX=[1:m]$, which can be thought of as the enumeration of the tokens.} $f(x,b^m)=b_x$ under which, the partition's probabilities are characterized by the distribution of the random variable $ Y\triangleq f(X,B^m)$.
To this end, we first solve the following optimization problem:
\begin{equation}\label{eq:ot_wm}
    \sup_{P_{S, Y}}\Pr(S= Y),\quad S\sim\mathsf{Unif}\left(\cS\right),  Y\sim P_{ Y}.
\end{equation}
This is a maximum coupling problem whose closed-form solution is given below. It is a direct consequence of the inf-representation of TV distance \cite{polyanskiy2025information}.
\begin{prop}\label{prop:optimal_coupling}
Let $S\sim\mathsf{Unif}[1:k]$ and $P_{ Y}=\{p_1,\dots,p_k\}\in\Delta_k$, $t = \TV(P_S, Y)$ and let $\Pi$ be the set of all couplings of $(P_S,P_{ Y})$.
Then, $\argmax_{\pi \in \Pi}\Pr(S= Y)$ is given by 
$$
\pi( Y=i, S=j) = 
\begin{cases}
    \min(\frac{1}{k},p_i),\qquad i=j,\\
    \frac{1}{t}(\frac{1}{k}-p_i)(p_j - \frac{1}{k}),\: (i\in A) \cap (j\in A^c),\\
    0, \qquad \text{otherwise},
\end{cases}
$$
where $A = \{i: p_i \geq \frac{1}{k}\}$, and $A^c = [k] \setminus A$.
\end{prop}
The resulting coupling can be thought of as a transition kernel that maps $P_{Y}$ to $P_S$ under maximum acceptance probability. When $k=2$, the optimal coupling boils down to a  binary asymmetric channel, known in information theory as the Z-channel \cite{CovThom06}.
% that is, a binary asymmetric channel used to model some asymmetric data storage problems where one input symbol can be flipped to the other, but not vice versa. 
That is, when $S=0$, the mapping always outputs $ Y=0$, but when $S=1$, the mapping may output either $ Y=1$ or $ Y=0$ with certain probabilities. This asymmetric structure is particularly effective for watermark detection because it creates a distinctive pattern that appears only in watermarked content. We therefore term this method as the correlated channel (CC) watermark.
We note that CC was previously considered, for example, in \cite{chao2024watermarking}.

\begin{algorithm}[!tb]
\caption{Correlated Channel Watermark (CC)}
\label{alg:correlated_Channel}
\begin{algorithmic}[1]
\Require LLM distribution $Q_X$, Side information $S$, shared randomness $B^m$.
\State \textbf{Alice:}
\State Generate $\tilde{Q}_{X|S,B^m}$ according to \eqref{eq:wm_dist_channel}
\State Flip a coin $C\sim\mathsf{Ber}(\frac{1}{2})$ and sample $A$ according to \eqref{eq:sample_a}.
\State \textbf{Bob:}
\State \textbf{if} {\( S=f(A,B^m) \)} --  Declare \textbf{Watermarked}
\State \textbf{else} -- Declare \textbf{Not watermarked}
\end{algorithmic}
\end{algorithm}

The CC scheme consists of the following steps: Both Alice and Bob observe $(s,b^m)$. Alice samples $C\sim\mathsf{Ber}(\frac{1}{2})$.
If $C=0$, she samples $a\sim Q_X$ and sends it. Otherwise, she samples and sends $a\sim\tilde{Q}_{X|S=s}$, which is given by the CC:
\begin{equation}\label{eq:wm_dist_channel}
    \tilde{Q}_{X|s,b^m}(x) = Q_X(x)\frac{P_{S| Y}(s|f(x,b^m))}{P_S(s)}.
\end{equation}
Bob performs the detection test by declaring that $a$ is watermarked if $s=f(a,b^m)$.
The complete list of steps is summarized in Algorithm \ref{alg:correlated_Channel}.
Note that by coupling $(P_{ Y},P_S)$, we result with a coupling of $(Q_X,P_S)$.
Consequently, we have
$Q_X = \mathbb{E}_{S}[ Q_{X|S}]=\bar{Q}_X$, which implies that the CC watermark has zero perception.

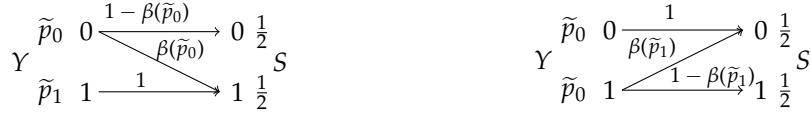
\begin{figure}[!t]
    \centering
    \begin{minipage}{0.4\textwidth}
    \centering
    \begin{tikzpicture}[scale=0.8]
      % Draw the Z-channel
      % For X axis
      \node at (-1.1,1) {\small$ Y$};
      \node at (0,1.5) {\small$0$};
      \node at (0,0.5) {\small$1$};
      \node at (-0.6,1.5) {\small$\tilde{p}_0$};
      \node at (-0.6,0.5) {\small$\tilde{p}_1$};
      % For Y axis
      \node at (3.2,1) {\small$S$};
      \node at (2.5,1.5) {\small$0$};
      \node at (2.5,0.5) {\small$1$};
      \node at (2.9,1.5) {\small$\frac{1}{2}$};
      \node at (2.9,0.5) {\small$\frac{1}{2}$};
    
      % Draw arrows representing transitions
      \draw[->] (0.2,1.5) -- (2.2,1.5);  % I_0 -> 0
      \draw[->] (0.2,1.5) -- (2.2,0.5);  % I_0 -> 1
      \draw[->] (0.2,0.5) -- (2.2,0.5);  % I_1 -> 1
      \node at (1,1.8) {\scriptsize$1-\beta(\tilde{p}_0)$}; 
      \node at (1.55,1.2) {\scriptsize$\beta(\tilde{p}_0)$}; 
      \node at (0.9,0.7) {\scriptsize$1$}; 
\end{tikzpicture}
    \end{minipage}
    \hspace{0.5cm}
    \begin{minipage}{0.4\textwidth}
    \centering
    \begin{tikzpicture}[scale=0.8]
      % Draw the Z-channel
      % For X axis
      \node at (-1.1,1) { \small$ Y$};
      \node at (0,1.5) {\small$0$};
      \node at (0,0.5) {\small$1$};
      \node at (-0.6,1.5) {\small$\tilde{p}_0$};
      \node at (-0.6,0.5) {\small$\tilde{p}_0$};
      % For Y axis
      \node at (3.2,1) {\small$S$};
      \node at (2.5,1.5) {\small$0$};
      \node at (2.5,0.5) {\small$1$};
      \node at (2.9,1.5) {\small$\frac{1}{2}$};
      \node at (2.9,0.5) {\small$\frac{1}{2}$};
    
      % Draw arrows representing transitions
      \draw[->] (0.2,1.5) -- (2.2,1.5);  % I_0 -> 0
      \draw[->] (0.2,0.5) -- (2.2,1.5);  % I_1 -> 0
      \draw[->] (0.2,0.5) -- (2.2,0.5);  % I_1 -> 1
      \node at (1,1.8) {\scriptsize$1$}; 
      \node at (0.7,1.2) {\scriptsize$\beta(\tilde{p}_1)$}; 
      \node at (1.7,0.71) {\scriptsize$1-\beta(\tilde{p}_1)$}; 
    \end{tikzpicture}
    % \caption*{(b)}
    \end{minipage}
    \caption{Optimal coupling between side information $S$ and random partition $ Y=f(X,B^m)$ for \( \tilde{p}_1 \leq 0.5 \) (\textbf{left}), \( \tilde{p}_0 \leq 0.5 \) (\textbf{right}), with $\beta(p)=\frac{2p-1}{2p}$.}
    \label{fig:Z-channel}
\end{figure}

\subsubsection{Theoretical Analysis of the CC scheme}

\noindent We provide a complete analysis of the CC scheme under $k=2$. Given the optimal coupling, we give a closed-form expression for $\rd$ in terms of the TV surrogate of mutual information in the resulting channel.
%\fc{Does the above only hold for $|\mathcal{S}|=2$?}
\begin{prop}\label{thm:detection_close_form}
    The CC watermark detection is given by 
    \begin{equation}
    \label{eq:Rd_cc}
        % \rd = \frac{1}{2}\left(1+\TV\left(P_{S},P_{S| Y}|P_{ Y}\right)\right) 
        % = 1-\frac{1}{2k}-\frac{1}{4}g(Q_X,B^m),
        \rd = \frac{1}{2}\left(1+\TV\left(P_{S},P_{S| Y}|P_{ Y}\right)\right) 
        = 1-\frac{1}{2k}-\frac{1}{2}\mathsf{TV}\left(P_{ Y},P_S\right).
    \end{equation}
\end{prop}
Proposition \ref{thm:detection_close_form} provides a closed-form characterization of Bob's detection probability as a function. Specifically, for $k=2$, we have $\rd = \frac{1}{2}(1+\tilde{p})$, where $\tilde{p}\triangleq\min\left(\tilde{p}_0,\tilde{p}_1\right)$.
This term is maximized when $ Y\sim\mathsf{Ber}(\frac{1}{2})$, with maximum value of $\frac{3}{4}$.
A consequence of Proposition \ref{thm:detection_close_form} is that we are interested in designing a partition that is as close as possible to $P_S$ as possible.
As $P_S$ is uniform over $\{1,\dots,k\}$, our aim is to obtain a uniform distribution, i.e., a balanced partition of the token vocabulary $\cX$, given the token distribution $Q_X$ and the partition randomness $P_{B^m}$.
\begin{remark}[Equivalence to the likelihood ratio test] When we consider the indicator test $\mathbf{1}\{f(x,b^m)=s\}$, the decision region obtained by the CC watermark is equivalent to the one attained by the LRT with threshold value of $\tau=1$. This follows from the observation that $\Pr[S|f(S,B^m)]\geq \frac{1}{2}$, if and only if $S=f(X,B^m)$. \label{thm:detection_test_optimal} 
\end{remark}
%\dt{prev 'thus' remark commented below}

Next, we discuss the design of randomness. Specifically, we analyze the dependence of the CC watermark detection probability on the distribution of $B^m$ and propose an optimal design of $P_{B^m}$.

\subsubsection{Optimizing the Partition}\label{sec:bm_design}

As seen in Equation \eqref{eq:Rd_cc}, the distribution of the resulting partition governs the detection power of the CC watermark.
The partition distribution is determined by the token distribution $Q_X$ and the distribution of $B^m$.
As $Q_X$ cannot be controlled by the watermark designer, we aim to characterize the class of distributions $P_{B^m}$ that maximizes $\rd$ under the worst-case adversarial distribution $Q_X$.
Due to the symmetry of the CC, we can restrict the optimization over permutation classes of $P_{B^m}$.
First, we show that the optimal distribution $P_{B^m}$ is permutation invariant. 

\begin{lemma}
\label{lem:permutation_invariance}
Let $F(P_{B^m})\triangleq \min_{\substack{Q_X\in \Delta_m\\ \|Q_X\|_\infty \leq \lambda}} \mathbb{E}_{P_{B^m}}\left[\rd(Q_X,B^m)\right]$. Let $P^\star_{B^m}$ be a distribution that maximizes $F(P_{B^m})$. Consider a permutation $\phi: \calS^m \to \calS^m$ and define $\tilde{P}_\phi(B^m) = P^\star_{B^m}(\phi \circ B^m)$. Then, $F(P^*_{B^m}) = F(\tilde{P}_{\phi})$.
\end{lemma}

Next, let $\calP_m = \{\calB_1, ..., \calB_K\}$ be the partition of $\calS^m$ into $K$ sets of sequences that are identical up to a permutation. We refer to each $\calB_i$ as a permutation class. We proceed to characterize the optimal mean detection probability $\rd^\star$ and the corresponding distribution $P^\star_{B^m}$.  %\dt{We don't end up using this term...}

\begin{thm}[Optimal max-min Detection]
\label{thm:optimal_maxmin_detection}
Let $|\calS|=k$ and $\calX = m$, and assume that $m$ is divisible by $k$. Given min-entropy constraint $\lambda \in [0,1]$, and let $t = \left\lfloor \frac{1}{\lambda} \right\rfloor$. The optimal minimax detection probability from Equation \ref{eq:maxmin} is given by:
\begin{align}
    R_d^*(\lambda)  = 1-\frac{1}{2k}-\frac{1}{4}\EE[g(Q_\lambda^*,B^m)],
\end{align}
where 
\begin{align*}
    \EE[g(Q_\lambda^*,B^m)]=k 
    \sum_{c=0}^t \frac{\binom{m/k}{c}\binom{m-m/k}{t-c}}{\binom{m}{t}}
    &\left( \left(\frac{ (m/k)-c}{m-t} \right)\left| c\lambda +(1-\lambda t) -\frac{1}{k} \right|\right. \\
    &+ \left. \left(1-\frac{ (m/k)-c}{m-t} \right)\left| c\lambda -\frac{1}{k} \right|\right).
\end{align*}
Furthermore, the optimal detection probability is achieved for $P^*_{B^m}$ corresponding to uniform sampling over the permutation class of the sequence with an equal number of each element. For $|\calS|=2$, $P_{B^m} = \mathsf{Unif}(B^\star)$, where $B^\star = \{b^m\in\{0,1\}^m|b^m\text{ has equal number of 1's and 0's}\}$. 
\end{thm}
Under additional assumptions, we can further simplify the optimal detection.
\begin{corollary}
    Under the setting of Theorem \ref{thm:optimal_maxmin_detection}, assume that $\lambda = \frac{1}{k}$. Then, we have 
\begin{align}
    R_d^*(\lambda)  = 1-\frac{1}{2k} - \frac{1}{2}\frac{\binom{(k-1)m/k}{k}}{\binom{m}{k}}.
\end{align}
Furthermore, if $k=2$ and $\lambda \in [\frac{1}{3},1]$ we have
\begin{equation}\label{eq:opt_b_thm}
    \rd^\star(\lambda) = 
    \begin{cases}
        \displaystyle\frac{3}{4}-\frac{m\lambda -1}{4(m-1)}, \quad~\mbox{for}\quad ~\frac{1}{2}\leq \lambda \leq 1\\
        \displaystyle\frac{3}{4} - \frac{m-2}{8(m-1)}, \quad~\mbox{for}\quad ~\frac{1}{3}\leq \lambda <\frac{1}{2}.
    \end{cases} 
\end{equation}
\end{corollary}

Here, we have characterized detection for the worst-case distributions $Q^\star_\lambda$, which lie at the extreme point of the feasible set --- probabilities with bounded inf norm $\|Q_X\|_\infty \leq \lambda$). For example, for $\lambda \in [0.5,1]$, the above minimax detection probability corresponds to token distributions with only two nonzero entries, i.e., $Q_X$ takes the form $[\lambda,1-\lambda,0,...,0]$;
for $\lambda \in [\frac{1}{3},\frac{1}{2}]$, the worst-case token distribution have 3 non-zero elements and has the form $[\lambda,\lambda, 1-2\lambda,0,...,0]$.
Furthermore, we note that due to Equation \eqref{eq:Rd_cc}, when $k=2$, $\rd$ is upper bounded by $\frac{3}{4}$.
Thus, the second term in \eqref{eq:opt_b_thm} serves as a penalty when considering the max-min setting.
Notably, for $\lambda \in [0.5,1]$ and when $m$ is large, this penalty equals $\frac{\lambda}{4}$, which implies that the cost of considering worst-case token distributions is lower bounded by $\frac{1}{8}$.

\begin{figure}[tb]
    \centering
    \includegraphics[width=0.45\linewidth]{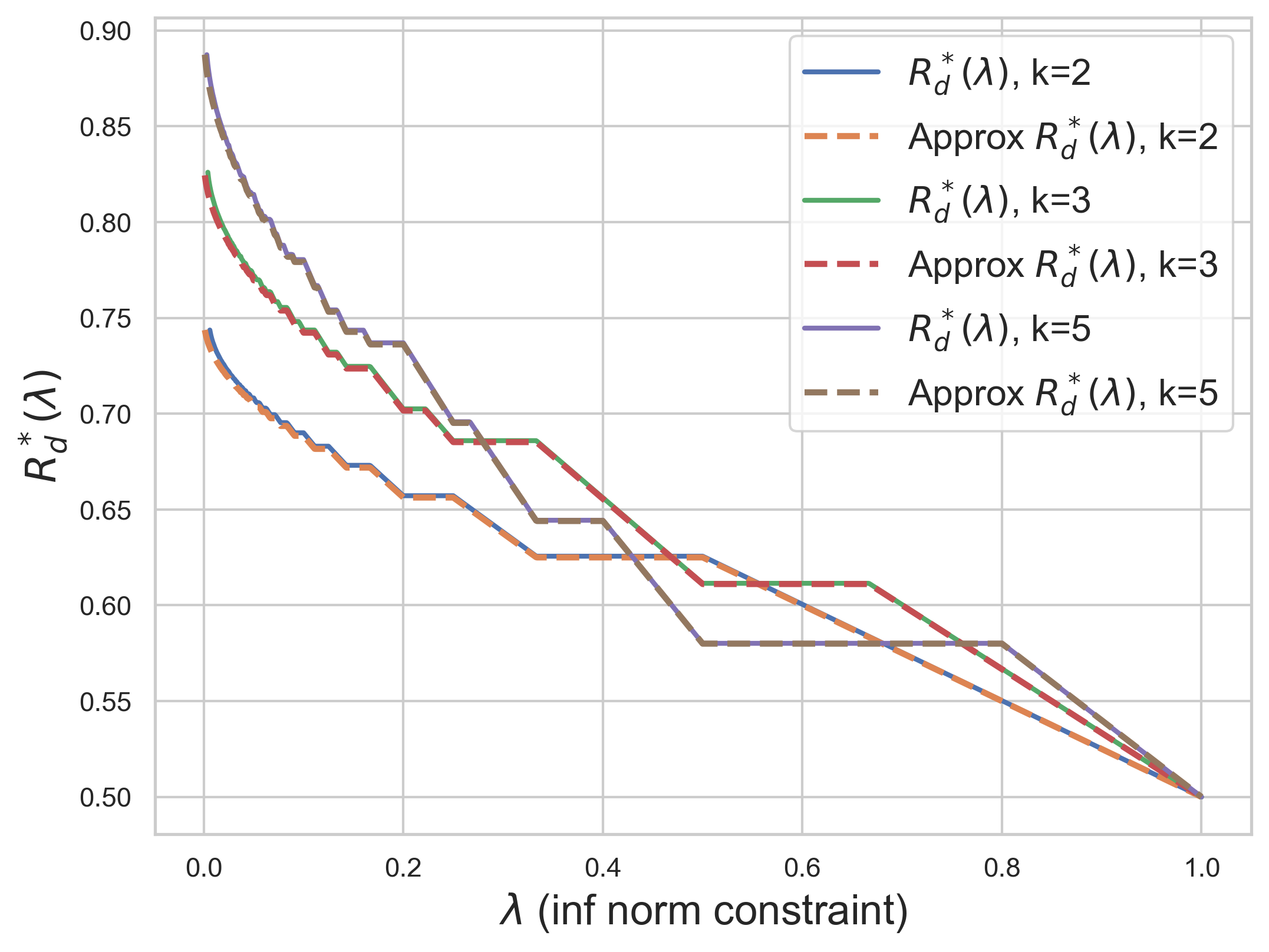}
    \caption{Optimal detection probability of CC in one-shot on the adversarial token distribution (Eq. \ref{eq:maxmin}) is plotted against the inf-norm constraint $\lambda$ (or equivalently, an entropy constraint) on $Q_X$$^3$.
    When $\lambda = 1$ (entropy $H(Q_X)=0$) , $Q_X$ is deterministic, and detection is random. As entropy of $Q_X$ grows (moves to smaller $\lambda$ values), single-token optimal detection probability reaches a maximum of around 0.75 for binary side information. If the side information one transmits contain a larger set of values, CC achieves a higher detection probability correspondingly. The actual detection rate (solid lines) and approximate solutions (dotted lines) overlap for large enough vocabulary size$^4$, and their exact forms are provided in Theorem \ref{thm:optimal_maxmin_detection} and \ref{thm:approx_maxmin_detection}.}
    \label{fig:Rd_star}
\end{figure}

In addition to characterizing the minimax detection rate, Theorem \ref{thm:optimal_maxmin_detection} shows that the optimal sampling strategy for token partition $B^m$ is to sample uniformly from a collection of sets with an equal number of each element in $k$. 
Next, we show that we can adopt a much simpler sampling strategy, sampling i.i.d. Bernoulli variables with probability $\frac{1}{k}$ and arrive at a near-optimal detection probability.
In Figure \ref{fig:Rd_star}, we plot the probability of detection of both sampling strategies and show that the Bernoulli sampling strategy results in negligible approximation error.
To motivate i.i.d. Bernoulli sampling, we start with an alternative view of the optimal sampling strategy in Theorem \ref{thm:optimal_maxmin_detection}. Sampling a $b^m$ uniformly over $\calB^*$ --- containing sequences with equal numbers of each element in $k$ --- can be equivalently defined as the following process: given $m$ elements with predefined proportions $[\frac{1}{k},...,\frac{1}{k}]$, sample $m$ times with replacement.
In the following theorem, we obtain an approximation of $R^\star_d$ for any $\lambda$ by sampling without replacement. We also show that, by applying de Finetti's theorem on finite exchangeable sequences\cite{diaconis1980finite}, the approximation error decays with $O(\frac{1}{m})$.

\footnotetext{For discrete probability $Q_X$, inf-norm and entropy are connected via $H(Q_X)\geq -\log\|Q_X\|_{\infty}$, and we have $\lambda = \|Q_X\|_{\infty}$.}
\addtocounter{footnote}{1}
\footnotetext{We take $m=100k$. Hence, existing LLMs with much larger vocabulary size would produce negligible approximation error.}

\begin{thm}[Approximation of Max-min Detection Rate]
\label{thm:approx_maxmin_detection}
    Given $|S|=k$, $|\calX| = m$, and the inf-norm constraint $\lambda \in [0,1]$. Let $t = \floor{\frac{1}{\lambda}}$, and $Y\sim Bin(t,\frac{1}{k})$ An approximation of the optimal minimax detection probability is given by: 
    \begin{align}
    \tilde{R}^\star_d(\lambda) = 1 - \frac{1}{2k}- 
    \frac{1}{4}\left[\sum_{c=0}^t \Pr[Y=c] \left(\left|(c-t)\lambda + (1-\frac{1}{k})\right| + (k-1)\left|c\lambda -\frac{1}{k}\right|\right)\right]
    \end{align}
    The approximation error decays as $O(\frac{1}{m})$. Specifically:
    \begin{align}
        \left|\tilde{R}^\star_d(\lambda) - R^\star_d(\lambda)\right| \leq \frac{2k\ceil{\frac{1}{\lambda}}}{m}
    \end{align}
\end{thm}

We plot the results of Theorem \ref{thm:optimal_maxmin_detection} and \ref{thm:approx_maxmin_detection} in Figure \ref{fig:Rd_star}. For all $\lambda$ and $k$ values, the approximated maxmin detection coincides with the closed-form $R^\star_d(\lambda)$. We choose $m=100*k$. The overlap between the actual and approximated $R^\star_d(\lambda)$ in the plot testifies our result that the approximation error decays with $m$. In practice, since LLMs have a much large vocabulary, where $m\approx 100,000$\cite{grattafiori2024llama}, the approximation error will be negligible.

\subsection{Sequential Watermarking}
\label{sec:sequential}
While this chapter focused on a single-shot analysis of token distribution watermarking, general text generation involves sequential prediction of long token sequences.
A common approach involves applying a token-level watermarking of the next token distribution and designing token-level test statistics.
This approach was shown to benefit from favorable performance \cite{kirchenbauer2023watermark,aaronson2023watermark}, albeit being theoretically suboptimal\cite{he2024universally}.
We note that our one-shot method readily extends to a sequential token-level scheme as we can treat each step as a one-shot problem, and considering an average test $\frac{1}{n}\sum \mathbf{1}[f_i(A_i,B^m_i) = S_i]$ which we them compare with some threshold $\tau\in[0,1]$.
We leave the theoretical analysis of the token-level extension of our scheme to future work, while showing empirical results in Section \ref{sec:CCexp}.
In the simplified case when $X^n$ are i.i.d., we provide the following bounds on the detection probability (a related result was given in \cite{chao2024watermarking} bounding mismatch proportion using entropy):
\begin{prop}
\label{prop:sequential_Rd_bound}
    Let $Q^n=Q^{\otimes n}_X$ be the an i.i.d. token distribution, let $S^n\sim P^{\otimes n}_S$ and apply the one-shot CC on each step $i\in[1:n]$, then
    \begin{equation*}
        1-2^{-\left(\frac{n}{2}+1\right)}\left(g(\tilde{p})\right)^n \leq \rd \leq \frac{1}{2}\left(1 + \sqrt{1-\left(\frac{\left(g(\tilde{p})\right)^2}{2}\right)^n}\right),
    \end{equation*}
    where $\tilde{p}=\min(\tilde{p}_0,\tilde{p}_1)$ is similarly defined as in the on-shot case, and $g(p) \triangleq p + \sqrt{\frac{1-p}{2}}\left(1+\sqrt{1-2p}\right), p\in[0,0.5].$
\end{prop}
The proof utilizes bounds on TV in terms of the Hellinger distance, which benefits from a tensorization. %\dt{Add proof to apdx}

\subsection{Experimental Results}\label{sec:CCexp}

We numerically evaluate the CC watermark on synthetic distributions with various inf-norm constraints. 
We compareCC with the solution of an exact GUROBI-based numerical solution \cite{gurobi} of Eq. \eqref{eq:curve_opt_iid} and the red/green watermark \cite{kirchenbauer2023watermark}.
\footnote{Full implementation details and code are given in \url{https://github.com/Carol-Long/CC_Watermark}.}

\subsubsection{One-Shot Performance Analysis}
\begin{figure}[tb]
  \centering
  % First subfigure
  \begin{subfigure}[b]{0.4\textwidth}
    \centering
    \includegraphics[width=\textwidth]{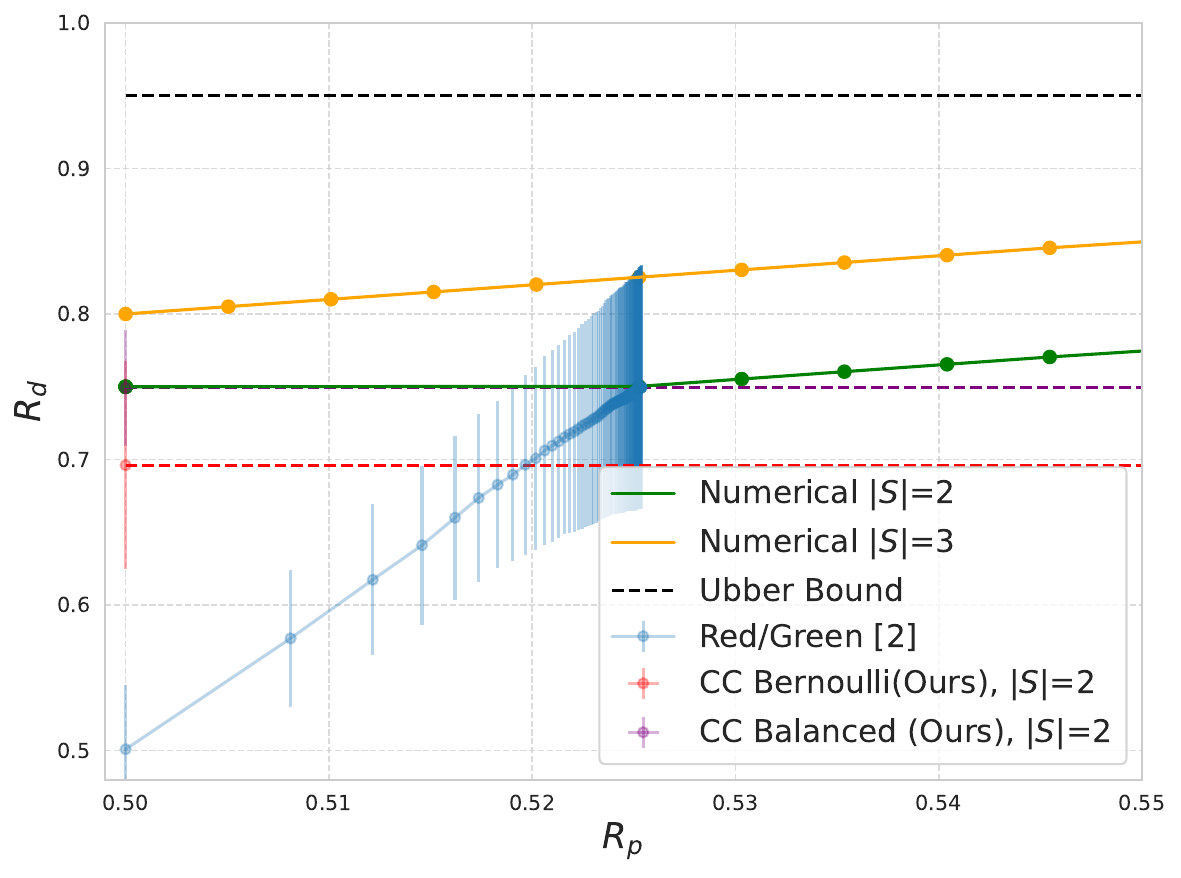}
    \caption{$Q_X=\mathsf{Unif}[1:m]$}
    \label{fig:exp_results_unif}
  \end{subfigure}
  \hspace{1cm}
  % Second subfigure
  \begin{subfigure}[b]{0.4\textwidth}
    \centering
    \includegraphics[width=\textwidth]{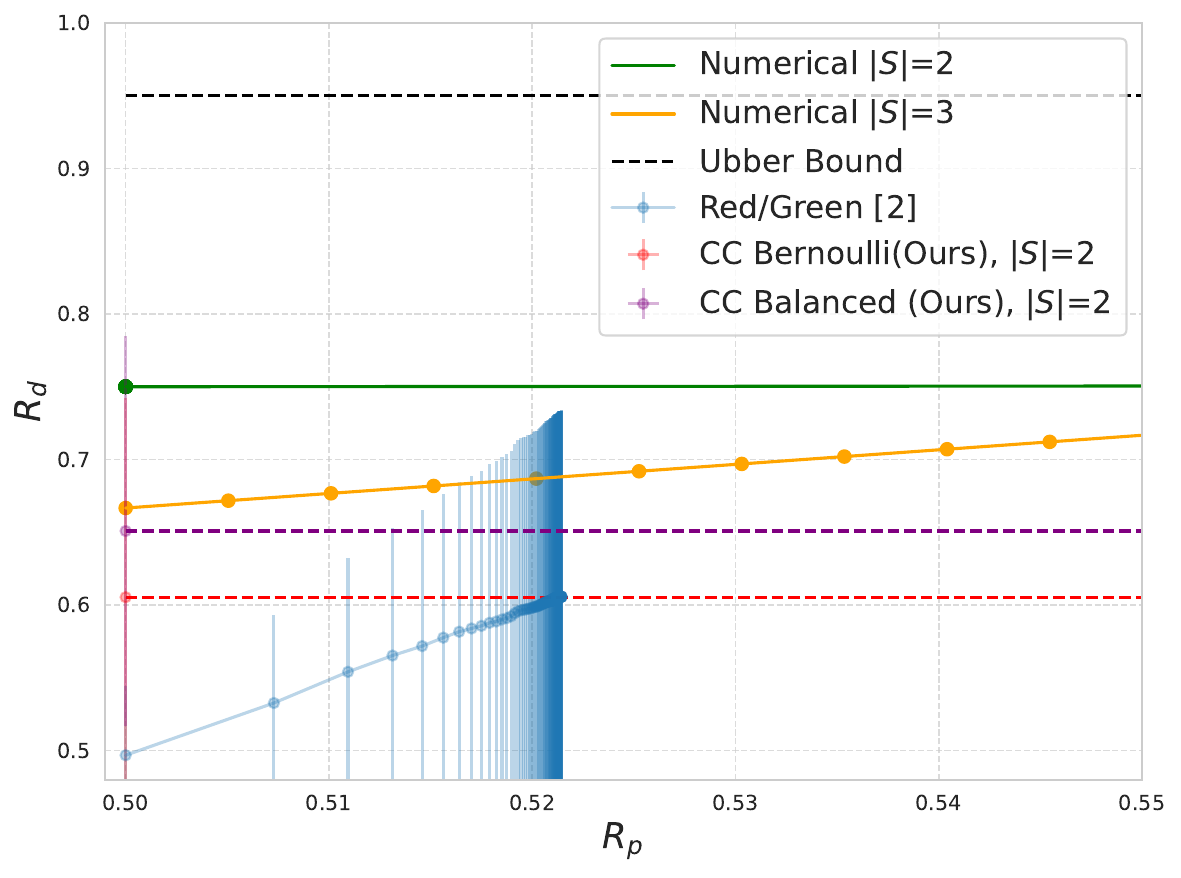}
    \caption{$Q_X=[0.5,0.5,0,\dots]$}
    \label{fig:subfig2}
  \end{subfigure}
  \caption{One-shot watermark detection results on $Q_X=\mathsf{Unif}(\cX)$. For $\alpha_p=0$, CC achieves a detection probability of $0.75$ and $0.7$ with balanced and Bernoulli partitions, respectively. CC Balanced achieves the optimal detection (Eq. \ref{eq:curve_opt_iid} with $\gamma = 1$ and $|\cS|=2$). Standard deviations plotted as two-sided bars.}
  %\label{fig:rd_rp}
\end{figure}

\begin{figure}[htbp]
  \centering
  % First subfigure
  \begin{subfigure}[b]{0.4\textwidth}
    \centering
    \includegraphics[width=\textwidth]{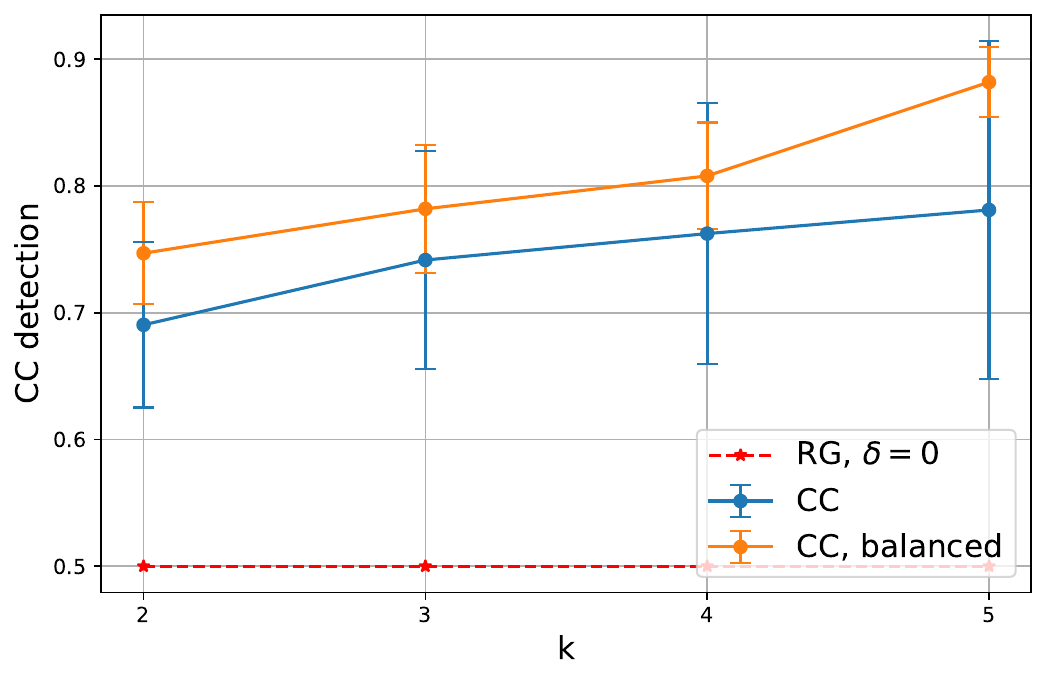}
    \caption{$m=10$}
    \label{fig:subfig11}
  \end{subfigure}
  \hspace{1cm}
  % Second subfigure
  \begin{subfigure}[b]{0.4\textwidth}
    \centering
    \includegraphics[width=\textwidth]{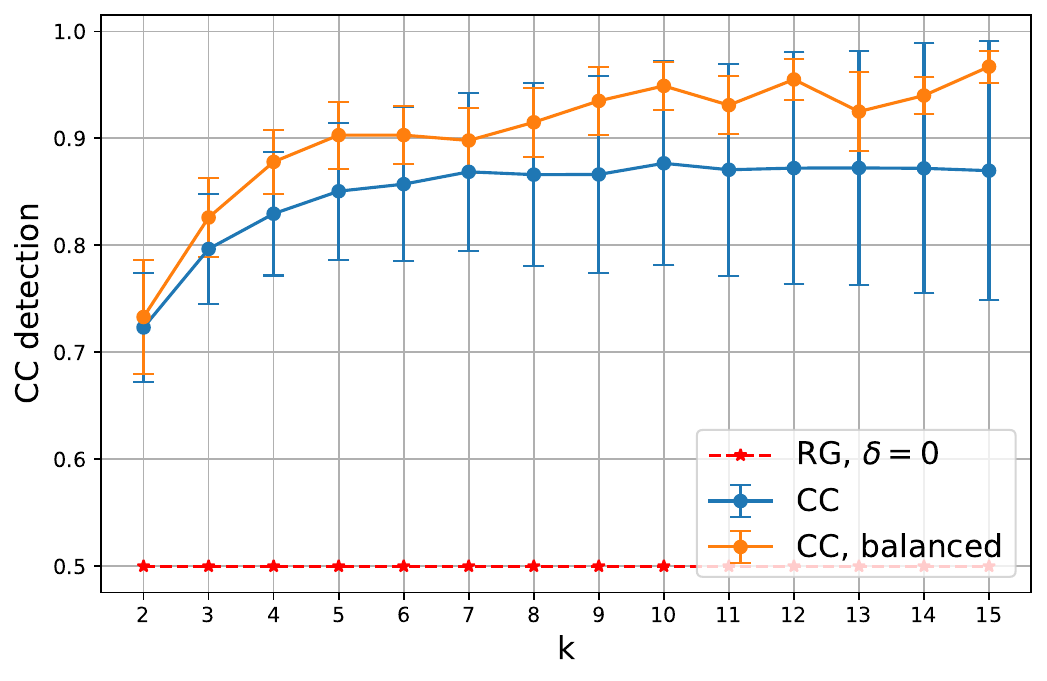}
    \caption{$m=60$}
    \label{fig:subfig22}
  \end{subfigure}
  \caption{Detection probability vs. $k$ for two values of $m$ and a uniform token distribution $Q_X$.}
  \label{fig:rd_vs_k}
\end{figure}

\textbf{Detection-Perception Tradeoff: }
We present the $(\rd,\rp)$ trade-off region for the one-shot watermarking setting. 
We consider the worst-case distribution within $\{Q_x, \|Q_X\|_\infty\leq\lambda\}$.
When $\lambda=\frac{1}{m}$, the resulting distribution is simply the uniform distribution over $\cX$ and when $\lambda\geq\frac{1}{2}$ it is given by a distribution with two nonzero entries valued $(\lambda, 1-\lambda)$. This distribution is representative of a next-token distribution in the low entropy regime (highly predictable next token).
As seen in Figure \ref{fig:exp_results_unif}, for uniform $Q_X$, when we apply the CC scheme with $P_{B^m}$ sampled over balanced partitions, we obtain a gain of $\approx 0.07$ over sampling $B^m\stackrel{i.i.d.}{\sim}\mathsf{Ber}(\frac{1}{2})$, meeting the upper bound from \eqref{eq:curve_opt_iid}.
In contrast, the red-green detection coincides with ours in the limit of $\delta\to\infty$, intersecting with the suboptimal i.i.d. Bernoulli sampling method at $\delta\approx7.6$.
When $\delta=\frac{1}{2}$ we observe a decrease in the gain of sampling from the balanced partition sets.
\textbf{Effect of $k$:}
Next, we analyze the effect of the side information alphabet size on the CC scheme performance.
We present a plot for $m=10$ which serves as an extension of the performance we present in Figure \ref{fig:exp_results_unif} and a plot for $m=60$, which allows us to further understand the effect.
As seen in Figure \ref{fig:rd_vs_k}, as $k$ increases, the detection rate of the CC watermark increases. However, the gain from increasing $k$ decreases as $k$ grows (or alternatively, as the ratio $m/k$ decreases). 
Furthermore, we note that the performance depends on the divisibility of $m$ by $k$; when $m/k$ is not an integer, we experience a degradation of performance. This follows from the inability to construct equally sized partitions of $\cX$, which, in turn, decreases the probability to result with a balanced partition.

\subsubsection{Sequential Watermarking}
\begin{figure}[htbp]
  \centering
  % First subfigure
  \begin{subfigure}[b]{0.35\textwidth}
    \centering
    \includegraphics[width=\textwidth]{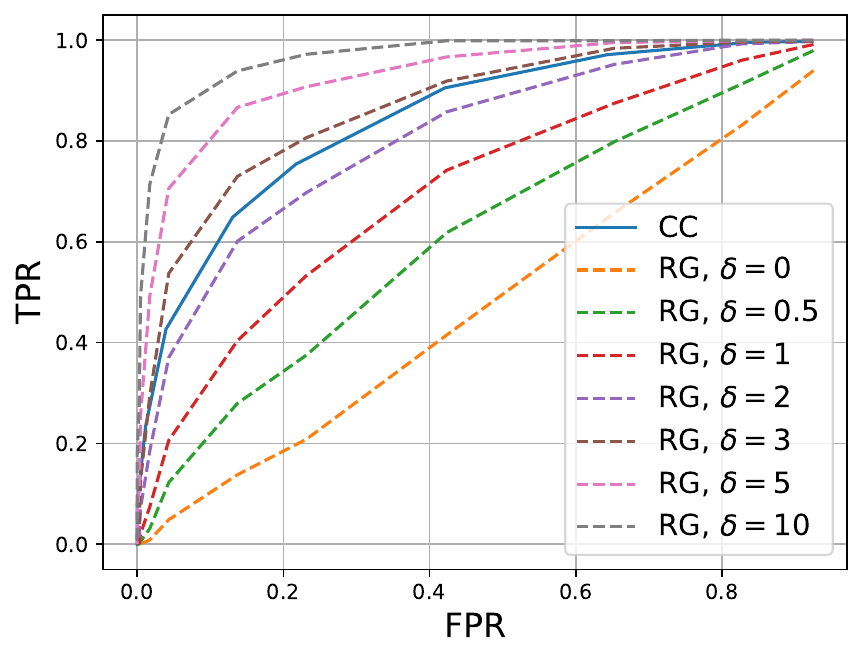}
    \caption{$\lambda=0.8$}
    \label{fig:subfig1_}
  \end{subfigure}
  \hspace{1cm}
  % Second subfigure
  \begin{subfigure}[b]{0.35\textwidth}
    \centering
    \includegraphics[width=\textwidth]{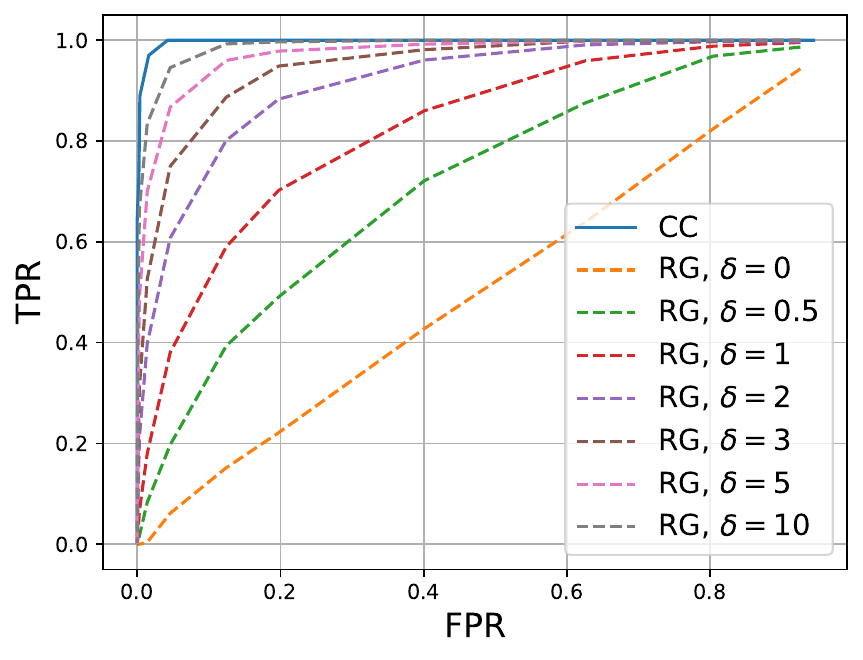}
    \caption{$\lambda=0.5$}
    \label{fig:subfig2_}
  \end{subfigure}
  \caption{ROC of the sequence-level watermarking scheme. We compare the red-green method \cite{kirchenbauer2023watermark} with the CC scheme (Section \ref{sec:cc_scheme}). We consider a range of $\delta$. An increase of $\delta$ increases detection, at the expense of higher perception (lower textual quality), while the CC method has fixed zero perception.}
  \label{fig:roc_deltas}
\end{figure}

We now present the performance of the CC watermark on a sequence level scheme. We present preliminary results on synthetically generated data, with the purpose of demonstrating the applicability of our method to a sequence-level test.
To that end, we consider the generation of $n$ tokens $A^n$, which are generated from a sequence of tokens $X^n\stackrel{i.i.d.}{\sim}Q_X$ using from $n$ i.i.d. samples of side information $s^n$ and randomness $(B^{m}(i))_{i=1}^n$.
We apply the token-level watermarking scheme to each element $X_i$ to generate $A_i$ and apply the following sequence-level threshold test
$$
r(A^n,S^n) = \left\{\frac{1}{n}\sum_{i=1}^n \mathbf{1}\left(f(A_i,B^m(i)) = S_i\right) \geq \tau\right\}
$$
for some threshold $\tau\in[0,1]$.
To understand the performance of the proposed sequence-level generalization, we analyze the ROC of the results scheme.
In out experiment, we consider $k=2$, $m=20$ and a sequence of $n=50$ tokens.
Figure \ref{fig:roc_deltas} compares the ROC of the CC scheme (sampling from balanced sets) with the red-green scheme for a range of $\delta$ values. 
We note that, while the CC method is perceptionless, it results in a better ROC than the red-green method.
Specifically, for $\lambda=0.5$, the CC method demonstrated better detection than the red-green method for the considered range of $\delta$ values. However, when $\lambda=0.8$, i.e., when the distribution is spikier, the red-green method with higher $\delta$ values result in a better ROC than the CC method, but at the cost of nonzero perception.

\begin{figure}[htbp]
  \centering
  % First subfigure
  \begin{subfigure}[b]{0.35\textwidth}
    \centering
    \includegraphics[width=\textwidth]{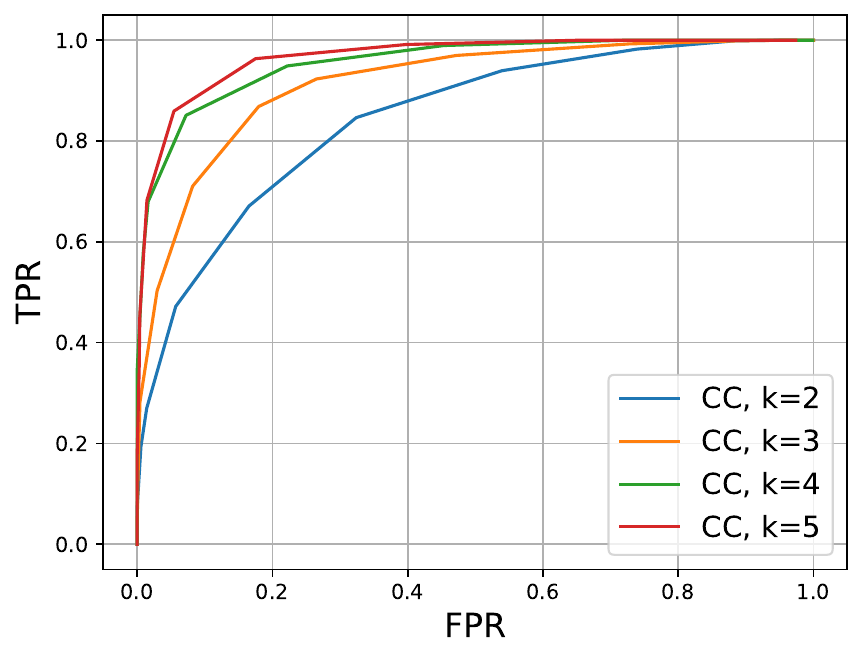}
    \caption{$Q_X$ has two nonzero values of $Q_1=0.8$ and $Q_2=0.2$.}
    \label{fig_}
  \end{subfigure}
  \hspace{1cm}
  % Second subfigure
  \begin{subfigure}[b]{0.35\textwidth}
    \centering
    \includegraphics[width=\textwidth]{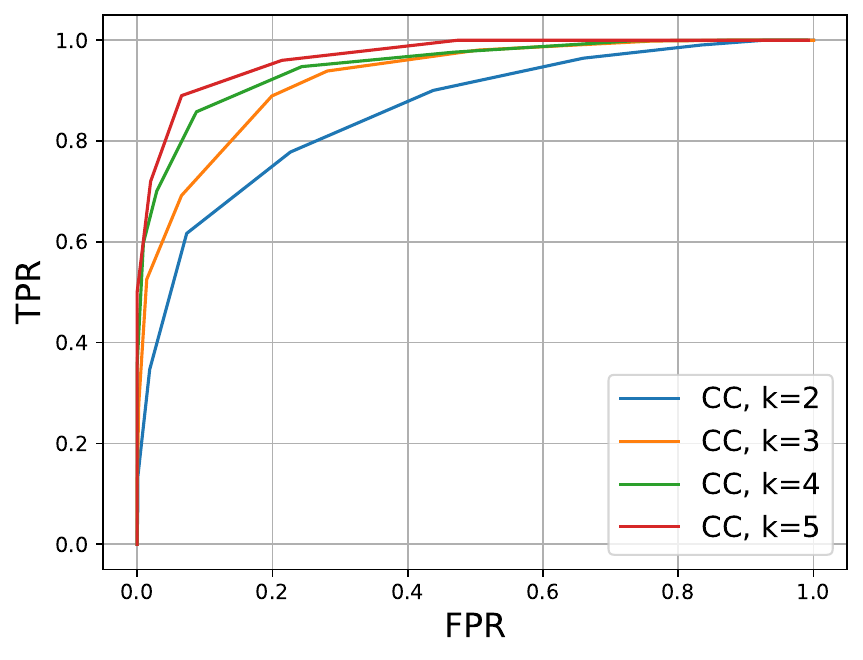}
    \caption{$Q_X$ has a spike of $\lambda=0.8$ and is uniform on the rest of its entries.}
    \label{fig:subfig22_}
  \end{subfigure}
  \caption{ROC of the sequence-level watermarking scheme under CC method for a range of $k$ values. }
  \label{fig:roc_vs_k}
\end{figure}

Finally, we analyze the effect of $k$ on performance in the sequential setting by observing the ROC for a range of $k$ values. 
Specifically, we consider $m=20$ and apply the sequential generalization of the CC watermark for $k\in\{2,3,4,5\}$.
We consider two distributions within the bounded infinity norm set with $\lambda=0.8$.
As can be seen in Figure \ref{fig:roc_vs_k}, as $k$ increases, the ROC improves. 

\subsection{Conclusion}

This section presents a rigorous analysis of text watermarking in a one-shot setting through the lens of hypothesis testing with side information. 
We analyze the fundamental trade-off between watermark detection power and distortion in generated textual quality.
A key insight of our approach is that effective watermark design hinges on generating a coupling between the side information shared with the watermark detector and a random partition of the LLM vocabulary.
We develop a perfect perception watermarking scheme -- the Correlated Channel Watermark (CC).
Our analysis identifies the optimal coupling and randomization strategy under the worst-case LLM next-token distribution that satisfies a min-entropy constraint. 
Under the proposed scheme, we derive a closed-form expression of the resulting detection rate, quantifying the cost in a max-min sense.
The CC scheme offers a framework that can potentially accommodate additional objectives of LLM watermarking, such as robustness against adversarial manipulations and embedding capacity.
Additionally, we envision future work implementing the scheme for sequential watermarking and extending it to the positive-perception regime, where minor adjustments to token probabilities are permitted in exchange for superior detection.

\section{\heavywater{} and \simplexwater{}: Distortion-free LLM Watermarks for Low-Entropy Distributions}

\subsection{Introduction}
Watermarking large language models (LLMs) consists of embedding a signal into the text generation process that allows reliable detection of machine-generated text. Over the past two years, there have been increasing calls for LLM watermarking by both policymakers and industry\cite{ncsl_ai_legislation_2024,rijsbosch2025adoption,eu_ai_act_2024}. Watermarks enable authentication of text provenance \cite{chandra2024reducing}, promote trust in AI systems \cite{rand2024watermarking}, and can address copyright and plagiarism issues \cite{sander2024watermarking,panaitescu2025can}. Ideally, watermarks should be detectable directly from text without access to the underlying LLM.
Watermarks should also be tamper-resistant -- i.e., robust to minor edits or paraphrasing of watermarked text \cite{kirchenbauer2023reliability} -- and incur little degradation in text quality. 

LLMs are watermarked by changing their next-token distributions according to random side information (see Fig. \ref{fig:system} for a visualization). 
Side information is typically generated  by hashing previous tokens and secret keys, then using the hash to seed a random number generator to produce a sample $s$ \cite{kirchenbauer2023reliability,dathathri2024scalable,zhaoprovable}.
The side information $s$ is then used to change the  distribution from which the next token $x$ is sampled. 
Watermarks are detected by mapping a token $x$ and corresponding side information $s$ to a score $f(x,s)$. 
By averaging scores across a sequence of tokens, we can perform statistical tests to decide if the text is watermarked. 
Watermarking is particularly challenging when the next-token distribution has low entropy, as is often the case in tasks such as code generation \cite{kuditipudi2023robust}.

\begin{figure}[!t]
    \centering
    \includegraphics[width=0.85\linewidth]{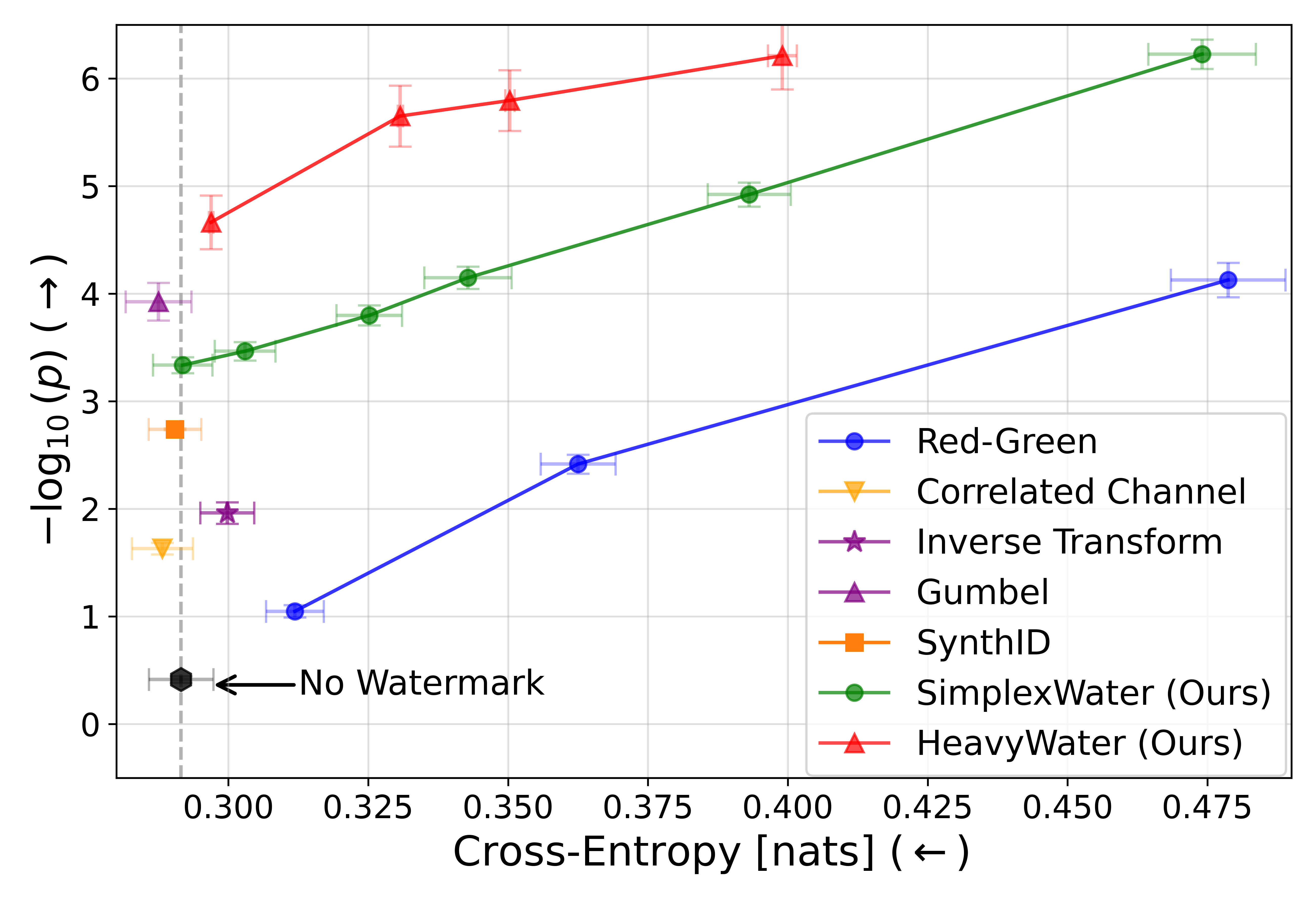}
    \caption{\heavywater{} and \simplexwater{} demonstrate favorable detection performance (measured by p-values) with minimal distortion to the base unwatermarked model (measured by Cross-Entropy). See Section \ref{sec:numerics} for details.}
    \label{fig:fig1}
\end{figure}

\begin{figure}[!t]
    \centering
    \includegraphics[trim={190pt 120pt 100pt 65pt}, clip, width=0.75\linewidth]{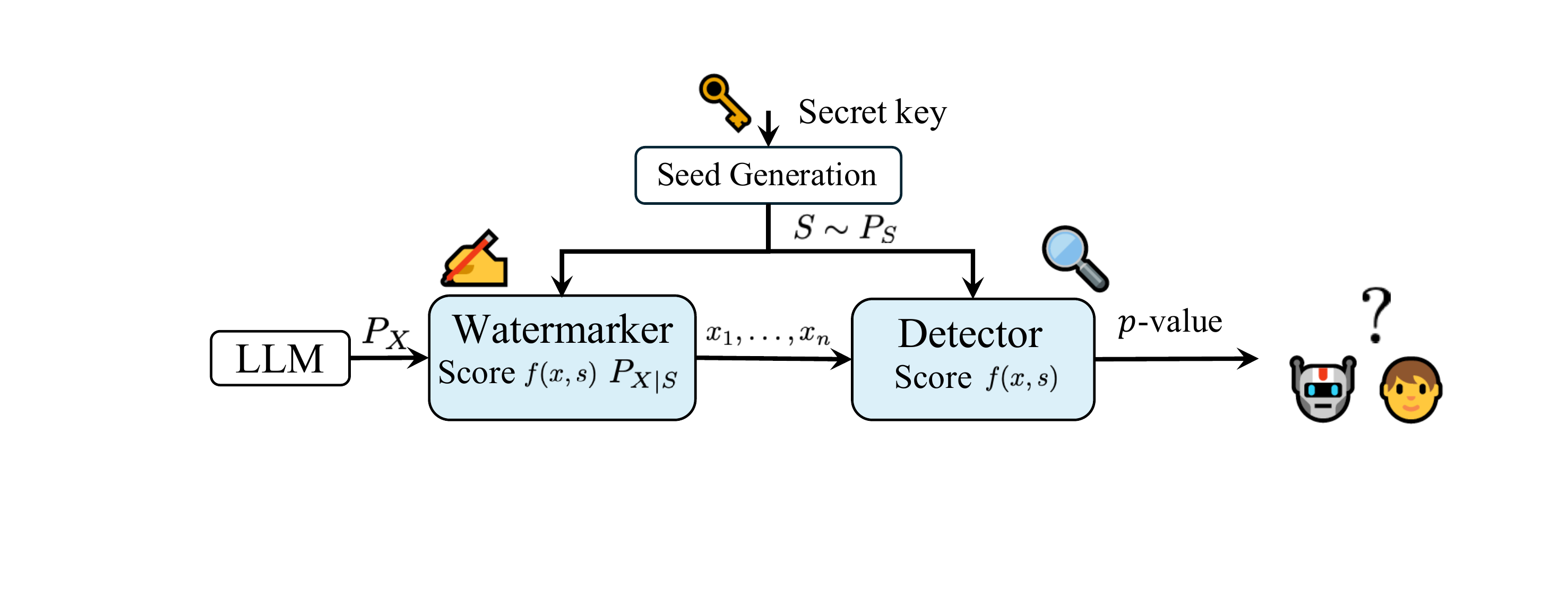}
    \vspace{-0.5em}
    \caption{Visualization of the components of watermarking design.}
    \label{fig:system}
    \vspace{-1em}
\end{figure}

\textbf{Our goal} \emph{is to design watermarks that optimally use  side information to maximize detection accuracy and minimize distortion of generated text.} 
Our starting point is a minimax optimization framework for watermark design. 
This framework allows us to jointly optimize how side information is embedded into next-token distributions and the score function used for watermark detection.
When restricting the optimization to binary-valued scores, we uncover  a surprising new connection between LLM watermarking and coding theory:  \textit{designing minimax-optimal watermarks is equivalent to constructing codes with large Hamming distance between codewords}.
We use this finding to design a new watermark called \simplexwater{} based on Simplex codes \cite{roth2006introduction}. In the low-entropy regime, 
\simplexwater{} is optimal across binary watermarks and empirically outperforms competing methods that use binary scores (cf. Red-Green \cite{kirchenbauer2023watermark}, and Correlated Channel \cite{long2025optimized} in Fig.~\ref{fig:fig1}).

Next, we relax the binary  restriction on the score function. We consider score functions whose outputs %the values of $f(x,s)$ 
are independently and identically (i.i.d.) drawn from a continuous distribution prior to the start of watermarking.
Interestingly, we show that the Gumbel watermark \cite{aaronson2023watermark} is a specific instantiation of our framework when scores are drawn from a Gumbel distribution.
We prove that, in the low-entropy regime, watermark detection depends on the weight of the tail of the distribution from which scores are sampled.
This observation leads to \heavywater{}, 
whose scores are drawn from a heavy-tailed distribution.
\heavywater{} outperforms competing state-of-the-art watermarks in terms of detection-distortion trade-offs (cf. Gumbel \cite{aaronson2023watermark} in Fig.~\ref{fig:fig1}).

In practice, \simplexwater{} and \heavywater{} are implemented by solving an optimal transport (OT) problem \cite{villan2006text} that maximizes the average score across all couplings between the side information and the next-token distributions. We efficiently solve the OT problem using Sinkhorn's algorithm  \cite{cuturi2013sinkhorn}. Since the OT preserves the average next-token distribution, both watermarks are \emph{distortion-free}, and a user without knowledge of the side information would not perceive an average change in generated text.   \simplexwater{} and \heavywater{} are also \emph{tunable}: we provide a simple scheme to further increase watermark detection accuracy at a small distortion cost by upweighting high-score tokens.

Our \textbf{main contributions} are:
\begin{itemize}[leftmargin=1em]
    % \item We introduce two new watermarks, \simplexwater{} and \heavywater{}, that are tunable, agnostic to side information generation, and applicable to any LLM.
    \item We introduce \simplexwater{}, a binary score watermark rooted in coding theory, and derive optimality guarantees across all binary-score watermarks. 
    \item We analyze watermarks that use randomly generated score functions drawn from a continuous distribution, and show that the Gumbel watermark \cite{aaronson2023watermark} is a special case of this construction.
    \item We prove that the detection power of watermarks with randomly generated score functions depends on the tail of the score function. This leads to \heavywater{}, a new watermark whose scores are randomly drawn from a heavy-tailed distribution.
    \item We demonstrate the favorable performance of \heavywater{} and \simplexwater{} across an array of models, datasets, and tasks, relative to state-of-the-art watermarks.
\end{itemize}

\subsubsection*{Related Works}
We discuss the related work most closely related to ours next, and provide a broader survey of the watermarking literature in Appendix \ref{apdx:related_work}.
The first LLM watermark was proposed in \cite{kirchenbauer2023watermark} -- referred here as the Red-Green watermark. This method partitions the token vocabulary into two lists, which are then used to reweigh the token distribution via exponential tilting. The Red-Green watermark was extended via different random side information generation schemes in \cite{kirchenbauer2023reliability}.
Watermarking has since been extensively studied, with more recent work introducing new sampling schemes to improve watermarking \cite{zhao2024permute} and multi-draft watermark generation methods \cite{giboulot2024watermax}. The use of threshold tests over average scores to detect watermarks -- a common paradigm in the literature, a design decision we make here -- was questioned in  \cite{fernandez2023three,li2024robust}. \cite{li2024robust}, in particular, shows that more sophisticated statistical tests can boost watermark detection. 
We note that our watermarks can be paired with multi-draft methods such as \cite{giboulot2024watermax} to boost detection, though we do not explore this here.

\textbf{Distortion-free Watermarking.} In watermarking, distortion quantifies how much a watermark distribution differs from the original LLM output and is a proxy for textual quality.
\cite{aaronson2023watermark} uses the Gumbel-max trick to design a distortion-free watermark. This scheme was relaxed to a soft reweigh of the next-token distribution \cite{fu2024gumbelsoft} following the idea from the Red-Green watermark. The Gumbel watermark achieves excellent performance in our benchmarks, though it is outperformed by \heavywater{}. 
Another popular distortion-free watermark is \cite{kuditipudi2023robust}, which uses inverse transform sampling. 
%We refer to it as the Inverse Transform Watermark in our experiments.  
SynthID \cite{dathathri2024scalable}, in turn, produces watermarked tokens via a strategy called tournament sampling. This watermark was extensively evaluated in real-world user tests, and we also select it as a competing benchmark. 
More recently, \cite{long2025optimized} proposed a binary-score watermark based on partitioning the token vocabulary into an arbitrary number of sets, followed by a simple binary test for watermark detection. The method in \cite{long2025optimized} is simple to implement and incurs little runtime overhead. However, it underperformed both \heavywater{} and \simplexwater{} in our benchmarks. Additional distortion-free watermarks include \cite{hetheoretically}, which combines a surrogate model with the Gumbel-max trick to boost watermark detection.

\textbf{Low-Entropy Watermarks.}
Several schemes were proposed to address the challenge of watermarking low-entropy distributions \cite{huang2023towards,kirchenbauer2023reliability}.
\cite{mao2024watermark} proposes a resampling method that is applied to adapt the Red-Green watermark to low-entropy distributions, while \cite{hou2024semstamp} turns to semantic sentence-level watermarks.
\cite{chang2024postmark} uses a logit-free method that injects the texts with words from an input-dependent set, and \cite{lu2024entropy} weights the tokens' scores according to their entropy. We share a similar design goal of optimizing the watermark for the low-entropy regime.

In practice, side information will not be perfectly random: it will depend on previous tokens and a secret shared key through various hashing schemes \cite{kirchenbauer2023watermark,yoo2023advancing}.
Hashing schemes range from simple $\mathsf{max}/\mathsf{min}$ operations to elaborate adaptive context windows \cite{dathathri2024scalable} and semantic hashing \cite{ren2023robust}.
We view side information generation as a separate yet no less important problem. Instead, we focus on \emph{how to optimally use} side information, and we develop a principled and theory-guided approach for watermark design.

\subsection{LLM Watermarking Framework}\label{framework}

In this section, we outline an optimization formulation for LLM watermarking. We denote random variables using capital letters (e.g., $X$ and $S$), their realizations by lower-case letters (e.g., $x$ and $s$), and their distributions by sub-scripted $P$ or $Q$ (e.g., $P_X$ and $P_S$). We consider an LLM with token vocabulary $\mathcal{X}=[1:m]\triangleq\{1,\dots,m\}$. During generation, the $t$-th token is drawn auto-regressively from a distribution $Q_{X_t|X_1,\dots,X_{t-1}}$ with support in $\mathcal{X}$. We assume watermarking is performed on a per-token basis, so we denote $P_X=Q_{X_t|X_1,\dots,X_{t-1}}$ for simplicity.

Consider two parties: the \textit{watermarker} and the \textit{detector} (see Fig. \ref{fig:system}). 
For each token generated by the model, both parties share 
random side information, represented by a random variable $S\sim P_S$ over a discrete alphabet $\mathcal{S}=[1:k]$. 
The watermarker has access to the next-token distribution $P_X$. Their goal is to change $P_X$ using the side information $S$. The detector, in turn, identifies the watermark from a sequence of observed tokens and side information pairs $(X,S)$ and does not know $P_X$.
We divide the watermarking procedure into three main components.

The first component is \textbf{random side information generation}. Both the watermarker and the detector must produce the same side information $S$. In practice, this is accomplished by employing hashing strategies: previous tokens and a secret key are combined and hashed to produce a seed for a random number generator \cite{kirchenbauer2023watermark,dathathri2024scalable}. This  generator then serves as the distribution $P_S$ from which side information $S$ is sampled. Popular hashing strategies consider $\mathsf{min}$, $\mathsf{max}$ or $\mathsf{sum}$ operations over some finite context window of previously generated tokens. Note that both the watermarker and the detector can produce the side information since it only depends on previously generated text and shared keys.

The second component deals with \textbf{how randomness is used}. 
We begin by designing a score function $f:\cX\times\cS\to\mathbb{R}$ that maps a token $x$ and corresponding side information $s$ onto a score $f(x,s)$.
The score function $f$ is fixed and known to both the watermarker and the detector.
In the watermarking stage, we use $f$ and the shared randomness $s$ to change the next-token distribution. 
Specifically, the watermarker samples from the conditional distribution $P_{X|S=s}$ instead of $P_X$. The distribution $P_{X|S=s}$ is designed to increase the probability of tokens with high scores $f(\cdot, s)$.
We call the altered distribution $P_{X|S=s}$ the \emph{watermarked distribution.}

The last component is \textbf{watermark detection}. In the detection stage, the detector receives a sequence of tokens and side information realizations $\{(x_t,s_t)\}_{t=1}^n$, from which they compute a sequence of scores $\{f(x_t,s_t) \}_{t=1}^n$. We assume the sequence of tokens is declared as watermarked if the averaged score $\frac{1}{n}\sum_{t=1}^nf(x_t,s_t)  \geq \tau$, where $\tau$ is a threshold that trades off between specificity and sensitivity of  watermark detection. See Figure \ref{fig:system} for a visualization of the watermarking procedure.

Most existing watermarks can be instantiated in terms of the three components above. For example\footnote{See Appendix \ref{appendix:instantiation} in which we instantiate additional watermarks within the proposed setting.}, in the Red-Green watermark \cite{kirchenbauer2023watermark}, $S$ is a sequence of $m$ bits ($k=2^m$), assigning one bit per token. A realization $s$ of side information is used to partition $\cX$ into two lists, corresponding to the entries that are drawn as 1 or 0. Tokens marked as 1 (``green list'') are upweighted, and tokens marked as 0 (``red list'') are downweighted, resulting in the watermarked distribution $P_{X|S=s}$ from which the next token is sampled. The score function is binary: $f(x,s)=1$ if $x$ is in the green list determined by $s$, or 0 otherwise. Detection is done by averaging binary scores and performing a  threshold test.

\paragraph{An Optimization Formulation for Watermarking.} Given the framework above, we present an optimization formulation from which we derive and analyze \simplexwater{} and  \heavywater{}. Our focus is on optimizing the watermarked distribution $P_{X|S}$ and the score function $f(x,s)$ when $P_S$ is the uniform distribution and we generate a sample $S\sim P_S$ for each token.
We also assume watermark detection is based on thresholding the average score $\frac{1}{n}\sum_{t=1}^nf(x_t,s_t)$ which, in turn, acts as a proxy for $\EE[f(X,S)]$.
When the text is watermarked, this expectation is with respect to $P_{XS}=P_{X|S}P_S$, and otherwise, it is taken with respect to $P_XP_S$.\footnote{Observe that $P_XP_S$ corresponds to the null hypothesis (non-watermarked text):  the generated tokens are independent of the side information $S$.}  Our goal is to design a pair $(f,P_{X|S})$ that maximizes the gap between average score when text is watermarked relative to when the text is not watermarked. %Maximizing this gap will optimize the average z-score in threshold tests for watermark detection. 

Watermark performance will inevitably depend on the entropy of the next-token distribution $P_X$ \cite{kirchenbauer2023reliability,kuditipudi2023robust}. We focus on low-entropy distributions, whose watermarking is considered to be challenging \cite{lu2024entropy}. We adopt the low-entropy constraint given the following definition.

\begin{definition}[Low-entropy distributions]
We use \emph{min-entropy} as our measure of uncertainty for the next-token distribution $P_X$, defined as
$H_{\min}(P_X) = -\log_2 \max_x P_X(x)$
A distribution $P_X$ is said to be \emph{low-entropy} if its min-entropy satisfies $H_{\text{min}}(P_X) \le 1$ bit/token, i.e., at least one token has probability mass greater than $\tfrac{1}{2}$.  
\label{def:low-entropy distributions}
\end{definition}

\begin{remark}[Watermarking in Low-Entropy Regime]
Our theoretical guarantees imply favorable watermarking performance in low-entropy scenarios. 
The ‘low-entropy’ constraint used in our analysis is not merely a theoretical assumption. It is based on the empirical observation that LLMs’ next-token distributions are \textbf{inherently low entropy}. Even though coding, Q\&A, and summarization tasks are commonly understood as having varying degrees of ‘entropy’ – with coding assumed to have lower entropy – we still observe $>90\%$ of next-token predictions across these tasks falling well within the low-entropy considered in our theoretical analysis for practical temperature values (see Appendix \ref{apdx:low_ent} Figures \ref{fig:histogram_stats_qa} and \ref{fig:histogram_stats_coding}). Refer to \cite[Page 13]{li2024statistical} and \cite[Page 4]{fairoze2025publicly} for a similar discussion.
\end{remark}

Given the aforementioned assumptions, the optimal score function $f$ and the joint distribution $P_{XS}$ that maximize detection performance for the worst-case distribution in $\cP_\lambda$ are the solution of the optimization  problem  
\begin{equation}\label{D_gap}
    \Dgap(m,k,\lambda,\mathcal{F}) = \underbrace{\max_{f \in \mathcal{F}}}_{\text{Score}} \hspace{0.3cm} \underbrace{\min_{P_X \in \mathcal{P}_\lambda}}_{\substack{\text{LLM}\\ \text{distribution}}} \underbrace{\max_{P_{XS}}}_{\substack{\text{Watermarked}\\ \text{distribution}}} \Big(\mathbb{E}_{P_{X|S}P_S}[f(X,S)] - \mathbb{E}_{P_X P_S}[f(X,S)]\Big).
\end{equation}

We call \eqref{D_gap} the \textit{maximum detection gap}. 
We restrict the inner maximization in \eqref{D_gap} to couplings between the marginals $(P_X,P_S)$, i.e., $P_{X|S}$ induces a joint distribution $P_{X,S}$ on $\cX\times\cS$ whose marginals are $P_X$ and $P_S$. As a result, the inner maximization is an OT problem \cite{villani2009optimal,peyre2019computational}. 
We denote the class of couplings of $(P_X,P_S)$ as $\Pi_{X,S}$. 
By limiting watermarked distributions to couplings, we ensure that the watermark is \textit{distortion-free}:  we have $\mathbb{E}_{S}[P_{X,S}]=P_X$. From the perspective of a user who does not know $S$, distortion-free watermarks incur (in theory) no perceptible change in text quality.  
In Section \ref{sec:binary} we propose a simple scheme that distorts the coupling solution to further increase $\Dgap$.

\textbf{Proposed method.} Following the optimization \eqref{D_gap}, we propose two watermarking schemes in Sections \ref{sec:binary} and \ref{sec:beyond_binary}, respectively.
Both watermarks differ on the nature of the score function $f$, but share the same underlying algorithmic structure given in Algorithm \ref{alg:watermark} and described next.

Our watermarks receive as input the next-token distribution $P_X$, side information sample $s\in\cS$, and a score function $f$.
Given a pair $(P_X,f)$, the inner maximization in \eqref{D_gap} amounts to an OT problem between $P_X$ and $P_S$, which is set to be uniform on $[1:k]$. 
The OT problem is given by
\begin{align}\label{eq:OT_problem}
P^*_{XS}=\arg\max_{P_{XS}\in\Pi_{X,S}}\left(\mathbb{E}_{P_{XS}}\left[f(X,S)\right]-\mathbb{E}_{P_{X}P_S}\left[f(X,S)\right] \right).
\end{align}
We solve \eqref{eq:OT_problem} using Sinkhorn's algorithm \cite{cuturi2013sinkhorn}. Note that the score function $f$ can be equivalently denoted as the $(m\times k)$-dimensional OT cost matrix $\rC$, which is defined as $\rC_{x',s'}=-f(x',s')$ for $(x',s')\in [1:m]\times[1:k]$. %This matrix is only generated once. 
Sinkhorn's algorithm yields a coupling $P_{X,S}$, from which the watermarked distribution is obtained via $P_{X|S=s}(\cdot|s)=k\cdot P_{X,S}(\cdot,s)$. The watermarked distribution is used to sample the next token.

\textbf{Tilting.} The watermarked distribution returned by the OT optimization does not change the average next-token probabilities. To further increase watermark detectability, we propose a tilting operation to $P_{X|S=s}$, which we denote by $\mathsf{tilt}$. 
The tilting operation increases the probability of higher scoring tokens, at the cost of incurring distortion to $P_{X|S=s}$.
This is done by increasing the probability of $x\in\cX$ with $f(x,s)>\mathbb{E}_{P_{X,S}}[f(X,S)]$ and decrease the probability of $x\in\cX$ with $f(x,s)<\mathbb{E}_{P_{X,S}}[f(X,S)]$.
The amount of tilting (and thereby the level is distortion) is determined by a parameter $\delta\in(0,1)$. 
The structure of tilting depends on the structure of $f$ and is later defined for each watermark separately.
See Algorithm \ref{alg:watermark} for the list of steps.

\textbf{Detection.} Watermark detection is performed as follows: Given a token sequence $(x_1,\dots,x_n)$ and a score $f$ we recover the set of side information samples $(s_1,\dots,s_n)$ and compute $f(x_t,s_t)$ for each observed pair $(x_t,s_t)$. We declare the text watermarked if the average score $\frac{1}{n}\sum_{t=1}^nf(x_t,s_t)$ exceeds a user-defined  threshold $\tau\geq0$.

\textbf{Limitations.} As we will see shortly, \eqref{D_gap} will inform the design of watermarking schemes with favorable empirical performance.
However, this optimization framework is not without limitations.
First, it is restricted to threshold tests applied to the averaged score, which are potentially suboptimal  \cite{xie2024debiasing}.
Second, we assume perfect randomness for side information generation, i.e., we can sample a  uniformly distributed and i.i.d. random variable $S$ for each token. When $X$ is not watermarked, we also assume it is independent of $S$. In sequential hashing schemes this is not necessarily the case, since both $S$ and $X$ will depend on previous tokens. 
We emphasize that generating high-quality side information remains an important practical challenge relevant to all existing watermarks, including ours. Nevertheless, from a design perspective, this challenge is somewhat orthogonal to
our guiding  question of \emph{how to optimally use} side information.
We examine the impact of non-i.i.d. side information on our watermark design in Appendix \ref{apdx:seeding}.

\begin{algorithm}[!t]
\caption{\simplexwater{} and \heavywater{} Watermark Generation}
\begin{algorithmic}[1]
    \State \textbf{Inputs}: Token distribution $P_X$, score function $f$, side information $s$, tilting parameter $\delta$.  
    \State \textbf{Outputs:} Watermarked distribution $P_{X|S=s}$
    \State Calculate OT cost matrix $\rC_{x',s'}=-f(x',s')$, for all $(x',s')\in[1:m]\times[1:k]$
    \State $P_{X,S} = \mathsf{Sinkhorn}\left( \rC, P_X, P_S \right)$, where $P_S=\mathsf{Unif}([1:k])$
    \State $P_{X|S=s}(x|s)= k\cdot P_{X,S}(x,s)$ 
    \If{$\delta>0$}
    \State $P_{X|S=s}\gets \mathsf{tilt}(P_{X|S=s},s,\delta)$
    \State Normalize $P_{X|S=s}$
    \EndIf
    \State \textbf{Return} $P_{X|S=s}$
\end{algorithmic}
\label{alg:watermark}
\end{algorithm}
% \vspace{-0.5em}

\subsection{\simplexwater{}: Watermark Design With Binary Scores}\label{sec:binary}

In this section, we consider the class of binary score functions, i.e.,  $\mathcal{F}=\mathcal{F}_{\mathsf{bin}}\triangleq \{f:\mathcal{X}\times\mathcal{S}\mapsto\{0,1\}\}$. Within this class, we solve \eqref{D_gap} to obtain the optimal binary score function $f$, and present the corresponding optimal watermark, which we call \simplexwater{}. We first present a simplification of the maximum detection gap in \eqref{D_gap} for $f\in \mathcal{F}_{\mathsf{bin}}$ and the low-entropy regime $\lambda\in[1/2,1)$.

\begin{prop}\label{prop:simplified_HD}
% Let $|\cX|=m$, $|\cS|=k$ and f 
Let $\lambda\in\left[\frac{1}{2},1\right)$.
For $f\in\fbin$, define the vector $f_i=[f(i,1),\dots,f(i,k)]\!\in\!\{0,1\}^k$ for each $i\in\mathcal{X}$. Then,
\begin{align}\Dgap(m,k,\lambda,\mathcal{F}_{\mathsf{bin}})=\max_{f\in\fbin}\quad\min_{\substack{\substack{i,j\in\mathcal{X},i\neq j}}} \quad \frac{(1-\lambda) d_H(f_i,f_j)}{k},\label{equivalent}
\end{align}
where $d_H(a,b)=\sum_{i=1}^k\mathbf{1}_{\{a_i\neq b_i\}}$ denotes the Hamming distance between $a,b\in\{0,1\}^k$ and $\mathbf{1}_{\{\cdot\}}$ is the indicator function.
\end{prop}

Remarkably, \eqref{equivalent} is equivalent to a classical problem in coding theory: the design of distance-maximizing codes! 
In this problem \cite{roth2006introduction}, the goal is to design a set of $m$ binary vectors of length $k$ -- called \emph{codewords} -- with maximum pairwise Hamming distance  \cite{roth2006introduction,van1998introduction}.
There is a one-to-one equivalence between designing a code with maximum distance between codewords and designing a binary score function for watermarking LLMs: for a fixed token $x$, the score function vector $[f(x,1),\dots,f(x,k)]$ can be viewed as a codeword. Conversely, a distance-maximizing code can be used to build a binary score function that maximizes the detection gap. 
This connection with coding theory allows us to use classic coding theory results -- specifically the Plotkin bound \cite{plotkin1960binary} -- to derive an upper bound for $\Dgap$ in  \eqref{equivalent}. This result is stated in the next theorem.

\begin{Theorem}[Maximum Detection Gap Upper Bound]\label{thm:converse_upperbound_binary} Consider the class of binary score functions $\mathcal{F}_{\mathsf{bin}}$ and uniform $P_S$. Then, for any $\lambda\in\left[\frac{1}{2},1\right)$, the maximum detection gap can be bounded as 
\begin{align}\label{eq:ub_binary}
\Dgap(m,k,\lambda,\mathcal{F}_{\mathsf{bin}})\leq\frac{m(1-\lambda)}{2(m-1)}
\end{align}
\vspace{-1em}
\end{Theorem}
The proof of Theorem~\ref{thm:converse_upperbound_binary} is given in Appendix~\ref{apdx:proof_thm_1}. For a fixed token vocabulary size $m$, this bound remains the same for any choice of $k$. Therefore, any score function constructed with any  $k$ that achieves this bound is optimal. This observation serves as the starting point for our optimal watermark design, leading to \simplexwater{}.

\paragraph{\simplexwater: An Optimal Binary-Score Watermark.} %\label{simplex_sec} 
%We introduce and leverage tools from coding theory to attain a provably optimal watermarking scheme.
The upper bound in \eqref{eq:ub_binary} is achievable when the score function is constructed using a \textit{simplex code} -- a family of codes 
that attain the Plotkin bound \cite{reed1954class,muller1954application}.
A simplex code is defined as follows:
\begin{definition}[Simplex Code]\label{simplex}
For any \(x,s \in [0:m-1]\), let \(\mathsf{bin}(x)\), \(\mathsf{bin}(s)\) denote their binary representations respectively using \(\log_2 m\) bits. 
A simplex code $\fsim:[0:m-1]\times[1:m-1]\to\{0,1\}$ is characterized by
\begin{align}\label{simplexdot}
    \fsim(x,s) \triangleq \mathrm{dot}(\mathsf{bin}(x), \mathsf{bin}(s)),
\end{align}
where $\mathrm{dot}(\mathsf{bin}(x), \mathsf{bin}(s)) \triangleq \sum_{i=1}^{\log_2 m} \mathsf{bin}(x)_i \cdot \mathsf{bin}(s)_i$ and \(\mathsf{bin}(v)_i\) denotes the \(i\)th bit in the binary representation of \(v\). 
\end{definition}

We present next \simplexwater{}, a watermark that employs the simplex code as its score function.
\simplexwater{} operates according to the steps of Algorithm~\ref{alg:watermark}, where the cost matrix is derived from $\fsim$. 
Specifically, given a token distribution, side information and the cost matrix derived from  $\fsim$, \simplexwater{} solves the OT via Sinkhorn's algorithm to compute the watermarked distribution $P_{X|S}$.
The optimality of \simplexwater{} follows directly from its one-to-one correspondence to the Simplex code that attains the Plotkin bound and is stated formally in  the following theorem. 
\begin{Theorem}[\simplexwater{} Optimality]\label{ach} 
For any $\lambda\in\left[\frac{1}{2},1\right)$ the maximum detection gap upper bound \eqref{eq:ub_binary} is attained by \simplexwater{}.

\end{Theorem}
The proof of Theorem~\ref{ach} is given in Appendix~\ref{apdx:proof_thm_2}. 
Together, Theorems~\ref{thm:converse_upperbound_binary} and~\ref{ach}~imply that \simplexwater{} is minimiax optimal among all watermarks with binary-valued score functions in the low-entropy regime.

As outlined in Algorithm \ref{alg:watermark}, we  further increase detection by tilting the watermarked distribution obtained by \simplexwater{}. For binary-valued scores, the tilting operation (Alg.~\ref{alg:watermark}, step 7) is 
\begin{equation}\label{eq:tilt_binary}
    \mathsf{tilt}(P^*_{X|S=s},s,\delta) = P^*_{X|S=s}(x,s)\left(1+ \delta\cdot( \mathbf{1}_{\{f(x,s)=1\}}-\mathbf{1}_{\{f(x,s)=0\}})\right).
\end{equation}
In Section \ref{sec:numerics} we show that by adding mild distortion we significantly increase the detection power of \simplexwater{}.

\subsection{\heavywater{}: Watermarking using Heavy-Tailed Distributions}\label{sec:beyond_binary}

So far, we restricted our analysis to binary scores, i.e., $f(x,s) \in \{0,1\}$. In this case, the detection gap is bounded by $m(1-\lambda)/(2(m-1))$  as shown in Theorem \ref{thm:converse_upperbound_binary} -- and \simplexwater{} achieves this bound.  
However, the binary constraint is quite restrictive. Watermarks such as \cite{aaronson2023watermark} are not restricted to binary scores -- and achieve excellent performance in practice (see Gumbel in Fig. \ref{fig:fig1}).
In what follows, we relax the binary constraint to improve variation in the score values, enabling us to break the binary detection barrier \eqref{eq:ub_binary} and significantly enhance detection performance.

We consider score functions where each score $f(x,s)$ is sampled independently from a continuous distribution prior to the watermarking process\footnote{A natural first step would be to extend \simplexwater{} by using $q$-ary instead of binary codes. This, however, results in marginal performance gain, as we discuss in Appendix \ref{apdx:qary}.}. Remarkably, under this formulation, the popular Gumbel watermark \cite{aaronson2023watermark} emerges as a special case when  scores are drawn from a Gumbel distribution \cite{huijben2022review}. Our framework allows us to explore a broader class of heavy-tailed distributions that improve upon the Gumbel approach (Figure \ref{fig:fig1}).

\paragraph{\heavywater{}: Heavy-Tailed Score Distributions for Improved Detection.}

Consider the watermarking setting where we draw each score value $f(x,s)$ i.i.d. from a continuous finite variance distribution $P_F$. This score is used in the OT problem in \eqref{eq:OT_problem} and Algorithm \ref{alg:watermark} to obtain the watermarked distribution. 
Next, we show that the Gumbel watermark \cite{aaronson2023watermark} falls into this category of watermarks when $P_F$ is a Gumbel distribution. 

\begin{Theorem}
    [Gumbel Watermark as OT]
\label{thm:gumbel_as_ot}
When the score random variables $f(x, s)$, are sampled i.i.d. from $\text{Gumbel}(0,1)$, the solution to the OT problem in \eqref{eq:OT_problem} converges to the Gumbel watermark \cite{aaronson2023watermark} as $|\mathcal{S}|=k \to \infty$. 
\end{Theorem}

This connection situates the Gumbel watermark within our broader framework \eqref{D_gap}. Can the Gumbel watermark be improved by selecting a different score distribution $P_F$? Intuitively, a heavy-tailed $P_F$ (which Gumbel is not) could increase the probability of sampling large values of $f(x,s)$, thus also increasing the  likelihood of watermark detection. 
To make this intuition precise, we analyze the maximum detection gap \eqref{D_gap} when the score function is randomly generated instead of selected from a set $\mathcal{F}$. % with a family of distributions from which $f(x,s)$ is sampled.

For fixed $|\mathcal{X}|=m$, $|\mathcal{S}|=k$, and $\lambda\in\left[\frac{1}{2},1\right)$, denote the worst case detection gap achieved by some fixed distribution $P_F$ as,
\begin{align}\label{d_gap_random}
    \Dgap^{[P_F]}(m,k,\lambda)=\min_{P_X\in\mathcal{P}_{\lambda}}\max_{P_{XS}}\left(\mathbb{E}_{P_{XS}}\left[f(X,S)\right]-\mathbb{E}_{P_{X}P_S}\left[f(X,S)\right] \right)
\end{align} 
Note that $\Dgap^{[P_F]}(m,k,\lambda)$ is a random variable due to the randomness of the $f(x,s)$ samples. We provide high-probability guarantees on the achievable maximum detection gap for any fixed distribution $P_F$ in the regime of large $k$. 

\begin{Theorem}[Detection Gap]\label{thm:detection-lb-informal}
Let $\lambda\in\left[\frac{1}{2},1\right)$, and consider the score difference random variable $\Delta = f(x,s) - f(x',s')$ for some $(x,s)\neq(x',s')$, where $f(x,s)$ and $f(x',s')$ are sampled i.i.d. from $P_F$. Let the cumulative distribution function of $\Delta$ be $F$, and let $Q = F^{-1}$ be its inverse. Then,
\begin{equation}
\lim_{k\to\infty}\Dgap^{[P_F]}(m,k,\lambda) =\int_{1-\lambda}^1 Q(u)du,
\label{eq:Qint}
\end{equation}
\end{Theorem}

Theorem~\ref{thm:detection-lb-informal} shows that detection performance improves when the distribution of $\Delta=f(x,s)-f(x',s')$ have heavier tails, where both $f(x,s),f(x',s')\sim P_F$.  We analyze several choices of heavy-tailed  $P_F$ for maximizing $\int_{1-\lambda}^1Q(u)du$ (see details in Appendix \ref{apdx:tails}).

\textbf{\heavywater{}: A watermark with heavy-tailed scores.} We propose \heavywater, a watermark that leverages Theorem~\ref{thm:detection-lb-informal}~and samples the score function  from  a heavy-tailed distribution.
\heavywater{} follows the steps of Algorithm \ref{alg:watermark} but uses scores that are generated by sampling i.i.d. from a heavy-tailed  $P_F$ prior to generation. We explored several options for selecting $P_F$. In theory, the Gamma distribution maximizes \eqref{eq:Qint} across several common heavy-tailed distribution (Appendix \ref{apdx:tails}). However, in practice, we observed that choosing $P_F$ to be lognormal achieved the highest detection accuracy.\footnote{Since detection is evaluated via $z$-scores and $p$-values which standardize by mean and variance, the specific mean and variance of the selected distributions do not affect detection accuracy.}
Assuming that $P_F$ is normalized to have zero-mean, the tilting operation in \heavywater{} is given by 
$$
\mathsf{tilt}(P^*_{X|S=s},s,\delta) =  P^*_{X|S=s}(x,s)\left(1+ \delta \cdot\mathsf{sign}(f(x,s))\right).
$$

\subsection{Numerical Experiments}\label{sec:numerics}

\paragraph{Experimental Setting.} 
We evaluate the detection-distortion-robustness tradeoff of \heavywater{} and \simplexwater{} following the experimental benchmarking guidelines in  Waterbench \cite{tu2023waterbench} and MarkMyWords \cite{piet2023mark}.
We use two popular prompt-generation datasets:  Finance-QA \cite{maia201818} (covering  Q\&A tasks)  and LCC~\cite{chen2021evaluating} (covering coding tasks). We evaluate on three open-weight models: Llama2-7B~\cite{touvron2023llama}, Llama3-8B~\cite{grattafiori2024llama}, and Mistral-7B~\cite{jiang2023mistral7b}. Implementation details are provided in Appendix \ref{apdx:more_imp}, and full results, including ablation studies, are presented in
Appendix \ref{appendix:results}.

\textbf{Baseline Watermarks.} We benchmark \simplexwater{} and \heavywater{} against the Gumbel watermark \cite{aaronson2023watermark}, the Inverse-transform watermark \cite{kuditipudi2023robust}, the Correlated Channel watermark \cite{long2025optimized}, and the SynthID watermark \cite{dathathri2024scalable} with binary scores and $K=15$ competition layers. We select these methods based on code availability or ease of implementation. These watermarks are distortion-free (i.e., they preserve the average next token distribution). To our knowledge, they also do not have an out-of-the-box method for trading off distortion with detection accuracy.   
We also consider the Red-Green watermark \cite{kirchenbauer2023reliability}, which can be tuned for such trade-offs.

\textbf{Evaluation Metrics.} 
We use \textbf{p-value} as the primary detection power metric (plotted as $-\log_{10} p$) because it offers a threshold-independent measure of statistical power of a watermark. Moreover, p-values are the metric of choice in  \cite{tu2023waterbench,piet2023mark,kirchenbauer2023watermark,kuditipudi2023robust,dathathri2024scalable, huang2024waterpool, bahri2024watermark}. The p-value measures the probability of obtaining a given test statistic -- or a more extreme one -- under the unwatermarked (null) distribution. This provides a direct connection to the false-alarm rate, avoids the need to choose arbitrary detection thresholds, and is well-suited for comparing different watermarking schemes on equal footing. When the distribution of the test statistic is not available, we use a z-test, which approximates the statistic as Gaussian under the Central Limit Theorem. For methods like Gumbel \cite{aaronson2023watermark}, where the test statistic distribution is known analytically, we use the exact form. At a generation length of $T=300$, we observe that the Gaussian approximation provided by the z-test is very accurate, ensuring that p-values remain reliable even when the exact distribution is unknown.

Distortion is measured using the estimated \textbf{cross entropy} (CE) between the watermarked and unwatermarked distribution, defined as $-\frac{1}{n}\sum_{t=1}^n\log P_X(\tilde{x}_t)$, where $(\tilde{x}_t)_{t=1}^n$ are watermarked tokens and $P_X$ is the LLM's distribution. This is proportional to relative perplexity -- a commonly used proxy for quality evaluation \cite{giboulot2024watermax,fernandez2023three}.
% The CE is defined as $-\frac{1}{n}\sum_{t=1}^n\log P_X(\tilde{x}_t)$, where $(\tilde{x}_t)_{t=1}^n$ are watermarked tokens and $P_X$ is the LLM distribution.
We also measure \textbf{watermark size}, proposed by \cite{piet2023mark} and defined as the number of tokens it takes to detect the watermark with a certain p-value.
We provide results for task-specific metrics for generation quality in Appendix \ref{apdx:waterbench}, including document summarization \cite{fabbri2019multi}, document and long-form QA \cite{zhong2021qmsum,fan2019eli5}, and code completion \cite{chen2021evaluating} as curated by \cite{tu2023waterbench}.

\begin{figure*}[!tb]
  \centering
  % Subfigure 1
  \begin{subfigure}[b]{0.49\linewidth}
    \centering
    \includegraphics[width=\linewidth]{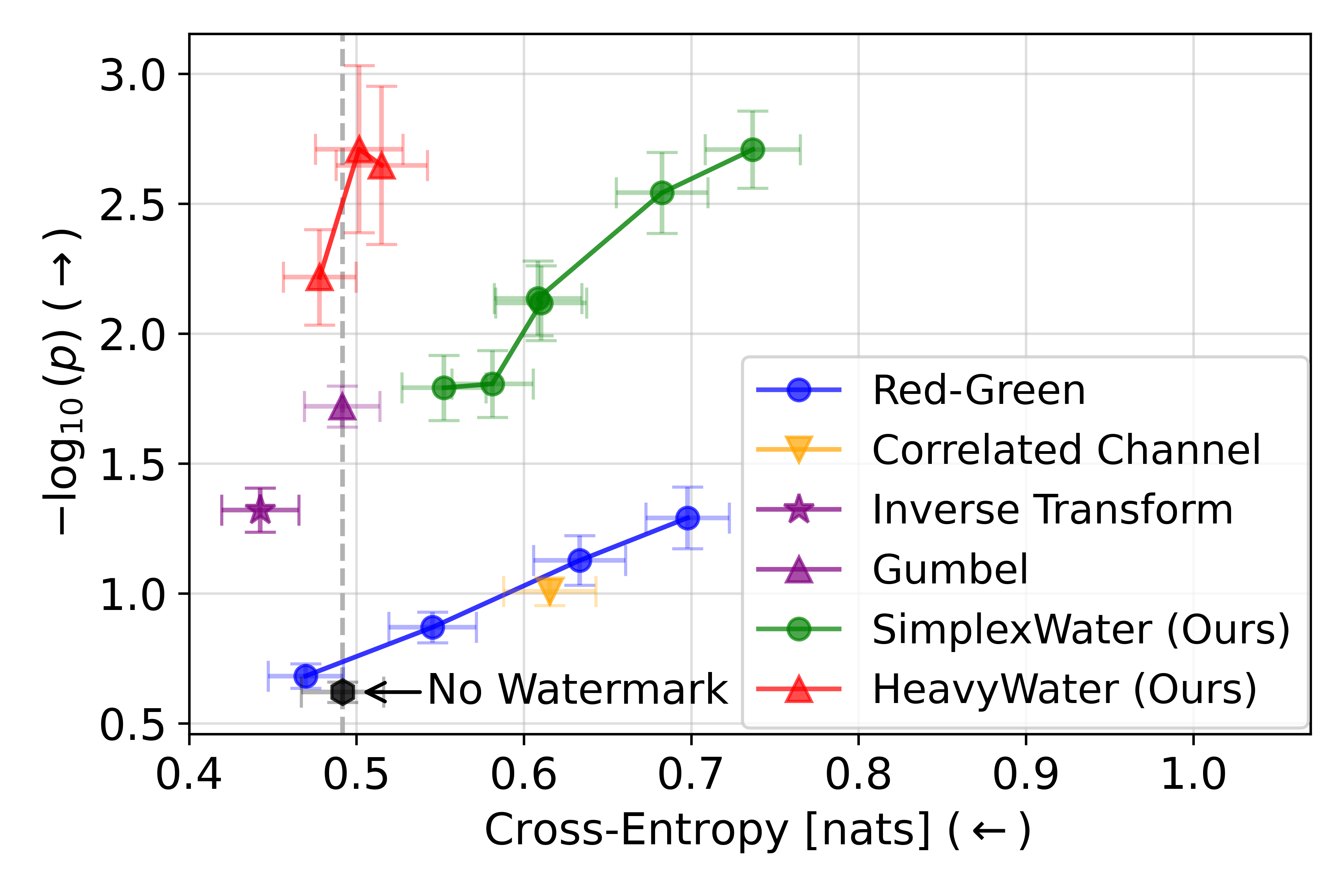}
    \caption{Detection-Distortion Tradeoff. Mistral-7B, Coding. }
    \label{fig:tradeoff_llama2}
  \end{subfigure}
  \hfill
  % Subfigure 2
  \begin{subfigure}[b]{0.49\linewidth}
    \centering
    % \vspace{-1em}
    \includegraphics[width=\linewidth]{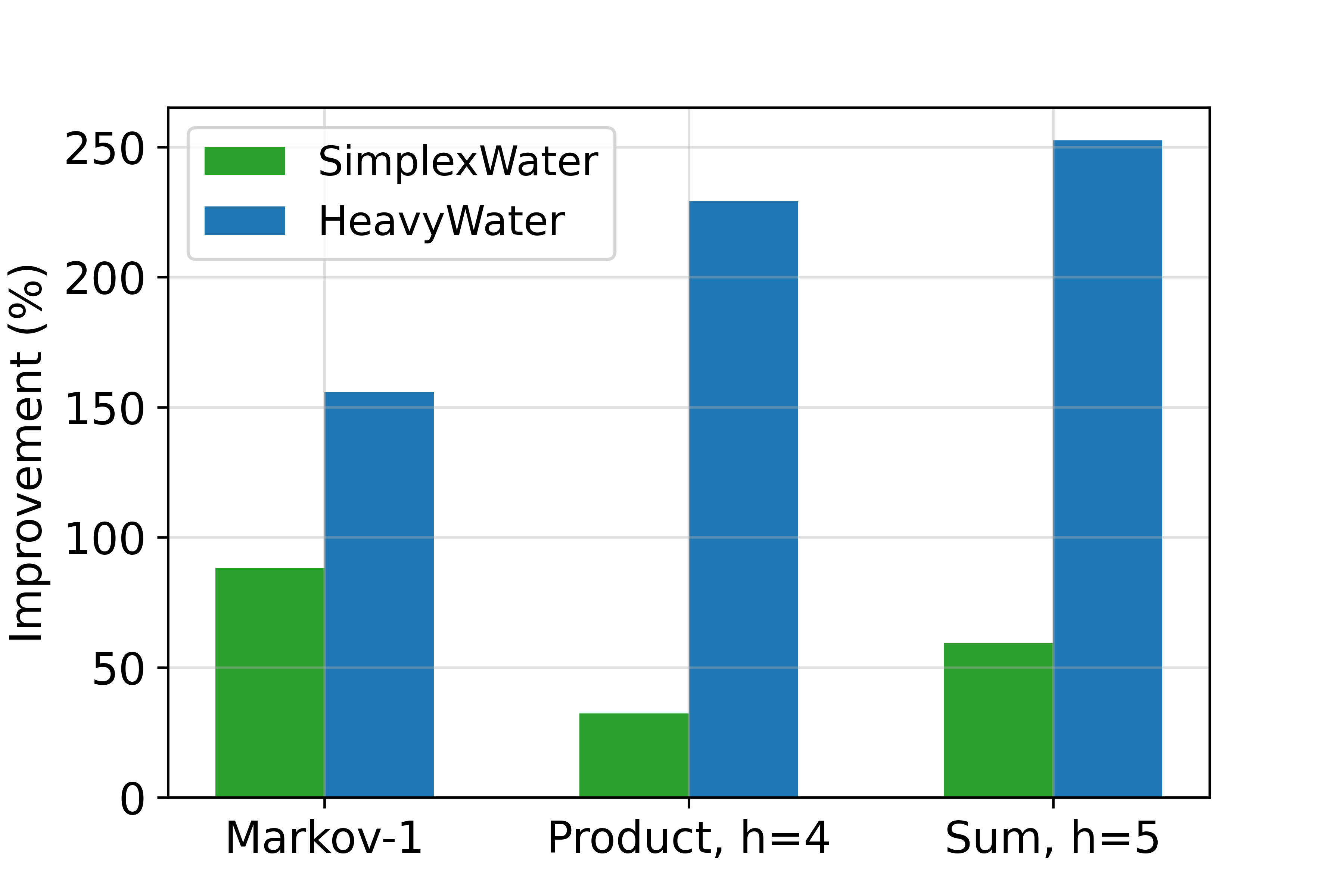}
    \caption{Watermark improvement over Red-Green \cite{kirchenbauer2023reliability} with several hashing strategies, $\delta=2$. Llama2-7B, Coding.}
    \label{fig:watermark_gain}
    % \vspace{-1em}
  \end{subfigure}%
  \caption{\textbf{Left:} Tradeoff between detection (measured by $p$-value) and distortion (measured by Cross-Entropy) --- \simplexwater{} and \heavywater{} achieve higher detection rates while preserving token distributions close to the base unwatermarked model. \textbf{Right:} Detection gained by employing our watermark under various randomness generation schemes and several sliding window sizes $h$. Both \simplexwater{} and \heavywater{} provide a significant improvement of up to $250\%$ and a decrease in distortion.}
  \label{fig:figs_llama2}
  \vspace{-1.5em}
\end{figure*}

\textbf{Detection-Distortion Tradeoff. }
We compare the tradeoff between watermark detection (p-value) against distortion (cross entropy).
The results are given in Figures \ref{fig:fig1} and \ref{fig:tradeoff_llama2}. We make two key observations. First, considering watermarks with binary scores, \simplexwater{} Pareto dominates the red-green method.
Second, \heavywater{} provides the top performance along the baseline watermarks, incurring zero distortion. When we tune the watermarked distribution, both $\simplexwater{}$ and $\heavywater{}$ obtain significant gains in detection, with mild distortion incurred.

\begin{table}[ht]
\centering
\caption{Results on HumanEval Functional Coding Dataset (pass@k, higher is better)}
\label{tab:humaneval-passk}
\begin{tabular}{lccc}
\toprule
\textbf{Watermark Scheme} & \textbf{pass@1 (\%) $\uparrow$} & \textbf{pass@5 (\%) $\uparrow$} & \textbf{pass@10 (\%) $\uparrow$} \\
\midrule
No Watermark            & 14.0 & 20.7 & 28.1 \\
\textbf{HeavyWater (Ours)}   & \textbf{13.1} & \textbf{18.9} & \textbf{27.8} \\
Gumbel                  & 14.3 & 20.5 & 25.6 \\
\textbf{SimplexWater (Ours)} & \textbf{13.7} & \textbf{22.7} & \textbf{25.3} \\
Inverse Transform       & 13.8 & 22.0 & 27.5 \\
Red/Green $\delta=3$    & 11.6 & 19.5 & 23.2 \\
\bottomrule
\end{tabular}
\end{table}

\textbf{Additional Quality Metrics for Textual Tasks.} In addition to cross entropy, we have reported additional performance metrics for text quality: ROUGE-L and F1 scores for four additional tasks in Appendix \ref{apdx:waterbench}. In Table \ref{tab:waterbench}, we assess the impact of watermarking on generation quality across four additional tasks from WaterBench \cite{tu2023waterbench}: LongformQA, Multi-news Summarization, Knowledge Memorization, and Knowledge Understanding. For these, we report ROUGE-L scores (LongformQA and Multi-news) and F1 scores (Knowledge Memorization and Knowledge Understanding). Our methods (SimplexWater and HeavyWater) maintain competitive detection performance across all datasets, with a generation metric comparable to unwatermarked text (see Table \ref{tab:waterbench}). Together, these results provide significant evidence that our watermarks preserve output quality in textual tasks.

\textbf{Additional Quality Metrics for Coding Tasks.}
We report additional tasks and metrics for low-entropy generation in coding. For LCC code completion tasks\cite{chen2021evaluating}, since human-written codes are used as the ground truth, \textbf{Edit\_Similarity} is the generation-quality metric. We benchmark our watermarking methods against prior work using this metric, and HeavyWater achieves the highest Edit\_Similarity score for the code completion task (see Table \ref{tab:lcc-editsim}). 
For the HumanEval dataset \cite{chen2021evaluating}, we take the standard metric \textbf{pass@$K$} with $K$=1,5,10 generation metric. 
As shown in Table \ref{tab:humaneval-passk}, SimplexWater achieves the highest score on pass@5 and HeavyWater performs best on pass@10. 
These results demonstrate that our watermarks preserve both functional correctness and syntactic quality of generated code.

\begin{wrapfigure}{r}{0.48\linewidth}
  \vspace{-3em}
  \includegraphics[width=\linewidth]{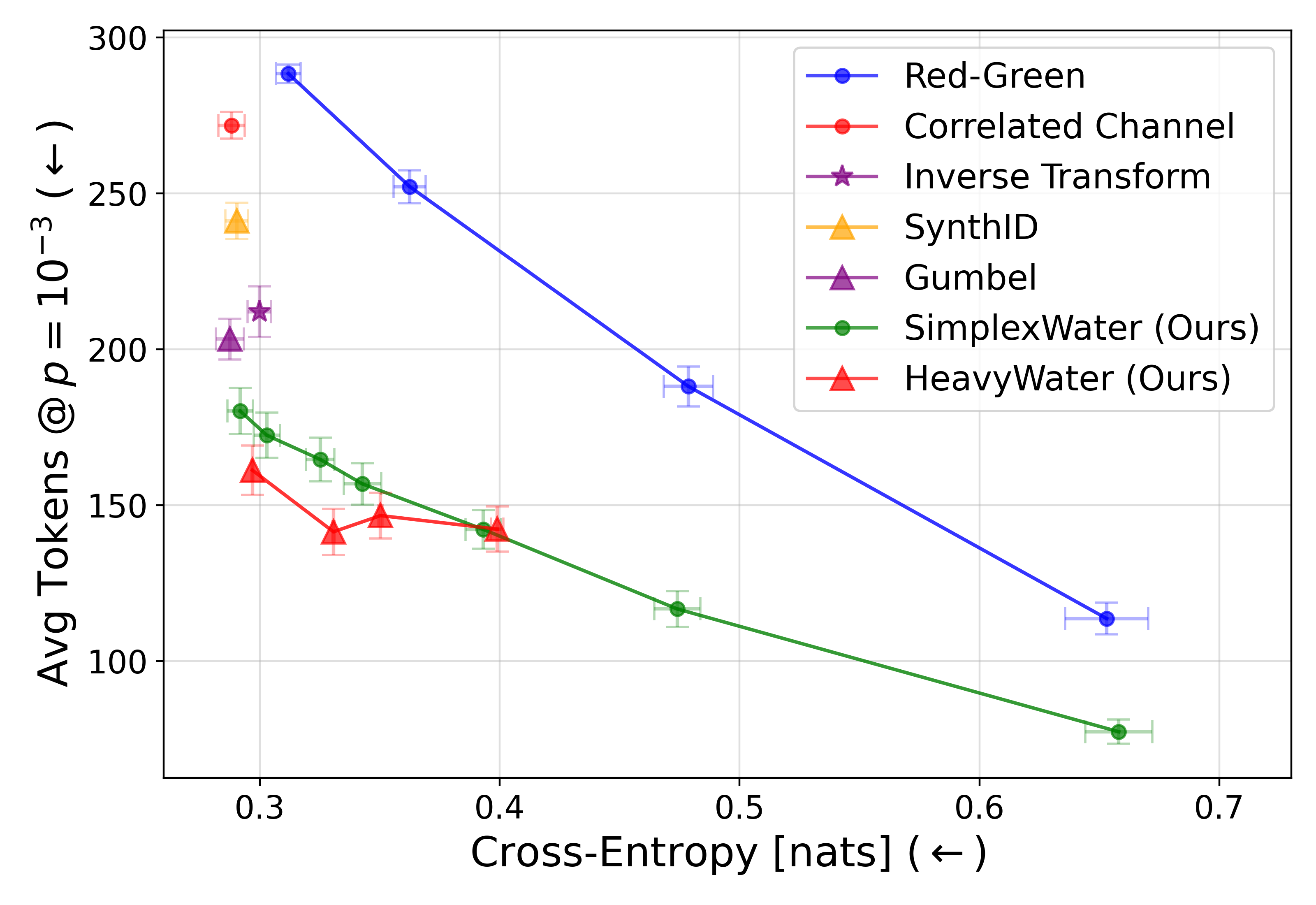}
  \caption{Our watermarks require fewer tokens to reach a given detection strength (p-value) with zero distortion.}
  \vspace{-1.8em}
  \label{fig:wm_size_figs_llama2}
\end{wrapfigure}

\textbf{Watermark Size.}
We consider $p=10^{-3}$, which corresponds to a $0.1\%$ false-positive rate and report the average generation length to reach this $p$ value.
As seen in Figure \ref{fig:wm_size_figs_llama2}, where Llama2-7B is run on Q\&A dataset, both the \simplexwater{} and \heavywater{} schemes provide the lowest watermark size in terms of the number of tokens required to attain a watermark detection strength, as measured by the p-value. 

\textbf{Gain and Performance Under Hashing.}
We  illustrate how \simplexwater{} and \heavywater{} can be coupled with different side information generation methods to boost watermark detection. 
We consider an experiment in which we replace the Red-Green watermark cost function and watermarked distribution with ours and report the gain in detection under several seed hashing methods adopted in \cite{kirchenbauer2023reliability}.
As seen in Figure \ref{fig:watermark_gain}, simply switching Red-Green ($\delta=2$) with our methods -- while keeping the same hashing scheme -- introduces significant detection gain (up to $250\%$ improvement in the aggregated $-\log_{10} p$ value).

\textbf{Additional Results.} 
The following experimental results and ablation studies are included in Appendix \ref{appendix:results}:
1) An evaluation on the WaterBench dataset \cite{tu2023waterbench} in which we show that our methods do not degrade textual quality across an array of generation metrics (Appendix \ref{apdx:waterbench}); 2) A robustness study in which evaluate tamper resistance of our watermarks (Appendix \ref{apdx:robustness}), and 3) computational overhead measurements, where we compare generation wall-clock times of various watermarking schemes (Appendix \ref{apdx:overhead}).

\subsection{Conclusion}
We developed two new watermarking schemes by characterizing and analyzing an optimization problem that models watermark detection in low-entropy text generation.
When restricting attention to binary scores, we draw connections to coding theory, which we then use to develop \simplexwater{}. We prove  that \simplexwater{} is minimax optimal within our framework. 
When we sample the scores i.i.d. at random, we show that watermark detection depends on the tail of the score distribution. This leads to \heavywater{}, where the scores are sampled from a heavy-tailed distribution.
We further show that the Gumbel watermark is a special case of this construction.
Both \simplexwater{} and \heavywater{} demonstrate favorable performance on an array of LLM watermarking experiments when compared to various recent schemes. An interesting direction for future work is further optimizing the score function. This could be done by exploring other heavy-tailed score distributions, or using alternative strategies to optimize $f(x,s)$ such as backpropagating through \eqref{D_gap}. We also restrict our analysis to threshold tests -- more powerful statistical tests could improve detection accuracy (see \cite{li2024robust}). 
Our work contributes to better authentication of text provenance and fostering trust in AI systems. Agnostic to how side information is generated, our watermarks can be paired with new research in side information generation strategies that are more robust to tampering and adversarial attacks.

\chapter{Autonomous Multi-Agent System for Supply Chain Management}
\label{ch:5}

\section{Introduction}

Less than a year ago, it seemed like that day when generative AI would bring about a new era of supply chain autonomy—one where AI could adeptly make all the inventory and logistics decisions—was still far off. But to the astonishment of many experts, including us, that day has arrived—at least in the lab.

In this chapter, we will share our findings on the capabilities of current generative AI models to manage supply chains autonomously and provide some high-level advice on how to build such a system.

\section{Automated vs. Autonomous Systems}
For a decade, supply chain leaders have raced to automate processes by deploying robots, building digital twins, and designing optimized data-driven, inventory-management policies. This wave of automation has enabled faster operations, reduced errors, and led to supply chains that operate according to carefully designed sets of rules. Yet automation has a ceiling. Humans still write the rules, coordinate across functions, and make management decisions. Automated supply chains adapt by applying the given rules but cannot learn, reason, or manage the fundamental tradeoffs that define supply chain operations. In contrast, supply-chain-management systems powered by generative AI could have the capability to operate autonomously.

Using a simulation model that we built in our lab, we tested whether newly released gen AI reasoning models could manage supply chains autonomously, coordinating demand forecasting, inventory planning, and replenishment decisions across multiple functions with minimal human oversight. The results (Appendix \ref{apdx:proofs_ch5}) were striking. A system comprised of multiple agents—each powered by the same state-of-the art gen AI model like either GPT-5 or Llama 4—that shared information turned in a stellar performance (Table \ref{tab:supply_chain_full}). Such systems even outperformed more than 100 undergraduate students at Georgia Tech’s Scheller College of Business, cutting total supply chain costs, which consist of backorder penalties for unfulfilled demand and holding costs for excessive inventory, by as much as 67\% compared to the performance of the students (Figure \ref{fig:supply_chain_bar_plots}). By comparison, older-generation large language models, including those many firms still use, often fail catastrophically in our simulations, generating supply chain costs up to five times higher than human teams achieve.

We also found that such a system can learn and adapt as conditions change with minimal human intervention: It can learn from their environment, anticipate bottlenecks, and adjust strategies in real time. This is the first evidence that gen AI can handle the cross-functional complexity that human supply-chain managers navigate daily.

\section{Our Simulation Model}
Most companies don't train their own AI models. They use frontier models like GPT-5, Claude, and Llama 4 off the shelf, accessed through standard interfaces with minimal customization. The question, therefore, that we explored wasn’t how to build better models. It was how to effectively deploy models that already exist. 

Our research asked the following fundamental questions: When gen AI models are used as is, with natural language prompts and no model modification, can autonomous agents effectively manage complex supply chain operations? And what strategies must supply chain managers master to orchestrate these off-the-shelf models successfully?

We built the first autonomous supply chain testbed around one of management education's most enduring simulations: the MIT Beer Distribution Game. For nearly 70 years, this deceptively simple exercise has humbled MBA students and seasoned executives alike. Developed in the 1950s by Jay Forrester to explain puzzling production swings at General Electric, the game captures the essential dynamics of any supply chain: information delays, coordination failures, and the human tendency to overreact under uncertainty.

The game works as follows. Four players—retailer, wholesaler, distributor, and factory—form a serial supply chain. Each week, every player makes one decision: how much to order from their upstream partner. The goal is straightforward: meet customer demand at the lowest total cost, balancing inventory expenses against costly backorders. The structure makes this deceptively difficult. Players operate in silos and cannot communicate. Only the retailer sees actual end-customer demand. Built-in shipping and ordering delays magnify uncertainty. When humans play, the result is almost always the same: chaos.

A minor spike in demand cascades into disaster. The retailer, seeing a short-term uptick, orders slightly more as a buffer. The wholesaler interprets this larger order as a lasting surge and orders even more in turn. The distributor and factory amplify the signal further. This chain reaction, known as the bullwhip effect, transforms small fluctuations into massive swings in inventory and cost. Delayed shipments eventually flood the system, leaving everyone drowning in excess stock. This cycle is nearly impossible to escape, even when you know it's coming.

In our testbed, four autonomous AI agents, each powered by one type of large language model such as GPT-5, operated the same supply chain under identical constraints. They faced the same information silos, delays, and pressure to avoid stockouts. Like humans, they had to anticipate demand, manage inventory, and coordinate implicitly across the chain. Unlike humans, they can be systematically orchestrated, with their decision-making guided by policies, data sharing, and carefully designed prompts.

We ran hundreds of simulations, testing different models and inference-time methods—techniques that optimize how models are used rather than how they are trained. These methods included designing better prompts, controlling what data to share, and implementing guardrails to limit the range of acceptable actions. We benchmarked AI performance against that of humans: We used data from 12 Georgia Tech cohorts with more than 100 undergraduate students in total who played the Beer Game over the past two years, all operating under the same system conditions as the gen AI testbed. In our best-performing setup—using Llama 4 Maverick 17B with optimized prompts, data-sharing rules, and guardrails—the AI agents reduced costs by as much as 67\% relative to the student teams.

\section{How Different Models Performed}
Our experiments revealed a sharp divide in the capability of current gen AI models. The release of GPT-5-class models in the summer of 2025 marked the widespread availability of a new generation of reasoning models that fundamentally differ from their non-reasoning predecessors. Earlier non-reasoning models solved problems by matching patterns from their training data; they could predict how to respond to queries linearly but were limited in their capability for making decisions with clear, structured logic. The new generation of reasoning models break down complex problems into manageable steps, solving them through explicit logical reasoning; they are guided by plan-execute-reflect loops, where reasoning continuously updates the plan as the model works toward a solution, enabling truly adaptive decision-making. 

Reasoning models outperformed non-reasoning models by wide margins in our testbed. In addition, providing each model with the right information—tailored to supplement its capabilities—and policies that constrain the range of acceptable decisions improved performance across the board.

\section{Important Factors Affecting Performance}

Four factors determine whether autonomous gen AI agents succeed or fail in supply chains.

\paragraph{1. A capable, reliable model.} Model selection matters most—no amount of orchestration can fix a model that cannot understand the task or follow instructions. An agent’s core reasoning capability directly drives supply chain costs and stability. Less-capable models amplify system noise (i.e., misleading signals about real demand) into costly bullwhip effects while more capable models can dampen it.

To test reliability of LLM models, we conducted many identical runs of the simulation for each model. In our decentralized setup—one in which no information is shared across the gen AI agents—we found many popular models like Llama 3.3 70B and GPT-4o mini are highly inefficient; they produce pronounced bullwhip effects, and their costs are an order of magnitude higher than those of human teams. All models showed instability in terms of their performance across identical runs (i.e., their outputs were unpredictable, inconsistent, and often degraded in quality or reliability over time), with total costs varying from 13\% to 46\% of the mean. Llama 4 Maverick 17B exhibited the greatest variability.

Worse, some models simply failed to follow instructions, causing systemic breakdowns. In our trials, models like Microsoft’s Phi-4 and DeepSeek-R1-0528 failed to generate their decision in the required format in over 25\% of cases.

However, the latest generation of models with advanced reasoning capabilities produced a clear leap in performance. For example, upgrading agents from GPT-4o mini to GPT-5 mini cut total supply-chain costs by 70\%. Similarly, the newer and smaller Llama 4 Maverick 17B model dramatically outperformed its much larger predecessor, Llama 3.3 70B, reducing costs by 82\%.

The superior performance of advanced reasoning models can be attributed to the policy they adopt to make decisions. A striking observation was that the newer reasoning models frequently applied the classic order-up-to policy—raising inventory positions to a target level—whereas older reasoning models often failed to articulate a coherent rationale for their decisions.

\paragraph{2. Guardrails to limit costly errors.} Policies that constrain a gen AI agent’s range of possible actions can materially improve both efficiency and reliability. For instance, a policy might cap order quantities or prevent new orders once inventory exceeds a set threshold.

In our experiment, a simple budget constraint proved highly effective. Each gen AI agent was given a fixed budget; orders could not exceed the available funds. In the real world, this hard guardrail works by preventing human purchasing agents from making panic buys. When an agent faces a stockout and attempts to place a massive order, the budget acts as a brake, forcing a more measured response and curtailing the amplification of misleading demand signals up the chain that can lead to bullwhip effects.
The results were dramatic: Total costs dropped 25\% for GPT-5 mini, 39\% for GPT-4o mini, and 41\% for Llama 4 Maverick 17B. For capable models like Llama 4 Maverick 17B whose performance without the guardrail had been unstable, variation in performance across runs fell from 46\% to 37\%.

\paragraph{3. Curated data shared through a central orchestrator.} LLMs don’t reason like humans. The data that helps your team can distract an AI agent, leading to worse decisions and higher costs. So, be selective and test what data you share with an AI agent. For more capable gen AI models, less is often more.

To test how information sharing affects agent performance, we introduced a central “orchestrator,” an agent with full visibility across the supply chain, responsible for sharing specific, curated data with the agents playing the game. We tested two information-sharing strategies, where the orchestrator shares information but makes no decision, and found that more data is not always better.

\begin{enumerate}
    \item Scenario 1: Share real-time customer demand. When the orchestrator shared only the current week’s end-customer demand, performance improved across the board. Total costs fell by approximately 18\% for GPT-5 mini, 25\% for Llama 4 Maverick 17B, and 38\% for GPT-4o mini.
    \item Scenario 2: Share demand history and analysis. When we also provided a five-week demand history and a volatility analysis, the results were mixed. This richer data significantly helped less-capable models (costs for GPT-4o mini fell by 69\%). But for more capable models, the extra information was a distraction, and they performed worse than when they only received information on real-time demand.
\end{enumerate}

Notably, other data points that typically help humans—such as inventory position or pipeline inventory—offered little benefit and often made the bullwhip effect worse.

\paragraph{4. Fine-tune performance with better prompts.} Prompt design can significantly improve the performance of less-capable models, but it may offer limited benefit for more-capable models. For more-capable models, robust guardrails and curated data matter more.

Because LLMs are probabilistic, the way you frame a task matters. Reframing the objective (i.e., the instruction given to LLM) from the general goal of “minimize total costs” to the more specific “minimize the weighted average of backlog and holding costs” produced large gains for less-capable models, cutting costs by 44\% for GPT-4o mini and 33\% for GPT-4.1 mini. For more capable models, the effect was negligible.

\begin{figure}[htbp]
    \centering
    % --- First Plot ---
    \begin{subfigure}[b]{0.8\textwidth}
        \centering
        \includegraphics[width=\textwidth]{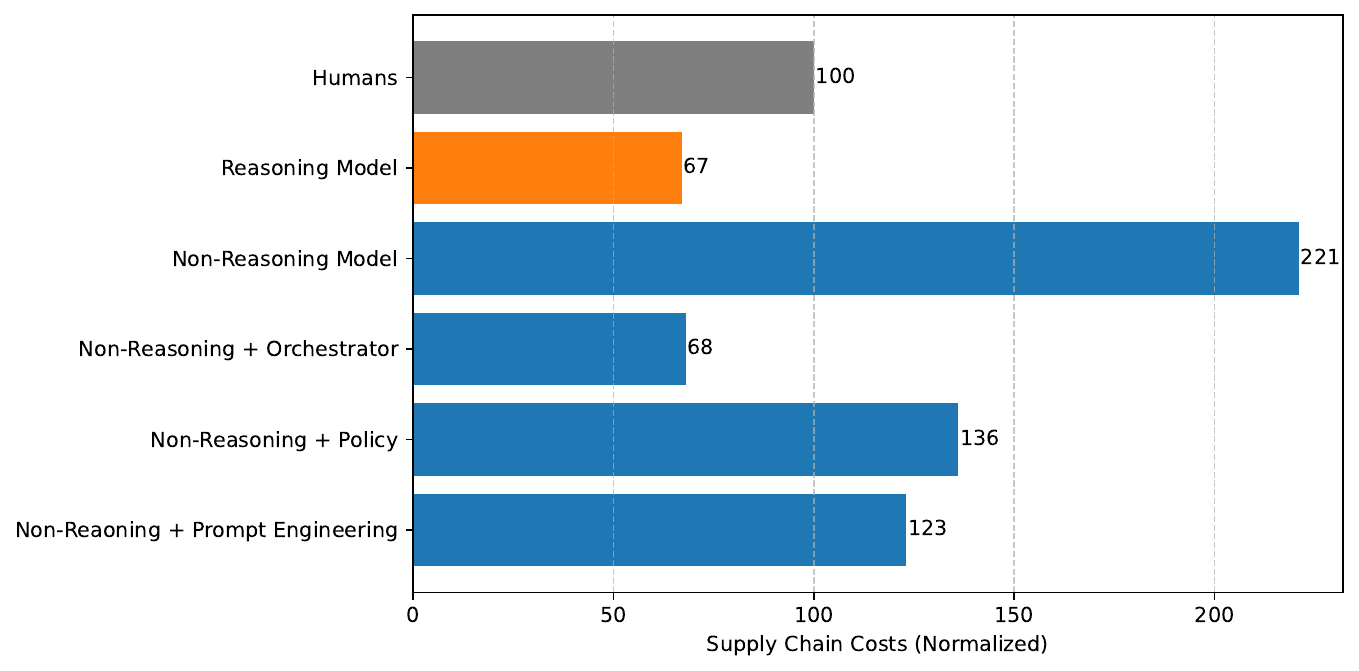}
        % \caption{Description of the top plot}
        \label{fig:top_plot}
    \end{subfigure}
    
    \vspace{1cm} % Adds vertical space between the two plots

    % --- Second Plot ---
    \begin{subfigure}[b]{0.8\textwidth}
        \centering
        \includegraphics[width=\textwidth]{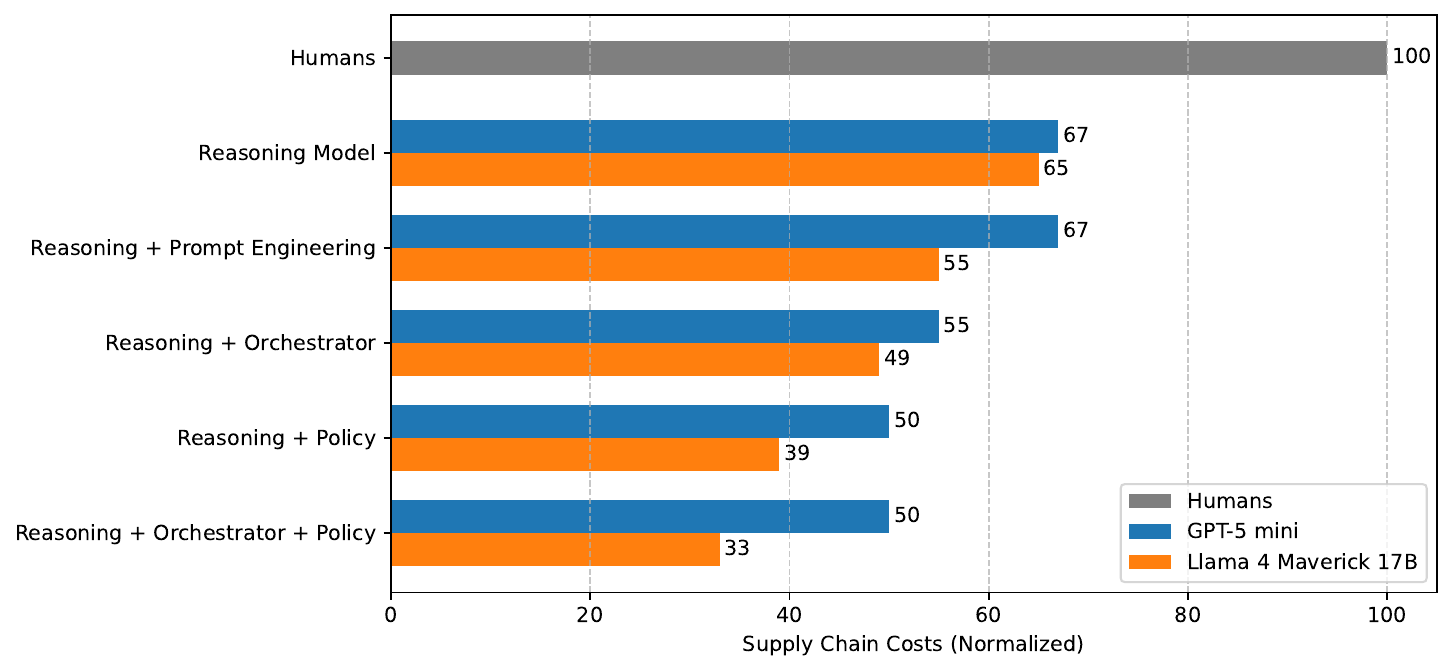}
        % \caption{Description of the bottom plot}
        \label{fig:bottom_plot}
    \end{subfigure}
    
    \caption{Performance gains of gen AI agents via model selection and inference-time techniques. Non-reasoning models (top, GPT-4o mini) required policy constraints, orchestration, and prompt engineering to close the performance gap with humans. In contrast, reasoning models (bottom, GPT-5 mini and Llama 4 Maverick 17B) started above human-level performance, and, when optimized with the same techniques, achieved up to a 67\% reduction in cost relative to human teams.}
    \label{fig:supply_chain_bar_plots}
\end{figure}

Figure \ref{fig:supply_chain_bar_plots} summarizes the effects of the four factors that determine the success or failure of the autonomous supply chains. Non-reasoning models (GPT-4o mini) performed poorly out of the box, generating costs 2x higher than that of human teams. However, when curated information is shared via a central orchestrator, the model surpassed human performance by 33\%. We demonstrate that reasoning models (GPT-5 mini and Llama 4 Maverick 17B) already outperformed human teams in baseline tests. When enhanced with inference-time techniques—such as information sharing, policy constraints (budget guardrails), and refined prompting—these agents delivered 50\% to 67\% reductions in total supply chain costs compared to human teams operating under identical conditions.

These results indicate that gen AI agents powered by state-of-the-art reasoning models can manage supply chains with proficiency that exceeds human teams, at least within the testbed.

\section{A New Paradigm for Supply Chain Management}
Our autonomous supply chain testbed demonstrates that gen AI agents, when provided with tailored information and policies that constraint the range of actions and orchestrate the flow of information among them, are capable of managing multi-function supply chain systems. It means that the capabilities of gen AI models have progressed to the point that autonomous systems—those that learn, adapt, and coordinate across functions in real time—are achievable and can replace both human-managed systems and automated systems that follow human-designed rules.

Critically, this approach has minimal development costs. Unlike traditional AI implementations that required expensive model retraining and specialized data science teams, properly configured gen AI agents can deliver substantial value straight out of the box. Moreover, barriers to adoption have all but disappeared. With the release of OpenAI’s AgentKit in October, even non-technical teams can design and deploy autonomous agents without writing a single line of code. World-class supply chain management is becoming a plug-and-play capability, accessible to any business that understands how to guide the gen AI agents with the right data and policies.

The implications extend beyond cost reduction. When autonomous agents handle operational coordination, human managers can redirect their expertise toward strategic challenges: network redesign, supplier relationships, cross-functional integration across the supply chain, finance, and marketing and sales. Thus, the role of supply chain leadership shifts from operator to orchestrator—from designing rigid rules to guiding intelligent agents.
The testbed also reveals a broader opportunity. Because gen AI agents can execute supply chain simulations in minutes rather than weeks, organizations can now rapidly test policies, benchmark strategies, and identify optimal approaches at unprecedented speed. This transforms supply chain strategy from experience-based intuition to data-driven experimentation.

This technological breakthrough arrives amid unprecedented volatility—from black-swan events to geopolitical shocks and fragile global networks that traditional forecasting models can’t handle. In this environment, gen AI’s ability to reason, simulate, and adapt dynamically makes it not just a technological advantage but a strategic imperative.

\subsection{How to Embark on the Journey}

Executives should take three steps to start experimenting with these new systems:

\paragraph{First, audit your AI infrastructure.} Identify which models currently power your supply chain systems. Many firms still rely on older, non-reasoning gen AI models that our research shows will fail at autonomous coordination of information across functions. They will have to upgrade to reasoning models.

\paragraph{Second, start with constrained pilots.} Deploy autonomous agents in bounded environments with clear guardrails. Test budget constraints, experiment with information sharing, and measure performance against human benchmarks. The methodology we used in our simulations llprovides a template for this experimentation. If your company operates a digital twin to simulate decisions in your supply chain, foow our methodology and embed gen AI agents in the digital twin. Doing so lets you quickly test, learn, and pinpoint how and what delivers real impact for your business.

\paragraph{Third, build orchestration capabilities.} The autonomous supply chain requires a new skillset: curating data flows between agents, designing policies that prevent systemic failures, and crafting prompts that align agent behavior with business objectives. These capabilities will differentiate leaders from followers.

\section{Concluding Remarks}
Our experiments suggest that the age of autonomous supply chains is at hand. Success will require more than deploying powerful models. It will demand a new form of leadership that orchestrates intelligence rather than executes tasks, one that designs systems for learning rather than compliance.

\chapter{Conclusion and Future Work}\label{ch:conclusion}

This thesis advances a unified vision of Trustworthy AI by ensuring reliability and accountability from models to agents. Grounded in information theory, optimization, and statistical learning, we develop principled foundations and algorithms that make AI systems more reliable, less biased, less arbitrary, and more accountable. Across fairness and arbitrariness in machine learning, watermarking of Large Language Models, and multi-agent decision-making of AI agents, prior assumptions in trustworthy AI are reexamined and new theoretical connections are introduced, reshaping how we evaluate and design AI systems for high-stakes domains such as education, healthcare, and supply chain management.

On model reliability, this thesis addresses two fundamental sources of unreliability: bias and arbitrariness. For model bias, we develop theory and algorithms that go beyond predefined demographic groups, and design data-driven methods to identify and reduce miscalibration on all subgroups of the population. These contributions reduce embedded human bias in fairness metrics and provide practical tools to pinpoint where models fail.

Furthermore, we introduce and formalize a second dimension of unreliability: arbitrariness. We prove that fairness-improving algorithms can paradoxically increase arbitrariness and may even exhibit maximal arbitrariness, revealing that the traditional fairness–accuracy tradeoff is incomplete. This work establishes arbitrariness as a necessary third axis for evaluating machine learning systems. These results are especially relevant in human-facing applications, where inconsistent predictions carry individual-level consequences.

On accountability, we establish an information-theoretic foundation for watermarking large language models. We analyze the detection-quality tradeoff given a fixed size of side information, and design the distortion-free Correlated Channel Watermark via optimizing the channel between output tokens and shared side information. We further connect LLM watermarking to coding theory, showing that designing optimally detectable watermarks with binary scores is equivalent to constructing codes with large Hamming distance between codewords. This connection leads to the SimplexWater algorithm based on Simplex codes. By unifying optimal transport, coding theory, and watermarking, our work lays the groundwork for reliable provenance-tracking mechanisms that promote trust in generative AI systems.

In the field of agentic AI, we develop a multi-agent testbed built on the MIT Beer Distribution Game to evaluate whether LLM-based agents can coordinate in a decentralized, multi-stage supply chain. Each facility in the supply chain is controlled by a generative AI agent, enabling the study of emergent system-level behavior under dynamic demand and built-in delays. Using advanced reasoning models with proper orchestration, we show that these agents can adapt to non-stationary conditions, reduce cumulative cost, and attenuate the bullwhip effect---a canonical failure mode in distributed supply chains. Importantly, reliability depends not only on model selection, but also on guardrail design, structured information sharing, and prompt specification. These findings demonstrate both the promise and fragility of autonomous LLM coordination.

We highlight the risks of tail events arising from LLM stochasticity and complex system dynamics. Therefore, a critical future direction is to ensure the reliability of LLM agents’ decision-making in multi-agent, multi-stage environments. Strong average performance is insufficient in high-stakes domains, where rare but consequential failures can dominate system outcomes. Trustworthy deployment requires guarantees of stability, robustness to distribution shifts, and resilience to cascading failures. By providing a controlled experimental framework to systematically study these dynamics, this line of research lays the foundation for principled mechanism designs that promote agent reliability.

In the long term, our research aims to build theoretically grounded mechanisms for reliable decision-making using AI. We seek to transform AI systems from isolated predictors and chatbots into dependable decision-making agents accessible to businesses and communities. Across models and agents, this thesis advances a cohesive research agenda that embeds accountability, mitigates bias, controls arbitrariness, and ensures reliable collective behavior in multi-agent systems, taking strides toward a safer and more trustworthy integration of AI into society.

%%%%%%%%%%%%%%%% BACK MATTER %%%%%%%%%%%%%%%%

% Put appendices, bibliography, and supplemental materials here

% The bibliography may be single spaced within each entry, but must be
% double-spaced between each entry. Most bibliography styles leave space between
% entries, so that shouldn't be a problem.
% \addcontentsline{toc}{chapter}{References}
\begin{singlespacing}
\renewcommand{\bibname}{References}
\bibliography{references}
\bibliographystyle{apalike}
\end{singlespacing}

% Appendices from all chapters should go at the end
\begin{appendices}

\chapter{Chapter \ref{ch:2}}
\label{apdx:proofs_ch2}

The appendix is divided into the following five parts. Appendix~\ref{sec:proof of orthogonality}: proof of the orthogonality of OAE and SP from ambiguity (Proposition~\ref{proposition: orthogonality OAE}); 
%\ref{sec:: proofs}: Proofs of theoretical results; 
Appendix~\ref{sec:proofs of concentration}: proofs from Section~\ref{sec:ensemble}; Appendix \ref{sec:: optimize param}: discussion on optimizing ensemble parameters; Appendix \ref{sec:: discussion fairness}: additional discussions on group fairness and fairness interventions; and Appendix \ref{sec:: additional experiments}: additional experiments and details on the experimental setup.

\section{Proof of Proposition~\ref{proposition: orthogonality OAE}: Orthogonality of OAE and SP from Ambiguity} \label{sec:proof of orthogonality}

    We divide the proof into two cases according to the group-fairness metric considered (OAE or SP).

    \subsection{Proof for the OAE Metric}
    
    Let $\calD \subset \BR^d \times [K]\times \{0,1\}$ be a dataset satisfying the two conditions listed in Remark~\ref{remark:OAE orthogonality}, namely, 
    \begin{enumerate}
        \item with $n_k$ denoting the number of samples in $\calD$ belonging to group $k\in [K]$, the greatest common divisor of the $n_k$ is at least $(m-1)/(m\epsilon - 1)$. 

        \item if $(\bx,s,y),(\bx,s',y')\in \calD$ share the same feature vector $\bx$, then $y=y'$ too.
    \end{enumerate}

    Consider the partition along group membership $\calD = \bigcup_{k\in [K]} \calD_k$, so we may write $\calD_k=\{(\bx_{k,i},k,y_{k,i})\}_{i\in [n_k]}$ for each $k\in [K]$, where $n_k \defined |\calD_k|$. By assumption on the $n_k$, we have 
    \begin{equation} \label{eq:g large}
        g \defined \mathrm{gcd}\left( n_1,\cdots,n_K \right) \ge \frac{m-1}{m\epsilon - 1}.
    \end{equation}

    Now, let $\theta \defined \lfloor \epsilon g \rfloor/g$, and note that we obtain the integers $\mu_k\defined \theta \cdot n_k$ for $k\in [K]$, because $g$ divides each $n_k$ by definition of the gcd. Let $\lambda \defined \left\lceil \frac{1-\epsilon}{\theta} \right\rceil$. Using $\lfloor t \rfloor > t-1$, we note that
    \begin{equation} 
        \frac{1-\epsilon}{\theta} + 1 < \frac{1-\epsilon}{\epsilon- \frac{1}{g}} + 1 \le m,
    \end{equation}
    where the last inequality follows by assumption on $g$ being large enough~\eqref{eq:g large}. Thus, taking the ceiling of both sides above, we obtain $\lambda + 1 \le m$. Hence, it suffices to prove that there are $\lambda+1$ classifiers $\calR = \{h_j\}_{j\in [\lambda+1]}$ satisfying the claim in the proposition (i.e., with $\lambda+1$ replacing $m$). 

    It is straightforward to check that $\lambda \mu_k \le n_k$ for each $k$. We divide each $\calD_k$ into $\lambda$ sets of size $\mu_k$ each, and collect the remainder into a separate subset. Thus, for each $k\in [K]$, consider any partition $\calD_k=\bigcup_{j\in [\lambda+1]} \calD_k^j$ where $|\calD_k^j| = \mu_k$ for each $j\in [\lambda]$ and $|\calD_k^{\lambda+1}|=n_k-\lambda \mu_k$. 
    
    Next, we define the classifiers based on the partitions of the $\calD_k$. For each $j\in [\lambda+1]$, define the classifier $h_j$ over $\calD$ as follows. Fix $(\bx,k,y)\in \calD$. We set the value $h_j(\bx)$ to be
    \begin{align}
        h_j(\bx) \defined 
        \begin{cases} 
          1-y & \text{ if } (\bx,k,y) \in \calD_k^j, \\
          y & \text{ otherwise.}
       \end{cases}
    \end{align}
    Note that this makes $h_j$ well-defined by the second property assumed on $\calD$ at the beginning of this proof. We show that the set $\calR=\{h_j\}_{j\in [\lambda+1]}$ satisfies the desired properties in the proposition.

    \paragraph{Accuracy.} First, we show that each $h_j$ incurs a 0-1 loss less than $\epsilon$. Indeed, we have that, for each $j\in [\lambda]$,
    \begin{equation}
        \ell(h_j;\calD) = \frac{1}{n} \sum_{k\in [K]} |\calD_k^j| = \frac{1}{n} \sum_{k\in [K]} \mu_k = \theta,
    \end{equation}
    whereas the error for the last classifier is
    \begin{equation}
        \ell(h_{\lambda+1};\calD) =  \frac{1}{n} \sum_{k\in [K]} |\calD_k^{\lambda+1}| = \frac{1}{n} \sum_{k\in [K]} n_k - \lambda \mu_k = 1 - \lambda \theta.
    \end{equation}
    Thus, it suffices to check that $\theta,1-\lambda \theta \le \epsilon$. Recall that we set $\theta = \lfloor \epsilon g \rfloor/g$, hence $\theta \le \epsilon$ is immediate. Further, as we take $\lambda = \left\lceil \frac{1-\epsilon}{\theta} \right\rceil \ge \frac{1-\epsilon}{\theta}$, the inequality $1-\lambda \theta \le \epsilon$ follows immediately too. Hence, we have that $\ell(h_j;\calD)\le \epsilon$ for all $j\in [\lambda+1]$. 

    \paragraph{Fairness.} Next, we check that each $h_j$ satisfies Overall Accuracy Equality (OAE) perfectly. Let $\hat{Y}_j$ be the prediction of the classifier $h_j$, so $P_{\hat{Y}_j|\bX=\bx}(1)=h_j(\bx)$. Then, for $j\in [\lambda]$ the predictions of the $j$-th classifier satisfy
    \begin{equation}
        \Pr\left( \hat{Y}_j \neq Y \mid S=k \right) = \frac{|\calD_k^j|}{n_k} = \frac{\mu_k}{n_k} = \theta \qquad \text{ for every group } k\in [K],
    \end{equation}
    and similarly the last classifier satisfies
    \begin{equation}
        \Pr\left( \hat{Y}_{\lambda+1} \neq Y \mid S=k \right) = \frac{|\calD_k^{\lambda+1}|}{n_k} = \frac{n_k-\lambda \mu_k}{n_k} = 1-\lambda \theta \qquad \text{ for every group } k\in [K].
    \end{equation}
    Hence, the $h_j$, for $j\in [\lambda+1]$, all satisfy OAE perfectly.

    \paragraph{Ambiguity.} Finally, we show that the set $\calR= \{h_j\}_{j\in [\lambda+1]}$ exhibits full ambiguity. Note that we have a partition $\calD = \bigcup_{(k,j)\in [K]\times [\lambda+1]} \calD_k^j$. Fix a sample $(\bx,s,y)\in \calD$ and consider the index $(k,j)\in [K]\times [\lambda+1]$ of the unique part including it in the partition, i.e., $(\bx,s,y)\in \calD_k^j$. By construction of $\calR$, we have that $h_j(\bx) = 1-y$ (so $\hat{Y}_j=1-y$) but $h_{j'}(\bx) = y$ (so $\hat{Y}_{j'}=y$) for any $j'\neq j$. In other words, for \emph{every} fixed sample in $\calD$, there is a pair of classifiers in $\calR$ assigning it conflicting predictions. Thus, the ambiguity is $\alpha(\calD,\calR)=100\%$.

\subsection{Proof for the SP Metric}

Consider now the SP group-fairness constraint instead. In this case, we define we construct the following alternative dataset $\calD \subset \BR^d \times [K] \times \{0,1\}$. Let $t \defined \lceil 1/\epsilon \rceil \ge 2$. % and $\epsilon' \defined  1/t \le \epsilon$. 
For each $(k,y)\in [K]\times \{0,1\}$, fix any dataset $\calD_{k,y}=\{(\bx_{k,y,i},k,y)\}_{i\in [t]} \subset \BR^d \times \{k\} \times \{y\}$, let $\calD \defined \bigcup_{(k,y)\in [K]\times \{0,1\}} \calD_{k,y} \subset \BR^d\times [K]\times \{0,1\}$, and assume that if $\bx_{k,y,i}=\bx_{k',y',i'}$ then $y=y'$. Denote $n\defined |\calD| = 2Kt$. 

%Let $i^\star:[K]\times \{0,1\} \to [t]$ be any function. 
For each $(k,y)\in [K]\times \{0,1\}$, let $\sigma_{k,y}$ be a permutation on $[t]$, i.e., $\{\sigma_{k,y}(j)\}_{j\in [t]} = [t]$. Define the set of classifiers $\calR = \{h_j\}_{j\in [t]}$ on $\calD$ as follows. For each $j\in [t]$ and each $(k,y,i)\in [K]\times \{0,1\}\times [t]$, we set
    \begin{align}
        h_j(\bx_{k,y,i}) \defined 
        \begin{cases} 
          1-y & \text{ if } i=\sigma_{k,y}(j), \\
          y & \text{ otherwise.}
       \end{cases}
    \end{align}
We show that the set $\calR=\{h_j\}_{j\in [t]}$ satisfies the desired properties in the proposition.

\paragraph{Accuracy.} Consider first the 0-1 loss incurred by the $h_j$. We have that, for each $j\in [t]$,
\begin{equation}
    \ell(h_j;\calD) = \frac{1}{n} \sum_{(k,y,i) \in [K]\times \{0,1\}\times [t]} \indicator\{ i = \sigma_{k,y}(j) \} = \frac{1}{n} \sum_{(k,y)\in [K]\times \{0,1\}} 1 = \frac{2K}{2Kt} = \frac{1}{t}.
\end{equation}
Therefore, $\ell(h_j;\calD)=1/t =1/\lceil 1/\epsilon \rceil \le \epsilon$. In other words, $\calR$ is a realization of the $\epsilon$-level empirical Rashomon set of size $t$.

\paragraph{Fairness.} Next, we check that each $h_j$ satisfies SP perfectly. In other words, we check that $h_j$ assigns to the class $y=1$ the same percentage of samples across the groups $k\in [K]$. Indeed, this is true as we are switching the class memberships of exactly one sample from each of the $\calD_{k,y}$. In particular, $h_j$ assigns the class membership $1$ to exactly $1+(|\calD_{k,1}|-1)=|\calD_{k,1}|=t$ samples out of the total $|\calD_{k,0}\cup \calD_{k,1}|=2t$ samples belonging group $k$. As the ratio $t/(2t)=1/2$ is independent of the group $k$, we obtain the desired result that $h_j$ satisfies SP.

\paragraph{Ambiguity.} Finally, we check that the set $\calR=\{h_j\}_{j\in [t]}$ suffers from $100\%$ ambiguity. Fix a sample $(\bx_{k,y,i},k,y)\in \calD$, and we will show that there are two classifiers in $\calR$ assigning conflicting predictions to it. Indeed, let $j=\sigma_{k,y}^{-1}(i)$ and let $j'\in [t] \setminus \{j\}$ be any other index. Then, $h_j(\bx_{k,y,i})=1-y$, whereas $h_{j'}(\bx_{k,y,i})=y$. Therefore, the sample $(\bx_{k,y,i},k,y)$ contributes to the overall ambiguity $\alpha(\calD,\calR)$. As this is true for all samples in $\calD$, we conclude that $\alpha(\calD,\calR)=100\%$, as desired. This completes the proof of the proposition.

\section{Proofs of Section~\ref{sec:ensemble}} \label{sec:proofs of concentration}

\subsection{Proof of Theorem~\ref{Thm:concentration}}

    We assume that $\|\blambda\|_2^2, \|\bgamma\|_2^2\le c/m$ for an absolute constant $c$, e.g., we have $c=1$ for the uniform ensembling $\blambda=(1/m,\cdots,1/m)$ as then $\|\blambda\|_2^2=1/m$. Fix $\bx$, and denote the mean of the classifiers $\mu_{\bx}=\BE_{h\sim \calT(\calD)}[h(\bx)]$. The mapping $(h_1,\cdots,h_m)\mapsto \bh^{\ens,\blambda}$ satisfies the bounded-difference condition in the McDiarmid inequality. Indeed, changing $h_i$ can change $\bh^{\ens,\blambda}$ by at most $\lambda_i$. Furthermore, $\bh^{\ens,\blambda}(\bx)$ has the mean
    \begin{equation}
        \BE\left[ \bh^{\ens, \blambda}(\bx) \right] = \sum_{i\in [m]} \lambda_i \BE[h_i(\bx)] = \mu_{\bx} \sum_{i\in [m]} \lambda_i = \mu_{\bx}. 
    \end{equation}
    Hence, by Mcdiarmid's inequality, we have the bound
    \begin{equation}
        \BP\left( \left| \bh^{\ens,\blambda}(\bx) - \mu_{\bx} \right| \ge \nu \right) \le 2 \exp\left( \frac{-2\nu^2}{\sum_{i\in [m]} \lambda_i^2} \right) \le  2 \exp\left( \frac{-2\nu^2m}{c} \right).
    \end{equation}
    The same inequality holds for $\bgamma$ in place of $\blambda$: 
    \begin{equation}
        \BP\left( \left| \tilde \bh^{\ens,\bgamma}(\bx) - \mu_{\bx} \right| \ge \nu \right)  \le  2 \exp\left( \frac{-2\nu^2m}{c} \right).
    \end{equation}
    Therefore, we obtain the bound
    \begin{align}
        1- \BP\left( \left| \bh^{\ens,\blambda}(\bx) -  \tilde\bh^{\ens,\bgamma}(\bx) \right| \ge \nu \right) &= \BP\left( \left| \bh^{\ens,\blambda}(\bx) - \mu_{\bx} + \mu_{\bx} -  \tilde\bh^{\ens,\bgamma}(\bx) \right| < \nu \right) \\
        &\ge \BP\left( \left| \bh^{\ens,\blambda}(\bx) - \mu_{\bx}\right| + \left|  \tilde\bh^{\ens,\bgamma}(\bx) - \mu_{\bx} \right| < \nu \right) \\
        &\ge \BP\left( \left| \bh^{\ens,\blambda}(\bx) - \mu_{\bx}\right| < \frac{\nu}{2} \cap  \left|  \tilde\bh^{\ens,\bgamma}(\bx) - \mu_{\bx} \right| < \frac{\nu}{2} \right) \\
        &= 1-\BP\left( \left| \bh^{\ens,\blambda}(\bx) - \mu_{\bx}\right| \ge \frac{\nu}{2} \cup  \left|  \tilde\bh^{\ens,\bgamma}(\bx) - \mu_{\bx} \right| \ge \frac{\nu}{2} \right) \\
        &\ge 1 - 4 \exp\left( \frac{-\nu^2m}{2c} \right),
    \end{align}
    where the first inequality comes from triangle inequality, the following from probability of subset of events ($\mathbb{P}(A)\geq \mathbb{P}(B)$ if $A\supseteq B$), the equality from taking complement, and the last line from applying McDiarmid's inequality along with the union bound. 

    Finally, applying the union bound on $\calD_{\text{valid.}}$ with $|\calD_{\text{valid.}}|=n$, we obtain the bound
    \begin{align}
        \BP\left( \bigcap_{\bx \in \calD_{\text{valid.}}} \left| \bh^{\textup{ens},\blambda}(\bx) - \tilde{\bh}^{\textup{ens},\bgamma}(\bx) \right| < \nu \right) &= 1- \BP\left( \bigcup_{\bx \in \calD_{\text{valid.}}} \left| \bh^{\textup{ens},\blambda}(\bx) - \tilde{\bh}^{\textup{ens},\bgamma}(\bx) \right| \ge \nu \right) \\
        &\ge 1 - 4 n  \exp\left( \frac{-\nu^2m}{2c} \right),
    \end{align}
    and the proof is complete.

\subsection{Proof of Theorem~\ref{thm::score variation}}

    The main idea is as follows: first observe that for the ensembled labels to disagree on a sample (given that the scores are bounded away from $\frac{1}{2}$ with high probability), the two models need to produce scores in the range $[0,\frac{1}{2}-\delta]\cup[\frac{1}{2}+\delta,1]$. This means that the scores need to deviate at least $2\delta$ which has an exponentially low probability given Theorem 4.1.

    We will show that
    \begin{equation}
        \BP\left( \bigcup_{\bx \in\calD_0} f( \bh^{\ens, \blambda}(\bx)) \neq f(\tilde{\bh}^{\ens,\bgamma}(\bx)) \right) \le \left( 4 e^{-2\delta^2 m/c}+2 \theta \right)n_0.
    \end{equation}
    Indeed, for each fixed $\bx\in \calD_0$, we may reduce the failure probability to the case of separation of scores:
    \begin{align}
        \BP\left(  f( \bh^{\ens, \blambda}(\bx)) \neq f(\tilde{\bh}^{\ens,\bgamma}(\bx))  \right) &\le \BP{\Bigg (} \left( \bh^{\ens,\blambda}(\bx) \in \left[ 0, \frac12 - \delta \right] \cap  \tilde\bh^{\ens, \bgamma}(\bx) \in \left[ \frac12+\delta, 1 \right] \right) \\
        &\qquad \cup \left(  \tilde\bh^{\ens,\bgamma}(\bx) \in \left[ 0, \frac12 - \delta \right] \cap \bh^{\ens, \blambda}(\bx) \in \left[ \frac12+\delta, 1 \right] \right) \\
        &\qquad \cup \bh^{\ens,\blambda}(\bx) \in \left[ \frac12-\delta,\frac12+\delta\right] \\
        &\qquad \cup  \tilde\bh^{\ens,\bgamma}(\bx) \in \left[ \frac12-\delta,\frac12+\delta\right] {\Bigg )} \\
        &\le \BP\left( \left| \bh^{\ens, \blambda}(\bx) -  \tilde\bh^{\ens, \bgamma}(\bx) \right| \ge 2\delta \right) + 2\theta \\
        &\le 4 \exp\left( \frac{-2\delta^2m}{c}  \right) + 2\theta.
    \end{align}
    Finally, applying the union bound, we obtain that
    \begin{align}
        \BP\left( \bigcap_{\bx \in\calD_0} f( \bh^{\ens, \blambda}(\bx)) = f(\tilde{\bh}^{\ens,\bgamma}(\bx)) \right) &= 1- \BP\left( \bigcup_{\bx \in\calD_0} f( \bh^{\ens, \blambda}(\bx)) \neq f(\tilde{\bh}^{\ens,\bgamma}(\bx)) \right) \\
        &\ge 1- \left( 4 e^{-2\delta^2 m/c}+2 \theta \right)n_0,
    \end{align}
    and the proof is complete.

\section{Discussion on optimizing ensemble parameters}
\label{sec:: optimize param}

We have taken the weights $\blambda \in \bDelta_m$ which determines the ensembled model $\bh^{\ens,\blambda}$ to be fixed. We explain here how $\blambda$ can be optimized according to a given cost. Specifically, given a loss function $\ell : [0,1]\times \{0,1\}\to \BR_+$, we can search for the optimal $\blambda \in \bDelta_m$ that minimizes the total cost
\begin{equation} \label{eq:SLT population}
    L_{\ens}(\blambda) \defined \BE\left[ \ell(\bh^{\ens,\blambda}(X),Y\right) ].
\end{equation}
For the above optimization problem, we think of the constituent models $h_1,\cdots,h_m$ as being fixed and the randomness is from that of $(X,Y)$. 

However, in practice, we have access to only samples $(\bx_i,y_i) \sim P_{\bX,Y}$. Thus, we consider minimizing the regularized sample mean (for fixed $\beta>0$)
\begin{equation} \label{eq:SLT sample}
    \hat{L}_{\ens}(\blambda) \defined \frac{1}{n} \sum_{i\in [n]} \ell\left( \bh^{\ens,\blambda}(\bx_i), y_i \right) + \frac{\beta}{\sqrt{n}} \|\blambda\|_2^2.
\end{equation}
The $2$-norm regularization is added to facilitate proving convergence. 
This convergence result can be obtained via known results from statistical learning theory, e.g., using Theorem~13.2 in~\citep{hajek2019statistical}. Specifically, consider the following two restrictions:
\begin{itemize}
    \item Consider only $\blambda\in \bDelta_m$ satisfying $\|\blambda\|_2\le \alpha $ for a prescribed $\alpha$. Note that we may take $\alpha=1$ to encapsulate the whole probability simplex. However, we may choose $\blambda$ to be a slight modification of the uniform ensembling, in which case we would have $\alpha$ of order $1/\sqrt{m}$. 
    \item Assume that the function $\blambda \mapsto \ell(\bh^{\ens,\blambda}(\bx),y)$ is convex and $A$-Lipschitz for each fixed $(\bx,y)$.
\end{itemize}
In this case, choosing $\beta = A/\alpha$ and denoting the optimizers
\begin{equation}
    \blambda^{(n)} \defined \underset{\|\blambda\|_2\le \alpha}{\textup{argmin}} \ \hat{L}_{\ens}(\blambda),
\end{equation}
we can bound the utility of these minimizers by
\begin{equation}
    \BP\left( L_{\ens}(\blambda^{(n)}) \le \inf_{\|\blambda\|_2\le \alpha} \ L_{\ens}(\blambda) + \frac{\beta\alpha^2}{\sqrt{n}} \cdot \left( 1 + \frac{1}{\delta} + \frac{8}{\delta \sqrt{n}} \right) \right) \ge 1-\delta
\end{equation}
for any $\delta\in (0,1)$.

\section{Additional discussion on group fairness and fairness interventions}
\label{sec:: discussion fairness}
In addition to OAE (Definition \ref{def::OAE}), two other important fairness criteria are Statistical Parity \citep{dwork2015preserving} and Equalized Odds \citep{hardt2016equality}.
\begin{defn}[SP]
    $\Pr(\hat{Y} = 1 | S = s) = \Pr(\hat{Y} = 1 | S = s')$ for all groups $s,s'\in [K]$ . 
\end{defn}
\begin{defn}[EO]
    $\Pr(\hat{Y} = 1 | S = s, Y = b) = \Pr(\hat{Y} = 1 | S = s', Y=y)$ for all groups $s,s'\in [K]$, and binary labels $y\in \{0,1\}.$ 
\end{defn}

Essentially, SP requires the predicted label $\hat{Y} \triangleq \arg\max h(\bX)$ to be independent of the group membership $S$~\citep{10.1145/2090236.2090255}. In comparison, EO conditions on both group and the true label~\citep{hardt2016equality}. 
EO improves upon SP in the sense that it does not rule out the perfect classifiers whenever the true label $Y$ is correlated with the group membership $S$~\citep{agarwal2018reductions}. In practice, we quantify EO violation by measuring Mean EO as in Equation \ref{eq::EO} (for two groups) and, more generally, in Equation \ref{eq::EO violation multi} below (beyond two groups). Similarly, we can measure SP violation as in Equation \ref{eq::SP violation}.
\begin{equation} \label{eq::EO violation multi}
    \textsc{Mean EO} \defined \max_{s,s' \in [K]}\frac{1}{2}\left( |\textsc{TPR}_{S=s}-\textsc{TPR}_{S=s'}| +|\textsc{FPR}_{S=s}-\textsc{FPR}_{S=s'}| \right).
\end{equation}
\begin{equation} \label{eq::SP violation}
    \textsc{SP Violation} \defined \max_{s,s' \in [K]} \frac{1}{2}\left( |\Pr(\hat{Y} = 1 | S = s) -\Pr(\hat{Y} = 1 | S = s')| \right).
\end{equation}

Next, we offer a brief discussion of various intervention mechanisms used in this paper. The fairness interventions can be categorized into two categories: in-processing and post-processing. In-processing mechanisms incorporate fairness constraints during training. It usually add the fairness constraint to the loss function and outputs a fair classifier. Post-processing mechanisms treat the model as a black box and update its predictions to achieve the desirable fairness constraints~\citep{caton2020fairness}. 

\textsc{Reduction}~\citep{agarwal2018reductions}, short for exponentiated gradient reduction, is an in-processing technique that reduces fair classification to a sequence of cost-sensitive classification problems, and yields a randomized classifier with the lowest empirical error subject to the desired constraints. This technique achieves fairness with a minimal decrease in accuracy, but it is computationally expensive since it requires re-training multiple models.

\textsc{Reject option classifier}~\citep{kamiran2012decision} is a postprocessing technique that achieves fairness constraints by modifying outcomes of samples in a confidence band of the decision boundary with the highest uncertainty. It gives favorable outcomes to unprivileged groups and unfavorable outcomes to privileged groups. It outputs a thresholded prediction rather than a probability over the binary labels.

\textsc{EqOdds}~\citep{hardt2016equality} is a post-processing technique that formulates empirical risk minimization with fairness constraint as a linear program and modifies predictions according to the derived probabilities to achieve equalized odds.

\textsc{Fair Projection}~\citep{alghamdi2022beyond} is a post-processing technique that can accommodate fairness constraints in a setting with multiple labels and multiple groups. The fair model is obtained from `projecting' a pre-trained (and potentially unfair) classifier onto the set of models that satisfy target group-fairness requirements.

\section{Additional experiments and details on the experimental setup}
\label{sec:: additional experiments}
\begin{figure}[!tb]
\centering
\includegraphics[width=1.0\textwidth]{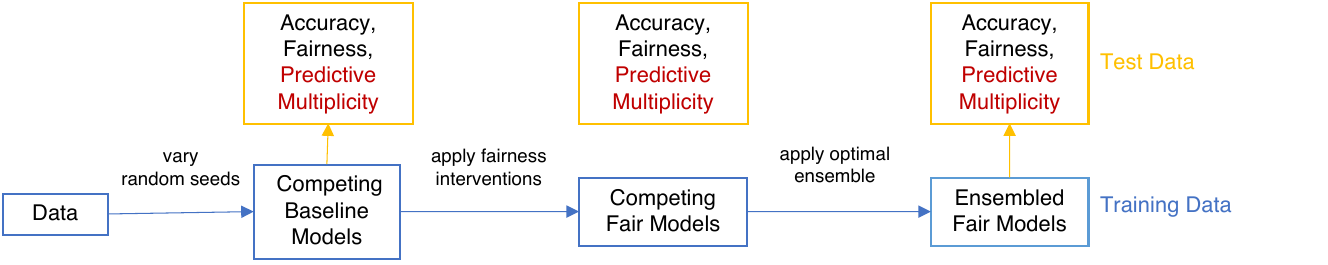}
 \caption{Flow chart of experimental procedure.}
 \label{fig::flowchart}
\end{figure}

Our proposed methodology can be summarized in the pipeline in Figure \ref{fig::flowchart}.

\subsection{Data}
The HSLS dataset~\citep{ingels2011high,jeong2022fairness} is an education dataset collected from 23,000+ students across high schools in the USA. Features of the dataset contain extensive information on students' demographic information, their parents' income and education level, schools' information, and students' academic performances across years. We apply the pre-processing techniques adopted by~\citet{alghamdi2022beyond}, with the number of samples reduced to 14,509. For the binary classification task with fairness constraints, the group attribute chosen is $\textsc{Race}\in \{\textsc{White},\textsc{Non-White}\}$  and the prediction label is students 9th-Grade $\textsc{GradeBin}\in\{0,1\}$, binarized according to whether a student's grade is higher or lower than the median. 

The ENEM dataset~\citep{cury2022instituto} is a Brazilian high school national exam dataset introduced by~\citet{alghamdi2022beyond}. It has 138 features containing students' demographic information, socio-economic questionnaire answers (e.g., parents' education level and if they own a computer), and students' exam scores. Adopting the preprocessing technique in~\citet{alghamdi2022beyond}, we sample 50K samples without replacement from the processed ENEM Year 2020 data. Identical to HSLS, the group attribute chosen is $\textsc{Race}\in \{\textsc{White},\textsc{Non-White}\}$  and the prediction label is students Grade binarized into  $\textsc{GradeBin}\in\{0,1\}$ according to whether a student's grade is higher or lower than the median. 

For the widely known Adult dataset~\citep{lichman2013uci}, also known as "Census Income" dataset, we choose the group attribute as  $\textsc{Sex}\in \{\textsc{Male},\textsc{Female}\}$ and predicted label to be $\textsc{Income}\in \{0,1\}$, where income bin denotes whether a person's income is higher or lower than 50K/yr.

\subsection{Competing Baseline Models}
We use the Scikit-learn implementation of logistic regression, gradient boosting, and random forest as baseline models. For logistic regression and gradient boosting, the default hyperparameter is used; for random forest, we set the number of trees and minimum number of samples per leaf to 10 to prevent over-fitting. To get 10 competing models for each hypothesis class, we use 10 random seeds (specifically 33--42).

In practice, the competing models, i.e., $h\in\hatRm$ can be obtained using different methodologies, such as sampling and adversarial weight perturbation~\citep{hsu2022rashomon,watson2022predictive}. We suggest one method for sampling. First, split the data into training, validation, and test dataset. We train a set of models by changing the randomized procedures in the training process, e.g., using different initializations, different cuts for cross-validation, data shuffling, etc. In this paper, we change the random seed feed into the baseline models to obtain competing models. We use the validation set to measure $\epsilon$ corresponding to this empirical Rashomon Set.

\subsection{Competing Fair Models}
For \textsc{EqOdds}, \textsc{Rejection}, and \textsc{Reduction}, we use the functions \textsc{EqOddsPostprocessing}, \textsc{RejectOptionClassification}, and \textsc{ExponentiatedGradientReduction} from AIF360 toolkits~\citep{bellamy2019ai}. For \textsc{Leveraging} and \textsc{Fair Projection}, we use the codes provided in the corresponding Github repositories of~\citet{chzhen2019leveraging} and~\citet{alghamdi2022beyond}.

\begin{figure}
    \centering
    \includegraphics[width=1\textwidth]{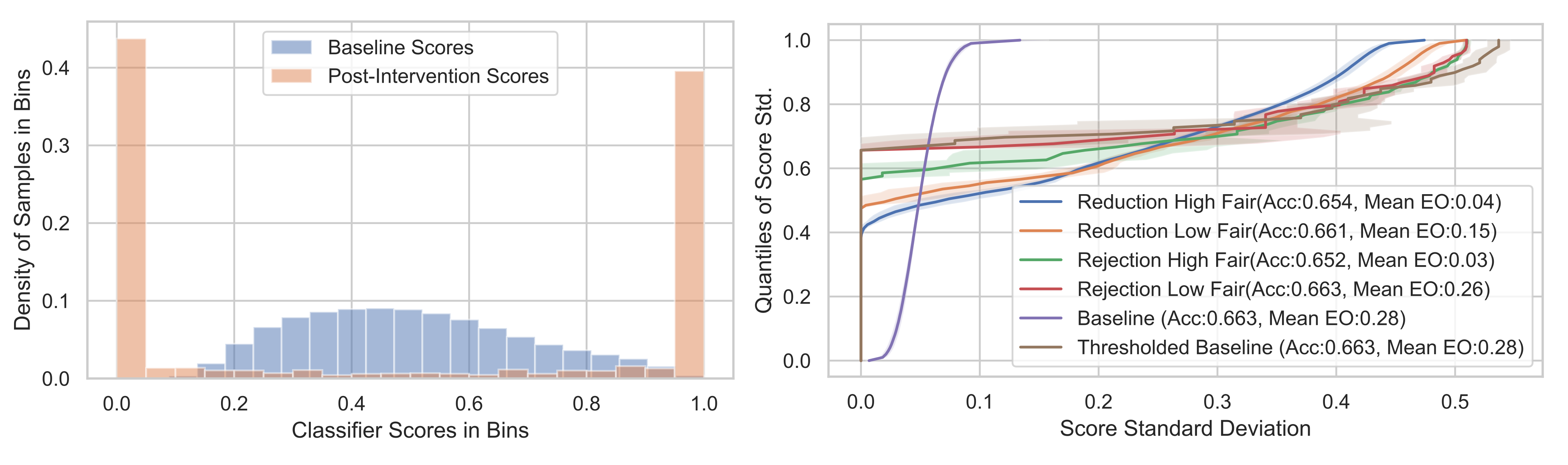}
    \caption{\textbf{Left}: We plot the score distribution of the \textsc{Reduction} approach as an example. The scores concentrate around 0 and 1, while scores of the baseline classifier are normally distributed. \textbf{Right}: Given this thresholded-like scores of fair classifiers, we include thresholed baseline in the quantile plot of score std.. The thresholded Baseline curve largely overlaps with Rejection curve with MEO 0.26 since they have the same MEO level. }
    \label{fig::score_distribution_thresholding}
\end{figure}
\begin{remark}
    We observe in practice that fairness classifiers are more confident and have scores that are thresholded-like (Figure \ref{fig::score_distribution_thresholding} \textbf{Left}). From the similarity in the shape of the thresholded baseline curve and the fair models’ curves (Figure \ref{fig::score_distribution_thresholding} \textbf{Right}), thresholding-like behavior of some interventions may explain some—but certainly not all (see Figure \ref{fig::Enem quantile} \textbf{Right})—increase in score std dev and the ensuing arbitrariness. Recall from the violin plot in Figure \ref{fig::Enem quantile} that the largest group (blue area) are those individuals for which std. increases from 0 to a large positive value (median around > 0.15). Hence, the blue area shows that significant arbitrariness is introduced by the fairness intervention, in addition to and separate from the effects of thresholding the baseline. 
\end{remark}

\subsection{Complete Experimental Plots}
In order to evaluate the predictive multiplicity of models with similar accuracy and fairness levels, we divide the accuracy-fairness frontier plots into 8x8 grids and put models in the corresponding bins. To compare the arbitrariness of models satisfying high/low fairness constraints, we select bins in two different MEO ranges and bins with baseline models. Then, we compute the standard deviation of scores of models corrected by \textsc{Reduction} in the three bins (high fairness/low fairness/baseline) and plot the quantile curves. We use \textsc{Pandas} package's quantile function with its default linear interpolation method.

Across three baseline model classes (random forest, gradient boosting, and logistic regression), fair models exhibit higher score arbitrariness. Especially at top quantiles, all fair models have standard deviations in scores going up to 0.5. This means that for the individuals at the top percentile, the model prediction can flip if another random seed is used in model training. 

\begin{figure}[!tb]
\centering
\includegraphics[width=1.0\textwidth]{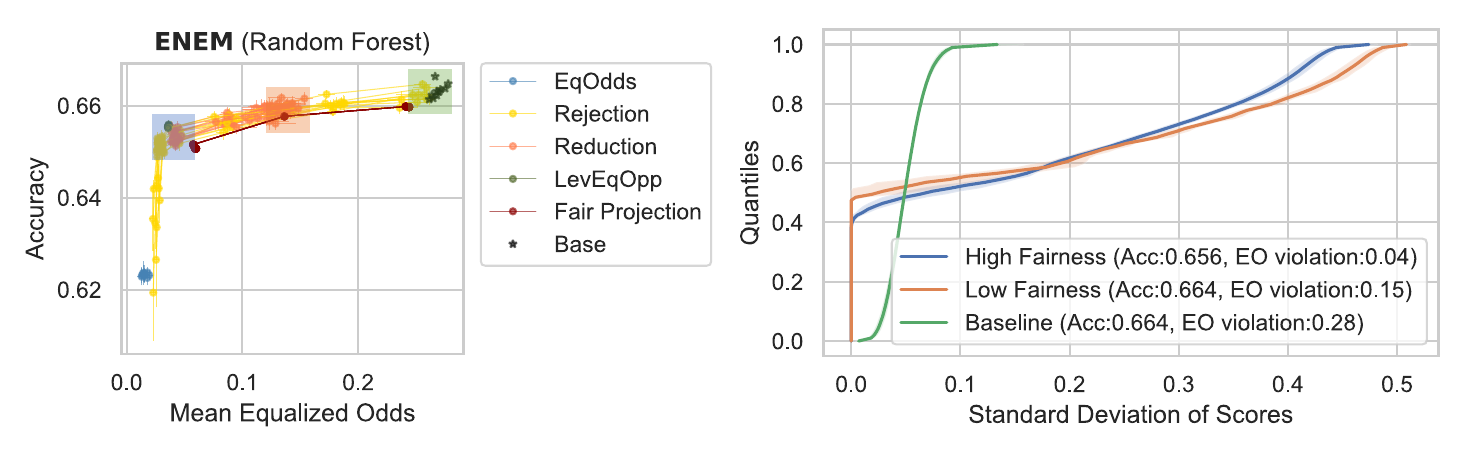}
\caption{\footnotesize \textbf{Left}: Accuracy-fairness curves of baseline random forest models v.s. fair models on the ENEM dataset.
\textbf{Right}: Quantiles of per-sample score std. across high/low fairness models and baseline.}
\label{fig::EnemRFHighLow}
\end{figure}

\begin{figure}[!tb]
\centering
\includegraphics[width=1.0\textwidth]{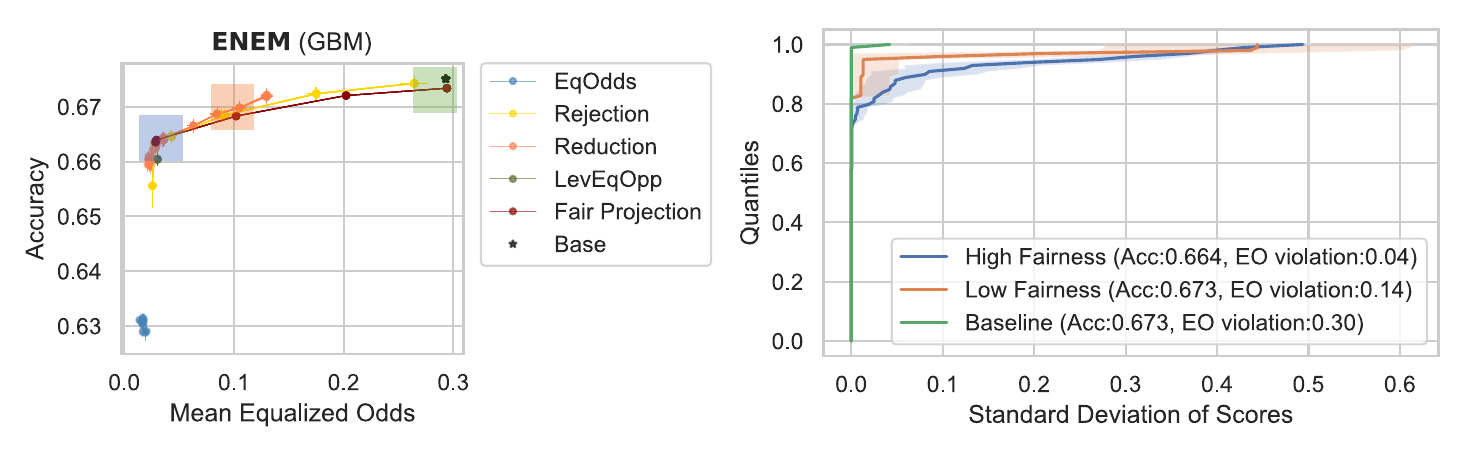}
\caption{\footnotesize \textbf{Left}: Accuracy-fairness curves of baseline gradient boosting models (GBM) v.s. fair models on the ENEM dataset.
\textbf{Right}: Quantiles of per-sample score std. across high/low fairness models and baseline.}
\end{figure}

\begin{figure}[!tb]
\centering
\includegraphics[width=1.0\textwidth]{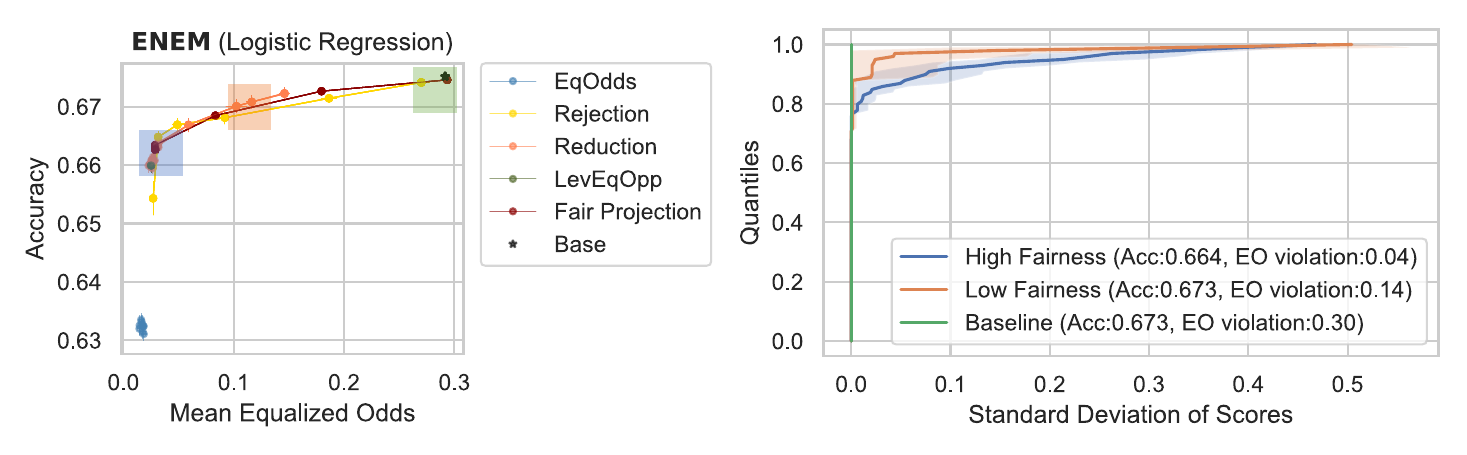}
\caption{\footnotesize \textbf{Left}: Accuracy-fairness curves of baseline logistic regression models v.s. fair models on the ENEM dataset.
\textbf{Right}: Quantiles of per-sample score std. across high/low fairness models and baseline.}
\end{figure}

\begin{figure}[!tb]
\centering
\includegraphics[width=1.0\textwidth]{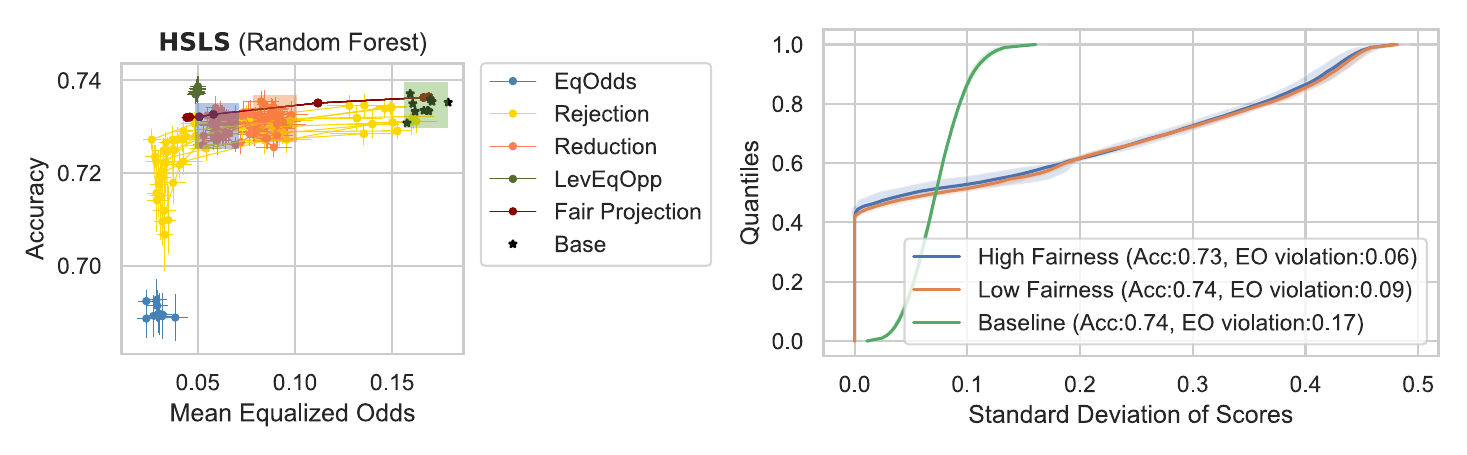}
\caption{\footnotesize \textbf{Left}: Accuracy-fairness curves of baseline random forest models v.s. fair models on the HSLS dataset.
\textbf{Right}: Quantiles of per-sample score std. across high/low fairness models and baseline.}
\end{figure}

\begin{figure}[!tb]
\centering
\includegraphics[width=1.0\textwidth]{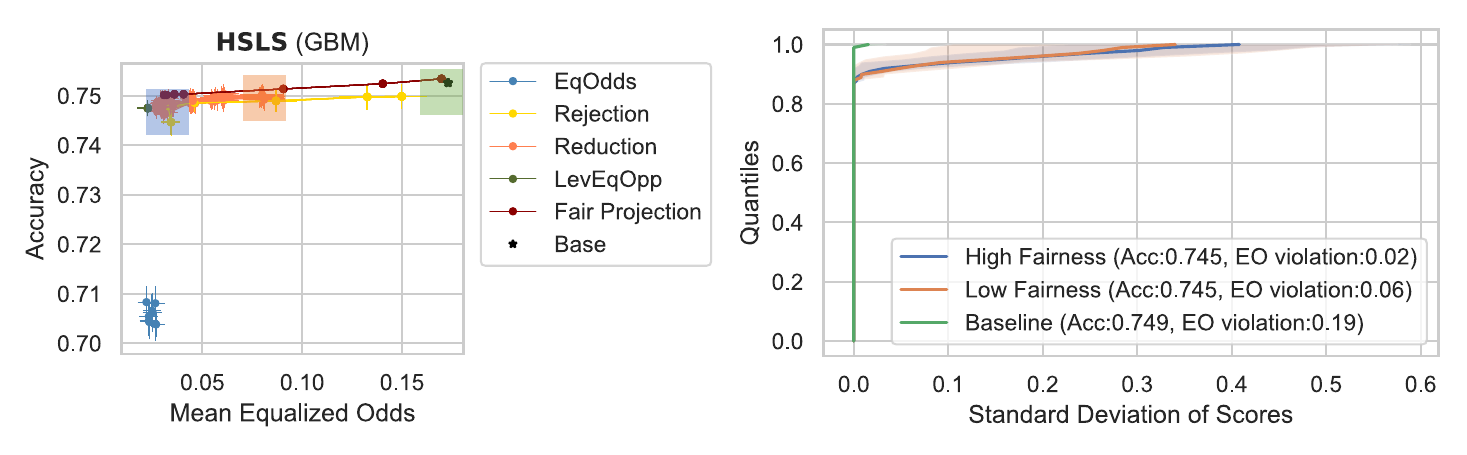}
\caption{\footnotesize \textbf{Left}: Accuracy-fairness curves of baseline gradient boosting models (GBM) v.s. fair models on the HSLS dataset.
\textbf{Right}: Quantiles of per-sample score std. across high/low fairness models and baseline.}
\end{figure}

\begin{figure}[!tb]
\centering
\includegraphics[width=1.0\textwidth]{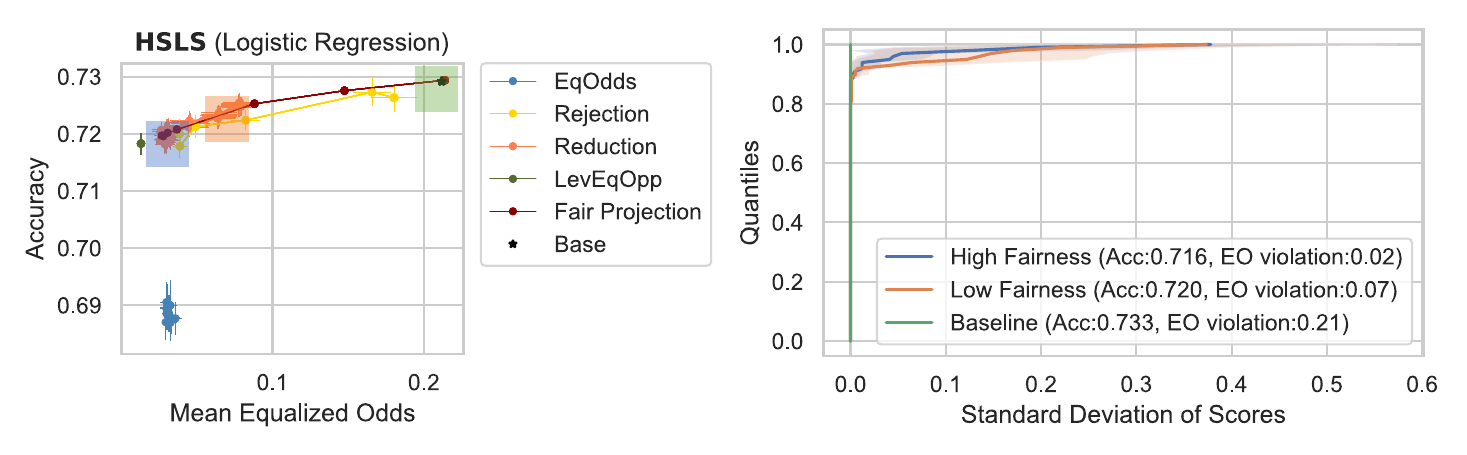}
\caption{\footnotesize \textbf{Left}: Accuracy-fairness curves of baseline logistic regression models v.s. fair models on the HSLS dataset.
\textbf{Right}: Quantiles of per-sample score std. across high/low fairness models and baseline.}
\end{figure}

\begin{figure}[!tb]
\centering
\includegraphics[width=1.0\textwidth]{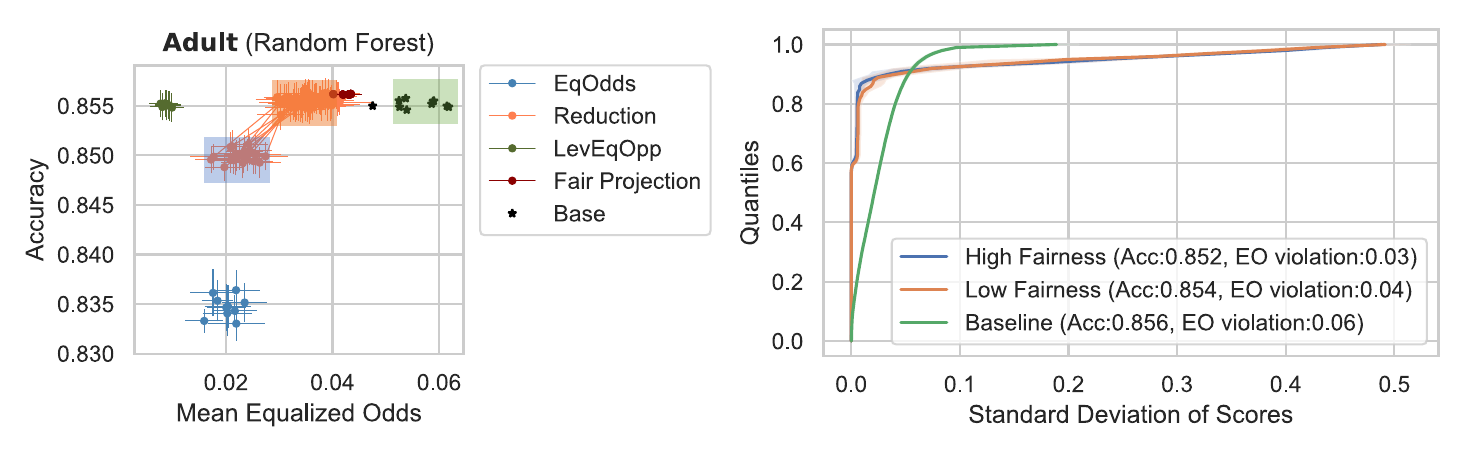}
\caption{\footnotesize \textbf{Left}: Accuracy-fairness curves of baseline random forest models v.s. fair models on the Adult dataset.
\textbf{Right}: Quantiles of per-sample score std. across high/low fairness models and baseline.}
\end{figure}

\begin{figure}[!tb]
\centering
\includegraphics[width=1.0\textwidth]{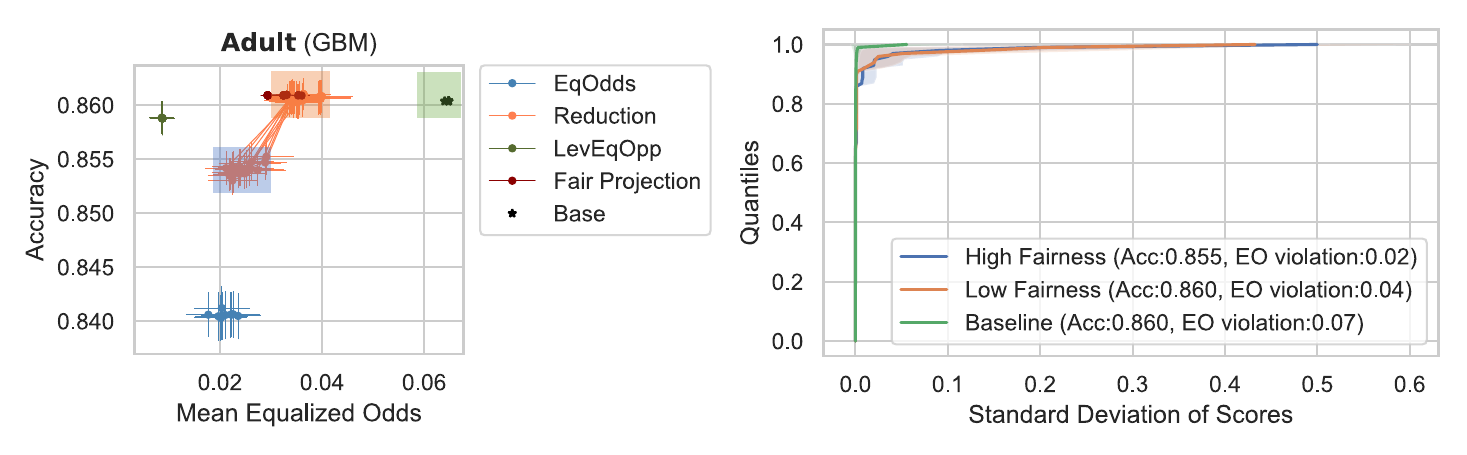}
\caption{\footnotesize \textbf{Left}: Accuracy-fairness curves of baseline gradient boosting models (GBM) v.s. fair models on the Adult dataset.
\textbf{Right}: Quantiles of per-sample score std. across high/low fairness models and baseline.}
\end{figure}

Furthermore, we evaluate the predictive multiplicity of models corrected by different fairness intervention methods. Across datasets, all fairness intervention methods exhibit maximal standard deviation of scores of 0.5 at top quantiles for random forest baseline methods. \textsc{Leveraging}~\citep{chzhen2019leveraging} exhibit score arbitrariness comparable to that of baseline for GBM and Logistic Regression methods. \textsc{Rejection} and \textsc{Leveraging} output thresholded scores directly, while \textsc{Reduction} outputs probabilities (with most scores close to 0 or 1). 

\begin{figure}[!tb]
\centering
\includegraphics[width=1.0\textwidth]{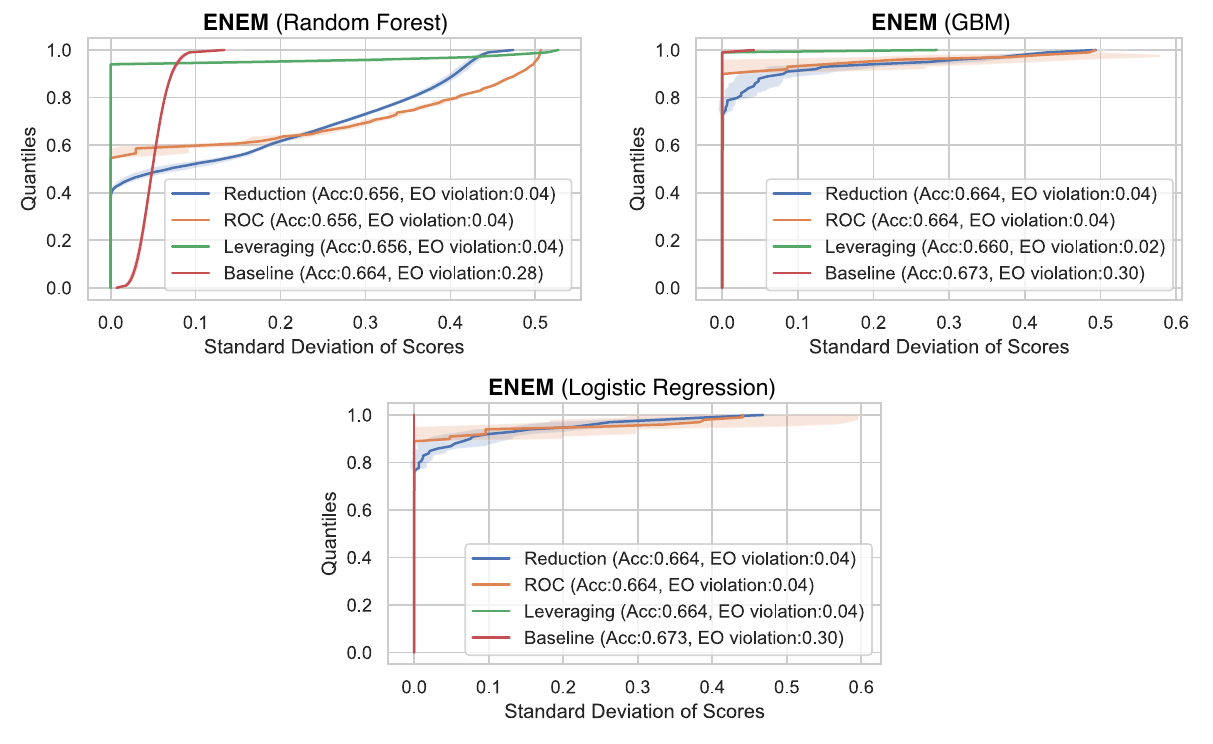}
\caption{\footnotesize
Quantile plot on models in high-fairness bin for various fairness interventions v.s. baseline models on ENEM. Fair models produce larger score std. at top percentiles compared to the baseline model.}
\end{figure}

\begin{figure}[!tb]
\centering
\includegraphics[width=1.0\textwidth]{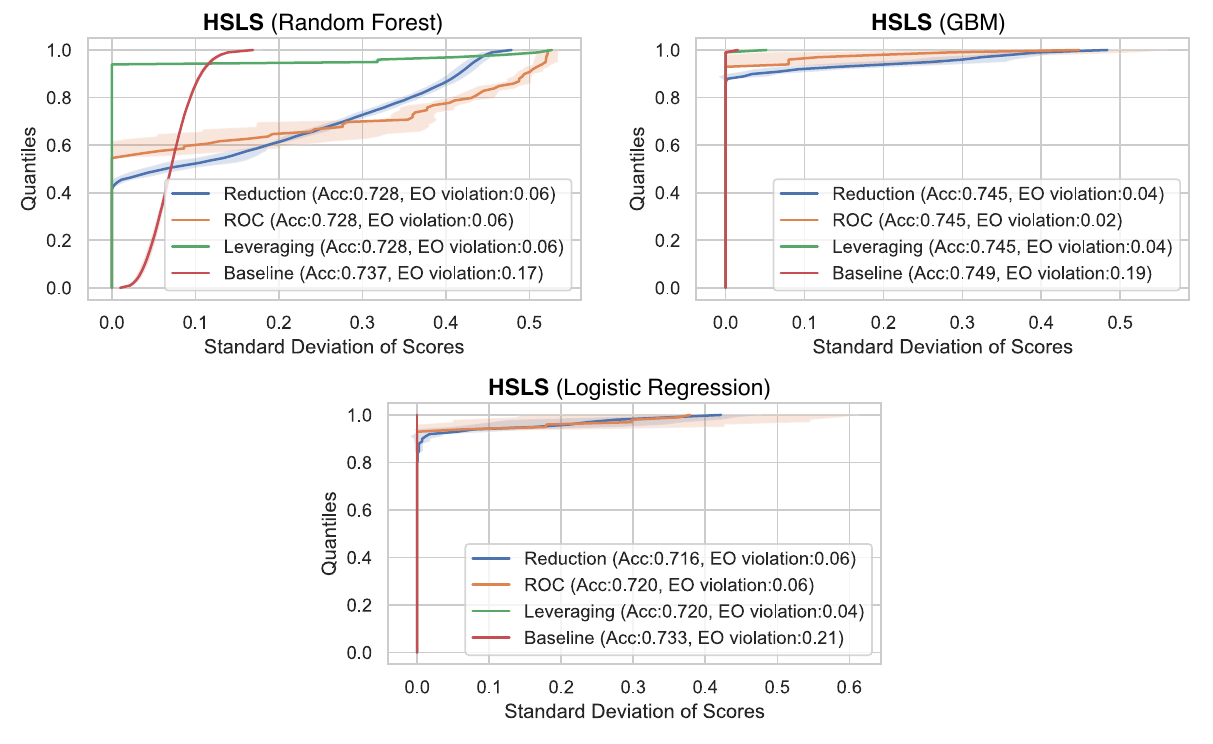}
\caption{\footnotesize
Quantile plot on models in high-fairness bin for various fairness interventions v.s. baseline models on HSLS. Fair models produce larger score std. at top percentiles compared to the baseline model.  }
\end{figure}

\chapter{Chapter~\ref{ch:3}} 
\label{apdx:proofs_ch3}

\section{Details of Theoretical Framework}
\label{appendix: QP}
From Equation~\eqref{eq:multiaccuracy optimization simplified}, we show that it is a quadratic program (QP).
To begin, writing $\calD = \{(\bx_j,y_j)\}_{j=1}^n$ and denoting
\begin{equation} \label{eq:QP constraint 0}
    \boldsymbol{f}_{\calD} =(f(\bx_1),\cdots,f(\bx_n))^T, \bg=(g(\bx_1),\cdots,g(\bx_n))^T,
\end{equation}
the objective function becomes the quadratic function $\frac12 \|\boldsymbol{f}_{\calD}- \bg\|_2^2$. Similarly, the constraint is a linear inequality in $\bg$, which we write as $\bA_{\calD_0,\calD} \ \bg \le \boldsymbol{b}_{\calD_0,\calD}$, 
where $\bA_{\calD_0,\calD} \in \BR^{(2n+2)\times n}$ and $\boldsymbol{b}_{{\calD_0,\calD}}\in \BR^{2n+2}$ are fixed and determined by $\calD_0$ and $\calD$ in view of equation~\eqref{eq:emp c star def} for the empirical witness function $c_{k,\calD_0,f}^\star$. Explicitly, denoting $\calD_0=\{(\tilde{\bx}_i,\tilde{y}_i)\}_{i=1}^m$, let us use the shorthands 
\begin{equation} \label{eq:QP constraint 1}
    \by_{\calD} = (y_1,\cdots,y_n)^T,  \bc_{\calD_0,\calD} = (c_{k,\calD_0,f}^\star(\bx_1),\cdots,c_{k,\calD_0,f}^\star(\bx_n))^T. %, \bK = (k(\bx_j,\tilde{\bx}_i))_{1\le i \le n; 1\le j \le m}. 
\end{equation}
Then, the multiaccuracy constraint in~\eqref{eq:multiaccuracy optimization simplified} can be written as $|\bc_{\calD_0,\calD}^T\bg/n - \bc_{\calD_0,\calD}^T \by_{\calD}/n |\le \alpha$. Taking the search space into consideration (i.e., $g$ evaluates to $[0,1]$), we see that~\eqref{eq:multiaccuracy optimization simplified} may be rewritten as the the following QP:
\begin{align}
    \underset{\bg \in \BR^n}{\text{minimize}} & \quad \frac12 \left\| \boldsymbol{f}_{\calD} - \bg \right\|_2^2 \label{eq:multiaccuracy optimization QP} \\
    \text{subject to} & \quad \bA_{\calD_0,\calD} \ \bg   \le \boldsymbol{b}_{\calD_0,\calD}, \nonumber
\end{align}
where we define the constraint's matrix and vector by
\begin{equation} \label{eq:QP constraint 2}
    \bA_{\calD_0,\calD} \define \begin{pmatrix}
        \bc_{\calD_0,\calD}^T/n \\
        -\bc_{\calD_0,\calD}^T/n \\
        \bI_n \\
        -\bI_n
    \end{pmatrix}, 
    \boldsymbol{b}_{\calD_0,\calD} \define \begin{pmatrix}
        \alpha +\bc_{\calD_0,\calD}^T\by/n \\
        \alpha - \bc_{\calD_0,\calD}^T\by/n \\
        \boldsymbol{1}_n \\
        \boldsymbol{0}_n
    \end{pmatrix}.
\end{equation}

Note that the witness function for a kernel $k:\calX^2\to \BR$, dataset $\calD_0=\{(\tilde{\bx}_j,\tilde{y}_j)\}_{j\in [m]}$, and predictor $g:\calX\to [0,1]$ is given by
\begin{equation}
    c_{k,\calD_0,g}^\star (\bx) = \frac{\theta_{k,\calD_0,g}}{m} (\tilde{\bg}-\tilde{\by})^T \tilde{\bk}(\bx),
\end{equation}
\begin{equation}
    c_{k,\calD_0,g}^\star (\bx) = \frac{(\tilde{\bg}-\tilde{\by})^T \tilde{\bk}(\bx)}{\sqrt{(\tilde{\bg}-\tilde{\by})^T \tilde{\bK}(\tilde{\bg}-\tilde{\by})}}
\end{equation}
where $\tilde{\bk}:\calX\to \BR^m$ is the vector-valued function defined by $\tilde{\bk}(\bx)\define (k(\bx,\tilde{\bx}_j))_{j\in [m]}$, $\tilde{\bg}\define (g(\tilde{\bx}_j))_{j\in [m]}$ and $\tilde{\by}\define (\tilde{y}_j)_{j\in [m]}$ are fixed vectors, and $\theta_{k,\calD_0,g}$ is a normalizing constant that is unique up to sign. We may compute $\theta_{k,\calD_0,g}$ by setting $\|c_{k,\calD_0,g}^\star \|_{\calH_k}=1$, namely, we have 
\begin{equation}
    \theta_{k,\calD_0,g}^2 = \frac{m^2}{(\tilde{\bg}-\tilde{\by})^T \tilde{\bK} (\tilde{\bg}-\tilde{\by})},
\end{equation}
where $\tilde{\bK}\define (k(\tilde{\bx}_i,\tilde{\bx}_j))_{i,j\in [m]}$ is a fixed matrix. Thus, the multiaccuracy constraint becomes
\begin{equation}
    \left| \overline{\bh}^T \overline{\bK} \tilde{\bh} \right| \le n \alpha \tau_{\tilde{\bh}}
\end{equation}
where $\tau_{\tilde{\bh}} \define \sqrt{\tilde{\bh}^T \tilde{\bK} \tilde{\bh}}$ and $\bh = (\overline{\bh}^T,\tilde{\bh}^T)^T$. With $\bh = \bg - \by$ and $\br = \boldsymbol{f} - \by$, the objective becomes $\frac{1}{2n}\| \br - \bh \|_2^2$. At each iteration of $\tau$, check if $\tau \le \tau_{\tilde{\bh}} $.

We may compute the KME of a predictor $g$ with respect to class $\frakC_k$ and dataset $\calD_1=\{(\bx_i,y_i)\}_{i\in [n]}$ via the equation 
\begin{equation}
\label{eq::normalization_factor}
    \textup{KME}(k,\calD_1,\calD_0,g) = \frac{1}{n} \cdot \frac{\left|(\bg - \by)^T\bK (\tilde{\bg}-\tilde{\by})\right|}{\sqrt{(\tilde{\bg}-\tilde{\by})^T \tilde{\bK}(\tilde{\bg}-\tilde{\by})}},
\end{equation}
where $\tilde{\bg},\tilde{\by}, \tilde{\bK}$ are computed on $\calD_0$ as above, $\bg \define (g(\bx_i))_{i\in [n]}$ and $\by\define (y_i)_{i\in [n]}$ are fixed vectors, and $\bK \define (k(\bx_i,\tilde{\bx}_j))_{(i,j)\in [|\calD_1|]\times [|\calD_0|]}$ is a fixed matrix. Note that if $\calD_0$ is used for computing $c_{k,\calD_0,f}^\star$ for a given predictor $f$ and then $g$ is obtained using $c_{k,\calD_0,f}^\star$ (so $\calD_0$ was used for deriving $g$), then one should report $\textup{KME}(k,\calD_1,\calD_0',g)$ for a freshly sampled $\calD_0'$ at the testing phase.

\section{Complete Experimental Results}
\label{appendix:experiments}

%leave ablation to the appendix to declutter the plot
\textbf{Ablation.} As we have presented evidence that isotonic calibration plus \KMultiAcc~can be an effective post-processing method, for the purpose of ablation we now analyze isotonic calibration being applied directly to the baseline classifier. We note that isotonic calibration tends to maintain an equivalent or higher AUC because the monotonic function preserves ranking of the samples up to tie breaking (which rarely has an influence) \cite{niculescu2005predicting}. Our ablation method frequently achieves a similar or better MSCE than \LSBoost{} (as discussed, the baseline plus isotonic calibration achieves a MSCE less than $.02$ in all benchmarks, while \LSBoost~only achieves this in $23$ of $40$ benchmarks), and a better average MSCE than \KMultiAcc~alone in all benchmarks. However, isotonic calibration alone has a significantly lower KME than \KMultiAcc~in $20$ of $40$ benchmarchs, confirming the utility of an algorithm targeting optimizing for multiaccuracy error as well.

\begin{figure*}[htbp]
     \centering
     \resizebox{\linewidth}{!}{\begin{tabular}{cccc}
    \includegraphics[width=0.2\textwidth]{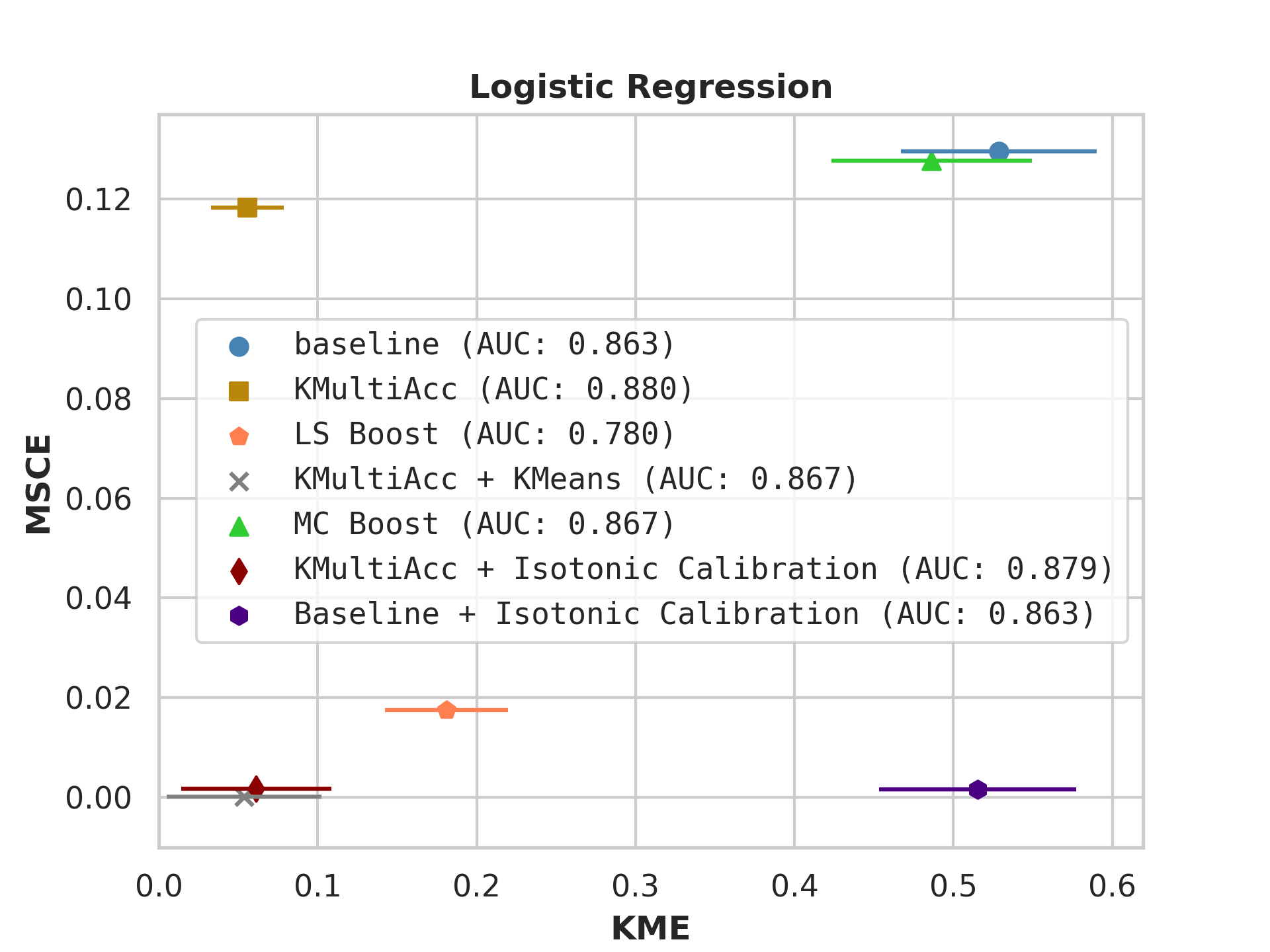} & 

     \includegraphics[width=0.2\textwidth]{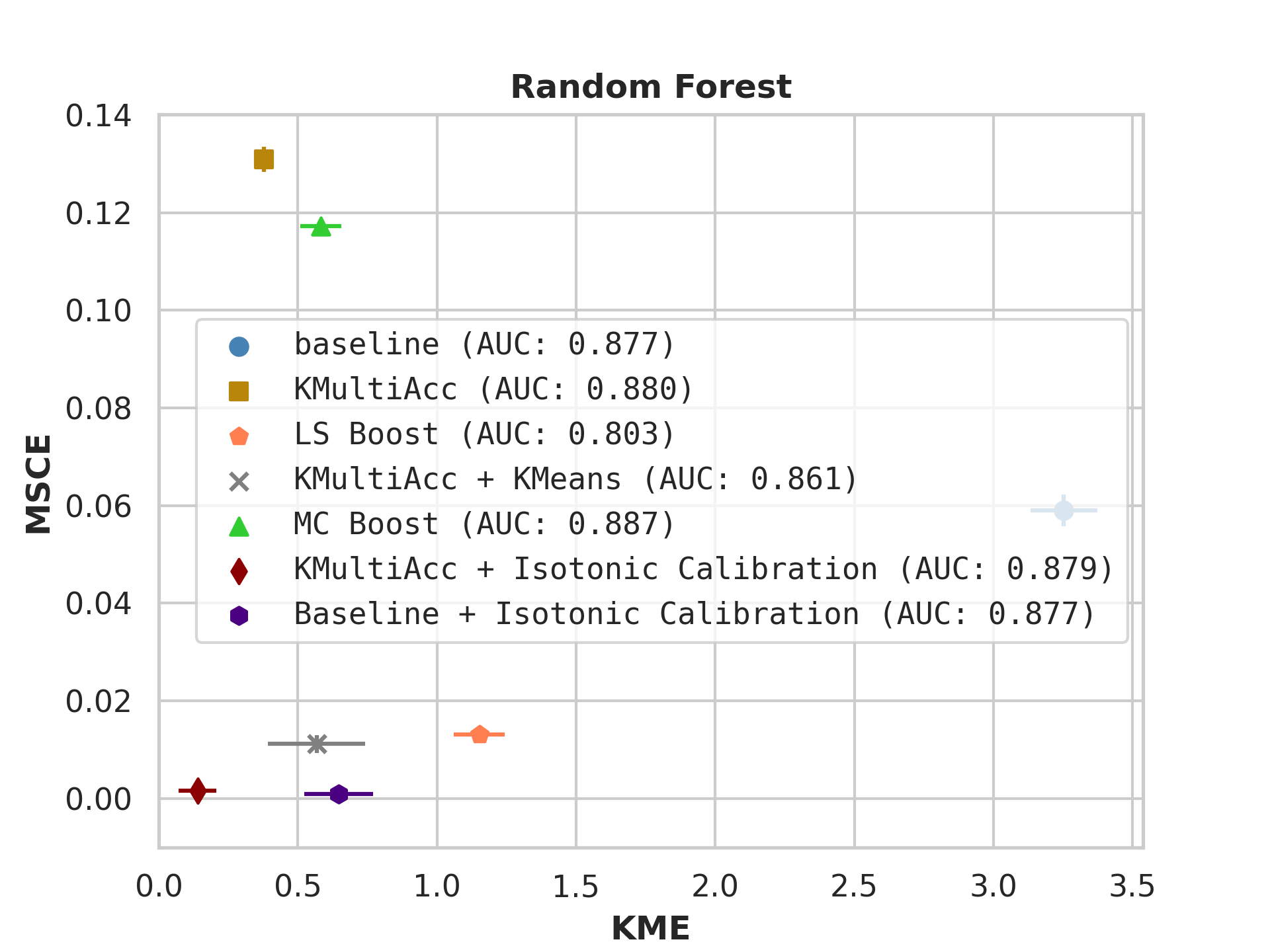}  &
    \includegraphics[width=0.2\textwidth]{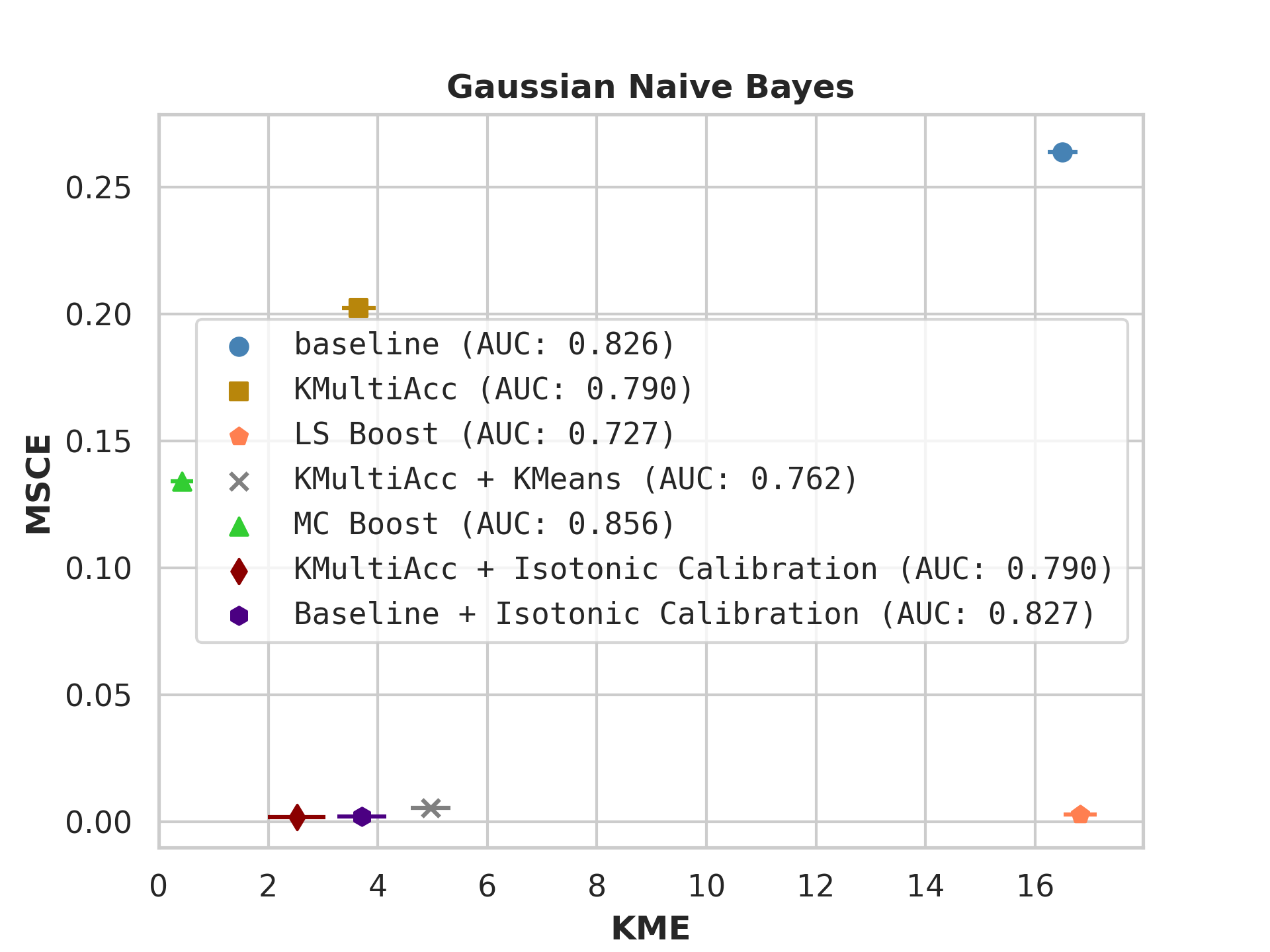}&
    \includegraphics[width=0.2\textwidth]{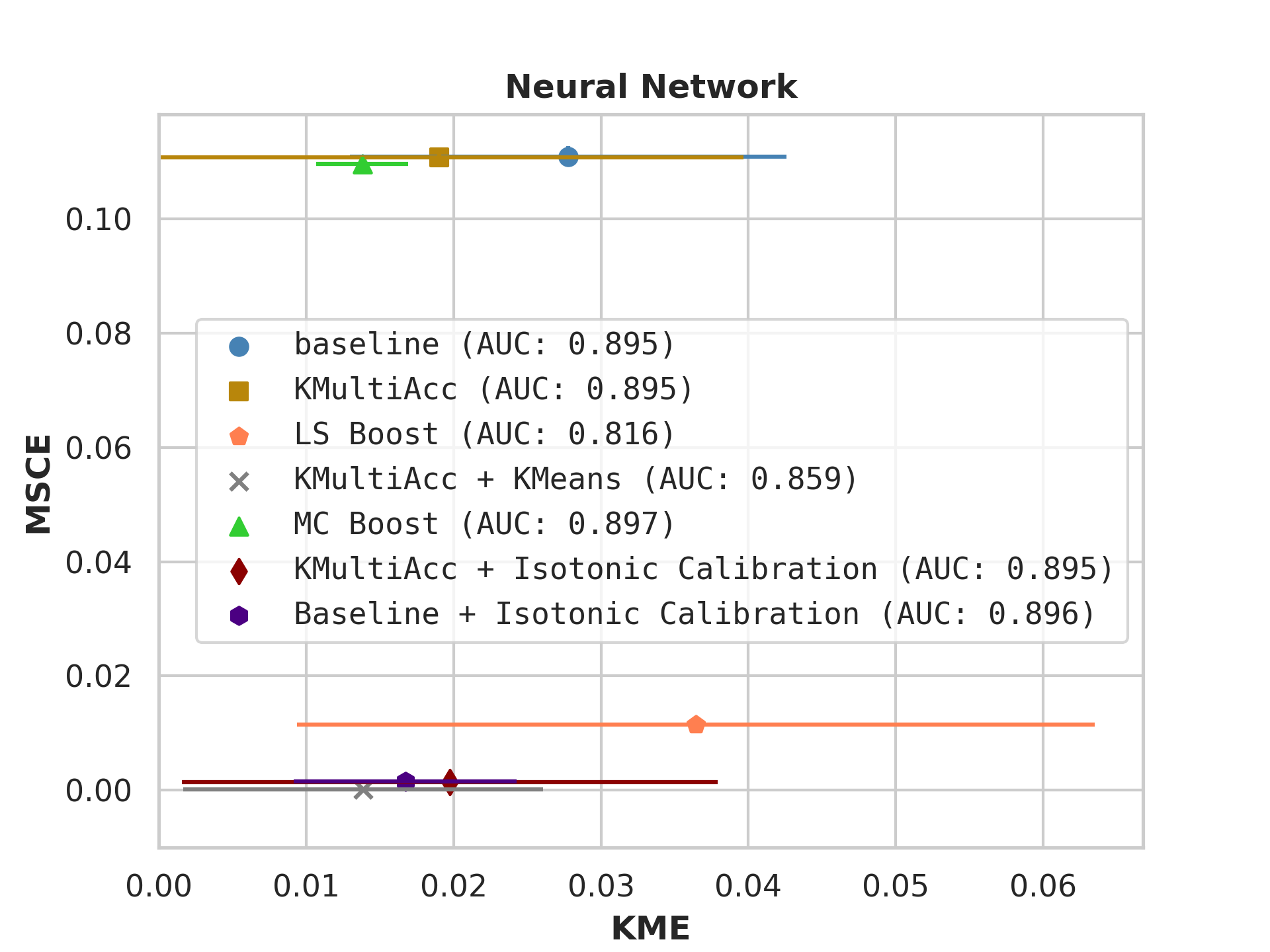} 
   
     \end{tabular}}
            \caption{The Folktables Employment Task with data from the state of Alabama.} \label{fig::emp_AL}
\end{figure*}

\begin{figure*}[htbp]
     \centering
     \resizebox{\linewidth}{!}{\begin{tabular}{cccc}
    
    \includegraphics[width=0.2\textwidth]{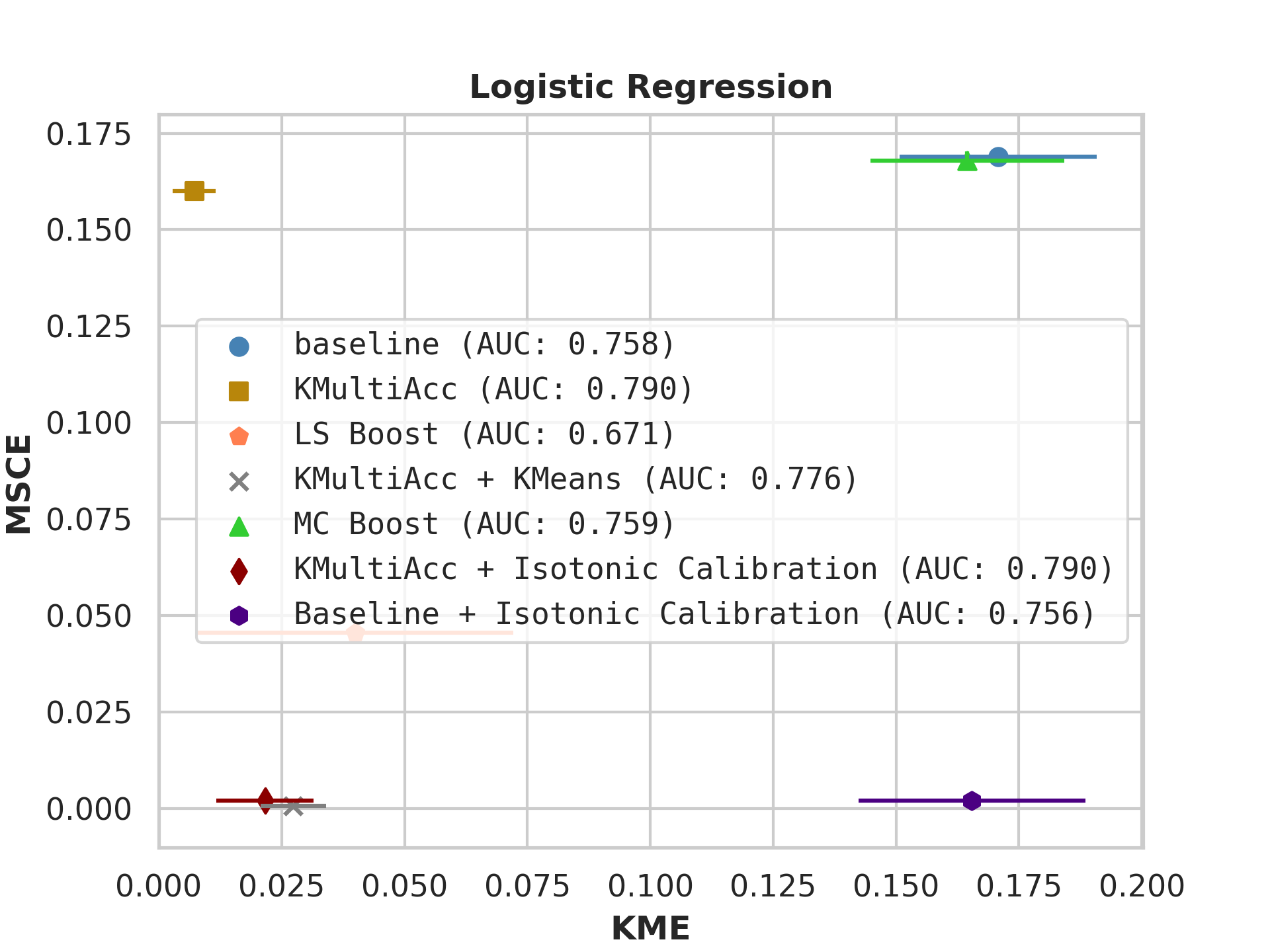} &

    \includegraphics[width=0.2\textwidth]{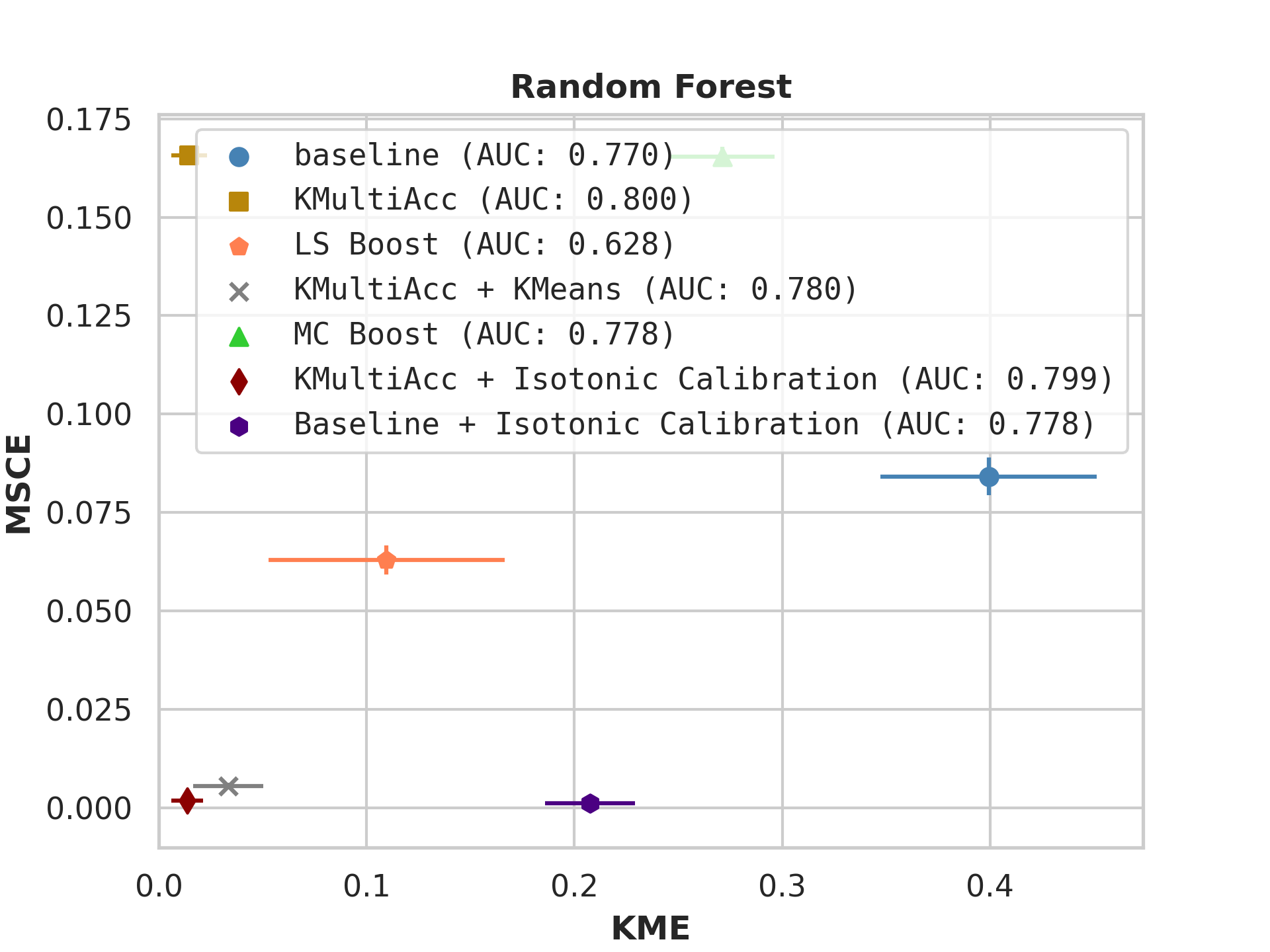}  &

    \includegraphics[width=0.2\textwidth]{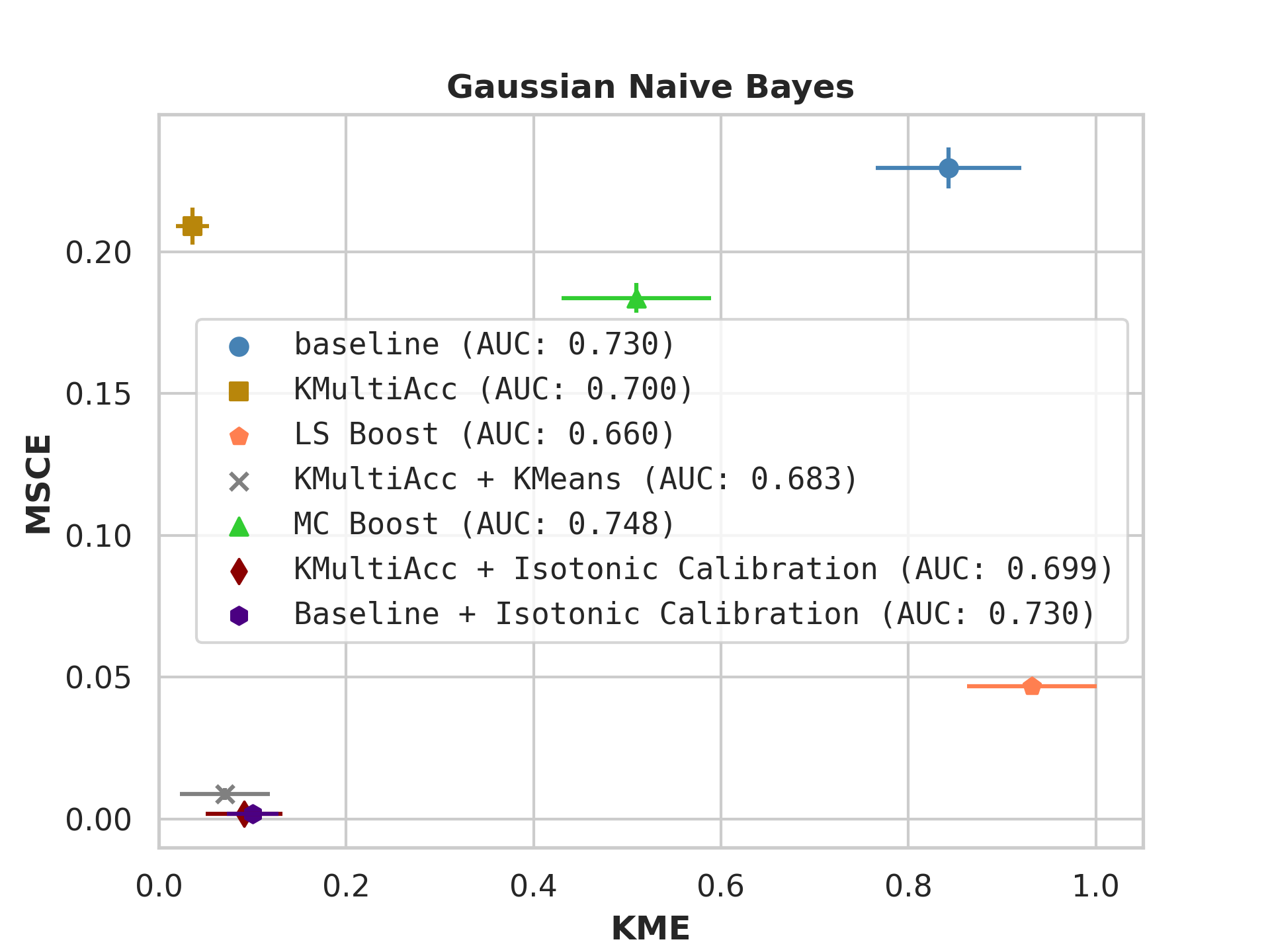}&

    \includegraphics[width=0.2\textwidth]{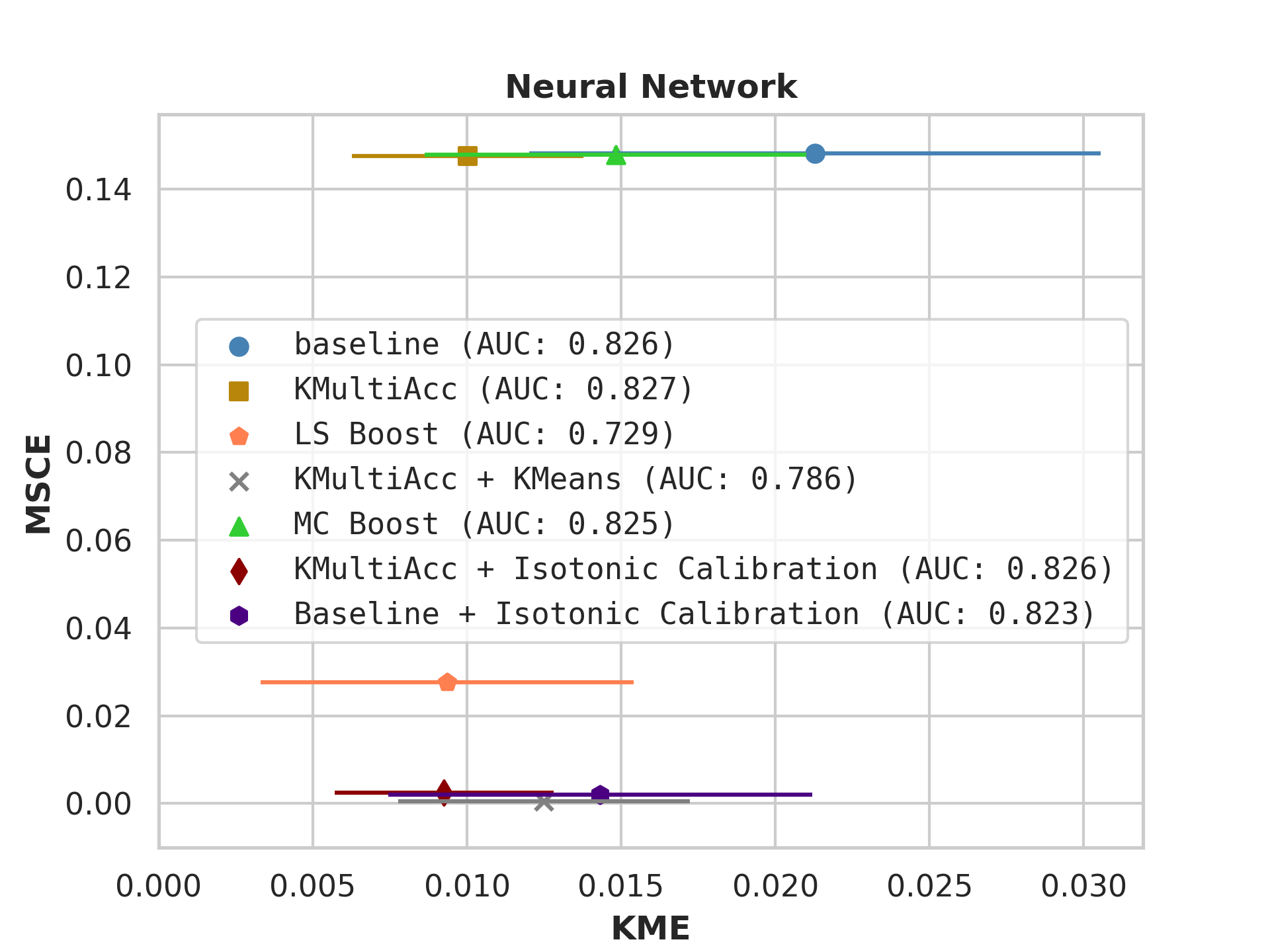}

     \end{tabular}}
      \caption{The Folktables Health Task with data from the state of Ohio.} \label{fig::Health_OH}
\end{figure*}

\begin{figure*}[htbp]
     \centering
     \resizebox{\linewidth}{!}{\begin{tabular}{cccc}
    \includegraphics[width=0.2\textwidth]{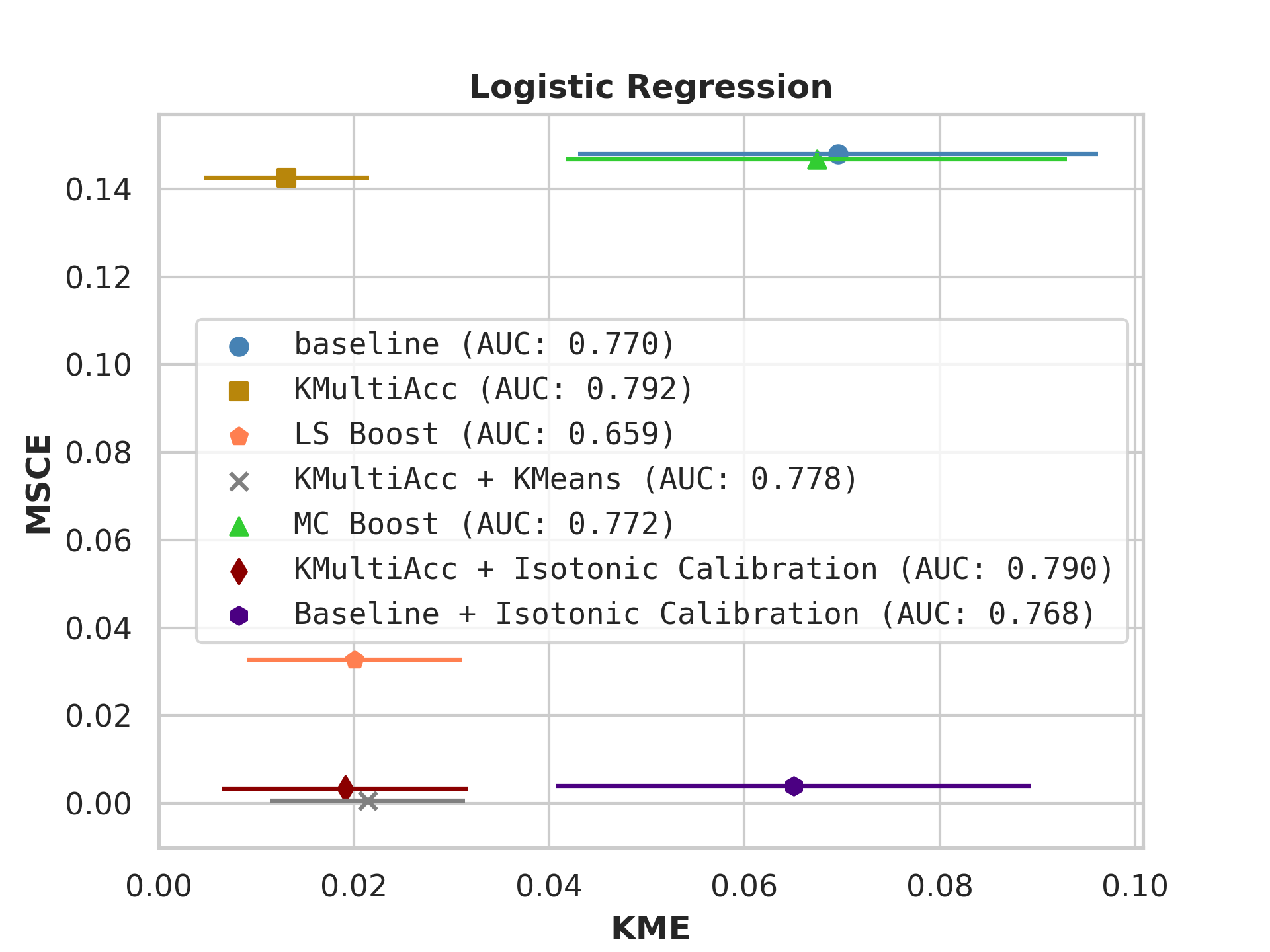} &

    \includegraphics[width=0.2\textwidth]{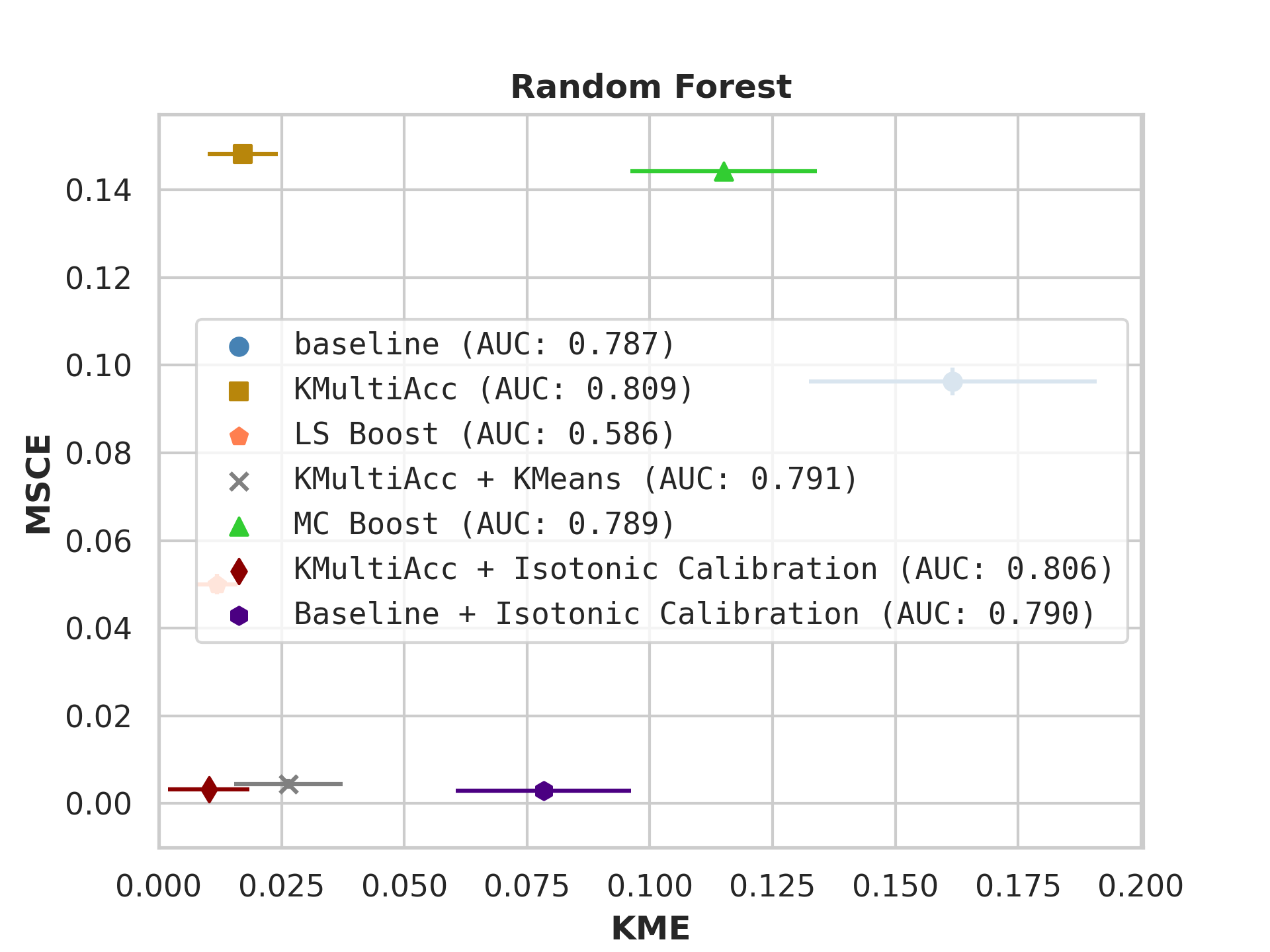}  &

    \includegraphics[width=0.2\textwidth]{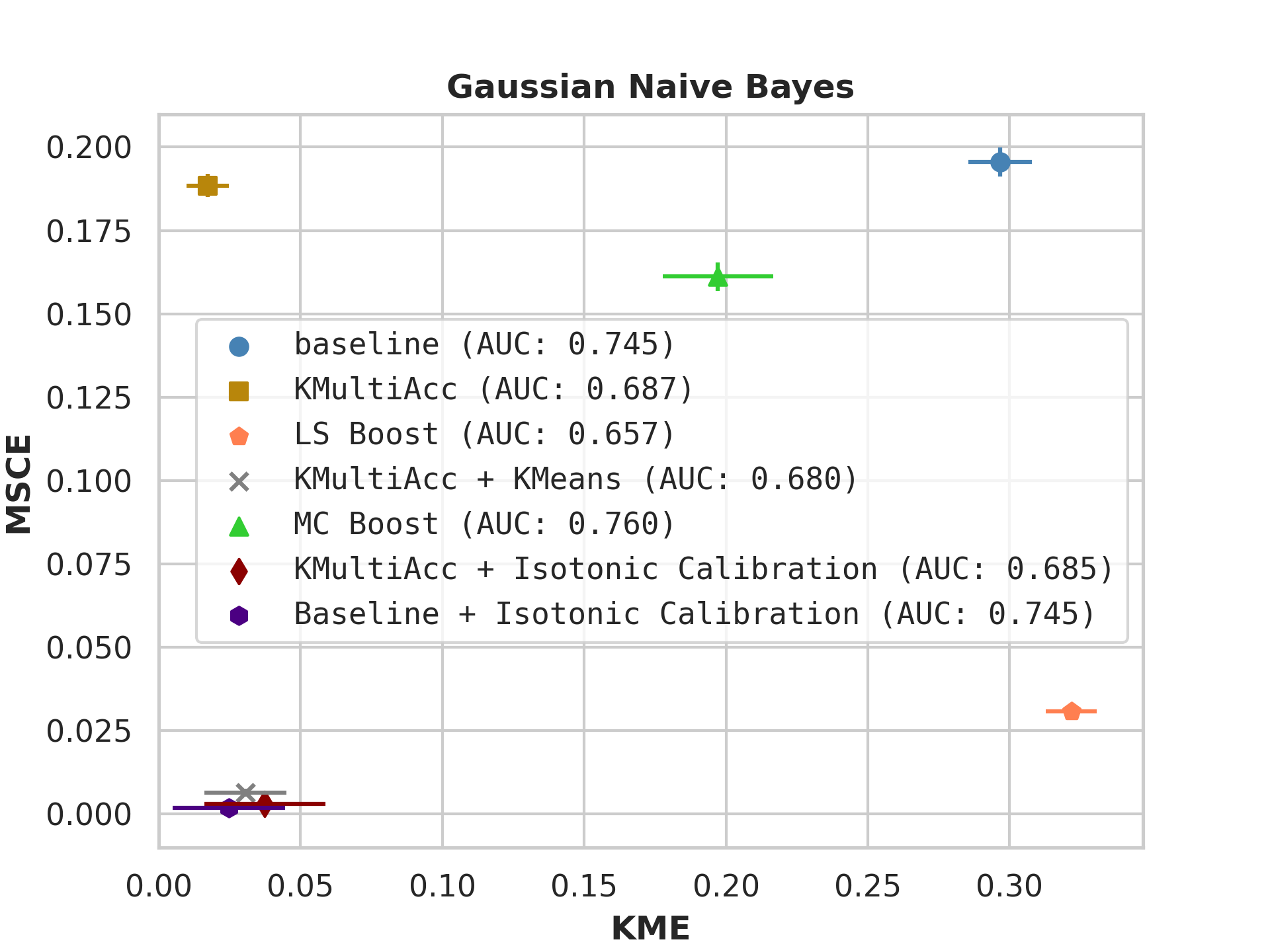}&

    \includegraphics[width=0.2\textwidth]{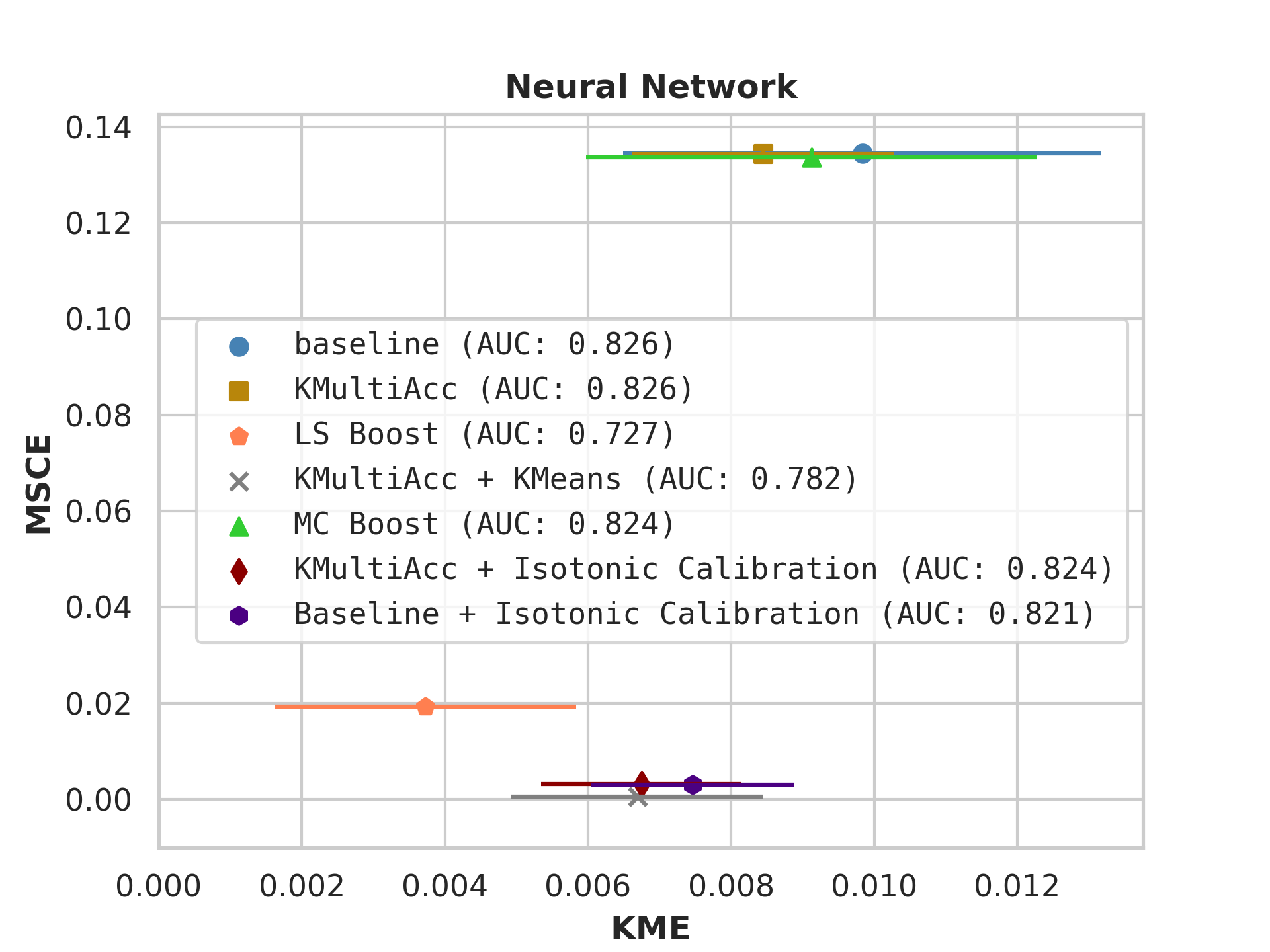} 
   
     \end{tabular}}
            \caption{The Folktables Health Task with data from the state of Wisconsin.} \label{fig::Health_WI}
\end{figure*}

\begin{figure*}[htbp]
     \centering
     \resizebox{\linewidth}{!}{\begin{tabular}{cccc}
    
 \includegraphics[width=0.45\textwidth]{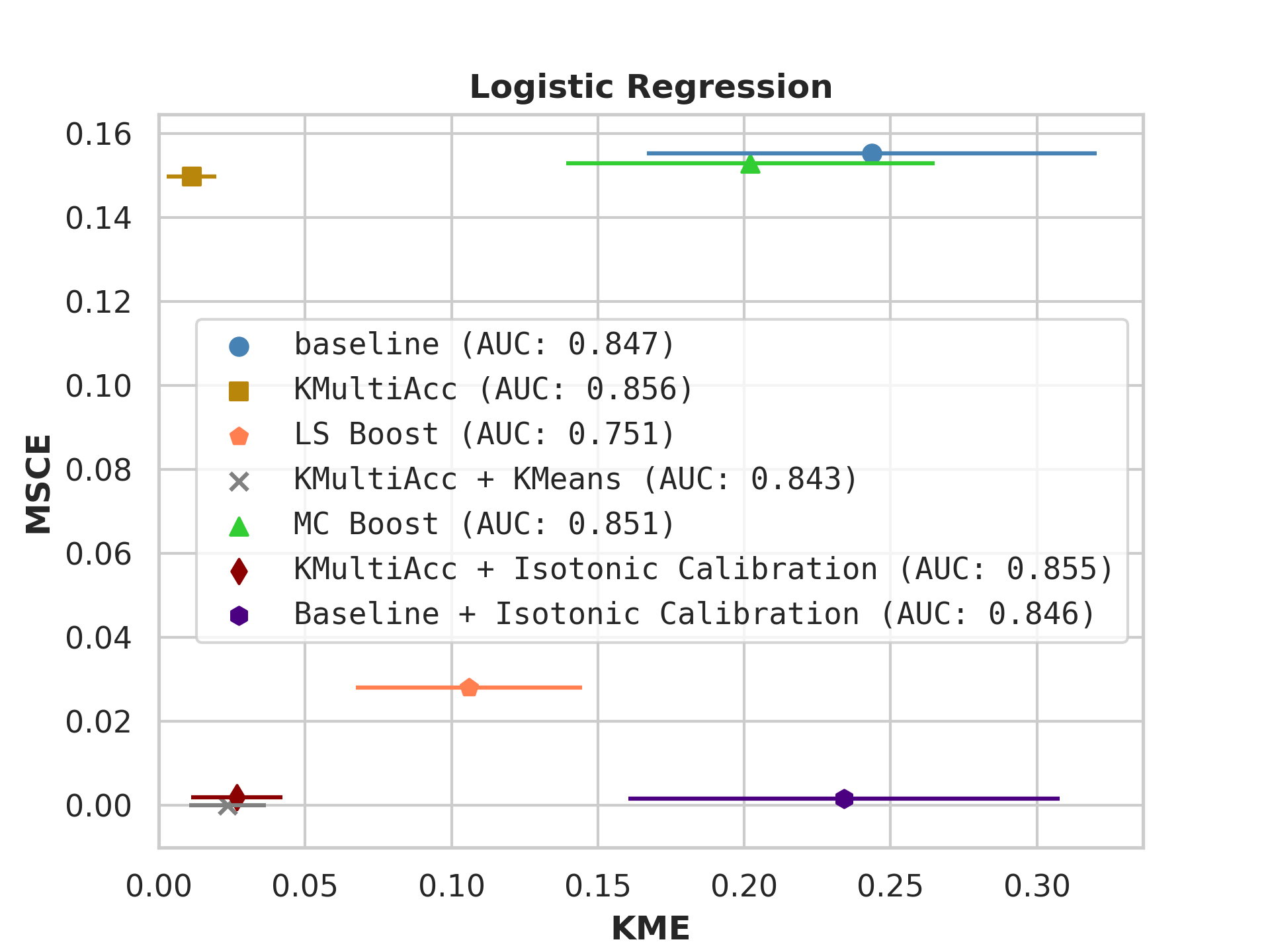} &

    \includegraphics[width=0.45\textwidth]{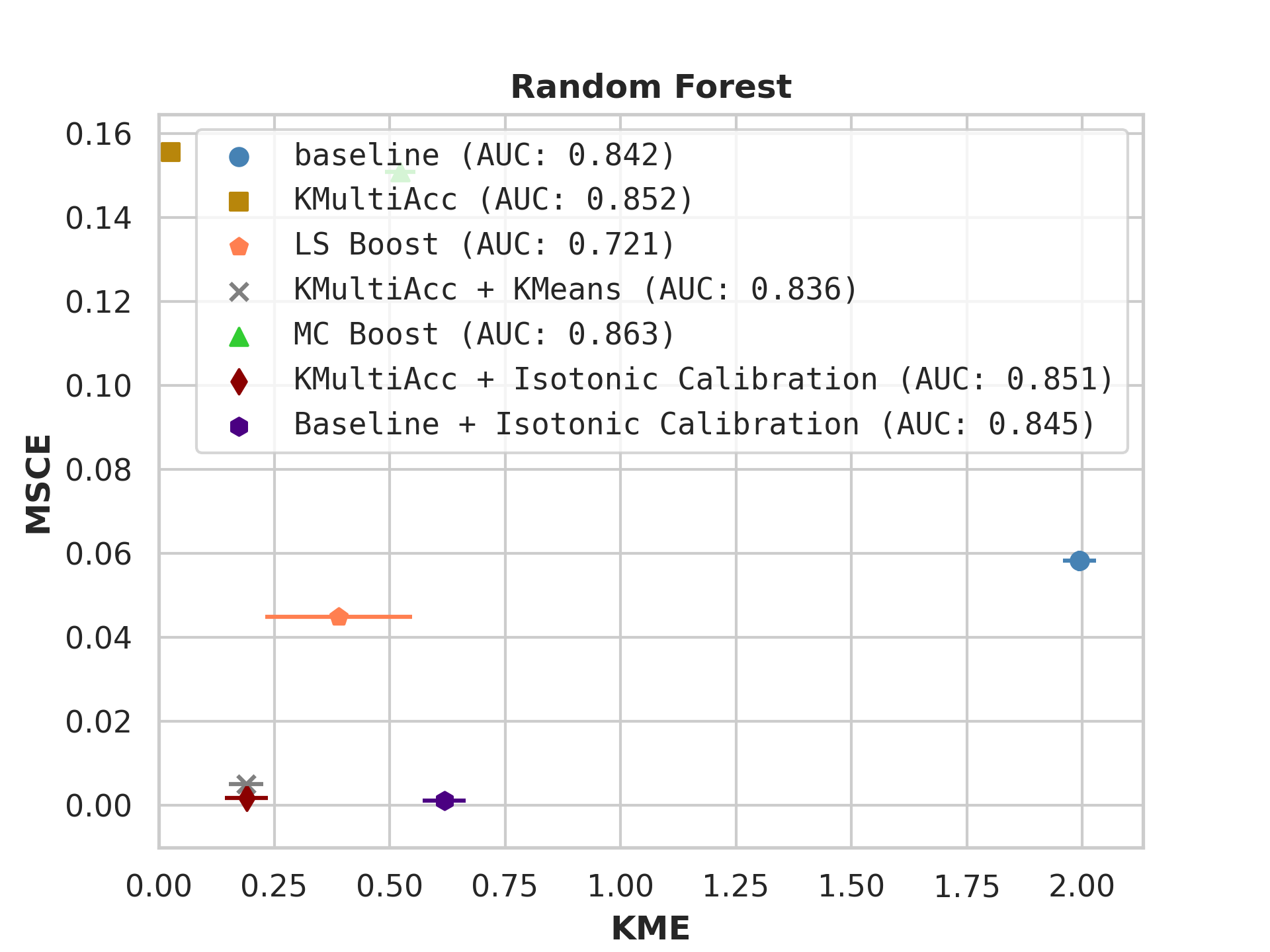}   &

    \includegraphics[width=0.45\textwidth]{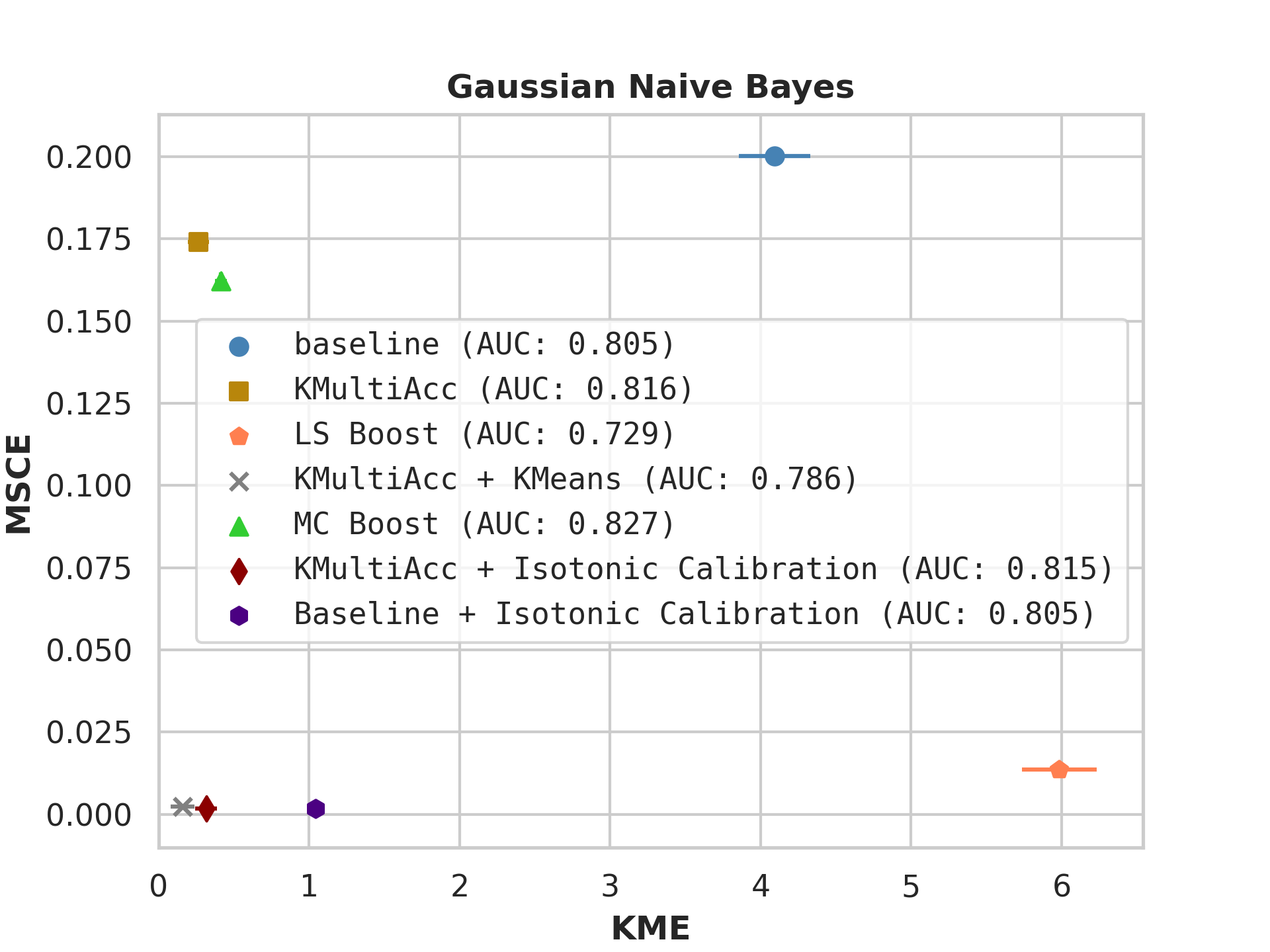}&
    
    \includegraphics[width=0.45\textwidth]{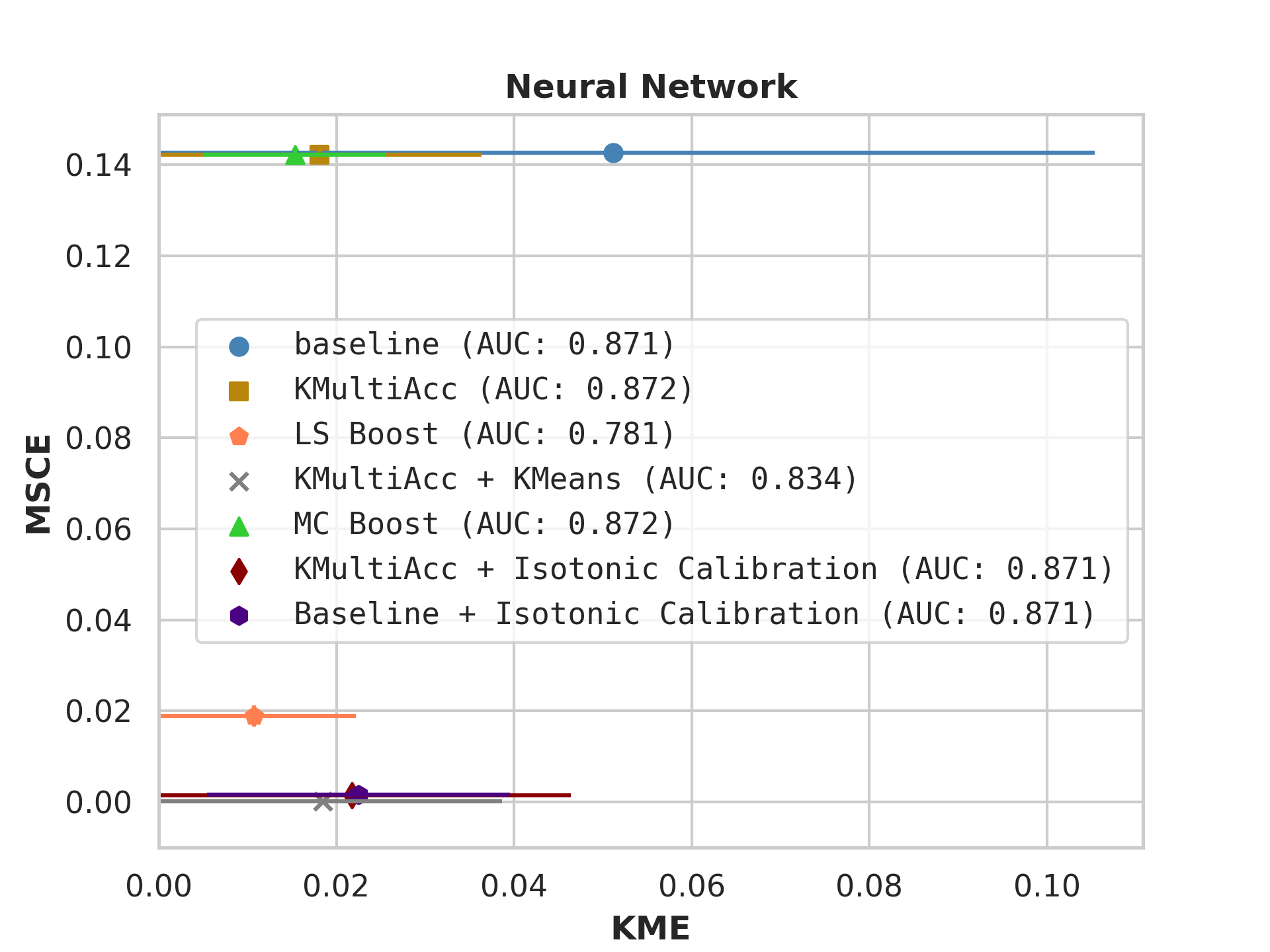}

     \end{tabular}}
      \caption{The Folktables Income Task with data from the state of Illinois.} \label{fig::Income_IL}
\end{figure*}

\begin{figure*}[htbp]
     \centering
     \resizebox{\linewidth}{!}{\begin{tabular}{cccc}
    \includegraphics[width=0.45\textwidth]{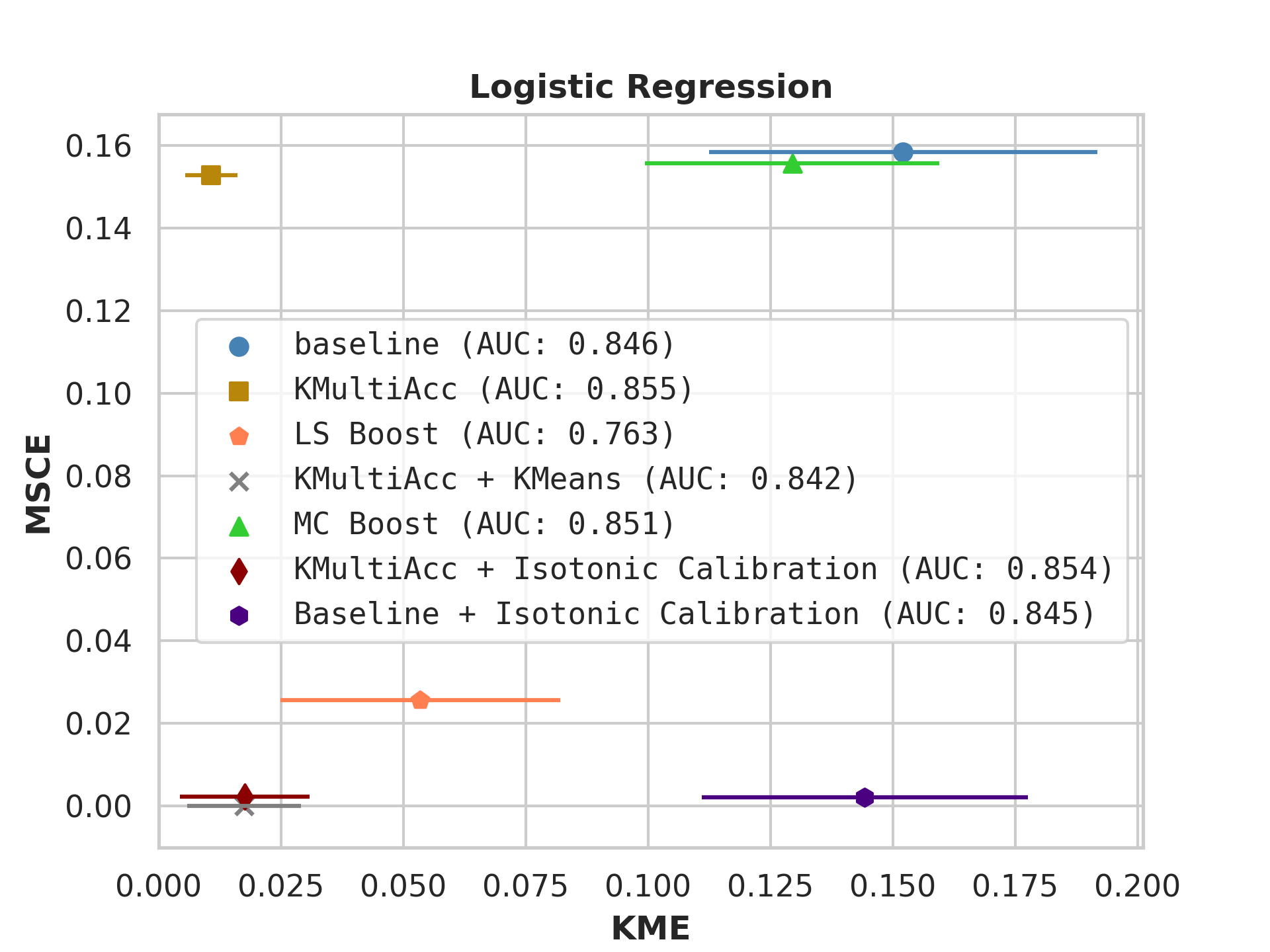} &

    \includegraphics[width=0.45\textwidth]{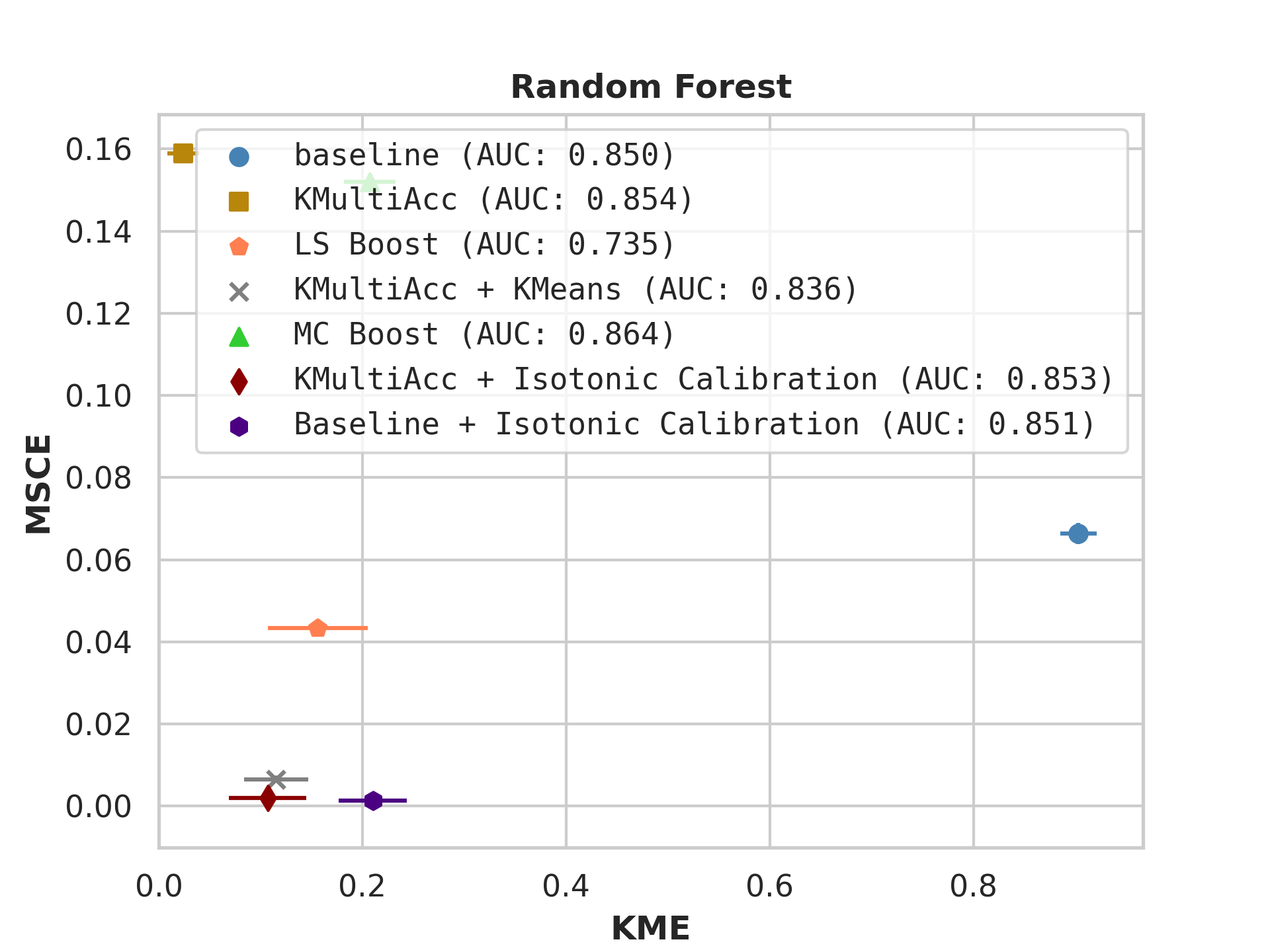}   &

    \includegraphics[width=0.45\textwidth]{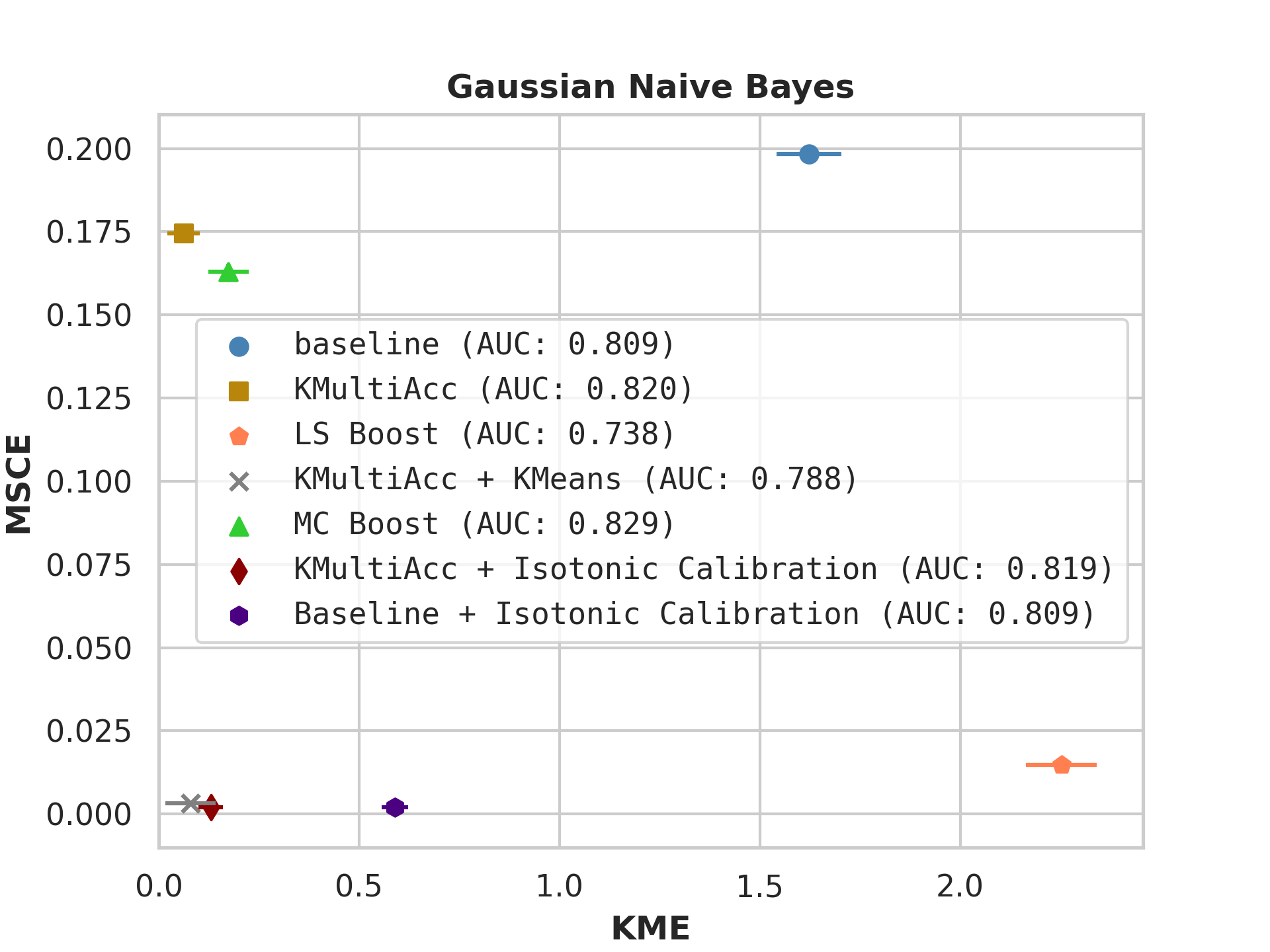}&
    
    \includegraphics[width=0.45\textwidth]{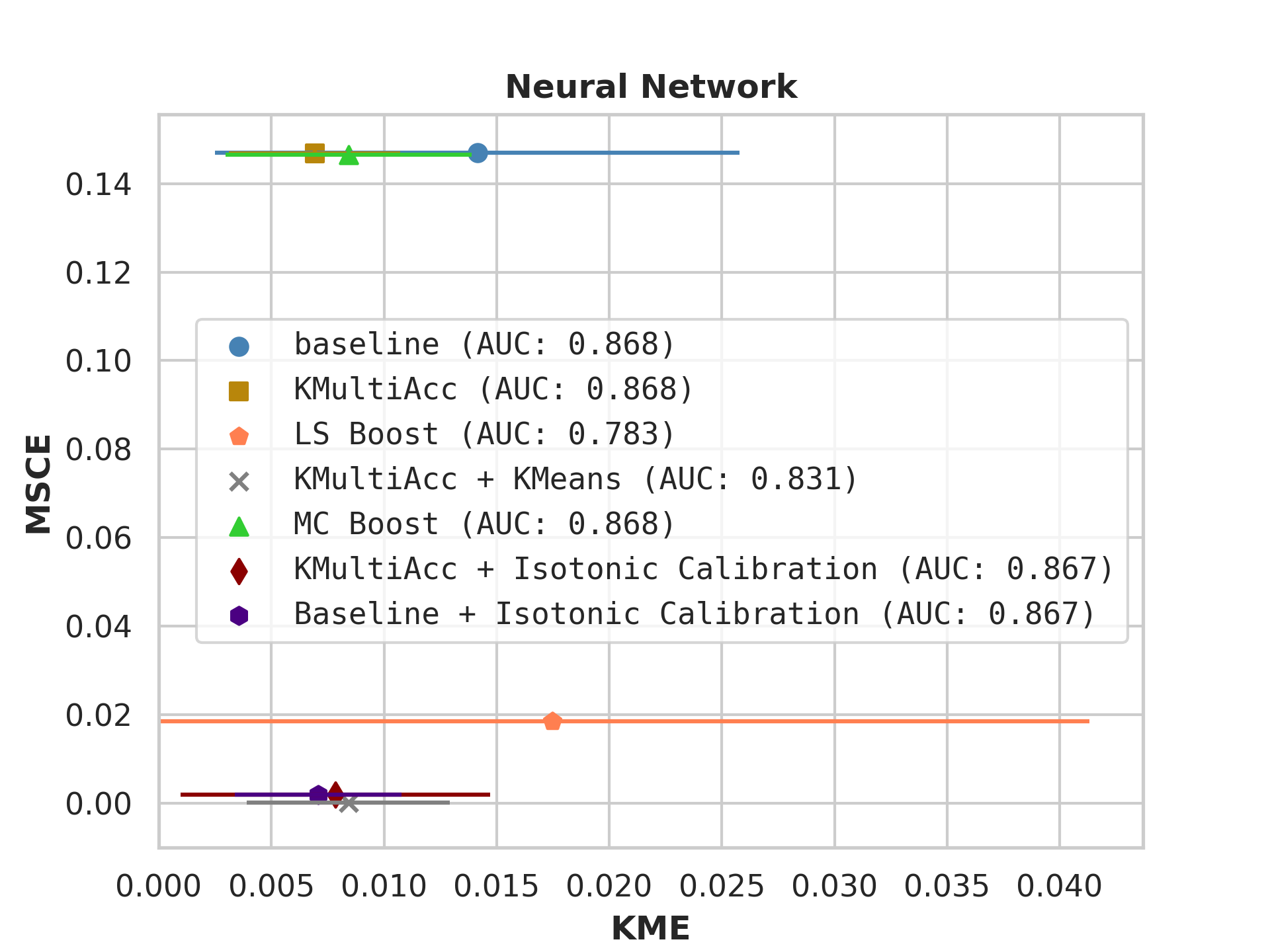}
   
     \end{tabular}}
            \caption{The Folktables Income Task with data from the state of Washington.} \label{fig::Income_WA}
\end{figure*}

\begin{figure*}[htbp]
     \centering
     \resizebox{\linewidth}{!}{\begin{tabular}{cccc}
    
 \includegraphics[width=0.45\textwidth]{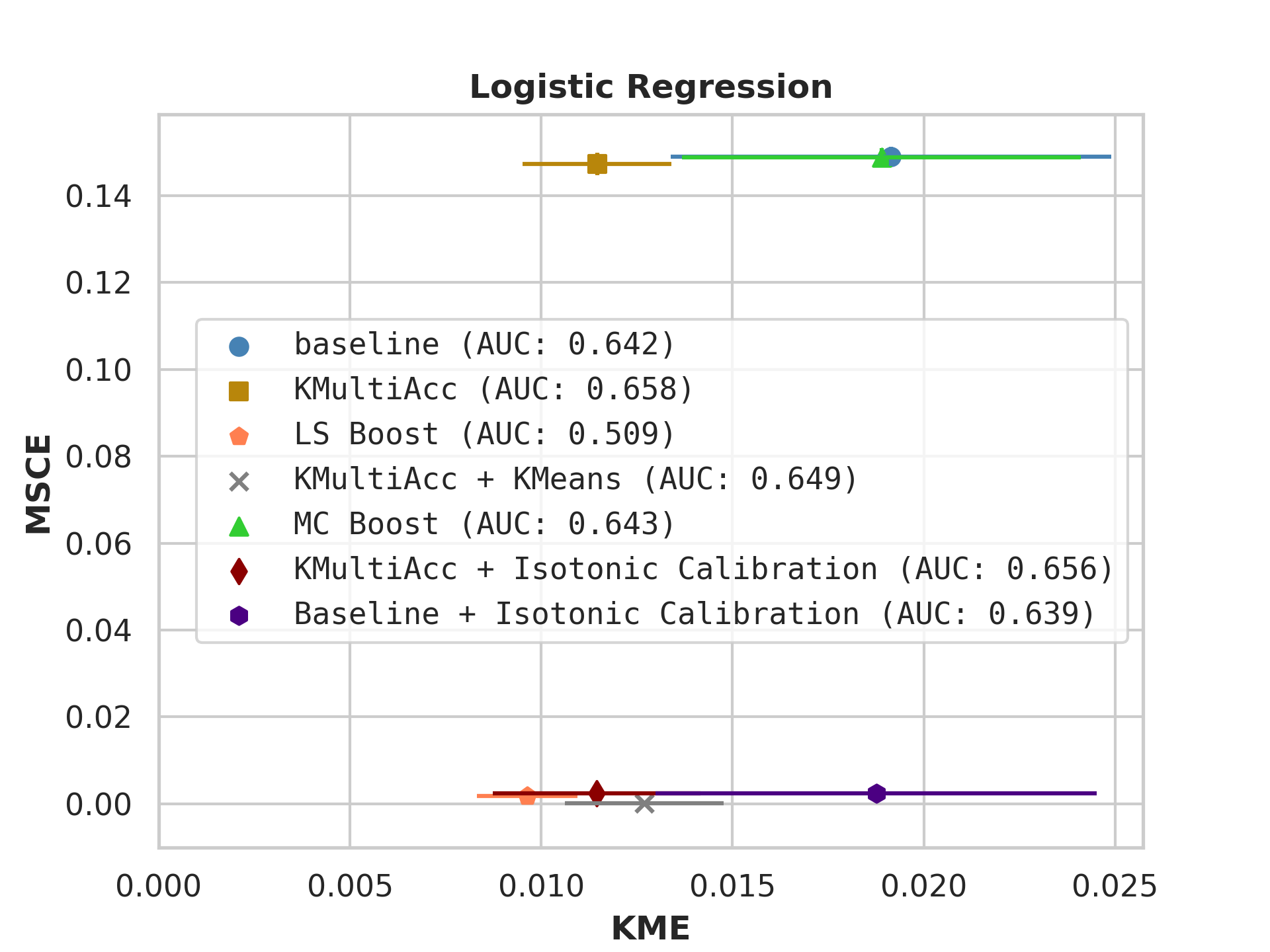} &

 \includegraphics[width=0.45\textwidth]{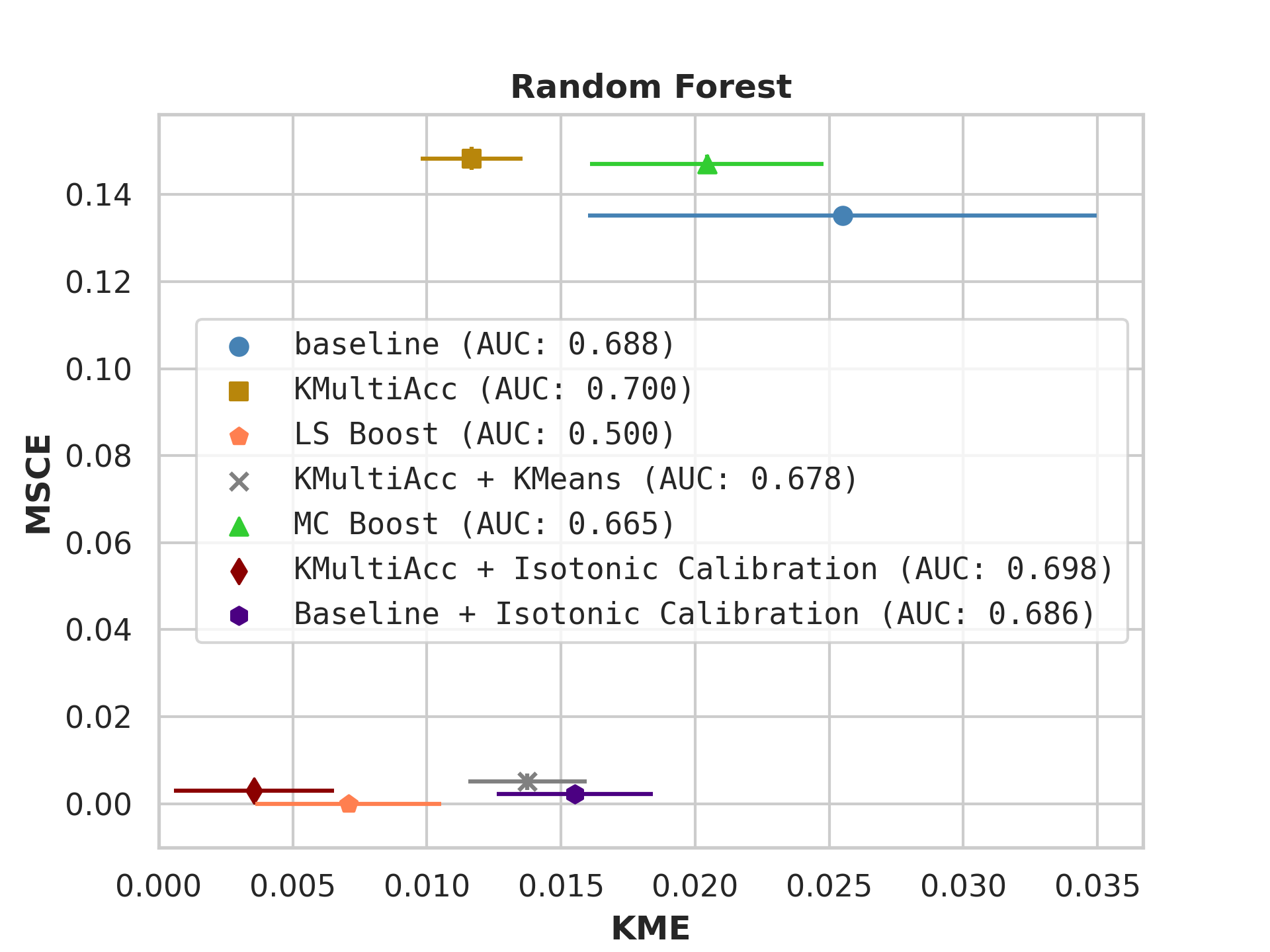}  &

 \includegraphics[width=0.45\textwidth]{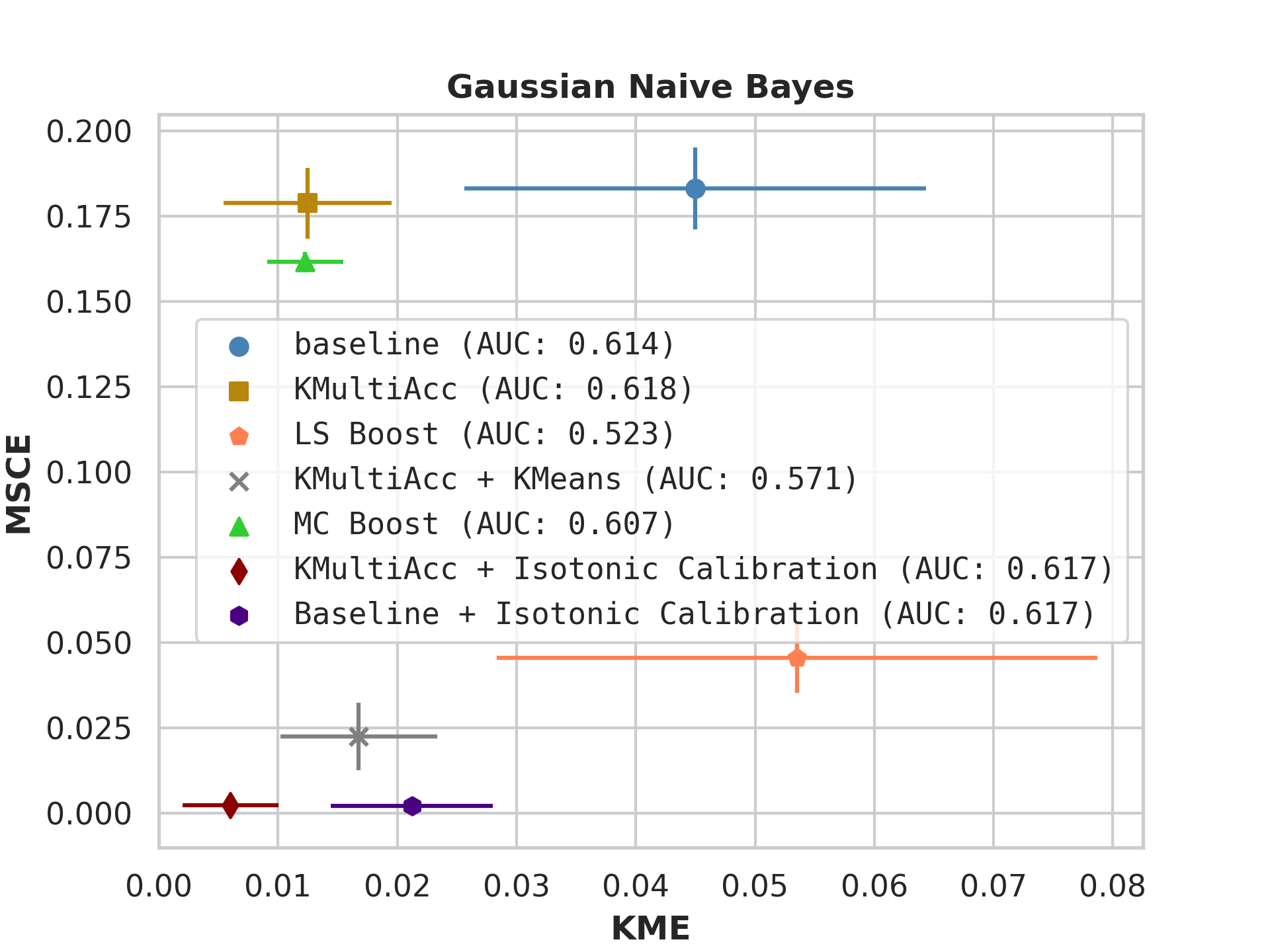}&
 
 \includegraphics[width=0.45\textwidth]{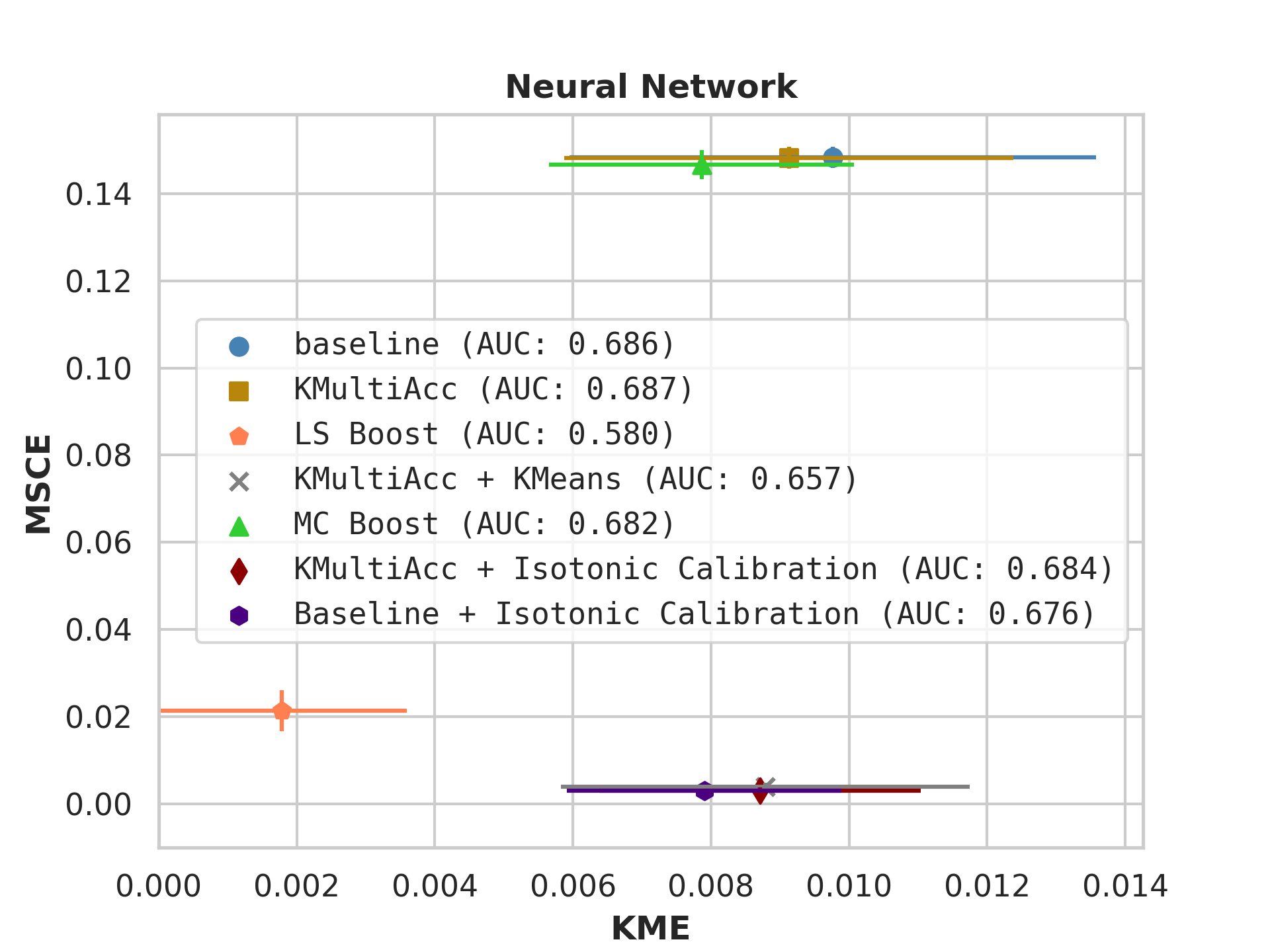}

     \end{tabular}}
      \caption{The Folktables Mobility Task with data from the state of New Jersey.} \label{fig::Mobility_NJ}
\end{figure*}

\begin{figure*}[htbp]
     \centering
     \resizebox{\linewidth}{!}{\begin{tabular}{cccc}
 \includegraphics[width=0.45\textwidth]{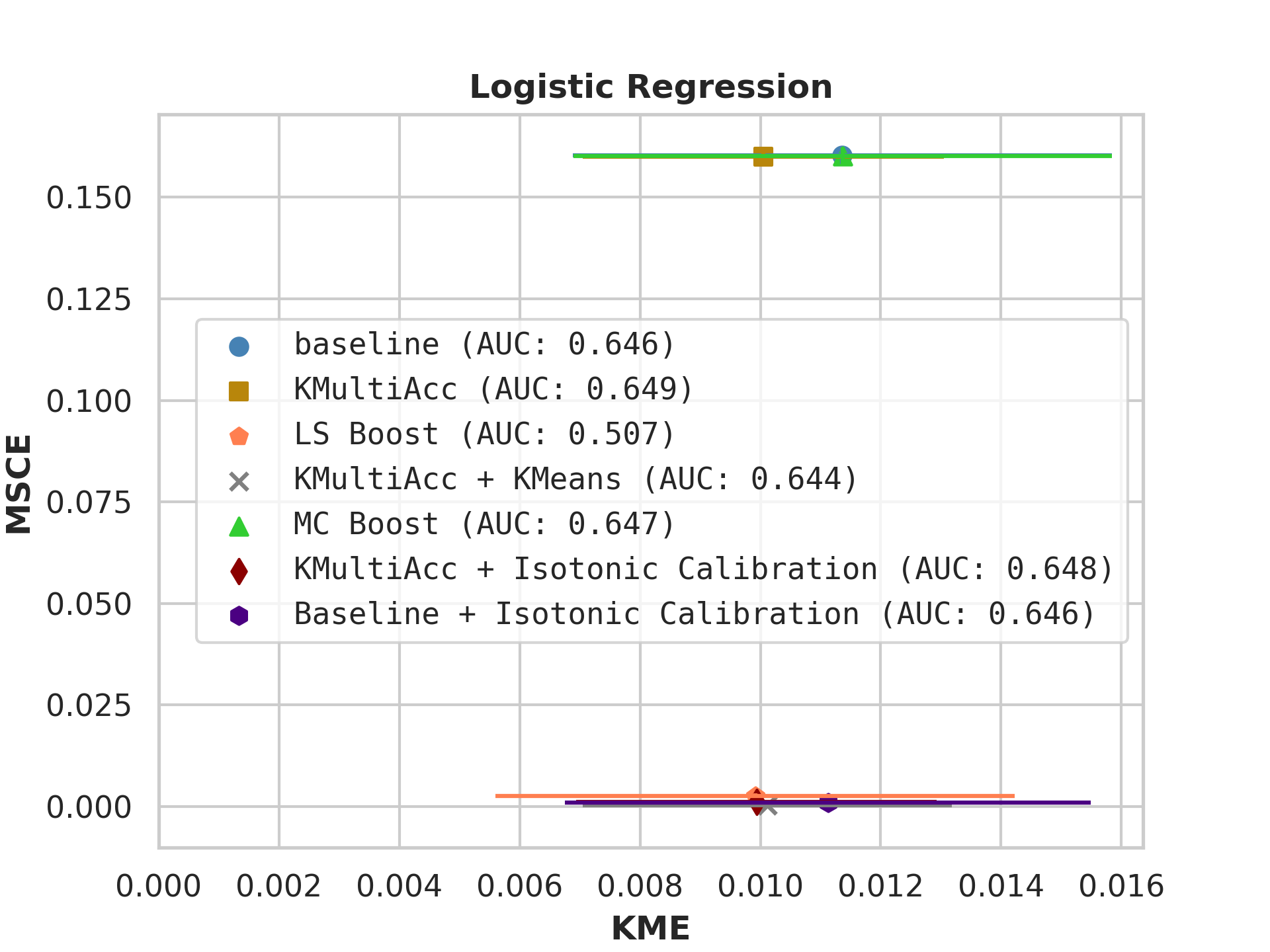} &

 \includegraphics[width=0.45\textwidth]{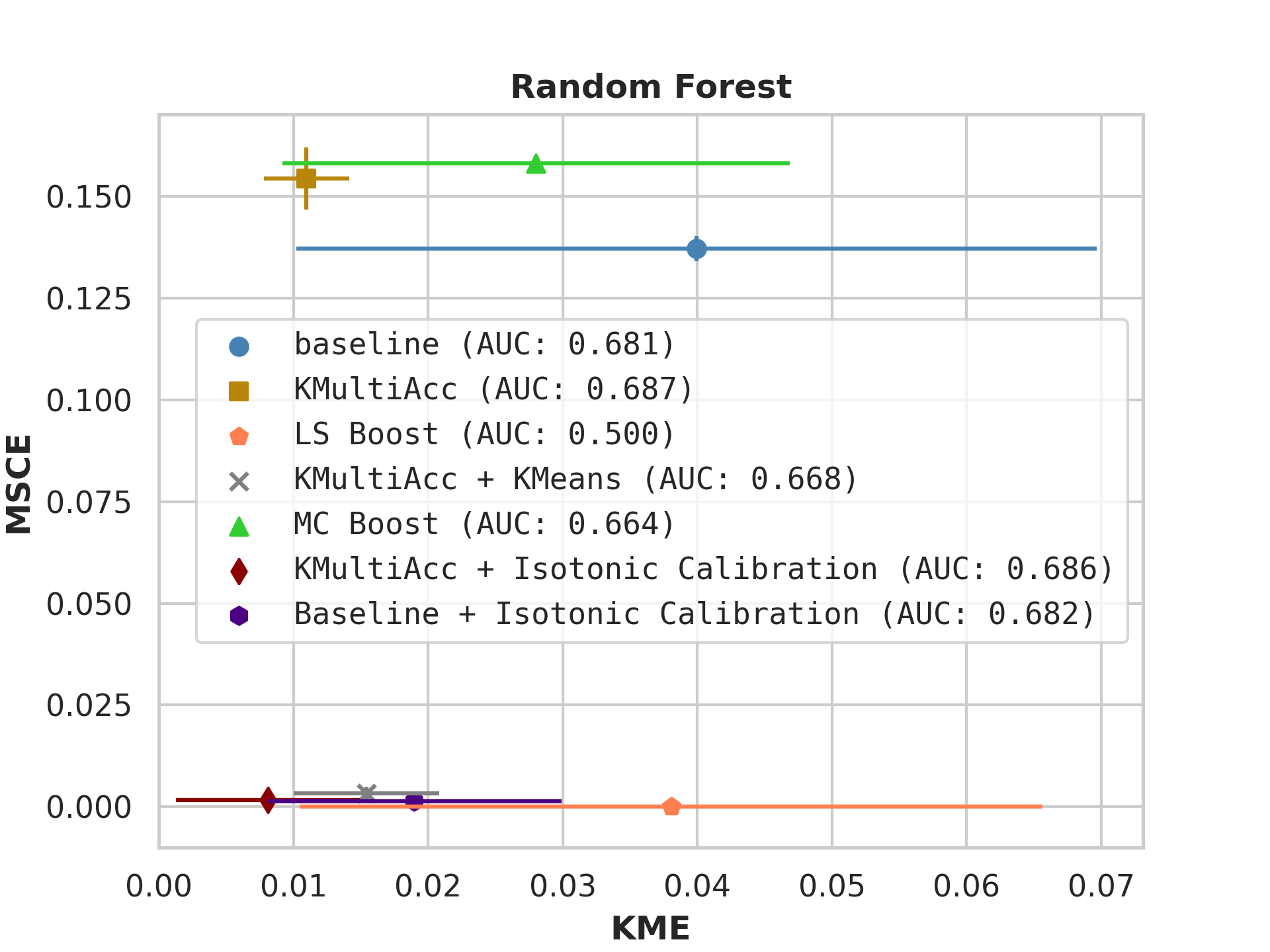}  &

 \includegraphics[width=0.45\textwidth]{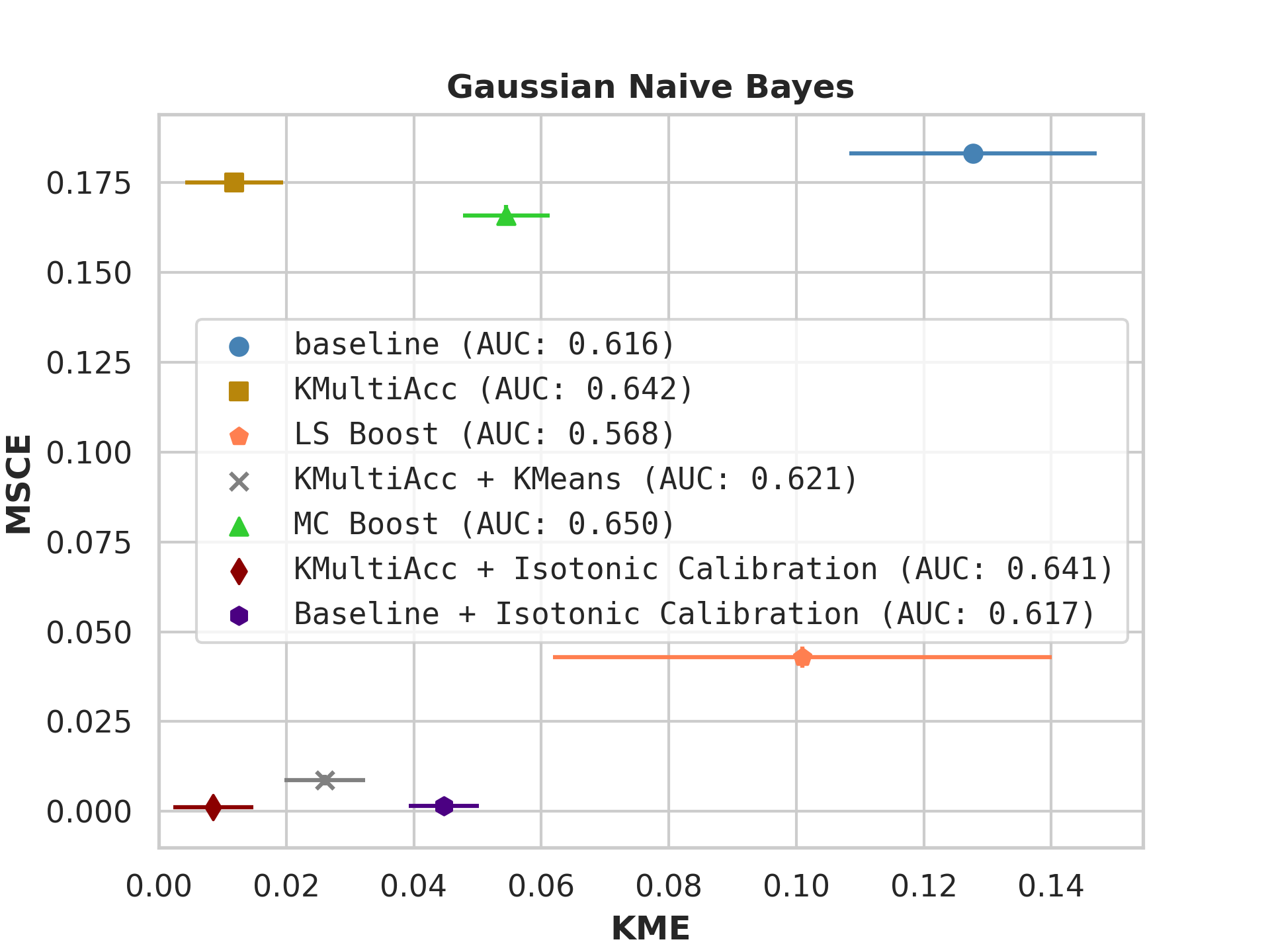}&
 
 \includegraphics[width=0.45\textwidth]{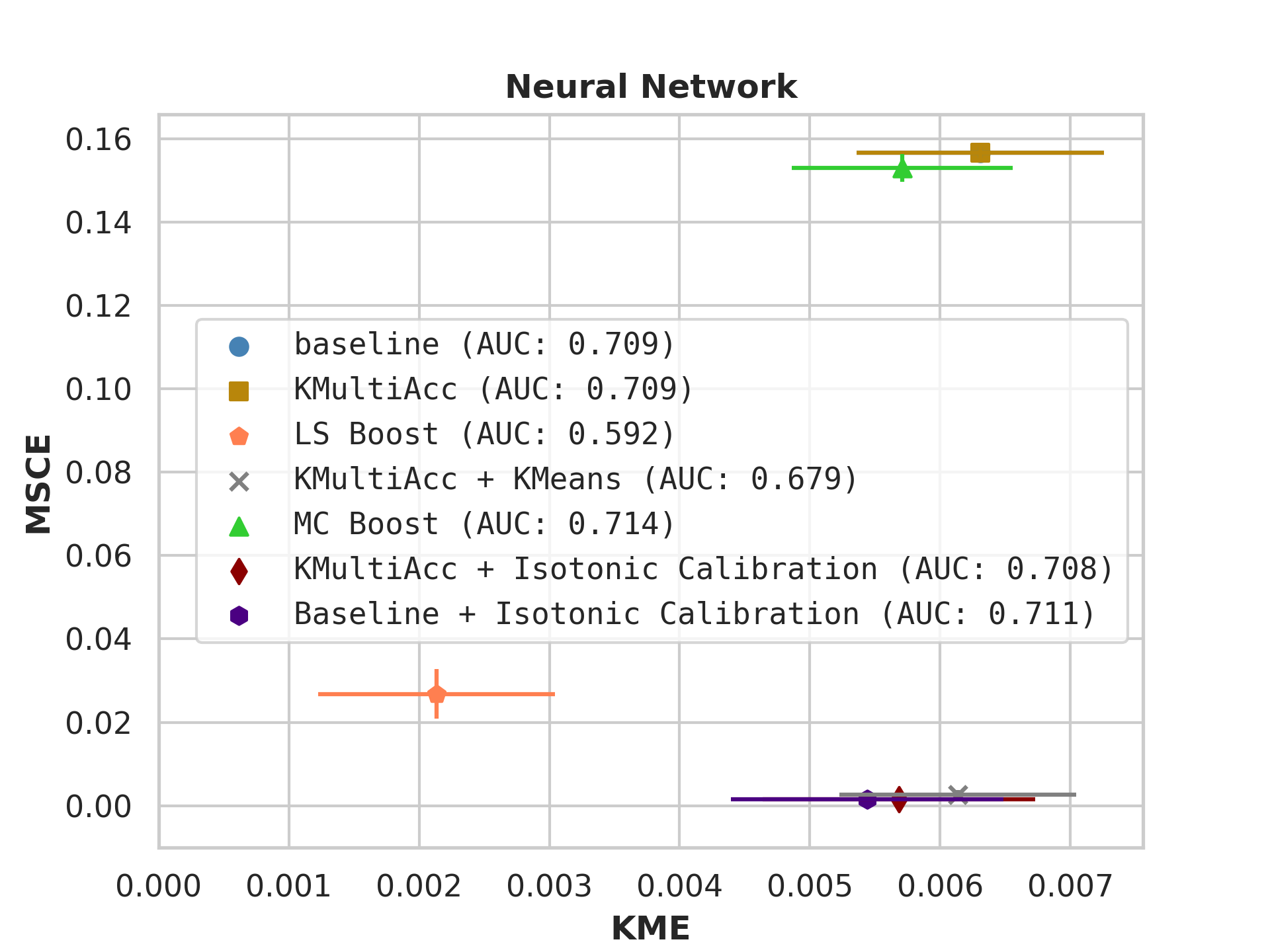}
   
\end{tabular}}
    \caption{The Folktables Mobility Task with data from the state of New York.} \label{fig::Mobility_NY}
\end{figure*}

\begin{figure*}[htbp]
     \centering
     \resizebox{\linewidth}{!}{\begin{tabular}{cccc}
 \includegraphics[width=0.45\textwidth]{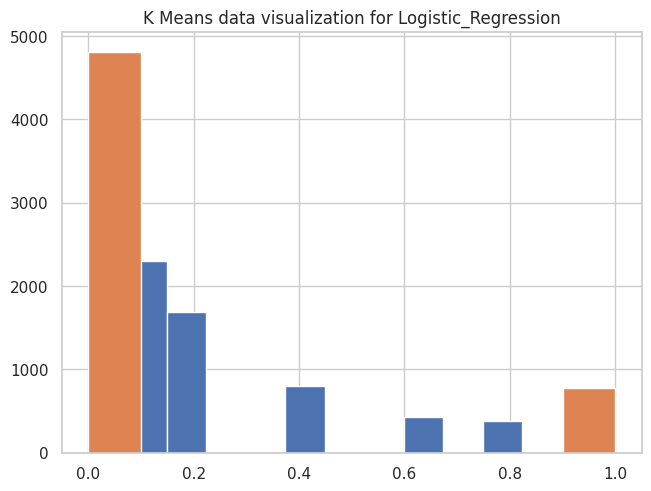} &

 \includegraphics[width=0.45\textwidth]{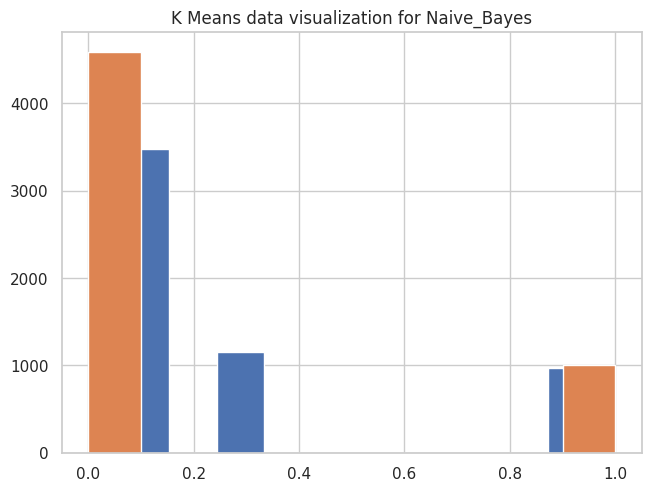} &

 \includegraphics[width=0.45\textwidth]{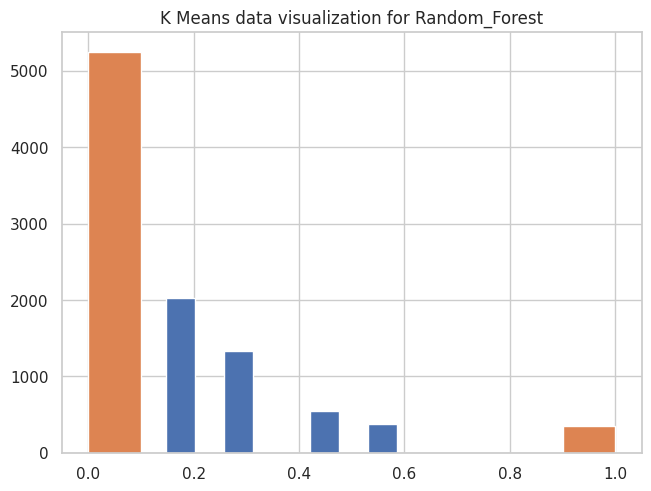} &

 \includegraphics[width=0.45\textwidth]{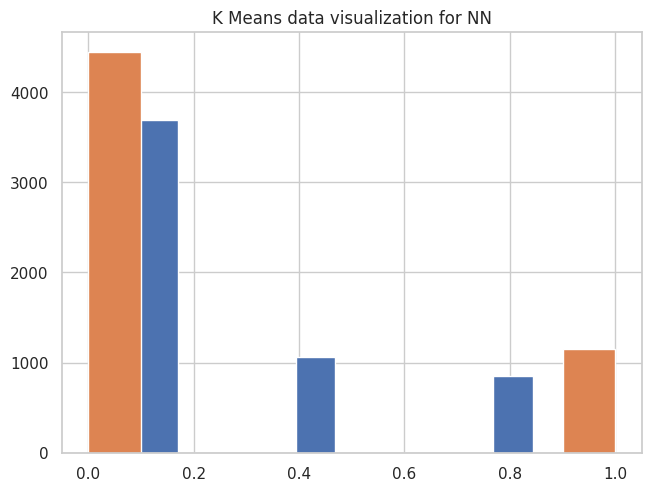}
 \\

 \includegraphics[width=0.45\textwidth]{Folktable_Results/Health_WI/KMeans_LR.png} &

 \includegraphics[width=0.45\textwidth]{Folktable_Results/Health_WI/KMeans_Naive_Bayes.png} &

 \includegraphics[width=0.45\textwidth]{Folktable_Results/Health_WI/KMeans_Random_Forest.png} &

 \includegraphics[width=0.45\textwidth]{Folktable_Results/Health_WI/KMeans_NN.png}
   
\end{tabular}}
    \caption{To understand the influence of \LSBoost{} on our predictions, we viewed it through the lens of histograms of \LSBoost{}'s estimates of the labels (Blue), and constructed K-means bins for the scores of functions (Orange) for each model type and the Health in Wisconsin and Wealth in Washington tasks (as examples of the empirical result guiding us to test the effect of these interventions).} \label{fig::KMeans}
\end{figure*}

\section{\KMultiAcc{}: Conditions for One-Step Sufficiency}
\label{appendix:one-step}

A one-step update using the witness function in \KMultiAcc leads to a $0$-multiaccurate predictor all functions in the RKHS $c \in \calH_k$ for the linear kernel and very specific constructions of nonlinear kernel. 
Running \KMultiAcc{} as an iterative procedure is redundant under these restricted settings. This is a property of RKHSs that follows from the Riesz Representation Theorem \cite{goodrich1970riesz}. %and Mercer's Theorem \cite{williams2006gaussian}~Chapter 4.

 \begin{remark}
    To gain an intuition on why multiaccuracy may be zero upon a one-step update, we can observe an analogous result in the Euclidean space $\mathbb{R}^d$. Given a linear subspace $\calH$, a prediction vector $\mathbf{f}$ and true labels $\by$, multiaccuracy can be similarly defined as 
    $$\max_{c \in \calH, \|c\|\leq 1} c^T (\by- \mathbf{f})$$
    The error of a classifier $\mathbf{e}=(\by-\mathbf{f})$ can be decomposed into two components: the projection of $\mathbf{e}$ onto the subspace $\calH$ and the residual. Hence, we have $\mathbf{e}=\mathbf{e}_\calH + \mathbf{e}_R$. Then, once we subtract away $\mathbf{e}_\calH$, multiaccuracy error boils down to the dot product $c^T \mathbf{e}_R$, which equals 0. \qed
\end{remark}

% \begin{proposition}[Sufficiency of One-Step Update in \KMultiAcc{}] \label{lemma:one-step sufficiency}
%     Given a base predictor $f:\calX \to [0,1]$, an RKHS $\calH_k$, the witness function $c_{k,f}^\star(\bx)$ and normalizing constant $\theta$ as defined in Eq. \eqref{eq:c star def}). We assume that $\calH_k\subset L^1(\calX)$\footnote{Again, $L^1(\calX)$ denotes the space of real-valued functions that are integrable against $P_{\bX}$, i.e. $L^1(\calX) \define \left\{ c:\calX \to \BR \Co \BE\left[\left|c(\bX)\right|\right] < \infty \right\}$.}, and that kernel $k$ is shift-invariant\footnote{Examples include the Gaussian and Laplacian kernel.}. Let the updated predictor be $g(x)=f(x)+\lambda c_{k,f}^\star(\bx)$, where $\lambda = \frac{1}{\theta}$. Upon the one-step update, the multiaccuracy error for all functions in the RKHS is zero: 
%     $$\EE{c(X)(Y-g(X))}=0 \quad \forall{c \in \calH_k}.$$
% \end{proposition}

Next, we proceed with an RKHS $H_k\subset L^1$. Given a base predictor $f:\calX \to [0,1]$, the kernel function $k$ w.r.t an RKHS $\calH_k$. Again, $L^1(\calX)$ denotes the space of real-valued functions that are integrable against $P_{\bX}$, i.e. $L^1(\calX) \define \left\{ c:\calX \to \BR \Co \BE\left[\left|c(\bX)\right|\right] < \infty \right\}$. Let the multiaccuracy error of one function $c\in\calH_k$ be $L(c,f)= \EE{c(X)(Y-f(X))}$. 
By the reproducing property of $\calH_k$, we have $c(x) = \langle c, k(\cdot, x)\rangle_k$ for all $ c\in \calH_k, x\in \calX$. We can thus rewrite the multiaccuracy error as the following:
\begin{align*}
    L(c,f) &= \EE{\langle c,k(\cdot,X)\rangle_k(Y-f(X))}\\
            &= \langle c, \EE{(Y-f(X))k(\cdot, X)}\rangle_k\\
            &= \langle c,h \rangle_k
\end{align*}
where $h(x) = \EE{(Y-f(X))k(x, X)}$. For the second step, since $H_k\subset L^1$, we can invoke the Fubini's Theorem to interchange expectation and inner product. Under the assumption of integrability, $h\in \calH_k$.
By the Riesz Representation Theorem, the linear functional $L(\cdot,f)$ has a unique representer $h\in \calH_k$, which is the function $h(x)$ defined above. Specifically, by the Riesz Representation Theorem, there exists a unique $c_{k,f}^\star \in \calH_k$ such that for all $c \in \calH_k$, 
    $$L(c,f) = \langle c, h\rangle_k,$$
where the function $c_{k,f}^*$ is defined as the normalized direction of $h$, i.e. $c_{k,f}^\star = \theta h$, and $\theta = \frac{1}{\|h\|_k}$ This is identical to $c_{k,f}^\star(\bx)$ as defined in \eqref{eq:c star def}.
From Proposition \ref{prop::witness_multiaccuracy}, $c_{k,f}^\star(\bx)$ achieves the supremum over $\calH_k$: 
$c_{k,f}^\star = \arg \sup_{c\in \calH_k, ||c||\leq 1} L(c,f)$. 
This simplifies $L(c_{k,f}^\star, f) = \langle \theta h,h \rangle_k=  \theta \|h\|^2_k = \|h\|$, where we substitute $\theta = \frac{1}{\|h\|_k}$. Let the updated predictor be:
$g(x) = f(x) + \lambda c_{k,f}^\star(x)$.

The multiaccuracy error after the one-step update is given by:
\begin{align*}
\EE{c(X)(Y - g(X))} &= \EE{c(X)(Y -  (f(x) + \lambda c_{k,f}^\star(x)))} \\
&= \EE{c(X)(Y - f(x))} -  \lambda \EE{c(X) c_{k,f}^\star(x)}\\
&= \EE{c(X)(Y - f(x))} -  \lambda \EE{c(X) \theta  h(X)}\\
&= L(c,f) - (\lambda\times \theta) \EE{c(X)h(X)}
\end{align*}

For the linear kernel (as we have observed in the remark), we operate in the Euclidean space, and $L(c,f)=\EE{c(X)h(X)}$. Hence, By taking $\lambda = \frac{1}{\theta}$, we have 
\begin{align*}
\EE{c(X)(Y - g(X))}= L(c) - (\lambda\times \theta) L(c)= 0.
\end{align*}

For non-linear kernels, $L(c,f)\neq \EE{c(X)h(X)}$ in general, and equality holds only when $$\BE_X [k(\cdot,X)k(X,X')] = \kappa k(\cdot, X')$$ where $\kappa = \BE_{X}[k(X,X)]$ is a scalar constant.

To see this, we need to simplify $\EE{c(X)h(X)}$ in the kernel space:
\begin{align*}
    \EE{c(X)h(X)} &= \BE_X[c(X) \BE_{X'}[(Y'-f(X'))k(X,X')]]\\
    &= \BE_{X,X'}[c(X)(Y'-f(X'))k(X,X')]\\
    &= \BE_{X,X'}[\langle c, k(\cdot, X)\rangle_k (Y'-f(X'))k(X,X')]\\
    &= \langle c, \BE_{X,X'}[(Y'-f(X'))k(\cdot,X)k(X,X')]\rangle_k\\
    &= \langle c, \BE_{X'}[(Y'-f(X'))\BE_X [k(\cdot,X)k(X,X')]]\rangle_k
\end{align*}
In the first equality, we substitute in the definition of $h(x)$. In the second equality, we apply Fubini's Theorem to swap the two expectations. In the third equality, we apply the reproducing property where $c(X) = \langle c, k(\cdot,X)\rangle$. In the fourth equality, we interchange the expectation and inner product by Fubini's theorem under integrability conditions. In the last equality, we expand into iterative expectations.

\chapter{Chapter \ref{ch:4}}
\label{apdx:proofs_ch4}

\section{Related Works}
Given the extensive volume of work in LLM watermarking, we focus our discussion on works that inform and contrast with our main contribution: theoretical frameworks for analyzing the limits of LLM watermarking.

\emph{Classical Information-Theoretic Approaches.} Post-process watermarking, where watermarks are embedded after content generation, has been extensively studied through information-theoretic lenses \cite{chen2000design,moulin2003information,martinian2005authentication}, particularly through the Gelfand-Pinsker (GP) channel \cite{gel1980coding,villan2006text,willems2000informationtheoretical}, which treats the LLM token $X\sim Q_X$ as the channel state for constructing the watermarked token. The GP scheme constructs auxiliary random variables $U\sim P(U|X)$ and encodes the watermarked token as $A=f(U,X)$. These approaches differ from our approach in two key aspects: (1) they typically require long sequences for joint typicality to hold, which leads to schemes that are intractable in the online setting with a large token vocabulary, while we focus on optimizing the one-shot minimax setting motivated by auto-regressive generation; and (2) they generally assume perfect knowledge of the underlying distributions, whereas our scheme is designed to work with the assumption that the underlying distribution is unknown, only the sampled token and side information are available.

\emph{Modern LLM Watermarking.} Kirchenbauer et al. \cite{kirchenbauer2023watermark} introduced the first watermarking scheme for LLMs, which divides the vocabulary into green and red lists and slightly enhances the probability of green tokens in the next token prediction (NTP) distribution. This seminal work sparked numerous developments \cite{aaronson2023watermark,he2024universally, bahri2024watermark,dathathri2024scalable,yang2023watermarking,ren2024subtle,huunbiased2024,zhao2024permute,chao2024watermarking,qu2024provably,xie2024debiasing,liuadaptive}, with several approaches focusing on distortion-free methods that maintain the original NTP distribution unchanged, e.g., \cite{kuditipudi2023robust,huunbiased2024,zhao2024permute,christ2024undetectable}. Unlike these methods which primarily focus on implementation strategies, our work provides a theoretical framework that characterizes optimal detection-perception trade-offs. Most related to our approach, Chao et al. \cite{chao2024watermarking} propose a watermark using optimal correlated channels, though our work differs by providing a complete characterization through joint optimization of the randomization distribution in the one-shot setting.

\emph{Theoretical Analysis of LLM Watermarking.} Recent work has advanced our theoretical understanding of LLM watermarking limitations. Huang et al. \cite{huang2023towards} designed an optimal watermarking scheme for a specific detector, but their approach requires knowledge of the original NTP distributions of the watermarked LLM, making it model-dependent. Li et al. \cite{li2024statistical} proposed detection rules using pivotal statistics, though their Type II error control relies on asymptotic techniques from large deviation theory and focuses on large-sample statistics, whereas our analysis addresses the fundamental one-shot case including explicit characterization of corner point cases and the development of an optimal correlated channel scheme. Most recently, He et al. \cite{he2024universally} characterizes the universal Type II error while controlling the worst-case Type-I error by optimizing the watermarking scheme and detector. In contrast to these approaches, we analyze optimal mean detection by formulating a minimax framework while balancing Type I and Type II errors through the use of an $E_\gamma$-information objective. In the minimax formulation, we provide the optimal mean detection in closed form and characterize the optimal distribution of randomness under adversarial token distributions. 

The development of the field is tracked through comprehensive benchmarks \cite{piet2023mark,tu2023waterbench,pan2024markllm,qiu2024evaluating} and surveys \cite{zhao2024sok, liu2024survey}. 

\section{Proofs}

\subsection{Proof for Proposition \ref{prop:tv_prob}}
% \begin{proof}
Fixed $(P_S, Q_X, \tilde{Q}_{X|S})$ and priors $(\pi_0, \pi_1)$. 

Eve's hypothesis testing problem can be formulated as distinguishing between $H_0: A \sim Q_X$ and $H_1: A \sim \tilde{Q}_X$. By the Neyman-Pearson Lemma, the optimal test statistic is given by the likelihood ratio $L(a) = Q_X(a)/\tilde{Q}_X(a)$. The optimal decision rule takes the form $\delta(a) = \mathbbm{1}\{L(a) > \eta\}$ for some threshold $\eta$. The probability of correct detection for Eve can be expressed as:
\begin{align*}
\Pr(\hat{H}_E = C) &= \frac{1}{2}\Pr(\delta(A) = 1|H_1) + \frac{1}{2}\Pr(\delta(A) = 0|H_0)
\end{align*}

For the optimal threshold $\eta = 1$, this probability becomes:
\begin{align*}
\Pr(\hat{H}_E = C) &= \frac{1}{2} + \frac{1}{2}\sum_{a \in \mathcal{X}}|\tilde{Q}_X(a) - Q_X(a)| \\
&= \frac{1}{2} + \frac{1}{2}\text{TV}(\tilde{Q}_X, Q_X)
\end{align*}

Now, we turn to Bob's detection probability. Bob's hypothesis testing problem differs from Eve's due to his access to the side information $S$. His testing problem can be formulated as distinguishing between $H_0: (A,S) \sim Q_{X|S} \times P_S$ and $H_1: (A,S) \sim \tilde{Q}_{X|S} \times P_S$. 

By the Neyman-Pearson Lemma, the optimal test statistic in this case is $L(a,s) = Q_{X|S}(a|s)/\tilde{Q}_{X|S}(a|s)$. Given priors $(\pi_0, \pi_1)$ and let $\gamma = \frac{\pi_1}{\pi_0}$, the conditional probability of correct detection given $S=s$ is:
\begin{align}
\Pr(\hat{H}_B = C|S=s) &= \pi_0 \Pr(\delta(A) = 0|H_0) + \pi_1 \Pr(\delta(A) = 1|H_1)\\
&=  \pi_0 Q_{X|S}[L(a,s)\geq \gamma] + \pi_1 \tilde{Q}_{X|S}[L(a,s) \leq \gamma]\\
&= \pi_1 + \pi_0 Q_{X|S}[L(a,s)\geq \gamma] - \pi_1 \tilde{Q}_{X|S}[L(a,s) \geq \gamma]\\
&= \pi_1 + \pi_0 \left[ Q_{X|S}[L(a,s)\geq \gamma] - \frac{\pi_1}{\pi_0}\tilde{Q}_{X|S}[L(a,s) \geq \gamma \right]\\
&= \pi_1 + \pi_0 E_{\gamma}(Q_{X|S}||\tilde{Q}_{X|S}).
\end{align}
The last equality comes from the alternative formula for $E_\gamma$ where $E_{\gamma}(P||Q) = \max_{\calA} [P(\calA)-\gamma Q(\calA)]$, and supremum is attained with $A = \{a|L(a,s)\geq \gamma\}$. 
% \end{proof}

\subsection{Proof of Theorem \ref{thm:opt_cornerpoints}}\label{proof:opt_cornerpoints}
By the assumption of a uniform prior, we are looking for bounds on the quantity 
$\frac{1}{2}(1+E_\gamma(\tilde{Q}_{X|S}\|Q_X|P_S))$, which boils down to bounding $E_\gamma(\tilde{Q}_{X|S}\|Q_X|P_S) = \mathbb{E}_S\left[E_\gamma(\tilde{Q}_{X|S}\|Q_X)\right]$.
First, note that under a uniform prior, this quantity is lower bounded by the performance of a random guess, i.e., $\frac{1}{2}\leq \rd$. In what follows, we develop an upper for $E_\gamma(\tilde{Q}_{X|S}\|Q_X|P_S)$.
For simplicity, denote $|\cX|=d$ and $|\cS|=m$.
Let $Q_{X|S=s_i}=p_i$ such that $p_1,...,p_m \in \Delta_d$, where $\Delta_d$ denotes the $d$-dimensional simplex.
Assume that $S\sim\mathsf{Unif}[m]$. Following the zero perception assumption, we have $\tilde{Q}_X=Q_X$, i.e., $\frac{1}{m}\sum_{i=1}^m p_i = Q_X$.
Consequently, our TV-optimization, when jointly optimized also over the marginal distribution $Q_X$ is of the form:
\begin{equation}
\label{eq:opt}
\max_{p_1,...,p_m\in \Delta_d} \frac{1}{m}\sum_{i=1}^m \left\|p_i - \frac{\gamma}{m}\sum_{i=1}^m p_i\right\|_+,
\end{equation}
where $\|x\|_+\triangleq \sum_{i}(x_i)_+$
for $d\geq m$.
We are maximizing a convex function over a polytope, so the optimal solution lies on the extreme points. 
Thus $p_i= e_j$ for some $j\leq d$, where $e_j$ is the indicator vector with $j$-th entry equal to one. The problem boils down to determining how many times each vector $e_j$ shows up.

Denote with $q$ the probability vector corresponding to the distribution $Q_X$. We note that $q$ can be rewritten as
\begin{equation}\label{eq:q_def_proof_opt}
q \triangleq \frac{1}{m}\sum_{i=1}^m p_i = \frac{1}{m}\sum_{j=1}^d n_j e_j, 
\end{equation}
where $\sum_j n_j = m$ and $n_j\in\NN$. Denote the $j$-th entry of $q$ by $q_j$. We have $\|e_j - q\|_+ = (1-q_j)_+  = 1-q_j.$
Therefore:
\begin{align*}
\frac{1}{m}\sum_{i=1}^m \left\|p_i - \gamma q\right\|_+&\stackrel{a}{=}\frac{1}{m}\sum_{j=1}^d n_j \left\|e_j - \gamma q\right\|_+\\
&= \frac{1}{m}\sum_{j=1}^d n_j (1-\gamma q_j)_+\\
&\stackrel{b}{=} \sum_{j=1}^d q_j (1-\gamma q_j)_+
\end{align*}
where (a) follows from from rewriting the sum in terms of $e_j$ and (b) follows from the relation $q_j = \frac{n_j}{m}$, as can be seen from \eqref{eq:q_def_proof_opt} and by the definition of the indicator.
Out optimization problem had therefore boiled down to maximizing on the quantity 
\begin{align}\label{eq:thm1_opt}
\sum_{j=1}^d q_j (1-\gamma q_j)_+~\mbox{such that } q_j = k/m, k\in \mathbb{Z}, \sum_{j=1}^d q_j=1.
\end{align}
To solve \eqref{eq:thm1_opt}, we will examine various settings of the value of $\gamma$.
\paragraph{$\gamma \leq 1$}
First, note that when $\gamma=0$ the objective sums up to $1$ by the constraints. 
Otherwise, note that whenever $\gamma\leq 1$, we have $(1-\gamma q_j)_+=1-\gamma q_j$. Thus, we have
$$
\sum_{j=1}^d q_j (1-\gamma q_j)_+ = 1 - \gamma\sum_{j=1}^n q_j^2.
$$
Thus, maximization of the objective, boils down to the minimization of the sum of squares.
We note that as $q$ is a probability vectors, the sum of square minimizes under the uniform distribution, with the minimum being $\frac{1}{m}$.
Thus, we have the upper bound 
$$
\frac{1}{2}(1+E_\gamma(\tilde{Q}_{X|S}\|Q_X|P_S))\leq \frac{1}{2}\left(1 + 1 - \frac{\gamma}{m}\right) = 1 - \frac{\gamma}{2m}.
$$
\paragraph{$\gamma>1$}
In this case, we are not guaranteed with the positivity of  $(1-\gamma q_j)$. We will look for a strategy to choose the values of $(q_j)_j$ such that the considered sum is maximized, while not passing the threshold that nullifies the terms $(1-\gamma q_j)$.
For each $j$, denote each summand as $f(q_j)$, whose value is
$$
f(q_j) = \begin{cases}
    q_j - \gamma q_j^2,\quad q_j \leq \frac{1}{\gamma}\\
    0,\quad \text{else}.
\end{cases}
$$
Consequently, as $q_j$ is constrained to the set $(\frac{k}{m})_{k=0}^m$, whenever $\gamma\geq m$, no positive value of $q_j$ will result in a positive value of $f(q_j)$. Thus, the resulting sum is $0$, which implies that $\rd = \frac{1}{2}$.
Thus we will focus on $\gamma\in(1,m)$.
In this case, there is at least one possible value for each $q_j$ that results in a nonnegative value of $f(q_j)$.
First, we note that the mapping $x \mapsto x-\gamma x^2$ is a concave function of $x$ for $\gamma>0$, whose maximum is obtained in $x^\star = \frac{1}{2\gamma}$.
Therefore, we would like to set $q_j=\frac{1}{2\gamma}$ as this will maximize a single summand.
However, in most cases $\frac{1}{2\gamma}\notin(\frac{k}{m})_{k=1}^m$.
To that end, we will set the closes possible value to $\frac{1}{2\gamma}$ within the allowed
set. Second, we we would like to set as many $q_j$'s to the value $\frac{1}{2\gamma}$ while following the constraint $\sum_{j=1}^d q_j=1$, we will choose the lower value.
To summarize, for each interval $\frac{k}{m}\leq\frac{1}{2\gamma}\leq \frac{k+1}{m}$, we will set $q_j = \frac{k}{m}$.
The maximal amount of such $q_j$ we can set while following the sum constraint is $\lfloor \frac{m}{k} \rfloor $.
Thus, we have the following
\begin{align*}
    E_\gamma(\tilde{Q}_{X|S}\|Q_X|P_S) &= \left\lfloor \frac{m}{k} \right\rfloor\left( \frac{k}{m} - \gamma \left(\frac{k}{m}\right)^2\right)\\
    &\leq  1 - \frac{\gamma k }{m}.
\end{align*}
The corresponding bound on $\rd$ is $1 - \frac{\gamma k }{2m}$.
The bound is achievable whenever $m$ is divisible by $k$ within the resulting interval.
Note that the interval $\frac{k}{m}\leq \frac{1}{2\gamma}\leq \frac{k+1}{m}$ corresponds to the interval $\frac{m}{2(k+1)}\leq \gamma\leq \frac{m}{2k}$.
However, we already know the resulting bounds for $\gamma\geq m$ and $\gamma\leq1$. Thus, the relevant values of $k$ that correspond to this case are $k\in[1:\frac{m}{2}]$.
Finally, when $\frac{1}{2m}<\frac{1}{2\gamma}<\frac{1}{m}$ we cannot take the lower value $(k=0)$, and will therefore take higher value $k=1$.
However, note that $\frac{1}{2m}<\frac{1}{2\gamma}$ corresponds to $\gamma>m$. Thus, this sub-case $(\frac{1}{2m}<\frac{1}{2\gamma}\leq \frac{1}{m})$ boils down to $\gamma<\frac{m}{2}$ with corresponding upper bound of $1-\frac{\gamma}{m}$, which will merge with the interval $\gamma\leq 1$.
This concludes the proof $\hfill\square$

\subsection{Proof of Theorem \ref{thm:universal_ub}}
Let $Q_i\triangleq Q_{X|S=s_i}$ The proof follows from analyzing the following steps:
\begin{align*}
    \sup_{\tilde{Q}_{X|S}}\sum_{s\in\cS} P_S(s) E_{\gamma}(\tilde{Q}_{X|S=s},Q_X) &= \sup_{\tilde{Q}_{X|S}}\frac{1}{2|\cS|}\sum_{i=1}^{|\cS|}\|Q_i - \gamma Q_x \|_1\\
    &= \frac{1}{2|\cS|}\sup_{f:\cS\to\cX}\sum_{i=1}^{|\cS|}\|Q_{f(i)} - \gamma Q_x \|_1\\
    &\leq \frac{1}{2} \sup_{i\in \cX} \|Q_{i} - \gamma Q_x \|_1\\
    &=\sup_{i\in \cX} \left|1 - \gamma Q_x(i)\right|\\
    &= 1 - \gamma Q_{\mathsf{min}}
\end{align*}
Therefore, 
$$
\rd\leq \frac{1}{2}\left(1 + 1 - \gamma Q_{\mathsf{min}}\right) = 1 - \frac{\gamma Q_{\mathsf{min}}}{2}
$$
For the second equality, note that argmax of a convex function lies in the corner of the probability simplex.
$\hfill\square$

\subsection{Proof of Correlated Channel (CC) with Perfect Perception}
We prove that CC is a perfect perception scheme, i.e. $\mathbb{E}_{S}\left[\tilde{Q}_{X|S}\right](x) = Q_X(x)$.
Recall that $S=(Y,B^m)$.We have the following
\begin{align*}
    \mathbb{E}_{S}\left[\tilde{Q}_{X|S}\right](x) &= \sum_{y,b^m} \mu_{B^m}(b^m)P_Y(y)Q_X(x)\frac{P_{Y|\tilde{Y}}(y|f(x,b^m))}{P_Y(y)}\\
    &= Q_X(x) \sum_{y,b^m}\mu_{B^m}(b^m)P_{Y|\tilde{Y}}(y|f(x,b^m)).
\end{align*}
\newcommand{\cB}{\mathcal{B}}
Denote by $\cB_1(x)\triangleq\left\{b^m: f(x,b^m)=1\right\}$ and denote $\cB_0(x)$ by the same token. We have
\begin{align*}
    &\mathbb{E}_{S}\left[\tilde{Q}_{X|S}\right](x) \\
    &=Q_X(x) \left(\sum_{b^m\in\cB_1(x)}\mu_{B^m}(b^m)\underbrace{\sum_{y=0,1}(b^m)P_{Y|\tilde{Y}}(y|1)}_{=1} + \sum_{b^m\in\cB_0(x)}\mu_{B^m}\underbrace{\sum_{y=0,1}\mu_{B^m}(b^m)P_{Y|\tilde{Y}}(y|0)}_{=1}\right)\\
    &=Q_X(x).
\end{align*}
This concludes the proof.$\hfill\square$

\subsection{Proof of Proposition \ref{prop:optimal_coupling}}

By the dual representation of the total variation
	\begin{equation}
	\TV(P,Q) = \min_{P_{XY}}\{ \mathbb{P}[X \not = Y]: P_X = P, P_Y = Q\},
	\label{eq:tv-dual}
\end{equation}

Given $S\sim \mathsf{Unif}[k]$ and $P_{\tilde{Y}}=\{p_1,...,p_k\}\in \Delta_k$. We have 
$\TV(P_S,P_{\tilde{Y}}) = 1- \sum_{i=1}^k \min(\frac{1}{k},p_i)$.

We propose a coupling and shows that it achieves $\TV(P_S,P_{\tilde{Y}})$. 

To simplify notation, let the distribution of $S$ and $\tilde{Y}$ be $P$ and $Q$. Let $t  = \TV(P,Q)$. Assume that $0<t<1$. Define three probability distributions 
	 $R = \frac{P \wedge Q}{1-t}$, $P'=\frac{P-P \wedge Q}{t}$ and $Q'=\frac{Q-P\wedge Q}{t}$.
	Construct $P_{XY}$ as follows: 
	\begin{enumerate}
	\item Generate $B \sim \text{Bernoulli}(t)$. 
	\item If $B=0$, draw $Z \sim R$ and set $S=\tilde{Y}=Z$. 
	\item If $B=1$, draw $S\sim P'$ and $\tilde{Y} \sim Q'$ independently. 
\end{enumerate} 

To show that this is a valid coupling, we verify the marginal distribution is kept the same.
We have: 
\begin{eqnarray*}
P_S(a) & = & \mathbb{P}(B=0)R(a) + \mathbb{P}(B=1)P'(a) \\
& = & (1-t) \left( \frac{P \wedge Q}{1-t}\right) (a) + t \left( \frac{P - P\wedge Q}{t}\right) (a) \\
& = & P(a)
\end{eqnarray*} 
Similarly, 
\begin{eqnarray*}
P_{\tilde{Y}}(a) & = & \mathbb{P}(B=0)R(a) + \mathbb{P}(B=1)Q'(a) \\
& = & (1-t) \left( \frac{P \wedge Q}{1-t}\right) (a) + t \left( \frac{Q - P\wedge Q}{t}\right) (a) \\
& = & Q(a)
\end{eqnarray*} 
Therefore $P_{S\tilde{Y}}$ is a valid coupling.

Lastly, we show that for the specific coupling, $\mathbf{P}(\tilde{Y} \neq S)=\TV(P_S,P_{\tilde{Y}})$

\begin{align*}
    \mathbf{P}(\tilde{Y} \neq S)&= 1-\mathbf{P}(\tilde{Y}=S)\\
    &= 1- (1-t)\\
    &= t\\
    &=\TV(P_S,P_{\tilde{Y}})
\end{align*}

Thus, we have constructed a coupling $P_{S\tilde{Y}}$ that minimizes $\mathbf{P}(\tilde{Y} \neq S)$, which means that it maximizes $\mathbf{P}(\tilde{Y} = S)$. $\hfill\square$

\subsection{Proof of Remark \ref{thm:detection_test_optimal}}
The hypothesis test is the following: $H_0: X \sim Q_X$ and $H_1: X \sim \tilde{Q}_{X|S,B^m}$, where $\tilde{Q}_{X|S,B^m}$ is the CC-watermark distribution shown in equation \eqref{eq:wm_dist_channel}, and side information $S\sim \mathsf{Ber}(0.5)$.  We show $H_0$ is rejected by the CC detection test $S=f(X,B^m)$ if and only if it is also rejected by the likelihood ratio test (LRT).

If $H_0$ is rejected by CC detection test, then $S=f(X,B^m)$. Then, consider the likelihood ratio:
\begin{align}
    \frac{Q_X(X)}{\tilde{Q}_{X|B^m,S}(X)}&= 
    \frac{Q(X)}{Q_X(X)\frac{1}{P_S(S)}P_{S|\tilde{Y}}(S|f(X,B^m)}\\
    &= \frac{2}{P_{S|\tilde{Y}}(S|f(X,B^m)}\\
    &<1,
\end{align}
The density of $\tilde{Q}_{X|B^m,S}(X)$ follows from the CC-watermark, side information $P_S(S)=0.5$. The last inequality come from the Z-S channel construction: $\Pr_{S|\tilde{Y}}(S|f(S,B^m)\geq \frac{1}{2}$, if and only if $S=f(X,B^m)$. Since the likelihood ratio is less than $1$, $H_0$ is rejected by the LRT.

If $H_0$ is rejected by the LRT with threshold 1, then we have 
$$ \frac{Q_X(X)}{\tilde{Q}_{X|B^m,S}(X)} <1.$$
Expanding the likelihood ratio as above, this implies: 
$P_{S|\tilde{Y}}(S|f(X,B^m)<\frac{1}{2}.$
By construction of the Z-S channel, $S=f(X,B^m)$. Hence, $H_0$ is rejected by CC detection test.

\subsection{Proof of Proposition \ref{thm:detection_close_form}}\label{proof:tvq_as_tvy}
We start by proving the following identity:
    $$
    \TV\left(Q_X,\tilde{Q}_{X|(S,B^m)}|P_{S,B^m}\right) = \TV\left(P_{S},P_{S|\tilde{Y}}|P_{\tilde{Y}}\right)
    $$

% \subsection{Proof of Proposition \ref{prop:corr_ch_bob_rate}}\label{proof:corr_ch_bob_rate}
Proof:
Recall that in the correlated channel watermark we have side information $S$ and partition bits $B^m$. By definition, we have
\begin{equation}\label{eq:tv_tower}
    \TV(Q_X,\tilde{Q}_{X|S,B^m}|P_{S,B^m}) = \sum_{b^m}\sum_{s=0,1}\mu(b^m)P_S(s)\TV(Q_X,\tilde{Q}_{X|b^m,s}).
\end{equation}
Next, we simplify the TV expression within the sum. For any $(b^m,s)$ we have
\begin{align*}
    \TV(Q_X,\tilde{Q}_{X|(b^m,s)}) &= \sum_{x}\left| Q_X(x) - Q_X(x)\frac{P_{S|\tilde{Y}}(s|f(x,b^m))}{P_S(s)} \right|\\
    &= 2\sum_{x} Q_X(x)\left| \frac{1}{2} - p_{S|\tilde{Y}}(s|\tilde{y})\right|,
\end{align*}
where recall that $\tilde{Y}=f(X,B^m)$, $p_{S|\tilde{Y}}(s|\tilde{y})$ is the corresponding coupling channel parameter, and
$S\sim \ber(\frac{1}{2})$.
We define the pre-image of $f$ for a fixed $b^m$ as $f^{-1}(\cdot,b^m):\{0,1\}\to 2^\cX$, with $f^{-1}(0),f^{-1}(1)\subseteq\cX$.
Plugging the simplified TV expression back into \eqref{eq:tv_tower}, we have 
\begin{align*}
    &\TV(Q_X,\tilde{Q}_{X|(b^m,s)}) \\
    &= \sum_{b^m}\mu(b^m)\sum_{s=0,1}\sum_{x} Q_X(x)\left| \frac{1}{2} - p_{S|\tilde{Y}}(s|\tilde{y})\right|\\
    &= \sum_{b^m}\mu(b^m)\sum_{s=0,1} \left( \sum_{x\in f^{-1}(0,b^m)} Q_X(x)\left| \frac{1}{2} - p_{S|\tilde{Y}}(s|0)\right| + \sum_{x\in f^{-1}(1,b^m)} Q_X(x)\left| \frac{1}{2} - p_{S|\tilde{Y}}(s|1)\right|  \right)\\
    &= \sum_{b^m}\mu(b^m) \left( P_{\tilde{Y}}(0)\sum_{s=0,1}\left| \frac{1}{2} - p_{S|\tilde{Y}}(y|0)\right| + P_{\tilde{Y}}(1) \sum_{s=0,1}\left| \frac{1}{2} - p_{S|\tilde{Y}}(s|1)\right| \right)\\
    &=\TV\left(P_S,P_{S|\tilde{Y}} | P_{\tilde{Y}} \right),
\end{align*}
where the randomness of $\tilde{Y}$ is determined by the pair $(Q_X,\mu)$. This concludes the proof. $\hfill\square$

With this, we proceed to showing CC's detection rate. By Theorem \ref{thm:detection_test_optimal}, CC's detection rate is equal to that of likelihood ratio test. By Proposition \ref{prop:tv_prob} and under equal priors on TPR and TNR, we have 
\begin{align}
    R_d &= \frac{1}{2}(1+\TV(Q_X,\tilde{Q}_{X|S,B^m}|P_{S,B^m}))\\
    &= \frac{1}{2}\left(1+\TV(P_{S},P_{S|\tilde{Y}}|P_{\tilde{Y}})\right),
\end{align}
where the last equality is due to the identity above.

Next, we obtain a closed form for $\TV(P_{S},P_{S|\tilde{Y}}|P_{\tilde{Y}})$. By definition, we have
$$
\TV\left(P_{S},P_{S|\tilde{Y}}|P_{\tilde{Y}}\right) = \tilde{p}_0 \TV\left(P_{S},P_{S|\tilde{Y}=0}\right)+\tilde{p}_1 \TV\left(P_{S},P_{S|\tilde{Y}=1}\right).
$$
Following Proposition \ref{prop:optimal_coupling}, the nature of the TV terms depends on wether $\tilde{p}_1\leq\frac{1}{2}$ or $\tilde{p}_0\leq\frac{1}{2}$ .
For $\tilde{p}_0\leq\frac{1}{2}$, the optimal coupling is given by a $Z$-channel, whose parameter is $\frac{2\tilde{p}_1-1}{2\tilde{p}_1}$. The TV terms are therefore given by
$$
    \TV\left(P_{S},P_{S|\tilde{Y}=0}\right) = \frac{1}{2}\left|\frac{1}{2}-1\right| + \frac{1}{2}\left|\frac{1}{2}\right| = \frac{1}{2}
$$
\begin{align*}
    \TV\left(P_{S},P_{S|\tilde{Y}=1}\right) &= \frac{1}{2}\left(\left|\frac{1}{2}-\frac{2\tilde{p}_1-1}{2\tilde{p}_1}\right| + \left|\frac{1}{2}-\frac{1}{2\tilde{p}_1}\right|\right)\\
    &= \frac{1}{2}\left(\left|\frac{1-\tilde{p}_1}{2\tilde{p}_1}\right| + \left|\frac{\tilde{p}_1-1}{2\tilde{p}_1}\right|\right)\\
    &= \frac{\tilde{p}_0}{2\tilde{p}_1}.
\end{align*}
Thus, we have
$$
\TV\left(P_{S},P_{S|\tilde{Y}}|P_{\tilde{Y}}\right) = \tilde{p}_0.
$$
By the symmetry of the optimal coupling, for $\tilde{p}_1\leq\frac{1}{2}$ we have 
$$
\TV\left(P_{S},P_{S|\tilde{Y}}|P_{\tilde{Y}}\right) = \tilde{p}_1.
$$

Hence, CC's detection rate is given by 
$R_d = \frac{1}{2}(1+\min(\tilde{p_0},\tilde{p_1}))$
. $\hfill\square$

\subsection{Proof of Theorem \ref{thm:optimal_maxmin_detection}}

We begin by proving Lemma \ref{lem:permutation_invariance}.
\subsubsection{Proof of Lemma \ref{lem:permutation_invariance}}
Let $\calS = [k]$ and $\calX=[m].$ For a given $Q_X = \bq = (q_1,\dots,q_m) \in \Delta_m$ and an $m$-length sequence $\mathbf{b}=(b_1,\dots,b_m)\in \calS^m$, we define the function $f:\calX \times \calS^m\to \calS$ as
\begin{equation}
    f(i,\mathbf{b}) = b_i.
\end{equation}
A sequence $\mathbf{b}$ induces a probability distribution  $\hat P(\mathbf{q},\mathbf{b}) $ over $\calS$ denoted as (with a slight abuse of notation)
\begin{equation}
    \hat P(s,\mathbf{q},\mathbf{b}) = \sum_{i=1}^m q_i \mathbf{1}\left[b_i=s\right]~\forall s\in[k].
\end{equation}
For a fixed $\mathbf{b}$ and $\bq$ and assuming that Alice uses the optimal coupling, Bob's probability of detection is given by the quantity
\begin{align}
    R_d(\bq, \mathbf{b})  &\triangleq  1 - \frac{1}{2}\mathsf{TV}\left(Q_S \| \hat P(\mathbf{q},\mathbf{b}) 
 \right) -\frac{1}{2k} \sum_{s=1}^k \hat P(s,\mathbf{q},\mathbf{b})\\
 &=  1-\frac{1}{2k} - \frac{1}{4} g(\bq,\mathbf{b}),
\end{align}
where
\begin{align}
 g(\bq,\mathbf{b}) \triangleq \sum_{s=1}^k \left|\hat P(s,\mathbf{q},\mathbf{b}) - \frac{1}{k}\right| 
\end{align}
where $Q_S$ is the uniform distribution. Our goal is to find a distribution over $P_{B^m}^*$ that maximizes the worst-case value of $R_d$ given a set of constraints on $\bq$.  Specifically, we analyze:
\begin{align}
    R_d^*(\lambda) &\triangleq \max_{P_{B^m}} \min_{\substack{\bq\in \Delta_m\\ \|\bq\|_\infty \leq \lambda}} \mathbb{E}\left[R_d(\bq,B^m)\right]\\
    &= 1 - \frac{1}{2k}-\frac{1}{4} \min_{P_{B^m}} \max_{\substack{\bq\in \Delta_m\\ \|\bq\|_\infty \leq \lambda}} \sum_{\mathbf{b}\in \mathcal{S}^m} P_{B^m}(\mathbf{b}) g(\bq, \mathbf{b}).
    \label{eq:rdstar}
\end{align}
The function
\begin{equation}
  H(P_{B^m}) \triangleq     \max_{\substack{\bq\in \Delta_m\\ \|\bq\|_\infty \leq \lambda}} \mathbb{E}\left[g(\bq, B^m) \right]
\end{equation}
is convex in the distribution $P_{B^m},$ since it is the maximum of linear functions. Let  $P_{B^m}^*$ be a distribution that minimized $H$ and consider the permutation $\pi:\mathcal{S}^m\to \mathcal{S}^m$, define $\tilde P_\pi(\mathbf{b}) = P_{B^m}^*(\pi \circ \mathbf{b})$.  

Since $ \mathbb{E}_{P_{B^m}^*}\left[g(\bq, B^m) \right] =  \mathbb{E}_{\tilde P_\pi}\left[g(\pi \circ \bq, B^m) \right]$ for all $\bq$,  $H(\tilde P_\pi)=H(P_{B^m})$ from the symmetry of the maximum. Hence, from the equality in \eqref{eq:rdstar} $F(\tilde P_\pi)=F(P_{B^m})$ for $F(P_{B^m})\triangleq \min_{\substack{\bq\in \Delta_m\\ \|\bq\|_\infty \leq \lambda}} \mathbb{E}_{P_{B^m}}\left[\rd(Q_X,B^m)\right]$.
$\hfill\square$

Next, we proceed with the proof of Theorem \ref{thm:optimal_maxmin_detection}.

Let $C=m!$ be the number of permutations of an $m$-length sequence, we have
\begin{equation}
    F\left(\frac{1}{C}\sum_\pi\tilde P_\pi  \right) \leq F(P_{B^m}^*).
\end{equation}
Consequently, it is sufficient to restrict the minimization in $P_{B^m}$ to distributions that assign equal probability mass to sequences that are identical up to a permutation.

Denote by $\mathcal{P}_m$ the partition of $\mathcal{S}^m$ into sets of sequences that are equal up to a permutation, with $|\mathcal{P}_m|=K$. For simplicity, we denote $\mathcal{P}_m = \left(\calB_{1},\dots,\calB_K \right)$ and refer to $\calB_i$ as a \emph{permutation class} (alternatively, we could have named it orbits or type classes). Then

\newcommand{\bw}{\mathbf{w}}
\begin{align}
    \min_{P_{B^m}} F(P_{B^m}) = \min_{\bw \in \Delta_{K} } \max_{\substack{\bq\in \Delta_m\\ \|\bq\|_\infty \leq \lambda}} \sum_{i=1}^K \frac{w_i}{|\calB_i|} \sum_{\bb\in \calB_i}g(\bq,\bb).
\end{align}
Observe that $g(\bq,\bb)$ is convex in $\bq$ (since it is the absolute value of a linear function in $\bq$), and thus the inner maximum is achieved at a vertex of the feasible set. The vertices of the polytope $\left\{\bq \in \Delta_m \mid \|\bq\|_\infty \leq \lambda \right\}$ are permutations of the vector
$$ \bq^*_\lambda = (\lambda,\dots,\lambda, 1-t\lambda, 0,\dots,0 ),$$
where $\bq^*$ has (i) exactly $t$ entries equal to $\lambda$ and $t$ is the largest integer such that $t\lambda \leq 1$ (assuming $\lambda \leq 1$), (ii) one entry equal to $1-t\lambda$, and (iii) the remaining entries equal to 0. 

Since the vertices are identical up to a permutation, and for any permutation $\pi$
\begin{equation}
 \sum_{\bb\in \calB_i}g(\bq,\bb) =  \sum_{\bb\in \calB_i}g(\pi \circ \bq,\bb),
\end{equation}
it is sufficient to select a vertex of the form $\bq^*_\lambda$. Thus, 
\begin{equation}
    \min_{P_{B^m}} F(P_{B^m}) = \min_{\bw \in \Delta_{K} } \sum_{i=1}^K \frac{w_i}{|\calB_i|} \sum_{\bb\in \calB_i}g(\bq^*_\lambda,\bb),
\end{equation}
and it sufficient to consider the optimal distribution $P_{B^m}^*$ as a distribution that selects a $\bb$ uniformly over a \emph{single} permutation class in $\calP_m$; namely the one that maximizes $\frac{1}{|\calB_i|} \sum_{\bb\in \calB_i}g(\bq^*_\lambda,\bb).$

Next, we aim to characterize $R_d^*(\lambda)$ for different values of $\lambda$. We denote by $P_{\calB}$ the distribution resulting from drawing a sequence at random from the permutation class $\calB\in \calP_m$.

Our goal is to compute
\begin{equation}
\label{eq:g_expectation}
    \mathbb{E}\left[g(\bq_\lambda^*,B^m )\right] = \sum_{s=1}^k \mathbb{E}\left[ \left|\hat P(s,\mathbf{q}_\lambda^*,B^m) - \frac{1}{k}\right| \right]
\end{equation}
Recall that the optimal choice of $P_{B^m}$ is to draw sequences uniformly from a single permutation class. Assuming w.l.o.g. that $\mathcal{S}=[k]$, fix a sequence $\bb\in \calS^m$ with $n_i$ entries equal to $i$, $i\in [k]$. For example, if $k=2$, $n_1$ is the number of entries equal to 1 and $n_2$ is the number of entries equal to 2. Naturally, $\sum_{i=1}^k n_i=m$.

Now, for a fixed $s\in \calS$, we can write
\begin{equation}
    P(s,\mathbf{q}_\lambda^*,B^m) = \lambda \sum_{i=1}^t X_i + (1-t\lambda)X_{t+1},
\end{equation}
where $t = \lfloor 1/\lambda \rfloor$ and $X_i \defined \ones\left(B_i = s \right)$. We can expand the expectation in the lhs of \eqref{eq:g_expectation} as
\begin{align}
    \mathbb{E}\left[ \left|\hat P(s,\mathbf{q}_\lambda^*,B^m) - \frac{1}{k}\right| \right] &= \mathbb{E}\left[\mathbb{E}\left[ \left|\hat P(s,\mathbf{q}_\lambda^*,B^m) - \frac{1}{k}\right| \left| \sum_{i=1}^t X_i \right. \right] \right]\\
    & = \sum_{c=0}^t \Pr\left(\sum_{i=1}^t X_i = c \right)\left( \Pr\left( X_{t+1} = 1 \left| \sum_{i=1}^t X_i = c\right. \right)\left| c\lambda +(1-\lambda t) -\frac{1}{k} \right| \right. \\
    &\qquad\qquad\qquad\qquad\qquad\qquad+ \left. \Pr\left( X_{t+1} = 0 \left| \sum_{i=1}^t X_i = c\right. \right)\left| c\lambda -\frac{1}{k} \right| \right). \label{eq:full_form}
\end{align}
For our sampling without replacement strategy, we have
\begin{align*}
    \Pr\left(\sum_{i=1}^t X_i = c \right) &= \frac{\binom{n_s}{c}\binom{m-n_s}{t-c}}{\binom{m}{t}},\\
     \Pr\left( X_{t+1} = 1 \left| \sum_{i=1}^t X_i = c\right. \right) &= \frac{n_s-c}{m-t}.
\end{align*}
Plugging these expressions in, we have:
\begin{equation}
     \mathbb{E}\left[g(\bq_\lambda^*,B^m )\right] = \sum_{s=1}^k \sum_{c=0}^t \frac{\binom{n_s}{c}\binom{m-n_s}{t-c}}{\binom{m}{t}}\left( \left(\frac{n_s-c}{m-t} \right)\left| c\lambda +(1-\lambda t) -\frac{1}{k} \right| + \left(1-\frac{n_s-c}{m-t} \right)\left| c\lambda -\frac{1}{k} \right|  \right)
\end{equation}
When we have an equal number of elements of each kind in the permutation class and $m$ is divisible by $k$, i.e., $n_1 = \dots=n_k = m/k$, the expression simplifies to:
\begin{equation}
     \mathbb{E}\left[g(\bq_\lambda^*,B^m )\right] = k \sum_{c=0}^t \frac{\binom{m/k}{c}\binom{m-m/k}{t-c}}{\binom{m}{t}}\left( \left(\frac{ (m/k)-c}{m-t} \right)\left| c\lambda +(1-\lambda t) -\\\frac{1}{k} \right| + \left(1-\frac{ (m/k)-c}{m-t} \right)\left| c\lambda -\frac{1}{k} \right|  \right)
\end{equation}

We can simplify this even further in the special case that $\lambda = 1/k$. In this case, $t=k$, and we don't have to consider the special case of $X_{t+1}$ -- $\bq_\lambda^*$ has $k$ entries equal to $\lambda$. In this case, denoting $Z_k = \sum_{i=1}^k X_i$ \eqref{eq:full_form}, simplifies to:
\begin{align}
    \mathbb{E}\left[ \left|\hat P(s,\mathbf{q}_\lambda^*,B^m) - \frac{1}{k}\right| \right] 
    & = \frac{1}{k} \sum_{c=0}^k \Pr\left(Z_k = c \right)\left| c -1 \right|\\
    &=\frac{1}{k} \left(\Pr\left(Z_k = 0 \right) + \sum_{c=1}^k \Pr\left(Z_k = c \right)\left( c-1 \right)\right)\\
    &= \frac{1}{k} \left(2\Pr\left(Z_k = 0 \right)-1+ \mathbb{E}[Z_k]\right)\\
    &= \frac{2}{k}\Pr\left(Z_k = 0 \right)\\
    &= \frac{2}{k} \times \frac{ \binom{ (k-1)m/k }{k}}{\binom{m}{k}} 
\end{align}
and, consequently, we arrive at the elegant expression
\begin{equation}
    \mathbb{E}\left[g(\bq_\lambda^*,B^m )\right] = 2\times \frac{ \binom{ (k-1)m/k }{k}}{\binom{m}{k}}.
\end{equation}

Hence, for any given $m,k,\lambda$, that satisfies $\lambda = \frac{1}{k}$ and $m$ divisible by $k$, we have (following Eq. \eqref{eq:rdstar}):
\begin{align}
    R_d^*(\lambda) 
    &= 1 - \frac{1}{2k}-\frac{1}{4} \mathbb{E}\left[g(\bq_\lambda^*,B^m )\right]\\
    &= 1 - \frac{1}{2k}-\frac{1}{2} \frac{ \binom{ (k-1)m/k }{k}}{\binom{m}{k}}
    \label{eq:rdstar_}
\end{align}

For $1/2\leq \lambda < 1$, $\bq^*_\lambda$ has two non-zero entries equal to $\lambda$ and $1-\lambda$. Consequently, $\hat P(\bq^*_\lambda,\bb)$ assigns probability 1 to one value of $S$ if $b_1=b_2$, otherwise assigns mass $1-\lambda$ and $\lambda$ to two separate values of $s$. Thus for a fixed distribution $P_\calB$ %\textbf{[Check]}
    \begin{equation}
    \label{eq:Rd_k2}
    \mathbb{E}_{P_\calB}\left[R_d(\bq^*_\lambda,B^m )\right] =1-\frac{1}{2k}- \Pr(B_1=B_2)\times \frac{k-1}{2k} -\frac{1}{4}  \Pr(B_1\neq B_2)\times \left(1-\frac{2}{k} + \left|\lambda-\frac{1}{k}\right| + \\\left|1-\lambda - \frac{1}{k}\right| \right).
    \end{equation}
    We need to select the set $\calB$ that maximizes $\Pr(B_1\neq B_2)$. For $m$ even and $k=2$ (i.e., $\calS$ binary), $\calB$ is the permutation class of the sequence of equal number of each element, we have
    $\Pr(B_1 = B_2) =   \frac{m-2 }{2(m-1)} $, $\Pr(B_1 \neq B_2) = \frac{m }{2(m-1)}$, which simplifies $R_d(\lambda)^*$ to 
    \begin{equation}
    \label{eq:binary-bound}
         R_d^*(\lambda) = \frac{3}{4}- \frac{m\lambda -1}{4(m-1)}~\mbox{for} ~ k=2,~\frac{1}{2}\leq \lambda \leq 1. 
    \end{equation}
As $m\to \infty$, $R^\star_d(\lambda) \to \frac{3}{4}-\frac{m}{4}$.

\begin{remark}
 We make precise why in the case for $\frac{1}{2}\leq \lambda <1$, $k=2$ and $m$ even, $\calB^* = \{b^m:\text{equal number of 0's and 1's}\}$. For $S = \{0,1\}$, i.e. $k=2$, permutation classes are characterized by the number of 1$'s$. Let $\alpha$ be the number of $1's$ in $\calB$ and $m-\alpha$ be the number of $0$'s. 
From Eq \eqref{eq:Rd_k2}, we need to select the set $\calB$ that maximizes $\Pr(B_1\neq B_2)$:
\begin{align}
    \alpha^* = \argmax_{\alpha\in [m]} \Pr[B_1\neq B_2] = \argmax_{\alpha \in [m]} 2\frac{\alpha(m-\alpha)}{m(m-1)} = \frac{m}{2}.
    \label{eq:alpha*_k2}
\end{align}
   
\end{remark}

Next, we consider the case for $\frac{1}{3}\leq\lambda < \frac{1}{2}$. $\bq^*_\lambda$ has three non-zero entries: $\bq^*_\lambda = (\lambda, \lambda, 1-2\lambda, 0, ...,0)$. Consequently, there are 4 cases with the corresponding $\hat{P}(\bq^*_\lambda,\bb)$ and $g(\bq^*_\lambda,\bb)$: 
\begin{align*}
    a. & B_1 = B_2 = B_3: \quad \hat{P}= [1, 0, ...,0] \quad g(\bq^*_\lambda,\bb) = 2(1-\frac{1}{k})\\
    b. & B_1 = B_2, B_3: \neq B_1 \quad \hat{P}= [2\lambda, 1-2\lambda, 0 ...,0]  \quad g(\bq^*_\lambda,\bb) = (2\lambda - \frac{1}{k})+|1-2\lambda -
    \frac{1}{k}| + \frac{1}{k}(k-2)\\
    c. & B_1 \neq B_2, B_3 = (B_1 \vee B_2): \quad \hat{P}= [1-\lambda, \lambda, 0 ...,0] \quad  g(\bq^*_\lambda,\bb) = |\lambda - \frac{1}{k}|+|1-\lambda -
    \frac{1}{k}| + \frac{1}{k}(k-2)\\
    d. & B_1 \neq B_2 \neq B_3: \quad \hat{P}= [\lambda, \lambda, 1-2\lambda, 0 ...,0] \quad g(\bq^*_\lambda,\bb) = 2|\lambda - \frac{1}{k}|+|1-2\lambda -
    \frac{1}{k}| + \frac{1}{k}(k-3)
\end{align*}
Recall that to maximize $\mathbb{E}_{P_\calB}\left[R_d(\bq^*_\lambda,B^m )\right]$, we need to minimize $\mathbb{E}_{P_\calB}\left[g(\bq^*_\lambda,B^m )\right]$. 

For k=2, case $d$ is invalid and case $c$ produces the minimum $g(\bq^*_\lambda,\bb)$. Hence, we select the set $\calB$ that maximizes $\Pr[B_1 \neq B_2, B_3 = (B_1 \vee B_2)]$, which is equivalent to maximizing $\Pr[B_1 \neq B_2]$. Following \eqref{eq:alpha*_k2}, $\calB^* = \{b^m:\text{equal number of 0's and 1's}\}$. We have $\Pr[B_1=B_2=B_3]= \frac{m-4}{4(m-1)}$, 
$\Pr[B_1=B_2, B_3\neq B_1]= \frac{m}{4(m-1)}$ and $\Pr[B_1\neq B_2, B_3 = (B_1 \vee B_2)]= \frac{m}{2(m-1)}$.

The resulting $R_d(\lambda)^*$ is:
\begin{align}
    R^*_d(\lambda) = \frac{3}{4} - \frac{m-2}{8(m-1)} ~\mbox{for} ~ k=2,~\frac{1}{3}\leq \lambda <\frac{1}{2}.
\end{align}
As $m\to \infty$, $R^*_d(\lambda)\to \frac{5}{8}$.

% \begin{align}
%     R^*_d(\lambda) = \frac{3}{4} - \frac{m-4+3m\lambda}{16(m-1)} ~\mbox{for} ~ k=2,~\frac{1}{3}\leq \lambda <\frac{1}{2}.
% \end{align}
% As $m\to \infty$, $R^*_d(\lambda)\to \frac{3}{4}-\frac{1+3\lambda}{16}$.

\subsection{Proof of Theorem \ref{thm:approx_maxmin_detection}}

Our results so far have been based on the discussion that it is sufficient to consider the optimal distribution $P_{B^m}^*$ as one that selects $\bb$ uniformly over a single permutation class $\calB^* \in \calP_m$. Recall that $\bb$ is a sequence of $m$ elements each take a value in $\calS$: $|\bb|=m$ and $\calS=k$. Recall as well that $\calB$ can be characterized by the proportion of each element of $S$: for $i \in [k]$, denote the proportions as $[p_1,...,p_k]$, where $$p_s = \frac{\sum_{i=1}^m \ones[\bb_i==s]}{m} \quad \forall \bb \in\calB.$$ 
Hence, sampling an $\bb$ uniformly over $\calB^*$ can be equivalently defined as the following process: given $m$ elements with predefined proportions $[p_1,...,p_k]$, sample $m$ times with replacement.

To generalize the analysis for other ranges of $\lambda$, $k$, and $m$, we consider an alternative process in which rather than fixing the proportions over $m$ elements, we take $[p_1,...,p_k]$ as probabilities. $\bb$ amounts to $m$ i.i.d samples from a categorical distribution: $\bb_i \stackrel{i.i.d}{\sim} \textsc{Categorical}(p_1,...,p_k)$. Recall that optimal $B^*$ amounts to having an equal number for each element in $\calS$. Hence,  for all $i\in [k]$, $p^*_i = \frac{1}{k}$. 

Furthermore, recall that the adversarial distribution for a given min-entropy constraint $\lambda$ is: $\bq^* = [\lambda, \lambda, ..., 1-t\lambda,0,...,0]$, where $t = \lfloor{\frac{1}{\lambda}}\rfloor$. For the purpose of characterizing $\mathbb{E}_{P_\calB} g(\bq^*,\bb)$, only the color of the first $t+1$ draws matter, because the rest have 0 probabilities. 

Let $X_i \defined \ones\left(B_i = s \right)$, for a fixed $s\in \calS$. $X_i \stackrel{i.i.d}{\sim}\textsc{Ber}(\frac{1}{k})$. We can compute $\mathbb{E}_{P_\calB} g(\bq^*,\bb)$ in closed form. Following \eqref{eq:g_expectation} and \eqref{eq:full_form}, for sampling with replacement, we have:  

\begin{align}
\mathbb{E}\left[g(\bq_\lambda^*,B^m )\right] 
    &= \sum_{s=1}^k \mathbb{E}\left[ \left|\hat P(s,\mathbf{q}_\lambda^*,B^m) - \frac{1}{k}\right| \right]\\
   &= \sum_{s=1}^k \sum_{c=0}^t \Pr\left(\sum_{i=1}^t X_i = c \right)\left( \Pr\left( X_{t+1} = 1 \left| \sum_{i=1}^t X_i = c\right. \right)\left| c\lambda +(1-\lambda t) -\frac{1}{k} \right| \right. \\
    &\qquad\qquad\qquad\qquad\qquad\qquad+ \left. \Pr\left( X_{t+1} = 0 \left| \sum_{i=1}^t X_i = c\right. \right)\left| c\lambda -\frac{1}{k} \right| \right)\\
    &= k \sum_{c=0}^t \Pr[Y=c] \left(\frac{1}{k} \left|c\lambda + (1-\lambda t)-\frac{1}{k}\right| + (1-\frac{1}{k})\left|c\lambda -\frac{1}{k}\right|\right)\\
    &= \sum_{c=0}^t \Pr[Y=c] \left(\left|(c-t)\lambda + (1-\frac{1}{k})\right| + (k-1)\left|c\lambda -\frac{1}{k}\right|\right)
\end{align}
where $Y\sim Bin(t,\frac{1}{k})$, and hence $\Pr[Y=c] = \binom{t}{c} (\frac{1}{k})^c (1-\frac{1}{k})^{t-c}$

By Eq. \ref{eq:rdstar}, the approximated minimax detection is given by: 
    \begin{align}
    \tilde{R}^\star_d(\lambda) = 1 - \frac{1}{2k}- 
    \frac{1}{4}\left[\sum_{c=0}^t \Pr[Y=c] \left(\left|(c-t)\lambda + (1-\frac{1}{k})\right| + (k-1)\left|c\lambda -\frac{1}{k}\right|\right)\right]
    \end{align}

% --------------------------------------------------------------------------------------------------------------------------------------
% \begin{align}
%     \mathbb{E}_{P_\calB} g(\bq^*,\bb) &= \Pr[\text{all same color}]* g([1,0])+\\
%     &\quad  \Pr[\text{n-1 same color}]* g([\lambda, 1-\lambda])+\\
%     &\quad  \Pr[\text{n-2 same color}]* g([2\lambda, 1-2\lambda])+\\
%     &\quad  ... +\\
%     &\quad  \Pr[\text{n-(n-1) same color}]* g([\lambda, 1-\lambda])\\
%     &=  2 \binom{n}{0} \left(\frac{1}{2}\right)^n * g([1,0])+\\
%     &\quad  \binom{n}{1} \left(\frac{1}{2}\right)^n* g([\lambda, 1-\lambda])+\\
%     &\quad  \binom{n}{2} \left(\frac{1}{2}\right)^n* g([2\lambda, 1-2\lambda])+\\
%     &\quad  ... +\\
%     &\quad  \binom{n}{n-1} \left(\frac{1}{2}\right)^n* g([\lambda, 1-(n-1)\lambda])\\
%     &= (\frac{1}{2})^{n-1} + 2(\frac{1}{2})^n \sum_{i=1}^{n-1} |\frac{1}{2}-i\lambda|
% \end{align}

% By Eq. \ref{eq:rdstar}, the approximated minimax detection is given by: 
%     \begin{align}
%     \tilde{R}^\star_d(\lambda) = \frac{3}{4}-
%     \frac{1}{4}\left[\left(\frac{1}{2}\right)^{n-1}\left(1 +  \sum_{i=1}^{n-1} \left|\frac{1}{2}-i\lambda\right|\right)\right]
%     \end{align}

Finally, we analyze the approximation error of $\tilde{R}^\star_d(\lambda)$. Define $H_\bb$ and $M_\bb$ as the distribution of $\bb$ when we sample without (which yields $R^\star_d(\lambda) $) and with replacement (which yields $\tilde{R}^\star_d(\lambda)$). First, notice that $g(\bq^*,\bb) \leq \frac{2(k-1)}{k}\leq 2$ by considering the TV between singleton distribution and uniform. Then, by triangular inequality, we have: 

\begin{align}
    \left|\tilde{R}^\star_d(\lambda) - R^\star_d(\lambda)\right|
    &= \frac{1}{4} \left|(\mathbb{E}_{\bb \sim H_\bb} g(\bq^*,\bb) -\mathbb{E}_{\bb \sim M_\bb} g(\bq^*,\bb))\right| \\
    &= \frac{1}{4}\left|\sum_{\bb} g(\bq^*,\bb) (H_\bb(\bb)- M_\bb(\bb))\right|\\
    &\leq \frac{1}{4}*2 \left|\sum_{\bb}  (H_\bb(\bb)- M_\bb(\bb))\right|\\
    &\leq \frac{1}{2} \sum_{\bb} \left| (H_\bb(\bb)- M_\bb(\bb))\right|\\
    &= \TV(M_\bb, H_\bb)\\
    &\leq \frac{2k\ceil{\frac{1}{\lambda}}}{m}
\end{align}
The last inequality follows from de Finetti's Finite Exchangeable Sequences\cite{diaconis1980finite}.

\subsection{Proof of Proposition \ref{prop:sequential_Rd_bound}}
Let $n<\infty$ and assume that $X^n\sim Q^{\otimes n}$, $S^n\sim P^{\otimes n}$ and $(B^m_i)_{i=1}^n\sim P^{\otimes n}_{B^m}$.
Consequently, the CC watermarked distribution is also i.i.d. distributed according $\tilde{Q} =  Q_{X|S}$.
On Bob's end, the detection probability is given by the expression
$$
\rd = \frac{1}{2}\left(1+ \TV\left( (PQ)^{\otimes n}, (P\tilde{Q})^{\otimes n}\right) \right),
$$
where $P\tilde{Q}(S,X)=P(S)\tilde{Q}(X|S)$
To that end, we focus on obtaining bounds on the aforementioned TV term. For a pair of distributions $P,Q$, we have the following Hellinger bounds on the TV distance  \cite{polyanskiy2025information}:
\begin{equation}\label{eq:hellinger_bounds}
    \frac{1}{2}H^2(P,Q) \leq \TV(P,Q) \leq H(P,Q)\sqrt{1-\frac{1}{4}H^2\left(P,Q\right)},
\end{equation}
where, for two measures $P,Q$ on a finite alphabet $\cX$, the squarred Hellinger divergence is given by the following equivalent forms
$$
H^2(P,Q)\triangleq \mathbb{E}_Q\left[\left(1 - \sqrt{\frac{P}{Q}}\right)^2\right]=\sum_{x\in\cX}\left(\sqrt{P(x)}-\sqrt{Q(x)}\right)^2=2-2\sum_{x\in\cX}\sqrt{P(x)Q(x)}.
$$
For a pair of product distributions $(P^{\otimes n},Q^{\otimes n})$, the squarred Hellinger divergence benefits from the relation \cite{polyanskiy2025information}
$$
H^2\left(P^{\otimes n},Q^{\otimes n}\right) = 2-\left(1-\frac{1}{2}H^2(P,Q)\right)^n.
$$
Our problem therefore boils down to characterize $H^2\left(PQ,P\tilde{Q}\right)$.
We have
\begin{align*}
    H^2\left(PQ,P\tilde{Q}\right) &= \sum_{x,s}P(s)\left(\sqrt{Q(x)}-\sqrt{\tilde{Q}(x|s)}\right)^2\\
    &= \mathbb{E}_S\left[H^2(Q(X),Q(X|S))\right].
\end{align*}
For a given $s,b^m)$,we have
\begin{align*}
    H^2(Q(X),Q(X|S=s) &= 2 - 2\sum_{x}\sqrt{Q(x)\tilde{Q}(x|s)}\\
    &= 2 - 2\sum_{x}Q(x)\sqrt{\frac{P_{S| Y}(s| y(x,b^m))}{P(s)}}\\
    &= 2\mathbb{E}_{X}\left[1 - \sqrt{\frac{P_{S| Y}(s| Y(X,b^m))}{P(s)}}\right],
\end{align*}
where $P(S| Y)$ is the correlated channel.
Assuming $S\sim\mathsf{Ber}\left(\frac{1}{2}\right)$, we have
\begin{align*}
    H^2\left(PQ,P\tilde{Q}\right) &= 2\mathbb{E}_{S,X}\left[1-\sqrt{\frac{P_{S| Y}(S| Y(X,b^m))}{P(S)}}\right]\\
    &= \mathbb{E}_{ Y}\left[ 1 -\sqrt{2P(0| Y)}\right] + \mathbb{E}_{ Y}\left[1 -\sqrt{2P(1| Y)}\right]\\
    &= 2 - \sqrt{2}\mathbb{E}_{ Y}\left[P(0| Y)+P(1| Y)\right]\\
    &= 2- \sqrt{2}\left(\tilde{p}_0\left(\sqrt{p(0|0)}+\sqrt{p(1|0)}\right) + \tilde{p}_1\left(\sqrt{p(0|1)}+\sqrt{p(1|1)}\right) \right),
\end{align*}
where $ Y\sim\mathsf{Ber}(\tilde{p}_0,\tilde{p}_1)$.
Due to the symmetry of the correlated channel, we have for $\tilde{p}\triangleq\min(\tilde{p}_0,\tilde{p}_1)$
$$
H^2\left(PQ,P\tilde{Q}\right) = 2-\sqrt{2}f(\tilde{p})
$$
where
$$
f(\tilde{p})\triangleq \tilde{p} + \sqrt{\frac{1-\tilde{p}}{2}}\left(1 + \sqrt{1-2\tilde{p}}\right),
$$
which implies that
$$
H^2\left(P^{\otimes n},Q^{\otimes n}\right) = 2 - 2^{1-\frac{n}{2}}\left(f(\tilde{p})\right)^n.
$$
The bounds on the detection probability then follow by plugging the squarred Hellinger distance into \eqref{eq:hellinger_bounds}.$\hfill\square$

\section{Supplementary material for  \heavywater{} and \simplexwater{}: Watermarking Low-Entropy Text Distributions}\label{appendix:related}

\def\theequation{A.\arabic{equation}}
\def\thetable{A.\arabic{table}}
\def\thefigure{A.\arabic{figure}}
\def\thelem{A.\arabic{lem}}
\def\thedefn{A.\arabic{defn}}
\def\theprop{A.\arabic{prop}}
% \dt{Things we promised in this Appendix:
% \begin{enumerate}
%     \item Broader survey of the  watermarking literature (In Related Work)
%     \item Instantiation of existing watermark through our lenses of side information, score function and watermarked distribution (IMO musts are - RG, SynthID, Inverse transform, CC is easy, Gumbel)
%     \item Discussion on how much randomness is used by each method (at least competing)
% \end{enumerate}
% }

\subsection{Additional Information on Related Work}\label{apdx:related_work}

\paragraph{Optimization Framework} Several optimization frameworks have been proposed for watermark analysis. \cite{hetheoretically, li2024statistical, huang2023towards, long2025optimized} adopts a hypothesis-testing framework for analyzing the statistical power of watermarking schemes. \cite{li2024statistical} goes beyond the vanilla threshold test to determine the optimal detection rule by solving a minimax optimization program. 
The authors of \cite{hetheoretically} consider an optimization under an additional constraint of controlled false-positive error. The watermarking scheme follows by learning a coupling that spreads the LLM distribution into an auxiliary sequence of variables (with alphabet size greater than $m$). While the proposed scheme is the optimal solution for the considered optimization, the scheme requires access to the (proxy of) LLM logits on both generation and detection ends. The authors of \cite{wouters2023optimizing} consider a multi-objective optimization --- maximization of the green list bias while minimizing the log-perplexity. Building on the hypothesis testing framework, we optimize over classes of score functions, by observing that the detection power of a watermark boils down to the separation of expected scores between the null and alternative hypotheses.

\paragraph{Hashing Schemes in LLM Watermarking}
Current LLM watermarking schemes derive their per‐token pseudorandom seed through five recurring hashing patterns. LeftHash hashes the immediately preceding token and feeds the digest to a keyed Psaudorandom function (PRF), yielding a light‑weight, self‑synchronising seed that survives single‑token edits~\cite{kirchenbauer2023watermark}. Window‑hash generalises this by hashing the last $h$ tokens, expanding the key‑space and hindering brute‑force list enumeration at the cost of higher edit sensitivity~\cite{kirchenbauer2023watermark,sander2024watermarking}. SelfHash (sometimes dubbed right‑hand hashing'') appends the \emph{candidate} token to the left context before hashing, so the seed depends on both history and the token being scored; this hardens the scheme against key‑extraction attacks and is used in multi‑bit systems such as MPAC~\cite{yoo2023advancing}. Orthogonally, counter‑based keystreams drop context altogether and set the seed to $\mathrm{PRF}(K,\text{position})$, a strategy adopted by inverse‑transform and Gumbel watermarks to preserve the original LM distribution in expectation~\cite{aaronson2023watermark,kuditipudi2023robust}. Finally, adaptive sliding‑window hashing—popularised by \emph{SynthID‑Text}—hashes the last $H\!=\!4$ tokens together with the secret key and \emph{skips watermarking whenever that exact window has appeared before} ($K$‑sequence repeated context masking''), thereby avoiding repetition artefacts while retaining the robustness benefits of a short window~\cite{dathathri2024scalable}. Semantic extensions build on these primitives: SemaMark quantises sentence embeddings~\cite{ren2023robust}, while Semantic‑Invariant Watermarking uses a learned contextual encoder~\cite{liuadaptive}. Collectively, these hashing families balance secrecy, robustness to editing or paraphrasing, and computational overhead, offering a rich design space for practical watermark deployments. Both \simplexwater{} and \heavywater{} are agnostic to and can be applied on top of any hashing scheme. We provide the detection gain over Red-Green using various hashing schemes in Fig. \ref{fig:watermark_gain}.

\paragraph{Additional Distortion-Free Watermarks}
An array of recent works propose distortion-free watermarks that preserve the original LLM distribution \cite{kuditipudi2023robust,aaronson2023watermark,long2025optimized, huunbiased2024, christ2024undetectable, zhao2024permute, chao2024watermarking, bahri2024watermark}.
In addition to the ones that we have introduced in Section \ref{sec:intro}, 
\cite{christ2024undetectable} constructs undetectable watermarks using one-way functions, which is a cryptography-inspired technique. \cite{chao2024watermarking} proposes a watermark using error-correcting codes that leverages double-symmetric binary channels to obtain the watermarked distribution. 
\cite{bahri2024watermark} designs a distortion-free watermark based on multiple draws from a black-box LLM, which involves fixing a score function, drawing multiple tokens in each step, and outputting the highest-score token.

\subsection{Instantiation of Existing Watermarks As Score, Distribution and Randomness Design}\label{appendix:instantiation}
Existing watermarks can be represented under the proposed outlook of side information, score function and watermarked distribution from Section \ref{framework}.
We next demonstrate that by instantiating several popular schemes through our lenses.

 The \textbf{Red-Green watermark} \cite{kirchenbauer2023watermark} randomly partitions $\cX$ into a green list and a red list. Here, $S$ is a set of $m$ binary random variables, representing the random list assignment.
The function $f$ has binary outputs, which are used to increase the probability of green list tokens though exponential tilting of $P_X$.

The \textbf{SynthID} watermark \cite{dathathri2024scalable} employs tournament sampling: a tournament between a set of $N^m$ token candidates along $m$-layers with $N$ competing token groups.
Each tournament is performed given a sample of shared randomness (denoted $r$ \cite{dathathri2024scalable}).
On the $\ell$th layer the winners are taken to be the token with highest score $f_\ell(x,s)$ within each $N$-sized set.
The value of the score function $f$ is obtained from the side information for each token candidate. Different domains of $f$ induce a different construction of the function (See \cite[Appendix~E]{dathathri2024scalable} for more information).
The watermarked distribution  $P_{X|S=s}$ is the obtained with a closed form in specific cases of $(f,N,m)$ and can be directly used instead of the instantiating a tournament. This form of watermarking is termed \textit{vectorized tournament sampling}, in which the tournament is not applied, but the induced conditional distribution is employed instead.
In this paper we consider vectorized tournament sampling with binary-valued scores, as this is the formulation given in \cite{dathathri2024scalable} with a closed form, and the one that was utilized in their proposed experiments.

In the \textbf{Gumbel watermark} \cite{aaronson2023watermark}, the side information consists of $m$ i.i.d., uniform random variables on $[0,1]$, each one corresponding to a single element in the vocabulary $x\in\cX$.
The score function $f$ is obtained by assigning each $x\in\cX$ with a value $l_x = -\frac{1}{P_X(x)}\log(u_x)$, where $u_x$ the uniform variable corresponding to $x$ and $P_X(x)$ is the probability of $x\in\cX$ under the unwatermarked model.
The watermarked distribution is then given by assigning probability $1$ to $argmax_{x\in\cX}l_x(x)$ and probability $0$ to the rest of $x\in\cX$ (i.e., a singleton distribution).

In the \textbf{Correlated Channel} watermark \cite{long2025optimized}, the side information corresponds to a partition of $\cX$ into $k\geq 2$ lists and a single shared uniform variables $S'$
 that is uniformly distributed on $[1:k]$.
The score function is then an indicator of the matching between the additional variable realization and the token random assignment, i.e., $f(x,s) = \mathbf{1}(B(x) = s)$, where $B(x)\in[1:k]$ is the assignment of $x$ into one of the $k$ lists. 
The watermarked distribution is then given in closed form by solving the maximum coupling problem.

\begin{table}[!t]
    \centering
    \begin{tabular}{|c|c|}
    \hline\textbf{Watermark} & \textbf{$\#$ Bits} \\ \hline
        Red-Green \cite{kirchenbauer2023watermark} & $m$\\ 
        Inverse Transform \cite{kuditipudi2023robust} & $m\mathrm{F}$ \\
        Gumbel \cite{aaronson2023watermark}  & $m\mathrm{F}$\\
        \simplexwater{} (\textbf{ours}) & $\log(m)$ \\
        \heavywater{} (\textbf{ours}) & $\log(k)$\\ \hline
    \end{tabular}
    \vspace{3pt}
    \caption{Amount of random bits generated per step in popular watermarks. $m$ is the vocabulary size, $\mathrm{F}$ is the floating point precision and $k$ is the side information alphabet size.}
    \label{tab:num_bits}
\end{table}

\subsection{Discussion on Randomness Efficiency}\label{apdx:randomness_eff}
Given a hashing procedure that determines the random number generator seed value, we sample the side information $S\sim P_S$ over $\cS$.
The size of $\cS$ determines the amount of side information we are required to sample.
As described in Appendix \ref{appendix:instantiation}, each watermarks corresponds to a different size fo $\cS$. We interpret that as \textit{randomness efficiency}. 
That is, the bigger $\cS$ is, the more bits of randomness we are required to extract from the random seed.
We argue that a byproduct of our method is randomness efficiency, i.e., \simplexwater{} and \heavywater{} require less random bits to sampled from the random seed compared to existing schemes. 
To that end, we compare with several popular watermarks (see Table \ref{tab:num_bits} for a summary).

The Red-Green watermark \cite{kirchenbauer2023watermark} corresponds to sampling a single bit for each element in $\cX$, whose value determines the token's list assignment (red or green), thus resulting in a total of $m$ bits.
\newcommand{\rF}{\mathrm{F}}
The inverse transform watermark \cite{kuditipudi2023robust} requires sampling a single uniform  and sampling a random permutation of $[1:m]$. Thus, it asymptotically requires $\rF \log(m!)$ bits, where $\rF$ is the resolution of the floating point representation used to sample the sampled uniform variable in bits (e.g. $32$ for $\mathsf{float32}$).
For the Gumbel watermark, we sample a uniform variable for each $x\in\cX$, resulting in a total of $m\rF$ bits.

In \simplexwater{}, the side information size is $|\cS|=m-1$. To that end, we required $\log(m)$ bits to sample a single $s\in[1:m-1]$.
Furthermore, \heavywater{} is not constrained to a specific size of $|\cS|=k$, and in the proposed experiments we take $k=1024$ which is significantly smaller than both $m$ and $2^\rF$.
The resulting amount of bits to be sampled from the random seed is $\log(1024)=10$ bits.
Consequently, we observe that \heavywater{} is the most randomness-efficient watermark across considered schemes, while also being the best-performing watermark across considered experiments (see Section \ref{sec:numerics}).

Minimizing the number of random bits extracted from a random number generator directly reduces computational and energy overhead in watermark embedding as less information is to be stored on the GPU. This eases implementation in resource‑constrained hardware by lowering entropy demands and memory usage.
Additionally, it limits side‑channel leakage by shrinking an adversary’s observable output \cite{mukhopadhyay2014hardware}, which may lead to more secure watermarking.
We leave an extensive study and quantification of the benefits of randomness efficiency to future work.

\section{Proofs for Theorems from Section \ref{sec:binary} and \ref{sec:beyond_binary}.}\label{appendix:proofs}
\def\theequation{B.\arabic{equation}}
\def\thetable{B.\arabic{table}}
\def\thefigure{B.\arabic{figure}}
\def\thelem{B.\arabic{lem}}
\def\thedefn{B.\arabic{defn}}
\def\theprop{B.\arabic{prop}}

We prove Proposition \ref{prop:simplified_HD}, Theorem \ref{thm:converse_upperbound_binary}, Theorem \ref{ach}, Theorem \ref{thm:gumbel_as_ot} and Theorem \ref{thm:detection-lb-informal}

\subsection{Proof of Proposition ~\ref{prop:simplified_HD}}\label{app:prop_HD}

In this section, we prove Proposition \ref{prop:simplified_HD}, which connects the watermark design problem to a coding theoretic problem when the score function class is chosen to be binary.

\textbf{Proposition 1 (restated):} Let $\lambda\in\left[\frac{1}{2},1\right)$.
For $f\in\fbin$, define the vector $f_i=[f(i,1),\dots,f(i,k)]\!\in\!\{0,1\}^k$ for each $i\in\mathcal{X}$. Then,
\begin{align}\Dgap(m,k,\lambda,\mathcal{F}_{\mathsf{bin}})=\max_{f\in\fbin}\quad\min_{\substack{\substack{i,j\in\mathcal{X},i\neq j}}} \quad \frac{(1-\lambda) d_H(f_i,f_j)}{k},
\end{align}
where $d_H(a,b)=\sum_{i=1}^k\mathbf{1}_{\{a_i\neq b_i\}}$ denotes the Hamming distance between $a,b\in\{0,1\}^k$ and $\mathbf{1}_{\{\cdot\}}$ is the indicator function.

\begin{proof}[Proof of Prop.1] Recall that we consider $P_S \sim \mathcal{U}(1\colon\!k)$. For any $P_X$, let $\Psi(P_X)=\max_{P_{XS}}\left(\mathbb{E}_{P_{XS}}\left[f(X,S)\right]-\mathbb{E}_{P_{X}P_S}\left[f(X,S)\right] \right)$. 

\begin{Claim}\label{concave}
$\argmin_{P\in\cP_\lambda}\Psi(P)\in\cP_{\mathsf{spike},\lambda}$, where
    \begin{align}\label{pspike}
        \cP_{\mathsf{spike},\lambda} =
\left\{ P \in \Delta_m  \mid \{P_X(x_1), P_X(x_2), \dots, P_X(x_m)\} = \{\lambda, 1-\lambda, 0, 0, \dots, 0\} \right\}.
    \end{align}
Here $P_X(x_i)$ denotes the $x_i$th element of the $P_X$ probability vector with $(x_1,\dotsc,x_m)$ representing any permutation of $(1,\dotsc,m)$. $\Delta_m$ is the $m$-dimensional probability simplex.
\end{Claim} 
\begin{proof}[Proof of Claim~\ref{concave}]
To prove Claim~\ref{concave}, we first show that $\Psi(P_X)$ is concave in $P_X$.
%We first show that for any two distributions \( P_{X}^{(1)} \) and \( P_{X}^{(2)} \), and any \( \theta \in [0,1] \), we have:
%\begin{equation}
%    \Psi(\theta P_{X}^{(1)} + (1-\theta) P_{X}^{(2)}) \geq \theta \Psi(P_{X}^{(1)}) + (1-\theta) \Psi(P_{X}^{(2)}).
%\end{equation}
By definition, for any given $P_X$ and uniform $P_S$, we have, 
$$\Psi(P_X) = \max_{P_{XS}} \left( \mathbb{E}_{P_{XS}}[f(X,S)] - \mathbb{E}_{P_X P_S}[f(X,S)] \right).$$

Define $P_{XS}^*$ as the optimal joint distribution that achieves the maximum for a given \( P_X \). Then, we have:
$$\Psi(P_X) = \mathbb{E}_{P_{XS}^*}[f(X,S)] - \mathbb{E}_{P_X P_S}[f(X,S)].$$
    
Now, consider the mixture \( P_X^\theta = \theta P_X^{(1)} + (1-\theta) P_X^{(2)} \) for some $\theta\in[0,1]$ and $P_X^{(1)},P_X^{(2)}\in\Delta_m$. Given \( P_{XS}^{(1)*} \) and \( P_{XS}^{(2)*} \), the maximizing couplings of \( \Psi(P_X^{(1)}) \) and \( \Psi(P_X^{(2)}) \) respectively, we define a mixed joint distribution:
$P_{XS}^{\theta*} = \theta P_{XS}^{(1)*} + (1-\theta) P_{XS}^{(2)*}$. Note that,
\begin{align}
\sum_sP_{XS}^{\theta*}&=\theta\sum_s P_{XS}^{(1)*}+(1-\theta)\sum_s P_{XS}^{(2)*}\\  
&=\theta P_X^{(1)}+(1-\theta) P_X^{(2)}\quad \text{(as $P_{XS}^{(i)*}$ is the optimal coupling of $P_X^{(i)}$ and $P_S$. )}\\
&=P_X^{(\theta)}\\
\sum_xP_{XS}^{\theta*}&=\theta\sum_x P_{XS}^{(1)*}+(1-\theta)\sum_x P_{XS}^{(2)*}\\  
&=\frac{\theta}{k}+(1-\theta) P_X^{(2)}\quad \text{(as $P_{XS}^{(i)*}$ is the optimal coupling of $P_X^{(i)}$ and $P_S$. )}\\
&=\frac{1}{k}
\end{align}
which shows that $P_{XS}^{\theta*}$ is a valid coupling of $P_X^{(\theta)}$ and uniform $P_S$. Using the linearity of the expectation operation, the expectation under the mixed distribution $P_{XS}^{\theta *}$ is:
$$    \mathbb{E}_{P_{XS}^{\theta*}}[f(X,S)] = \theta \mathbb{E}_{P_{XS}^{(1)*}}[f(X,S)] + (1-\theta) \mathbb{E}_{P_{XS}^{(2)*}}[f(X,S)].$$

Similarly, for the independent case:
$$    \mathbb{E}_{P_{X}^{\theta} P_S}[f(X,S)] = \theta \mathbb{E}_{P_{X}^{(1)} P_S}[f(X,S)] + (1-\theta) \mathbb{E}_{P_{X}^{(2)} P_S}[f(X,S)].$$

Since $\Psi(P_X^\theta)$ by definition takes the maximum coupling among all $P_{XS}$, it upper bounds the value attained by the specific mixture $P_{XS}^{\theta*}$. Thus,
\begin{align*}
    \Psi(P_X^\theta)&\geq \mathbb{E}_{P_{XS}^{\theta*}}[f(X,S)] - \mathbb{E}_{P_{X}^{\theta} P_S}[f(X,S)] \\
    &= \theta \mathbb{E}_{P_{XS}^{(1)*}}[f(X,S)] + (1-\theta) \mathbb{E}_{P_{XS}^{(2)*}}[f(X,S)] -\theta \mathbb{E}_{P_{X}^{(1)} P_S}[f(X,S)]\nonumber\\
    &\qquad- (1-\theta) \mathbb{E}_{P_{X}^{(2)} P_S}[f(X,S)])\\
    &= \theta \Psi(P_X^{(1)}) + (1-\theta) \Psi(P_X^{(2)}),
\end{align*}
which proves concavity.

We now show that for any \( \lambda \in \left[\frac{1}{2},1\right] \), the minimizer of \( \Psi(P_X) \) lies in the set \( \mathcal{P}_{\mathsf{spike},\lambda} \),where
\begin{align}\label{two-token}
        \cP_{\mathsf{spike},\lambda} =
\left\{ P \in \Delta_m  \mid \{P_X(x_1), P_X(x_2), \dots, P_X(x_m)\} = \{\lambda, 1-\lambda, 0, 0, \dots, 0\} \right\},
    \end{align}

Recall that $\cP_\lambda=\{P\in\Delta_m,\|P\|_\infty\leq\lambda\}$. This is a convex set and the extreme points of this set are precisely the spike distributions in \( \mathcal{P}_{\mathsf{spike},\lambda} \), which correspond to permutations of \( (\lambda, 1-\lambda, 0, \dots, 0) \).
Since \( \Psi(P_X) \) is concave, its minimum over the convex set \( \mathcal{P}_\lambda \) occurs at an extreme point of \( \mathcal{P}_\lambda \). 
%In fact, any extreme point of \( \mathcal{P}_\lambda \) yields the same value as the function $\Psi(P_X)$ is symmetric with respect to permutations of the coordinates of $P_X$. Given any bijection $\sigma: [1:m]\mapsto [1:m]$ which correspond to a permutation, we have $F(\sigma(P_X)) = F(P_X)$, since a permutation only relabels $X$ while maintaining the same structure:
%\begin{align*}
%    F(\sigma(P_X)) &= \max_{P_S \circ  \sigma(P_{X|S})} \left(\mathbb{E}_{P_{S}} \mathbb{E}_{\sigma(P_{X|S})}[f(X,S)] - \mathbb{E}_{\sigma(P_X) P_S}[f(X,S)] \right)\\
 %   &= \max_{P_{XS}} \left( \mathbb{E}_{P_{XS}}[f(X,S)] - \mathbb{E}_{P_X P_S}[f(X,S)] \right)\\
 %   &= F(P_X)
%\end{align*}
Hence, the minimizer of \( \Psi(P_X) \) belongs to \( \mathcal{P}_{\mathsf{spike},\lambda} \), which completes the proof.
\end{proof}
Using Claim~\ref{concave}, we prove Proposition~\ref{prop:simplified_HD} as follows. Let $p^\star\in\cP_{\mathsf{spike},\lambda}$ be the minimizer in the minimization of $\Dgap(m,k,\lambda,\mathcal{F}_{\mathsf{bin}})$ in \eqref{D_gap}, and let $(i,j)$ be its non-zero entries, i.e., $p^\star_i=\lambda$ and $p^\star_j=1-\lambda$.
Under such $p^\star$ and uniform $P_S$, any coupling can be written as 
\[
P_{XS}(x,s)=\begin{cases}
\lambda Q_{X|S}(s\mid i) & \text{if } x=i,\\[1mm]
(1-\lambda)Q_{X|S}(s\mid j) & \text{if } x=j\\
0 & \text{if } x\neq i,j
\end{cases}
\]
where $Q_{S|X}=P_{XS}/P_X$ denotes the conditional distribution of the shared randomness $S$ given the selected token $X$. The coupling constraint for each $s\in\cS$ becomes
\[
\lambda Q_{S|X}(s\mid i)+(1-\lambda)Q_{S|X}(s\mid j)=\frac{1}{k},
\]
To that end, we represent such coupling as 
$$
Q_{S|X}(s\mid i)=a_s,\quad Q_{S|X}(s\mid j)=\frac{\frac{1}{k}-\lambda a_s}{1-\lambda}
$$
for some set of parameters $(a_s)_{s\in\cS}$, such that $a_s\in[0,\frac{1}{k\lambda}]$.
Under this construction, considering the worst case $(i,j)$ pair of non-zero $P_X$ indices, we have,
\begin{align}
    \Dgap\left(m,k,\lambda,\mathcal{F}_{\mathsf{bin}}\right)&=\max_{f:\mathcal{X}\times\mathcal{S}\mapsto[0,1]}\min_{i\neq j}\max_{Q_{X|S}} \sum_{s=1}^k\lambda Q_{S|X}(s|i)f(i,s)+(1-\lambda)Q_{S|X}(s|j)f(j,s)\nonumber\\
    &\qquad-\left(\frac{\lambda}{k}f(i,s)+\frac{1-\lambda}{k}f(j,s)\right)\\
    &=\max_{f:\mathcal{X}\times\mathcal{S}\mapsto[0,1]}\min_{i\neq j}\max_{a_s\in[0,\frac{1}{k\lambda}]} \sum_{s=1}^k\lambda a_sf(i,s)+(1-\lambda)\left(\frac{\frac{1}{k}-\lambda a_s}{1-\lambda}\right)f(j,s)\nonumber\\
    &\qquad-\left(\frac{\lambda}{k}f(i,s)+\frac{1-\lambda}{k}f(j,s)\right)\\
    &=\max_{f:\mathcal{X}\times\mathcal{S}\mapsto[0,1]}\min_{i\neq j}\max_{a_s\in[0,\frac{1}{k\lambda}]} \sum_{s=1}^k \lambda\left(a_s-\frac{1}{k}\right)(f(i,s)-f(j,s))\label{last2}
\end{align}
The design of the coupling $P_{XS}$ boils down to choosing $\{a_s\}_{s\in\cS}$ such that \eqref{last2} is maximized. Moreover, since $\sum_{s=1}^kQ_{S|X}(s|x)=1$ for $x=i$ and $x=j$, we have,
\begin{align}
    \sum_{s=1}^kQ_{S|X}(s|x)=1\quad\implies\quad\sum_{s=1}^ka_s=1
\end{align}
For a given $f$ and any two indices $(i,j)$, we characterize optimal $a_s$ as follows. Let $m_+^{i,j}=\#\{s:f(i,s)-f(j,s)=1\}$ and $m_-^{i,j}=\#\{s:f(i,s)-f(j,s)=-1\}$ where $\#$ denotes the cardinality of a set. Assume that the score functions $f$ satisfy $m_+^{i,j}\leq k\lambda$ and $m_-^{i,j}+m_-^{i,j}<k$. These assumptions are needed to eliminate trivial edge cases for which the inner minimization in \eqref{last2} is zero, which leads to zero detection (see Appendix~\ref{edge_case}). The inner-most maximum in \eqref{last2} is obtained when:
\begin{align}
    a_s&=\begin{cases}
        \frac{1}{k\lambda},& s:f(i,s)-f(j,s)=1\\
        0, & s:f(i,s)-f(j,s)=-1\\
        \frac{1-\frac{1}{k\lambda}m_+^{i,j}}{k-m_+^{i,j}-m_-^{i,j}},& s:f(i,s)-f(j,s)=0
    \end{cases}
\end{align}
The optimal values of $a_s$ are obtained by allocating the maximum probability mass to the highest scores $f(i,s)-f(j,s)$ in \eqref{last2}.
%where $m_+^{i,j}=|s:f(i,s)-f(j,s)>0|$ and
Continuing from \eqref{last2}, we have,

\begin{align}
    \Dgap\left(m,k,\lambda,\mathcal{F}_{\mathsf{bin}}\right)&=\max_{f:\mathcal{X}\times\mathcal{S}\mapsto[0,1]}\min_{i\neq j} \frac{1}{k}\sum_{s:f_i-f_j=-1} \lambda|f(i,s)-f(j,s)|\nonumber\\
    &\qquad+\frac{1}{k}\sum_{s:f_i-f_j=1} (1-\lambda)|f(i,s)-f(j,s)|\\
&=\max_{f:\mathcal{X}\times\mathcal{S}\mapsto[0,1]}\min_{i\neq j} \frac{1}{k}(\lambda m_-^{i,j}+(1-\lambda)m_+^{i,j})\label{cont2}
\end{align}
Let $d_{ij}=\#\{s:f(i,s)\neq f(j,s)\}$. Then, $m_{-}^{ij}+m_{+}^{ij}=d_{ij}$. Let $m_{-}^{ij}=\beta d_{ij}$ and $m_{+}^{ij}=(1-\beta) d_{ij}$ for some $\beta\in[0,1]$. From \eqref{cont2}, we have,
\begin{align}
   \Dgap\left(m,k,\lambda,\mathcal{F}_{\mathsf{bin}}\right)&=\max_{f:\mathcal{X}\times\mathcal{S}\mapsto[0,1]}\min_{i\neq j}\min_{\beta\in[0,1]} \frac{1}{k}(\lambda \beta+(1-\lambda)(1-\beta))d_{i,j}\\
    &=\max_{f:\mathcal{X}\times\mathcal{S}\mapsto[0,1]}\min_{i\neq j}\min_{\beta\in[0,1]} \frac{1}{k}(\beta(2\lambda-1)+(1-\lambda))d_{i,j}\\
    &=\max_{f:\mathcal{X}\times\mathcal{S}\mapsto[0,1]}\min_{i\neq j} \frac{d_{i,j}}{k}(1-\lambda)
\end{align}
since the inner minimum is achieved when $\beta=0$, as $\lambda\geq\frac{1}{2}$. The proof is complete as $d_{ij}$ is exactly the Hamming distance between $f(i,\cdot)$ and $f(j,\cdot)$.

\end{proof}

\subsubsection{Edge Cases}\label{edge_case}

In Appendix~\ref{app:prop_HD}, we assumed that the score functions $f$ satisfy $m_+^{i,j}\leq k\lambda$ and $m_+^{i,j}+m_-^{i,j}<k$. In this section, we show what happens when these constraints are violated.

\textbf{Case 1: $m_+^{i,j}> k\lambda$: } In this case, we obtain the optimal values of $a_s$ as in the proof of Proposition ~\ref{prop:simplified_HD} above, by assigning the probability masses to the high score cases as follows. Let $\mathcal{J}\subset m_+^{i,j}$, $|\mathcal{J}|=k\lambda$ be any subset of $k\lambda$ values of $s$, each satisfying $f(i,s)-f(j,s)=1$. In this case, similar to case 1, the inner maximum in \eqref{last2} is obtained when: 
\begin{align}
    a_s&=\begin{cases}
        \frac{1}{k\lambda},& s\in\mathcal{J}\\
        0, & s\notin\mathcal{J}
    \end{cases}
\end{align}
Continuing from \eqref{last2}, we have,
\begin{align}
    \Dgap\left(m,k,\lambda,\mathcal{F}_{\mathsf{bin}}\right)&=\max_{f:\mathcal{X}\times\mathcal{S}\mapsto[0,1]}\min_{i\neq j} \lambda(1-\lambda)-\frac{\lambda(m_{+}^{i,j}-k\lambda)}{k}+\frac{\lambda m_{-}^{i,j}}{k}\\
    &=\max_{f:\mathcal{X}\times\mathcal{S}\mapsto[0,1]}\min_{i\neq j} \lambda+\frac{\lambda}{k}(m_-^{i,j}-m_+^{i,j})\\
    &=\max_{f:\mathcal{X}\times\mathcal{S}\mapsto[0,1]}\min_{i\neq j}\min_{\beta\in[0,1]} \lambda+\frac{\lambda}{k}(2\beta-1)d_{ij}\\
    &=\max_{f:\mathcal{X}\times\mathcal{S}\mapsto[0,1]}\min_{i\neq j}\lambda\left(1-\frac{1}{k}d_{ij}\right)\\
    &=0
\end{align}
where the last equality follows from the fact that the inner minimum is achieved when $d_{ij}=k$, i.e., $f(i,\cdot)$ is the all zeros vector and $f(j,\cdot)$ is the all ones vector. The score functions $f$ that result in such cases are uninteresting as the detection can not be improved.

\textbf{Case 2: $m_+^{i,j}\leq k\lambda$ and $m_+^{i,j}+m_-^{i,j}=k$: } In this case, we obtain the optimal values of $a_s$ as in Appendix~\ref{app:prop_HD}, by assigning the probability masses to the high score cases as follows. Note that in this case $|s:f(i,s)-f(j,s)=0|=0$. Therefore,
\begin{align}
    a_s&=\begin{cases}
        \frac{1}{k\lambda},& s:f(i,s)-f(j,s)=1\\
        \frac{1-\frac{1}{k\lambda}m_+^{i,j}}{k-m_+^{i,j}},& s:f(i,s)-f(j,s)=-1
    \end{cases}
\end{align}
Continuing from \eqref{cont2}, we have,
\begin{align}
   \Dgap\left(m,k,\lambda,\mathcal{F}_{\mathsf{bin}}\right)&=\max_{f:\mathcal{X}\times\mathcal{S}\mapsto[0,1]}\min_{i\neq j} \frac{(1-\lambda)m_+^{ij}}{k}-\lambda\left(\frac{1}{k}-\frac{1-\frac{1}{k\lambda}m_+^{ij}}{k-m_+^{i,j}}\right)(k-m_+^{ij})\\
    &=\max_{f:\mathcal{X}\times\mathcal{S}\mapsto[0,1]}\min_{i\neq j}\frac{2m_+^{ij}}{k}(1-\lambda)\\
    &=0
\end{align}
where the last equality follows from the fact that the inner minimum is achieved when $m_+^{ij}=0$, i.e., $f(i,\cdot)$ is the all zeros vector and $f(j,\cdot)$ is the all ones vector. The score functions $f$ that result in such cases are uninteresting as the detection can not be improved.

\subsection{Proof of Theorem~\ref{thm:converse_upperbound_binary}.}\label{apdx:proof_thm_1}

In this section, we derive an upper bound for the detection gap in \eqref{D_gap} using the Plotkin bound from coding theory \cite{roth2006introduction}.

\textbf{Theorem 1 (restated):}
Consider the class of binary score functions $\mathcal{F}_{\mathsf{bin}}$ and uniform $P_S$. Then, for any $\lambda\in\left[\frac{1}{2},1\right)$, the maximum detection gap can be bounded as 
\begin{align}
\Dgap(m,k,\lambda,\mathcal{F}_{\mathsf{bin}})\leq\frac{m(1-\lambda)}{2(m-1)}
\end{align}
\begin{proof}[Proof of Thm.~\ref{thm:converse_upperbound_binary}]
Thm.~\ref{thm:converse_upperbound_binary} follows directly from the Plotkin bound \cite{roth2006introduction}, which provides an upper bound on the minimum normalized Hamming distance between any two codewords, considering any code construction. Indeed,
\begin{align}\Dgap(m,k,\lambda,\mathcal{F}_{\mathsf{bin}})=\max_{f\in\fbin}\quad\min_{\substack{\substack{i,j\in\mathcal{X},i\neq j}}} \quad \frac{(1-\lambda) d_H(f_i,f_j)}{k}\leq (1-\lambda)\frac{m}{2(m-1)}.\label{ub_pf}
\end{align}

\end{proof}

\subsection{Proof of Theorem ~\ref{ach}.}\label{apdx:proof_thm_2}

In this section, we show that \simplexwater{} achieves the upper bound in \eqref{ub_pf}.

Theorem~\ref{ach} (restated): For any $\lambda\in\left[\frac{1}{2},1\right)$ the maximum detection gap upper bound \eqref{eq:ub_binary} is attained by \simplexwater{}.

\begin{proof}
    \simplexwater{} uses the Simplex code construction in Definition~\ref{simplex} as the score function. The simplex code achieves the Plotkin bound \cite{roth2006introduction}, i.e.,
    \begin{align}
        \min_{i\neq j} \frac{d_H(\fsim(i,\cdot),\fsim(j,\cdot))}{k}=\frac{m}{2(m-1)}
    \end{align}
    Therefore,
    \begin{align}\Dgap(m,k,\lambda,\mathcal{F}_{\mathsf{bin}})\geq\min_{\substack{\substack{i,j\in\mathcal{X},i\neq j}}}\frac{(1-\lambda) d_H(\fsim(i,\cdot),\fsim(j,\cdot))}{k}= (1-\lambda)\frac{m}{2(m-1)}.\label{lb_pf}
\end{align}
Considering the upper and lower bounds in \eqref{ub_pf} and \eqref{lb_pf}, we have,
\begin{align}\Dgap(m,k,\lambda,\mathcal{F}_{\mathsf{bin}})= (1-\lambda)\frac{m}{2(m-1)},
\end{align}
which is achieved by \simplexwater{}.
\end{proof}

\subsection{Proof of Theorem \ref{thm:gumbel_as_ot}.}\label{sec:appendix_proof_gumbel}

In this section, we establish that the Gumbel watermark scheme \cite{aaronson2023watermark} can be understood within our optimal transport framework, i.e., we prove Theorem \ref{thm:gumbel_as_ot}, restated below.

\begin{Theorem}[Gumbel Watermark as OT]
When the score random variables $f(x, s)$, are sampled i.i.d. from $\text{Gumbel}(0,1)$, the solution to the OT problem in \eqref{eq:OT_problem} converges to the Gumbel watermark \cite{aaronson2023watermark} as $|\mathcal{S}|=k \to \infty$. %Specifically, the argmax sampling strategy in Gumbel watermarking corresponds to the optimal coupling derived from our optimal transport formulation.
%\begin{equation}
%P_X(i) = \mathbb{P}(\arg\max_j [u_j/T + G_j] = i) = \mathbb{E}[\mathbbm{1}\{i \in \arg\max_j [f(j,s) - \alpha^*_j]\}]
%\end{equation}
%where $u_j/T$ are the scaled logits and $G_j$ are Gumbel-distributed random variables.
\end{Theorem}

To prove this, we first write down the Kantorovich dual of our optimal transport formulation and identify the dual potentials $\{\alpha_x,\beta_s\}$ that generate the arg-max coupling. We then analyze the behavior of these potentials in the limit $k \rightarrow \infty$, showing that the resulting arg-max rule converges to the classical Gumbel-Max sampling procedure. A final concentration argument ensures that the random coupling concentrates around its expectation, thereby establishing that the OT‐derived sampler coincides with the Gumbel watermarking scheme.

To understand this connection, we first review the Gumbel watermarking scheme. The Gumbel-Max trick states that sampling from a softmax distribution can be equivalently expressed as:
\begin{equation}
y_t = \arg\max_{y \in \mathcal{V}} \frac{u_t(y)}{T} + G_t(y)
\end{equation}
where $u_t(y)$ are the logits, $T$ is the temperature parameter, and $G_t(y) \sim \text{Gumbel}(0,1)$ independently for each token position $t$ and vocabulary element $y$. A Gumbel$(0,1)$ random variable can be generated via:
\begin{equation}
G_t(y) = -\log(-\log(r_t(y)))
\end{equation}
where $r_t(y) \sim \text{Uniform}([0,1])$.

In the Gumbel watermarking scheme, the uniform random variables are replaced with pseudo-random values:
\begin{align}
r_t(y) &\sim \text{Uniform}([0,1]) \text{ (in unwatermarked model)} \\
r_t(y) &= F_{y_{t-m:t-1},\mathbf{k}}(y) \text{ (in watermarked model)}
\end{align}

Here, $F_{y_{t-m:t-1},\mathbf{k}}(y)$ uses a secret key $\mathbf{k}$ and previously generated tokens $y_{t-m:t-1}$ to deterministically generate values that appear random without knowledge of $\mathbf{k}$.

To establish the connection to our OT framework \eqref{eq:OT_problem}, in the case when randomness is i.i.d. generated, we analyze the dual formulation of the optimal transport problem. In this formulation, we seek to minimize $\sum_x P_X(x)\alpha_x + \frac{1}{k}\sum_s \beta_s$ subject to $\alpha_x + \beta_s \geq f(x,s)$. The optimal values for these dual variables satisfy specific conditions that link to the Gumbel-Max construction.
The key insight comes from the optimality condition in our framework:
\begin{equation}
P_X(i) = \mathbb{P}(\arg\max_j [f(j, s) - \alpha^*_j] = i)
\end{equation}
This can be directly mapped to the Gumbel-Max trick by identifying that $f(j, s)$ corresponds to the Gumbel noise $G_t(y)$ and $\alpha^*_j$ corresponds to $-\frac{u_t(y)}{T}$ (the negative normalized logits). With these identifications, the Gumbel-Max sampling expression:
\begin{equation}
y_t = \arg\max_{y \in \mathcal{V}} \frac{u_t(y)}{T} + G_t(y) = \arg\max_{y \in \mathcal{V}} [G_t(y) - (-\frac{u_t(y)}{T})]
\end{equation}

Takes exactly the same form as our framework's expression $\arg\max_j [z(j, s) - \alpha^*_j]$. Further, the probability that this argmax equals $i$ is precisely $P_X(i)$ in our framework. Therefore, when $f(j,s) \sim \text{Gumbel}(0,1)$, we will show below that our optimal transport solution exactly recovers the Gumbel watermarking scheme.

The detailed proof proceeds by analyzing the convergence of the discretized dual problem to its continuous limit as $k \to \infty$. The optimality conditions in the limit confirm that our OT framework generalizes the Gumbel watermark as a special case when scores are drawn from the Gumbel distribution. We start by defining a general optimal transport problem and its Kantorovich duality. 

\begin{definition}[Optimal Transport Problem]
Given probability measures $\mu$ and $\nu$ on spaces $\mathcal{X}$ and $\mathcal{Y}$ respectively, and a cost function $c: \mathcal{X} \times \mathcal{Y} \rightarrow \mathbb{R}$, the optimal transport problem seeks to find a coupling $\pi$ (a joint distribution with marginals $\mu$ and $\nu$) that minimizes the expected cost:
\begin{align}
\inf_{\pi \in \Pi(\mu, \nu)} \int_{\mathcal{X} \times \mathcal{Y}} c(x,y) \, d\pi(x,y)
\end{align}
where $\Pi(\mu, \nu)$ is the set of all probability measures on $\mathcal{X} \times \mathcal{Y}$ with marginals $\mu$ and $\nu$.
\end{definition}

The Kantorovich duality is a fundamental result that provides an equivalent formulation of this problem. For more details, see \cite[Chapter 5]{villani2009optimal}.

\begin{thm}[Kantorovich Duality]
The optimal transport problem is equivalent to:
\begin{align}
\sup_{(\varphi, \psi) \in \Phi_c} \left\{ \int_{\mathcal{X}} \varphi(x) \, d\mu(x) + \int_{\mathcal{Y}} \psi(y) \, d\nu(y) \right\}
\end{align}
where $\Phi_c = \{(\varphi, \psi) : \varphi(x) + \psi(y) \leq c(x,y) \}$ is the set of functions satisfying the c-inequality constraint.
\end{thm}

This duality relationship (i.) transforms a non-trivial constrained optimization problem over probability measures into an optimization over functions, (ii.) provides a way to certify optimality through complementary slackness conditions and (iii.) enables us to analyze the convergence properties of OT problems through the convergence of dual objective functions. In many applications, the Kantorovich duality is rewritten with the constraint reversed (potential functions sum to $\geq c$), transforming the problem into a minimization rather than maximization. We adopt this convention in our formulation below.

\textbf{Primal Formulation.} Let us consider an optimal transport problem described by the inner maximization of \eqref{D_gap}, that is, given by Equation \eqref{eq:OT_problem}. Recall that given a pair $(P_X,f)$, the inner maximization in \eqref{D_gap} amounts to an OT problem between $P_X$ and $P_S$, which is set to follow an uniform distribution on $[1:k]$. There, the score function $f$ can be equivalently denoted as the $(m\times k)$-dimensional OT cost matrix $\rC$, which is defined as $\rC_{x',s'}=-f(x',s')$ for $(x',s')\in [1:m]\times[1:k]$. This matrix is only generated once. 

Now, let $P_X$ be a probability distribution over a finite set $\{1,2,\ldots,m\}$ and $P_S$ be a uniform distribution with mass $1/k$ at each point in $\{1,2,\ldots,k\}$. We have a cost function $f(x,s)$ with zero mean and unit variance, i.e., $\mathbb{E}_P[z] = 0$ and $\mathbb{E}_P[z^2] = 1$. The inner optimization in \eqref{D_gap} aims to find a coupling $P_{XS}$ that maximizes:
\begin{align}
C_{P,k}^* = \max_{P_{XS}} \sum_{x,s} P_{XS}(x,s) \cdot f(x,s)
\end{align}
\begin{align}
\text{s.t.} \sum_s P_{XS}(x,s) &= P_X(x) \quad \forall x\\
\sum_x P_{XS}(x,s) &= \frac{1}{k} \quad \forall s\\
P_{XS}(x,s) &\geq 0 \quad \forall x,s
\end{align}

\textbf{Dual Formulation.} In order to analyze this primal optimal transport problem we can look at the dual by using a Kantorovich duality argument. The equivalent dual problem is given by:
\begin{align}
C_{D,k}^* = \min_{\alpha_x, \beta_s} \sum_x P_X(x)\alpha_x + \sum_s \frac{1}{k}\beta_s
\end{align}
\begin{align}
\text{Subject to:} \ \alpha_x + \beta_s \geq f(x,s) \quad \forall x,s
\end{align}

In this dual formulation, $\alpha_x$ and $\beta_s$ are the Kantorovich potentials (i.e., Lagrange multipliers) enforcing the marginal constraints $\sum_s P_{XS}(x,s) = P_X(x)$ and $\sum_x P_{XS}(x,s) = \frac{1}{k}$ respectively.
For any fixed values of $\alpha_x$, we need to determine the optimal values of $\beta_s$ that minimize the dual objective function. Since we aim to minimize the objective and the coefficients of $\beta_s$ are positive ($\frac{1}{k} > 0$), we want to make each $\beta_s$ as small as possible while still satisfying the constraints. For a given $s$, the constraint becomes:
\begin{align}
\beta_s \geq f(x,s) - \alpha_x \quad \forall x
\end{align}

This means that $\beta_s$ must be at least as large as $f(x,s) - \alpha_x$ for every value of $x$. To ensure all constraints are satisfied while keeping $\beta_s$ minimal, we set:
\begin{align}
\beta_s = \max_{x'}[f(x',s) - \alpha_{x'}]
\end{align}

This is the smallest value of $\beta_s$ that satisfies all constraints for a given $s$. Any smaller value would violate at least one constraint, and any larger value would unnecessarily increase the objective function. Substituting this expression back into the dual objective yields:
\begin{align}
C_{D,k}^* = \min_{\alpha_x} \sum_{x=1}^m P_X(x)\alpha_x + \frac{1}{k}\sum_{s=1}^k \max_{x'}[f(x',s) - \alpha_{x'}]
\end{align}

This reformulation reduces the dual problem to an unconstrained minimization over $\alpha_x$ only. 

To establish convergence of the dual problem, which is the key to embedding the Gumbel watermarking scheme in our framework, we will first recall a few fundamental concepts from variational analysis and convex optimization.

\begin{definition}[Epi-convergence]
A sequence of functions $f_k: \mathbb{R}^n \to \mathbb{R} \cup \{\pm \infty\}$ is said to epi-converge to a function $f: \mathbb{R}^n \to \mathbb{R} \cup \{\pm \infty\}$ if the following two conditions hold:
\begin{enumerate}[label=(\roman*)]
\item For every $x \in \mathbb{R}^n$ and every sequence $x_k \to x$, $\liminf_{k \to \infty} f_k(x_k) \geq f(x)$.
\item For every $x \in \mathbb{R}^n$, there exists a sequence $x_k \to x$ such that $\limsup_{k \to \infty} f_k(x_k) \leq f(x)$.
\end{enumerate}
\end{definition}

Epi-convergence is particularly important in optimization because it guarantees that minimizers and minimal values converge appropriately. For more details on epi-convergence, see \cite{rockafellar2009variational}.

\begin{definition}[Equi-lower semicontinuity]
A family of functions $\{f_k\}$ is equi-lower semicontinuous if for every $x \in \mathbb{R}^n$ and every $\varepsilon > 0$, there exists a neighborhood $V$ of $x$ such that
\begin{align}
\inf_{y \in V} f_k(y) > f_k(x) - \varepsilon
\end{align}
for all $k$.
\end{definition}

The following result connects pointwise convergence with epi-convergence for convex functions:

\begin{thm}[{\cite[Theorem 2]{salinetti1977relations}}]
\label{lemma:epi-convergence}
If $\{f_k\}$ is a sequence of convex, continuous, and equi-lower semicontinuous functions that converge pointwise to a function $f$ on $\mathbb{R}^n$, then $\{f_k\}$ epi-converges to $f$.
\end{thm}

As a first step toward applying our framework to the Gumbel watermarking scheme, we now characterize how the optimal transport cost behaves in the limit $k \to \infty$.

\begin{thm}\label{thm:minimizer_convergence}
As $k \to \infty$, the optimal value of the discrete problem converges to the expected value problem:
\begin{align}
C_{D,k}^* \to C_D^* = \min_{x \in \mathbb{R}^m} p^T x + \mathbb{E}[\max_j(f(j,s) - x_j)]
\end{align}
where $p_j = P_X(j)$ and $f(j,s)$ are random variables with the distribution matching the problem's cost function.
\end{thm}

\begin{proof}[Proof of Theorem \ref{thm:minimizer_convergence}]
    
Let us rewrite the objective function in vector notation:
\begin{align}
f_k(x) = p^T x + \frac{1}{k}\sum_{s=1}^k \max_j[f(j,s) - x_i]
\end{align}
where $\vec{z}_j = [z_{1,j}, \ldots, z_{m,j}]$ and $z_{i,j}:=f(i,s_j)$ denote the sampled scores, i.e., each column $\vec{z}_j$ corresponds to the vector of scores across tokens for a fixed side-information sample $s_j$, and $x = [\alpha_1, \ldots, \alpha_m]$.

We need to establish that the functions $f_k$ are continuous and convex. Note that

\begin{enumerate}
\item The term $p^T x$ is linear and therefore continuous and convex.
\item The function $g_s(x) = \max_j[f(j,s) - x_j]$ is continuous for each $j$ because:
   \begin{enumerate}
   \item The functions $h_j(x) = f(j,s) - x_j$ are continuous for each $j$.
   \item The maximum of a finite number of continuous functions is continuous. It is also convex as the maximum of linear functions.
   \end{enumerate}
\item The sum $\frac{1}{k}\sum_{s=1}^k g_s(x)$ is continuous as a linear combination of continuous functions and convex as a positive linear combination of convex functions.
\end{enumerate}

As $k \to \infty$, by the Law of Large Numbers, for any fixed $x$, the sample average converges to the expected value:
\begin{align}
\frac{1}{k}\sum_{s=1}^k \max_j[f(j,s) - x_j] \to \mathbb{E}[\max_j(f(j,s) - x_j)]
\end{align}

Therefore, 
\begin{align}
f_k(x) \to f(x) = p^T x + \mathbb{E}[\max_i(z_i - x_i)]
\end{align}

This establishes pointwise convergence of $f_k$ to $f$. Next, we need to prove that the family of functions $\{f_k\}$ is equi-lower semicontinuous.

\begin{lemma}
The family of functions $\{f_k\}$ defined above is equi-lower semicontinuous under mild assumptions on the boundedness of the data points $\vec{z}_j$.
\end{lemma}

\begin{proof}
First, observe that each $f_k$ is continuous (and hence lower semicontinuous). For any $x \in \mathbb{R}^n$ and $\varepsilon > 0$, consider the neighborhood 
\begin{align}
V = \{y : \|y - x\|_\infty < \varepsilon/2\}
\end{align}

For any $y \in V$ and any $j, i$, we have
\begin{align}
z_{i,j} - y_i > z_{i,j} - x_i - \varepsilon/2
\end{align}

Therefore,
\begin{align}
\max_i[z_{i,j} - y_i] \geq \max_i[z_{i,j} - x_i - \varepsilon/2] = \max_i[z_{i,j} - x_i] - \varepsilon/2
\end{align}

This implies
\begin{align}
\frac{1}{k}\sum_{j=1}^k \max_i[z_{i,j} - y_i] \geq \frac{1}{k}\sum_{j=1}^k \max_i[z_{i,j} - x_i] - \varepsilon/2
\end{align}

Combined with the linear term, we get
\begin{align}
f_k(y) &= p^T y + \frac{1}{k}\sum_{j=1}^k \max_i[z_{i,j} - y_i]\\
&\geq p^T x - \|p\|_1 \cdot \varepsilon/2 + \frac{1}{k}\sum_{j=1}^k \max_i[z_{i,j} - x_i] - \varepsilon/2\\
&= f_k(x) - \|p\|_1 \cdot \varepsilon/2 - \varepsilon/2
\end{align}

By choosing $\varepsilon$ small enough, we ensure that $\|p\|_1 \cdot \varepsilon/2 + \varepsilon/2 < \varepsilon$, which gives us
\begin{align}
\inf_{y \in V} f_k(y) > f_k(x) - \varepsilon
\end{align}

This holds for all $k$, establishing the equi-lower semicontinuity of the family $\{f_k\}$.
\end{proof}

Since we have established that the functions $f_k$ are convex, continuous, and equi-lower semicontinuous, and that they converge pointwise to $f$, we can apply Lemma \ref{lemma:epi-convergence} to conclude that $f_k$ epi-converges to $f$. By the fundamental properties of epi-convergence, if $f_k$ epi-converges to $f$, then $\min f_k \to \min f$. Also, every limit point of minimizers of $f_k$ is a minimizer of $f$. This establishes that $C_{D,k}^* \to C_D^*$ as $k \to \infty$.
\end{proof}

\textbf{Optimality Conditions and Characterization.} Now, we analyze the optimality conditions for the dual problem. Consider our objective function:
\begin{align}
f_k(\alpha) = \sum_{i=1}^m P_X(i)\alpha_i + \frac{1}{k}\sum_{s=1}^k \max_j[f(j,s) - \alpha_j]
\end{align}

To find the derivative with respect to $\alpha_i$, we note that the derivative of the first term is simply $P_X(i)$. For the second term, we need to understand how the max function behaves. At points where a unique index $j$ achieves the maximum value of $f(j,s) - \alpha_j$, the derivative is:
\begin{align}
\frac{\partial}{\partial \alpha_i}\max_j[f(j,s) - \alpha_j] = 
\begin{cases}
-1 & \text{if } i = \argmax_j[f(j,s) - \alpha_j] \\
0 & \text{otherwise}
\end{cases}
\end{align}

The negative sign appears because $\alpha_i$ has a negative coefficient in the expression inside the max.For simplicity, we often write the optimality condition using the indicator function $\mathbbm{1}[i \in \argmax_j[f(j,s) - \alpha_j]]$, which equals 1 when $i$ is in the argmax set and 0 otherwise.
\begin{align}
\frac{\partial}{\partial \alpha_i}\max_j[f(j,s) - \alpha_j] = -\mathbbm{1}[i = \argmax_j[f(j,s) - \alpha_j]]
\end{align}

If there are ties (multiple indices achieve the maximum), then the max function is not differentiable. Instead, we use the concept of subdifferential. A valid subgradient can be written as $-\mathbbm{1}[i \in \argmax_j[f(j,s) - \alpha_j]] \cdot w_i$, where $w_i \geq 0$ are weights such that $\sum_{i \in \argmax} w_i = 1$.

The derivative (or subgradient) of the entire objective function with respect to $\alpha_i$ is:
\begin{align}
\frac{\partial f_k}{\partial \alpha_i} = P_X(i) - \frac{1}{k}\sum_{s=1}^k \mathbbm{1}[i \in \argmax_j[f(j,s) - \alpha_j]]
\end{align}

Then, for the optimal dual variables $\alpha^*$, the first-order optimality condition gives:
\begin{align}
\frac{\partial f_k}{\partial \alpha_i}(\alpha^*) = P_X(i) - \frac{1}{k}\sum_{s=1}^k \mathbbm{1}[i \in \argmax_j[f(j,s) - \alpha^*_j]] = 0
\end{align}

We interpret this optimality condition as follows: the probability $P_X(i)$ equals the empirical probability that $i$ is in the argmax set of $f(j,s) - \alpha^*_j$ across all samples $s$. Rearranging, we have:
\begin{align}\label{eq:empirical_average}
\frac{1}{k}\sum_{s=1}^k \mathbbm{1}[i \in \argmax_j[f(j,s) - \alpha^*_j]] = P_X(i)
\end{align}

To fully understand the emergence of the Gumbel-Max connection, we must carefully analyze how the discrete optimality condition converges to its continuous counterpart as $k \to \infty$. The key step is understanding how the empirical average on Equation \ref{eq:empirical_average} becomes the probability statement:
\begin{align}
\Pr[\argmax_j(z_j - \alpha^*_j) = i] = P_X(i)
\end{align}
In other words, it remains to show that 
\begin{equation}\label{eq:empirical_avg_conv}
\frac{1}{k}\sum_{s=1}^k 
  \mathbbm{1}\!\bigl\{\,i \in \arg\max_{j}\!\bigl[f(j,s) - \alpha_j^*\bigr]\bigr\}
\;\longrightarrow\;
\mathbb{E}\!\Bigl[\,
  \mathbbm{1}\!\bigl\{\,i \in \arg\max_{j}\!\bigl[f(j,s) - \alpha_j^*\bigr]\bigr\}
\Bigr].
\end{equation}

In order to show this convergence, let $\alpha^*_k$ denote the optimal dual variables for the problem with $k$ samples. From the epi-convergence of $f_k$ to $f$, we know that $\alpha^*_k \to \alpha^*$ as $k \to \infty$. 

For any fixed $\alpha$, the indicator functions $\mathbbm{1}\{i \in \arg\max_j[f(j,s) - \alpha_j]\}$ are independent and identically distributed across $s$, since $f(j,s)$ are sampled independently. Therefore, by the Strong Law of Large Numbers:

\begin{equation}
\frac{1}{k}\sum_{s=1}^k \mathbbm{1}\{i \in \arg\max_j[f(j,s) - \alpha_j]\} \xrightarrow{a.s.} \mathbb{E}[\mathbbm{1}\{i \in \arg\max_j[f(j,s) - \alpha_j]\}]
\end{equation}

To address the case where $\alpha$ is replaced by $\alpha^*_k$ which depends on the samples, we decompose:

\begin{align}
&\left|\frac{1}{k}\sum_{s=1}^k \mathbbm{1}\{i \in \arg\max_j[f(j,s) - \alpha^*_{k,j}]\} - \mathbb{E}[\mathbbm{1}\{i \in \arg\max_j[f(j,s) - \alpha^*_j]\}]\right| \\
&\leq \left|\frac{1}{k}\sum_{s=1}^k \mathbbm{1}\{i \in \arg\max_j[f(j,s) - \alpha^*_{k,j}]\} - \frac{1}{k}\sum_{s=1}^k \mathbbm{1}\{i \in \arg\max_j[f(j,s) - \alpha^*_j]\}\right| \\
&+ \left|\frac{1}{k}\sum_{s=1}^k \mathbbm{1}\{i \in \arg\max_j[f(j,s) - \alpha^*_j]\} - \mathbb{E}[\mathbbm{1}\{i \in \arg\max_j[f(j,s) - \alpha^*_j]\}]\right|
\end{align}

The second term converges to zero by the Strong Law of Large Numbers. For the first term, we exploit the fact that the argmax function is stable under small perturbations except at ties, which occur with probability zero for continuous distributions of $f$. 

Specifically, let $\Delta_k = \|\alpha^*_k - \alpha^*\|_\infty$. For any realization of $f(j,s)$, the argmax changes only if the perturbation $\Delta_k$ exceeds the minimum gap between the maximum value and the second-largest value. Let $G_s = \min_{j \neq j^*} [f(j^*,s) - \alpha^*_{j^*}] - [f(j,s) - \alpha^*_j]$, where $j^* = \arg\max_j [f(j,s) - \alpha^*_j]$. Then:

\begin{equation}
\mathbbm{1}\{i \in \arg\max_j[f(j,s) - \alpha^*_{k,j}]\} \neq \mathbbm{1}\{i \in \arg\max_j[f(j,s) - \alpha^*_j]\} \implies \Delta_k > G_s
\end{equation}

Since $\alpha^*_k \to \alpha^*$, we have $\Delta_k \to 0$ as $k \to \infty$. The probability $\mathbb{P}(\Delta_k > G_s)$ converges to zero because $G_s > 0$ almost surely for continuous distributions. By dominated convergence, the first term also converges to zero.

Consequently:

\begin{equation}
\frac{1}{k}\sum_{s=1}^k \mathbbm{1}\{i \in \arg\max_j[f(j,s) - \alpha^*_{k,j}]\} \xrightarrow{p} \mathbb{E}[\mathbbm{1}\{i \in \arg\max_j[f(j,s) - \alpha^*_j]\}]
\end{equation}

Combined with our optimality condition in equation \eqref{eq:empirical_average}, we obtain:

\begin{equation}
P_X(i) = \mathbb{E}[\mathbbm{1}\{i \in \arg\max_j[f(j,s) - \alpha^*_j]\}] = \mathbb{P}(\arg\max_j [f(j,s) - \alpha^*_j] = i)
\end{equation}
In other words, the random mapping $ S \mapsto \argmax_j [f(j,s) - \alpha^*_j]$ reproduces the original law $P_X$ exactly. This fact is precisely what Theorem \ref{thm:gumbel_as_ot} asserts: the Gumbel‐Max procedure constitutes the optimal‐transport coupling between $P_X$ and the side‐information mechanism.

\textbf{Connection to Gumbel Watermarking.}  The Gumbel watermarking scheme described above can be directly interpreted within our optimal transport framework by noticing that this same arg-max coupling is exactly what underlies the Gumbel-Max trick: adding Gumbel noise $G_t(j)$ to (negative) logits $-\frac{u_t(y)}{T}$ and taking argmax samples from the softmax.  In our formulation, the cost function $f(j,s)$ corresponds to the Gumbel noise $G_t(j)$, while the dual variables $\alpha_j$ correspond to $-\frac{u_t(j)}{T}$. The optimality condition
\begin{equation}\label{gumbel-equiv}
P_X(i) = \mathbb{P}(\arg\max_j [f(j,s) - \alpha^*_j] = i)
\end{equation}
is precisely the Gumbel-Max trick, which states that sampling from a softmax distribution is equivalent to adding Gumbel noise to logits and taking the argmax.\footnote{To see this precisely, substitute the $j$th logit as $-\alpha^*_j$ on the RHS of \eqref{gumbel-equiv}, and simplify the RHS using the same steps as in Appendix B.1 of \cite{fu2024gumbelsoft}. This shows that \eqref{gumbel-equiv} holds when $-\alpha^*_j=j$th logit. Which means that the Gumbel watermark is a special case of our construction in sec.~\ref{sec:beyond_binary} (which boils down to \eqref{gumbel-equiv}), with $f(x,s)$ specifically chosen has Gumbel$(0,1)$.} 
In practice, under the watermarking process, the Gumbel scheme replaces the uniform variables $r_t(y)$ with pseudo-random values $F_{y_{t-m:t-1},k}(y)$ that depend on the secret key and previously generated tokens. This creates a coupling between the original token distribution and the side information, exactly as prescribed in our optimal transport approach. Thus, when we assume i.i.d. scores, the Gumbel watermarking scheme represents a specific instantiation of our general optimal transport framework, where the coupling is designed to preserve the original distribution in expectation while maximizing detectability through heavy-tailed score distributions.

\subsection{Proof of the Detection Gap, Theorem~\ref{thm:detection-lb-informal}.}\label{apdx:proof_thm_4}

In Section \ref{sec:beyond_binary}, we introduced the \heavywater{} scheme by generalizing the scores beyond the binary case. Since $f(x,s)$ is random, $\Dgap^{[P_F]}(m,k,\lambda)$ given by
\begin{align}\label{d_gap_new}
    \Dgap^{[P_F]}(m,k,\lambda)=\min_{P_X\in\mathcal{P}_{\lambda}}\max_{P_{XS}}\left(\mathbb{E}_{P_{XS}}\left[f(X,S)\right]-\mathbb{E}_{P_{X}P_S}\left[f(X,S)\right] \right)
\end{align} 
is also a random variable. Now, we will show that we can go beyond Theorem \ref{thm:gumbel_as_ot}, improving on the watermarking scheme like Gumbel, and prove Theorem \ref{thm:detection-lb-informal} that connects the asymptotic detection gap with the quantiles of the distribution of the score difference $\Delta = f(x,s) - f(x',s')$.

\begin{thm}[Detection Gap, asymptotic randomness]\label{thm:detection_gap_asymptotic}
Let $\lambda\in\left[\frac{1}{2},1\right)$, and consider the score difference random variable $\Delta = f(x,s) - f(x',s')$ for some $(x,s)\neq(x',s')$, where $f(x,s)$ and $f(x',s')$ are sampled i.i.d. from $P_F$. Let the cumulative distribution function of $\Delta$ be $F$, and let $Q = F^{-1}$ be its inverse. Then,
\begin{equation}
\lim_{k\to\infty}\Dgap^{[P_F]}(m,k,\lambda) =\int_{1-\lambda}^1 Q(u)du.
\end{equation}
\end{thm}

\begin{proof}

We begin by characterizing the structure of the optimal coupling $P_{XS}$ that attains the maximum in \eqref{OT_neww} below.

\begin{align}\label{OT_neww}
        &\max_{P_{XS}} \mathbb{E}^{[k]}_{P_{XS}}[f(X,S)]-\mathbb{E}^{[k]}_{P_XP_S}[f(X,S)]\nonumber\\
        \text{s.t. } &\sum_{s=1}^k P_{x,s}=p_x,\quad x\in\{1,\dotsc,m\}\nonumber\\
        & \sum_{x=1}^m P_{x,s}=\frac{1}{k},\quad s\in\{1,\dotsc,k\}
\end{align}

\begin{Lemma}
The minimum in \eqref{d_gap_new} is achieved by the $P_X\in\mathcal{P}_\lambda$ satisfying
\begin{align}
    P_X(x)=\begin{cases}
    \lambda, & x=i,\\
    1-\lambda & x=j,\\
    0, & x\neq i,j.
\end{cases}
\end{align}
as shown in \eqref{two-token}. Assume that $\lambda \ge \frac{1}{2}$. 
Then, the optimal $P_{XS}$, solution of Equation \eqref{OT_neww}, is given by
\[
    \begin{bmatrix}
         & s_1 & \cdots & s_{r} & s_{r+1} & s_{r+2}  & \cdots & s_k\\
        \lambda & \frac{1}{k} & \cdots & \frac{1}{k} & \lambda-\frac{r}{k}& 0 & \cdots & 0\\
        1-\lambda & 0 & \cdots & 0 & \frac{r+1}{k}-\lambda& \frac{1}{k} & \cdots & \frac{1}{k}
    \end{bmatrix}
\]
where $r$ is the integer satisfying $\frac{r}{k}\le \lambda < \frac{r+1}{k}$.
\end{Lemma}

The optimal coupling assigns the probability mass of $x=i$ sequentially to the first $r$ side-information bins until its total mass $\lambda$ is exhausted, after which the remaining bins are filled by $x=j$. This is intuitive as $P_X(i)\geq P_X(j)$.

For a given $k$, $\max_{P_{XS}}\mathbb{E}^{[k]}_{P_{XS}}[f(X,S)]$ is given by
\begin{align}
    &\max_{P_{XS}}\mathbb{E}^{[k]}_{P_{XS}}[f(X,S)]\\
    &=\frac{1}{k}\sum_{s=1}^rf(1,s)+\left(\lambda-\frac{r}{k}\right)f(1,r+1)+\left(\frac{r+1}{k}-\lambda\right)f(2,r+1)+\frac{1}{k}\sum_{s=r+2}^kf(2,s)\\
    &=\frac{1}{k}\sum_{s=1}^kf(2,s)+\left(\lambda-\frac{r}{k}\right)\Delta_{r+1}+\frac{1}{k}\sum_{s=1}^r\Delta_{s}
\end{align}
For $\mathbb{E}^{[k]}_{P_X P_S}[f(X,S)]$, we have
\begin{equation}
\mathbb{E}^{[k]}_{P_X P_S}[f(X,S)] = \frac{\lambda}{k}\sum_{s=1}^k f(1,s) + \frac{1-\lambda}{k}\sum_{s=1}^k f(2,s).
\end{equation}
Therefore, the difference is given by
\begin{align}
&\max_{P_{XS}}\mathbb{E}^{[k]}_{P_{XS}}[f(X,S)]-\mathbb{E}_{P_X P_S}^{[k]}[f(X,S)] \\
&= \frac{1}{k}\sum_{s=1}^kf(2,s)+\left(\lambda-\frac{r}{k}\right)\Delta_{r+1}+\frac{1}{k}\sum_{s=1}^r\Delta_{s} - \frac{\lambda}{k}\sum_{s=1}^k f(1,s) - \frac{1-\lambda}{k}\sum_{s=1}^k f(2,s) \\
&= \frac{\lambda}{k}\sum_{s=1}^kf(2,s) + \left(\lambda-\frac{r}{k}\right)\Delta_{r+1}+\frac{1}{k}\sum_{s=1}^r\Delta_{s} - \frac{\lambda}{k}\sum_{s=1}^k f(1,s)
\end{align}

Let us define the following terms that we will analyze precisely:
\begin{align}
\bar{f}_{1,k} = \frac{1}{k}\sum_{s=1}^{k}f(1,s), \quad
\bar{f}_{2,k} = \frac{1}{k}\sum_{s=1}^{k}f(2,s) \quad \text{and} \quad 
I_k = \frac{1}{k}\sum_{s=1}^r\Delta_{s}.
\end{align}

With these definitions, we can rewrite the detection gap as:
\begin{align}\label{eq:detection_gap}
\max_{P_{XS}}\mathbb{E}^{[k]}_{P_{XS}}[f(X,S)]-\mathbb{E}_{P_X P_S}^{[k]}[f(X,S)] = \lambda\bar{f}_{2,k} + \left(\lambda-\frac{r}{k}\right)\Delta_{r+1} + I_k - \lambda\bar{f}_{1,k}
\end{align}

To establish the exact limit, we need to analyze each term with greater precision.

\textbf{Exact characterization of $\lambda(\bar{f}_{2,k} - \bar{f}_{1,k})$:} Both $f(1,s)$ and $f(2,s)$ are i.i.d. sub-exponential random variables with parameters $(\nu^2, b)$. By the strong law of large numbers, we have almost surely:
\begin{align}
\lim_{k\to\infty}\bar{f}_{1,k} = \mathbb{E}[f(1,1)] \quad \text{and} \quad \lim_{k\to\infty}\bar{f}_{2,k} = \mathbb{E}[f(2,1)]
\end{align}

Since $f(x,s)$ are i.i.d. with zero mean, we have $\mathbb{E}[f(1,1)] = \mathbb{E}[f(2,1)] = 0$. Therefore, almost surely:
\begin{align}
\lim_{k\to\infty}\lambda(\bar{f}_{2,k} - \bar{f}_{1,k}) = 0
\end{align}

\textbf{Exact characterization of $\left(\lambda-\frac{r}{k}\right)\Delta_{r+1}$:} By definition of $r = \lfloor\lambda k\rfloor$, we have $0 \leq \lambda - \frac{r}{k} < \frac{1}{k}$. Since $\Delta_{r+1}$ is a sub-exponential random variable with parameters $(4\nu^2, 2b)$, it is almost surely finite. Combining this with the above inequality:
\begin{align}
\lim_{k\to\infty}\left(\lambda-\frac{r}{k}\right)\Delta_{r+1} = 0 \quad \text{almost surely}.
\end{align}

\textbf{Exact characterization of $I_k$:} For this term, we use order statistics. Let $\Delta_{(1)}\leq\Delta_{(2)}\leq\cdots\leq\Delta_{(k)}$ denote the ordered values for the difference of scores $\Delta_s$. Then:
\begin{equation}
I_k = \frac{1}{k}\sum_{j=k-r+1}^{k}\Delta_{(j)}.
\end{equation}
Let $F$ be the CDF of $\Delta_s$ and $F_k$ be the empirical CDF given by \begin{equation}
F_k(x)\;:=\;\frac1k\sum_{s=1}^{k}\mathbf 1\!\{\Delta_s\le x\},
\end{equation} By the Glivenko-Cantelli theorem, we have:
\begin{equation}
\sup_{x\in\mathbb{R}}|F_k(x) - F(x)| \leq \eta_k \quad \text{where } \eta_k \to 0 \text{ almost surely as } k \to \infty
\end{equation}

For each order statistic $\Delta_{(j)}$, the empirical CDF by definition gives $
F_k(\Delta_{(j)}) = \frac{j}{k}$. The bound from Glivenko-Cantelli gives:
\begin{equation}
\frac{j}{k} - \eta_k \leq F(\Delta_{(j)}) \leq \frac{j}{k} + \eta_k.
\end{equation}

Let $Q = F^{-1}$ be the quantile function. By definition of the quantile function, if $p \leq F(x)$ then $Q(p) \leq x$, and if $F(x) \leq q$ then $x \leq Q(q)$. Applying these relationships:
\begin{equation}
Q\left(\frac{j}{k} - \eta_k\right) \leq \Delta_{(j)} \leq Q\left(\frac{j}{k} + \eta_k\right)
\end{equation}

Now we perform a change of index. We want to rewrite the sum in $I_k$ which uses index $j$ ranging from $k-r+1$ to $k$. Let's set $j = k-i+1$, so $i$ ranges from $1$ to $r$. This gives:
\begin{equation}
\frac{j}{k} = \frac{k-i+1}{k} = 1 - \frac{i-1}{k}
\end{equation}

Substituting this into our bounds on $\Delta_{(j)}$:
\begin{equation}
Q\left(1-\frac{i-1}{k} - \eta_k\right) \leq \Delta_{(k-i+1)} \leq Q\left(1-\frac{i-1}{k} + \eta_k\right)
\end{equation}

Summing over $i$ from $1$ to $r$ and dividing by $k$:
\begin{align}
\frac{1}{k}\sum_{i=1}^{r}Q\left(1-\frac{i-1}{k} - \eta_k\right) &\leq \frac{1}{k}\sum_{i=1}^{r}\Delta_{(k-i+1)}\\
&= \frac{1}{k}\sum_{j=k-r+1}^{k}\Delta_{(j)}\\
&= I_k\\
&\leq \frac{1}{k}\sum_{i=1}^{r}Q\left(1-\frac{i-1}{k} + \eta_k\right)
\end{align}

As $k \to \infty$, we know that $\eta_k \to 0$ almost surely by the Glivenko-Cantelli theorem. The points $u_i := 1-\frac{i-1}{k}$ for $i = 1,\ldots,r$ where $r = \lfloor \lambda k \rfloor$, form a partition of the interval $[1-\lambda, 1]$ as follows:
\begin{align}
u_1 = 1, \
u_2 = 1-\frac{1}{k}, \
u_3 = 1-\frac{2}{k} ,\
\dots, \ 
u_r = 1-\frac{r-1}{k} \approx 1-\lambda.
\end{align}

As $k \to \infty$, the number of points $r = \lfloor \lambda k \rfloor$ also increases, and the distance between adjacent points $\frac{1}{k} \to 0$. Therefore, these points $\{u_i\}_{i=1}^r$ form an increasingly fine partition of the interval $[1-\lambda,1]$. For any fixed $u \in [1-\lambda,1]$, as $k \to \infty$, there exists a sequence of indices $i_k$ such that $u_{i_k} \to u$. Specifically, we can take $i_k = \lceil k(1-u) + 1 \rceil$, which ensures $u_{i_k} \to u$ as $k \to \infty$.
Our bounds for $I_k$ can be written as:
\begin{align}
\frac{1}{k}\sum_{i=1}^{r}Q(u_i-\eta_k) \leq I_k \leq \frac{1}{k}\sum_{i=1}^{r}Q(u_i+\eta_k)
\end{align}

Since $Q$ is non-decreasing, it is bounded on the compact interval $[1-\lambda-\beta, 1+\beta]$ for some $\beta > 0$. Let
\begin{equation}
M := \sup_{t \in [1-\lambda-\beta, 1+\beta]} |Q(t)| < \infty,
\end{equation}
then $|Q(u_i-\eta_k)| \leq M$ and $|Q(u_i+\eta_k)| \leq M$ for all $i$ and sufficiently large $k$.

The lower sum $\frac{1}{k}\sum_{i=1}^{r}Q(u_i-\eta_k)$ is a perturbed lower Riemann sum for the integral $\int_{1-\lambda}^{1}Q(u)\,du$, and similarly the upper sum is a perturbed upper Riemann sum. As $k \to \infty$, two things happen simultaneously, namely, (i) the mesh width $\frac{1}{k} \to 0$, so the Riemann sums converge to the integral and (ii) the perturbation $\eta_k \to 0$, so $Q(u_i \pm \eta_k) \to Q(u_i)$. By the Dominated Convergence Theorem, we have
\begin{align}
\lim_{k\to\infty} \frac{1}{k}\sum_{i=1}^{r}Q(u_i-\eta_k) &= \int_{1-\lambda}^{1}Q(u)\,du \\
\lim_{k\to\infty} \frac{1}{k}\sum_{i=1}^{r}Q(u_i+\eta_k) &= \int_{1-\lambda}^{1}Q(u)\,du
\end{align}

Since $I_k$ is bounded between these two quantities that converge to the same limit, we have
\begin{equation}
\lim_{k\to\infty} I_k = \int_{1-\lambda}^{1}Q(u)du \quad \text{almost surely}.
\end{equation}

\textbf{Combining all terms:} From our asymptotic analysis of each term, we have:
\begin{align}
\lim_{k\to\infty}\max_{P_{XS}}\mathbb{E}^{[k]}_{P_{XS}}[f(X,S)]-\mathbb{E}_{P_X P_S}^{[k]}[f(X,S)] = \lim_{k\to\infty} I_k = \int_{1-\lambda}^{1}Q(u)du \quad \text{almost surely}.
\end{align}

Therefore:
\begin{align}
\lim_{k\to\infty}D_{\text{gap}}^{[P_F]}(m,k,\lambda) = \int_{1-\lambda}^{1}Q(u)du.
\end{align}

\end{proof}

Theorem \ref{thm:detection_gap_asymptotic} implies that distributions with heavier tails imply larger values of the integral $\int_{1-\lambda}^{1}Q(u)du$, which in turn imply higher detection gaps. Indeed, fix $\lambda$. We say that a distribution $F_2$ is \emph{(right-)heavier-tailed} than $F_1$ when
\(
1-F_2(x)\le 1-F_1(x)\text{ for all large }x,
\)
equivalently $Q_{2}(u)\ge Q_{1}(u)$ for every $u$ in a neighbourhood of
$1$.  In particular, for all $u \in [1-\lambda,1]$:
\[
Q_{2}(u) \;\ge\; Q_{1}(u)
\]
which implies:
\[
\int_{1-\lambda}^{1}Q_{2}(u)du \;\ge\; \int_{1-\lambda}^{1}Q_{1}(u)du
\]

Therefore, keeping the same confidence levels:
\[
\text{heavier tail }\Longrightarrow
\text{larger }\int_{1-\lambda}^{1}Q(u)du
\Longrightarrow
\text{larger guaranteed detection gap}.
\]

The asymptotic detection gap, given by $\int_{1-\lambda}^{1}Q(u)du$, represents the average value of the quantile function over the upper $\lambda$ fraction of the distribution. This integral captures how much signal can be extracted from the tail of the score differences. Distributions whose upper tail places more mass far from zero (resulting in larger values of the integral $\int_{1-\lambda}^{1}Q(u)du$ directly increase the detection capability of the watermark. This explains why the choice of score distribution fundamentally impacts watermark detectability. Consequently, among distributions with the same mean and variance, the
heavier-tailed ones yield strictly higher guaranteed detection rates. Therefore, if we constrain ourselves to the ensemble of sub-exponential probability distributions, by choosing something with a heavier tail than Gumbel, e.g., Log-Normal, we can achieve a better detection scheme, as we show in our experiments, cf. Figure \ref{fig:fig1}.

\begin{remark}
By using Bernstein's inequality for sub-exponential variables and Dvoretzky–Kiefer–Wolfowitz inequality instead of Glivenko-Cantelli, it is possible to show a non-asymptotic version of the theorem above:

\begin{thm}[Detection gap, formal]\label{thm:detection-lb}
Let \(0<\lambda<1\) be fixed, write \(r=\lfloor\lambda k\rfloor\) and define  
\[
I_k=\frac1k\sum_{s=1}^{r}\Delta_s,\qquad  
\bar f_{1,k}=\frac{1}{k}\sum_{s=1}^{k}f(1,s),\quad  
\bar f_{2,k}=\frac{1}{k}\sum_{s=1}^{k}f(2,s),
\]
where $\Delta_s=f(1,s)-f(2,s)$. For any confidence levels \(\delta,\delta',\delta^{\ddagger}\in(0,1)\) set
\[
\varepsilon_k(\delta)=\sqrt{\frac{2\nu^{2}\log(1/\delta)}{k}}
                      +\frac{b\log(1/\delta)}{k},\qquad
t_{\!*}(\delta')=\max\!\Bigl\{2\nu\sqrt{\log\tfrac1{\delta'}},
                             \;2b\log\tfrac1{\delta'}\Bigr\},
\]
\[
\eta_k(\delta^{\ddagger})=\sqrt{\frac{\log(2/\delta^{\ddagger})}{2k}}.
\]
Then, with probability at least \(1-\delta-\delta'-\delta^{\ddagger}\),
\begin{equation}\label{eq:detection-lb}
\max_{P_{XS}}\;
\mathbb{E}^{[k]}_{P_{XS}}\!\bigl[f(X,S)\bigr]
-\mathbb{E}^{[k]}_{P_XP_S}\!\bigl[f(X,S)\bigr]
\;\ge
\frac{1}{k}\sum_{i=1}^{r}
Q\!\Bigl(1-\frac{i-1}{k}-\eta_k(\delta^{\ddagger})\Bigr).
-\Bigl[2\varepsilon_k(\delta)+\tfrac{t_{\!*}(\delta')}{k}\Bigr],
\end{equation}
where $Q=F^{-1}$ is the quantile function of $\Delta_s$, and $\mathbb{E}^{[k]}$ denotes an expectation where the side information alphabet is of size $k$.
\end{thm}

\end{remark}

\section{Additional Information and Implementation Details}\label{appendix:our_method}
\def\theequation{C.\arabic{equation}}
\def\thetable{C.\arabic{table}}
\def\thefigure{C.\arabic{figure}}
\def\thelem{C.\arabic{lem}}
\def\thedefn{C.\arabic{defn}}
\def\theprop{C.\arabic{prop}}

\subsection{Low-Entropy Distributions in LLMs}\label{apdx:low_ent}
To motivate the low-entropy regime, we compute summary statistics of token distributions of popular open-weight LLMs. We consider the densities of three statistics: infinity norm (connects directly to min-entropy\footnote{An infinity norm of $\lambda$, i.e. $\max_s P(x) < \lambda$, translates directly to min-entropy constraint of $-\log \lambda$.}), entropy, and L-2 norm, all calculated on the next token prediction along a collection of responses.

We observe that 90\% LLM token distributions fall into the low-entropy regime we consider with infinity norm greater than $1/2$, i.e. $\max_x P(x)\geq \frac{1}{2}$, across the three open LLMs (Llama2-7B\cite{touvron2023llama}, Llama3-8B\cite{touvron2024llama3}, Mistral-7B) and two on popular prompt-generation datasets (Q\&A tasks from Finance-QA\cite{maia201818} and coding tasks from LCC\cite{chen2021evaluating}). We show the histogram and CDF plots in Fig. \ref{fig:histogram_stats_qa} and \ref{fig:histogram_stats_coding}.

\subsection{Theoretical effect of different tails of distributions}\label{apdx:tails}
\begin{figure}[!ht]
    \centering
    \includegraphics[width=0.65\linewidth]{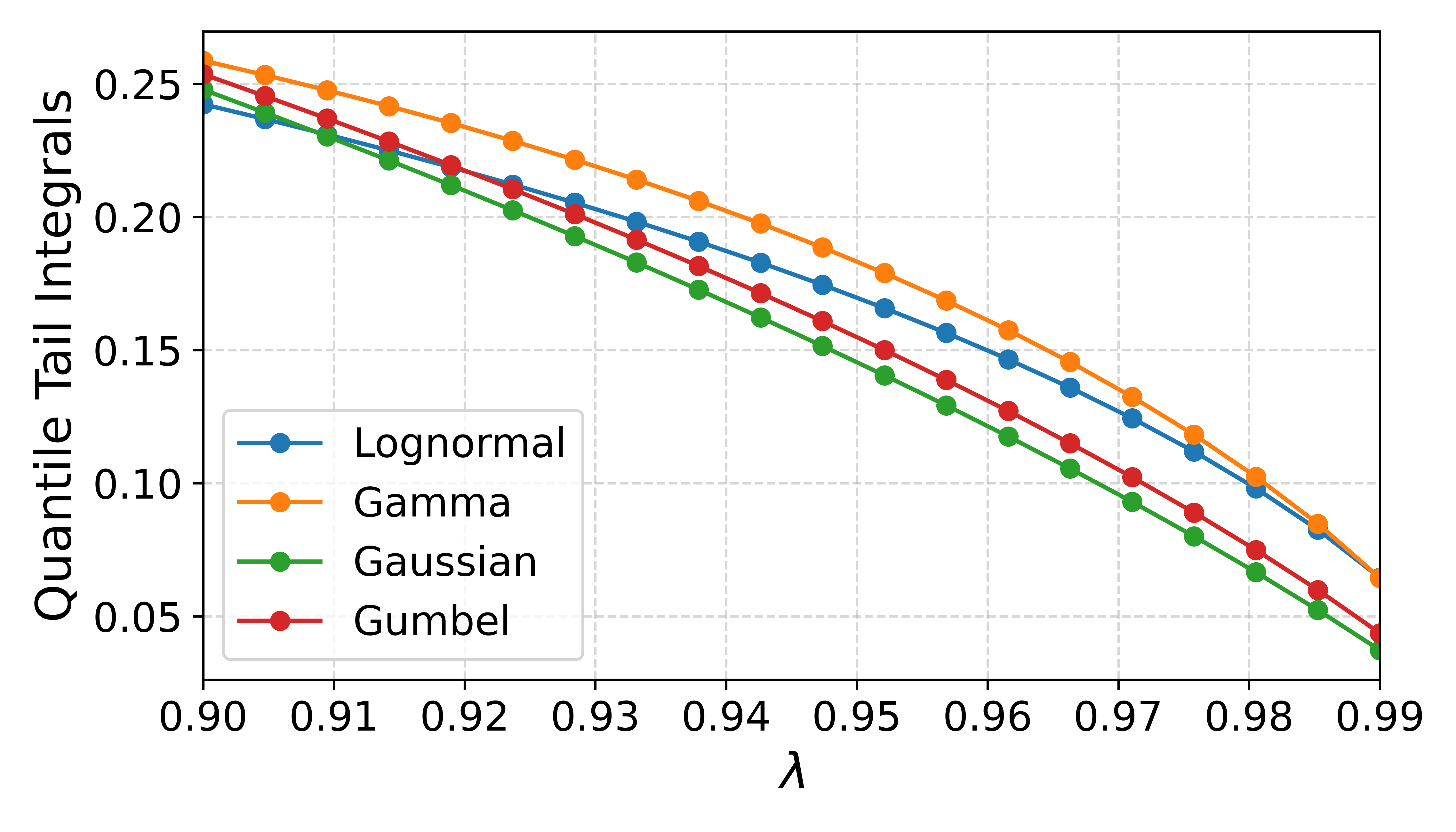}
    \caption{Tail integrals of different \emph{score difference distributions}: Higher the tail integral, better the detection.}
    \label{fig:quantile_tail_plot}
\end{figure}

To further improve the detection performance, we go beyond binary score functions to explore the flexibility in the design space that continuous score distributions offer. Motivated by the Gumbel watermark \cite{aaronson2023watermark}, we observe that we can significantly improve detection by using continuous score distributions, particularly those with heavy tails. 
Recall that our minimax formulation considers the low-entropy regime ($\lambda \in [1/2, 1]$), where the worst-case token distribution $P_X$ has only two non-zero elements with values $\{\lambda, 1-\lambda\}$. Working with this distribution, consider score matrices where each entry $f(x,s)$ is sampled independently from a distribution $P_F$ with zero mean and unit variance. We additionally assume that $f$ (and hence every $\Delta_s=f(1,s)-f(2,s)$) is \emph{sub-exponential} with parameters $(\nu^{2},b)$, i.e., $\mathbb{E}[e^{\lambda f}]\le \exp\!\bigl(\tfrac{\nu^{2}\lambda^{2}}2\bigr)$ for all $|\lambda|\le 1/b$. Many distributions, such as Gamma, Gaussian, and Lognormal satisfy this property. 
This formulation leads to Theorem \ref{thm:detection-lb-informal}, which formally characterizes the achievable maximum detection gap for any $P_F$ for large $k$, in terms of quantile tail integrals of various candidate distributions as its score distribution $P_F$. We visualize the result in Fig. \ref{fig:quantile_tail_plot}.

In Fig. \ref{fig:quantile_tail_plot}, we present the quantile tail integrals of four different distributions: Lognormal, Gamma, Gaussian, and Gumbel. 
A higher value on the y-axis (quantile tail integral) indicates a greater detection gap under the low-entropy regime, which translates to a greater probability of detection under adversarial token distributions.
In theory and as seen from Figure \ref{fig:quantile_tail_plot}, drawing score functions i.i.d from either the Lognormal or Gamma distribution outperforms that from Gumbel. Recall that the significance of adopting the Gumbel score function is that we have established its equivalence with the Gumbel watermark by \cite{aaronson2023watermark} in Theorem \ref{thm:gumbel_as_ot}. Although the Gamma distribution maximizes the detection gap in the worst-case regime, we observe that choosing $P_F$ to be lognormal achieves the highest detection accuracy in practice, which is what we eventually adopt for \heavywater{}.

\subsection{Q-ary Code}\label{apdx:qary}
We provide a discussion of the direct extension of \simplexwater{} to go beyond binary-valued scores. Recall that \simplexwater{} uses the binary Simplex Code as its score function. For a given prime field-size $q$, a Q-ary \simplexwater{} adopts the corresponding Q-ary Simplex Code \cite{roth2006introduction}, which we define next. Besides the score function, the algorithm for Q-ary \simplexwater{} is identical to the binary case, which we have provided in the main text.

Given an alphabet of size $m$ and field size $q>2$, the size of a Q-ary codeword is $q$ to the power of the ceiling of log $m$ with base $q$, i.e. $n = q^{\lceil \log_q m \rceil}$. For any $x,s \in [0:n-1]$, let \(\mathsf{qary}(x)\), \(\mathsf{qary}(s)\) denote their Q-ary representations respectively using \(n\) bits. 
A Q-ary simplex code $\fsim:[0:n-1]\times[1:n-1]\to[0:q-1$ is characterized by
\begin{align}
    \fsim(x,s) \triangleq \mathrm{dot}(\mathsf{qary}(x), \mathsf{qary}(s)),
\end{align}
where $\mathrm{dot}(\mathsf{qary}(x), \mathsf{qary}(s)) \triangleq \sum_{i=1}^{n} \mathsf{qary}(x)_i \cdot \mathsf{qary}(s)_i$ and \(\mathsf{qary}(v)_i\) denotes the \(i\)th bit in the Q-ary representation of \(v\).

There are two main limitations of Q-ary \simplexwater{}, which motivated us to explore a continuous score function directly and ultimately led to \heavywater{}. The first limitation is that we do not have optimality guarantees for the Q-ary code. Recall that to maximize watermark detection, the optimal code maximizes the $L_1$ distance. Simplex Code achieves the Plotkin bound, which maximizes the pair-wise Hamming distance between codewords. In the binary case, maximizing the Hamming distance and $L_1$ distance are equivalent, where $\mathbf{1}_{i,j} = |i-j|$; in the Q-ary case, this equivalence doesn't hold. Hence, we do not have an optimality guarantee in detection for Q-ary \simplexwater{}.
The second limitation is that the size of  Q-ary codewords is potentially very large, leading to memory issues in actual implementation on a GPU. 
Recall that in the binary case, we have \( n = m - 1 \). In the Q-ary case, however, the use of the ceiling function when converting \( m \) to base-\( q \) artificially inflates \( n \), often far beyond the actual vocabulary size.  
For example, with \( q = 7 \) and a vocabulary size of \( m = 100{,}000 \), we compute
\(\lceil \log_7(100{,}000) \rceil \approx \lceil 5.92 \rceil = 6\),
which corresponds to \(n = 7^6 = 117{,}649 \) codewords—more than 18\% larger than the original vocabulary.
Furthermore, the required alphabet size to implement a Q-arry code grows with $q$.

\subsection{Implementation Details}\label{apdx:more_imp}

In this section we provide additional implementation details for \simplexwater{} and \heavywater{}. Our code implementation employs the code from the two benchmark papers, namely WaterBench\footnote{\texttt{https://github.com/THU-KEG/WaterBench}} \cite{tu2023waterbench} and MarkMyWords\footnote{\texttt{https://github.com/wagner-group/MarkMyWords}} \cite{piet2023mark}.

\paragraph{Score Matrix Instantiation} We sample the score matrix once during the initialization stage of \heavywater{} generation and it has shape $(m, k)$. Each row maps a vocabulary token to a cost vector over the side information space $\cS$. This matrix is stored as a \texttt{torch.Tensor} and reused across watermarking calls. During generation, the score matrix is used as the cost matrix for Sinkhorn-based optimal transport computations.

\paragraph{Normalization of $f$} As described before, for each element in the vocabulary, $k$ samples for the score $f$ are drawn from a heavy-tailed distribution (log-normal in the experiment). To ensure numerical stability and suitability for optimal transport computations, each row of the score matrix (size $\text{vocab\_size}*k$) is normalized to have zero mean and unit variance. 

\paragraph{Top-$p$ Filtering} To reduce the computational cost of watermarking, top-$p$ filtering is applied prior to watermarking. Identical to the common definition of top-$p$, we take the minimal set of tokens whose cumulative softmax probability exceeds a threshold (e.g., $p = 0.999$). The watermarking algorithm is then restricted to this filtered subset. 
We emphasize that, in the considered experiments, top-$p$ filtering was applied to all considered watermarks to maintain consistency in the experimental setting.

\paragraph{Detection Algorithm} We outline in Algorithm \ref{fig:detection_algo} a standard watermark detection algorithm that employs a threshold-test with a score matrix $\rF$. Given the sampled token $x$ and side information $s$, the corresponding score is $\rF_{x,s}=f(x,s)$, which is obtained in a similar fashion to its construction in generation: If \simplexwater{} is used, then it is obtained from the Simplex code, and if \heavywater{} it is randomly sampled using the shared secret key as the initial seed. This maintains the generation of the same cost matrix $\rF$ on both ends of generation and detection.
A watermark is detected if the sum of scores exceeds a certain predetermined threshold.

\begin{algorithm}[H]
\caption{\textbf{Detection} using a threshold-test with a score matrix}
\begin{algorithmic}[1]
  \State \textbf{Input:} Token sequence $x^n$, side information $s$, $\mathsf{seed}$, score matrix $\rF\in \mathbb{R}^{m \times k}$
  \State \textbf{Outputs:} $p$-value based detection outcome
  \For{$t = 1$ to $T$}
    \State Compute score $\phi_t := \rF_{x_t,s}$
  \EndFor
  \State $Z \gets \sum_{t=1}^T \phi_t$
  \State \textbf{if} $Z > \tau$ \textbf{then return} ``Watermark Detected''
  \State \textbf{else return} ``No Watermark''
\end{algorithmic}
\label{fig:detection_algo}
\end{algorithm}

\paragraph{Fresh Randomness Generation} Fresh randomness is crucial to ensure side information is sampled independently from previous tokens to avoid seed collision. Our implementation supports several seeding strategies. In majority of our experiment, we use the 'fresh' strategy, which generates a unique seed for each token by incrementing a counter and combining it with the shared secret key. This allows both ends to share the same seed that dynamically changes, but is independent of previously generated tokens. Our implementation of \heavywater{} and \simplexwater{} also allows for various forms of temporal or token-based encoding strategies, such as sliding-window hashing.

\subsection{Information on Optimal Transport and Sinkhorn's Algorithm}\label{apdx:ot}
In this section, we provide preliminary information on the considered OT problem and its solution through Sinkhorn's algorithm.
Let $p\in\Delta_m$ and $q\in\Delta_k$ be two probability vectors.
For a given cost matrix $\rC\in\mathbb{R}^{m\times k}$, the OT between $p$ and $q$ is given by
\newcommand{\sC}{\mathsf{C}}
$$
\mathsf{OT}(p,q)=\min_{P\in\Pi_{p,q}}\sum_{i=1}^m\sum_{j=1}^k \sC_{i,j}P_{i,j},
$$
where $\Pi_{p,q}$ is the set of joint distributions with marginals $p$ and $q$, which we call \textit{couplings}.
Efficiently solving the optimization $\mathsf{OT}(p,q)$ is generally considered challenging \cite{peyre2019computational}. To that end, a common approach to obtain a solution efficiently is through entropic regularization. An entropic OT (EOT) with parameter $\epsilon$ is given by
$$
\mathsf{OT}_\epsilon(p,q)=\min_{P\in\Pi_{p,q}}\left(\sum_{i=1}^m\sum_{j=1}^k \sC_{i,j}P_{i,j}-\epsilon H(P)\right),
$$
where $H(P)\triangleq-\sum_{i,j}P_{i,j}\log(P_{i,j})$ and $\rC$ is the OT cost.
The EOT is an $\epsilon$-strongly convex problem, which implies its fast convergence to the \textit{unique} optimal solution.
However, the EOT provides an approximate solution which converges to the unregularized solution with rate $O(\epsilon\log(1/\epsilon))$.

One of the main reasons EOT has gained its popularity is due to Sinkhorn's algorithm \cite{sinkhorn1967diagonal}, which is a matrix scaling algorithm that has found its application to solve the dual formulation of the EOT problem \cite{cuturi2013sinkhorn}.
Sinkhorn's algorithm looks for a pair of vectors $u,v$ that obtain the equality 
$$
P^*=\mathsf{diag}(u)K\mathsf{diag}(v),
$$
where $P^*$ is the EOT solution and $K=\exp(-\rC/\epsilon)$ is called the \textit{Gibbs Kernel}.
Consequently, Sinkhorn's algorithm follows from a simple iterative procedure of alternately updating $u$ and $v$. The steps of Sinkhorn's algorithm are given in Algorithm \ref{alg:sinkhorn}.
Having solved Sinkhorn's algorithm, we obtain the optimal EOT coupling.
When using Sinkhorn's algorithm, the stopping criteria is often regarding the marginalization of the current coupling against the corresponding marginal, i.e., we check wether $\bigl\|\,u\odot(Kv)\;-\;p\bigr\|_1\leq\delta$ where  $\triangleq \sum_{i=1}^n \bigl|u_i\,(Kv)_i \;-\;p_i\bigr|$ and $\delta>0$ is some threshold. 
In our implementation, we solve Sinkhorn's algorithm using the Python optimal transport package \cite{flamary2021pot}.

\begin{algorithm}[ht]
\caption{Sinkhorn’s Algorithm for Entropic OT}
\label{alg:sinkhorn}
\begin{algorithmic}[1]
  \State \textbf{Input:}
    Marginals $P_X\in\Delta_m$, $P_S\in\Delta_k$, 
    cost matrix $C\in\mathbb{R}^{m\times k}$   
    regularization parameter $\varepsilon>0$,  
    Threshold $\delta>0$.
  \State \textbf{Output:} Optimal coupling
    $P\in\mathbb{R}_{\ge0}^{n\times m}$ 
  \State Calculate kernel $K \gets \exp\bigl(-C / \varepsilon\bigr)$
  \State $u \gets \mathbf{1}_m,\quad v \gets \mathbf{1}_k$
  \While{$\bigl\|u\odot(K\,v) - P_X\bigr\|_1 > \delta$}
    \State $u \gets a /(K\,v)$    \Comment{element-wise division}
    \State $v \gets b / (K^{\intercal}u)$
  \EndWhile
  \State $P \gets \mathrm{diag}(u)\;K\;\mathrm{diag}(v)$
  \State \Return $P$
\end{algorithmic}
\end{algorithm}

\section{Additional Numerical Results and Ablation Study}\label{appendix:results}
\def\theequation{D.\arabic{equation}}
\def\thetable{D.\arabic{table}}
\def\thefigure{D.\arabic{figure}}
\def\thelem{D.\arabic{lem}}
\def\thedefn{D.\arabic{defn}}
\def\theprop{D.\arabic{prop}}

\subsection{Ablation Study}\label{apdx:ablation}
We perform an ablation study to investigate the effect of various hyperparameters set in \simplexwater{} and \heavywater{}.
The ablation study is performed in a curated subset of prompts from the Finance-QA dataset \cite{maia201818} considered in Section \ref{sec:numerics}.

\begin{figure*}[!ht]
  \centering
  % Subfigure 1
  \begin{subfigure}[t]{0.32\linewidth}
    \centering
    \includegraphics[width=\linewidth]{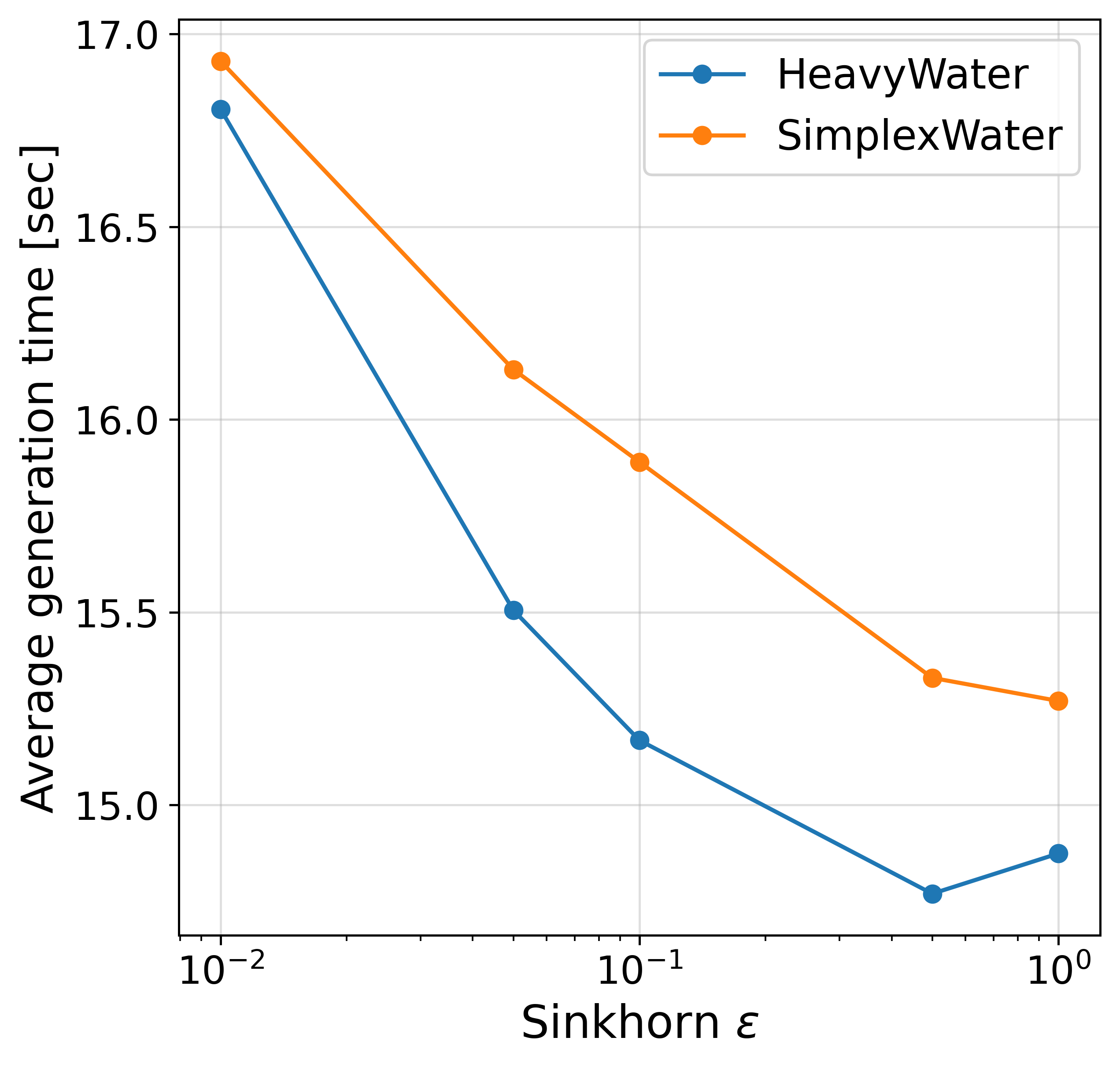}
    \caption{Generation time vs. Sinkhorn regularization parameter $\epsilon$.}
    \label{fig:sink_eps_time}
  \end{subfigure}%
  \hfill
  % Subfigure 2
  \begin{subfigure}[t]{0.295\linewidth}
    \centering
    \includegraphics[trim={30pt 5pt 5pt 5pt}, clip,width=\linewidth]{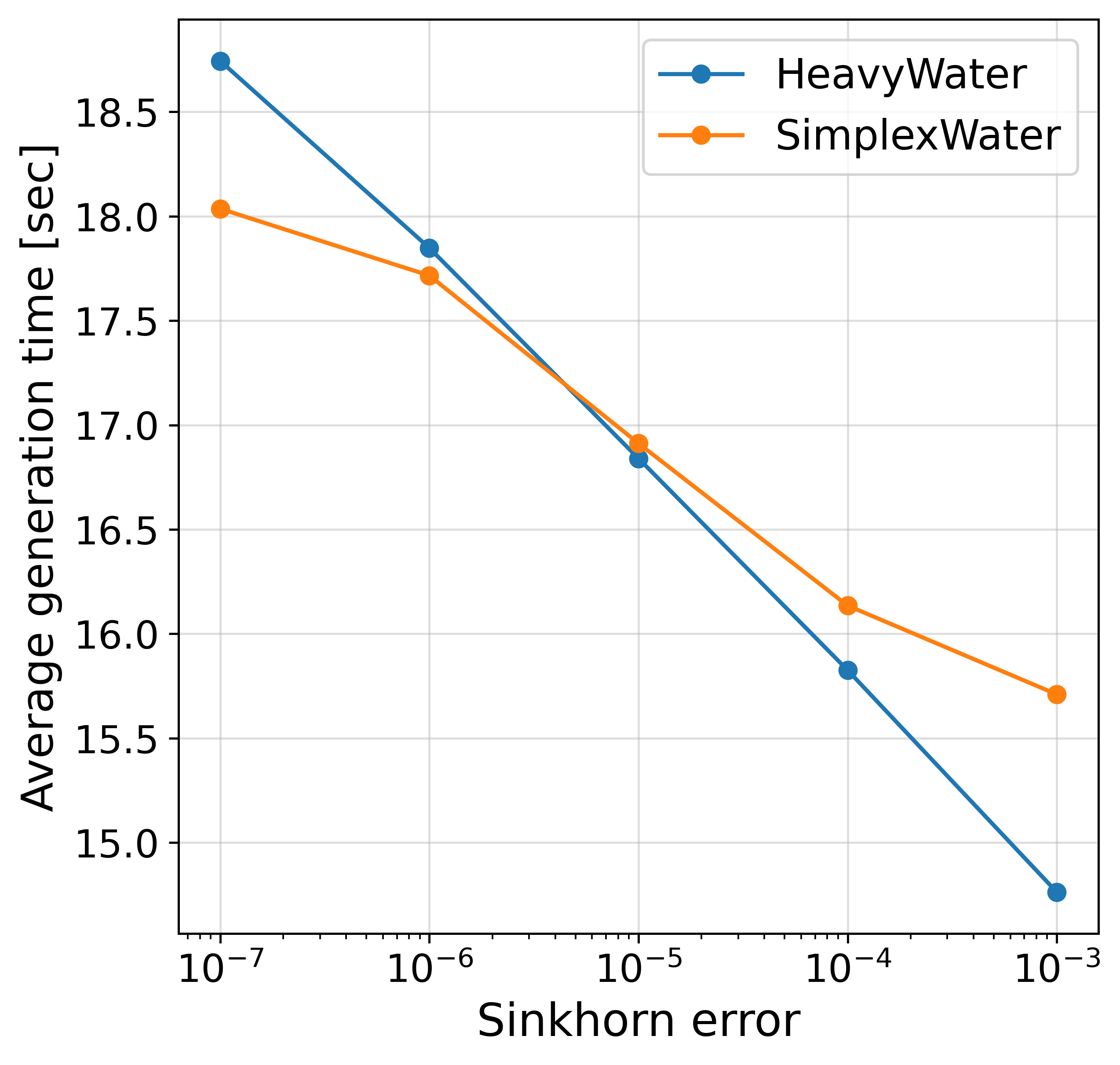}
    \caption{Generation time vs. Sinkhorn error threshold.}
    \label{fig:sink_time_err}
  \end{subfigure}
  \hfill
  \begin{subfigure}[t]{0.325\linewidth}
    \centering
    \includegraphics[width=\linewidth]{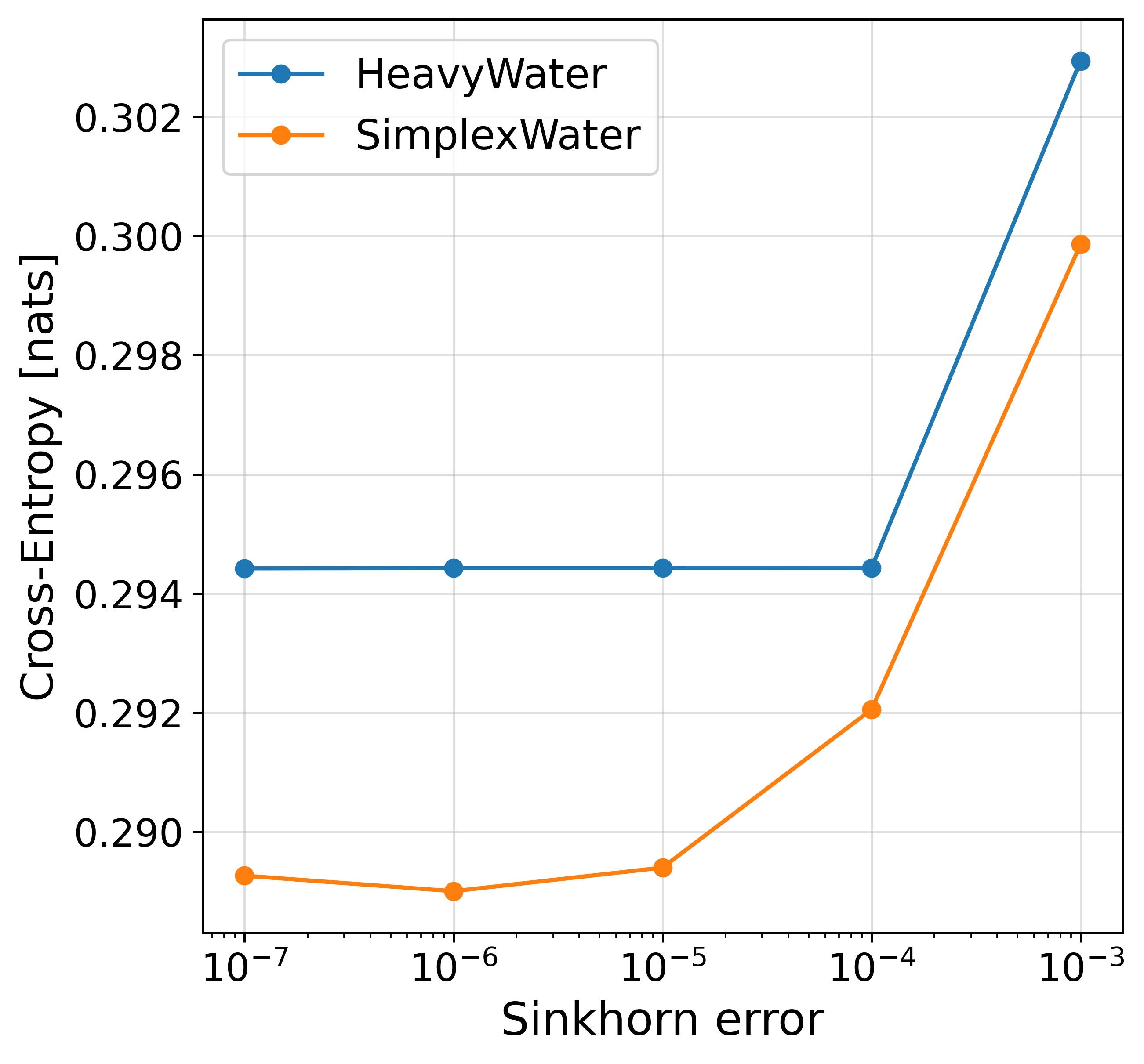}
    \caption{Cross Entropy of watermarked text  vs. Sinkhorn error threshold.}
    \label{fig:sink_ce_err}
  \end{subfigure}
  \caption{Effect of Sinkhorn Algorithm's parameters on Runtime  and distortion.}
  % \hfill
  \label{fig:figs_sinkhorn_ablation}
\end{figure*}

\paragraph{Sinkhorn Algorithm Parameters.}
We study the effect of Sinkhorn's algorithm's parameter on the performance of the proposed watermarking scheme.
We note that, while the Sinkhorn's algorithm is set with a predetermined maximum iterations parameters, in the considered experiments the algorithm runs until convergence.
We study this effect through three cases. 
\begin{enumerate}
    \item We analyze the effect of Sinkhorn's algorithm's regularization parameter $\epsilon$ on the overall runtime. While lower values of $\epsilon$ provide solution that are closer to the underlying OT solution, often a smaller value of $\epsilon$ required more time for convergence of the algorithm. As seen from Figure \ref{fig:sink_eps_time}, as expected, smaller $\epsilon$ increase overall runtime. In our experiments we chose $\epsilon=0.05$ which resulted in overall satisfactory performance, while incurring mild runtime overhead.
    \item We analyze the effect of the error threshold on runtime.
    The lower the error threshold, the higher the higher the accuracy in the solution and the higher the overall algorithm runtime. This is indeed the case, and the effect on the watermarking procedure runtime is visualized in Figure \ref{fig:sink_time_err}.
    \item We analyzed the effect of the error threshold on the watermarked distribution cross entropy (see Section \ref{sec:numerics} for definition). As seen from Figure \ref{fig:sink_ce_err}, while a lower threshold results with a higher runtime, the improvement on the cross entropy, which is a proxy for textual quality, becomes negligible from some point.
    In our experiments, we chose a threshold value of $10^{-5}$, which demonstrated the best performance between runtime, cross entropy and detection.
\end{enumerate}

\subsection{Impact of Non i.i.d. Side Information Generation}\label{apdx:seeding}
As we previously mentioned, most of the considered experiments in Section \ref{sec:numerics} operate under a 'fresh randomness' scheme, in which we try to replicate independence between the random side information and the LLM net token distribution.
However, in practice various hashing scheme are employed, often with the purpose of increasing the overall watermarking scheme's robustness to attacks.
Such hashing schemes aggregate previous tokens (using some sliding window with context size $h$) and a shared secret key $r\in\NN$.
We are interested in verifying that indeed, robustness-driven seed generation scheme do not degrade the performance of our methods.

\begin{figure*}[!ht]
  \centering
  % Subfigure 1
  \begin{subfigure}[t]{0.32\linewidth}
    \centering
    \includegraphics[width=\linewidth]{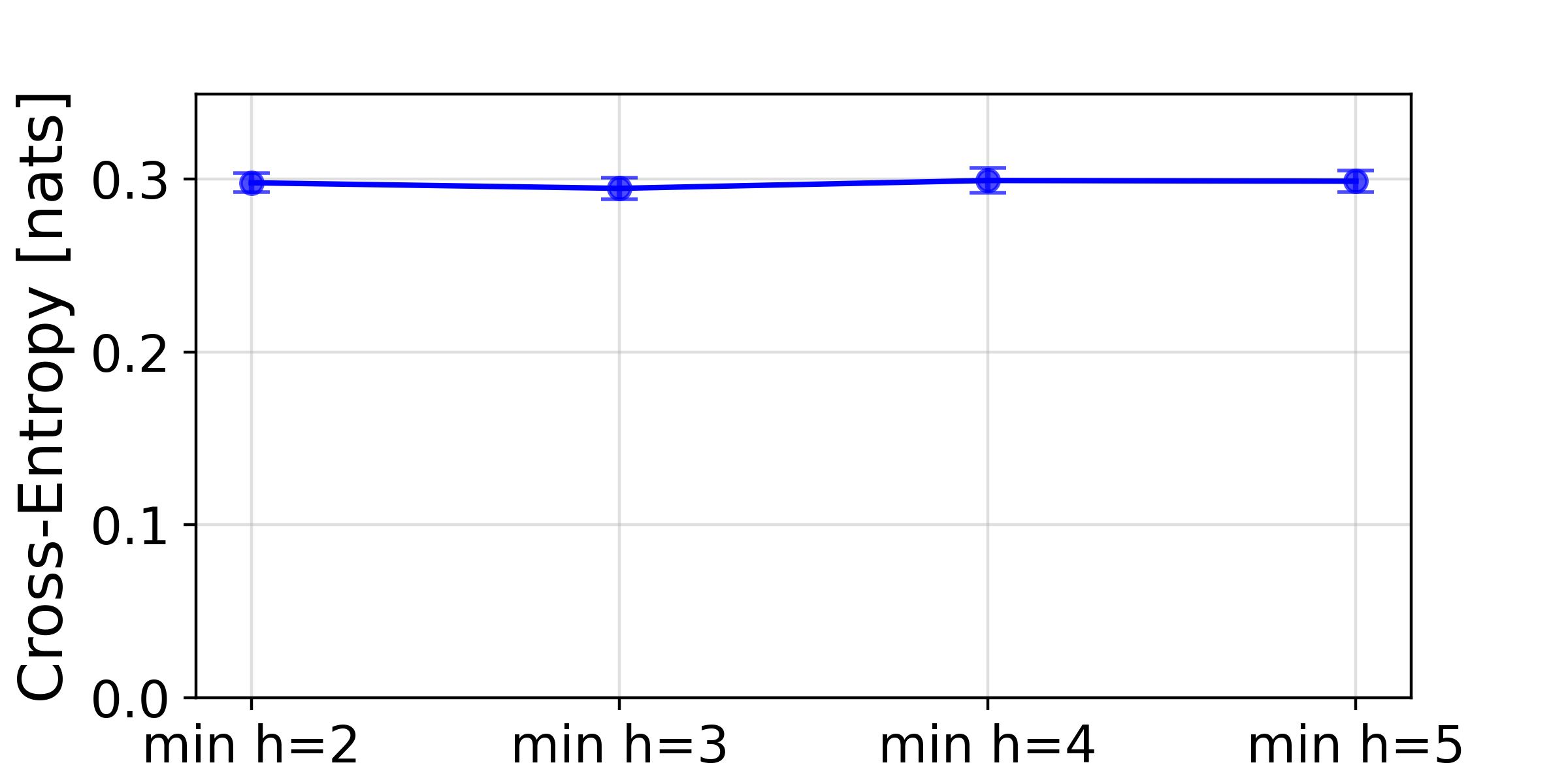}
    \caption{Cross entropy vs. context window size, $\mathsf{min}$-hash.}
    \label{fig:sink_eps}
  \end{subfigure}%
  \hfill
  % Subfigure 2
  \begin{subfigure}[t]{0.32\linewidth}
    \centering
    \includegraphics[width=\linewidth]{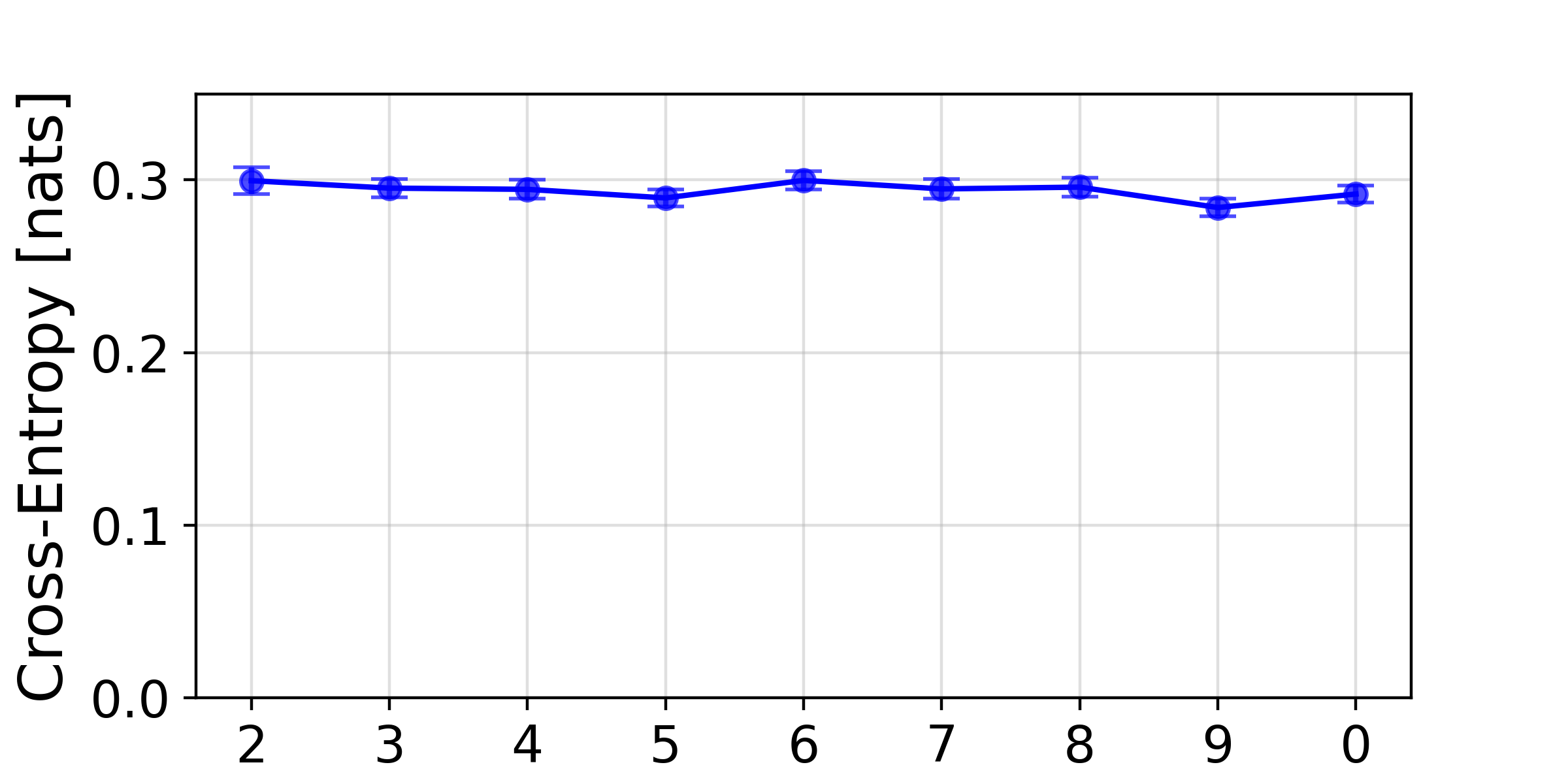}
    \caption{Cross entropy vs. context window size, $\mathsf{sum}$-hash.}
    \label{fig:ce_vs_context_len}
  \end{subfigure}
  \hfill
  \begin{subfigure}[t]{0.32\linewidth}
    \centering
    \includegraphics[width=\linewidth]{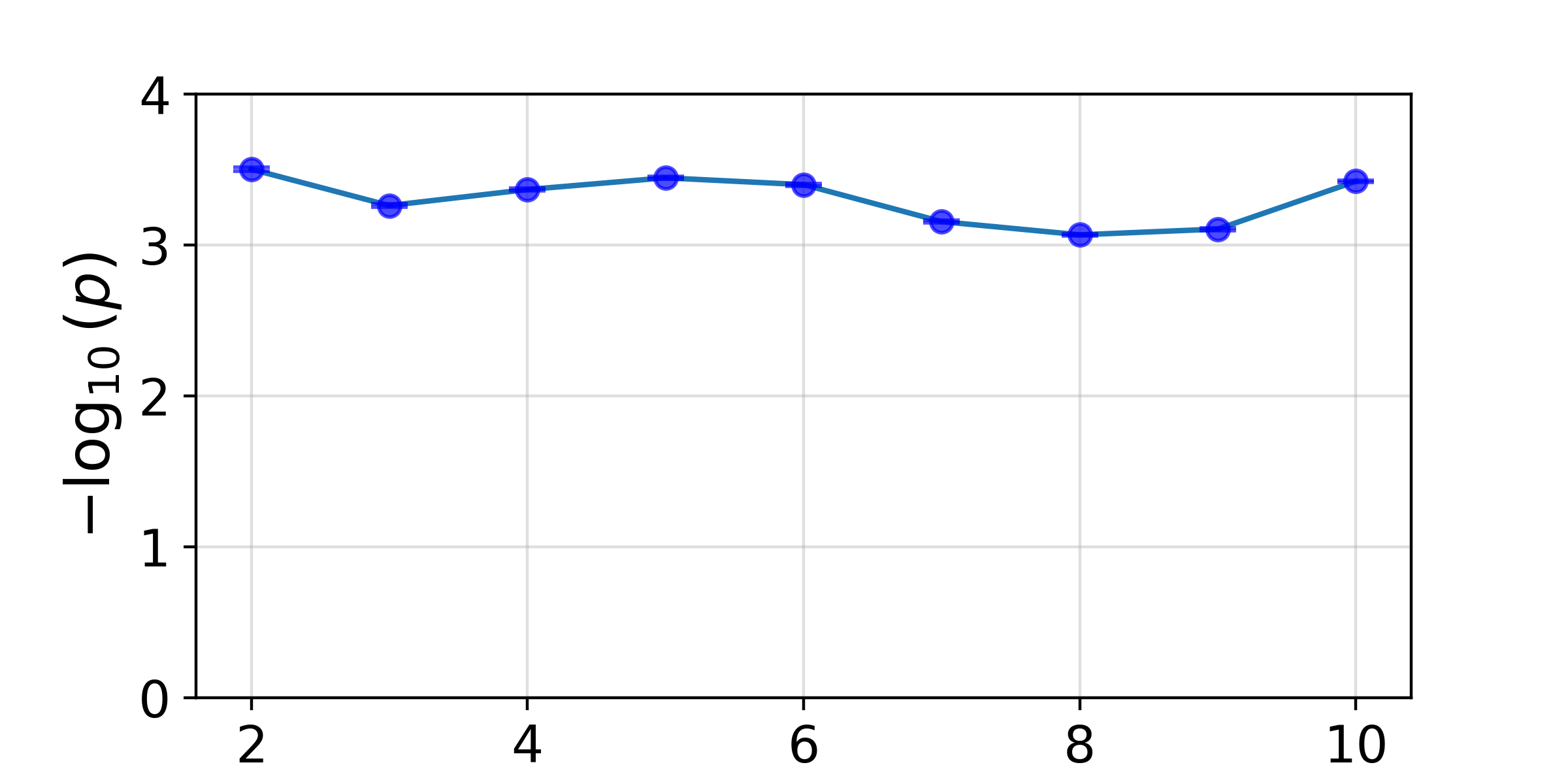}
    \caption{$-\log_{10}(p)$ vs. context window size, $\mathsf{min}$-hash.}
    \label{fig:p_vs_h}
  \end{subfigure}
  \caption{Seed Ablation: The effect of the context window is is negligible on the performance of \simplexwater{}.}
  % \hfill
  \label{fig:figs_seed_ablation}
\end{figure*}

\begin{figure*}[!ht]
  \centering
  % Subfigure 1
  \begin{subfigure}[t]{0.45\linewidth}
    \centering
    \includegraphics[width=\linewidth]{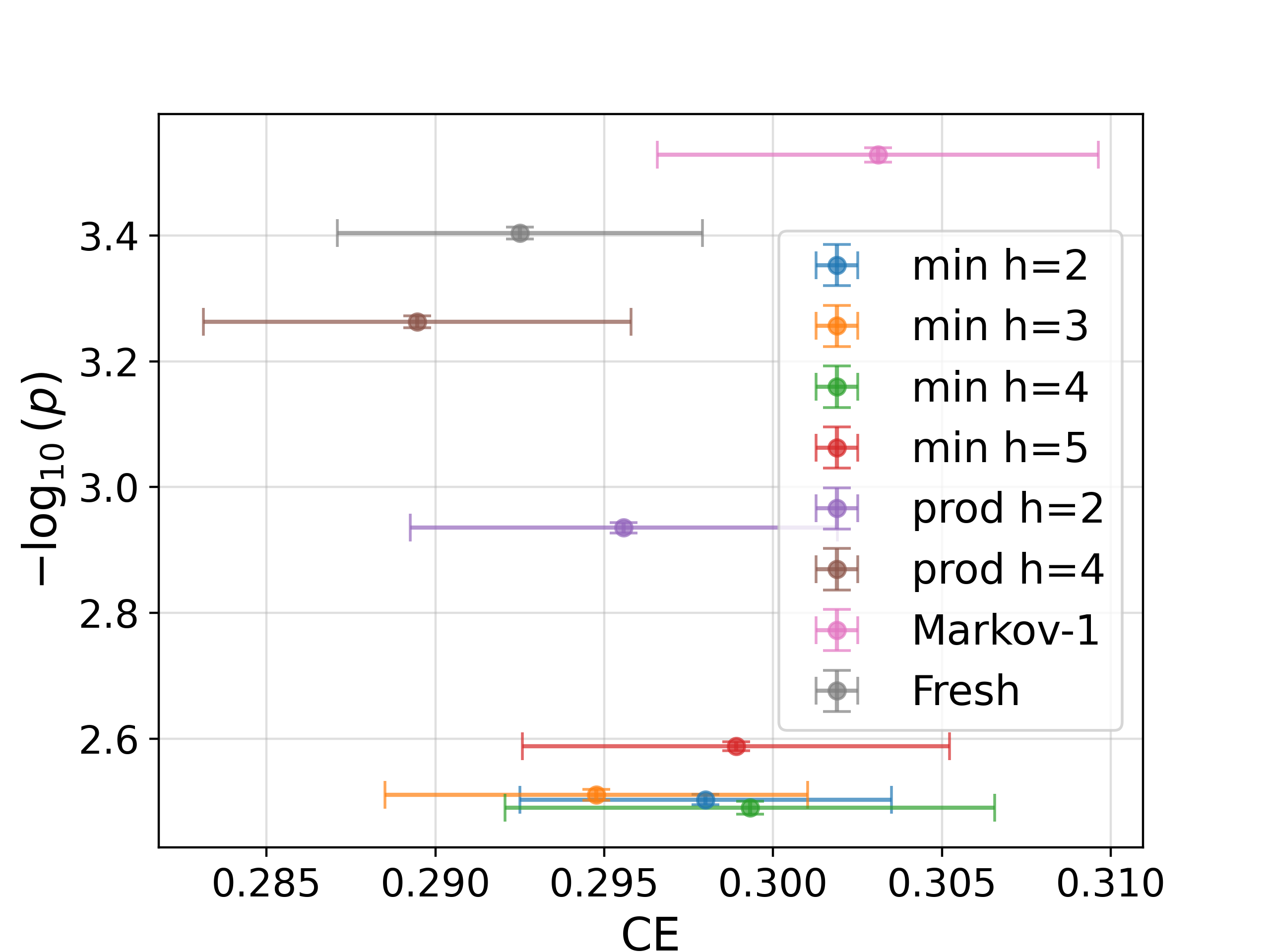}
    \caption{Cross Entropy vs $-\log_{10}(p)$.}
    \label{fig:sink_eps_ce}
  \end{subfigure}%
  % \hspace{1.5cm}
  \hfill
  % Subfigure 2
  \begin{subfigure}[t]{0.45\linewidth}
    \centering
    \includegraphics[width=\linewidth]{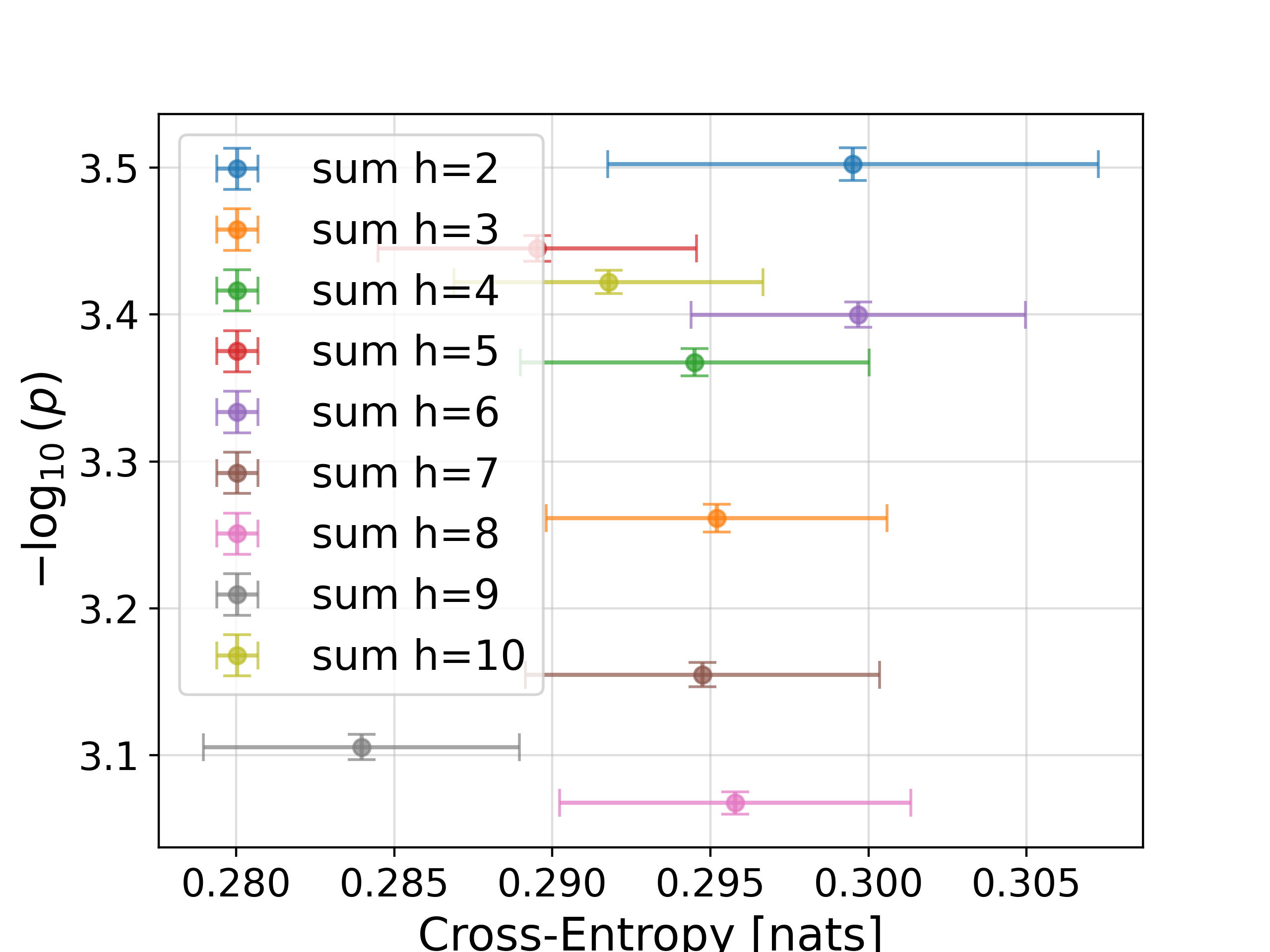}
    \caption{Cross Entropy vs $-\log_{10}(p)$.}
    \label{fig:ce_vs_p_seeding}
  \end{subfigure}
  \caption{Seed Ablation visualized in the detection-distortion plane. It is visible that the performance of our watermark is consistent across an array of hashing scheme and context window sizes.}
  % \hfill
  \label{fig:figs_seed_ablation_plane}
\end{figure*}

To that end, in this section we test the effect of various popular hashing schemes in the performance of our watermarks. As both \simplexwater{} and \heavywater{} follow the same watermarking algorithm we anticipate them to demonstrate similar dependence on the hashing scheme. We therefore prioritize an extensive study on a single watermark - \simplexwater{}.
For a given sliding window size $h$, we consider the following seed generation functions:
\begin{enumerate}
    \item $\mathsf{min}$-hash, which takes the minimum over token-ids and multiplies it with the secret key, i.e. $\mathsf{seed} = \mathsf{min}(x_{t-1},\dots,x_{t-h})\cdot r$.
    \item $\mathsf{sum}$-hash, which takes the sum of the token-ids and multiplies it with the secret key, i.e. $\mathsf{seed} = \mathsf{sum}(x_{t-1},\dots,x_{t-h})\cdot r$.
    \item $\mathsf{prod}$-hash, which takes the product of the token-ids and multiplies it with the secret key, i.e. $\mathsf{seed} = \mathsf{prod}(x_{t-1},\dots,x_{t-h})\cdot r$.
    \item Markov-1 scheme, which considers $h=1$.
\end{enumerate}
We present performance across the aforementioned schemes, considering several values of $h$. As seen from Figure \ref{fig:figs_seed_ablation_plane}, the change of seed generation scheme is does have a significant effect on the overall detection-distortion tradeoff. Furthermore, as emphasize in Figure \ref{fig:figs_seed_ablation}, the size of the sliding window also results in a negligible effect on the watermark performance.

\subsection{Experiment: Robustness To Textual Attacks}\label{apdx:robustness}
The watermarks in this paper are obtained by optimizing a problem that encodes the tradeoff between detection and distortion under worst case distribution. 
To that end, the proposed watermarks are not theoretically optimized for robustness guarantees. However, robustness is often a byproduct of the considered randomness generation scheme, as text edit attacks mainly effect the context from which the seed is generated. A discrepancy in the seed results in a discrepancy in the shared side information sample $s$.
However, regardless of the seed generation scheme, one has to choose a score function and a watermarked distribution design.

In this section, we show that, while not optimized for robustness directly, \simplexwater{} and \heavywater{} demonstrate competitive performance in terms on robustness to common textual edit attacks.
We consider the setting from the watermarking benchmark MarkMyWords \cite{piet2023mark}.
We compare our performance with the Red-Green watermark and the Gumbel watermark.
We choose the value of $\delta$ for the Red-Green watermark such that its cross-entropy distortion is comparable with \simplexwater{}, \heavywater{} and Gumbel ($\delta=1$).

\begin{figure}[!ht]
    \centering
    \includegraphics[width=0.6\linewidth]{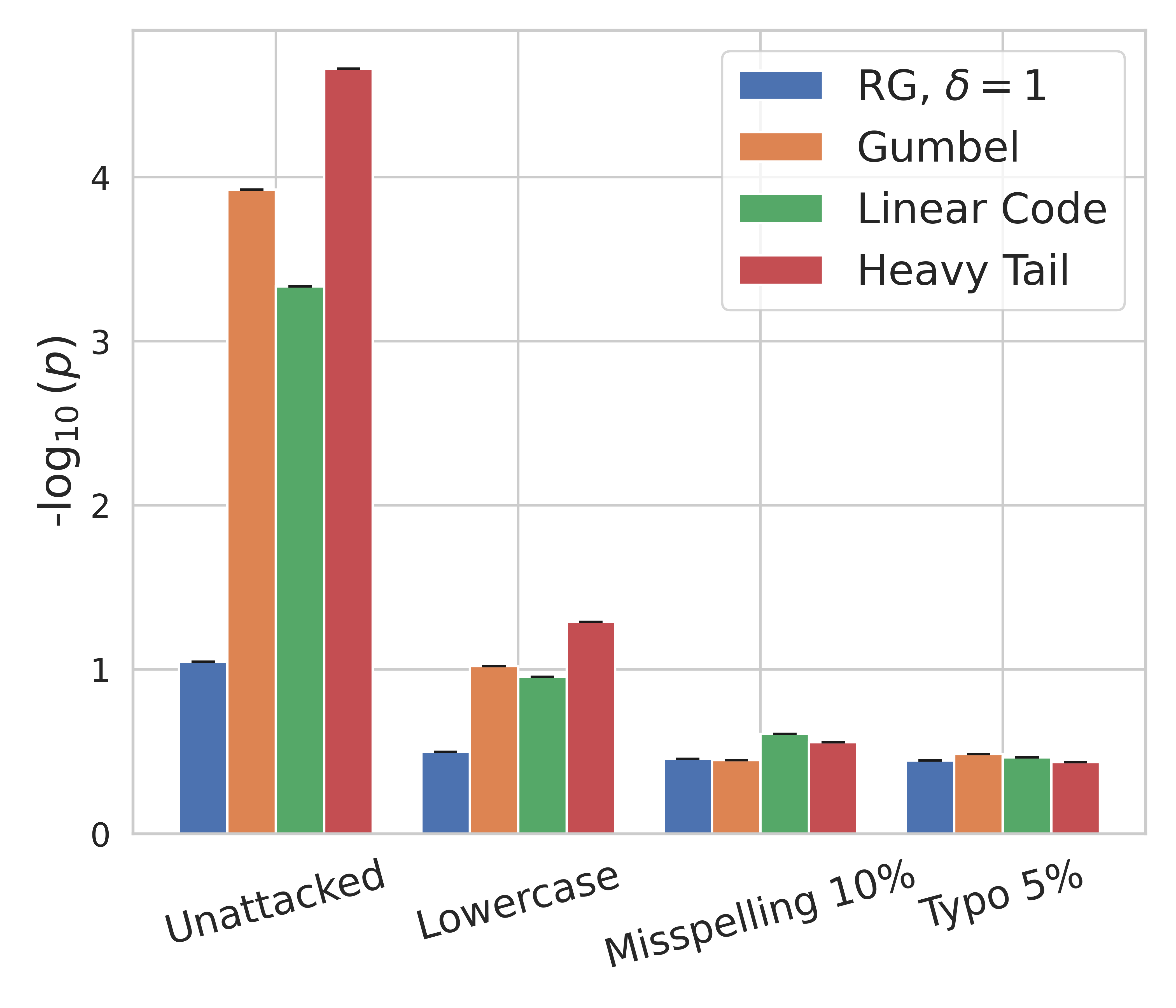}
    \caption{Robustness to attacks --- \heavywater{} demonstrates equal or superior detection performance, as measured by $-\log_{10}(p)$, across a variety of attacks involving edits to generated outputs.}
    \label{fig:robustness}
\end{figure}

We consider three attacks:
\begin{enumerate}
    \item A $\mathsf{Lowercase}$ attack, in which all the characters are replaced with their lowercase version.
    \item A $\mathsf{Misspelling}$ attack, in which words are replaced with a predetermined misspelled version. Each word is misspelled with probability $0.1$.
    \item A $\mathsf{Typo}$ attack, in which, each character is replaced with its neighbor in the $\mathsf{QWERTY}$ keyboard. A character is replaced with probability $0.05$.
\end{enumerate}
As seen in Figure \ref{fig:robustness}, our schemes demonstrate strong robustness under the considered attacks, resulting in the highest detection capabilities in $3$ out of $4$ cases and competitive detection power in the $4$th. This implies that, even though \simplexwater{} and \heavywater{} are not designed to maximize robustness, the resulting schemes show competitive resilience to common text edit attacks.

\subsection{Computational Overhead}\label{apdx:overhead}
We analyze the computational overhead induced by the considered watermarking scheme. Theoretically, Sinkhorn's algorithm has an iteration computational complexity of $O(km)$ for token vocabulary of size $|\cX|=m$ and side information of alphabet $|\cS|=k$ due to its vector-matrix operations.
In practice, watermarking is a single step within the entire next token generation pipeline. 

\begin{figure}[!ht]
    \centering
    \includegraphics[width=0.5\linewidth]{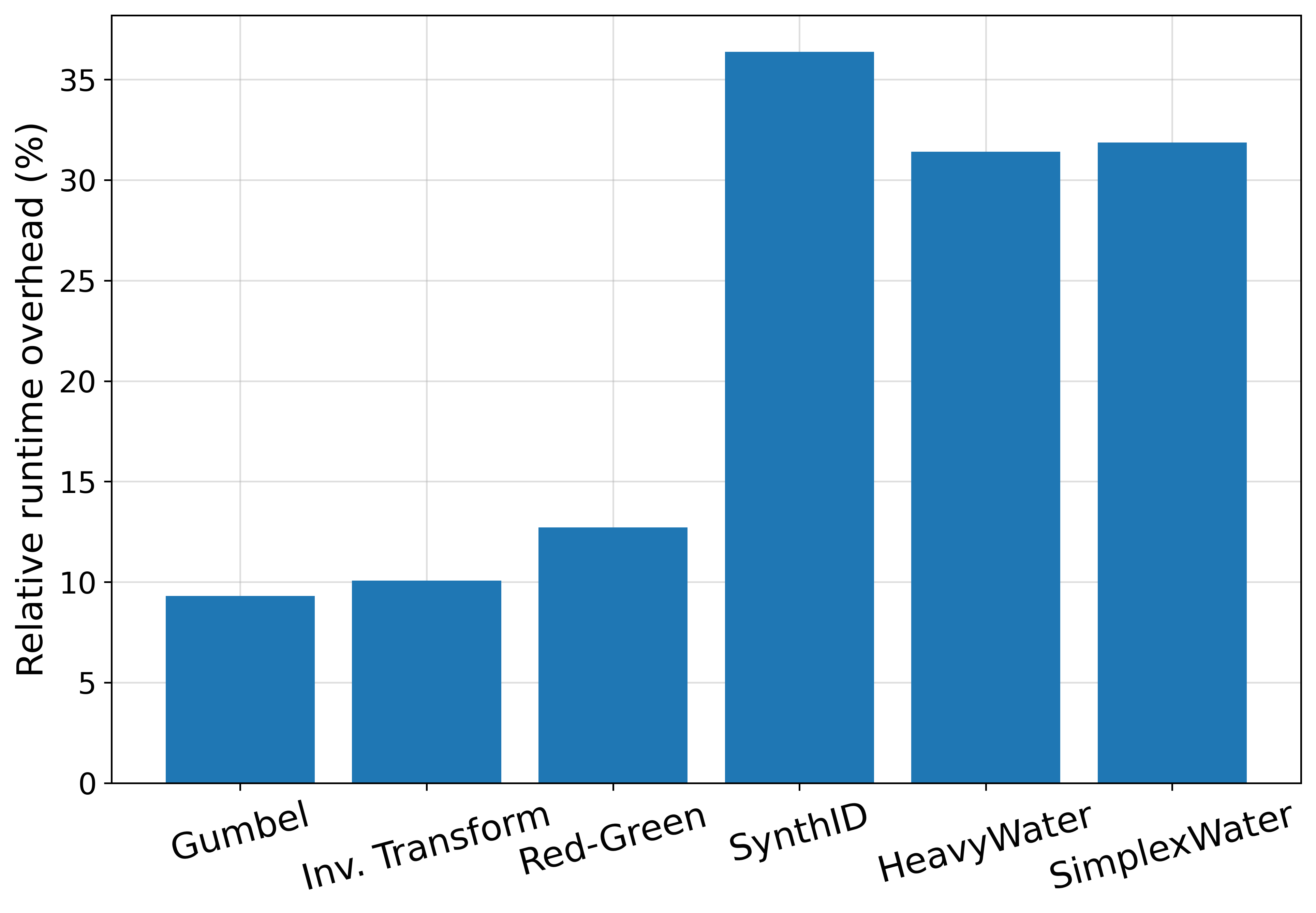}
    \caption{Computational overhead over unwatermarked text generation, Llama2-7b.}
    \label{fig:time_overhead}
\end{figure}

We analyze the computational overhead induced by applying \simplexwater{} and \heavywater{}. Figure \ref{fig:time_overhead} shows the overhead of watermarking in a few common watermarks - Red-Green \cite{kirchenbauer2023watermark}, Gumbel \cite{aaronson2023watermark}, Inverse-transform \cite{kuditipudi2023robust}, SynthID \cite{dathathri2024scalable} and our watermarks.
It can be seen that The Gumbel, Inverse transform and Red-Green watermarks induce a computational overhead of $\sim 10\%$, while \simplexwater{}, \heavywater{} and SynthID induce an overhead of $\sim 30 \%$.
While this overhead is not negligible, our methods demonstrate superior performance over considered methods. However, replacing a 'fast', yet 'weaker' watermark with ours boils down to a difference in $\sim 20\%$ increase in generation time.
We consider an implementation of the SynthID through vectorized tournament sampling with a binary score function and $15$ tournament layers, which is the method reported in the main text experiments.
As we previously mentioned, we consider top-$p$ sampling with $p$=0.999.
We note that, in many text generation schemes, lower top-$p$ values, which accelerate Sinkhorn's algorithm's runtime, thus further closing the computational gap.

\subsection{Experiment: Alternative Quality Metrics for Textual and Coding Tasks}\label{apdx:waterbench}
This paper focused on distortion as the proxy for textual quality. This is a common practice in watermarking (e.g. \cite{aaronson2023watermark,fu2024gumbelsoft,kuditipudi2023robust,dathathri2024scalable,long2025optimized,hetheoretically}).
Distortion is measured by the discrepancy between the token distribution $P_X$ and the expected watermarked distribution $\mathbb{E}_{S}[P_{X|S}]$.
In practice, distortion and textual quality are often measured with some perplexity-based measure (e.g. cross-entropy in this paper).
However, as explored in the WaterBench benchmark \cite{tu2023waterbench}, such measures are not guaranteed to faithfully represent degradation in textual quality.

To that end, WaterBench proposed an array of alternative generation metrics, whose purpose us to evaluate the quality of generated watermarked text, and are tailored for specific text generation tasks.
We consider $4$ datasets from the WaterBench benchmark \cite{tu2023waterbench}:
\begin{enumerate}
    \item Longform QA \cite{fan2019eli5}: A dataset of 200 long questions-answer generation prompts. The considered generation metric is the ROGUE-L score.
    \item Knowledge memorization: A closed-ended entity-probing benchmark drawn from KoLA \cite{yu2023kola}, consisting of 200 triplets sampled at varying frequencies from Wikipedia the test an LLM’s factual recall. The considered generation metric is the F1 score as it is a factual knowledge dataset.
    \item Knowledge understanding \cite{peng2022copen}: A dataset of 200 questions that demonstrate the LLM's understanding of various concepts. The considered generation metric is the F1 score as it is a factual knowledge dataset.
    \item Multi-news summarization: A collection of 200 long news clusters, coupled with summarization prompts. The score here is the ROGUE-L score.
\end{enumerate}

As seen in Table \ref{tab:waterbench}, our watermarks maintain competitive performance in the considered set of textual generation tasks, even under alternative text generation evaluation metrics.

To better evaluate quality of generated watermarked code, we include two coding-centric evaluation metrics that assess different aspects of code quality.

\textbf{Pass@K Evaluation (Functional Correctness)} We evaluate functional correctness using the HumanEval dataset \cite{chen2021evaluating}, which is a popular benchmark for code generation. Pass@K measures the empirical probability of a solution passing all unit tests among K generated solutions. We measure pass@k with $k \in {1, 5, 10}$, where we execute the generated code against unit tests provided in the dataset in a sandbox environment. As shown in the table below, our methods outperformed the competitors: (i) SimplexWater achieves the highest pass@5 (22.7\%), outperforming all baselines including unwatermarked text (20.7\%) and (ii) HeavyWater achieves the highest pass@10 (27.8\%), nearly matching unwatermarked performance (28.1\%). In all cases, performance is comparable to unwatermarked code (see Table \ref{tab:humaneval-passk}).

\textbf{Edit\_Sim (Edit Similarity)} We use this metric on the LCC coding dataset \cite{chen2021evaluating} which focuses on code completion: generating the next few lines of code given a long context. Since the dataset is sourced from GitHub, human-written ground-truth completions are available. Edit\_Sim is defined as the Levenshtein similarity between generated and ground-truth code.

Let $s_1, s_2 \in \Sigma^*$ be two strings over an alphabet $\Sigma$. 
The \emph{Levenshtein distance} $d_L(s_1, s_2)$ is defined as the minimum number of 
single-character edit operations (insertions, deletions, or substitutions) required 
to transform $s_1$ into $s_2$. The \emph{Levenshtein similarity} (normalized form) is then defined as:
\[
\mathrm{LevSim}(s_1, s_2) = 1 - \frac{d_L(s_1, s_2)}{\max(|s_1|, |s_2|)}
\]
where $|s|$ denotes the length of string $s$. 
Thus, $\mathrm{LevSim}(s_1, s_2) \in [0,1]$, with 
$\mathrm{LevSim}(s_1, s_2) = 1$ if and only if $s_1 = s_2$.

We benchmark our watermarking methods against prior work using this metric. The results are provided in the Table \ref{tab:lcc-editsim}. Specifically, HeavyWater ties for best performance 0.52. 
These two metrics support that HeavyWater and SimplexWater have no significant impact on the quality of generated code relative to unwatermarked code, and perform favorably when compared to competing methods both in terms of detection accuracy and generation quality. 

\textbf{Code Examples.} We also include an example of generated code before and after watermarking with HeavyWater.

\textbf{Unwatermarked Output:}

\begin{lstlisting}
def has_close_elements(numbers: List[float], threshold: float) -> bool:
    # Sort the list of numbers
    numbers = sorted(numbers)

    # Find the first two numbers in the list 
    that are closer to each other than the threshold
    min_distance = float('inf')
    for i in range(1, len(numbers)):
        distance = abs(numbers[i] - numbers[i - 1])
        if distance < min_distance and distance < threshold:
            min_distance = distance

    # If we found a pair of numbers that are 
    closer to each other than the threshold, return True
    return min_distance < threshold
\end{lstlisting}

\noindent\textbf{Unit Tests result:} Pass $\rightarrow$ Pass@1 = 100\% on this example prompt.

\vspace{1em}

\textbf{HeavyWater’s Watermarked Output:}

\begin{lstlisting}
def has_close_elements(numbers: List[float], threshold: float) -> bool:
    close_pairs = []
    for i in range(len(numbers)):
        closest = None
        closest_distance = float("inf")
        for j in range(i+1, len(numbers)):
            distance = abs(numbers[i] - numbers[j])
            if distance < closest_distance:
                closest = numbers[j]
                closest_distance = distance
        if closest and closest_distance <= threshold:
            close_pairs.append((numbers[i], closest))
    return len(close_pairs) > 0
\end{lstlisting}

\noindent\textbf{Unit Tests result:} Pass $\rightarrow$ Pass@1 = 100\% on this example prompt.

\noindent\textbf{Example: Code Completion under Long Context}

\noindent This example was shortened for brevity. 

\vspace{0.5em}
\noindent\textbf{Prompt: Please complete the code given below.}

\begin{lstlisting}
#!/usr/bin/env python
# -- coding: utf-8 --

from HttpUtils import App, buildOpener

class Device(object):
    def __init__(self, token):
        self.token = token
        self.app = App()

    def check_inspection(self):
        data = self.app.check_inspection()
        return data
    [abbreviated]

class Exploration(object):
    def __init__(self, app):
        self.app = app

    def getAreaList(self):
        data = self.app.exploration_area()
        return data
    [abbreviated]

class User(object):
    [abbreviated]

class RoundTable(object):
    [abbreviated]

class Menu(object):
    [abbreviated]

if __name__ == "__main__":
    from config import deviceToken, loginId, password
\end{lstlisting}

\vspace{0.5em}
\noindent\textbf{Answers provided by different methods:}

\noindent\textbf{Human Answer / Ground Truth:}
\begin{lstlisting}[frame=none]
device = Device(deviceToken)
\end{lstlisting}

\noindent\textbf{No Watermark} $\rightarrow$ Edit\_Sim = 0.94
\begin{lstlisting}[frame=none]
dev = Device(deviceToken)
\end{lstlisting}

\noindent\textbf{HeavyWater / Inverse-Transform / SimplexCode} $\rightarrow$ Edit\_Sim = 1
\begin{lstlisting}[frame=none]
device = Device(deviceToken)
\end{lstlisting}

\noindent\textbf{Red/Green} $\rightarrow$ Edit\_Sim = 0.9
\begin{lstlisting}[frame=none]
d = Device(deviceToken)
\end{lstlisting}

\vspace{0.5em}

\subsection{Alternative Detection Metric}\label{apdx:pdpfa}
In this section we provide results on an additional detection metric.
We consider the detection probability under a false-alarm (FA) constraint.
As we consider watermarks from which $p$-values can be calculated, we can impose such a FA constraint.
For a given set of responses obtained from a dataset of prompts, we are interested in calculating an estimate of the detection probability at some FA constraint, given a set of $p$-values, each calculated for one of the responses.
We obtain an estimate of the detection probability at a given FA constraint by taking the ratio of responses whose $p$-value is lower than the proposed $p$-value threshold, over the total number of responses.

\begin{figure}[!ht]
    \centering
    \begin{subfigure}[b]{0.45\textwidth}
        \centering
        \includegraphics[trim={0pt 0pt 0pt 13pt}, clip,width=\textwidth]{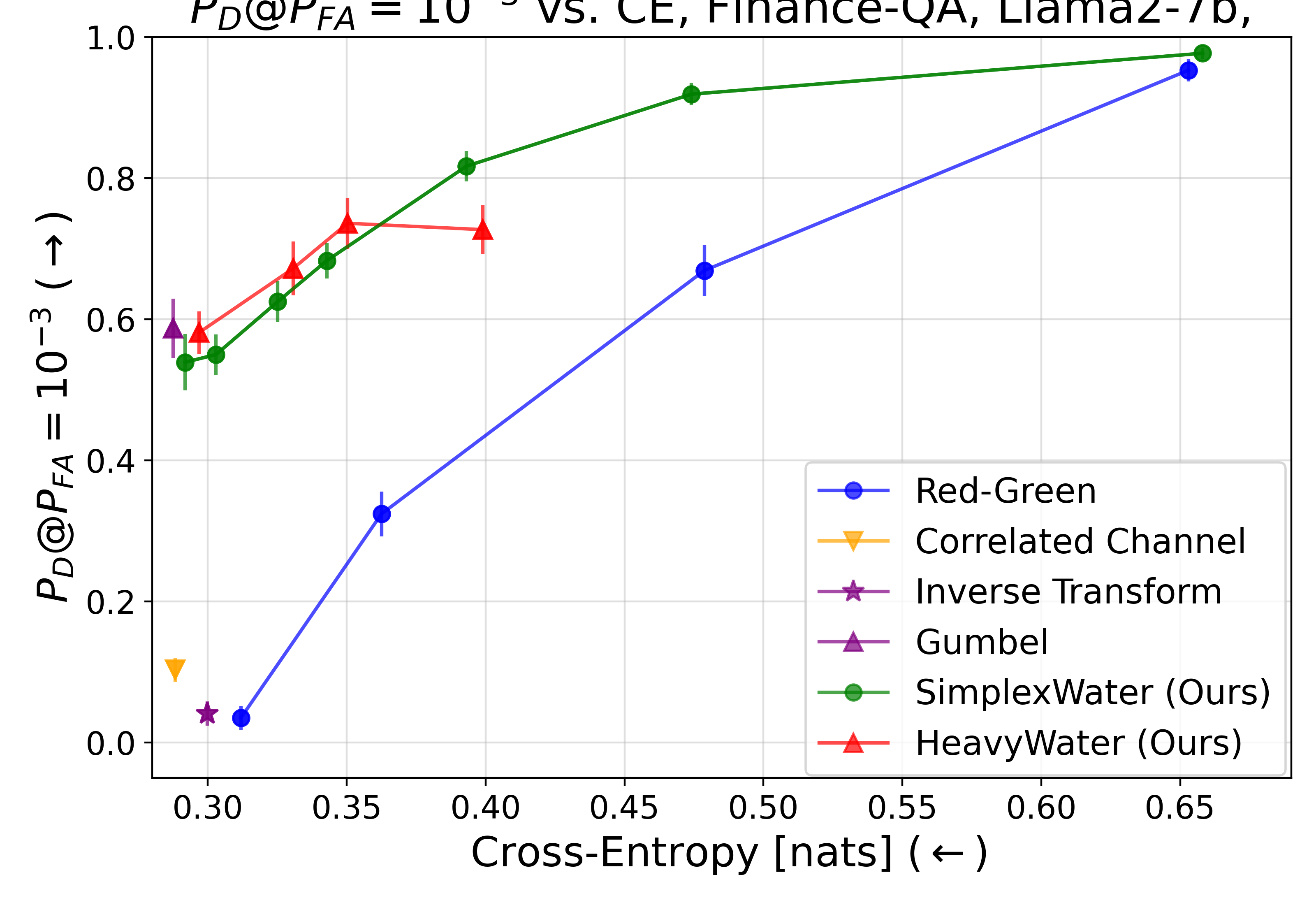}
        \label{fig:1e3}
        \caption{$P_D$ at $P_{FA}=10^{-3}$}
    \end{subfigure}
    \hfill
    \begin{subfigure}[b]{0.45\textwidth}
        \centering
        \includegraphics[width=\textwidth]{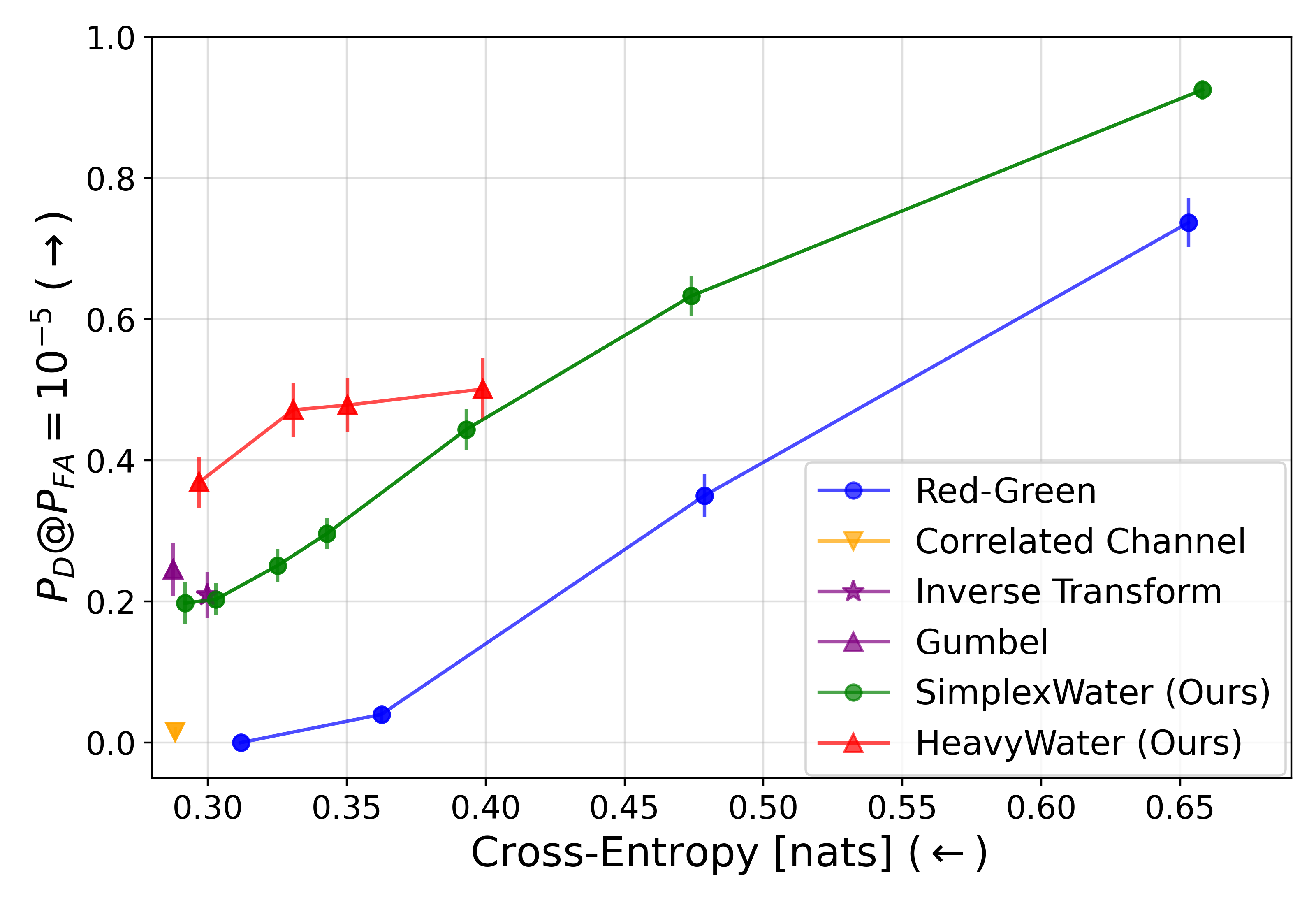}
        \label{fig:1e5}
        \caption{$P_D$ at $P_{FA}=10^{-5}$}
    \end{subfigure}
    \caption{Detection probability at a given false alarm constraint. LLama2-7b, Finance-QA dataset.}
    \label{fig:1e31e5}
\end{figure}

\begin{figure}
    \centering
    \includegraphics[width=0.45\linewidth]{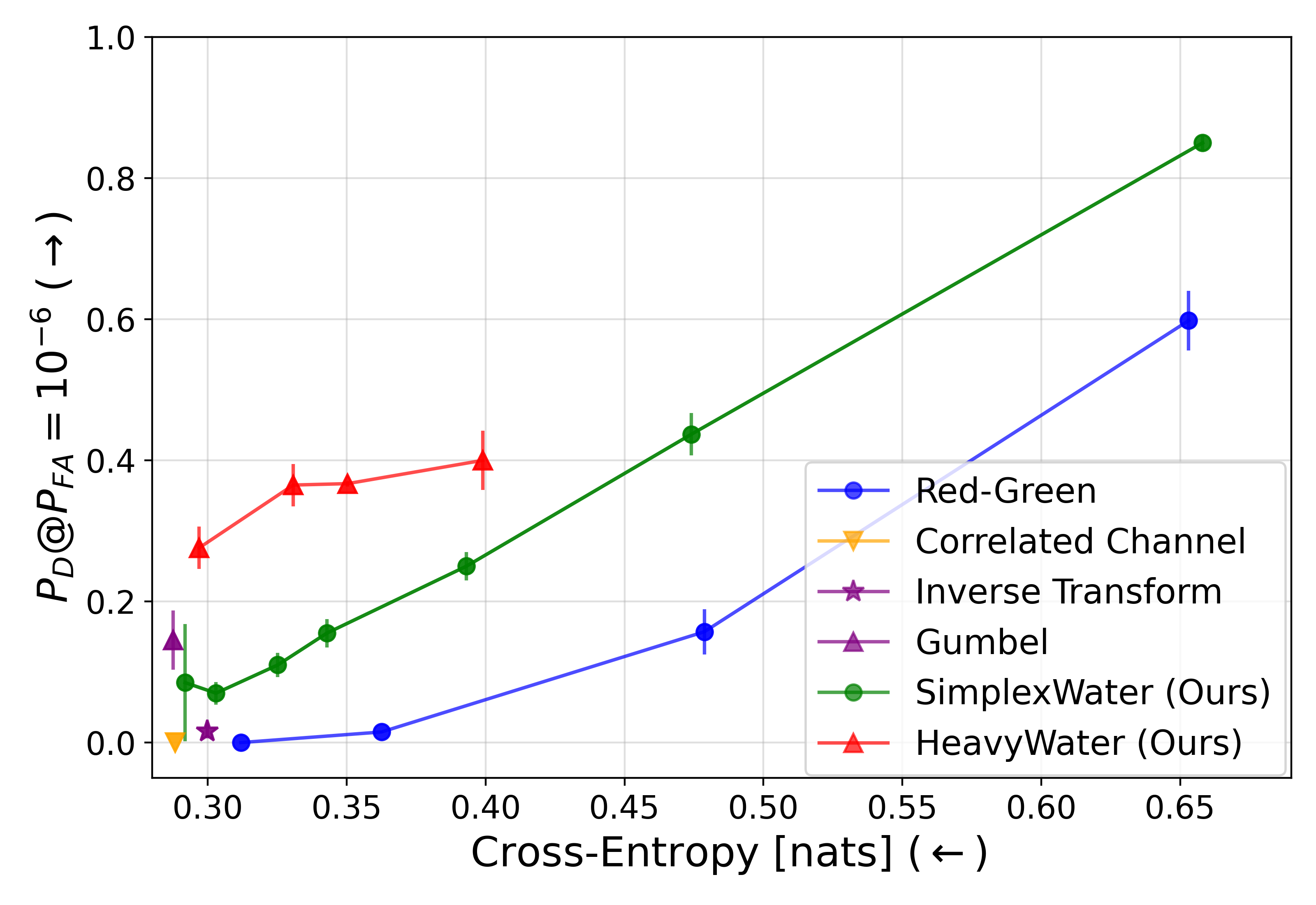}
    \caption{$P_D$ at $P_{FA}=10^{-6}$,  LLama2-7b, Finance-QA dataset.}
    \label{fig:1e6}
\end{figure}

We provide results on the FinanceQA dataset using Llama2-7b. 
We consider several FA values and visualize the resulting tradeoff curves in Figures \ref{fig:1e31e5} and \ref{fig:1e6}.
To obtain error-bars, we consider the following bootstrapping technique: Out of the $200$ responses, we randomly sample 200 subsets with $150$ responses and calculate the corresponding metric.
From the set of $150$ results we provide error-bars, considering the average value and standard deviation.
It can be seen that the trends presented in Figures \ref{fig:fig1} and \ref{fig:tradeoff_llama2} are preserved under the considered detection metric.

\subsection{Additional Detection-Distortion Tradeoff Results}\label{apdx:additional_tradeoff}
\begin{figure}[htbp]
    \centering
    \begin{subfigure}[b]{0.4\textwidth}
        \centering
        \includegraphics[width=\textwidth]{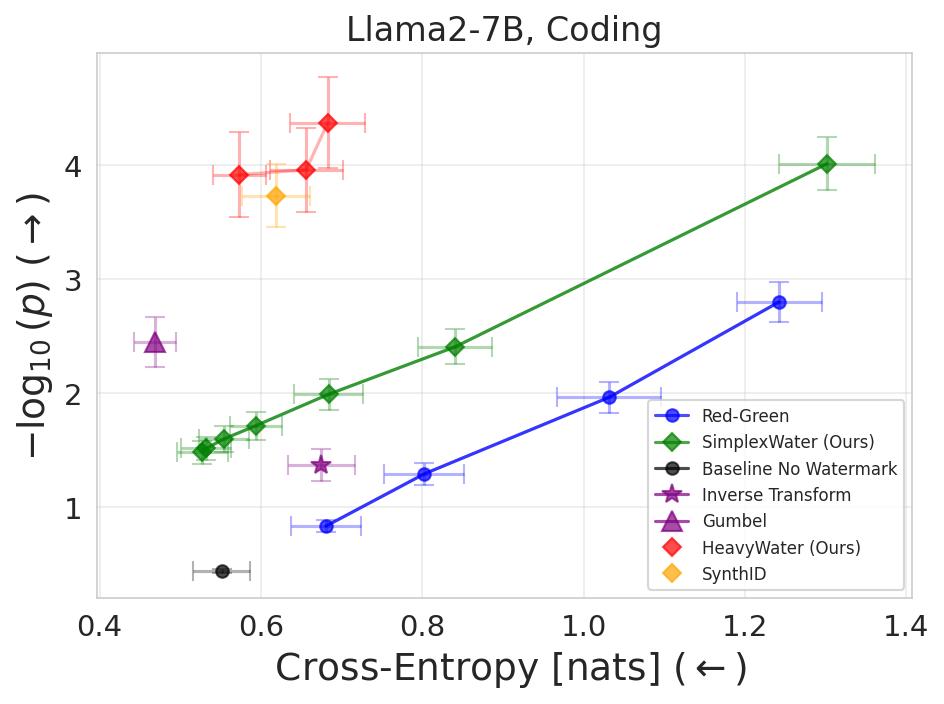}
        \label{fig:1}
    \end{subfigure}
    \hfill
    \begin{subfigure}[b]{0.4\textwidth}
        \centering
        \includegraphics[width=\textwidth]{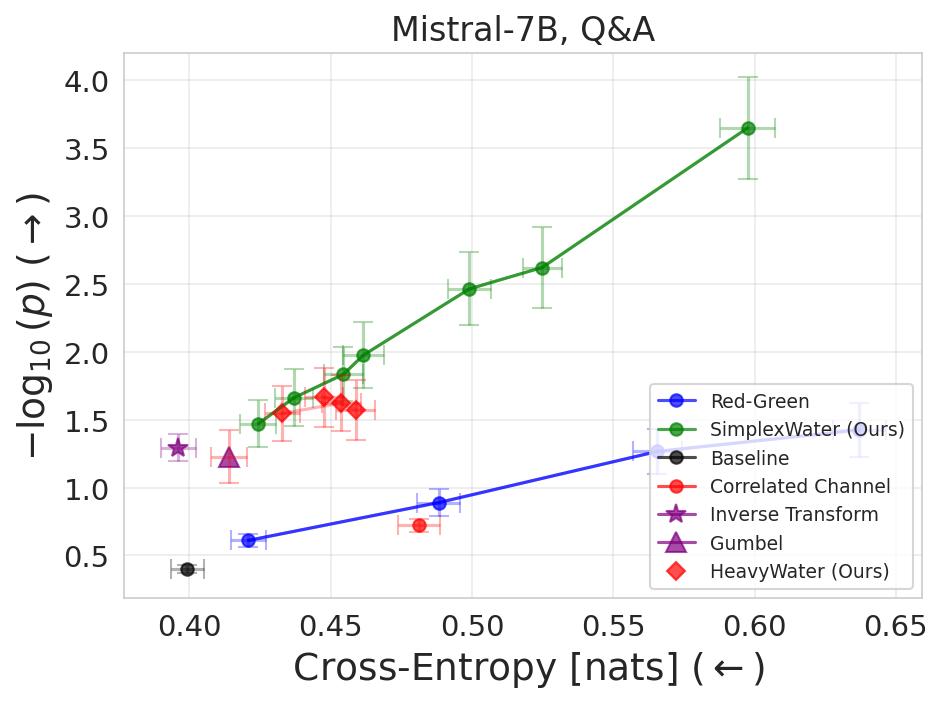}
        \label{fig:2}
    \end{subfigure}
    
    \vspace{0.5cm}
    
    \begin{subfigure}[b]{0.4\textwidth}
        \centering
        \includegraphics[width=\textwidth]{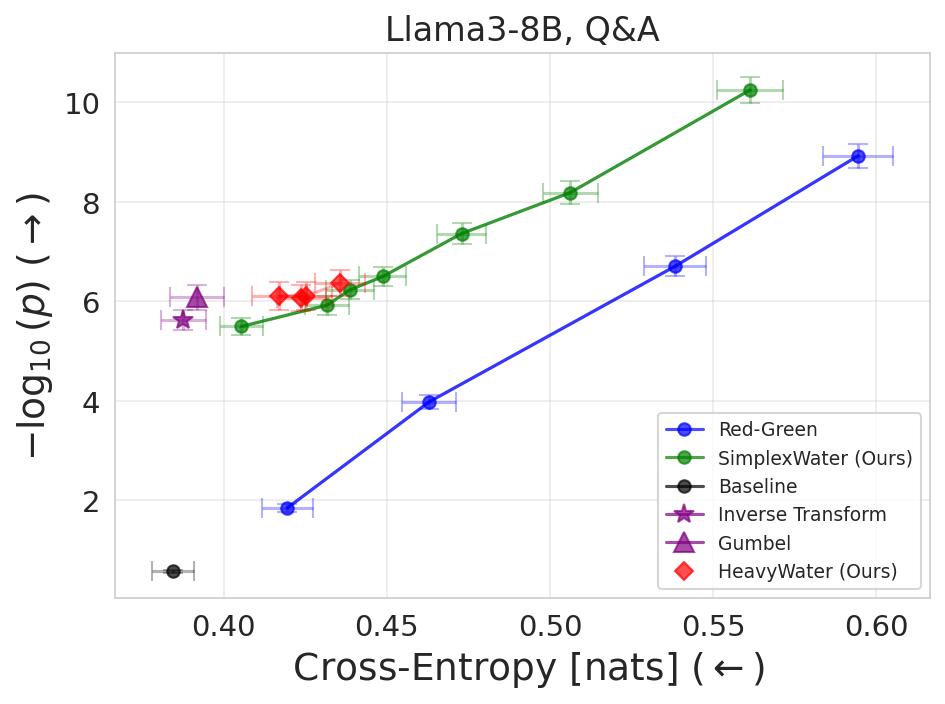}
        \label{fig:3}
    \end{subfigure}
    \hfill
    \begin{subfigure}[b]{0.4\textwidth}
        \centering
        \includegraphics[width=\textwidth]{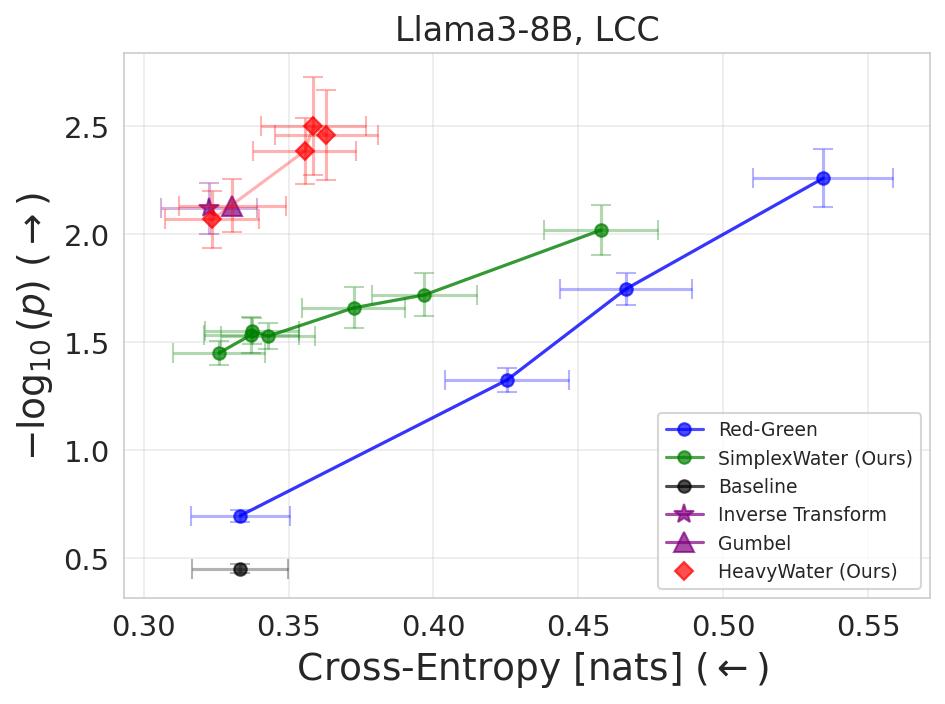}
        \label{fig:4}
    \end{subfigure}
    
    \caption{Detection-distortion tradeoffs on multiple models and tasks.}
    \label{fig:additional_detection_distortion_tradeoff}
\end{figure}

\begin{figure}[htbp]
  \centering
  \begin{subfigure}{1\textwidth}
    \includegraphics[width=\linewidth]{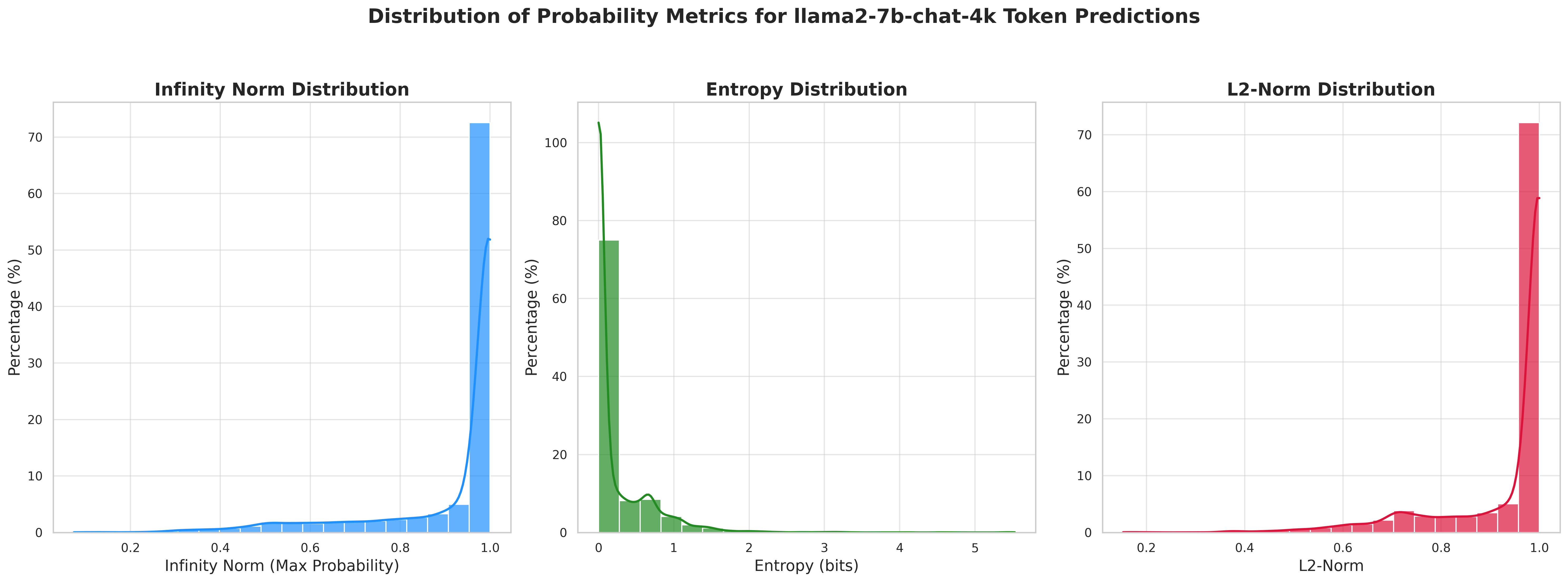}
  \end{subfigure}

  \begin{subfigure}{1\textwidth}
    \includegraphics[width=\linewidth]{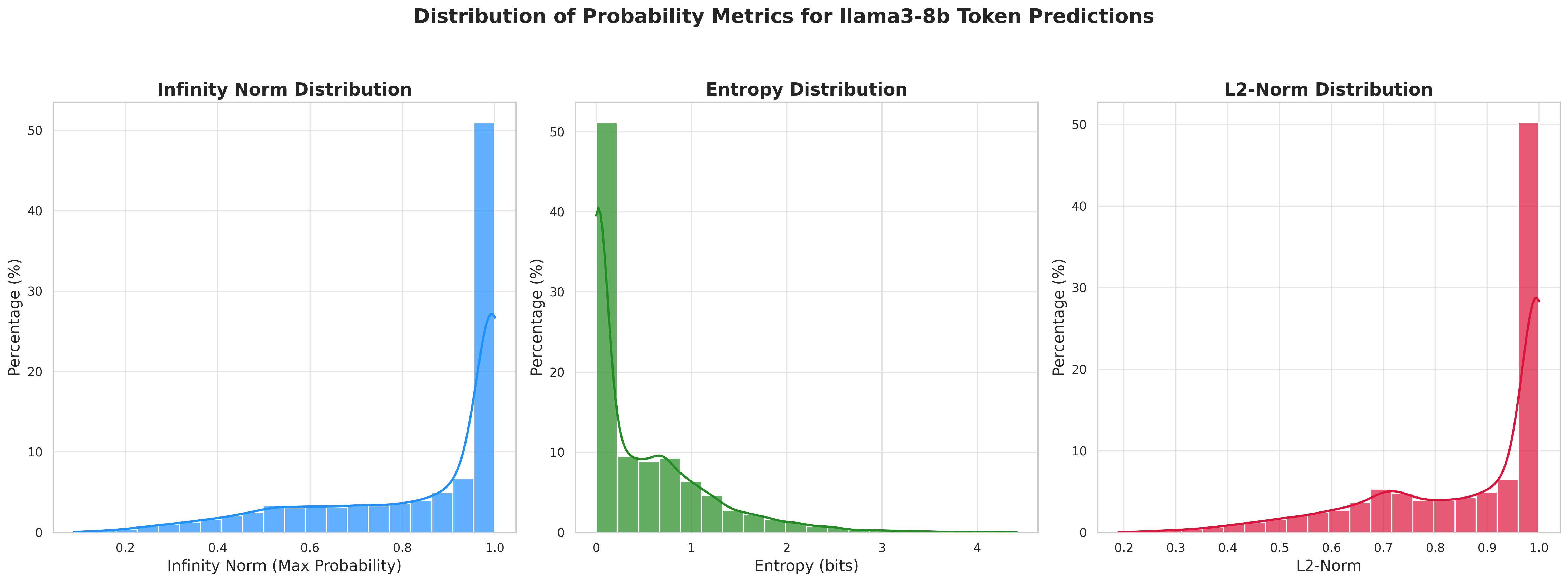}
  \end{subfigure}

  \begin{subfigure}{1\textwidth}
    \includegraphics[width=\linewidth]{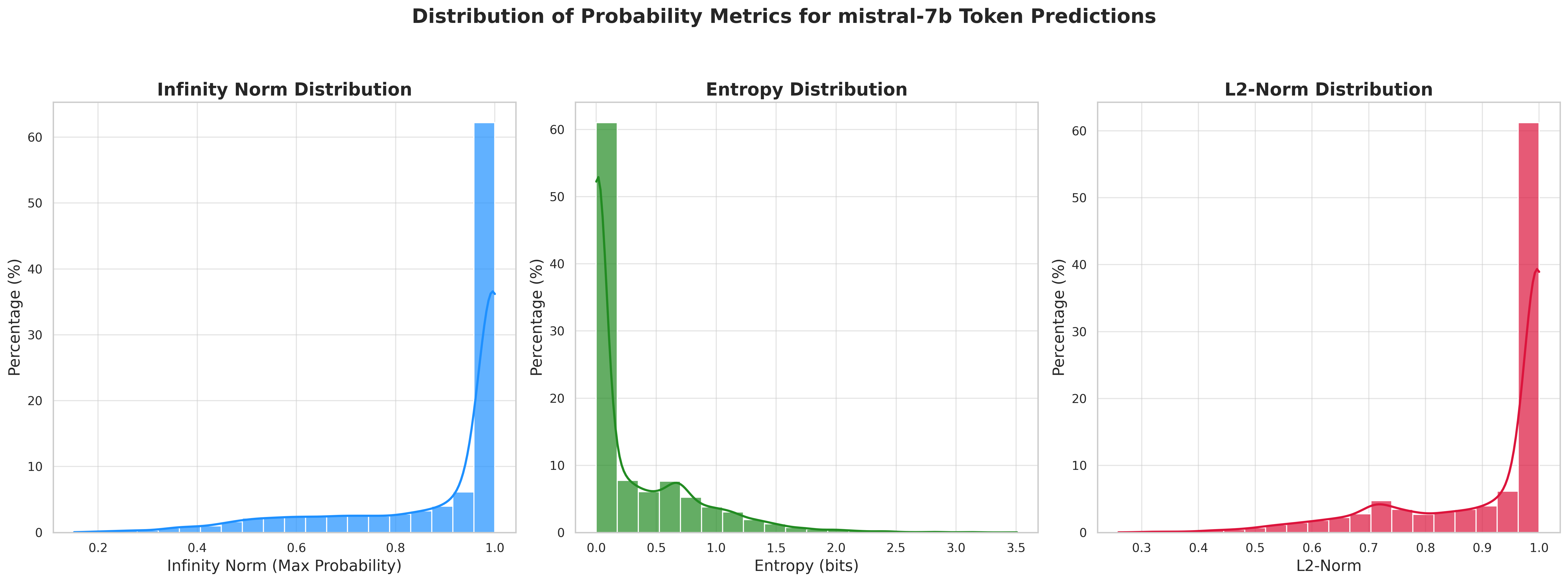}
  \end{subfigure}

  \caption{Histograms of statistics of token distributions on \textbf{Q\&A} dataset. 90\% token distributions fall into the low-entropy regime with infinity norm greater than $1/2$, i.e. $\max_x P(x)\geq \frac{1}{2}$.}
  \label{fig:histogram_stats_qa}
\end{figure}

\begin{figure}[htbp]
  \centering

  \begin{subfigure}{1\textwidth}
    \includegraphics[width=\linewidth]{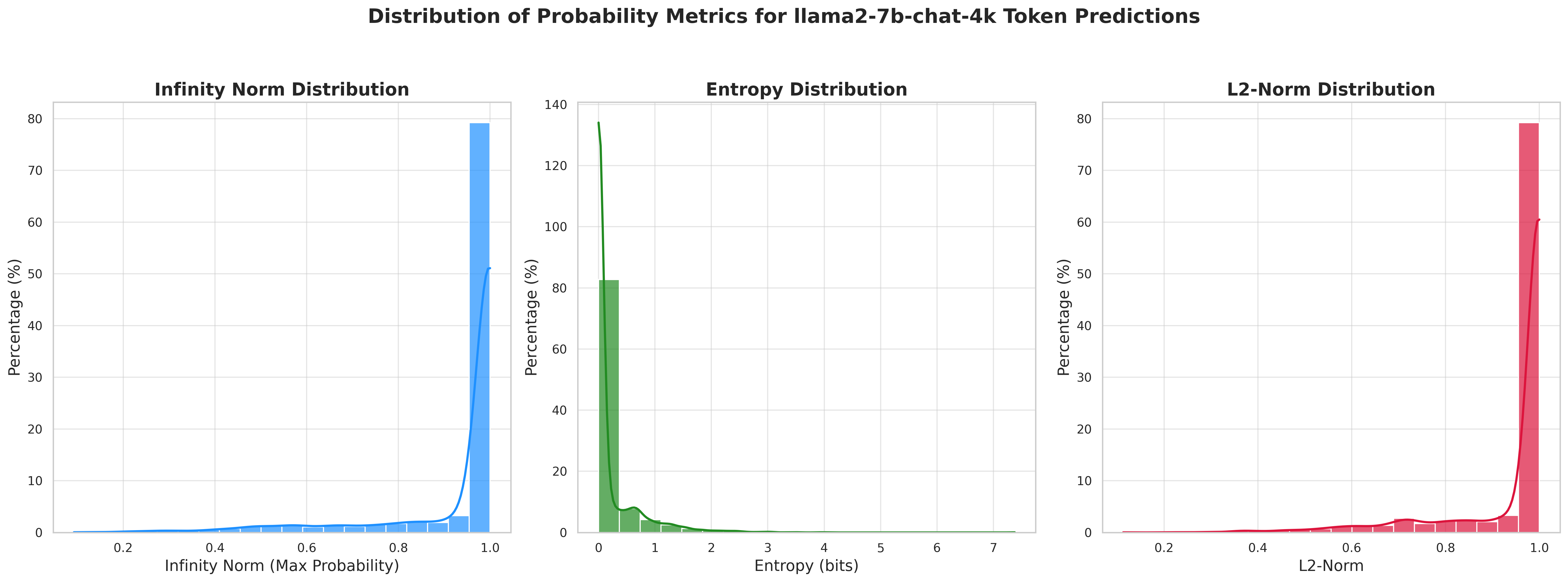}
  \end{subfigure}

  \begin{subfigure}{1\textwidth}
    \includegraphics[width=\linewidth]{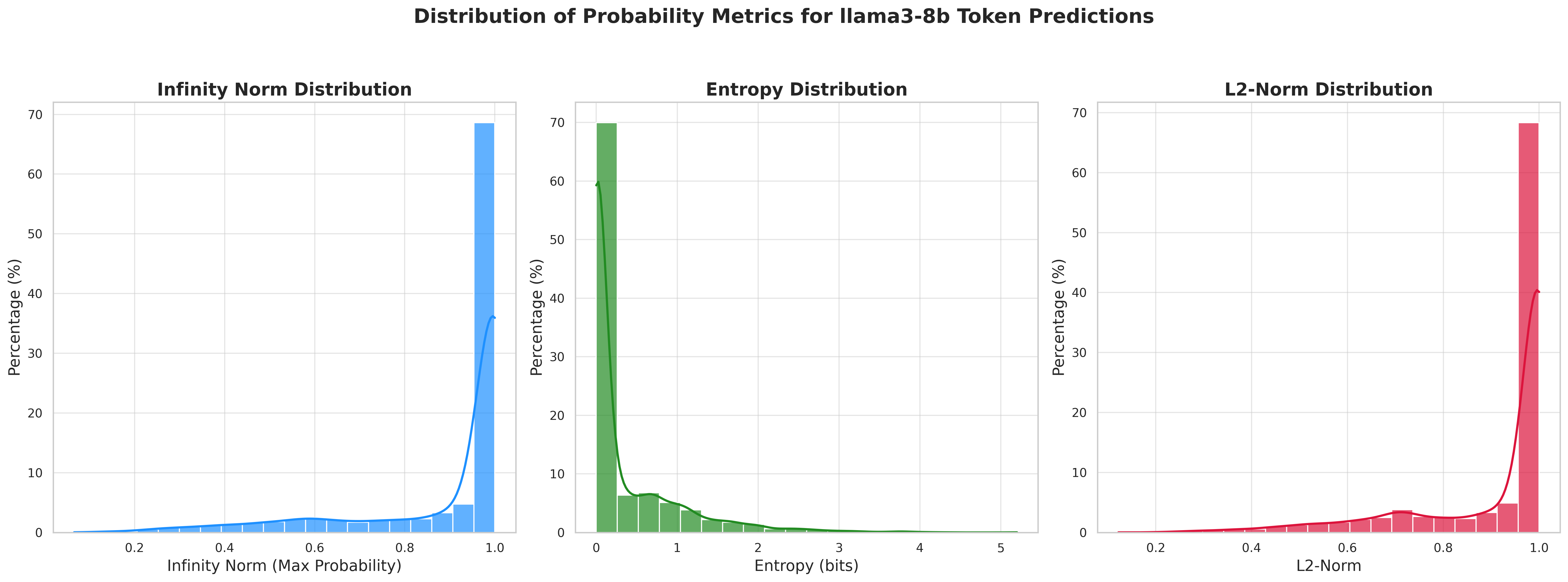}
  \end{subfigure}

  \begin{subfigure}{\textwidth}
    \includegraphics[width=\linewidth]{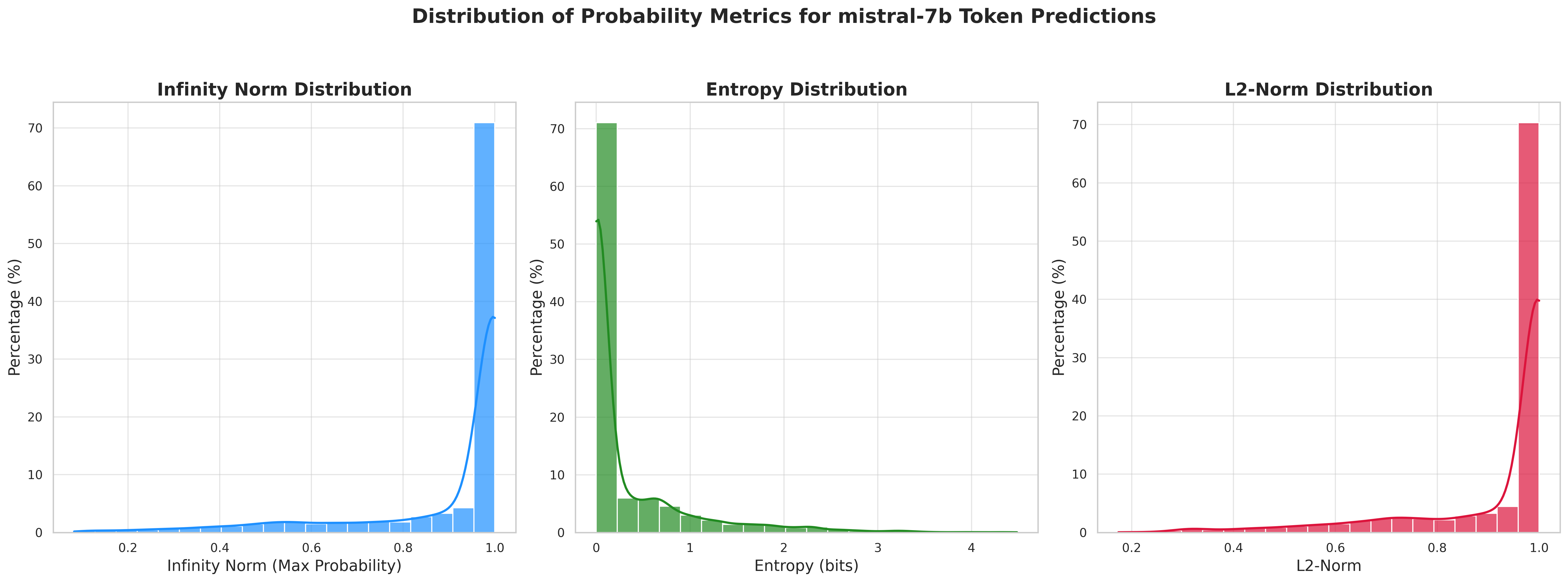}
  \end{subfigure}

  \caption{Histograms of statistics of token distributions on \textbf{coding} dataset. 93\% token distributions fall into the low-entropy regime with infinity norm greater than $1/2$, i.e. $\max_x P(x)\geq \frac{1}{2}$.}
  \label{fig:histogram_stats_coding}
\end{figure}

We provide results that explore the detection-distortion tradeoff, in addition to ones presented in Fig. \ref{fig:fig1} and Fig. \ref{fig:tradeoff_llama2}.
We run three models (Llama2-7b, Llama3-8b, Mistral-7b) on two tasks (Q\&A and coding). We employ the popular Q\&A dataset, FinanceQA, and code-completion dataset LCC. Fig. \ref{fig:fig1} shows the result for Llama2-7b on Q\&A, while Fig. \ref{fig:tradeoff_llama2} shows the result for Mistral-7B on coding.  In Fig. \ref{fig:additional_detection_distortion_tradeoff}, we present this tradeoff over the remaining datasets and LLMs. 

\begin{table}[!ht]
  \centering
  \small
  \caption{Performance of Watermarking Methods across Four Datasets}
  \label{tab:waterbench}
  \begin{tabular}{@{} l l c c c @{}} 
    \toprule
    \textbf{Dataset}      & \textbf{Watermark}   & \textbf{Gen.\ Metric $\uparrow$} 
                         & \textbf{\% Drop in GM $\downarrow$} & \textbf{$-\log_{10}p$ $\uparrow$} \\
    \midrule
    \multirow{6}{*}{\textbf{Longform}}
      & Gumbel            & 21.20 &  -0.856 & 8.006 \\
      & HeavyWater        & 21.48 &  -2.188 & \textbf{8.089} \\
      & Simplex           & 21.90 &  \textbf{-4.186} & 4.985 \\
      & Inv.\ Tr.         & 21.27 &  -1.189 & 3.687 \\
      & RG, $\delta=1$    & 21.25 &  -1.094 & 1.456 \\
      & RG, $\delta=3$    & 21.19 &  -0.809 & 7.078 \\
    \midrule
    % \hline
    \multirow{6}{*}{\textbf{Memorization}}
      & Gumbel            &  5.66 &  -2.536 & 1.085 \\
      & HeavyWater        &  5.73 &  \textbf{-3.804} & \textbf{1.605} \\
      & Simplex           &  5.71 &  -3.442 & 0.977 \\
      & Inv.\ Tr.         &  5.38 & 2.536 & 0.792 \\
      & RG, $\delta=1$    &  5.35 & 3.080 & 0.482 \\
      & RG, $\delta=3$    &  5.82 &  5.435 & 0.912 \\
    % \hline
    \midrule
    \multirow{6}{*}{\textbf{Understanding}}
      & Gumbel            & 33.42 &  \textbf{-9.574} & 0.396 \\
      & HeavyWater        & 32.59 &  -6.852 & 0.308 \\
      & Simplex           & 31.50 &  -3.279 & 0.920 \\
      & Inv.\ Tr.         & 27.93 & 8.426 & \textbf{1.045} \\
      & RG, $\delta=1$    & 32.96 &  8.066 & 0.184 \\
      & RG, $\delta=3$    & 33.83 & 10.918 & 0.300 \\
    % \hline
    \midrule
    \multirow{6}{*}{\textbf{MultiNews}}
      & Gumbel            & 25.69 & 2.579 & 3.172 \\
      & HeavyWater        & 25.67 & 2.655 & 3.491 \\
      & Simplex           & 25.86 & \textbf{1.934} & 2.701 \\
      & Inv.\ Tr.         & 25.74 & 2.389 & 1.586 \\
      & RG, $\delta=1$    & 25.85 & 1.940 & 0.963 \\
      & RG, $\delta=3$    & 25.74 & 2.389 & \textbf{3.781} \\

    \bottomrule
  \end{tabular}
\end{table}

\begin{table}[ht]
\centering
\caption{Results on LCC Code Completion Dataset (Edit\_Similarity, higher is better)}
\label{tab:lcc-editsim}
\begin{tabular}{lc}
\toprule
\textbf{Watermark Method} & \textbf{Edit\_Similarity $\uparrow$} \\
\midrule
No Watermark              & 0.45 \\
Gumbel                    & 0.52 \\
\textbf{HeavyWater (Ours)}     & \textbf{0.52} \\
\textbf{SimplexWater (Ours)}   & 0.44 \\
Inverse Transform         & 0.45 \\
Red/Green $\delta=3$      & 0.43 \\
\bottomrule
\end{tabular}
\end{table}

\chapter{Chapter \ref{ch:5}}
\label{apdx:proofs_ch5}

The following tables present the underlying numerical data that support the findings
discussed in the main text. Table~\ref{tab:human_teams_costs} reports the total supply chain costs recorded across eleven runs of the Beer Game played by
human student teams (4-8 students per team, 100+ students in total) from two Georgia Tech cohorts (April 2025 and April 2024),
together with the average cost of \$3,207 that serves as the human performance
benchmark throughout the study. Table~\ref{tab:supply_chain_full} summarises the
aggregate performance of all gen AI configurations tested under the classical Beer Game setting played by students (20-week, 2-2-2 lead times), listing total costs and normalized costs relative to the human benchmark
for each combination of model type and inference-time technique; values below 100 indicate that the agent outperformed the average human team. Table~\ref{tab:supply_chain_costs} provides descriptive statistics over 30 identical runs of the simulation for each model---average total supply chain costs, standard deviation, and coefficient of
variance---illustrating the differences in both cost efficiency and run-to-run stability across models and interventions. Table~\ref{tab:supply_chain_analysis} identifies the specific model comparisons underlying the numerical results reported in the main text, together with the corresponding percentage changes in total supply chain costs and differences in the coefficient of variance.

% The supplementary tables in this appendix provide a granular breakdown of the supply chain performance metrics, beginning with a stability analysis that uses the Coefficient of Variance (CV) to demonstrate that reasoning-capable models, particularly when augmented with a policy layer, achieve significantly lower order volatility than human baselines. The augmentation analysis quantify the specific impact of architectural interventions, revealing that the integration of both Policy and Orchestrator layers yields a synergistic effect, reducing costs by up to 52.1\% compared to base models. Finally, the human baseline data offers an aggregated look at the control group's performance across multiple teams, establishing the 3,207-unit total cost mean that serves as the 100\% normalization benchmark for all model comparisons throughout the study.

\begin{table}[ht]
    \centering
    \caption{Human Teams: Total Costs per Run}
    \label{tab:human_teams_costs}
        \begin{tabular}{lr}
            \toprule
            \textbf{Human Teams} & \textbf{Total Costs} \\
            \midrule
            4/6/2025, 50 students &   867.5  \\
                                  &   854.0  \\
                                  &  1091.0  \\
                                  &  8784.5  \\
            4/3/2024, 60 students &  1049.5  \\
                                  &   695.5  \\
                                  &  4732.5  \\
                                  &  7182.5  \\
                                  &  6258.5  \\
                                  &  2024.5  \\
                                  &  1735.0  \\
            \midrule
            Average Costs         & 3206.82  \\
            \bottomrule
        \end{tabular}
\end{table}

\begin{table}[ht]
    \centering
    \caption{Supply Chain Performance: 20 Weeks Beer Game}
    \label{tab:supply_chain_full}
    \begin{adjustbox}{width=\textwidth}
        \begin{tabular}{llrr}
            \toprule
            \textbf{Scenario} & \textbf{Model/Human} & \textbf{Total Costs} & \textbf{Normalized Costs} \\
            \midrule
            Human                              & Human Teams                            & 3207  & 100.00  \\
            Reasoning Model                    & GPT-5 mini default                     & 2142  & 66.79   \\
            Reasoning + Orchestrator           & GPT-5 mini demand                      & 1765  & 55.04   \\
            Reasoning + Policy                 & GPT-5 mini budget                      & 1608  & 50.14   \\
            Reasoning + Prompt                 & GPT-5 mini prompt                      & 2620  & 81.70   \\
            Reasoning + Orchestrator + Policy  & GPT-5 mini demand+budget               & 1596  & 49.77   \\
            Reasoning Model                    & Llama 4 Maverick 17B default           & 2080  & 64.86   \\
            Reasoning + Orchestrator           & Llama 4 Maverick 17B demand            & 1559  & 48.61   \\
            Reasoning + Policy                 & Llama 4 Maverick 17B budget            & 1235  & 38.51   \\
            Reasoning + Prompt                 & Llama 4 Maverick 17B prompt            & 1758  & 54.82   \\
            Reasoning + Orchestrator + Policy  & Llama 4 Maverick 17B demand+budget     & 1063  & 33.15   \\
            Non-Reasoning Model                & GPT-4o mini default                    & 7093  & 221.17  \\
            Non-Reasoning + Orchestrator       & GPT-4o mini demand+volatility          & 2171  & 67.70   \\
            Non-Reasoning + Policy             & GPT-4o mini budget                     & 4351  & 135.67  \\
            Non-Reasoning + Prompt             & GPT-4o mini prompt                     & 3956  & 123.36  \\
            Non-Reasoning + Orchestrator       & GPT-4o mini demand                     & 4383  & 136.67  \\
            Non-Reasoning Model                & GPT-4.1 mini default                   & 7093  & 221.17  \\
            Non-Reasoning + Prompt             & GPT-4.1 mini prompt                    & 4734  & 147.61  \\
            Non-Reasoning Model                & Llama 3.3 70B default                  & 11743 & 366.17  \\
            \bottomrule
        \end{tabular}
    \end{adjustbox}
\end{table}

\begin{table}[ht]
    \centering
    \caption{Supply Chain Cost and Coefficient of Variance Across Runs by Model Type}
    \label{tab:supply_chain_costs}
    \begin{adjustbox}{width=\textwidth}
        \begin{tabular}{llrrr}
            \toprule
            \textbf{Scenario} & \textbf{Model} & \textbf{Total Costs} & \textbf{Standard Deviation} & \textbf{Coefficient of Variance} \\
            \midrule
            Reasoning Model         & GPT-5 mini default           & 2142 & 906.24 & 0.4231 \\
            Reasoning + Policy      & GPT-5-mini budget            & 1608 & 631.34 & 0.3926 \\
            Non-Reasoning Model     & GPT-4o mini default          & 7093 & 890.70 & 0.1256 \\
            Non-Reasoning + Policy  & GPT-4o-mini budget           & 4351 & 242.58 & 0.0558 \\
            Reasoning Model         & Llama 4 Maverick 17B default & 2080 & 956.31 & 0.4598 \\
            Reasoning + Policy      & Llama 4 Maverick 17B budget  & 1235 & 453.69 & 0.3674 \\
            \bottomrule
        \end{tabular}
    \end{adjustbox}
\end{table}

\begin{table}[ht]
    \centering
    \caption{Supply Chain Analysis Summary}
    \label{tab:supply_chain_analysis}
    \begin{adjustbox}{width=\textwidth}
        \begin{tabular}{llrr}
            \toprule
            \textbf{Analysis} & \textbf{Comparison} & \textbf{Numbers} & \textbf{Note} \\
            \midrule
            Human vs AI + Orchestration              & Llama 4 Maverick 17B demand+budget vs Human Teams           & $-67\%$ & Percentage Change in Total Costs \\
            Stability in Runs                        & Min: GPT-4o mini default; Max: Llama 4 Maverick 17B default & 13\%, 46\%  & Coefficient of Variance          \\
            Reasoning vs Non-Reasoning               & GPT-4o mini default vs GPT-5 mini default                   & $-70\%$ & Percentage Change in Total Costs \\
           & Llama 3.3 70B default vs Llama 4 Maverick 17B default       & $-82\%$ & Percentage Change in Total Costs \\
            Effect of Guardrail                      & GPT-5 mini default vs GPT-5 mini budget                     & $-25\%$ & Percentage Change in Total Costs \\
             & GPT-4o mini default vs GPT-4o mini budget                   & $-39\%$ & Percentage Change in Total Costs \\
          & Llama 4 Maverick 17B default vs Llama 4 Maverick 17B budget & $-41\%$ & Percentage Change in Total Costs \\
            Improvement in Stability                 & Llama 4 Maverick 17B default vs Llama 4 Maverick 17B budget & 46\%, 37\%  & Coefficient of Variance          \\
            Orchestration: Demand Sharing            & GPT-5 mini default vs GPT-5 mini demand                     & $-18\%$ & Percentage Change in Total Costs \\
        & Llama 4 Maverick 17B default vs Llama 4 Maverick 17B demand & $-25\%$ & Percentage Change in Total Costs \\
          & GPT-4o mini default vs GPT-4o mini demand                   & $-38\%$ & Percentage Change in Total Costs \\
            Orchestration: Demand Sharing + Analysis & GPT-4o mini default vs GPT-4o mini demand+volatility        & $-69\%$ & Percentage Change in Total Costs \\
            Effect of Prompts                        & GPT-4o mini default vs GPT-4o mini prompt                   & $-44\%$ & Percentage Change in Total Costs \\
        & GPT-4.1 mini default vs GPT-4.1 mini prompt                 & $-33\%$ & Percentage Change in Total Costs \\
            \bottomrule
        \end{tabular}
    \end{adjustbox}
\end{table}

\end{appendices}

\end{document}